\documentclass[abstract=on, 11pt]{scrreprt}
\addtokomafont{disposition}{\rmfamily}
\usepackage{tikz}
\usetikzlibrary{positioning, arrows.meta, fit, backgrounds}
\usetikzlibrary{shapes.geometric}
\usepackage{makecell}
\usepackage{amsmath,amsfonts}
\usepackage{amsthm,amssymb}
\usepackage{subcaption}
\usepackage{wrapfig}

\usepackage{newpxmath}
\usepackage{newpxtext}

\usepackage{hyperref}
\hypersetup{
     colorlinks   = true,
     linkcolor = black,
     urlcolor    = teal,
	 citecolor = teal
}
\usepackage{authblk}
\usepackage[margin=1in]{geometry}

\usepackage[utf8]{inputenc} 
\usepackage[T1]{fontenc}    
\usepackage{url}            
\usepackage{booktabs}       
\usepackage{nicefrac}       
\usepackage{microtype}      
\usepackage{xcolor}         

\usepackage{bm}
\usepackage{enumitem,array,booktabs}
\usepackage{algorithm}
\usepackage{algorithmic}
\newcounter{protocol}
\makeatletter
\newenvironment{protocol}[1][htb]{%
  \let\c@algorithm\c@protocol
  \renewcommand{\ALG@name}{Protocol}
  \begin{algorithm}[#1]%
  }{\end{algorithm}
}
\makeatother

\usepackage{bbm}
\usepackage{tabulary,multirow}
\usepackage{dsfont}
\usepackage{tcolorbox}
\usepackage{thmtools}
\usepackage{xpatch}
\xpatchcmd{\proof}{\itshape}{\scshape}{}{}
\renewenvironment{proof}[1][]{\par\noindent{\textbf{Proof #1}\ }}{\hfill$\blacksquare$\\[2mm]}
\usepackage{tablefootnote}

\newtheorem{theorem}{Theorem}[chapter]
\newtheorem{definition}{Definition}[chapter]
\newtheorem{lemma}{Lemma}[chapter]
\newtheorem{proposition}{Proposition}[chapter]
\newtheorem{example}{Example}[chapter]
\newtheorem{corollary}{Corollary}[chapter]

\newtheorem{problem}{Problem}[chapter]
\newtheorem{remark}{Remark}[chapter]
\newtheorem{observation}{Observation}[chapter]
\newtheorem{assumption}{Assumption}[chapter]
\newtheorem{innercustomasp}{Assumption}[chapter]
\newenvironment{customasp}[1]
  {\renewcommand\theinnercustomasp{#1}\innercustomasp}
  {\endinnercustomasp}

\renewcommand{\vec}[1]{\boldsymbol{\mathbf{#1}}}
\let\top\intercal

\DeclareMathOperator*{\argmax}{arg\,max}
\DeclareMathOperator*{\argmin}{arg\,min}

\DeclareMathOperator{\Log}{Log}
\DeclareMathOperator{\poly}{poly}
\DeclareMathOperator{\polylog}{polylog}
\DeclareMathOperator{\loglog}{loglog}

\newcommand{\inner}[2]{\left\langle #1, #2 \right\rangle}

\DeclareMathOperator{\st}{s.t.}

\newcommand{\ceil}[1]{\left\lceil#1\right\rceil}
\newcommand{\floor}[1]{\left\lfloor#1\right\rfloor}

\newcommand{\II}[1]{\mathds{1}_{\left\{#1\right\}}}

\newcommand{\Ber}{\mathrm{Ber}}
\newcommand{\Unif}{\mathrm{Unif}}
\newcommand{\Bin}{\mathrm{Bin}}

\newtheorem{conj}{Conjecture}
\newtheorem{lm}{Lemma}
\newtheorem{thm}{Theorem}
\newtheorem{dfn}{Definition}
\newtheorem{crl}{Corollary}
\newtheorem{rmk}{Remark}
\newtheorem{claim}{Claim}
\newtheorem{eg}{Example}

\newcommand{\R}{\mathbb{R}}

\newcommand{\nats}{\mathbb{N}}

\newcommand{\NN}{{\mathbb N}}
\newcommand{\1}[1]{\mathds{1}(#1)}
\newcommand{\ind}[1]{\mathds{1}_{#1}}
\newcommand{\bOne}{{\bm 1}}
\newcommand{\bZero}{{\bm 0}}
\renewcommand{\P}{\mathbb{P}}
\newcommand{\E}{\mathbb{E}}
\newcommand{\EE}[1]{\mathbb{E}\left[#1\right]}

\newcommand{\EEs}[2]{\mathbb{E}_{#1}\left[#2\right]}
\newcommand{\EEc}[2]{\mathbb{E}\left[#1\left|#2\right.\right]}

\newcommand{\PP}[1]{\mathbb{P}\left(#1\right)}
\newcommand{\PPs}[2]{\mathbb{P}_{#1}\left(#2\right)}

\newcommand{\norm}[1]{\left\|#1\right\|}

\newcommand{\abs}[1]{\left|#1\right|}

\newcommand*{\eqdef}{\triangleq}

\newcommand{\cA}{\mathcal{A}}
\newcommand{\cB}{\mathcal{B}}

\newcommand{\cD}{\mathcal{D}}
\newcommand{\cE}{\mathcal{E}}

\newcommand{\cF}{\mathcal{F}}
\newcommand{\cG}{\mathcal{G}}
\newcommand{\cH}{\mathcal{H}}
\newcommand{\cI}{\mathcal{I}}

\newcommand{\cK}{\mathcal{K}}
\newcommand{\cL}{\mathcal{L}}

\newcommand{\cM}{\mathcal{M}}

\newcommand{\cO}{\mathcal{O}}

\newcommand{\cP}{\mathcal{P}}
\newcommand{\cQ}{\mathcal{Q}}
\newcommand{\cR}{\mathcal{R}}

\newcommand{\cS}{\mathcal{S}}
\newcommand{\cT}{\mathcal{T}}

\newcommand{\cU}{\mathcal{U}}
\newcommand{\cV}{\mathcal{V}}

\newcommand{\cX}{\mathcal{X}}
\newcommand{\X}{\cX}
\newcommand{\cY}{\mathcal{Y}}

\newcommand{\bb}{\textbf{b}}

\newcommand{\bC}{\textbf{C}}

\newcommand{\bg}{\textbf{g}}
\newcommand{\bh}{\textbf{h}}

\newcommand{\bP}{\textbf{P}}

\newcommand{\be}{\textbf{e}}
\newcommand{\bff}{\textbf{f}}

\newcommand{\bq}{\textbf{q}}
\newcommand{\bu}{\textbf{u}}

\newcommand{\bv}{\textbf{v}}

\newcommand{\by}{\textbf{y}}

\newcommand{\bx}{\textbf{x}}

\newcommand{\bZ}{\textbf{Z}}

\newcommand{\eps}{\varepsilon}
\renewcommand{\epsilon}{\varepsilon}
\renewcommand{\hat}{\widehat}
\renewcommand{\tilde}{\widetilde}
\renewcommand{\bar}{\overline}

\newcommand{\balpha}{{\boldsymbol \alpha}}

\newcommand{\btheta}{{\boldsymbol \theta}}

\newcommand{\bdelta}{{\boldsymbol \delta}}

\newcommand{\bphi}{{\boldsymbol \phi}}
\newcommand{\bmu}{{\boldsymbol \mu}}

\newcommand{\nothere}[1]{}

\newcommand{\loss}[1]{\ell^{(#1)}}
\newcommand{\Loss}[1]{L^{(#1)}}
\newcommand{\hy}{\hat{y}}
\newcommand{\reg}[1]{R^{(#1)}}
\newcommand{\err}{\mathrm{err}}
\newcommand{\vcd}{\mathrm{VCdim}}
\newcommand{\ld}{\mathrm{Ldim}}

\newcommand{\VS}{\mathrm{VS}}
\newcommand{\SVM}{\mathrm{SVM}}

\newcommand{\DIS}{\mathrm{DIS}}
\newcommand{\ATK}{\mathrm{ATK}}
\newcommand{\Adv}{\mathrm{Adv}}
\newcommand{\atk}{\mathrm{atk}}

\newcommand{\trn}{\mathrm{trn}}
\newcommand{\conv}{\mathrm{Conv}}
\newcommand{\sign}{\mathrm{sign}}

\newcommand{\ERM}{{\mathrm{ERM}}}

\newcommand{\Proj}{{\mathrm{Proj}}}
\newcommand{\Rfl}{{\mathrm{Ref}}}
\newcommand{\A}{\mathcal{A}}

\newcommand{\Major}{{\mathrm{Major}}}

\newcommand{\supp}{{\mathrm{supp}}}
\newcommand{\po}{\leq_f^\cH}
\newcommand{\clsr}{\mathrm{Closure}}
\newcommand{\sph}{\Gamma}
\newcommand{\rbst}{{\mathrm{rbst}}}

\newcommand{\wt}[1]{\widetilde{#1}}
\newcommand{\diag}{\mathrm{diag}}
\newcommand{\CR}{\mathrm{CR}}

\newcommand{\MB}{\mathrm{MB}}
\newcommand{\SC}{\mathrm{SC}}
\newcommand{\MW}{\mathrm{MWMR}}

\newcommand{\TV}{\mathrm{TV}}
\newcommand{\KL}[2]{\mathrm{D}_\mathrm{KL}(#1\|#2)}
\newcommand{\kl}[2]{\mathrm{kl}(#1, #2)}
\newcommand{\out}{\mathrm{out}}

\newcommand{\lossstr}{\ell^{\textrm{str}}}

\newcommand{\fpr}{\mathrm{fpr}}

\newcommand{\fnr}{\mathrm{fnr}}

\newcommand{\Rad}{\mathrm{Rad}}

\newcommand{\true}[1]{\mathds{1}(#1)}

\newcommand{\lstr}{\ell^{\textrm{str}}}
\newcommand{\Lstr}{\cL^{\textrm{str}}}
\newcommand{\hatlstr}{\hat \cL^{\textrm{str}}}
\newcommand{\lgraph}{\cL_{\textrm{neighborhood}}}
\newcommand{\lproxy}{\cL_{\textrm{proxy}}}
\newcommand{\hatlproxy}{\hat\cL_{\textrm{proxy}}}
\newcommand{\hatlgraph}{\hat\cL_{\textrm{neighborhood}}}

\newcommand{\regret}{\textrm{Regret}}

\newcommand{\restatableeq}[3]{\label{#3}#2\gdef#1{#2\tag{\ref{#3}}}}

\newcommand{\bvartheta}{{\boldsymbol \vartheta}}
\newcommand{\optlocal}{\mathrm{opt}}
\newcommand{\bFlocal}{\mathbf{f}}
\newcommand{\bfm}{\mathbf{m}}
\newcommand{\ef}{\mathrm{ef}}
\newcommand{\eq}{\mathrm{eq}}

\renewcommand\c[0]{\boldsymbol{c}}

\renewcommand\u[0]{\boldsymbol{u}}
\renewcommand\v[0]{\boldsymbol{v}}

\newcommand\x[0]{\boldsymbol{x}}
\newcommand\y[0]{\boldsymbol{y}}

\newcommand\ee[0]{\mathbb{E}}

\newcommand\ii[0]{\mathbb{I}}

\renewcommand\ll[0]{\mathbb{L}}

\newcommand\rr[0]{\mathbb{R}}

\newcommand\aaa[0]{\mathcal{A}}

\newcommand\ddd[0]{\mathcal{D}}

\newcommand\fff[0]{\mathcal{F}}
\renewcommand\ggg[0]{\mathcal{G}}

\newcommand\mmm[0]{\mathcal{M}}

\newcommand\ooo[0]{\mathcal{O}}
\newcommand\ppp[0]{\mathcal{P}}

\newcommand\www[0]{\mathcal{W}}

\newcommand\rb[1]{\left(#1\right)}
\renewcommand\sb[1]{\left[#1\right]}
\newcommand\cb[1]{\left\{#1\right\}}

\usepackage[makeroom]{cancel}

\usepackage{xcolor}

\usepackage{lipsum}

\newcommand{\ngrad}[1]{\nabla \tilde F_{#1}}

\newcommand{\OPT}{\textrm{OPT}}

\newcommand{\gbs}{\textrm{GBS}}

\newcommand{\Xset}{X_1,\ldots,X_k}
\newcommand{\xset}{\{\Xset\}}

\newcommand{\ir}{\textrm{B2IR}}

\newcommand{\pip}{\pi}

\newcommand{\brp}{p^*}
\newcommand{\brr}{a^*}

\newcommand{\nr}{\textrm{NegReg}}

\newcommand{\eint}{\epsilon_\textrm{int}}

\newcommand{\eneg}{\epsilon_\textrm{neg}}

\newcommand{\sig}{\varphi}
\newcommand{\init}{\textrm{init}}
\newcommand{\pit}{p_t^\init}
\newcommand{\rit}{r_t^\init}

\newcommand{\ugap}{\textrm{UGap}}

\newcommand{\Span}{\mathrm{span}}
\newcommand{\tj}{\tau}
\newcommand{\spl}{\mathrm{Simplex}}
\newcommand{\rank}{\mathrm{rank}}

\newcommand{\eacc}{\epsilon_{\text{acc}}}
\newcommand{\ealpha}{\epsilon_{\alpha}}
\newcommand{\calpha}{C_{\alpha}}
\newcommand{\cw}{C_w}
\newcommand{\cv}{C_V}
\newcommand{\Lsup}[1]{L^{(#1)}}
\newcommand{\Esup}[1]{E^{(#1)}}
\newcommand{\xsup}[1]{x^{(#1)}}
\newcommand{\Qsup}[1]{F^{(#1)}}
\newcommand{\Jsup}[1]{J^{(#1)}}
\newcommand{\betasup}[1]{\beta^{(#1)}}
\newcommand{\append}{\circ}

\newcommand{\pssp}{P(s' \vert s, \pi(s))}
\newcommand{\basis}{b}

\newcommand{\hatwd}{\hat w ^{(\delta)}}
\newcommand{\hatAd}{\hat A^{(\delta)}}
\newcommand{\hatwo}{\hat w_0}

\newcounter{relctr} 
\everydisplay\expandafter{\the\everydisplay\setcounter{relctr}{0}} 

\newcommand\labelrel[2]{%
  \begingroup
    \refstepcounter{relctr}%
    \stackrel{\textnormal{(\alph{relctr})}}{\mathstrut{#1}}%
    \originallabel{#2}%
  \endgroup
}
\AtBeginDocument{\let\originallabel\label} 

\newcommand{\tloss}[1]{\tilde{\ell}^{(#1)}}
\newcommand{\tLoss}[1]{\tilde{L}^{(#1)}}

\newcommand{\sreg}[1]{\mathrm{SleepReg}^{(#1)}}

\newcommand{\SL}{\mathrm{SL}}

\newcommand{\sd}{\mathrm{SD}}
\newcommand{\fll}{\mathrm{FLL}}
\newcommand{\ts}{I_{h^*}}

\newcommand{\la}{\leftarrow}
\newcommand{\vcao}{\mathrm{VC_{ao}}}
\newcommand{\vco}{\mathrm{VC_{o}}}

\newcommand{\inv}{\mathrm{INV}}
\newcommand{\re}{\mathrm{RE}}
\newcommand{\ag}{\mathrm{AG}}
\newcommand{\da}{\mathrm{DA}}
\newcommand{\Sym}{\mathrm{Sym}}
\newcommand{\BL}{\mathrm{BL}}
\newcommand{\Majority}{{\mathrm{Majority}}}

\usepackage[
backref=true,
natbib=true,
style=apa,
maxalphanames=8,
sorting=nyt
]{biblatex}
\begin{document}

\title{Trustworthy Machine Learning under Social \\and Adversarial Data Sources}
\date{\today}
\author{Han Shao}

\begin{titlepage}
   \begin{center}
       \vspace*{4cm}

       {\huge \textbf{Trustworthy Machine Learning under Social \\and Adversarial Data Sources}}

       \vspace{2cm}
BY\\
    HAN SHAO

         \vspace{4cm}   
       A thesis submitted
       \\
in partial fulfillment of the requirements for
\\
the degree of
\\
Doctor of Philosophy in Computer Science
\\
at the
\\
TOYOTA TECHNOLOGICAL INSTITUTE AT CHICAGO
\\
Chicago, Illinois
\\
July, 2024
       \vspace{1cm}
     
       \textbf{Thesis Committee:}\\
       Avrim Blum (Thesis Advisor)\\
       Nika Haghtalab\\
       Yishay Mansour\\
       Nati Srebro         
   \end{center}
\end{titlepage}

\begin{abstract}
Machine learning has witnessed remarkable breakthroughs in recent years.
As machine learning permeates various aspects of daily life, individuals and organizations increasingly interact with these systems, exhibiting a wide range of social and adversarial behaviors. These behaviors may have a notable impact on the behavior and performance of machine learning systems. Specifically, during these interactions, data may be generated by \emph{strategic individuals}, collected by \emph{self-interested data collectors}, possibly poisoned by \emph{adversarial attackers}, and used to create predictors, models, and policies satisfying \emph{multiple objectives}.
As a result, the machine learning systems' outputs might degrade, such as the susceptibility of deep neural networks to adversarial examples~\citep{szegedy2013intriguing,shafahi2018poison} and the diminished performance of classic algorithms in the presence of strategic individuals~\citep{ahmadi2021strategic}. Addressing these challenges is imperative for the success of machine learning in societal settings. 

This thesis is organized into two parts: learning under social data sources and adversarial data sources. For social data sources, we consider problems including: (1) learning with strategic individuals for both finite and infinite hypothesis classes, where we provide an understanding of learnability in both online and PAC strategic settings, (2) incentives and defections of self-interested data collectors in single-round federated learning, multi-round federated learning, and collaborative active learning, (3) learning within games, in which one of players runs a learning algorithm instead of best responding, and (4) multi-objective learning in both decision making and online learning. For adversarial data sources, we study problems including: (1) robust learning under clean-label attacks, where the attacker injects a set of correctly labeled points into the training set to mislead the learner into making mistakes at targeted test points, and (2) learning under transformation invariances and analyzing the popular method of data augmentation. 
\end{abstract}
\chapter*{Acknowledgements}
I couldn't have been more fortunate to be advised by the best advisor in the world, Avrim Blum. Avrim is an incredible mentor---wise, smart, knowledgeable, kind, supportive, and trustworthy. From him, I learned how to conduct research, develop my taste in problems, model and formulate questions, and find solutions. Avrim gave me the freedom to explore different problems, encouraged me whenever I got stuck, and helped me grow as an independent researcher. He provided valuable and insightful suggestions whenever I faced uncertainty in my decisions. Avrim is my role model as a researcher, advisor, and human being. Avrim, thank you. I've truly enjoyed my PhD journey under your guidance and wouldn't have come this far in academia without you. I aspire to be an advisor like you in the future, though I know it's a really high bar to reach.

I learned a lot from working with Yishay Mansour and Shay Moran. They are both incredibly smart and supportive. I was lucky to have worked with them, especially during my job search season. The highlight of my week was always our meeting time, where I received enlightening research ideas, job search suggestions, and mental support. I was very stressed during that time, and things wouldn't have gone as smoothly without your help.

I'd like to thank Aaron Roth for hosting me as a visiting student during the summer of 2023. I had a wonderful time in Philly. Aaron is smart, supportive, and responsive. Thank you, Aaron, for introducing me to calibration and helping me complete my first pure game theory project.

I am grateful for the opportunity to work with Nika Haghtalab during my Ph.D. She is the most energetic person I have ever met and always has amazing ideas. She is another role model to me.

I would like to thank my committee member Nati Srebro, not only for being on my committee and providing insightful feedback but also for his reading group, the best I've ever attended. I read many papers outside my area in his reading group and learned a lot from the discussions. If I organize my own reading group in the future, I would definitely want it to be in the same format.

I would also like to thank all my other collaborators and coauthors: Lee Cohen, Natalie Collina, Minbiao Han, Steve Hanneke, Omar Montasser, Kumar Kshitij Patel, Richard Lanas Phillips, Jian Qian (my boyfriend; our collaboration happened because of COVID, which was a very fun experience), Aadirupa Saha, Matthew R Walter, and Lingxiao Wang. Working with you has been enjoyable and taught me valuable lessons. Beyond my coauthors, I'd also like to thank Freda Shi and Xiao Zhang for discussing with me and sharing perspectives from the applied side.

TTIC is an amazing place to conduct research. I am grateful to the faculty and staff members for their efforts to make TTIC even better. I have benefited greatly from the excellent courses and/or conversations with Zhiyuan Li, Yury Makarychev, Greg Shakhnarovich, Madhur Tulsiani, and Matthew Turk, as well as the administrative support from Adam Bohlander, Rose Bradford, Erica Cocom, Chrissy Coleman, Jessica Jacobson, Celeste Ki, Deree Kobets, Mary Marre, and Amy Minick.

I would like to express my gratitude to my friends throughout Chicago, who have made my life significantly more enjoyable while pursuing my PhD. Here is a non-exhaustive list of friends I would like to thank: Freda, Omar, Sudarshan, Xiaodan, Jiading, Gene, Shengjie, Naren, Keziah, Ankita, Kshitij, Kavya, Pushkar, Shashank, Chung-Ming, Melissa, Anmol, Amin, Donya, Luzhe, Shuo, Hongyuan, Liren, Jiawei, and Kunhe. I have had a lot of fun hanging out with you. I am especially grateful to Lee, Naren, Hongyuan, Shuo, and Luzhe for supporting me during my toughest times. You are incredibly kind and supportive, and so good at being there for friends that I have to reflect on whether I can be as good as you. I've always found therapy in our conversations. Special thanks to Naren, who was my first local friend after I came to the U.S. and introduced me to many new things. Adapting to life in a new country would have been much more difficult without you, though possibly "safer" (I will never forget our "red line adventure." As I write this acknowledgment, I am still waiting for my ankle to recover after spraining it while climbing with you last weekend). Freda---not much to say, my sis.
I'd also like to thank the anonymous friends from the list for sharing gossip with me and adding a bit of spice to my life. I am passionate about exercise and really enjoyed running from Hyde Park to downtown with Pushkar and Chung-Ming and then having meals with Naren and Anmol there. I would like to thank those on the list who have been ``tortured'' by me at the gym and hope you enjoyed our training time as much as I did.

I would also like to thank the women in computer science who have inspired and encouraged me, including the great female researchers I've met and my female researcher friends. You made me braver and more confident. Special thanks to Lee---you are like a patient, rational, and considerate big sister to me, always there to help. Additionally, I'd like to thank Ning, who taught me some interesting lessons I never knew before.

Last but not least, I'd like to thank my parents for their unconditional love and support and for respecting my decision to pursue what I want. 

\newpage
\tableofcontents
\newpage
\listoffigures
\newpage

\chapter{Introduction}
Machine learning has witnessed remarkable breakthroughs in recent years. As machine learning permeates various aspects of daily life, individuals and organizations increasingly interact with these systems, exhibiting a wide range of social and adversarial behaviors that may significantly impact the performance of machine learning systems. 

\paragraph{Strategic Individuals}
Across many domains, machine learning is applied to inform decisions about applicants for a variety of resources. However, when individuals have incentives to benefit from specific predictive outcomes, they may act to obtain favorable predictions by modifying their features. Since this can harm predictive performance, learning becomes susceptible to a classic principle in financial policy-making known as Goodhart's law, which states, "\textit{If a measure becomes the public's goal, it is no longer a good measure}." This natural tension between learning systems and those to whom the system is applied is widespread, spanning loan approvals, university admissions, job hiring, and insurance. In these scenarios, learning systems aim for accurate predictions, while individuals, regardless of their true labels, have an incentive to be classified positively. For instance, in college admissions, applicants might retake the SAT or take easier courses to boost their GPA to fool the classifier.

\paragraph{Self-Interested Data Collectors}
In many real-world applications, datasets are distributed across different silos, such as hospitals, schools, and banks, necessitating collaborations among them. In recent years, collaborative learning, such as federated learning, has been embraced as an approach for facilitating collaboration across large populations of data collectors. However, what will ultimately determine the success and impact of collaborative learning is the ability to recruit and retain large numbers of data collectors. There is an inherent tension between the collaborative learning protocol and the data collectors. The learning protocol aims to find a model that is beneficial for all data collectors, while each data collector's goal is to find a model that is good for their own local data with minimal data contribution. Consequently, if the learning protocol requires data collectors to provide more data than necessary to fulfill their own objectives, they will not contribute as the protocol requires.

\paragraph{Multi-Objective Users}
While machine learning problems typically involve optimizing a single scalar reward, there are many domains where it is desirable or necessary to optimize over multiple (potentially conflicting) objectives simultaneously. For example, safety, speed, and comfort are all desired objectives for autonomous car users, but speed could negatively impact safety (e.g., taking longer for the vehicle to stop suddenly) or comfort (e.g., causing discomfort during fast turns). Consequently, when a learning system optimizes a scalar loss, it might overlook these multiple objectives and produce an unsatisfactory model or policy for users. Additionally, there could be more than one stakeholder in the learning process, each with a different objective. Focusing solely on one objective might lead to drastic performance drops for the others.

\paragraph{Adversarial Attackers}
Adversarial attacks play a significant role in exposing the vulnerabilities of machine learning systems. Many popular models lack robustness in real-world scenarios. For example, in image tasks, adding imperceptible noise to training images~\citep{szegedy2013intriguing} or poisoning the training set with additional images~\citep{shafahi2018poison} can severely compromise the performance of deep neural networks.

Due to these social and adversarial data factors, the outputs of machine learning systems might degrade. Addressing these challenges is imperative for the success of machine learning.

This thesis makes contributions that address, from a theoretical perspective, the trustworthiness challenges arising from these social and adversarial data factors. Such data factors have not been well modeled by the existing theory. Hence, this thesis focuses on modeling the social and adversarial aspects inherent in machine learning interactions, analyzing their impact on predictors, and developing methods and insights to enhance performance.

The central theme of this thesis is to
\begin{quote}
\textit{develop the theoretical foundations for trustworthy machine learning under social and adversarial data sources.}
\end{quote}

\section{Overview of Thesis Contributions and Structure}
This thesis is organized in two parts: learning under social and adversarial data sources.

\subsection*{Learning under Social Data Sources}

Within large interactive systems, humans frequently demonstrate strategic behaviors, including strategic manipulations by individuals and defections by self-interested data collectors, or have a variety of different objectives. 

\paragraph{Chapters~\ref{chap:strategic-finite} and \ref{chap:strategic-infinite}: Learning with Strategic Individuals.} In these two chapters, we consider the problem of strategic classification, where agents can strategically manipulate their feature vector up to an extent in order to be predicted as positive. 

In Chapter~\ref{chap:strategic-finite}, 
we start by considering learning a finite hypothesis class $\cH$.
It is well-known that in the standard (non-strategic) realizable online learning, the mistake bound of $\log(\abs{\cH})$ can be achieved by the Halving algorithm while in the standard PAC learning, the sample complexity of $\cO(\log(\abs{\cH}))$ can be achieved by the empirical risk minimizer (ERM). Then we have the following question: \textit{under realizability in the strategic setting, can we also achieve a logarithmic dependency on $\abs{\cH}$ in the strategic setting?}

As the problem reduces to a standard learning problem when manipulation abilities are known, we focus on the scenario where the manipulation abilities are unknown. 
We show that in the case of ball manipulations, when the original feature vector is revealed prior to choosing the implementation, we can achieve logarithmic mistake bound and sample complexity via a variant of Halving.
When the original feature vector is not revealed beforehand, the problem becomes significantly more challenging. 
Specifically, any learner will experience a mistake bound that scales linearly with $\abs{\cH}$, and any proper learner will face sample complexity that also scales linearly with $\abs{\cH}$.
Then we propose an improper learner with $\polylog(\abs{\cH})$ sample complexity.
For arbitrary (non-ball) manipulations, the situation worsens. Even in the simplest setting where the original feature is observed before implementation and the manipulated feature is observed afterward, any learner will encounter a linear mistake bound and sample complexity. 

Then in Chapter~\ref{chap:strategic-infinite}, we further consider learning an infinite hypothesis class, i.e., $|\cH|=\infty$. In classical binary classification, it is well-known that Probably Approximately Correct (PAC) learnability is characterized by the Vapnik–Chervonenkis (VC) dimension $\vcd(\cH)$~\citep{vapnik:74} and that online learnability is characterized by the Littlestone dimension $\ld(\cH)$~\citep{littlestone1988learning}. Then we have the following question: \textit{for every learnable hypothesis class, is it also learnable in the strategic setting? If so, how?} 

We essentially show that \emph{every learnable class is also strategically learnable when each agent can only manipulate to a bounded number of feature vectors}.
We consider various settings based on what and when information is revealed to the learner, including: (1) known graph and pre-manipulation feature beforehand, (2) known graph but unknown pre-manipulation feature, and (3) unknown graph but known pre-manipulation feature.
In the online setting, we provide reduction algorithms that reduce strategic learning problems into standard learning problems.
In the PAC setting, we develop entirely new algorithms.
We also show that having a bounded number of feature vectors each agent can manipulate is a necessary condition for strategic learnability.

\paragraph{Chapters~\ref{chap:incentives-single}, \ref{chap:incentives-multi} and \ref{chap:incentives-active}: Incentives in Collaborative Learning} In these chapters, we consider the defection behaviors of self-interested data collectors in various collaborative learning settings.

In Chapter~\ref{chap:incentives-single}, we start by considering \textit{single-round federated learning} setting, where all collectors have to decide their contribution level at the beginning. Inspired by game theoretic notions, we introduce a framework for incentive-aware learning and data sharing in federated learning. We define stable and envy-free equilibria, which capture notions of collaboration in the presence of collectors interested in meeting their learning objectives while keeping their own data contribution level low. For example, in an envy-free equilibrium, no collector would wish to swap their contribution level with any other collector, and in a stable equilibrium, no collector would wish to unilaterally reduce their contribution. In addition to formalizing this framework, we characterize the structural properties of such equilibria, proving when they exist, and showing how they can be computed. Furthermore, we compare the sample complexity of incentive-aware collaboration with that of optimal collaboration when one ignores collectors' incentives.

However, in many practical algorithms, the interaction between the learning protocol and the data collectors usually occurs in multiple rounds. For example, prevalent federated learning methods like \textsc{FedAvg} employ an intermittent communication scheme. In Chapter~\ref{chap:incentives-multi}, we model and study one type of self-interested behavior of data collectors in \textit{multi-round federated learning}.
In this scenario, the learning protocol shares the current model with all data collectors, each collector provides their local information, and the protocol then aggregates this information to update the model. Data collectors will drop out of the training procedure once the current model sent to them is good enough for their local data. Then we reveal that the prevalent method \textsc{FedAvg} will fail when data collectors demonstrate such a self-interested behavior and offer a new algorithm that can prevent collectors' defections.

Beyond federated learning, we also consider another collaborative learning setting --- collaborative active learning in Chapter~\ref{chap:incentives-active}. Here, rational collectors aim to obtain labels for their data sets while keeping label complexity at a minimum. We focus on designing (strict) individually rational (IR) collaboration protocols, ensuring that collectors cannot reduce their expected label complexity by acting individually. We first show that given any optimal active learning algorithm, the collaboration protocol that runs the algorithm as is over the entire data is already IR. However, computing the optimal algorithm is NP-hard. We therefore provide collaboration protocols that achieve (strict) IR and are comparable with the best known tractable approximation algorithm in terms of label complexity.

Both strategic classification and incentivized collaboration are learning problems in which participants demonstrate strategic behaviors and attempt to game the learning systems. On the other hand, as machine learning becomes more accessible and popular, an increasing number of individuals rely on AI interfaces like ChatGPT for decision-making in their interactions with others. Consequently, learning within games emerges as a crucial research area.

\paragraph{Chapter~\ref{chap:games}: Learning within Games} In this chapter, we study a repeated principal agent problem between a long-lived principal and agent pair in a prior free setting.  In our setting, the sequence of realized states of nature may be adversarially chosen, the agent is non-myopic, and the principal aims for a strong form of policy regret. 
Following~\citet{camara2020mechanisms}, we model the agent's long-run behavior with behavioral assumptions that relax the common prior assumption (for example, that the agent is running a no-swap-regret learning algorithm). Within this framework, we revisit the mechanism proposed by~\citet{camara2020mechanisms}, which informally uses calibrated forecasts of the unknown states of nature in place of a common prior. We give two main improvements. First, we give a mechanism that has an exponentially improved dependence (in terms of both running time and regret bounds) on the number of distinct states of nature. To do this, we show that our mechanism does not require truly calibrated forecasts, but rather forecasts that are unbiased subject to only a polynomially sized collection of events --- which can be produced with polynomial overhead. 
Second, instead of constructing a policy and assuming that the policy is ``stable'' (informally requiring that under such a policy, all approximately optimal actions lead to approximately the same Principal payoff), we propose a general framework given access to a stable policy oracle. We then instantiate the oracle by developing efficient algorithms in several significant special cases, including the focal linear contracting setting.
Taken together, our new mechanism makes the compelling framework proposed by \citet{camara2020mechanisms} more powerful, now able to be realized over polynomially sized state spaces. 

Back to learning problems, users commonly possess multiple objectives that may not be encompassed by a single scalar loss function. Consequently, when the learning system optimizes a scalar loss, it could potentially overlook these multiple objectives and produce an unsatisfactory model or policy for users. 

\paragraph{Chapter~\ref{chap:momdp} and \ref{chap:primary}: Multi-Objective Learning} In these chapters, we consider two multi-objective learning problems.

In Chapter~\ref{chap:momdp}, we consider multi-objective decision making problem. In the area of multi-objective decision making, take, for instance, the scenario of recommending self-driving cars to users concerned about various objectives like speed, safety, and comfort, among others. These users possess diverse and undisclosed preferences regarding these objectives.
Our goal is to efficiently find a personalized optimal policy for each user.
We employ the established model of multi-objective decision making, utilizing Markov decision processes (MDPs) with a vector reward, where each element represents one of the multiple objectives. Our focus revolves around solving two fundamental questions: 1) \textit{How can we interact with a user to learn their preferences?} and 2) \textit{How can we minimize user queries?} The former question focuses on human user feedback modeling--while quantitative questions pose challenges to users, comparative questions are more manageable, and interpretability--how to describe a policy (a mapping from states to actions) to a user. The latter question addresses the efficiency aspect--how to elicit a user's preference without bombarding them with numerous queries.

Aside from decision making, we have also explored the multi-objective issue in the classic online learning setting in Chapter~\ref{chap:primary}. 
In many online learning scenarios, there are two losses concerned, a primary one and a secondary one. For example, a recruiter making decisions about which job applicants to hire might weigh false positives and false negatives equally (the primary loss) but the applicants might weigh false negatives much higher (the secondary loss).
Notably, it's well-established that achieving no regret concerning both losses simultaneously is unattainable. Consequently, we've redirected our aim slightly: \textit{Can we achieve no regret concerning the primary loss while maintaining performance not significantly worse than the worst expert concerning the secondary loss?} Our investigation reveals that achieving this milder objective necessitates additional assumptions, under which we design algorithms to achieve this goal.

\subsection*{Learning under Adversarial Data Sources}

For this part, we aim to provide a theoretical exploration of some adversarial attackers, which have been shown empirically to be surprisingly powerful. We strive to comprehend these empirical phenomena from a theoretical viewpoint and suggest methodologies and insights to enhance both accuracy and robustness.

\paragraph{Chapter~\ref{chap:clean-label}: Robust Learning under Clean-Label Attack}
Data poisoning is a widely studied type of adversarial attack where the attacker adds examples to the training set with the goal of causing the algorithm to produce a classifier that makes specific mistakes the attacker wishes to induce at test time.
Clean label attacks, involving the injection of \emph{correctly-labeled} data points, have been empirically verified to be effective on deep neural networks by~\citet{shafahi2018poison}. There has been a lack of understanding regarding \textit{whether this susceptibility primarily stems from the complexity of deep networks or if it could also impact much simpler learning rules, such as linear classifiers}. We address this question in this chapter. 
Our findings illustrate that a single injected data point can render max-margin linear separators vulnerable. Additionally, simple linear classifiers are susceptible when attackers can inject an arbitrary amount of clean poisoning data. However, when the attacker is limited to injecting a finite number of points, we provide a generic robust method and prove its optimality.

\paragraph{Chapter~\ref{chap:transformation}: Learning under Transformation Invariances and Data Augmentation}
Transformation invariances are present in many real-world problems. For example, image classification is usually invariant to rotation and color transformation: a rotated car in a different color is still identified as a car. A label-invariant transformation attacker perturbs the test data by applying label-invariant transformations, such as rotations and altering colors, in an attempt to mislead the classifier into making errors.
To improve accuracy and robustness against such attackers, a universally applicable and easy way is data augmentation (DA), which adds the transformed data into the training set and trains a model on the augmented data. 
Empirical findings have substantiated the effectiveness of DA in improving accuracy and robustness, leading to its widespread adoption in modern deep learning, such as in AlexNet~\citet{krizhevsky2012imagenet}. Nevertheless, it remains uncertain under what circumstances DA provides a benefit.
As a result, it is natural to pose the following theoretical questions. \emph{How does data augmentation perform theoretically? And what is the optimal algorithm in terms of sample complexity under transformation invariances?} We provide an extensive theoretical study centering on these two questions in this chapter. We find that when the ground truth labeling function is contained in the hypothesis class, DA indeed ``helps'' but is not optimal. Then we propose an alternative algorithm that achieves optimality by implementing transformation invariances not only within the training data but also across the test data. This finding underscores the importance of implementing data augmentation not only during training but also during testing.
However, when the ground truth lies outside the hypothesis class, our analysis reveals that DA can even ``hurt'', performing worse than the standard empirical risk minimization.

\section{Bibliographical Remarks}
The research presented in this thesis is based on joint work with several co-authors, described
below. This thesis only includes works for which this author was the, or one of the, primary
contributors.
Chapter~\ref{chap:strategic-finite} is based on \citet*{shao2024strategic}.
Chapter~\ref{chap:strategic-infinite} is based on \citet*{cohen2024learnability}. Chapter~\ref{chap:incentives-single} is based on \citet*{blum2021one}.
Chapter~\ref{chap:incentives-multi} is based on
\citet*{han2023effect}.
Chapter~\ref{chap:incentives-active} is based on \citet*{Cohen2023incentivized}.
Chapter~\ref{chap:games} is based on \citet*{collina2024efficient}.
Chapter~\ref{chap:momdp} is based on \citet*{shao2024eliciting}.
Chapter~\ref{chap:primary} is based on \citet*{blum2020online}.
Chapter~\ref{chap:clean-label} is based on \citet*{blum2021robust}.
Chapter~\ref{chap:transformation} is based on \citet*{shao2022theory}.

\chapter{Strategic Classification for Finite Hypothesis Class}\label{chap:strategic-finite}
\section{Introduction}
Strategic classification addresses the problem of learning a classifier robust to manipulation and gaming by self-interested agents~\citep{hardt2016strategic}.
For example, given a classifier determining loan approval based on credit scores, applicants could open or close credit cards and bank accounts to increase their credit scores. In the case of a college admission classifier, students may try to take easier classes to improve their GPA, retake the SAT or change schools in an effort to be admitted.
In both cases, such manipulations do not change their true qualifications.
Recently, a collection of papers has studied strategic classification in both the online setting where examples are chosen by an adversary in a sequential manner~\citep{dong2018strategic,chen2020learning,ahmadi2021strategic, ahmadi2023fundamental}, and the distributional setting where the examples are drawn from an underlying data distribution~\citep{hardt2016strategic,zhang2021incentive,sundaram2021pac,lechner2022learning}.
Most existing works assume that manipulation ability is uniform across all agents or is known to the learner. However, in reality, this may not always be the case.
For instance, low-income students may have a lower ability to manipulate the system compared to their wealthier peers due to factors such as the high costs of retaking the SAT or enrolling in additional classes, as well as facing more barriers to accessing information about college~\citep{milli2019social} and it is impossible for the learner to know the highest achievable GPA or the maximum number of times a student may retake the SAT due to external factors such as socio-economic background and personal circumstances.

We characterize the manipulation of an agent by a set of alternative feature vectors that she can modify her original feature vector to, which we refer to as the \textit{manipulation set}.
\textit{Ball manipulations} are a widely studied class of manipulations in the literature, where agents can modify their feature vector within a bounded radius ball.
For example, \citet{dong2018strategic,chen2020learning,sundaram2021pac} studied ball manipulations with distance function being some norm and \citet{zhang2021incentive,lechner2022learning,ahmadi2023fundamental} studied a manipulation graph setting, which can be viewed as ball manipulation w.r.t. the graph distance on a predefined known graph.


In the online learning setting, the strategic agents come sequentially and try to game the current classifier.
Following previous work, we model the learning process as a repeated Stackelberg game over $T$ time steps.
In round $t$, the learner proposes a classifier $f_t$ and then the agent, with a manipulation set (unknown 
 to the learner), manipulates her feature in an effort to receive positive prediction from $f_t$.
There are several settings based on what and when the information is revealed about the original feature vector and the manipulated feature vector in the game.
The simplest setting for the learner is observing the original feature vector before choosing $f_t$ and the manipulated vector after.
In a slightly harder setting, the learner observes both the original and manipulated vectors after selecting $f_t$.
An even harder setting involves observing only the manipulated feature vector after selecting $f_t$.
The hardest and least informative scenario occurs when neither the original nor the manipulated feature vectors are observed.

In the distributional setting, the agents are sampled from an underlying data distribution.
Previous work assumes that the learner has full knowledge of the original feature vector and the manipulation set, and then views learning as a one-shot game and solves it by computing the Stackelberg equilibria of it.
However, when manipulations are personalized and unknown, we cannot compute an equilibrium and study learning as a one-shot game. In this work, we extend the iterative online interaction model from the online setting to the distributional setting, where the sequence of agents is sampled i.i.d. from the data distribution.
After repeated learning for $T$ (which is equal to the sample size) rounds, the learner has to output a strategy-robust predictor for future use.

In both online and distributional settings, examples are viewed through the lens of the current predictor and the learner does not have the ability to inquire about the strategies the previous examples would have adopted under a different predictor.



\paragraph{Related work}
Our work is primarily related to strategic classification in online and distributional settings.
Strategic classification was first studied in a distributional model by~\cite{hardt2016strategic} and subsequently by~\cite{dong2018strategic} in an online model.
\cite{hardt2016strategic} assumed that agents manipulate by best response with respect to a uniform cost function known to the learner.
Building on the framework of \citep{hardt2016strategic}, \cite{lechner2022learning,sundaram2021pac,zhang2021incentive,hu2019disparate,milli2019social} studied the distributional learning problem, and all of them assumed that the manipulations are predefined and known to the learner, either by a cost function or a predefined manipulation graph.
For online learning, \cite{dong2018strategic} considered a similar manipulation setting as in this work, where manipulations are personalized and unknown. However, they studied linear classification with ball manipulations in the online setting and focused on finding appropriate conditions of the cost function to achieve sub-linear Stackelberg regret.
\cite{chen2020learning} also studied Stackelberg regret in linear classification with uniform ball manipulations. \cite{ahmadi2021strategic} studied the mistake bound under uniform (possbily unknown) ball manipulations, and \cite{ahmadi2023fundamental} studied regret under a pre-defined and known manipulation.
The most relevant work is a recent concurrent study by~\cite{lechner2023strategic}, which also explores strategic classification involving unknown personalized manipulations but with a different loss function. 
In their work, a predictor incurs a loss of $0$ if and only if the agent refrains from manipulation and the predictor correctly predicts at the unmanipulated feature vector. 
In our work, the predictor's loss is $0$ if it correctly predicts at the manipulated feature, even when the agent manipulates.
As a result, their loss function serves as an upper bound of our loss function.

There has been a lot of research on various other issues and models in strategic classification. 
Beyond sample complexity, \cite{hu2019disparate,milli2019social} focused on other social objectives, such as social burden and fairness. 
Recent works also explored different models of agent behavior, including
proactive agents~\cite{zrnic2021leads}, non-myopic agents~\citep{haghtalab2022learning} and noisy agents~\citep{jagadeesan2021alternative}.
\cite{ahmadi2023fundamental} considers two agent models of randomized learners: a randomized algorithm model where the agents respond to the realization, and a fractional classifier model where agents respond to the expectation, and our model corresponds to the randomized algorithm model.
Additionally, there is also a line of research on agents interested in improving their qualifications instead of gaming~\citep{kleinberg2020classifiers, haghtalab2020maximizing,ahmadi2022classification}.
Strategic interactions in the regression setting have also been studied (e.g.,~\cite{bechavod2021gaming}).





\section{Model}\label{sec:model-strategic}
\paragraph{Strategic classification} 
We consider the binary classification task.
Let $\cX$ denote the feature vector space, $\cY = \{+1,-1\}$ denote the label space, and $\cH\subseteq \cY^\cX$ denote the hypothesis class.
In the strategic setting, instead of an example being a pair $(x,y)$, an example, or \textit{agent}, is a triple $(x,u,y)$ where $x \in \cX$ is the original feature vector, $y \in \cY$ is the label, and $u \subseteq \cX$ is the manipulation set, which is a set of feature vectors that the agent can modify their original feature vector $x$ to. 
In particular, given a hypothesis $h\in \cY^\cX$, the agent will try to manipulate her feature vector $x$ to another feature vector $x'$ within $u$ in order to receive a positive prediction from $h$.
The manipulation set $u$ is \textit{unknown} to the learner.
In this work, we will be considering several settings based on what the information is revealed to the learner, including both the original/manipulated feature vectors, the manipulated feature vector only, or neither, and when the information is revealed.



More formally, for agent $(x,u,y)$, given a predictor $h$, if $h(x) = -1$ and her manipulation set overlaps the positive region by $h$, i.e., $u \cap \cX_{h,+}\neq \emptyset$ with $\cX_{h,+}:= \{x\in \cX|h(x)=+1\}$, the agent will manipulate $x$ to $\Delta(x,h,u)\in u\cap \cX_{h,+}$\footnote{For ball manipulations, agents break ties by selecting the closest vector. When there are multiple closest vectors, agents break ties arbitrarily. For non-ball manipulations, agents break ties in any fixed way.
} to receive positive prediction by $h$.
Otherwise, the agent will do nothing and maintain her feature vector at $x$, i.e., $\Delta(x,h,u)=x$.
We call $\Delta(x,h,u)$ the manipulated feature vector of agent $(x,u,y)$ under predictor $h$.

A general and fundamental type of manipulations is \textit{ball manipulations}, where agents can manipulate their feature within a ball of \textit{personalized} radius.
More specifically, given a metric $d$ over $\cX$, the manipulation set is a ball $\cB(x;r) = \{x'|d(x,x')\leq r\}$ centered at $x$ with radius $r$ for some $r\in \R_{\geq 0}$.
Note that we allow different agents to have different manipulation power and the radius can vary over agents.
Let $\cQ$ denote the set of allowed pairs $(x,u)$, which we refer to as the feature-manipulation set space.
For ball manipulations, we have
$\cQ = \{(x, \cB(x;r))|x\in \cX, r\in \R_{\geq 0}\}$ for some known metric $d$ over $\cX$. In the context of ball manipulations, we use $(x,r,y)$ to represent $(x,\cB(x;r),y)$ and $\Delta(x,h,r)$ to represent $\Delta(x,h,\cB(x;r))$ for notation simplicity.

For any hypothesis $h$, let the strategic loss $\lossstr(h,(x,u,y))$ of $h$ be defined as the loss at the manipulated feature, i.e., $\lossstr(h,(x,u,y)) := \1{h(\Delta(x,h,u))\neq y}$. According to our definition of $\Delta(\cdot)$, we can write down the strategic loss explicitly as 
\begin{align}
    \lossstr(h,(x,u,y)) = \begin{cases}
        1 & \text{ if } y = -1, h(x) = +1 \\
        1 & \text{ if } y = -1, h(x)=-1 \text{ and } u \cap \cX_{h,+}\neq \emptyset\,,  \\
        1 & \text{ if } y = +1, h(x) = -1\text{ and } u \cap \cX_{h,+}= \emptyset\,,  \\
        0 & \text{ otherwise.} 
    \end{cases}
\end{align}
For any randomized predictor $p$ (a distribution over hypotheses), the strategic behavior depends on the realization of the predictor and the strategic loss of $p$ is $\lossstr(p,(x,u,y)) := \EEs{h\sim p}{\lossstr(h,(x,u,y))}$.

\paragraph{Online learning}
We consider the task of sequential classification where the learner aims to classify a sequence of agents $(x_1,u_1,y_1),(x_2,u_2,y_2),\ldots, (x_T,u_T,y_T)\in \cQ\times \cY$ that arrives in an online manner.
At each round, the learner feeds a predictor to the environment and then observes his prediction $\hat y_t$, the true label $y_t$ and possibly along with some additional information about the original/manipulated feature vectors.
We say the learner makes a mistake at round $t$ if $\hat y_t\neq y_t$ and the learner's goal is to minimize the number of mistakes on the sequence.
The interaction protocol (which repeats for $t=1,\ldots,T$) is described in the following.
\vspace{-1mm}
\begin{protocol}[H]
    \caption{Learner-Agent Interaction at round $t$}
    \label{prot:interaction-finite}
        \begin{algorithmic}[1]
            \STATE The environment picks an agent $(x_t,u_t,y_t)$ and reveals some context $C(x_t)$. In the online setting, the agent is chosen adversarially, while in the distributional setting, the agent is sampled i.i.d.
            \STATE The learner $\cA$ observes $C(x_t)$ and picks a hypothesis $f_t\in \cY^\cX$.
            \STATE The learner $\cA$ observes the true label $y_t$, the prediction $\hat y_t = f_t(\Delta_t)$, and some feedback $F(x_t,\Delta_t)$, where $\Delta_t = \Delta(x_t,f_t,u_t)$ is the manipulated feature vector.
        \end{algorithmic}
    \end{protocol}
\vspace{-4mm}
The context function {$C(\cdot)$} and feedback function {$F(\cdot)$} reveals information about the original feature vector $x_t$ and the manipulated feature vector $\Delta_t$. {$C(\cdot)$} reveals the information before the learner picks $f_t$ while {$F(\cdot)$} does after.
We study several different settings based on what and when information is revealed.
\begin{itemize}[leftmargin = *]
    \item The simplest setting for the learner is observing the original feature vector $x_t$ before choosing $f_t$ and the manipulated vector $\Delta_t$ after. Consider a teacher giving students a writing assignment or take-home exam. The teacher might have a good knowledge of the students' abilities (which correspond to the original feature vector $x_t$) based on their performance in class, but the grade has to be based on how well they do the assignment. The students might manipulate by using the help of ChatGPT / Google / WolframAlpha / their parents, etc. The teacher wants to create an assignment that will work well even in the presence of these manipulation tools. In addition, If we think of each example as representing a subpopulation (e.g., an organization is thinking of offering loans to a certain group), then there might be known statistics about that population, even though the individual classification (loan) decisions have to be made based on responses to the classifier.
    This setting corresponds to {$C(x_t) =x_t$} and {$F(x_t,\Delta_t) = \Delta_t$}.
    We denote a setting by their values of $C,F$ and thus, we denote this setting by $(x,\Delta)$.
    \item In a slightly harder setting, the learner observes both the original and manipulated vectors after selecting $f_t$ and thus, $f_t$ cannot depend on the original feature vector in this case. For example, if a high-school student takes the SAT test multiple times, most colleges promise to only consider the highest one (or even to "superscore" the test by considering the highest score separately in each section) but they do require the student to submit all of them. Then {$C(x_t) =\perp$} and {$F(x_t,\Delta_t) = (x_t,\Delta_t)$}, where $\perp$ is a token for ``no information'', and this setting is denoted by $(\perp, (x,\Delta))$.
    \item An even harder setting involves observing only the manipulated feature vector after selecting $f_t$ (which can only be revealed after $f_t$ since $\Delta_t$ depends on $f_t$). Then {$C(x_t) =\perp$} and {$F(x_t,\Delta_t) = \Delta_t$} and this setting is denoted by $(\perp, \Delta)$.
    \item The hardest and least informative scenario occurs when neither the original nor the manipulated feature vectors are observed. Then {$C(x_t) =\perp$} and {$F(x_t,\Delta_t) = \perp$} and it is denoted by $(\perp, \perp)$.


\end{itemize}
Throughout this work, we focus on the \textit{realizable} setting, where there exists a perfect classifier in $\cH$ that never makes any mistake at the sequence of strategic agents. 
More specifically, there exists a hypothesis $h^*\in \cH$ such that for any $t\in [T]$, we have $y_t = h^*(\Delta(x_t,h^*,u_t))$\footnote{It is possible that 
there is no hypothesis $\bar h\in \cY^\cX$ s.t. $y_t = \bar h(x_t)$ for all $t\in [T]$.}.
Then we define the mistake bound as follows.
\begin{definition}
For any choice of $(C,F)$, let $\cA$ be an online learning algorithm under Protocol~\ref{prot:interaction-finite} in the setting of $(C,F)$.
Given any realizable sequence $S = ((x_1,u_1, h^*(\Delta(x_1,h^*,u_1))),\ldots, (x_T,u_T, h^*(\Delta(x_T,h^*,u_T)))\in (\cQ\times \cY)^T$, where $T$ is any integer and $h^*\in \cH$, 
let $\cM_{\cA}(S)$ be the number of mistakes $\cA$ makes on the sequence $S$.
The mistake bound of $(\cH,\cQ)$, denoted $\MB_{C,F}$, is the smallest number $B\in \NN$ such that there exists an algorithm $\cA$ such that $\cM_{\cA}(S)\leq B$ over all realizable sequences $S$ of the above form.
\end{definition}
According the rank of difficulty of the four settings with different choices of $(C,F)$, the mistake bounds are ranked in the order of $\MB_{{x},{\Delta}} \leq \MB_{{\perp},{(x,\Delta)}} \leq \MB_{{\perp},{\Delta}} \leq \MB_{{\perp},{\perp}}$.


\paragraph{PAC learning} 
In the distributional setting, the agents are sampled from an underlying distribution $\cD$ over $\cQ\times \cY$.
The learner's goal is to find a hypothesis $h$ with low population loss $\err_\cD(h) := \EEs{(x,u,y)\sim \cD}{\lossstr(h,(x,u,y))}$. 
One may think of running empirical risk minimizer (ERM) over samples drawn from the underlying data distribution, i.e., returning $\argmin_{h\in \cH} \frac{1}{m}\sum_{i=1}^m \lossstr(h,(x_i,u_i,y_i))$, where $(x_1,u_1,y_1), \ldots,(x_m,u_m,y_m)$ are i.i.d. sampled from $\cD$.
However, ERM is unimplementable because the manipulation sets $u_i$'s are never revealed to the algorithm, and only the partial feedback in response to the implemented classifier is provided.  In particular, 
in this work we consider using the same interaction protocol as in the online setting, i.e., Protocol~\ref{prot:interaction-finite}, with agents $(x_t,u_t,y_t)$ i.i.d. sampled from the data distribution $\cD$.
After $T$ rounds of interaction (i.e., $T$ i.i.d. agents), the learner has to output a predictor $f_\out$ for future use. 

Again, we focus on the \textit{realizable} setting, where the sequence of sampled agents (with manipulation) can be perfectly classified by a target function in $\cH$.
Alternatively, there exists a classifier with zero population loss, i.e., there exists a hypothesis $h^*\in \cH$ such that $\err_\cD(h^*) = 0$. Then we formalize the notion of PAC sample complexity under strategic behavior as follows.
\begin{definition}
For any choice of $(C,F)$, let $\cA$ be a learning algorithm that interacts with agents using Protocol~\ref{prot:interaction-finite} in the setting of $(C,F)$ and outputs a predictor $f_\out$ in the end.
For any $\epsilon,\delta \in (0,1)$, the sample complexity of realizable $(\epsilon,\delta)$-PAC learning of $(\cH,\cQ)$, denoted $\SC_{C,F}(\epsilon,\delta)$, is defined as the smallest $m\in \NN$ for which there exists a learning algorithm $\cA$ in the above form such that for any distribution $\cD$ over $\cQ\times \cY$ where there exists a predictor $h^*\in \cH$ with zero loss, $\err_\cD(h) = 0$, with probability at least $1-\delta$ over $(x_1,u_1, y_1),\ldots,  (x_m,u_m,y_m)\stackrel{\text{i.i.d.}}{\sim} \cD$, $\err_\cD(f_\out)\leq \epsilon$.
\end{definition}
Similar to mistake bounds, the sample complexities are ranked in the same order $\SC_{x,\Delta} \leq \SC_{\perp,(x,\Delta)} \leq \SC_{\perp,\Delta} \leq \SC_{\perp,\perp}$ according to the rank of difficulty of the four settings.

\section{Overview of Results}\label{sec:overview-strategic}
In classic (non-strategic) online learning, the Halving algorithm achieves a mistake bound of $\log(\abs{\cH})$ by employing the majority vote and eliminating inconsistent hypotheses at each round.
In classic PAC learning, the sample complexity of $\cO(\frac{\log(\abs{\cH})}{\epsilon})$ is achievable via ERM.
Both mistake bound and sample complexity exhibit logarithmic dependency on $\abs{\cH}$.
This logarithmic dependency on $\abs{\cH}$ (when there is no further structural assumptions) is tight in both settings, i.e., there exist examples of $\cH$ with mistake bound of $\Omega(\log(\abs{\cH}))$ and with sample complexity of $\Omega(\frac{\log(\abs{\cH})}{\epsilon})$.
In the setting where manipulation is known beforehand and only $\Delta_t$ is observed, \cite{ahmadi2023fundamental} proved a lower bound of $\Omega(\abs{\cH})$ for the mistake bound.
Since in the strategic setting we can achieve a linear dependency on $\abs{\cH}$ by trying each hypothesis in $\cH$ one by one and discarding it once it makes a mistake, the question arises:
\begin{center}
    \textbf{Can we achieve a logarithmic dependency on $\abs{\cH}$ in strategic classification?}
\end{center}
In this work, we show that the dependency on $\abs{\cH}$ varies across different settings and that in some settings mistake bound and PAC sample complexity can exhibit different dependencies on $\abs{\cH}$.
We start by presenting our results for ball manipulations in the four settings.
\begin{itemize}[leftmargin = *]
    \item Setting of $(x,\Delta)$
    (observing $x_t$ before choosing $f_t$ and observing $\Delta_t$ after)
    : For online learning, we propose an variant of the Halving algorithm, called Strategic Halving (Algorithm~\ref{alg:halving}), which can eliminate half of the remaining hypotheses when making a mistake. The algorithm depends on observing $x_t$ before choosing the predictor $f_t$.
    Then by applying the standard technique of converting mistake bound to PAC bound, we are able to achieve sample complexity of 
    $\cO(\frac{\log(\abs{\cH})\loglog(\abs{\cH})}{\epsilon})$.
    \item Setting of $(\perp,(x,\Delta))$
    (observing both $x_t$ and $\Delta_t$ after selecting $f_t$)
    : 
    We prove that, there exists an example of $(\cH,\cQ)$ s.t. the mistake bound is lower bounded by $\Omega(\abs{\cH})$. 
   This implies that no algorithm can perform significantly better than sequentially trying each hypothesis, which would make at most $\abs{\cH}$ mistakes before finding the correct hypothesis.
   However, unlike the construction of mistake lower bounds in classic online learning, where all mistakes can be forced to occur in the initial rounds, we demonstrate that we require $\Theta(\abs{\cH}^2)$ rounds to ensure that all mistakes occur.
    In the PAC setting, we first show that, any learning algorithm with proper output $f_\out$, i.e., $f_\out\in\cH$, needs a sample size of $\Omega(\frac{\abs{\cH}}{\epsilon})$. 
    We can achieve a sample complexity of $O(\frac{\log^2(\abs{\cH})}{\epsilon})$ by executing Algorithm~\ref{alg:end-iid-ball}, which is a randomized algorithm with improper output.

    \item Setting of $(\perp,\Delta)$
    (observing only $\Delta_t$ after selecting $f_t$)
    : The mistake bound of $\Omega(\abs{\cH})$ also holds in this setting, as it is known to be harder than the previous setting. For the PAC learning, we show that any conservative algorithm, which only depends on the information from the mistake rounds, requires $\Omega(\frac{\abs{\cH}}{\epsilon})$ samples.
    The optimal sample complexity is left as an open problem.

    \item Setting of $(\perp,\perp)$
    (observing neither $x_t$ nor $\Delta_t$)
    : Similarly, the mistake bound of $\Omega(\abs{\cH})$ still holds.
    For the PAC learning, we show that the sample complexity is $\Omega(\frac{\abs{\cH}}{\epsilon})$ by reducing the problem to a stochastic linear bandit problem.
\end{itemize}

Then we move on to non-ball manipulations. However, we show that even in the simplest setting of observing $x_t$ before choosing $f_t$ and observing $\Delta_t$ after, there is an example of $(\cH,\cQ)$ such that the sample complexity is $\tilde\Omega(\frac{\abs{\cH}}{\epsilon})$. This implies that in all four settings of different revealed information, we will have sample complexity of $\tilde\Omega(\frac{\abs{\cH}}{\epsilon})$ and mistake bound of $\tilde\Omega(\abs{\cH})$.
We summarize our results in Table~\ref{tab:res}.

\begin{table}[htb!]
    \centering
    {
    \renewcommand{\arraystretch}{1.5}
    \begin{tabular}{c|c|c|c}
        & setting & mistake bound & sample complexity\\\hline
       \multirow{5}{*}{ball} & $(x,\Delta)$
       &  $ \Theta(\log(\abs{\cH}))$ (Thm~\ref{thm:halving}) & $\tilde \cO(\frac{\log(\abs{\cH})}{\epsilon})$$^a$ (Thm~\ref{thm:x-first-pac}), $\Omega(\frac{\log(\abs{\cH})}{\epsilon})$\\\cline{2-4}
        &\multirow{2}{*}{ $(\perp,(x,\Delta))$
        } & $\cO(\min(\sqrt{ \log(\abs{\cH})T},\abs{\cH}))$ (Thm~\ref{thm:mw})
         & $\cO(\frac{\log^2(\abs{\cH})}{\epsilon})$ (Thm~\ref{thm:x_end_pac}), $\Omega(\frac{\log(\abs{\cH})}{\epsilon})$\\
        & &
        $\Omega(\min(\frac{T}{\abs{\cH}\log(\abs{\cH})}, \abs{\cH}))$(Thm~\ref{thm:x-end-online})&  $\SC^{\text{prop}}= \Omega(\frac{\abs{\cH}}{\epsilon})$ (Thm~\ref{thm:x-end-pac-proper})
        \\\cline{2-4}
        & $(\perp,\Delta)$
        & $\Theta(\abs{\cH})$ (implied by Thm~\ref{thm:x-end-online}) & $\SC^{\text{csv}} = \tilde\Omega(\frac{\abs{\cH}}{\epsilon})$ (Thm~\ref{thm:delta-csv})
        \\\cline{2-4}
       & $(\perp,\perp)$
       & $\Theta(\abs{\cH})$ (implied by Thm~\ref{thm:x-end-online}) & $\tilde \cO(\frac{\abs{\cH}}{\epsilon})$ , $ \tilde\Omega(\frac{\abs{\cH}}{\epsilon})$ (Thm~\ref{thm:x-delta-never})
        \\\hline
       nonball & all & $ \tilde \Omega(\abs{\cH})$(Cor~\ref{cor:non-ball-all}) , $\cO(\abs{\cH})$ & $\tilde \cO(\frac{\abs{\cH}}{\epsilon})$ , $ \tilde\Omega(\frac{\abs{\cH}}{\epsilon})$ (Cor~\ref{cor:non-ball-all})\\\hline
    \end{tabular}}
    
    \raggedright \footnotesize{$^a$ A factor of $\loglog(\abs{\cH})$ is neglected.}
    \caption{The summary of results. $\tilde\cO$ and $\tilde \Omega$ ignore logarithmic factors on $\abs{\cH}$ and $\frac{1}{\epsilon}$.
    The superscripts prop stands for proper learning algorithms and csv stands for conservative learning algorithms. 
    All lower bounds in the non-strategic setting also apply to the strategic setting, implying that $\MB_{C,F}\geq \Omega(\log(\abs{\cH}))$ and $\SC_{C,F}\geq \Omega(\frac{\log(\abs{\cH})}{\epsilon})$ for all settings of $(C,F)$.
    In all four settings, a mistake bound of $\cO(\abs{\cH})$ can be achieved by simply trying each hypothesis in $\cH$ while the sample complexity can be achieved as $\tilde \cO(\frac{\abs{\cH}}{\epsilon})$ by converting the mistake bound of $\cO(\abs{\cH})$ to a PAC bound using standard techniques.}
    \label{tab:res}
\end{table}

\section{Ball manipulations}
In ball manipulations, when $\cB(x;r) \cap \cX_{h,+}$ has multiple elements, the agent will always break ties by selecting the one closest to $x$, i.e., $\Delta(x,h, r) = \argmin_{x'\in \cB(x;r)  \cap \cX_{h,+}} d(x,x')$.
In round $t$, the learner deploys predictor $f_t$, and once he knows $x_t$ and $\hat y_t$, he can calculate $\Delta_t$ himself without needing knowledge of $r_t$ by

\begin{align*}
    \Delta_t = \begin{cases}
    \argmin_{x'\in \cX_{f_t,+}} d(x_t,x') & \text{ if } \hat y_t = +1\,,\\
    x_t & \text{ if } \hat y_t = -1\,.
    \end{cases}
\end{align*}
Thus, for ball manipulations, knowing $x_t$ is equivalent to knowing both $x_t$ and $\Delta_t$.
\subsection{Setting $(x,\Delta)$: Observing $x_t$ Before Choosing $f_t$} 
\paragraph{Online learning} We propose a new algorithm with mistake bound of $\log(\abs{\cH})$ in setting $(x,\Delta)$.
To achieve a logarithmic mistake bound, we must construct a predictor $f_t$ such that if it makes a mistake, we can reduce a constant fraction of the remaining hypotheses.
The primary challenge is that we do not have access to the full information, and predictions of other hypotheses are hidden.
To extract the information of predictions of other hypotheses, we take advantage of ball manipulations, which induces an ordering over all hypotheses.
Specifically, for any hypothesis $h$ and feature vector $x$, we define the distance between $x$ and $h$ by the distance between $x$ and the positive region by $h$, $\cX_{h,+}$, i.e.,
\begin{equation}
    d(x,h) := \min\{ d(x,x')|x'\in \cX_{h,+}\}\,.\label{eq:dist}
\end{equation}
At each round $t$, given $x_t$, the learner calculates the distance $d(x_t,h)$ for all $h$ in the version space (meaning hypotheses consistent with history) and selects a hypothesis $f_t$ such that $d(x_t,f_t)$ is the median among all distances $d(x_t,h)$ for $h$ in the version space.
We can show that by selecting $f_t$ in this way, the learner can eliminate half of the version space if $f_t$ makes a mistake.
We refer to this algorithm as Strategic Halving, and provide a detailed description of it in Algorithm~\ref{alg:halving}.
\begin{theorem}\label{thm:halving}
    For any feature-ball manipulation set space $\cQ$ and hypothesis class $\cH$, Strategic Halving achieves mistake bound $\MB_{x,\Delta} \leq \log(\abs{\cH})$.
\end{theorem}
\vspace{-7pt}
\begin{algorithm}[H]\caption{Strategic Halving}\label{alg:halving}
    \begin{algorithmic}[1]
    \STATE Initialize the version space $\VS=\cH$.
    \FOR{$t=1,\ldots,T$}
    \STATE pick an $f_t\in \VS$ such that $d(x_t,f_t)$ is the median of $\{d(x_t,h)|h\in \VS\}$.\label{algline:pickf}
    \STATE \textbf{if} {$\hat y_t \neq y_t$ and $y_t = +$} \textbf{then}
     $\VS\leftarrow \VS\setminus \{h\in \VS|d(x_t,h)\geq d(x_t,f_t)\}$;
    \STATE \textbf{else if} {$\hat y_t \neq y_t$ and $y_t = -$} \textbf{then} $\VS\leftarrow  \VS\setminus \{h\in \VS|d(x_t,h)\leq d(x_t,f_t)\}$.
    \ENDFOR
    \end{algorithmic}
\end{algorithm}
\vspace{-11pt}
To prove Theorem~\ref{thm:halving}, we only need to show that each mistake reduces the version space by half.
Supposing that $f_t$ misclassifies a true positive example $(x_t,r_t, +1)$ by negative, then we know that $d(x_t, f_t)> r_t$ while the target hypothesis $h^*$ must satisfy that $d(x_t, h^*)\leq r_t$.
Hence any $h$ with $d(x_t,h)\geq d(x_t,f_t)$ cannot be $h^*$ and should be eliminated. Since $d(x_t,f_t)$ is the median of $\{d(x_t,h)|h\in \VS\}$, we can elimate half of the version space.
It is similar when $f_t$ misclassifies a true negative.
The detailed proof is deferred to Appendix~\ref{app:halving}.

\paragraph{PAC learning} 
We can convert Strategic Halving to a PAC learner by the standard technique of converting a mistake bound to a PAC bound \citep{10008965845}. Specifically, the learner runs Strategic Halving until it produces a hypothesis $f_t$ that survives for $\frac{1}{\epsilon}\log(\frac{\log(\abs{\cH})}{\delta})$ rounds and outputs this $f_t$.
Then we have Theorem~\ref{thm:x-first-pac}, and the proof is included in Appendix~\ref{app:x_firtst_pac}.
\begin{theorem}\label{thm:x-first-pac}
    For any feature-ball manipulation set space $\cQ$ and hypothesis class $\cH$, we can achieve $\SC_{x,\Delta}(\epsilon,\delta) = \cO(\frac{\log(\abs{\cH})}{\epsilon}\log(\frac{\log(\abs{\cH})}{\delta}))$ by combining Strategic Halving and the standard technique of converting a mistake bound to a PAC bound.
\end{theorem}



\subsection{Setting $(\perp, (x,\Delta))$: Observing $x_t$ After Choosing $f_t$} 
When $x_t$ is not revealed before the learner choosing $f_t$, the algorithm of Strategic Halving does not work anymore. 
We demonstrate that it is impossible to reduce constant fraction of version space when making a mistake, and 
prove that the mistake bound is lower bounded by $\Omega(\abs{\cH})$ by constructing a negative example of $(\cH,\cQ)$.
However, we can still achieve sample complexity with poly-logarithmic dependency on $\abs{\cH}$ in the distributional setting.
\subsubsection{Results in the Online Learning Model}
To offer readers an intuitive understanding of the distinctions between the strategic setting and standard online learning, we commence by presenting an example in which no deterministic learners, including the Halving algorithm, can make fewer than $\abs{\cH} -1$ mistakes.

\begin{example}\label{eg:x-after-det}
Consider a star shape metric space $(\cX,d)$, where $\cX = \{0,1,\ldots,n\}$, $d(i,j) = 2$ and $d(0,i)=1$  for all $i,j\in [n]$ with $i\neq j$.
The hypothesis class is composed of singletons over $[n]$, i.e., $\cH = \{2\ind{\{i\}}-1|i\in [n]\}$.
When the learner is deterministic, the environment can pick an agent $(x_t, r_t, y_t)$ dependent on $f_t$.
If $f_t$ is all-negative, then the environment picks $(x_t,r_t,y_t) = (0,1,+1)$, and then the learner makes a mistake but no hypothesis can be eliminated. 
If $f_t$ predicts $0$ by positive, the environment will pick $(x_t,r_t,y_t) = (0,0,-1)$, and then the learner makes a mistake but no hypothesis can be eliminated. 
If $f_t$ predicts some $i\in [n]$ by positive, the environment will pick $(x_t,r_t,y_t) = (i,0,-1)$, and then the learner makes a mistake with only one hypothesis $2\ind{\{i\}} -1$ eliminated.
Therefore, the learner will make $n-1$ mistakes.
\end{example}
\vspace{-3pt}

In this chapter, we allow the learner to be randomized. When an $(x_t,r_t,y_t)$ is generated by the environment, the learner can randomly pick an $f_t$, and the environment does not know the realization of $f_t$ but knows the distribution where $f_t$ comes from. 
It turns out that randomization does not help much.
We prove that there exists an example in which any (possibly randomized) learner will incur $\Omega(\abs{\cH})$ mistakes.

\begin{theorem}\label{thm:x-end-online}
     There exists a feature-ball manipulation set space $\cQ$ and hypothesis class $\cH$ s.t. the mistake bound $\MB_{\perp,(x,\Delta)}\geq \abs{\cH}-1$.
     For any (possibly randomized) algorithm $\cA$ and any $T\in \NN$,  there exists a realizable sequence of $(x_t,r_t,y_t)_{1:T}$ such that with probability at least $1-\delta$ (over randomness of $\cA$), $\cA$ makes at least $\min(\frac{T}{5\abs{\cH}\log(\abs{\cH}/\delta)}, \abs{\cH}-1)$ mistakes.
\end{theorem}
Essentially, we design an adversarial environment such that the learner has a probability of $\frac{1}{\abs{\cH}}$ of making a mistake at each round before identifying the target function $h^*$. The learner only gains information about the target function when a mistake is made.
The detailed proof is deferred to Appendix~\ref{app:x-end-online}.
Theorem~\ref{thm:x-end-online} establishes a lower bound on the mistake bound, which is $\abs{\cH}-1$. 
However, achieving this bound requires a sufficiently large number of rounds, specifically $T = \tilde \Omega (\abs{\cH}^2)$.
This raises the question of whether there exists a learning algorithm that can make $o(T)$ mistakes for any $T\leq \abs{\cH}^2$.
In Example~\ref{eg:x-after-det}, we observed that the adversary can force any deterministic learner to make $\abs{\cH}-1$ mistakes in $\abs{\cH}-1$ rounds. Consequently, no deterministic algorithm can achieve $o(T)$ mistakes.

To address this, we propose a randomized algorithm that closely resembles Algorithm~\ref{alg:halving}, with a modification in the selection of $f_t$. Instead of using line~\ref{algline:pickf}, we choose $f_t$ randomly from $\VS$ since we lack prior knowledge of $x_t$. This algorithm can be viewed as a variation of the well-known multiplicative weights method, applied exclusively during mistake rounds. For improved clarity, we present this algorithm as Algorithm~\ref{alg:mw} in Appendix~\ref{app:mw} due to space limitations.
\begin{theorem}\label{thm:mw}
    For any $T\in \NN$, Algorithm~\ref{alg:mw} will make at most $\min(\sqrt{ 4\log(\abs{\cH})T},\abs{\cH}-1)$ mistakes in expectation in $T$ rounds.
\end{theorem}
Note that the $T$-dependent upper bound in Theorem~\ref{thm:mw} matches the lower bound in Theorem~\ref{thm:x-end-online} up to a logarithmic factor when $T = \abs{\cH}^2$. This implies that approximately $\abs{\cH}^2$ rounds are needed to achieve $\abs{\cH}-1$ mistakes, which is a tight bound up to a logarithmic factor. Proof of Theorem~\ref{thm:mw} is included in
Appendix~\ref{app:mw}.

\subsubsection{Results in the PAC Learning Model}
In the PAC setting, the goal of the learner is to output a predictor $f_\out$ after the repeated interactions. 
A common class of learning algorithms, which outputs a hypothesis $f_\out \in \cH$, is called proper. 
Proper learning algorithms are a common starting point when designing algorithms for new learning problems due to their natural appeal and ability to achieve good performance, such as ERM in classic PAC learning.
However, in the current setting, we show that proper learning algorithms do not work well and require a sample size linear in $\abs{\cH}$.
The formal theorem is stated as follows and the proof is deferred to Appendix \ref{app:x-end-pac-proper}.
\begin{theorem}\label{thm:x-end-pac-proper}
     There exists a feature-ball manipulation set space $\cQ$ and hypothesis class $\cH$ s.t. $\SC_{\perp, \Delta}^\text{prop}(\epsilon, \frac{7}{8}) = \Omega(\frac{\abs{\cH}}{\epsilon})$, where $\SC_{\perp, \Delta}^\text{prop}(\epsilon,\delta)$ is the $(\epsilon,\delta)$-PAC sample complexity achievable by proper algorithms.
\end{theorem}
Theorem~\ref{thm:x-end-pac-proper} implies that any algorithm capable of achieving sample complexity sub-linear in $\abs{\cH}$ must be improper. As a result, we are inspired to devise an improper learning algorithm. Before presenting the algorithm, we introduce some notations.
For two hypotheses $h_1,h_2$, let $h_1 \vee h_2$ denote the union of them, i.e., $(h_1 \vee h_2)(x) = +1$ iff. $h_1(x)=+1$ or $h_2(x)=+1$.
Similarly, we can define the union of more than two hypotheses. 
Then for any union of $k$ hypotheses, $f= \vee_{i=1}^k h_i$, the positive region of $f$ is the union of positive regions of the $k$ hypotheses and thus, we have $d(x,f) = \min_{i\in [k]} d(x,h_i)$.
Therefore, we can decrease the distance between $f$ and any feature vector $x$ by increasing $k$. 
Based on this, we devise a new randomized algorithm with improper output, described in Algorithm~\ref{alg:end-iid-ball}.
\begin{theorem}\label{thm:x_end_pac}
For any feature-ball manipulation set space $\cQ$ and hypothesis class $\cH$, we can achieve $\SC_{\perp, (x,\Delta)}(\epsilon,\delta) = \cO(\frac{\log^2(\abs{\cH}) +\log(1/\delta)}{\epsilon}\log(\frac{1}{\delta}))$ by combining Algorithm~\ref{alg:end-iid-ball} with a standard confidence boosting technique. Note that the algorithm is improper.
\end{theorem}
\begin{algorithm}[H]\caption{}\label{alg:end-iid-ball}
    \begin{algorithmic}[1]
    \STATE Initialize the version space $\VS_0=\cH$.
    \FOR{$t=1,\ldots,T$}
    \STATE randomly pick $k_t\sim\Unif(\{1, 2, 2^2,\ldots,2^{\floor{\log_2(n_t)-1}}\})$ where $n_t = \abs{\VS_{t-1}}$;\label{algline:choosef1}
    \STATE sample $k_t$ hypotheses $h_1,\ldots,h_{k_t}$ independently and uniformly at random from $\VS_{t-1}$;
    \STATE let $f_t = \vee_{i=1}^{k_t} h_i$.\label{algline:choosef2}
    \STATE \textbf{if} {$\hat y_t \neq y_t$ and $y_t = +$} \textbf{then} $\VS_t= \VS_{t-1}\setminus \{h\in \VS_{t-1}|d(x_t,h)\geq d(x_t,f_t)\}$;\label{algline:vsupdate1}
    \STATE \textbf{else if} {$\hat y_t \neq y_t$ and $y_t = -$} \textbf{then} $\VS_t= \VS_{t-1}\setminus\{h\in \VS_{t-1}|d(x_t,h)\leq d(x_t,f_t)\}$;\label{algline:vsupdate2}
    \STATE \textbf{else} $\VS_t = \VS_{t-1}$.
    \ENDFOR
    \STATE randomly pick $\tau$ from $[T]$ and randomly sample $h_1, h_2$ from $\VS_{\tau-1}$ with replacement.\label{algline:fout1}
    \STATE \textbf{output} $h_1 \vee h_2$\label{algline:fout2}
    \end{algorithmic}
\end{algorithm}
Now we outline the high-level ideas behind Algorithm~\ref{alg:end-iid-ball}.
In correct rounds where $f_t$ makes no mistake, the predictions of all hypotheses are either correct or unknown, and thus, it is hard to determine how to make updates.
In mistake rounds, we can always update the version space similar to what was done in Strategic Halving. To achieve a poly-logarithmic dependency on $\abs{\cH}$, 
we aim to reduce a significant number of misclassifying hypotheses in mistake rounds. The maximum number we can hope to reduce is a constant fraction of the misclassifying hypotheses.
We achieve this by randomly sampling a $f_t$ (lines~\ref{algline:choosef1}-\ref{algline:choosef2}) s.t. $f_t$ makes a mistake, and 
$d(x_t,f_t)$ is greater (smaller) than the median of $d(x_t,h)$ for all misclassifying hypotheses $h$ for true negative (positive) examples.
However, due to the asymmetric nature of manipulation, which aims to be predicted as positive, the rate of decreasing misclassifications over true positives is slower than over true negatives.
To compensate for this asymmetry, we output a $f_\out = h_1 \vee h_2$ with two selected hypotheses $h_1,h_2$ (lines \ref{algline:fout1}-\ref{algline:fout2}) instead of a single one to increase the chance of positive prediction.

We prove that Algorithm~\ref{alg:end-iid-ball} can achieve small strategic loss in expectation as described in Lemma~\ref{lmm:alg-exp}. 
Then we can achieve the sample complexity in Theorem~\ref{thm:x_end_pac} by boosting Algorithm~\ref{alg:end-iid-ball} to a strong learner.
This is accomplished by running Algorithm~\ref{alg:end-iid-ball} multiple times until we obtain a good predictor.
The proofs of Lemma~\ref{lmm:alg-exp} and Theorem~\ref{thm:x_end_pac} are deferred to Appendix~\ref{app:x_end_pac}.
\begin{lemma}\label{lmm:alg-exp}
Let $S = (x_t,r_t,y_t)_{t=1}^T\sim \cD^T$ denote the i.i.d. sampled agents in $T$ rounds and let $\cA(S)$ denote the output of Algorithm~\ref{alg:end-iid-ball} interacting with $S$.
For any feature-ball manipulation set space $\cQ$ and hypothesis class $\cH$, when $T\geq \frac{320\log^2(\abs{\cH})}{\epsilon}$, we have $\EEs{\cA,S}{\err(\cA(S))}\leq \epsilon$.
\end{lemma}


\subsection{Settings $(\perp, \Delta)$ and $(\perp, \perp)$
}\label{sec:ball-delta}
\paragraph{Online learning}
As mentioned in Section~\ref{sec:model-strategic}, both the settings of $(\perp,\Delta)$ and ${(\perp,\perp)}$ are harder than the setting of $(\perp,(x,\Delta))$, all lower bounds in the setting of $(\perp,(x,\Delta))$ also hold in the former two settings.
Therefore, by Theorem~\ref{thm:x-end-online}, we have $\MB_{\perp,\perp} \geq \MB_{\perp,\Delta}\geq \MB_{\perp,(x,\Delta)} =\abs{\cH}-1$.

\paragraph{PAC learning} 
In the setting of $(\perp,\Delta)$, Algorithm~\ref{alg:end-iid-ball} is not applicable anymore since the learner lacks observation of $x_t$, making it impossible to replicate the version space update steps in lines \ref{algline:vsupdate1}-\ref{algline:vsupdate2}.
It is worth noting that both PAC learning algorithms we have discussed so far fall under a general category called conservative algorithms, depend only on information from the mistake rounds.
Specifically, an algorithm is said to be conservative if for any $t$, the predictor $f_t$ only depends on the history of mistake rounds up to $t$, i.e., $\tau<t$ with $\hat y_\tau \neq y_\tau$, and the output $f_\out$ only depends on the history of mistake rounds, i.e., $(f_t,\hat y_t, y_t, \Delta_t)_{t:\hat y_t \neq y_t}$.
Any algorithm that goes beyond this category would need to utilize the information in correct rounds. As mentioned earlier, in correct rounds, the predictions of all hypotheses are either correct or unknown, which makes it challenging to determine how to make updates.
For conservative algorithms, we present a lower bound on the sample complexity in the following theorem, which is $\tilde\Omega(\frac{\abs{\cH}}{\epsilon})$, and its proof is included in Appendix~\ref{app:delta-csv}. The optimal sample complexity in the setting $(\perp,\Delta)$ is left as an open problem.



\begin{theorem}\label{thm:delta-csv}
There exists a feature-ball manipulation set space $\cQ$ and hypothesis class $\cH$ s.t. $\SC_{\perp,\Delta}^\text{csv}(\epsilon, \frac{7}{8}) = \tilde\Omega(\frac{\abs{\cH}}{\epsilon})$, where $\SC_{\perp,\Delta}^\text{csv}(\epsilon,\delta)$ is $(\epsilon,\delta)$-PAC the sample complexity achievable by conservative algorithms.
\end{theorem}




In the setting of $(\perp,\perp)$, our problem reduces to a best arm identification problem in stochastic bandits. We prove a lower bound on the sample complexity of $\tilde\Omega(\frac{\abs{\cH}}{\epsilon})$ in Theorem~\ref{thm:x-delta-never} by reduction to stochastic linear bandits and applying the tools from information theory. The proof is deferred to Appendix~\ref{app:x-delta-never}.
\begin{theorem}\label{thm:x-delta-never}
    There exists a feature-ball manipulation set space $\cQ$ and hypothesis class $\cH$ s.t. $\SC_{\perp,\perp}(\epsilon, \frac{7}{8})= \tilde\Omega(\frac{\abs{\cH}}{\epsilon})$.
\end{theorem}

\section{Non-ball Manipulations}
In this section, we move on to non-ball manipulations. In ball manipulations, for any feature vector $x$, we have an ordering of hypotheses according to their distances to $x$, which helps to infer the predictions of some hypotheses without implementing them.
However, in non-ball manipulations, we don't have such structure anymore.
Therefore, even in the simplest setting of observing $x_t$ before $f_t$ and $\Delta_t$, we have the PAC sample complexity lower bounded by $\tilde\Omega(\frac{\abs{\cH}}{\epsilon})$.
\begin{theorem}\label{thm:non-ball}
    There exists a feature-manipulation set space $\cQ$ and hypothesis class $\cH$ s.t. $\SC_{x,\Delta}(\epsilon, \frac{7}{8})= \tilde\Omega(\frac{\abs{\cH}}{\epsilon})$.
\end{theorem}
The proof is deferred to Appendix~\ref{app:non-ball}. It is worth noting that in the construction of the proof, we let all agents to have their original feature vector $x_t = \bZero$ such that $x_t$ does not provide any information.
Since $(x,\Delta)$ is the simplest setting and any mistake bound can be converted to a PAC bound via standard techniques (see Section~\ref{app:mistake2pac} for more details), we have the following corollary.
\begin{corollary}\label{cor:non-ball-all}
    There exists a feature-manipulation set space $\cQ$ and hypothesis class $\cH$ s.t. for all choices of $(C,F)$, $\SC_{C,F}(\epsilon, \frac{7}{8})= \tilde\Omega(\frac{\abs{\cH}}{\epsilon})$ and $\MB_{C,F}= \tilde \Omega(\abs{\cH})$.
\end{corollary}

\section{Discussion and Open Problems}
In this chapter, we investigate the mistake bound and sample complexity of strategic classification over finite hypothesis class across multiple settings.
Unlike prior work, we assume that the manipulation is personalized and unknown to the learner, which makes the strategic classification problem more challenging.
In the case of ball manipulations, when the original feature vector $x_t$ is revealed prior to choosing $f_t$, the problem exhibits a similar level of difficulty as the non-strategic setting (see Table~\ref{tab:res} for details).
However, when the original feature vector $x_t$ is not revealed beforehand, the problem becomes significantly more challenging. 
Specifically, any learner will experience a mistake bound that scales linearly with $\abs{\cH}$, and any proper learner will face sample complexity that also scales linearly with $\abs{\cH}$.
In the case of non-ball manipulations, the situation worsens. Even in the simplest setting, where the original feature is observed before choosing $f_t$ and the manipulated feature is observed afterward, any learner will encounter a linear mistake bound and sample complexity. 

We primarily concentrate on the realizable setting. However, investigating the sample complexity and regret bounds in the agnostic setting would be an interesting avenue for future research.

\chapter{Strategic Classification for Infinite Hypothesis Class}\label{chap:strategic-infinite}
In the previous chapter, we discuss strategic classification for finite hypothesis class, i.e., $\abs{\cH}<\infty$. In this chapter, we consider infinite hypothesis class, $\abs{\cH}=\infty$.
\section{Introduction}
When modeling strategic classification, the manipulation power of the agents is considered limited. 
In this model, each agent has a feature vector $x$, represented as a node in a graph, and a binary label, $y$. An arc from $x$ to~$x'$ indicates that an agent with the feature vector $x$ can adjust their feature vector to $x'$—essentially, an arc denotes a potential manipulation.\footnote{The neighborhood is actually identical to manipulation set in the previous chapter. We use different names in this chapter.}
Similar to the assumptions made in \cite{ahmadi2023fundamental}, we posit that the manipulation graph has a bounded degree, $k$. This constraint adds a realistic dimension to the model, reflecting the limited manipulation capabilities of the agents within the strategic classification framework. In addition, as we discuss in Theorems~\ref{thm:fi-pac-lb} and~\ref{thm:fi-online-lb}, this bounded degree constraint is also necessary in the sense that relaxing it leads to impossibility results for extremely easy-to-learn hypothesis classes. 
%


In this chapter, we aim to address a fundamental question:
\begin{center}
    \textbf{Does learnability imply strategic learnability?}
\end{center}
We study the above question within both PAC and online learning frameworks, addressing it across both realizable and agnostic contexts.  In each case, we analyze various levels of information received by the learner as feedback. 

%
\vspace{1mm}

We start with a \textit{fully informative} setting.  Here, the learner has full information of both the manipulation graph as well as the pre- and post-manipulation features: at each time step the learner (1) observes the pre-manipulation features of the current agent, (2)  selects and implements a hypothesis 
(according to which the agent manipulates), and (3) observes both the manipulated features and the true label of the agent. 

We are mainly interested in this model because it is the simplest and therefore a natural first step from a theoretical perspective. Nevertheless, it still could be interesting from a practical perspective because it aligns with real-world situations like customers inquiring about mortgage rates and providing their credit score or income as input, or in college admissions where colleges can access sophomore-year grades. 
In other scenarios, such as those introduced by  by~\cite{shao2023strategic}, pre-manipulation features are known. 
For instance, when a teacher assigns a writing task or take-home exam to students. While the teacher may have knowledge of students' abilities (pre-existing features) from class performance, grading must objectively evaluate the work submitted for that specific assignment.
 Students might attempt manipulation using resources like ChatGPT. The teacher aims to design assignments resilient to such manipulation tools. 
This basic setting already deviates from the standard learning paradigm, as the learner needs to consider the potential manipulation of the current agents. 


\vspace{1mm}

We consider the fully informative setting as a baseline. The other two settings we consider deviate from this baseline by introducing two pragmatic elements of uncertainty regarding the learner's knowledge: 
one concerning the pre-manipulation features (previously studied for finite hypotheses class in \cite{ahmadi2023fundamental}), and the other regarding the manipulation graph (this type of uncertainty is introduced in this work).


In the second setting, \textit{post-manipulation feedback}, the learner selects the hypothesis before observing any information about the agent. Then, after the agent's manipulation and its classification, 
the learner observes either (i) the original feature vector, or (ii) the manipulated feature vector (we consider both variants). 
Thus, significantly less information is provided to the learner compared to the first setting, wherein the learner could compute the set of potential manipulations for each feature vector $x$ for every hypothesis. 
This setting was previously studied in \cite{ahmadi2023fundamental} which focused on finite hypotheses classes. 
There are also scenarios in which the original features are observed afterward--- for example, when a high-school student takes the SAT test multiple times, most colleges commit to only consider the highest one (or even to ``superscore'' the test by considering the highest score separately in each section), and some colleges require students to submit all SAT scores (\href{https://blog.collegeboard.org/what-is-an-sat-superscore}{reference}).


We highlight that while we assume knowledge of the manipulation graph for the first two settings, our algorithms exclusively depend on local information concerning the manipulation set of the current feature vector. Thus, our algorithms remain applicable in scenarios where global information may not be readily available.

\vspace{1mm}

In the last setting, the \textit{unknown manipulation graph}, we relax the fully informative baseline by introducing uncertainty about the manipulation graph. So, the learner observes both the pre- and post-manipulation feature vectors, but no longer has access to the entire manipulation graph. Instead, the learner has prior knowledge in the form of a finite class of graphs $\cG$. The class $\cG$ is assumed to contain the true manipulation graph in the realizable setting, and in the agnostic setting, $\cG$ serves as a comparator class for competitive/regret analysis.
This setting is a natural variant of the 
fully informative feedback model. For instance, a lender can observe the credit score of a customer but does not know the (at most $k$) easiest and most effective types of manipulations for the agent to pursue —whether they involve obtaining more credit cards, increasing their salary, or even changing employment status from unemployed to employed. 


From a technical perspective, the last setting gives rise to a natural graph learning task: consider an unknown target graph. At training time, we observe random examples of the form (vertex $v$, and a random arc from $v$). During testing, we receive a random vertex $u$ (drawn from the same distribution as the training nodes), and our objective is to predict the entire neighborhood of $u$. 
This problem is motivated by applications such as recommendation systems like Netflix, where a learner observes a random user (vertex) and a random movie they watched (random arc). During testing, the learner observes a user and aims to predict a small set of movies to recommend to this user.
Our algorithm in the unknown manipulation graph setting can be applied to solve this neighborhood learning problem.


\subsection{Results Overview} 

In classical binary classification, it is well-known that Probably Approximately Correct (PAC) learnability is characterized by the Vapnik–Chervonenkis (VC) dimension \citep{vapnik:74} and that online learnability is characterized by the Littlestone dimension \citep{littlestone1988learning} (see Appendix~\ref{app:background} for their definitions). As detailed below, we essentially show that \emph{every learnable class is also strategically learnable}. For certain scenarios, we efficiently transform standard learning algorithms into their strategic counterparts, while for others, we develop entirely new algorithms. Please see Table~\ref{tb:results} for a summary according to different settings. 

\textbf{Fully Informative Setting}
We establish upper and lower bounds for the sample complexity and mistake bounds, which are tight up to logarithmic factors. 
In particular, our bounds imply that in this basic setting, learnability implies strategic learnability. Quantitatively, we prove that the learning complexity of strategic learning is larger by at most a multiplicative logarithmic factor of the maximal degree of the manipulation graph. We further prove that this logarithmic factor is necessary, which implies that manipulation indeed makes learning harder.


In this setting, both PAC and online learnability are captured by the VC and Littlestone dimension of the corresponding classes of strategic losses (i.e.\ the 0-1 loss under manipulations). In the PAC setting we show that the strategic VC dimension is $\tilde \Theta(d_1\log k)$, where $d_1$ is the standard VC dimension of~$\cH$. This bound is tight up to a logarithmic factor in $d_1$ (see Theorems~\ref{thm:fi-pac} and~\ref{thm:fi-pac-lb}). 
In the online setting we provide a tight bound of $d_2\log k$ on the strategic Littlestone dimension of $\cH$, where $d_2$ is the standard Littlestone dimension of $\cH$ (see Theorems~\ref{thm:fi-online} and ~\ref{thm:fi-online-lb}).

Algorithmically, in the PAC setting any empirical risk minimizer w.r.t.\ the strategic losses achieves near optimal sample complexity in both the realizable and agnostic setting. In the online setting we construct an efficient transformation that converts an online algorithm in the standard setting to a strategic online algorithm. Our transformation involves using the algorithm from the standard setting to generate multiple experts, and running a weighted-majority vote scheme over these experts.

\textbf{Post-Manipulation Feedback Setting}
In the PAC setting, we derive sample complexity bounds identical to those that appear in the fully informative setting. 
This occurs because, when the graph is known, the timing of observing the original feature---whether beforehand or afterward---does not make any difference. Even if the learner can only observe manipulated features, by implementing the all-positive or all-negative hypothesis at each round, one can obtain the same information feedback as observing original features.

However, in online learning,  this setting is more subtle because we cannot separate exploration (training) from exploitation (test). To handle this we again devise an efficient transformation that that converts online learners from the classical setting to strategic online learners.
Our algorithm here is similar to the one from the fully informative setting, yet the optimal mistake bound here is $d_2 k$, which has an exponentially worse dependence on $k$ (see Theorems~\ref{thm:fu-online} and~\ref{thm:fu-online-lb}). This optimal mistake bound also resolves an open question posed by \cite{ahmadi2023fundamental}.
 

\textbf{Unknown Manipulation Graph Setting}
We answer our overarching question affirmatively for this setting too. 
The main challenge of PAC strategic learning with an unknown graph setting is that we cannot obtain an unbiased estimate of the loss of each graph, and simply running ERM does not work. We tackle this challenge by using the expected degree as a regularizer over the graphs.
Then, in the realizable setting, by finding the consistent minimal empirical degree graph-hypothesis pair, we can obtain sample complexity 
$\tilde O({(d_1\log k + k\log |\cG|)}/{\epsilon})$ (Theorem~\ref{thm:ug-pac-rel}).

In the agnostic PAC setting, although we cannot obtain an unbiased estimate of the graph loss, we propose a proxy loss, which can approximate the graph loss up to a multiplicative factor $k$. By minimizing the empirical proxy loss to find an approximate graph and running ERM over the strategic loss under this approximate graph, we are able to obtain a hypothesis that is at most a factor of $k$ from the best hypothesis-graph pair (see Theorem~\ref{thm:ug-pac-agn}).

In the realizable online setting, we design an algorithm that runs the online learning algorithm from the post-manipulation feedback setting over the majority vote of the graphs and achieves a mistake bound of $O(d_2k\log(k)+ \log |\cG|)$ (see Theorem~\ref{thm:ug-online}). In both realizable PAC and online settings, we provide a lower bound with logarithmic dependency on $|\cG|$, demonstrating that $\log|\cG|$ is inherent (see Theorems~\ref{thm:ug-pac-rel-lb} and~\ref{thm:ug-online-lb}).
We further discuss the application of our results in the unknown graph setting to the multi-label learning problem, which is of independent interest. 

\begin{table}[t]
\centering
\resizebox{\textwidth}{!}{%
\renewcommand{\arraystretch}{2}
\begin{tabular}{|c|c|c|c|}
\hline
\multirow{2}{*}{Setting}& \multicolumn{2}{c|}{Sample complexity} & Regret\\\cline{2-4}
 & Realizable & Agnostic  & Realizable$^{(*)}$   \\ \hline
\makecell{Fully informative} & \multirow{2}{*}{\makecell{
$\tilde O(\frac{d_1\log k}{\epsilon})$ 
(Thm~\ref{thm:fi-pac})\\
$\Omega(\frac{d_1\log k}{\epsilon})$ (Thm~\ref{thm:fi-pac-lb})
}}  & \multirow{2}{*}{\makecell{
$\tilde O(\frac{d_1\log k}{\epsilon^2})$ (Thm~\ref{thm:fi-pac})\\
$\Omega(\frac{d_1\log k}{\epsilon^2})$ (Thm~\ref{thm:fi-pac-lb})}} & \makecell{$O(d_2 \log k)$
(Thm~\ref{thm:fi-online})\\ $\Omega(d_2\log k )$ (Thm~\ref{thm:fi-online-lb})} \\ \cline{1-1}\cline{4-4}
\makecell{Post-manipulation\\ feedback} &   &   & \makecell{ $\tilde O(d_2 k)$ (Thm~\ref{thm:fu-online})\\
$\Omega(d_2k)$  
 (Det, Thm~\ref{thm:fu-online-lb})}  \\ \hline
\makecell{Unknown\\ manipulation graph} & \makecell{$\tilde O(\frac{d_1\log k + k\log(|\cG|)}{\epsilon})$ (Thm~\ref{thm:ug-pac-rel})\\ $\Omega(\log|\cG|)$ (Ada, Thm~\ref{thm:ug-pac-rel-lb})} & \makecell{$O(\frac{k^2\log(|\cG|) + d_1\log k}{\epsilon^2})$\\
 (Mul, Thm~\ref{thm:ug-pac-agn})} & \makecell{$O(d_2k\log k+ \log |\cG|)$ (Thm~\ref{thm:ug-online})\\
$\tilde\Omega(\log|\cG|)$ (Det, Thm~\ref{thm:ug-online-lb})} \\ \hline
\end{tabular}
}
\caption{Summary of results, where $k$ is the upper bound of the maximum out-degree of the underlying manipulation graph $G^\star$ as well as all graphs in $\cG$, $d_1 = \vcd(\cH)$ is the VC dimension of $\cH$, and $d_2 = \ld(\cH)$ is the Littlestone dimension of $\cH$. \textit{Det} means a lower bound for all deterministic algorithms. \textit{Ada} means in this lower bound, the true graph is generated adaptively. \textit{Mul} means that there is a multiplicative factor of $k$ in the loss. 
$ ^{(*)}$ Some results can be extended to the agnostic case by standard techniques as discussed in Appendix~\ref{app:online-agn}.}
\label{tb:results}
\end{table}

\section{Model}

\subsection{Strategic Classification}
Throughout this work, we focus on the binary classification task in both adversarial online setting and PAC setting.
Let $\cX$ denote the feature vector space, $\cY=\{0,1\}$ denote the label space, and $\cH\subset \cY^\cX$ denote the hypothesis class.
In the strategic setting, an example $(x,y)$, referred to as an \textit{agent} in this context, consists of a pair of feature vector and label.
An agent will try to receive a positive prediction by modifying their feature vector to some reachable feature vectors. 
Following prior works (e.g., ~\cite{zhang2021incentive,lechner2022learning,ahmadi2023fundamental,lechner2023strategic}), we model the set of reachable feature vectors as a manipulation graph $G = (\cX, \cE)$, where the nodes are all feature vectors in $\cX$. 
For any two feature vectors $x, x'\in \cX$, there is a directed arc $(x,x')\in \cE$ from $x$ to $x'$ if and only if an agent with feature vector $x$ can manipulate to $x'$. 
We let $N_G(x) = \{x'|(x,x')\in \cE\}$ denote the out-neighborhood of $x$ in graph $G$ and $N_G[x] = \{x\}\cup N_G(x)$ denote the closed out-neighborhood. 
Let $k(G)$ denote the maximum out-degree of graph $G$. 
When a hypothesis $h\in \cY^\cX$ is implemented, if an agent $(x,y)$ is classified by $h$ as negative at their original feature vector $x$ but has a neighbor which is classified as positive by $h$, she will manipulate to such neighbor.
Otherwise, she will stay at her original feature vector, $x$. Either way, the label of the agent remains~$y$.
Formally,
\begin{definition}[Manipulated feature]
    Given a manipulation graph $G=(\cX,\cE)$ and a hypothesis $h$, the manipulated feature vector of $x$ induced by $(G,h)$ is
    \vspace{-0.5em}
    \begin{equation*}
        \pi_{G,h}(x) =
        \begin{cases}
            v \in \cX_{h,+}\cap N_G(x) & \text{if } h(x) = 0 \text{ and }\cX_{h,+}\cap N_G(x)\neq \emptyset,  \\
            x &\text{otherwise,}
        \end{cases}
    \end{equation*}
    \vspace{-0.5em}
    
    where $\cX_{h,+}=\{x'\in \cX| h(x')=1\}$ is the positive region by $h$, and the tie-breaking here is assumed to be arbitrary unless specified otherwise.
\end{definition}
The post-manipulation labeling of the agent with original feature vector $x$ by a hypothesis $h$ is determined by their prediction at the resulting manipulated feature vector, $\pi_{G,h}(x)$.
\begin{definition}
[Manipulation-induced labeling]
    Given a manipulation graph $G=(\cX,\cE)$ and a hypothesis~$h$, the labeling of $x$ induced by $(G,h)$ is defined as $\bar h_G(x) = h(\pi_{G,h}(x))$.
We define $\bar \cH_{G} = \{\bar h_G|h\in \cH\}$ to be the $(G,\cH)$-induced labeling class of hypothesis class $\cH$.
\end{definition}
When implementing a hypothesis $h$, the loss incurred at the agent $(x,y)$ is the misclassification error when the agent manipulates w.r.t.\ $h$. We define the strategic loss as follows.
\begin{definition}[Strategic loss]\label{def:lstr}
    Given the manipulation graph $G$, for any hypothesis $h\in \cY^\cX$, the strategic loss of $h$ at agent $(x,y)$ is
    \vspace{-0.5em}
    \begin{equation*}
        \lstr_G(h, (x,y)) := \1{\bar h_G(x)\neq y}\,.
    \end{equation*}
    \vspace{-0.5em}
\end{definition}
\vspace{-1em}

%


\subsection{Learning with Strategic Agents}
Let $G^\star$ denote the underlying true manipulation graph and let $k = k(G^\star)$ denote the maximum out-degree of~$G^\star$.
In the strategic setting, a sequence of agents arrives and interacts with the learner.
The learner perceives agents through the lens of the currently implemented hypothesis, and 
lacks the ability to counterfactually inquire about the manipulations previous agents would have adopted under a different hypothesis. 
More formally,
consider a sequence of $T$ agents $(x_1,y_1),\ldots,(x_T,y_T)$ arriving sequentially.
At each round $t$, an agent $(x_t,y_t)$ arrives. Then, the learner implements a hypothesis $h_t\in \cY^{\cX}$ and reveals $h_t$ to the agent.
After observing $h_t$, the agent responds by manipulating her feature vector from $x_t$ to $v_t =\pi_{G^\star,h_t}(x_t)$.
At the end of the round, the learner 
receives the result of her prediction on the manipulated feature, $\hat y_t = h_t(v_t)$,
and the true label, $y_t$. 
Again, we consider both the PAC and adversarial online setting. 

In the PAC setting, the agents are sampled from an underlying distribution $\cD$ over $\cX\times \cY$.
For any data distribution $\cD$ over $\cX\times \cY$, the strategic population loss of $h$ is defined as
    \begin{equation*}
        \Lstr_{G^\star, \cD}(h) := \EEs{(x,y)\sim \cD}{\lstr_{G^\star}(h,(x,y))}\,.
    \end{equation*}
In the \textit{realizable} case, there exists a perfect classifier $h^\star\in \cH$ with strategic population loss $\Lstr_{G^\star,\cD}(h^\star) = 0$.

In the online setting, the agents are generated by an adversary. The learner's goal is to minimize the total number of mistakes, $\sum_{t=1}^T\1{\hat y_t \neq y_t}$.
More specifically, the learner's goal is to minimize the Stackelberg regret\footnote{This is 
Stackelberg regret since the learner selects a hypothesis $h_t$ and assumes that the agent will best respond to $h_t$.}  with respect to the best hypothesis $h^\star\in \cH$ in hindsight, had the agents manipulated given $h^\star$:

\begin{equation*}
    \regret(T) := \sum_{t=1}^T \lstr_{G^\star}(h_t,(x_t,y_t)) - \min_{h^\star\in \cH} \sum_{t=1}^T\lstr_{G^\star}(h^\star,(x_t,y_t))
\end{equation*}

In the \textit{realizable} case, for the sequence of agents $(x_1,y_1),\ldots,(x_T,y_T)$,  there exists a perfect hypothesis $h^\star\in \cH$ that makes no mistakes, i.e., $\lstr_{G^\star}(h^\star,(x_t,y_t)) = 0$ for all $t=1,\ldots, T$.
In this case, the Stackelberg regret is reduced to the number of mistakes.


\subsection{Uncertainties about Manipulation Structure and Features}\label{sec:model-uncertainty}
We begin with the simplest fully informative setting, where the underlying manipulation graph $G^\star$ is known, and the original feature vector $x_t$ is revealed to the learner before the implementation of the hypothesis. 
We use this fully informative setting as our baseline model and investigate two forms of uncertainty: either the original feature vector $x_t$ is not disclosed beforehand or the manipulation graph $G^\star$ is unknown.
\begin{enumerate}
    \item \textbf{Fully Informative Setting:} Known manipulation graph and known original feature beforehand. More specifically, the true manipulation graph $G^\star$ is known, and $x_t$ is observed before the learner implements~$h_t$.  This setting is the simplest and serves as our baseline, providing the learner with the most information.
    Some may question the realism of observing $x_t$ beforehand, but there are scenarios in which this is indeed the case. For instance, when customers inquire about mortgage rates and provide their credit score or income as input, or in college admissions where colleges can access sophomore-year grades. 
    In cases where each agent represents a subpopulation and known statistics about that population are available, pre-observation of $x_t$ can be feasible.

    \item \textbf{Post-Manipulation Feedback Setting:} 
    Here the underlying graph $G^\star$ is still known, but the original feature vector $x_t$ is not observable before the learner implements the hypothesis. Instead, either the original feature vector $x_t$ or the manipulated feature vector $v_t$ are revealed afterward (we consider both variants). It is worth noting that, since the learner can compute $v_t= \pi_{G^\star,h_t}(x_t)$ on her own given $x_t$, knowing $x_t$ is more informative than knowing~$v_t$. \cite{ahmadi2023fundamental} introduced this setting in the context of online learning problems and offered preliminary guarantees on Stackelberg regret, which depend on the cardinality of the hypothesis class $\cH$ and the maximum out-degree of graph~$G^\star$.
    
    \item \textbf{Unknown Manipulation Graph Setting:} 
    In real applications, the manipulation graph may not always be known. 
    We therefore
    introduce the unknown manipulation graph setting in which we relax the assumption of a known graph. Instead, we assume that the learner has prior knowledge of a certain graph class $\cG$; we consider both the realizable setting in which $G^\star\in\cG$ and the agnostic setting in which $\cG$ serves as a comparator class.
    When the true graph $G^\star$ is undisclosed, the learner cannot compute $v_t$ from $x_t$. 
    Thus, there are several options depending on when the learner gets the feedback: observing $x_t$ before implementing $h_t$ followed by $v_t$ afterward, observing $(x_t,v_t)$ after the implementation, and observing $x_t$ only (or $v_t$ only) after the implementation, arranged in order of increasing difficulty. In this work, we mainly focus on the most informative option, which involves observing $x_t$ beforehand followed by~$v_t$ afterward. However, we show that in the second easiest setting of observing $(x_t,v_t)$ afterward, no deterministic algorithm can achieve sublinear (in $|\cH|$ and $|\cG|$) mistake bound in the online setting.
\end{enumerate}

Note that the first two settings encompass all possible scenarios regarding the observation timing of $x_t$ and~$v_t$ when $\cG^\star$ is known, given that $v_t$ depends on the implemented hypothesis.
It is important to emphasize that our algorithms do not necessitate knowledge of the entire graph. Instead, we only require specific local manipulation structure information, specifically, knowledge of the out-neighborhood of the currently observed feature.


\section{Highlights of Some Main Results}
Due to the page limit, we only highlight a few of our main results and technical ideas in this section and defer other results and complete proofs to the appendices.

\subsection{PAC Learning in Fully Informative and Post-Manipulation Feedback Settings}
When the manipulation graph $G^\star$ is known, and we have an i.i.d. sample $S = ((x_1, y_1),\ldots, (x_T, y_T))$, we can obtain a hypothesis $\hat h$ by minimizing the empirical strategic loss,

\begin{align}
\restatableeq{\eqERM}{\hat h =\argmin_{h\in \cH} \sum_{t=1}^T \lstr_{G^\star}(h,(x_t,y_t))\,.    }{eq:erm}
\end{align}
\vspace{-1em}


Although in these two settings we assume the learner knows the entire graph $G^\star$, the learner only needs access to the out-neighborhood $N_{G^\star}(x_t)$ to implement ERM. If only the pair $(v_t, N_{G^\star}(v_t))$ is observed, we can obtain the knowledge of $(x_t, N(x_t))$ by implementing $h_t = \1{\cX}$ so that $v_t =x_t$.

We can then guarantee that $\hat h$ achieves low strategic population loss by VC theory. Hence, all we need to do 
for generalization is to bound the VC dimension of $\bar \cH_{G^\star}$. We establish that $\vcd(\bar \cH_{G^\star})$ can be both upper bounded (up to a logarithmic factor in $\vcd(\cH)$) and lower bounded by $\vcd(\cH) \log k$.
This implies that for any hypothesis class $\cH$ that is PAC learnable, it can also be strategically PAC learnable, provided that the maximum out-degree of $G^\star$ is finite. Achieving this only necessitates a sample size that is $\log k$ times larger and this logarithmic factor is necessary in some cases.
\begin{restatable}{theorem}{vcdinduced}\label{thm:vcd-induced}
For any hypothesis class $\cH$ and graph $G^\star$, we have $\vcd(\bar \cH_{G^\star}) \leq d\log(kd)$ where $d = \vcd(\cH)$. Moreover, for any $d,k>0$, there exists a class  $\cH$ with $\vcd(\cH) = d$ and a graph $G^\star$ with maximum out-degree $k$, such that $\vcd(\bar \cH_{G^\star})\geq d\log k$.
\end{restatable}

\begin{proof}[sketch of lower bound]
Let us start with the simple case of $d=1$. 
Consider $(k+1)\log k$ nodes $\cX =\{x_{i,j}|i = 1,\ldots,\log k, j=0,\ldots, k\}$, where there is an arc from $x_{i,0}$ 
to each of $\{x_{i,j}|j=1,\ldots, k\}$ for all $i\in [\log k]$ in $G^\star$. 
For $j=1,\ldots,k$, let $\text{bin}(j)$ denote the $\log k$-bit-binary representation of $j$. 
Let hypothesis $h_j$ be the hypothesis that labels $x_{i,j}$ by $\text{bin}(j)_i$ and all the other nodes by $0$.
Let $\cH = \{h_j|j\in [k]\}$.
Note that $\vcd(\cH) = 1$. This is because for any two points $x_{i,j}$ and $x_{i',j'}$, if $j\neq j'$, no hypothesis will label them as positive simultaneously; if $j=j'$, they can only be labeled as $(\text{bin}(j)_i,\text{bin}(j)_{i'})$ by $h_j$ and $(0,0)$ by all the other hypotheses.
However, $\bar \cH_{G^\star}$ can shatter $\{x_{1,0},x_{2,0},\ldots,x_{\log k, 0}\}$ since $h_j$'s labeling over  $\{x_{1,0},x_{2,0},\ldots,x_{\log k, 0}\}$ is $\text{bin}(j)$. Hence, we have $\vcd(\bar \cH_{G^\star}) = \log k$. We extend the example to the case of $\vcd(\cH) = d$ by making $d$ independent copies of $(\mathcal X,\cH, G^\star)$.
The detailed proof of Theorem~\ref{thm:vcd-induced} appears in Appendix~\ref{app:fullInfo}.
\end{proof}
\vspace{-1.75em}
\begin{remark}
    We have similar results for Littlestone dimension as well. More specifically, that $\ld(\bar \cH_{G^\star}) \leq \ld(\cH)\cdot\log k$ for any $\cH, G^\star$ (Theorem ~\ref{thm:fi-online}). 
    We also show that $\ld(\bar \cH_{G^\star}) \geq \ld(\cH)\cdot\log k$ using the same example we construct for VC dimension (Theorem~\ref{thm:fi-online-lb}). In contrast with the PAC setting where we derive the sample complexity bounds by a combinatorial analysis of the VC dimension, in the online setting we obtain the upper bound using an explicit algorithm. 
\end{remark}

\subsection{PAC Learning in the Unknown Manipulation Graph Setting}\label{sec:highlight-ug-pac}
Without the knowledge of $G^\star$ (either the entire graph or the local structure), we can no longer implement ERM.
For any implemented predictor $h$, if $h(x)=0$ and $N_{G^\star}(x)\cap \cX_{h,+}$ is non-empty, 
we assume that the agent will break ties randomly by manipulating to\footnote{This is not necessary. All we need is a fixed tie-breaking way that is consistent in both training and test time.} $\pi_{G^\star,h}(x)\sim \Unif(N_{G^\star}(x)\cap \cX_{h,+})$.
Since each agent can only manipulate within her neighborhood, we define a graph loss $\lgraph$ for graph $G$ as  the 0-1 loss of neighborhood prediction, 

\vspace{-1em}
\begin{equation*}
    \lgraph (G) := \Pr_{(x,y)\sim \cD}(N_{G^\star}(x)\neq N_{G}(x))\,.
\end{equation*}
\vspace{-1em}

Then we upper bound the strategic loss by the sum of the graph loss of $G$ and the strategic loss under the graph $G$, i.e.,

\vspace{-1em}
\begin{equation}
    \Lstr_{G^\star,\cD}(h) \leq \lgraph (G) + \Lstr_{G,\cD}(h), \text{ for all } G\,.\label{eq:l-decompose}
\end{equation}
\vspace{-1em}

This upper bound implies that if we can find a good approximate graph, we can then solve the problem by running ERM over the strategic loss under the approximate graph.
The main challenge of finding a good approximation of $G^\star$ is that \textbf{we cannot obtain an unbiased estimate of $\lgraph (G)$}.
This is because we cannot observe $N_{G^\star}(x_t)$. The only information we could obtain regarding $G^\star$ is a random neighbor of $x_t$, i.e., $v_t \sim \Unif(N_{G^\star}(x_t)\cap \cX_{h_t,+})$. While one may think of finding a consistent graph whose arc set contains all arcs of $(x_t,v_t)$ in the historical data, such consistency is not sufficient to guarantee a good graph. For example, considering a graph $G$ whose arc set is a superset of $G^\star$'s arc set, then $G$ is consistent but its graph loss could be high.
We tackle this challenge through a regulation term based on the empirical degree of the graph. 

In the realizable case, where there exists a perfect graph and a perfect hypothesis, i.e., there exists $G^\star$ such that $\lgraph(G^\star)=0$ and $\Lstr_{G^\star,\cD}(h^\star)=0$, 
our algorithm is described as follows.
\begin{itemize}[topsep = 2pt, itemsep=3pt, parsep=1pt, leftmargin = *]
    \item \underline{Prediction:} At each round $t$, after observing $x_t$, we implement the hypothesis $h(x)=\1{x\neq x_t}$, which labels all every point as positive except $x_t$. This results in a manipulated feature vector $v_t$ sampled uniformly from $N_{G^\star}(x_t)$. 

    \item \underline{Output:} 
    Let $(\hat G, \hat h)$ be a pair that holds

    \vspace{-2em}
    \begin{align}
    (\hat G, \hat h) &\in \argmin_{(G,h)\in (\cG,\cH)} \sum_{t=1}^T \abs{N_G(x_t)}\notag \\
    \text{s.t.}\quad &\sum_{t=1}^T \1{v_t\notin N_G(x_t)} = 0\,.\quad \label{eq:lossG} \\
    &\sum_{t=1}^T \1{\bar{h}_G(x_t)\neq y_t} = 0\,.\label{eq:lossGH}
\end{align}
\vspace{-1.5em}

Notice that Eq~\eqref{eq:lossG} guarantees $G$ is consistent with all arcs of $(x_t,v_t)$ in the historical data and Eq~\eqref{eq:lossGH} guarantees that $h$ has zero empirical strategic loss when the manipulation graph is $G$. $(\hat G, \hat h)$ is the graph-hypothesis pair that satisfies Eqs~\eqref{eq:lossG} and~\eqref{eq:lossGH} with \textbf{minimal empirical degree}.
Finally, we output $\hat h$.
\end{itemize}
Next, we derive sample complexity bound to guarantee that $\hat h$ will have a small error.

\begin{restatable}{theorem}{ugPacRel}\label{thm:ug-pac-rel}
For any hypothesis class $\cH$ with $\vcd(\cH)=d$, the underlying true graph $G^\star$ with maximum out-degree at most $k$, finite graph class $\cG$ in which all graphs have maximum out-degree at most $k$, any $(\cG,\cH)$-realizable data distribution, and any $\epsilon,\delta\in (0,1)$, with probability at least $1-\delta$ over $S\sim \cD^T$ and $v_{1:T}$ where $T = O(\frac{d \log(kd)\log(1/\epsilon) + \log|\cG|\cdot (\log(1/\epsilon) + k) + k\log(1/\delta)}{\epsilon})$, the output $\hat h$ satisfies $\Lstr_{G^\star,\cD}(\hat h)\leq \epsilon$.
\end{restatable}

\vspace{-0.5em}
\begin{proof}[sketch]
Finding a graph-hypothesis pair satisfying Eq~\eqref{eq:lossGH} would guarantee small $\Lstr_{G,\cD}(h)$ by uniform convergence. 
According to the decomposition of strategic loss in Eq~\eqref{eq:l-decompose}, it is sufficient to show the graph loss $\lgraph(\hat G)$ of $\hat G$ is low. 
Note that satisfying Eq~\eqref{eq:lossG} does not guarantee low graph loss. As aforementioned, a graph $G$ whose arc set is a superset of $G^\star$'s arc set satisfies Eq~\eqref{eq:lossG}, but its graph loss could be high. 
Hence, returning a graph-hypothesis pair 
that satisfies Eqs~\eqref{eq:lossG} and~\eqref{eq:lossGH} is not enough and it is crucial to have empirical degree as a regularize.

Now we show the graph loss $\lgraph(\hat G)$ is small.
First, the 0-1 loss of the neighborhood prediction can be decomposed as

\vspace{-1em}
\begin{equation*}
    \1{N_{G^\star}(x)\neq N_{G}(x)}
    \leq \abs{N_{G}(x)\setminus N_{G^\star}(x)} + \abs{N_{G^\star}(x)\setminus N_{G}(x)}\,.
\end{equation*}
\vspace{-1em}

For any $G$ satisfying Eq~\eqref{eq:lossG}, $G$ will have a small $\EEs{x}{\abs{N_{G^\star}(x)\setminus N_{G}(x)}}$. This is because

\vspace{-1em}
\begin{align*}
    \abs{N_{G^\star}(x)\setminus N_{G}(x)} 
    =\abs{N_{G^\star}(x)}\Pr_{v\sim \Unif(N_{G^\star}(x)) }( v\notin N_{G}(x_t))
    \leq k\Pr_{v\sim \Unif(N_{G^\star}(x)) }( v\notin N_{G}(x))\,.
\end{align*}
\vspace{-1em}

Note that $\1{v_t\notin N_{G}(x_t)}$ is an unbiased estimate of $\Pr_{x, v\sim \Unif(N_{G^\star}(x))}(v\notin N_G(x))$. 
Hence, for any graph $G$ satisfying Eq~\eqref{eq:lossG}, $\Pr_{x, v\sim \Unif(N_{G^\star}(x))}(v\notin N_G(x))$ is small by uniform convergence and thus, $\EEs{x}{\abs{N_{G^\star}(x)\setminus N_{G}(x)}}$ is small.

Then we utilize the ``minimal degree'' to connect $\EEs{x}{\abs{N_{G}(x)\setminus N_{G^\star}(x)}}$ and $\EEs{x}{\abs{N_{G^\star}(x)\setminus N_{G}(x)}}$.
For any graph with expected degree $\EEs{x}{|N_{G}(x)|} \leq \EEs{x}{|N_{G^\star}(x)|}$, subtracting $\EEs{x}{|N_{G^\star}(x)\cap N_{G}(x)|}$ from both sides implies
\vspace{-0.5em}
\begin{equation}
    \EEs{x}{|N_{G}(x)\setminus N_{G^\star}(x)|}\leq \EEs{x}{|N_{G^\star}(x)\setminus N_{G}(x)|}\,,\label{eq:neighborhood-difference}
\end{equation}

\vspace{-0.5em}

By minimizing the empirical degree, we get $\EEs{x}{|N_{G}(x)\setminus N_{G^\star}(x)|}\lesssim \EEs{x}{|N_{G^\star}(x)\setminus N_{G}(x)|}$. The formal proof can be found in Appendix~\ref{app:unknownGraph}.
\end{proof}
So, our algorithm can find a good hypothesis and needs at most $k\log|\cG|\epsilon^{-1}$ more training data. However, as we show in the next theorem, the $\log|\cG|$ factor is necessary.

\vspace{-0.5em}

\begin{restatable}{theorem}{thmUgPacRelLb}\label{thm:ug-pac-rel-lb}
There exists a hypothesis class $\cH$ with $\vcd(\cH) = 1$ and a graph class $\cG$ in which all graphs have a maximum out-degree of $1$. For any algorithm $\cA$, there exists a data distribution $\cD$ such that for any i.i.d. sample of size $T$, there exists a graph $G \in \cG$ consistent with the data history such that when $T \leq \frac{\log|\cG|}{\epsilon}$, we have $\Lstr_{G,\cD}(\hat{h}) \geq \epsilon$.
\end{restatable}

\vspace{-0.5em}

\textbf{Agnostic case } 
Similar to the realizable case, the challenge lies in the fact that  $\lgraph (G)$ is not estimable. Moreover, in the agnostic case 
there might not exist a consistent graph satisfying Eqs~\eqref{eq:lossG} and \eqref{eq:lossGH} now. Instead, we construct the following alternative loss function as a proxy:

\vspace{-0.5em}
\begin{equation*}
    \lproxy(G) := 2\mathbb{E}_{x}[P_{v\sim \Unif(N_{G^\star})(x)}(v\notin N_G(x))] + \frac{1}{k}\mathbb{E}[|N_G(x)|] - \frac{1}{k}\mathbb{E}[|N_{G^\star}(x)|]\,.
\end{equation*}


Note that in this proxy loss, $2\cdot\1{v_t\notin N_G(x_t)}+\frac{1}{k}N_G(x_t)$ is an unbiased estimate of the first two terms, 
and the third term is a constant. Hence, we can find a graph with low proxy loss by minimizing its empirical value.
We then show that this proxy loss is a good approximation of the graph loss when $k$ is small.

\vspace{-0.5em}

\begin{restatable}{lemma}{lmmproxyLoss}\label{lmm:proxy-loss}
    Suppose that $G^\star$ and all graphs $G\in \cG$ have maximum out-degree at most $k$. Then, we have
    \vspace{-0.5em}
    \begin{equation*}
        \frac{1}{k} \lgraph(G) \leq \lproxy(G) \leq 3 \lgraph(G)\,.
    \end{equation*}
    \vspace{-1em}
\end{restatable}
\vspace{-1em}

By minimizing the proxy loss $\lproxy(\cdot)$, we can obtain an approximately optimal graph up to a multiplicative factor of $k$ due to the gap between the proxy loss and the graph loss. See details and proofs in Appendix~\ref{app:unknownGraph}. \footnote{Different tie-breaking rules can impact the estimation factor of $\lproxy(\cdot)$ and consequently the sample complexity.}

\paragraph{Application of the Graph Learning Algorithms in Multi-Label Learning }
Our graph learning algorithms have the potential to be of independent interest in various other learning problems, such as multi-label learning. To illustrate, let us consider scenarios like the recommender system, where we aim to recommend movies to users. In such cases, each user typically has multiple favorite movies, and our objective is to learn this set of favorite movies.

This multi-label learning problem can be effectively modeled as a bipartite graph denoted as $G^\star = (\cX, \cY, \cE^\star)$, where $\cX$ represents the input space (in the context of the recommender system, it represents users), $\cY$ is the label space (in this case, it represents movies), and $\cE^\star$ is a set of arcs. In this graph, the presence of an arc $(x, y) \in \cE^\star$ implies that label $y$ is associated with input $x$ (e.g., user $x$ liking movie $y$). Our primary goal here is to learn this graph, i.e., the arcs $\cE^\star$. More formally, given a marginal data distribution $\cD_\cX$, our goal is to find a graph $\hat G$ with minimum neighborhood prediction loss $\lgraph (G)= \Pr_{x\sim \cD_\cX}(N_{G}(x)\neq N_{G^\star}(x))$. Suppose we are given a graph class $\cG$. Then, our goal is to find the graph $\argmin_{G\in \cG}\lgraph (G)$.

However, in real-world scenarios, the recommender system cannot sample a user along with her favorite movie set.
Instead, at each time $t$, the recommender recommends a set of movies $h_t$ to the user and the user randomly clicks one of the movies in $h_t$ that she likes (i.e., $v_t\in N_{G^\star}(x_t)\cap h_t$). Here we abuse the notation a bit by letting $h_t$ represent the set of positive points labeled by this hypothesis.
This setting perfectly fits into our unknown graph setting and our algorithms can find a good graph.


\subsection{Online Learning}\label{sec:highlight-online}
In this section, we mainly present our algorithmic result in the post-manipulation feedback setting. 
The post-manipulation setting is introduced by \cite{ahmadi2023fundamental}, where they exclusively consider the finite hypothesis class $\cH$ and derive Stackelberg regret bounds based on the cardinality $|\cH|$. In contrast, our work extends to dealing with infinite hypothesis classes and offers algorithms with guarantees dependent on the Littlestone dimension. 

In the post-manipulation feedback setting, the learner observes either $x_t$ or $v_t$ after implementing $h_t$. It might seem plausible that observing $x_t$ provides more information than observing $v_t$ since we can compute $v_t$ given $x_t$. However, we show that there is no gap between these two cases. 
We provide an algorithm based on the ``harder'' case of observing $v_t$ and a lower bound based on the ``easier'' case of observing $x_t$ and show that the bounds are tight.

Our algorithm is based on a reduction from strategic online learning to standard online learning. More specifically, given any standard online learning algorithm $\cA$, we construct a weighted expert set $E$ which is initialized to be $\{\cA\}$  with weight $w_\cA =1$. At each round, we predict $x$ as $1$ iff the weight of experts predicting $x$ as $1$ is higher than $1/2(k+1)$ of the total weight.
Once a mistake is made at the observed node $v_t$, we differentiate between the two types of mistakes.
\begin{itemize}[topsep = 2pt, itemsep=3pt, parsep=1pt, leftmargin = *]
    \item \underline{False positive:} Namely, the  labeling of $x_t$ induced by $h_t$ is $1$ but the true label is $y_t = 0$. 
    Since the true label is $0$, the entire neighborhood of $x_t$ should be labeled as $0$  by the target hypothesis, including the observed feature $v_t$. Hence, we proceed by updating all experts incorrectly predicting $v_t$ as $1$ with the example $(v_t,0)$ and halve their weights. 
    \item \underline{False negative:} Namely, the labeling of $x_t$ induced by $h_t$ is $0$, but the true label is $y_t = 1$.
    Thus, $h^\star$ must label some neighbor of $x_t$, say $x^\star$, by $1$. For any expert $A$ labeling the entire neighborhood by $0$, they make a mistake. But we do not know which neighbor should be labeled as $1$. So we split $A$ into $|N_{G^\star}[x_t]|$ number of experts, each of which is fed by one neighbor $x\in N_{G^\star}[x_t]$ labeled by $1$. Then at least one of the new experts is fed with the correct neighbor $(x^\star,1)$. We also split the weight equally and halve it.
\end{itemize}
The pseudo-code is included in Algorithm~\ref{alg:reduction2online-pmf}.
\begin{algorithm}[t]\caption{Red2Online-PMF: Reduction to online learning 
 for post-manipulation feedback}\label{alg:reduction2online-pmf}
    \begin{algorithmic}[1]
        \STATE \textbf{Input: } a standard online learning algorithm $\cA$, maximum out-degree upper bound $k$
        \STATE \textbf{Initialization: } expert set $E = \{\cA\}$ and weight $w_{\cA}=1$
        \FOR{$t=1,2,\ldots$}
        \STATE \underline{Prediction}: at each point $x$, $h_t(x) =1$ if $\sum_{A \in E: A(x) = 1}w_{A} \geq \frac{\sum_{A\in E} w_{\cA_H}}{2(k+1)}$\label{alg-line:k-fraction-weight}
        \STATE \underline{Update}: \textit{//when we make a mistake at the observed node $v_t$}
        \IF{$y_t = 0$}
        \STATE for all $A\in E$ satisfying $A(v_t) = 1$, feed $A$ with $(v_t,0)$ and update $w_{A} \leftarrow \frac{1}{2} w_{A}$
        \ELSIF{$y_t = 1$}
        \STATE for all $A \in E$ satisfying $\forall x\in N_{G^\star}[v_t], A(x) =0$
        \STATE for all $x\in N_{G^\star}[v_t]$, by feeding $(x,1)$ to $A$, we can obtain a new expert $A(x,1)$
        \STATE remove $A$ from $E$ and add $\{A(x,1)|x\in N_{G^\star}[x_t]\}$ to $E$ with weights $w_{A(x,1)} = \frac{w_{A}}{2\abs{N_{G^\star}[x_t]}}$
        \ENDIF
        \ENDFOR
    \end{algorithmic}
\end{algorithm}
Next, we derive a mistake bound for Algorithm~\ref{alg:reduction2online-pmf} that depends on the mistake bound of the standard online learning algorithm, and by plugging in the Standard Optimal Algorithm (SOA), we derive a mistake bound of $O(k \log k \cdot M)$. 

\vspace{-0.5em}
\begin{restatable}{theorem}{fuOnline}\label{thm:fu-online}
For any hypothesis class $\cH$, graph $G^\star$ with maximum out-degree $k$, and a standard online learning algorithm with mistake bound $M$ for $\cH$, for any realizable sequence, we can achieve mistake bound of $O(k \log k \cdot M)$ by Algorithm~\ref{alg:reduction2online-pmf}.
By letting the SOA algorithm by~\cite{littlestone1988learning} as the input standard online algorithm $\cA$,  Red2Online-PMF(SOA)
makes at most $O(k \log k\cdot \ld(\cH))$ mistakes.
\end{restatable}
Compared to the fully informative setting (in which the dependency on $k$ of the optimal mistake bound is $\Theta(\log k)$), the dependency of the number of mistakes made by our Algorithm~\ref{alg:reduction2online-pmf} on $k$ is $O(k\log k)$. However, as we show in the next theorem, 
no deterministic algorithm can make fewer than $\Omega(\ld(\cH)\cdot k)$ mistakes. Hence, our algorithm is nearly optimal (among all deterministic algorithms) up to a logarithmic factor in $k$.

\vspace{-0.7em}

\begin{restatable}{theorem}{fuOnlineLb}\label{thm:fu-online-lb}
For any $k,d>0$, there exists a graph $G^\star$ with maximum out-degree $k$ and a hypothesis class $\cH$ with $\ld(\cH) = d$ such that for any deterministic algorithm, there exists a realizable sequence for which the algorithm will make at least $d(k-1)$ mistakes.
\end{restatable}

\vspace{-1.5em}
\begin{remark}
    \cite{ahmadi2023fundamental} leave the gap between the upper bound of $O(k \log(\abs{\cH}))$ and the lower bound of $\Omega(k)$ as an open question. Theorems~\ref{thm:fu-online} and \ref{thm:fu-online-lb} closes this gap up to a logarithmic factor in $k$.
\end{remark}


\textbf{Extension to Unknown Graph Setting }
In the unknown graph setting, we do not observe $N_{G^\star}(x_t)$ or $N_{G^\star}(v_t)$ anymore.
We then design an algorithm by running an instance of Algorithm~\ref{alg:reduction2online-pmf} over the majority vote of the graphs. 
More specifically, after observing $x_t$,  let $\tilde N(x_t)$ denote the set of nodes $x$ satisfying that $(x_t,x)$ is an arc in at least half of the consistent graphs. 
\begin{itemize}[topsep = 2pt, itemsep=3pt, parsep=1pt, leftmargin = *]
    \item \underline{Reduce inconsistent graphs}: For any $x\notin \tilde N(x_t)$, which means $(x_t,x)$ is an arc in at most half of the current consistent graphs, we predict $h_t(x) =1$. If we make a mistake at such a $v_t = x$, it implies that $(x_t,x)$ is an arc in the true graph, so we can eliminate at least half of the graphs.
    \item \underline{Approximate $N_{G^\star}(x_t)$ by $\tilde N(x_t)$}: For any $x\in \tilde N(x_t)$, we predict $h_t(x)$ by following Algorithm~\ref{alg:reduction2online-pmf}. 
    We update Algorithm~\ref{alg:reduction2online-pmf} in the same way when we make a false positive mistake at $v_t$ as it does not require the neighborhood information. However, when we make a false negative mistake at $v_t$, it implies that $v_t =x_t$ and the entire neighborhood $N_{G^\star}(x_t)$ is labeled as $0$. We then utilize $\tilde N(x_t)$ as the neighborhood feedback to update  Algorithm~\ref{alg:reduction2online-pmf}.
    Note that this majority vote neighborhood $\tilde N(x_t)$ must be a superset of $N_{G^\star}(v_t)=N_{G^\star}(x_t)$ since the entire $N_{G^\star}(v_t)$ is labeled as $0$ by $h_t$ and every $x\notin \tilde N(x_t)$ is labeled as $1$. Moreover, the size of $\tilde N(x_t)$ will not be greater than $2k$ since all graphs have maximum degree at most $k$.
\end{itemize}
We defer the formal description of this algorithm to Algorithm~\ref{alg:ug-online} that appears in Appendix~\ref{app:unknownGraph}. In the following, we derive an upper bound over the number of mistakes for this algorithm.
\begin{restatable}{theorem}{ugOnline}\label{thm:ug-online}
    For any hypothesis class $\cH$, the underlying true graph $G^\star$ with maximum out-degree at most $k$, finite graph class $\cG$ in which all graphs have maximum out-degree at most $k$, for any realizable sequence, Algorithm~\ref{alg:ug-online} will make at most $O(k\log(k)\cdot \ld(\cH)+ \log |\cG|)$ mistakes.   
\end{restatable}
So when we have no access to the neighborhood information, we will suffer $\log |\cG|$ more mistakes. However, we show that this $\log |\cG|$ term is necessary for all deterministic algorithms.

\vspace{-0.5em}
\begin{restatable}{theorem}{ugOnlineLb}
    \label{thm:ug-online-lb}
    For any $n\in \NN$, there exists a hypothesis class $\cH$ with $\ld(\cH) =1$, a graph class $\cG$ satisfying that all graphs in $\cG$ have maximum out-degree at most $1$ and $|\cG| = n$ such that for any deterministic algorithm, there exists a graph $G^\star\in \cG$ and a realizable sequence for which the algorithm will make at least $\frac{\log n}{\loglog n}$ mistakes.  
\end{restatable}
\vspace{-0.5em}

Finally, as mentioned in Section~\ref{sec:model-uncertainty}, there are several options for the feedback, including observing $x_t$ beforehand followed by $v_t$ afterward, observing $(x_t,v_t)$ afterward, and observing either $x_t$ or $v_t$ afterward, arranged in order of increasing difficulty. We mainly focus on the simplest one, that is, observing $x_t$ beforehand followed by $v_t$ afterward. We show that even in the second simplest case of observing $(x_t,v_t)$ afterward, any deterministic algorithm suffers mistake bound linear in $|\cG|$ and $|\cH|$.

\vspace{-0.5em}
\begin{restatable}{proposition}{uglessinfo}\label{prop:online-v-lb}
    When $(x_t,v_t)$ is observed afterward, for any $n\in \NN$, there exists a class $\cG$ of graphs of degree $2$ and a hypothesis class $\cH$ with $|\cG| = |\cH| = n$ such that for any deterministic algorithm, there exists a graph $G^\star\in \cG$ and a realizable sequence for which the algorithm will make at least $n-1$ mistakes.
\end{restatable}
\vspace{-0.5em}
\section{Discussion}
In this work, we have investigated the learnability gaps of strategic classification in both PAC and online settings. We demonstrate that learnability implies strategic learnability when the manipulation graph has a finite maximum out-degree, $k<\infty$. Additionally, strategic manipulation does indeed render learning more challenging.
In scenarios where we consider both the true graph information and pre-manipulation feedback, manipulation only results in an increase in both sample complexity and regret by a $\log k$ multiplicative factor. However, in cases where we only have post-manipulation feedback, the dependence on $k$ in the sample complexity remains $\log k$, but increases to $k$ in the regret.
When the true graph is unknown and only a graph class $\cG$ is available, there is an additional increase of a $\log|\cG|$ additive factor in both sample complexity and regret. Our algorithms for learning an unknown graph are of independent interest to the multi-label learning problem.

Several questions remain open. The first pertains to agnostic online learning. We have explored extending some realizable online learning results to the agnostic case in Appendix~\ref{app:online-agn} through a standard reduction technique. However, it is unclear how to address this issue, especially in the case where there is no perfect graph in the class.
Additionally, several other technical questions remain open, such as improving the lower bounds presented in the work and eliminating the multiplicative factor in Theorem~\ref{thm:ug-pac-agn}.

Another interesting open question is gaps in computational complexity. In the fully informative PAC setting, our algorithm minimizes the empirical strategic loss. 
Computing $\bar h_{G^\star}(x)$ for each $x$ requires $\cO(k)$ time, by checking if there is any neighbor of $x$ that is labeled as positive by $h$. Hence, minimizing the empirical strategic loss would take $O(k)$ times more computational complexity than minimizing the standard $0-1$ loss. In the unknown graph PAC learning setting, we optimize over all graph-hypothesis pairs, which makes the computational complexity scale linearly with the cardinality of the graph class.

\chapter{Incentives in Single-Round Federated Learning}\label{chap:incentives-single}
\section{Introduction}

In recent years, federated learning has been embraced as an approach for enabling large numbers of learning agents to collaboratively accomplish their goals using collectively fewer resources, such as smaller data sets.
Indeed, collaborative protocols are starting to be used across networks of hospitals \citep{wen2019federated,NVIDIA2} and devices~\citep{mcmahan_ramage_2017} and are behind important breakthroughs such as understanding the
biological mechanisms underlying schizophrenia in a large scale  collaboration of more than 100 agencies~\citep{bergen2012genome}.

This promise of creating large scale impact from mass participation has led to federated learning 
receiving substantial interest in the machine learning research community,
and has resulted in faster and more communication-efficient collaborative systems.
But, what will ultimately decide the success and impact of collaborative federated learning is the ability to recruit and retain large numbers of learning agents --- a feat that requires collaborative algorithms to
\begin{quote}
\emph{
help agents accomplish their learning objectives while ``equitably'' spreading the data contribution responsibilities among agents who want a lower sample collection burden.
}
\end{quote}
This is to avoid the following inequitable circumstances that
may otherwise arise in collaborative learning.
First, when part of an agent's data is exclusively used to accomplish another agent's learning goals; for example, if an agent's learning task can be accomplished even when she (unilaterally) lowers her data contribution.
Second, when an agent envies another agent; for example, if an agent's learning goal can be accomplished even when she swaps her contribution burden with another agent who has a lower burden.

In this paper, we introduce the first comprehensive game theoretic framework for collaborative federated learning in the presence of agents who are interested in accomplishing their learning objectives while keeping their individual sample collection burden low. 
Our framework introduces two notions of equilibria
that avoid the aforementioned inequities. 
First, analogous to the concept of Nash equilibrium~\citep{nash1951non}, our \emph{stable equilibrium} requires that no agent could unilaterally reduce her data contribution responsibility and still accomplish her learning objective. 
Second, inspired by the concept of envy-free allocations~\citep{foley1967resource,VARIAN197463},  our \emph{envy-free equilibrium} requires that no agent could swap her data contribution with an agent with lower contribution level and still accomplish her learning objective. 
In addition to capturing what is deemed as an ``equitable'' collaboration to agents, using stable and envy-free equilibria is essential for keeping learning participants fully engaged in ongoing collaborations. 

Our framework is especially useful for analyzing how the sample complexity of federated learning may be affected by the agents' desire to keep their individual sample complexities low.
To demonstrate this, we work with three classes as running examples of agent learning objectives: random discovery (aka linear) utilities, random coverage utilities, and general PAC learning utilities. Our results answer the following qualitative and quantitative questions:

\vspace*{-6pt}
\paragraph{Existence of Equilibria.}
In Section~\ref{sec:exist}, we show that the existence of a stable equilibrium depends on whether agents' learning objectives are ``well-behaved''.
In particular, we see that in the PAC learning setting, there may not exist a stable equilibrium, but under mild assumptions, a stable equilibrium exists in the random discovery (aka linear) and random coverage settings.
On the other hand, an envy-free equilibrium with equal agent contribution trivially exists. 

\vspace*{-6pt}
\paragraph{Sample Complexity of Equilibria.} In Section~\ref{sec:sample-complexity}, we show that even for well-behaved learning objectives, such as random discovery and random coverage examples, there is a large gap between the socially optimal sample complexity and the optimal sample complexity achieved by any equilibrium.
In particular, we show that there is a factor $\Omega(\sqrt{k})$ gap between the socially optimal sample complexity and that of optimal stable or envy-free equilibria for $k$ agents.

\vspace*{-6pt}
\paragraph{Algorithmic and Structural Properties.}
The main result of Section~\ref{sec:algs} shows that in the random discovery (aka linear) setting, in every
optimal stable equilibrium there is a core-set of agents for whom \emph{the equilibrium happens to also be socially optimal}, and agents who do not belong to this set make $0$ contribution in the equilibrium.
This result allows us to characterize classes of problems where the optimal stable equilibria are also socially optimal.
We further show that in some cases, linear or convex programs can be used to compute socially optimal or optimal stable equilibria.

\vspace*{-6pt}
\paragraph{Empirical Analysis.}
We show that some commonly used federated algorithms produce solutions that are very far from being an equilibrium. 
We show that the Federated-Averaging (FedAvg) algorithm of \citet{mcmahan2017communication}  lead to solutions where a large number of agents would rather reduce their contribution to as little as 25\% to 1\%.
We also work with the Multiplicative Weight Update style algorithm (MW-FED) of \citet{blum2017collaborative} and show that this algorithm produces allocations that are closer to being an equilibrium,
but more work is needed for designing algorithms  that further close this gap.

\subsection{Related Work.}

Federated learning and the model aggregation algorithm FedAvg were proposed by~\citet{mcmahan2017communication}. The collaborative learning framework of~\citet{blum2017collaborative} studied heterogeneous learning objectives in federated learning
and quantified how sample complexity improves with more collaboration.
However, except for a few recent works discussed below, agents' incentives  have not been addressed in these frameworks.

\citet{lyu2020collaborative,yu2020fairness,zhang2020hierarchically} proposed several fairness metrics for federated learning that reward high-contributing agents with higher payoffs, however, they do not consider strategicness of agents and the need for equilibrium.
\citet{li2019fair} empirically studied a different fairness notion of uniform accuracy across devices without discussing data contribution, 
while our work allows for different accuracy levels so long as every learners objective is accomplished and focuses data contribution. 
Recently, \citet{donahue2020model} studied
individual rationality in federated learning when global models may be worse than an agent's local model and used concepts from hedonic game theory to discuss coalition formation. 
Other works have discussed issues of free-riders and reputation~\citep{lin2019free,kang2019incentive} as well as markets and credit sharing in machine learning ~\citep{ghorbani2019data,agarwal2019marketplace,balkanski2017statistical,jia2019towards}.

\section{Problem Formulation}\label{sec:model-examples}

Let us start this section with a motivating example before introducing our general model in Section~\ref{sec:gen-model}.

Consider the collaborative learning problem with $k$ agents with distributions $\cD_1,\ldots,\cD_k$. For each agent $i\in[k]$, her goal is to satisfy the constraint of low expected error, i.e.,
\begin{align}
    \EEs{\{S_j\sim \cD_j^{m_j}\}_{j\in[k]}}{\err_{\cD_i}(h_{S})}\leq \epsilon\,,
    \label{eq:ind-opt}
\end{align}
for some $\epsilon>0$, where each agent $j$ takes $m_j\geq 0$ random samples $S_j\sim \cD_j^{m_j}$ and $h_S$ is a prediction rule based on $S=\cup_{j\in [k]} S_j$.
For example, $h_S$ can be the prediction output by performing ERM or gradient descent over the set $S$. Then the sample complexity of federated learning is the optimal allocation $(m_1, \dots, m_k)$ that minimizes the total number of samples conditioned on satisfying every agent's accuracy constraint,  that is,
\begin{equation}
    \begin{array}{l}
        \min \sum\limits_{i=1}^k m_i\\
        \st\,  \EEs{\{S_j\sim \cD_j^{m_j}\}_{j\in[k]}}{\err_{\cD_i}(h_{S})}\leq \epsilon,\forall i\in[k]\,.
    \end{array}
    \label{eq:social-opt}
\end{equation}
It is not surprising that optimizing Equation~\eqref{eq:social-opt} requires collectively fewer samples than the total number of samples agents need to individually solve Equation~\eqref{eq:ind-opt}\footnote{\citet{blum2017collaborative} upper and lower bound this improvement.}, but the optimal solution to Equation~\eqref{eq:social-opt} may unfairly require one or more agents  to contribute larger sample sets than could reasonably be expected from them. 
Our notion of a \emph{stable equilibrium} requires that no agent $j$, conditioned on keeping her constraint satisfied, can unilaterally reduce $m_j$. On the other hand, our \emph{envy-free equilibrium} requires that no agent $j$,
conditioned on keeping her constraint satisfied, can swap $m_j$ with another agent's contribution $m_i<m_j$.
Taking stability and envy-freeness as constraints, we ask whether such notions of equilibria always exist and whether stable and envy-free sample complexities are significantly worse than the optimal solution to Equation~\eqref{eq:social-opt}.

\subsection{The General Framework} \label{sec:gen-model}

In this paper, we study this problem in a more general setting where there are $k$ agents and all agents collaboratively select a strategy $\btheta= (\theta_1,\ldots,\theta_k)$ from a strategy space $\Theta\subseteq \R_+^k$. Each agent $i$ selects a number $\theta_i$ as her contribution level, e.g., the number of samples.
We define $u_i:\Theta\mapsto \R$ as the utility function for each agent $i$ and her goal is to achieve 
\[
u_i(\btheta)\geq \mu_i
\]
for some $\mu_i$. This utility function is a generalization of the expected accuracy in our motivating example and $\mu_i = 1-\epsilon$ is the minimum accuracy required by the agent.

For any $\btheta\in \R^k$, $x\in \R$, let $(x,\btheta_{-i})\in\R^k$ denote the vector with the $i$-th entry being $x$ and the $j$-th entry being $\theta_j$ for $j\neq i$. 
Without loss of generality, we assume that every agent can satisfy her constraint individually, i.e., $\forall i\in[k], \exists \vartheta_i\in \R_+^k$ such that $u_i(\vartheta_i,\bZero_{-i})\geq \mu_i$ and $u_i$ is non-decreasing with $\theta_j$ for any $j$.

We say that $\btheta$ is \emph{feasible} if  $u_i(\btheta)\geq \mu_i$ for all $i\in[k]$.
We define the socially optimal solution analogously to Equation~\ref{eq:social-opt} as the optimal feasible solution that does not consider agents' incentives.

\begin{definition}[Optimal solution (OPT)]
$\btheta^\optlocal$ is a \emph{socially optimal solution} in $\Theta$ if it is the optimal solution to the following program
\begin{equation} \label{eq:gen-opt}
    \begin{array}{l}
        \min_{\btheta\in \Theta} \bOne^\top \btheta\\
        \st u_i(\btheta)\geq \mu_i,\forall i\in[k]\,.
    \end{array}
\end{equation}

\end{definition}

A stable equilibrium is a feasible solution where no player has incentive to unilaterally decrease her strategy.
\begin{definition}[Stable equilibrium (EQ)]
     A feasible solution $\btheta^\eq$ is a \emph{stable equilibrium} over $\Theta$ if for any $i\in [k]$, there is no $(\theta_i',\btheta^\eq_{-i})\in \Theta$ such that $\theta_i'<\theta^\eq_i$ and $u_i(\theta_i',\btheta^\eq_{-i})\geq \mu_i$.
\end{definition}

    An envy-free equilibrium is a feasible solution where no agent has an incentive to swap their sampling load with another agent.
For any $\btheta$, let $\btheta^{(i,j)}$ denote the $\btheta$ when the $i$-th and the $j$-th entries are swapped,
i.e., $\theta^{(i,j)}_i=\theta_j$, $\theta^{(i,j)}_j=\theta_i$ and $\theta^{(i,j)}_l=\theta_l$ for $l\neq i,j$. 
\begin{definition}[Envy-free equilibrium (EF)]
A feasible solution $\btheta^\ef$ is \emph{envy-free} if for any $i\in [k]$, there is no $\btheta^{\ef(i,j)}\in \Theta$ such that $\theta^\ef_j<\theta^\ef_i$ and $u_i(\btheta^{\ef(i,j)})\geq \mu_i$.

\end{definition}
We call an equilibrium $\btheta$ optimal if it is an equilibrium with minimal resources, i.e., minimizes $\bOne^\top\btheta$.

We use the game theoretic quantities known as the \emph{Price of Stability}~\citep{anshelevich2008price} and the \emph{Price of Fairness}~\citep{caragiannis2012efficiency} to quantify the impact of equilibria on the efficiency of collaboration.

\begin{definition}[Price of Stability]
Price of Stability (PoS) is defined as the ratio of the value of the optimal stable equilibrium to that of the socially optimal solution. That is, letting  $\Theta^\eq\subseteq \Theta$ be the set of all stable equilibria, $\mathrm{PoS} = \min_{\btheta\in \Theta^\eq} \bOne^\top \btheta / \bOne^\top \btheta^\optlocal$.
\end{definition}

\begin{definition}[Price of Fairness]
Price of Fairness (PoF) is defined as the ratio of the value of the optimal envy-free equilibrium to that of the socially optimal solution. That is, letting  $\Theta^\ef\subseteq \Theta$ be the set of all envy-free equilibria, $\mathrm{PoF} = \min_{\btheta\in \Theta^\ef} \bOne^\top \btheta / \bOne^\top \btheta^\optlocal$.
\end{definition}

\subsection{Canonical Examples and Settings}\label{sec:canonicalexamples}
We use the following three canonical settings as running examples throughout the paper.
\paragraph{Random Discovery aka Linear Utilities.} 
We start with a setting where any agent's utility is a linear combination of the efforts other agents put into solving the problem. As a general setting, we let $\bu(\btheta)= W\btheta$ for matrix $W\in [0,1]^{k\times k}$, where $W_{ij}$ denotes how the effort of agent $j$ affects the utility of agent $i$. We commonly assume that $W$ is a symmetric PSD matrix with an all one diagonal.

As an example, consider a setting where each agent $i$ has a distribution $\bq_i$ over the instance space $\cX$ with $\abs{\cX}=n$, and where the agent $i$ receives a reward proportional to the density of $q_{ix}$ every time an instance $x$ is realized (or discovered) by any agent's sampling effort. Formally, the utility of agent $i$ in strategy $\btheta$ is her expected reward:
\[ u_i(\btheta) = \bq_i Q^\top \btheta,
\]
where $Q=[q_{ix}]\in \R_+^{k\times n}$ denote the matrix with the $(i,x)$-th entry being $q_{ix}$, we have that $\bu(\btheta) = QQ^\top \btheta$ is a linear function.  Note that in this case, $W=  QQ^\top$ is indeed a symmetric PSD matrix.

\paragraph{Random Coverage.} 
While in our previous example an agent draws utility \emph{everytime} an instance $x$ is discovered, in many classification settings, the utility of an agent is determined by whether $x$ has been observed at all (and not the number of its observations).
This gives rise to the non-linear utilities we define below.

Consider a simple binary classification setting where the label of each point is uniformly labeled positive or negative independently of all others.
More specifically, assume that the domain $\cX$ is labeled according to a target function $f^*$ that is chosen uniformly at random from  $\{\pm 1\}^{\cX}$. Note that given any set of observed points $S = \{x_1, \dots, x_m\} \subseteq \cX$ and their corresponding revealed labels $f^*(x)$ for $x\in S$. The optimal classifier $h_S$ classifies each $x\in S$ correctly as $f^*(x)$ and misclassifies each $x\notin S$  with probability $1/2$.
Let $u_i(\btheta)$ be the expected accuracy of the optimal classifier where agent $i$ took an integral value $\theta_i$ number of samples, i.e.,
\[
u_i(\btheta) = 1 - \frac 12 \sum_{x\in \cX} q_{ix} \prod_{j=1}^k \left(1-q_{jx} \right)^{\theta_j}\,.
\]
Throughout the paper, we consider the general random coverage setting introduced here and its simpler variants where all agents' distributions are uniform over equally-sized sets.

As opposed to the linear utilities, non-integral values of $\theta_i$ (as mean of a distribution over integers)  are not as easily interpretable. Indeed, the same $\theta_i$ may refer to distributions with different expected utilities.
Here we consider one natural interpretation of a real-valued $\theta_i$: randomized rounding over $\lfloor \theta_i\rfloor$ and $\lceil \theta_i \rceil$ with mean of $\theta_i$. See Appendix~\ref{app:integral} for more information.

\paragraph{General PAC Learning.} Now we consider a general learning setting, where the labels of points are not necessarily independent. In this case, the optimal classifier can improve its accuracy on unobserved points based on those points' dependence on observed points. For example, consider a scenario where an input space $\cX$ where $\abs{\cX} =2$ and a hypothesis class that always labels points in $\cX$ either both positive or negative. Then if only one point is observed, the classifier will classify the unobserved point the same as the label of the observed one.

Generally, given input space $\cX$, hypothesis class $\cH$ and agent $i$'s distribution $\cD_i$ over $\cX$, we let utility function $u_i(\btheta)$ be the expected accuracy of any consistent function $h_S\in \cH$ given training data set $S=\cup_{j\in[k]}S_j$ when agent $i$ takes an integral value $\theta_i$ number of samples,
\[
u_i(\btheta) = 1 -{\EEs{\{S_j\sim \cD_j^{\theta_j}\}_{j\in[k]}}{\err_{\cD_i}(h_{S})}}\,.
\]
Similar to the random coverage settings, we interpret real values $\theta_i$ as the appropriate distribution over  $\lfloor \theta_i\rfloor$ and $\lceil \theta_i \rceil$ whose mean is $\theta_i$.

\section{Existence of Equilibria}\label{sec:exist}
In this section, we discuss the existence of stable and envy-free equilibria in collaborative federated learning.
Clearly, any solution with equal allocation among all agents is an envy-free allocation. That is, any feasible allocation $\btheta$ can be converted to an envy-free allocation $\btheta^\ef$ by letting $\forall i \in [k], \theta^\ef_i = \max_j \theta_j$.

\begin{theorem}\label{thm:efexist}
An envy-free solution always exists in a feasible collaborative learning problem.%
\end{theorem}
In the aforementioned envy-free solution $\btheta^\ef$, however, all agents (except for those with the maximum allocation) could unilaterally reduce their allocations while meeting their constraints, so $\btheta^\ef$ is not an equilibrium.
Indeed, in the remainder of this section we show that existence of an equilibrium in collaborative learning depends on the precise setting of the problem. In particular, we show that an equilibrium solution exists when unilateral deviations in an agent's contribution has a bounded impact on the utility of any agent.
On the other hand, an equilibrium solution may not exist if infinitesimally small changes to an agent's contribution has an outsized effect on other agents' utilities (or if an agent's strategy space is not even continuous).

We will formalize this in the next definition. Broadly, this definition states that an agent's utility increases at a positive (and bounded away from zero) rate when the agent unilaterally increases her contribution. Moreover, an agent's utility does not increase at an infinite rate. In other words, it is bounded above by a constant when other agents unilaterally increase their contributions.

\begin{definition}[Well-behaved Utility Functions]\label{assp:eqgeneral}
We say that a set of utility functions $\{u_i: \Theta \rightarrow \R \mid i\in[k]\}$ is well-behaved over $\bigtimes_{i=1}^k[0,C_i] \subseteq \Theta$ for some $C_i$s, if
and for each agent $i\in[k]$ there are constants $c^i_1 \geq 0$ and $c^i_2>0$ such that for any $\btheta\in \bigtimes_{i=1}^k[0,C_i]$,
\begin{enumerate}
    \item $\partial u_i(\btheta)/\partial \theta_i\geq c^i_2$; and
    \item for all $j\in[k]$ and $j\neq i$, $0\leq \partial u_i(\btheta)/\partial \theta_j\leq c^i_1$.
\end{enumerate} 
\end{definition}

We emphasize that the utility functions that correspond to many natural learning settings and domains, such as in the linear case and random coverage, are well-behaved.
That being said, it is also not hard to construct natural learning settings where the utility functions are not well-behaved, e.g., when an agent is restricted to taking an integral number of samples and therefore its utility is not continuous.
In the remainder of this section, we prove that, when agent utilities are well-behaved, an equilibrium exists.

\begin{theorem}\label{thm:eqexist}
For any collaborative learning problem with utility functions $u_i$s and $\mu_i$s, let $\vartheta_i$ represent the individually satisfying strategy such that $u_i(\vartheta_i,\bZero_{-i})\geq \mu_i$. If $u_i$s are well-behaved over $\bigtimes_{i=1}^k[0,\vartheta_i]$, then there exists an equilibrium.
\end{theorem}
We complement this positive result by constructing a natural learning setting that corresponds to ill-behaved utility functions and show that this problem has no equilibrium.
\begin{theorem}\label{thm:eqintexist}
There is a feasible collaborative learning problem in the general PAC learning setting that does not have an equilibrium.

\end{theorem}

\subsection{Are Canonical Examples Well-behaved?}
Recalling the three canonical examples introduced in Section~\ref{sec:canonicalexamples}, here we discuss whether they are well-behaved or not.
It is not hard to see that linear utilities are well-behaved as $u_i$ increases at a constant rate $W_{ij}$ when agent $j$ increases her strategy unilaterally, $W_{ii} = 1$, and $W_{ij}\leq 1$.

{In the random coverage case, the utilities are well-behaved over $\bigtimes_{i=1}^k[0,\vartheta_i]$ as long as 
$ u_i(\vartheta_i + 1, \bvartheta_{-i})-u_i(\bvartheta)$ is bounded away from $0$.
For example, this is the case when $\bmu \in [\frac 12, C]^k$ for $C<1$ that is bounded away from $1$.

At a high level, the smallest impact that an additional sample by agent $i$ has on $u_i$ is when $\btheta \rightarrow \bvartheta$. This impact is at least $ u_i(\vartheta_i + 1, \bvartheta_{-i})-u_i(\bvartheta) > 0$.
On the other hand, 
$\partial u_i(\btheta)/\partial \theta_j$ is bounded above, because the marginal impact of any one sample on $u_i$ is largest when no agent has yet taken a sample. Therefore, this impact is at least $u_i(1, \bZero_{-j}) - u_i(\bZero) = 1/2\sum_{x\in \cX} q_{ix} q_{jx} \leq 1/2$. 
This shows that under mild assumption the random coverage utilities are well-behaved.
}

{We note that the range of $\bigtimes_{i=1}^k[0,C_i]$ and the continuity of $\Theta$ plays an important role in determining the behavior. For example,  none of these utility functions are well-behaved over the set of integers, since $\partial u_i(\btheta)/\partial \theta_j$ is undefined.} More detail can be found in Appendix~\ref{app:wellbehave}.

\subsection{Proof of Theorem~\ref{thm:eqexist}}

In this section, we prove Theorem~\ref{thm:eqexist} and show that an equilibrium exists when utility functions are well-behaved.
Our main technical tool is to show that the best-response dynamic has a fixed point.
We define a \emph{best-response} function ${\bFlocal:} \bigtimes_{i=1}^k[0,\vartheta_i] \mapsto \bigtimes_{i=1}^k[0,\vartheta_i]$ that maps any $\btheta$ to $\btheta'$, where $\theta'_i$ is the minimum contribution agent $i$ has to make so that $u_i(\theta'_i, \btheta_{-i})\geq \mu_i$. This is formally defined  by
$\bFlocal(\btheta) := (f_i(\btheta))_{i\in[k]}$, where
    \begin{align*}
            f_i(\btheta) = \arg\min_{x\geq 0} u_i(x, \btheta_{-i}) \geq \mu_i\,.
    \end{align*}
Due to the monotonicity of $u_i$s and the definition of $\vartheta_i$s, it is easy to show that $f_i(\btheta)\leq \vartheta_i$.

Fixed points of function $\bFlocal$, i.e., those $\btheta$ for which $\bFlocal(\btheta) = \btheta$, refer to the equilibria of the collaborative learning game.
This is because, by definition, $f_i(\btheta)$ is the smallest contribution from agent $i$ that can satisfy agent $i$'s constraint in response to other agents' contributions $\btheta_{-i}$. Therefore, when $\theta_i = f_i(\btheta)$ for all $i\in [k]$, no agent can unilaterally reduce their contribution and still satisfy their constraint. That is, such $\btheta$ is an equilibrium. Therefore, to prove Theorem~\ref{thm:eqexist}, it suffices to show that the best-response function {$\bFlocal$} has a fixed point.

\begin{restatable}{lemma}{rstfixedpt}\label{lmm:fixedpt}
If utilities are well-behaved over $\bigtimes_{i=1}^k[0,\vartheta_i]$, the best-response function $\bFlocal$ has a fixed point, i.e., $\exists \btheta\in \bigtimes_{i=1}^k[0,\vartheta_i], \bFlocal(\btheta) = \btheta$.
\end{restatable}

We defer the proof of Lemma~\ref{lmm:fixedpt} to Appendix~\ref{app:fixedpt}. At a high level, we show that $f$ is continuous because, for well-behaved utility functions, a small change in other agents' contributions affects the utility of agent $i$ only by a small amount. Thus, a small adjustment to agent $i$'s contribution will be sufficient to re-establish her constraint when other agents make infinitesimally small adjustments to their strategies.
Then, combining this with the celebrated Brouwer fixed-point theorem proves this lemma.

\subsection{Proof of Theorem~\ref{thm:eqintexist}}\label{sec:proofofeqexist}

In this section, we prove Theorem~\ref{thm:eqintexist} and show that an equilibrium might not exist if the utility functions are not well-behaved.
We demonstrate this using a simple example where the utility function corresponds to the accuracy of classifiers in a general PAC learning setting with integral value strategies. We give a more general construction in Appendix~\ref{app:general-consruction}.

We consider the problem in the binary classification setting where one agent's marginal distribution reveals information about the optimal classifier for another agent.

Consider the domain $\cX = \{0,\ldots,5\}$ and the label space $\cY = \{0, 1\}$. We consider agents $\{0,1,2\}$ with distributions $\cD_0,\cD_1,\cD_2$ over $\cX\times \cY$. Let $\oplus$ and $\ominus$ denote addition and subtraction modulo $3$. 

We give a probabilistic construction for  $\cD_0,\cD_1,\cD_2$. Take independent  random variables $Z_0, Z_1, Z_2$ that are each uniform over $\{0,1\}$. For each $i\in\{0,1,2\}$, distribution $\cD_i$ is a point distribution over a single instance-label pair $(2i + z_i, z_{i\ominus 1})$.
In other words, the marginal distribution of $\cD_i$ is equally likely to be the point distribution on $2i$ or $2i+1$. Moreover, the labels of points in distribution $\cD_{i\oplus 1}$ are decided according to the marginal distribution of $\cD_{i}$: If the marginal distribution of $\cD_{i}$ is a point distribution supported on $2i$ then any point in $\cD_{i\oplus 1}$ is labeled $0$, and 
if the marginal distribution of $\cD_{i}$ is a point distribution on $2i+1$ then any point in $\cD_{i\oplus 1}$ is labeled $1$.

Consider the optimal classifier conditioned on the event where agent $i$  takes a sample $(2i + z_i, z_{i\ominus 1})$ from $\cD_i$ and no other agents takes any samples. This reveals $z_i$ and $z_{i\ominus 1}$. 
Therefore, the optimal classifier conditioned on this event achieves an accuracy of $1$ for agent $i$ (by classifying $2i$ and $2i+1$ as $z_{i\ominus 1}$) 
and agent $i\oplus 1$ (by classifying $2(i\oplus 1)$ and $2(i\oplus 1)+1$ as $z_i$).
On the other hand, the optimal label for instances owned by agent $i\ominus 1$, is $Z_{i\oplus 1}$.
By the independence of random variables $Z_0, Z_1$, and $Z_2$, we have that $Z_{i\oplus 1}$ is uniformly random over $\{0,1\}$ even conditioned on $z_i$ and $z_{i\ominus 1}$. 
Therefore, the optimal classifier has an expected error of $1/2$ for agent $i\ominus 1$.
Using a similar analysis, if any two agents each take a single sample from their distributions, the accuracy of the optimal classifier for all agents is $1$.

We now formally define the strategy space and utility functions that correspond to this setting. Let $\Theta = \{0,1\}^3$ to be the set of strategies in which each agent takes zero or one sample.
Let $\bmu = \mathbf{1}$.
Let $u_i(\btheta)$ be the expected accuracy of the optimal classifier given the samples taken at random under $\btheta$.
As a consequence of the above analysis, 
$$u_i(\btheta) = \begin{cases}
    1 & \theta_i = 1 \text{ or } \theta_{i\ominus 1} = 1\\
    \frac 12 & \text{otherwise}
\end{cases}
$$

Note that any $\btheta\in \Theta$ for which $\|\btheta\|_1 \geq 2$ is a feasible solution, while no $\|\btheta\|_1 \leq 1$ is a feasible solution.
Now consider any $\btheta$ for which $\|\btheta\|_1 \geq 2$. Without loss of generality, there must be an agent $i$ such that $\theta_i = \theta_{i\ominus 1}=1$. Since $\theta_{i\ominus 1}=1$, we also have that $u_i(0, \btheta_{-i}) = 1$. That is agent $i$ can deviate from the strategy and still meet her constraint. Therefore, no feasible solution is a stable equilibrium. This proves Theorem~\ref{thm:eqintexist}.

\section{Quantitative Bounds on Price of Stability and Price of Fairness}
\label{sec:sample-complexity}

As shown in Section~\ref{sec:exist}, while an envy-free solution always exists, the existence of stable equilibria depends on the properties of the utility function.
In this section, we go beyond existence and give quantitative bounds on the sub-optimality of these equilibria notions even when they exists in the presence of (very) well-behaved functions.

\begin{restatable}{theorem}{rstflowergap}\label{thm:flowergap}
There is a collaborative learning setting with well-behaved utility functions such that the Price of Stability and Price of Fairness are at least $\Omega(\sqrt{k})$.
Moreover, these utilities correspond to two settings: a) a random domain coverage example with uniform distributions over equally sized subsets and b) a linear utility setting with $W_{ii}=1$ and $W_{ij}\in O(1/\sqrt{k})$ for $j\neq i$.
\end{restatable}

We provide an overview of the proof of Theorem~\ref{thm:flowergap} here and defer the details of this proof to Appendix~\ref{app:flowergap}.
Our construction for the random coverage and linear utility settings are very similar, here we only discuss the random coverage setting.
The crux of our approach is to build a set structure where
one agent, called the core, overlaps with all other agents and no two agent sets intersect outside of the core.
We use a relatively small $\mu_i$s so that  every agent only needs to observe one of the points in her set.
In our construction, the core is the most  ``efficient'' agent in reducing the error of all other agents and optimal collaboration puts a heavy sampling load (of about $\sqrt{k})$ on the core.
Moreover, because the core includes all the points on which two other agents intersect, the core's constraint is also easily satisfied when any other agent's constraint is satisfied. This means that in no stable or envy-free equilibrium the core can take more samples than another  agent. Therefore, most of the work has to be done by other agents in any equilibrium allocation, which requires a total of $k$ samples.
This tradeoff between being both the most ``efficient'' at sampling to reduce error and having an ``easy-to-satisfy constraint'' leads to a large Price of Stability and Price of Fairness.

\section{Structural and Algorithmic Perspectives}
\label{sec:algs}
In this section, we take a closer look at the stable equilibria of the two canonical example where they are guaranteed to exist, i.e.,  the linear utilities and the coverage utilities, and study their structural and computational aspects.

\subsection{Algorithms for Linear Utility}\label{sec:linearcvx}
Recall that linear utility functions are functions $\bu(\btheta)= W\btheta$ where $W\in [0,1]^{k\times k}$, where $W_{ij}$ denotes how the efforts of agent $j$ affects the utility of agent $i$. In this section, we assume that $W$ is a symmetric PSD matrix~\footnote{This matches our motivating use-case defined in Section~\ref{sec:model-examples}} with an all $1$ diagonal.

An immediate consequence of linear utilities is that the optimal collaborative solution can be computed using the following linear program efficiently
\begin{equation} \label{eq:LP}
\begin{array}{ll}
\min &\sum\limits_{i=1}^k \theta_i \\
\st &W\btheta \geq \bmu\\
&\btheta \geq \bZero.
\end{array}
\tag{LP 1}
\end{equation}
Interestingly, the set of stable equilibria of linear utilities are also convex and the optimal stable equilibrium can be computed using a convex program.
To see this, note that any solution to \ref{eq:LP} satisfies the constraints $\theta_i(W_i^\top \btheta -\mu_i)\geq 0,\forall i\in[k]$, where $W_i$ denotes the $i$-th column of $W$.
Hence, adding the constraints $\theta_i(W_i^\top \btheta - \mu_i)\leq 0,\forall i\in[k]$ to \ref{eq:LP} will further restrict the solution to be a stable equilibrium where $\theta_i = 0$ or $W_i^\top \btheta =\mu_i$.
Given that any stable equilibrium meets both of these constraints with tight equality of $0$, they can be equivalently represented by the following convex program.
\begin{theorem}
The following convex program computes an optimal stable equilibrium of collaborative learning with linear utility functions
\begin{equation}\label{eq:lineareq}
\begin{array}{ll}
\min &\sum\limits_{i=1}^k \theta_i \\
\st &W\btheta \geq \bmu \\
&\btheta \geq \bZero\\
&\btheta^\top W \btheta-\mu^{\top} \btheta \leq 0,
\end{array}
\tag{CP 1}
\end{equation}
where the last inequality is convex when $W$ is PSD.
\end{theorem}

\subsection{Structure of Equilibria for Linear Utility}
In this section, we take a closer look at the structural properties of stable and envy-free equilibria and provide a qualitative comparison between them and the optimal solutions.
The main result of this section is that in any optimal stable equilibrium, there is a core subset of $k$ agents for which the equilibrium is also a socially optimal collaboration, while all other agents' contributions are fixed at $0$.

\begin{restatable}{theorem}{rstthmlinearopt}\label{thm:lineareqopt}
Let $\btheta^\eq$ be an optimal stable equilibrium for linear utilities $u_i(\btheta) = W^\top_i\btheta$ and $\mu_i = \mu$ for $i\in [k]$, where $W$ is a symmetric PSD matrix.
Let $I_{\btheta^\eq} = \{ i\mid \theta_i^\eq = 0\}$ be the set of non-contributing agents and let $\bar W$ and $\bar\btheta^\eq$ be the restriction of $W$ and $\btheta^\eq$ to  $[k]\setminus I_{\btheta^\eq}$.
Then $\bar\btheta^\eq$ is \emph{a socially optimal solution} for the set of agents $i\in [k]\setminus I_{\btheta^\eq}$, i.e., agents with utilities $u_i(\bar\btheta) = \bar{W}_i^\top \bar\btheta$ for $i\in [k]\setminus I_{\btheta^\eq}$.

Furthermore, let $\tilde{\btheta}$ represent the extension of $\bar{\btheta}$ by padding $0$s at $I_{\btheta^\eq}$, i.e., $\tilde{\theta}_i=0$ for $i\in I_{\btheta^\eq}$ and $\tilde{\theta}_i = \bar{\theta}_i$ for $i\in [k]\setminus I_{\btheta^\eq}$. For any $\bar\btheta$ that is a socially optimal solution for agents $[k]\setminus I_{\btheta^\eq}$, $\tilde{\btheta}$ is an optimal stable equilibrium for agents $[k]$.
\end{restatable}

This theorem implies that any equilibrium in which all agents have non-zero contribution has to be socially optimal.

\begin{restatable}{corollary}{rstcrllinearopt}\label{cor:lineareqopt}
Consider an optimal equilibrium $\btheta^\eq$. If $\btheta^\eq > \bZero$, then $\btheta^\eq$ is socially optimal.
\end{restatable}

An advantage of Corollary~\ref{cor:lineareqopt} is that in many settings it is much simpler to verify that every agent has to contribute a non-zero amount at an equilibrium without computing the equilibrium directly. One such class of examples is when matrix $W$ is a \emph{diagonally dominant} matrix, i.e., $\sum_{j\neq i}W_{ij}< W_{ii}$ for all $i\in [k]$, in addition to satisfying the requirements of Theorem~\ref{thm:eqexist}.
In this case, every agent can satisfy their own constraint in isolation using $\vartheta_i = 1/\mu$ contribution. Therefore, in any stable equilibrium the total utility an agent will receive from all others (even at their maximum contribution of $1/\mu$) is not sufficient to meet her constraint. Therefore, 
every agent has a non-zero contribution in an equilibrium. This shows that the Price of Stability corresponding to diagonally dominant matrices is $1$. 

We defer the proofs of Theorem~\ref{thm:lineareqopt} and Corollary~\ref{cor:lineareqopt} to Appendix~\ref{app:lineareqopt}. At a high level, our proofs use the duality framework and the linear program~\eqref{eq:LP} and convex program~\eqref{eq:lineareq}. 
At a high level, the first part of Theorem~\ref{thm:lineareqopt}  follows from the observation that the dual problem of the linear program~\eqref{eq:LP} for the set of agents $i\in [k]\setminus I_{\btheta^\eq}$ is
\begin{equation*}
\begin{array}{ll}
\max_\by &\bOne^\top \by \\
\st &\bar{W} \by\leq \mu \bOne\\
&\by \geq \bZero\,.
\end{array}
\end{equation*}
Since $\btheta^\eq$ is a stable equilibrium with positive entries in $\bar \btheta^\eq$, it is not hard to see that $\bar{W} \bar \btheta^\eq = \mu \bOne$. Then we know that $\bar \btheta^\eq$ is not only a feasible solution to \eqref{eq:LP} for $[k]\setminus I_{\btheta^\eq}$ but also a feasible solution to its dual with the same value. Therefore, $\bar \btheta^\eq$ is a socially optimal solution for the set of agents $i\in [k]\setminus I_{\btheta^\eq}$.
A closer look at this dual also proves that any $0$ padding of a socially optimal solution for the set of agents $[k]\setminus I_{\btheta^\eq}$ is a stable equilibrium for the set of agents $[k]$ as well.

Lastly, in the linear utilities case, it is not hard to show that any stable equilibrium is also envy-free. 

\begin{restatable}{theorem}{rstlineareqef}\label{thm:lineareqef}
When $W_{ij}< W_{ii}$ for all $i,j\in [k]$, any stable equilibrium is also envy-free.
\end{restatable}

We defer the proof of Theorem~\ref{thm:lineareqef} to Appendix~\ref{app:lineareqef}. Theorem~\ref{thm:lineareqef} and Corollary~\ref{cor:lineareqopt} together highlight an advantage of optimal stable equilibria. Not only are these equilibria are socially optimal for a subset of agents (and in some cases for all agents) but also they satisfy the additional property of being envy-free.

\subsection{Coverage Utilities}
We complement the algorithmic and structural perspective of equilibria in the linear utility case with those for the random coverage utilities. Unlike the linear utility case, both the stable feasible set and the envy-free feasible set for the random coverage utilities are non-convex, which indicates that either optimal stable equilibrium or optimal envy-free equilibrium is intractable.
\begin{restatable}{theorem}{rstthmcvrgeqnoncvx}\label{thm:cvrgeqnoncvx}
There exists a random coverage example with strategy space $\Theta=\R_+^k$ such that $\Theta^\eq$ is non-convex, where $\Theta^\eq\subseteq \Theta$ is the set of all stable equilibria. 
\end{restatable}

We defer the proof to Appendix~\ref{app:ncvxef} and provide an overview of the proof of Theorem~\ref{thm:cvrgeqnoncvx} here. Consider an example where there are $2$ agents and both are with a uniform distribution over the instance space $\cX=\{0,1\}$ and $\mu_i=3/4$ for all $i\in [2]$.
Note that both $\be_1$ and $\be_2$ are stable equilibria, since both agents receive $3/4$ utility if either of them observe any one of the instances.
Now consider a convex combination of these two strategies $(\be_1+\be_2)/2$, i.e, each agent takes one sample with probability $1/2$. In this case, there is a small probability that when both agents sample they both uncover the same point. Thus they do not receive any marginal utility from the second sample. This means that the utility that both agents receive from 
$(\be_1+\be_2)/2$ is strictly less than $3/4$, that is, $(\be_1+\be_2)/2$ is not even a feasible solution let alone a stable equilibrium. For more details refer to Appendix~\ref{app:ncvxef}.

\begin{restatable}{theorem}{rstthmcvrgefnoncvx}\label{thm:cvrgefnoncvx}
There exists a random coverage example with strategy space $\Theta=\R_+^k$ such that $\Theta^\ef$ is non-convex, where $\Theta^\ef\subseteq \Theta$ is the set of all envy-free equilibria. 
\end{restatable}

We defer the proof to Appendix~\ref{app:ncvxef}. At a high level, considering a complete graph on $4$ vertices, we let each edge correspond to one agent and put one point in the middle of every edge and one point on every vertex. Then we let each agent's distribution be a uniform distribution over $\cX_i$, which is the $3$ points on agent $i$'s edge. In this example, we can obtain a envy-free equilibrium $\btheta^\ef$ by picking any perfect matching on this complete graph and then letting $\theta_i^\ef=1$ if edge $i$ is in this matching and $\theta_i^\ef=0$ otherwise. However, we can show that there exists a convex combination of two envy-free equilibria corresponding to two different perfect matchings such that it is not envy-free.

\section{Experimental Evaluation}
To demonstrate potential issues with not considering incentives in federated learning, we compare two federated learning algorithms that account for these incentives to different extents. We consider both federated averaging \citep{mcmahan2017communication} and a collaborative PAC-inspired algorithm based on \citet{blum2017collaborative, nguyen2018improved,chen:tight2018} called MW-FED. Federated averaging is envy-free as agents take the same number of samples in expectation. Unfortunately, FedAvg may find solutions that are far from any stable equilibrium. MW-FED does not explicitly guarantee envy-freeness or stability, however, we demonstrate that it produces solutions that are closer to being a stable equilibrium. This is due to the fact that it implicitly reduces the sample burden of those agents who are close to having satisfied their constraints.

\paragraph{Federating Algorithms} At a high level, FedAvg involves sending a global model to a set of clients and requesting an updated model (from some number of updates performed by the client) based on the client’s data. The server then calculates a weighted average of these updates and sets this as the new set of parameters for the model. MW-FED uses the Multiplicative Weight Update meta-algorithm and adjusts the number of samples that each agent has to contribute over multiple rounds.
MW-FED takes a fixed number of samples at each round, but distributes the load across agents proportional to weights $w_i^t$. In the first iteration, the load is distributed uniformly between the agents, i.e., $w_i^1 = 1$. At every new  iteration, the current global model is tested on each agent's holdout set. Distributions that do not meet their accuracy objective increase their $w_i^t$ according to the Multiplicative Weight Update.
A more detailed statement of the algorithm can be found in Appendix \ref{experimental_section}.

\paragraph{EMNIST Dataset} We study the balanced split of the EMNIST \citep{DBLP:journals/corr/CohenATS17}, a character recognition dataset of 131,600 handwritten letters. EMNIST provides a variety of heterogenous data while still remaining accessible enough to run a sufficient number of trials. We encourage further heterogeneity via a sampling technique that identifies difficult and easy points. Each agent is assigned 2000 points from some mixture of these two sets.
Implicitly, this creates agents who have varying degrees of difficulty in achieving their learning objectives.
From these $2000$ points, 1600 are selected as the training and 400 as a validation set. During the course of training, we say that a distribution's contribution is the fraction of its 1600 points that it will use during learning. That is, if an agent's contribution level is $0.01$, it will take a sample of $16$ points at the beginning of the optimization procedure and only uses those data when creating mini-batches.

\begin{figure}[H]
    \centering
 \includegraphics[width=.4\textwidth]{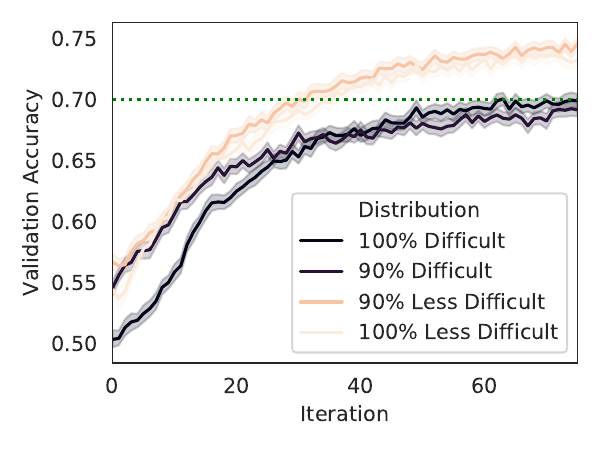}
    \caption[Validation accuracy of distributions with different difficulty levels.]{Each line here represents the average of 100 non-federated runs of a distribution used in this experiment. Note that the less difficult distributions reach the threshold quickly, whereas the more difficult distributions take nearly three times as long.}
    \label{fig:solo}
\end{figure}

For clarity of presentation, in these experiments we use four agents, two that have harder distributions and two that have easier distributions. Figure~\ref{fig:solo} shows the average performances of the four distributions without federation. Our observations and trends hold across larger sample sizes and with additional agents as shown in Appendix~\ref{experimental_section}. For training, we use a four-layer neural network with two convolutional layers and two fully-connected layers. For efficiency and to mirror real-world federated learning applications, we pre-train this model on an initial training set for 40 epochs to achieve 55\%  accuracy and then use federated training to achieve a 70\% accuracy level for all agents. More details on the dataset and model used can be found in Appendix~\ref{experimental_section}.

\paragraph{Results.}
To compare the two algorithms, we consider the resulting likelihood of any agent's constraint remaining satisfied when they unilaterally reduce their contribution level. Specifically, each agent wants to attain an accuracy of 70\% on their individual validation set. We chose this threshold as the easy distributions readily, individually converge above this level whereas, in our time horizon, the difficult distributions took, on average, nearly three times as long. See Figure~\ref{fig:solo} for the averaged individual performance trajectories. 
If an agent can drop their contribution level significantly and still attain this accuracy consistently during the optimization process, then either (a) other agents are oversampling and this agent is able to benefit from their over-allocation or (b) the agent was sampling too much to begin with relative to their requirements.

FedAvg makes no distinction between these cases. All agents contribute at an equal rate to convergence. On the other hand, MW-FED quickly reduces an agent's contribution level when she has met her constraints, reducing her ability to oversample.

\begin{figure}[H]
    \centering
     \includegraphics[width=.4\textwidth]{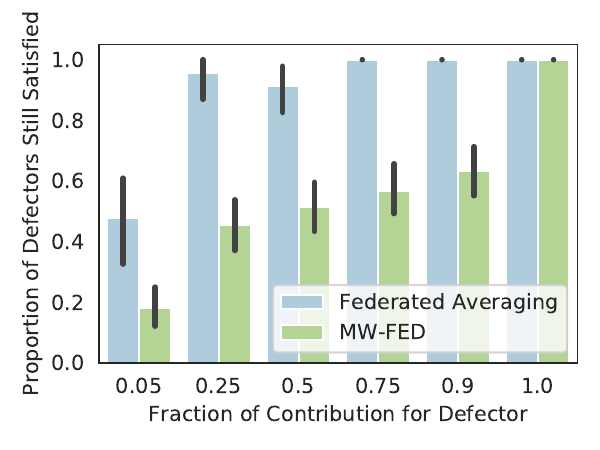}
    \caption[Comparing the likelihood that a single defector will reach their accuracy threshold at various contributions for federated averaging and MW-FED after 10 epochs.]{Comparing the likelihood that a single defector will reach their accuracy threshold at various contributions for federated averaging and MW-FED after 10 epochs. The result shows that MW-FED results in allocations that are closer to an equilibrium compared to FedAvg.
    }
    \label{fig:Doublebar}
\end{figure}

Figure~\ref{fig:Doublebar} shows the results of FedAvg and MW-FED run on the dataset 100 times. When everyone fully contributes, 100\% of these FedAvg runs satisfy the requirements of all agents by the tenth epoch. This figure compares the probability that, if a random single agent defected to a given contribution level, they would expect to have met their accuracy threshold at this point. For instance, if a single random agent only contributed 25\% of their data in FedAvg, they still have a 94\% chance of being satisfied by the tenth epoch. By comparison, only 45\% of agents at the same contribution level would succeed in MW-FED. This is striking as is discussed further in Appendix~\ref{experimental_section} where, even with pre-training, none of the agents in the individual (non-federated) setting reaches 70\% accuracy with 50\% or less of their data.
In Appendix~\ref{experimental_section}, we give one possible explanation for the performance of MW-FED by drawing parallels to algorithms in Section~\ref{sec:algs} that work in the linear setting.

\section{Conclusion}

Our paper introduced a comprehensive game theoretic framework for collaborative federated learning that considers agent incentives. Our theoretical results and empirical observations form the first steps in what we hope will be a collective push towards  designing equitable  collaboration protocols that will be essential for recruiting and retaining large numbers of participating agents.

\chapter{Incentives in Multi-Round Federated Learning}\label{chap:incentives-multi}
\section{Introduction}\label{sec:linear-defect}
Collaborative machine learning protocols have fueled significant scientific discoveries \citep{bergen2012genome} and are also gaining traction in diverse sectors such as healthcare networks \citep{li2019privacy, powell2019nvidia, roth2020federated}. A key factor propelling this widespread adoption is the emerging field of federated learning \citep{mcmahan2016federated} that allows multiple agents (also called devices) and a central server to tackle a learning problem collaboratively without exchanging or transferring any agent's raw data. Federated learning (FL) comes in various forms \cite{kairouz2019advances}, ranging from models trained on millions of peripheral devices \citep{mcmahan_ramage_2017, apple, paulik2021federated} like Android smartphones (a.k.a. cross-device FL) to those trained on a limited number of large data repositories (cross-silo FL). In this paper, we concentrate on a scenario that appears frequently when multiple organizations representative of an underlying population collaborate to get a consensus model. For instance, consider a medical study led by a government agency that selects several dozen hospitals to partake in the study to develop a model for the entire national populace. 

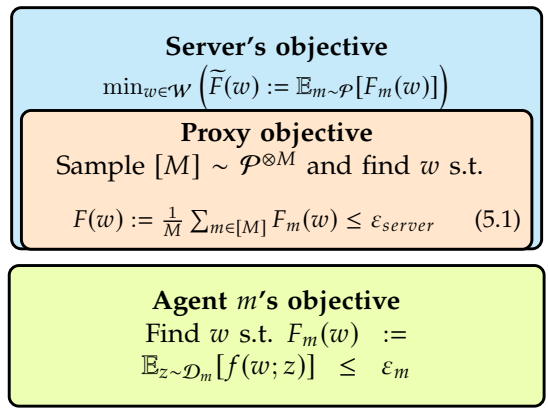
\begin{figure}
    \centering
    \resizebox{0.45\textwidth}{!}{

\begin{tikzpicture}[node distance=0.5cm,
                    serverbox/.style={draw, very thick, rounded corners, minimum size=3.5cm, align=center, fill=cyan!20, inner sep=9pt, text width=7cm, text depth=2.4cm},
                    proxybox/.style={draw, very thick, rounded corners, align=center, fill=orange!20, inner sep=5pt, text width=7cm},
                    agentbox/.style={draw, very thick, rounded corners, align=center, fill=lime!30, inner sep=10pt, text width=7cm},
                    >={Latex[width=2mm,length=2mm]}]

\node[serverbox] (server) {
    \textbf{Server's objective}\\ \begin{small}$\min_{w\in \www}\left(\tilde F(w) := \mathbb{E}_{m \sim \mathcal{P}}[F_m(w)]\right)$\end{small}
};

\node[proxybox, below=-2.05cm of server.south] (proxy) {
    \textbf{Proxy objective}\\ Sample $[M]\sim \mathcal{P}^{\otimes M}$ and find $w$ s.t. \begin{small} \begin{equation}\label{eq:relaxed} \textstyle F(w) := \frac{1}{M}\sum_{m\in[M]}F_m(w) \leq \epsilon_{server} \end{equation}\end{small}
};






\node[agentbox, below=2mm of server] (agents) {\textbf{Agent $m$'s objective}\\ Find $w$ s.t. $F_m(w):= \mathbb{E}_{z\sim \mathcal{D}_m}[f(w;z)] \leq \epsilon_m$};

\end{tikzpicture}
}
    \caption{Objectives of the server v/s the agents.}
    \label{fig:rational}
\end{figure}

Specifically, as depicted in Figure \ref{fig:rational}, the server (the governmental agency) has a distribution $\cP$ over agents (hospitals), and each agent $m$ maintains a local data distribution $\ddd_m$ (over its local patients). The server's goal is to find a model $w$ with low \textbf{population loss}, $\tilde F(w) := \mathbb{E}_{m\sim \cP, z\sim \ddd_m}[f(w;z)]$, where \(f(w;z)\) represents the loss of model \(w\) at datum \(z\).  Due to constraints like communication overhead, latency, and limited bandwidth, training a model across all agents (say, all the hospitals in a country) is infeasible. The server, therefore, aims to achieve its goal by sampling $M$ agents from $\cP$ and minimizing the proxy \textbf{average loss} $F(w) := \frac{1}{M}\sum_{m\in[M]} F_m(w)$ to precision $\epsilon_{server}$.


When all the $M$ agents share some optima---meaning a model exists that minimizes all the participating hospitals' objectives---such a model is desirable to the agents as well as the server since it also minimizes $F$. This paper reveals that this collaborative/federated learning approach falters when the agents, in addition to minimizing their objective, also aim to reduce their workload or, in the context of hospitals, aim to minimize data collection from their patients. More specifically, prevalent federated optimization techniques like federated averaging (\textsc{FedAvg}) \citep{mcmahan2016communication} and mini-batch SGD \citep{dekel2012optimal, woodworth2020minibatch} (refer to Appendix~\ref{app:setup}), employ a strategy of intermittent communication where in each round $r = 1, \ldots, R$:
\begin{itemize}
    \item \textbf{Server-to-Agent Communication: The server sends a synchronized model $w_{r-1}$ to all participating agents;}
    \item \textbf{Local Computation:} Every agent initiates local training of its model from $w_{r-1}$ using its own dataset;
    \item \textbf{Agent-to-Server Communication:} Each agent transmits the locally computed updates back to the server;
    \item \textbf{Model Update:} The server compiles these updates and revises the synchronized model to $w_r$.
\end{itemize}


If all agents provide their assigned model updates and collaborate effectively, the synchronized model should converge to a final model $w_R$, with a low average loss \( F(w_R) \). However, since the agents face various costs, rational agents might exit the process once they are content with the performance of the current model on their local data. For instance, in our example, hospitals may be satisfied with a model that performs well for their patient data.

\begin{figure}[t]
    \centering
    \includegraphics[width=0.5\textwidth]{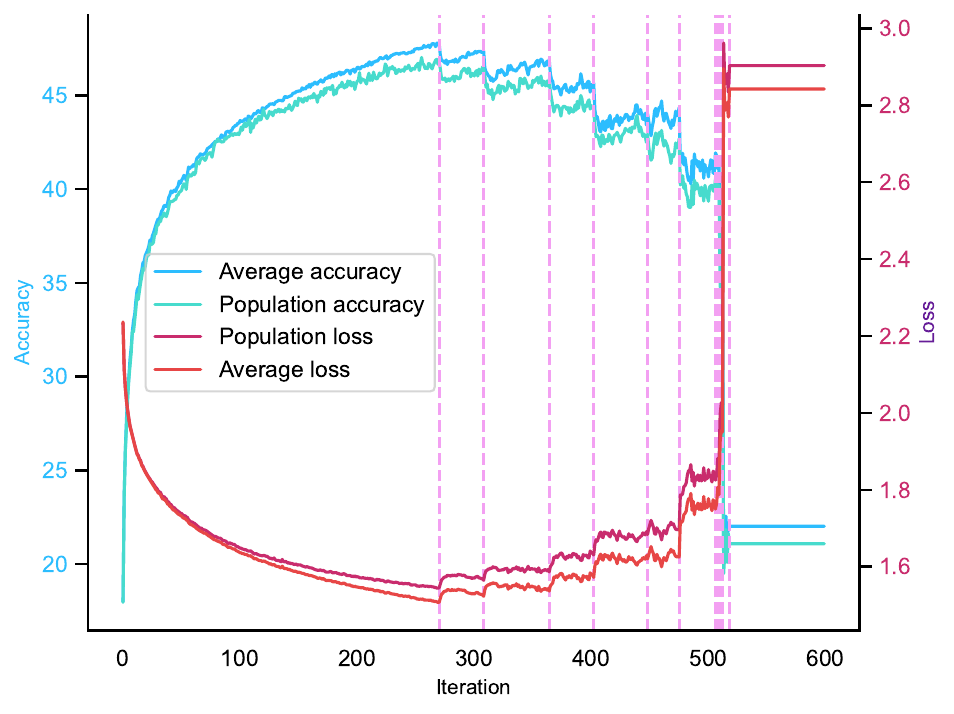}
    \caption[Impact of defections on both average and population accuracy metrics when using federated averaging with local update steps $K=5$ and step size $\eta = 0.1$.]{
    Impact of defections on both average and population accuracy metrics when using federated averaging with local update steps $K=5$ and step size $\eta = 0.1$. The CIFAR10 dataset \citep{krizhevsky2009learning} is processed to achieve a heterogeneity level of $q = 0.9$ (refer to appendix \ref{sec:exp} for more details). Agents utilizing a two-layer fully connected neural network with a softmax activation function to achieve a precision/loss threshold of $\epsilon = 0.2$. Dashed lines mark the iterations when an agent defects. It is evident that each defection adversely affects the model's accuracy. For example, the peak average accuracy drops from approximately 46\% prior to any defections to around 22\% after 500 iterations. A similar decline is observed in population accuracy.   
     }
    \label{fig:defection_time}
    \vspace{-1.5em}
\end{figure}

\begin{figure}
        \centering
        \includegraphics[width=0.3\textwidth]{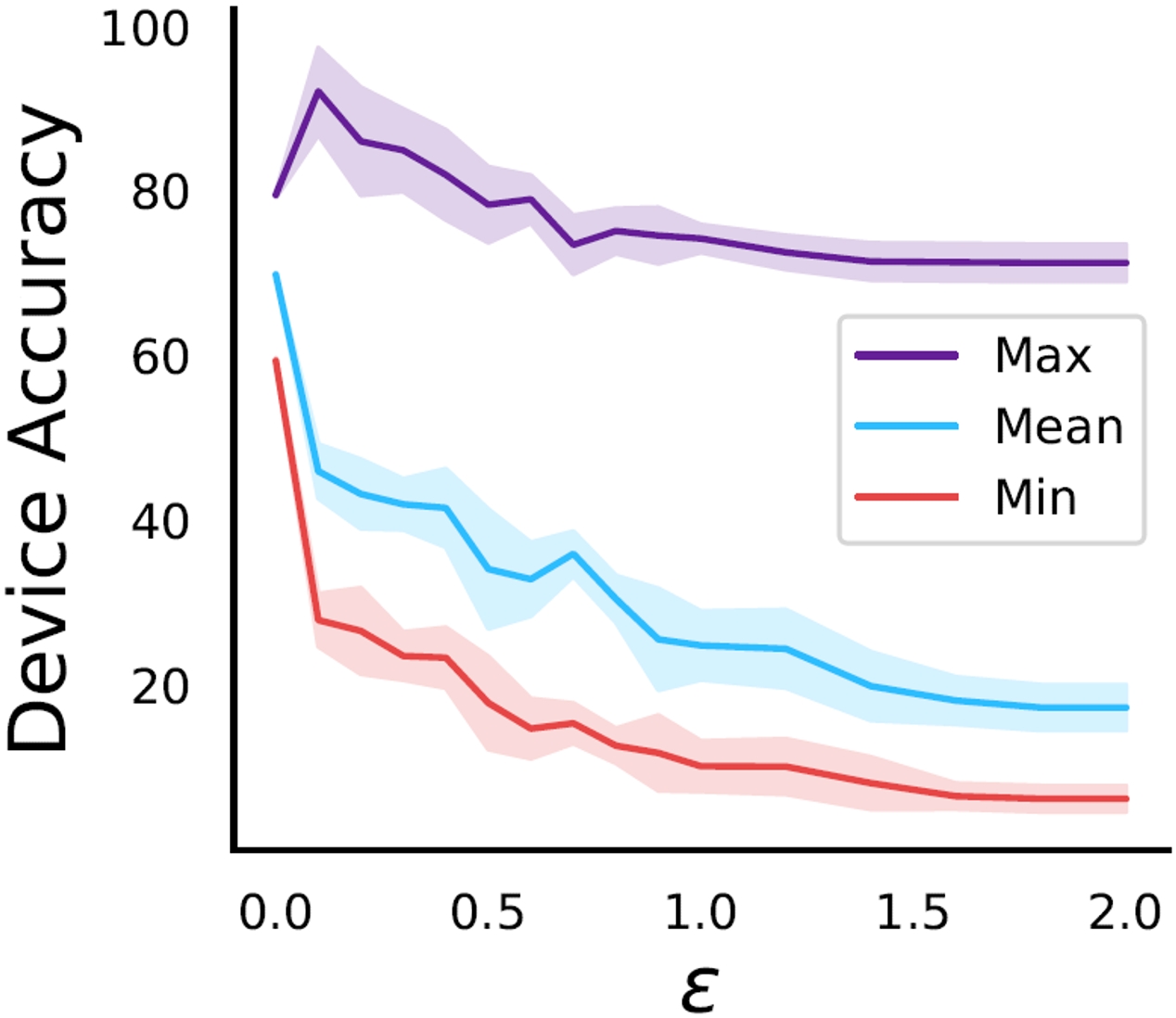}
        \caption[The minimum, maximum, and mean device accuracy out of the $M$ devices for the final \textsc{FedAvg} model as a function of the required precision $\epsilon$.]{In the above figure, we plot the minimum, maximum, and mean device accuracy out of the $M$ devices for the final \textsc{FedAvg} model as a function of the required precision $\epsilon$ (on the $x$-axis in the plot). As $\epsilon$ increases, the likelihood of each device defecting increases, so all the curves almost always decrease. The task is multi-class classification on CIFAR-10, and we simulate data heterogeneity by over-representing different classes on different agents (see Appendix \ref{sec:exp}). All experiments report accuracy averaged across 10 runs, along with error bars for $95\%$ confidence level.}
        \label{fig:defection_worker}
        \vspace{-1.5em}
\end{figure}

\textbf{\textit{Defections}}, or the act of permanently exiting before completing $R$ rounds, can adversely affect the quality of the final model $w_R$. This impact is particularly significant if the defecting agent had a large or diverse dataset or was the sole contributor to a specific type of data (for an illustration, see Figure \ref{fig:defection_cartoon}). Consequently, the final model $w_R$ may fail to achieve a low average loss $F(w_R)$ (for empirical evidence, see Figures \ref{fig:defection_time}, \ref{fig:defection_worker}, and \ref{fig:defection_train_full}). In general, defections can lead to:
\begin{itemize}
    \item \textbf{Suboptimal Generalization. Due to defections, the dataset may become imbalanced when the remaining agents do not adequately represent the broader population or their updates are too similar. The model may overfit this dataset (not representative of $\ppp$ anymore), leading to a higher generalization loss $\tilde F$ (see Figures \ref{fig:defection_time} and \ref{fig:defection_test_full}).}
    \item \textbf{Disparate Final Performance.} When agents belong to protected groups (e.g., based on gender or race), it is crucial for the model to be fair across these groups \citep{ezzeldin2021fairfed}. However, defections can result in a poor model for some agents (see Figure \ref{fig:defection_worker}).
    \item \textbf{Disproportionate Workload.} Even if defections do not directly harm the model's performance, they increase the workload for non-defecting agents, who need to provide additional updates to compensate.
\end{itemize}

Besides the aforementioned issues, defections can also lead to unpredictable outcomes when agents generate data in real-time \citep{huang2021federated, wei2023incentivized, patel2022distributed}, or when they undergo distribution shifts. Inspired by these observations, we aim to address the following questions:
\begin{center}
    \textbf{1. Under what conditions are defections detrimental to widely-used FL algorithms like \textsc{FedAvg}? \\
    2. Is it possible to develop an algorithm that mitigates defections while still optimizing effectively?}
\end{center}

\paragraph{Contributions.} Our contributions are as follows:
\begin{itemize}
    \item In Section \ref{sec:defections_hurt}, we distinguish between benign and harmful defections and explore the influence of (i) \textbf{initial conditions}, (ii) \textbf{learning rates}, and (iii) \textbf{aggregation methods}. Our findings indicate that simply averaging local updates is usually insufficient, and we show in Theorem \ref{obs:aggre} that \textsc{FedAvg} can not avoid harmful defections.
    \item In Section \ref{sec:alg}, we introduce our novel algorithm, \textsc{ADA-GD}. Under mild (and possibly necessary) conditions, we show in Theorem \ref{thm:two_agents} that \textsc{ADA-GD} can provably avoid agent defections while converging to a model that attains both agents' and server's objectives (see Figure \ref{fig:rational}).
    \item We extensively empirically validate our observations. We measure the effect of data-heterogeneity and target precision parameters on the final model obtained by \textsc{FedAvg} (in Figures \ref{fig:defection_time}, \ref{fig:defection_worker}, \ref{fig:defection_train_full}, \ref{fig:defection_test_full}, \ref{fig:defection_worker_full}). These experiments confirm that defections can substantially degrade the final performance of \textsc{FedAvg}. We also show that our algorithm \textsc{ADA-GD} effectively prevents defections and results in a superior final model compared to \textsc{FedAvg} (in Figure \ref{fig:our_algorithm}).   
\end{itemize}


\section{Related Work}\label{sec:related}
\begin{table*}
\vspace{-1em}
    \centering
     \renewcommand{\arraystretch}{1.25}
     \caption{Summary of existing work as discussed in Section \ref{sec:related} and Appendix \ref{app:related}. 
    $^\ddagger$ They consider a different model of agent rationality, where the agent compares the consensus model against the best model the agent can come up with individually. $^\star$Since in principle these works allow for adversarial defections, there could be repeated decision-making albeit without any rationality model.}
    \label{tab:related_work}
\end{table*}

The complexity of managing defections in federated learning arises primarily from two factors: (i) the repeated interactions between the server and the agents, allowing agents to access all intermediate models, and (ii) agents not disclosing their raw data to the server. 
Factor (i) motivates the study of repeated decision-making, where agents can decide how to join the collaboration at every round instead of making a decision at the beginning and committing to it.
These two factors together result in an information asymmetry, as the server can only speculate about potential defections and can not retract an intermediate model already exposed to an agent. This makes the problem challenging, and no current theoretical research can mitigate defections of rational agents, which interact over several rounds while also optimizing the server objective $F(\cdot)$ (see Table \ref{tab:related_work}). 

Most existing studies concentrate on single-round decision-making \citep{karimireddy2022mechanisms, blum2021one,cho2022federate,donahue2021model}, in which agents decide how much data they would like to contribute or how to collaborate with other agents at the beginning and commit to their decision in the whole training process.
\citet{blum2021one, karimireddy2022mechanisms} study the data sharing game where the action space of agents is the number of samples they would like to contribute and their utility is the accuracy of the model on their local data. 
While \citet{blum2021one} characterize the properties of the equilibrium of such a game, \citet{karimireddy2022mechanisms} provide a sample maximizing mechanism that is allowed to provide some fraction of the final samples to each agent. \citet{huang2023evaluating} also provide a sample maximizing mechanism, but they also incentivize the submission of samples that are most informative for minimizing the global objective. 
\textbf{Data sharing mechanisms}  offer an implicit guarantee for the average objective, as the server, having received the samples, can run a centralized algorithm on these samples. These also account for the individual rationality of agents, who want to minimize data contribution levels. 
But these mechanisms are unimplementable as they require writing the accuracy as an explicit function of data contribution levels.

\citet{cho2022federate,donahue2021model} consider the \textbf{core-stability} problem where the agents might want to join the global model or stay with their local
model.
\citet{cho2022federate}, design an alternative optimization objective to $F(w)=\frac{1}{M}\sum_{m\in[M]}F_m(w)$ in order to ensure that each agent ends up with a model $\hat w$ that is better than a model they could have come up with on their own. While their proposed algorithm is in the form of repeated interactions, the behavioral modeling of agents is still in a single round, i.e., the agents only decide to join or not once by speculating about the accuracy of the collaboratively trained model and their local best model. 
Overall, the paper by \citet{cho2022federate} lacks theoretical guarantees, including one for minimizing the average objective $F$, which is the goal of the server. \cite{donahue2021model} provide theoretical guarantees on the conditions for stable partitions of agents, and they don't provide algorithms to incentivize collaboration.

Another line of work on \textbf{collaborative PAC learning} \citep{blum2017collaborative, nguyen2018improved, haghtalab2022demand} considers multi-round interactions between the agents and the server. The aim of these works is to minimize the total number of samples collected across the agents while guaranteeing that each agent can attain its objective. To do so, a typical algorithm is \textsc{MW-FED}, which deliberately decelerates the advancement of agents nearing their target accuracy levels, thus demanding more samples from agents with harder objectives and vice versa. These works do not model agent rationality or try to prevent defections, so there are no formal guarantees against defections. However, as a side effect of minimizing sample complexity, \textsc{MW-FED} has fewer defections than \textsc{FedAvg} \citep{blum2021one}. 

Finally, there is a line of work in federated optimization that considers stochastic agent dropouts through \textbf{partial participation} models \citep{karimireddy2020mime, patel2022towards} as well as adversarial/\textbf{arbitrary device dropouts} \citep{wang2022friends, xu2023stabilizing}. These works provide guarantees for minimizing the average objective $F$ over repeated interactions in the face of these dropouts. However, this line of work is orthogonal as individual rationality is not behind these dropouts, making it impossible to disincentivize them. We provide an extensive literature review in Appendix \ref{app:related}.

\section{Problem Setup}\label{sec:setup-defect}
Recall that our (server's/government agency's) learning goal is to minimize $F(w) = \frac{1}{M}\sum_{m\in[M]}F_m(w)$ over all $w\in \www \subseteq \rr^d$, where $F_m(w):=\ee_{z\sim\ddd_m}[f(w;z)]$ is the loss on agent $m$'s data distribution. We assume the functions $F_m$'s are differentiable, convex, Lipschitz, and smooth. 
\begin{assumption}\label{ass:cvx_smth}
    Differentiable $F_m:\www\to\rr$ is convex and $H$-smooth, if for all $u, v\in \www$, 
    \begin{align*}
      F_m(v) &+ \inner{\nabla F_m(v)}{u-v} \leq F_m(u)\\
      &\leq F_m(v) + \inner{\nabla F_m(v)}{u-v} + \frac{H}{2}\norm{u-v}^2.  
    \end{align*}
\end{assumption}

\begin{assumption}[Lipschitzness]\label{ass:lip}
    Function $F_m:\www\to\rr$ is $L$-Lipschitz, if for all $u, v\in \www$, $$|F_m(u)-F_m(v)|\leq L\norm{u-v}.$$
\end{assumption}
Furthermore, the data distributions $\ddd_m$'s have to be ``similar'' so that all agents would benefit from collaboratively learning a single consensus model. There is no reason for two agents to collaborate if their data is labeled contrarily. Following~\cite {blum2017collaborative}, we capture the similarity among the agents using the following assumption.  
\begin{assumption}[Simultaneous Optimality]\label{ass:realizable}
    There exists $w^\star\in \www$ such that $w^*$ is the shared minima for all agents, i.e., $F_m(w^*) = \min_{w\in \www} F_m(w)$ for all $m\in [M]$. We denote the set of shared minima by $\www^\star$.
    For simplicity, we assume that $F_m(w^\star) =0$ for all $m\in [M]$, and all results also hold in the general case where the minimum value of each agent $F_m(w^\star) >0$.
\end{assumption}

This assumption holds when using over-parameterized models that easily fit the combined training data on all the agents. Furthermore, when this assumption is not satisfied, it is unclear if returning a single consensus model is even reasonable as opposed to a personalized model for each agent. 
We denote the $\epsilon_m$-sub level sets of the function $F_m$ by $S_m^\star$, i.e., $S_m^\star = \{w\in \www| F_m(w)\leq \epsilon_m\}$. Then agent $m$ wants to return a model in the set $S_m^\star$. Our simultaneous optimality assumption implies that $S^\star := \cap_{m\in[M]}S_m^\star \neq \emptyset$ and all the agents' learning goals can be achieved by outputting $w_R\in S^*$. Furthermore, outputting $\hat w\in S^\star$ will also ensure that, $F(\hat w) \leq \frac{1}{M}\sum_{m\in[M]}\epsilon_m$, which would satisfy the server as long as $\epsilon_{server}\geq \frac{1}{M}\sum_{m\in[M]}\epsilon_m$.
\vspace{-0.8em}
\begin{algorithm}[tbh]
    \caption{ICFO algorithm $\aaa(w_0, \eta_{1:R})$}\label{alg:icfo}
    \textbf{Parameters:} Initialization $w_0$, step sizes $\{\eta_{1}, \dots, \eta_R\}$
    
    \begin{algorithmic}[1]
    \STATE Initialize with all agents $\cM_0 = [M]$\\
    \FOR{$r = 1,\ldots,R$}{
        \STATE The server sends out $w_{r-1}$ to agents in $\cM_{r-1}$\\
        \FOR{$m\in \cM_{r-1}$}{
        \STATE Agent $m$ decides to drop out or to stay.
        }
        \ENDFOR
        \STATE Let $\cM_r\subseteq \cM_{r-1}$ denote the set of agents who stay\\
        \FOR{$m\in \cM_r$}{
        \STATE Agent $m$ sends its oracle output to the server\\
            $\ooo_m(w_{r-1}) = ( F_m(w_{r-1}), \nabla F_m(w_{r-1}))$
        }
        \ENDFOR
        \STATE The server aggregates the model by\\ $w_r = w_{r-1} - \eta_r\cdot h_r\rb{\{\ooo_m(w_{r-1})\}_{m\in \cM_r}}$
    }
    \ENDFOR
        
    \end{algorithmic}
    \textbf{Return:} $\hat{w}= w_R$
\end{algorithm}
\paragraph{Intermittently Communicating First-order (ICFO) Algorithms.}
We focus on intermittently communicating first-order (ICFO) algorithms, described in Algorithm~\ref{alg:icfo}. One example of such algorithms is \textsc{FedAvg} (see Algorithm \ref{alg:fedavg}). More specifically, at each round $r\in[R]$, the server sends the current model $w_{r-1}$ to all agents participating in the learning in this round. Every participating agent $m$ queries their first-order oracle \footnote{In practical implementations of these algorithms, usually the agent sends back an unbiased estimate $(\hat F_m(w_{r-1}), \nabla \hat F_m(w_{r-1}))$ of $(F_m(w_{r-1}), \nabla F_m(w_{r-1}))$. 
This is attained by first sampling data-points $\{z^m_{r,k}\sim \ddd_m\}_{k\in K_r^m}$ on machine $m$ and returning $\rb{\frac{1}{K_r^m} \sum_{k\in[K_r^m]}f(w_{r-1}; z_{r,k}^m), \frac{1}{K_r^m} \sum_{k\in[K_r^m]}\nabla f(w_{r-1}; z_{r,k}^m) }$. We focus on the setting with exact gradients and function values in our theoretical results, but our experiments consider stochastic oracles. When the agents do local update steps, as in \textsc{FedAvg}, they do not send gradient updates but update directions accumulating all local gradients.} at the current model $\ooo_m(w_{r-1}) = (F_m(w_{r-1}), \nabla F_m(w_{r-1}))$ and send it to the server, which updates the model. Note that each ICFO algorithm $\aaa$ has an aggregation rule $h_r(\cdot)$ for each round, which takes input $\{\ooo_m(w_{r-1})\}_{m\in \cM_r}$ and outputs a vector in $Span\{\nabla F_m(w_{r-1})\}_{m\in \cM_r}$. A widespread rule is \textit{uniform aggregation} where the aggregation rule $h_r\rb{\{\ooo_m(w_{r-1})\}_{m\in \cM_r}} = \frac{1}{\abs{\cM_r}}\sum_{m\in \cM_r}\nu(\{\ooo_m(w_{r-1})\}_{m\in \cM_r})\cdot \nabla F_m(w_{r-1})$ puts the same weight $\nu(\{\ooo_m(w_{r-1})\}_{m\in \cM_r})$ on all the gradients, where $\cM_r$ is the set of participating agents in round $r$. The simplest and most popular example is \textsc{FedAvg}/\textsc{FedSGD} which uses uniform aggregation, and sets $h_r\rb{\{\ooo_m(w_{r-1})\}_{m\in \cM_r}} = \frac{1}{\abs{\cM_r}}\sum_{m\in \cM_r}\nabla F_m(w_{r-1})$.

Furthermore, each algorithm has two parameters: (i) the initialization $w_0$, and (ii) step sizes $\eta_{1:R}$.  We will denote an instance of an ICFO algorithm by $\aaa(w_0, \eta_{1:R})$. 
We say an instance of an ICFO algorithm is \textbf{convergent} if it will converge to optima in $S^\star$ (in the limit $R\to \infty$) when agents are irrational, i.e., when agents never drop. When we refer to an ICFO algorithm, we are referring to the algorithmic procedure and not an instance of the algorithm. This would be crucial when we discuss the effect of initialization, step size, and aggregation rules in the next section.    

\begin{figure}
\begin{small}
    \centering
    \hspace{-1em}
    \scalebox{0.85}{
\tikzstyle{startstop} = [rectangle, rounded corners=5mm, 
text centered, 
text width=4cm,
minimum height=1.75cm,
draw=black, 
fill=orange!30,
very thick]

\tikzstyle{io} = [rectangle, rounded corners=4mm, 
text centered, 
draw=black, 
fill=blue!20,
very thick]

\tikzstyle{process} = [rectangle, rounded corners=5mm, 
text centered, 
text width=4cm,
minimum height=1.4cm,
draw=black, 
fill=teal!30,
very thick]

\tikzstyle{process2} = [rectangle, rounded corners=5mm, 
text centered, 
text width=3cm,
minimum height=1cm,
draw=black, 
fill=purple!30,
very thick]

\tikzstyle{decision} = [ellipse, 
text centered,
minimum height=1cm,
draw=black, 
fill=green!20,
very thick]

\tikzstyle{arrow} = [very thick,->,>=stealth]

\begin{tikzpicture}[node distance=0.8cm]

\node (start) [startstop] {The server sends the synchronized model \( w_{r-1} \) to all remaining agents \( \mmm_{r-1} \)};
\node (pro1) [process, left=of start] {Agents \( m \in \mmm_{r-1} \) evaluate their loss \( F_m(w_{r-1}) \)};
\node (dec1) [decision, below=of pro1, yshift=-0.2cm] {\( F_m(w_{r-1}) \leq \epsilon_m \)?};
\node (pro2a) [process, below=of dec1, yshift=-0.2cm] {Let \( \cM_r \subseteq \cM_{r-1} \) denote the set of agents who want to stay. For each \( m \in \cM_r \), agent \( m \) sends \( \ooo_m(w_{r-1}) \) back to the server};
\node (stop) [startstop, right=of pro2a] {The server aggregates the model by \( w_r = w_{r-1} - \eta_r \cdot h_r\rb{\{\ooo_m(w_{r-1})\}_{m\in \cM_r}} \)};
\node (pro2b) [process2, right=of dec1] {Agent \( m \) defects};

\draw [arrow] (start) -- (pro1);
\draw [arrow] (pro1) -- (dec1);
\draw [arrow] (dec1) -- node[anchor=east] {no} (pro2a);
\draw [arrow] (dec1) -- node[anchor=south] {yes} (pro2b);
\draw [arrow] (pro2a) -- (stop);

\end{tikzpicture}}
    \caption{A flowchart illustrating the steps in round $r$ of an ICFO algorithm when rational agents are present.}
    \label{fig:flowchart}
    \vspace{-1.5em}
\end{small}
\end{figure}
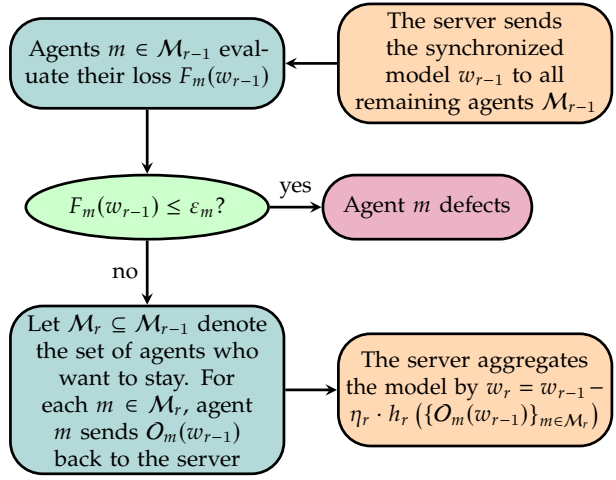

\paragraph{Rational Agents.}
Since agents have computation/communication costs to join this collaboration, they will defect when they are happy with the instantaneous model's performance on their local data. In particular, in the $r$-th round, after receiving a synchronized model $w_{r-1}$ from the server, agent $m$ will defect permanently if the current model $w_{r-1}$ is satisfactory, i.e.,  $F_m(w_{r-1})\leq \epsilon_m$, i.e., $w_{r-1}\in S_m^\star$. Then, agent $m$ will not participate in the remaining process of this iteration, including local training, communication, and aggregation. Thus, $m\notin \cM_{r'}$ for $r'\geq r$, where $\cM_{r'}$ is the set of participating agents in round $r'$. The detailed ICFO algorithm framework in the presence of rational agents is described in Figure~\ref{fig:flowchart}. Note that a sequence of agents may defect for an algorithm $\aaa$, running over a set of rational agents. We say the defection behavior (of this sequence of agents) \textbf{is harmful} (for algorithm $\aaa$) if the final output $w_R\notin S^\star$, i.e., we don't find a model in the $\epsilon$ sub-level set.

\begin{remark}
In the field of game theory and economics, the rationality of agents is commonly modeled as aiming to maximize their net utility, which is typically defined as the difference between their payoffs and associated costs, such as the difference between value and payment in auctions or revenue and cost in markets \citep{myerson1981optimal,huber2001gaining,borgers2015introduction}.
In our model, agent $m$ aims to obtain a loss below $\epsilon_m$ over their local data distribution (being indifferent to any loss below $\epsilon_m$), and therefore, their payoff upon receiving a model $w_r$ is $v(w_r) =V\cdot \ii[F_m(w_r)\leq \epsilon_m]$
for some $V>0$.
Their costs increase linearly with the number of training rounds they participate in, so the cost of obtaining the model $w_r$ is given by $c(w_r) = C\cdot r$ for some constant $C< V$. Thus, the optimal choice for agent $m$ is to defect \textbf{as soon as} they receive a model $w_r\in S_m^\star$. Note that there are other ways of modeling rational agents, such as in \citet{cho2022federate}, and we include a discussion on some other possible models in Appendix \ref{app:rational}.   
\end{remark}

\section{\textsc{FedAvg} Fails to Avoid Harmful Defections}\label{sec:defections_hurt}

One might naturally wonder whether defections in an optimization process are universally detrimental, consistently harmless, or fall somewhere in between for any given algorithm. Our experiments, confirm the existence of harmful defections. However, it's worth noting that not all defections have a negative impact; some are actually benign.
\begin{proposition}[Benign Defections]
    There exists a learning problem $\{F_1,F_2\}$ with two agents such that for any convergent ICFO algorithm instance $\cA(w_0, \eta_{1:R})$, any defection will be benign, i.e., $\cA(w_0, \eta_{1:R})$ will output a $w_R\in S^*$.
\end{proposition}
\begin{proof}
    Consider a $2$-dimensional linear classification problem where each point $x\in \R^2$ is labeled by $w^* = \mathbbm{1}[\be_1^\top x\leq 0]$. There are two agents where the marginal data distribution of agent $1$ is uniform over the unit ball $\{x|\norm{x}^2 \leq 1\}$ while agent $2$'s marginal data distribution is a uniform distribution over $\{\pm \be_1\}$. In this case, any linear model $w$ satisfying agent $1$, i.e., $F_1(w)\leq \epsilon$ (for any appropriate loss function) will also satisfy agent $2$. Therefore, for any convergent ICFO algorithm,  if there is a defection, it must be either agent 2's defection or the defection of both agents. In either case, the defection is benign for any convergent ICFO algorithm. This example can be extended to a more general case where agent $1$'s $\epsilon$-sub level set is a subset of agent $2$, i.e., $S^*_1\subset S^*_2$. 
\end{proof}

In practice, the ideal scenario where defections are benign is uncommon, as most defections tend to have a negative impact (refer to Figure~\ref{fig:defection_cartoon} for an illustration). Thus, the extent to which defections are harmful is largely algorithm-dependent. To gain insights into which algorithms can mitigate the adverse effects of defections, we will explore the roles of initialization, step size, and aggregation methods that characterize an ICFO algorithm (c.f., Algorithm \ref{alg:icfo}). To begin understanding the role of initialization, we characterize a \textit{``bad region''} for a given algorithm and problem instance.

\begin{figure}[t]
     \centering
     \begin{subfigure}{0.35\textwidth}
        \centering    \includegraphics[width=\textwidth]{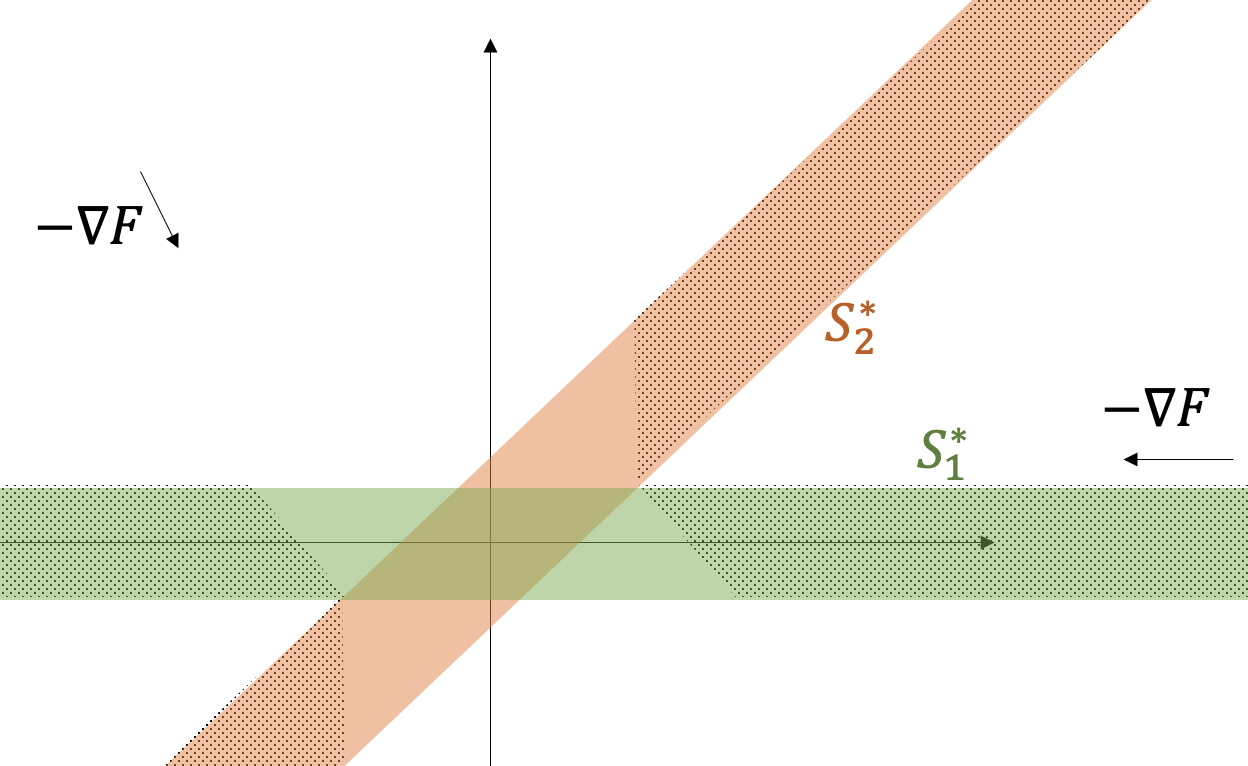}
         \caption{Contour illustration for Observation~\ref{obs:bad_region}.}
         \label{fig:example-def}
     \end{subfigure}
     \hfill
     \begin{subfigure}{0.35\textwidth}
         \centering
         \includegraphics[width=\textwidth]{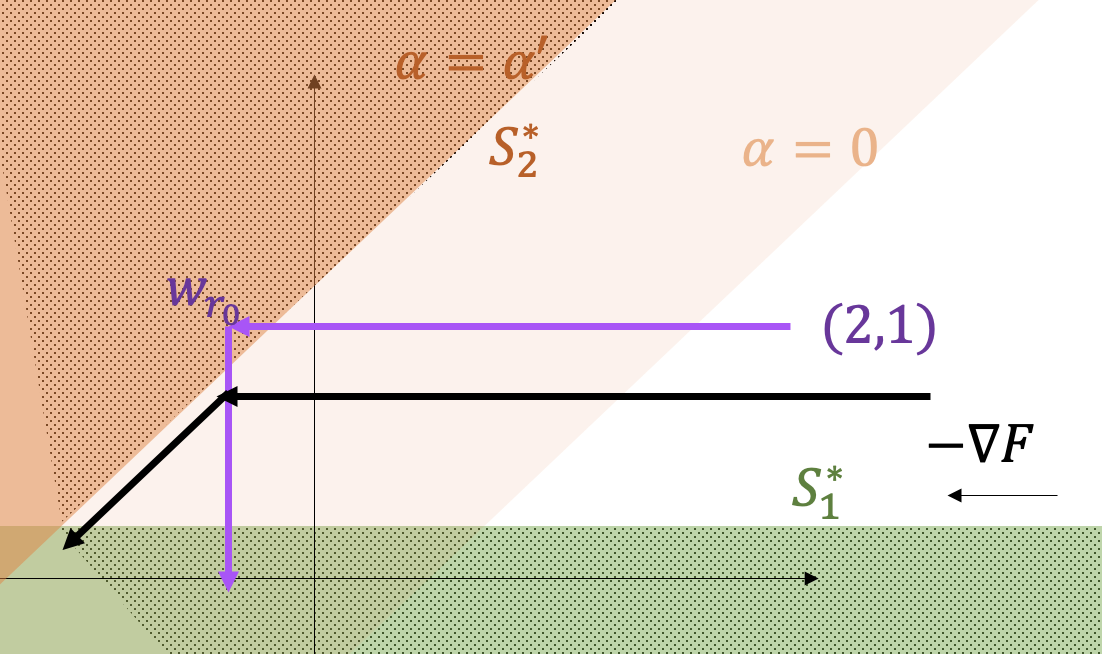}
         \caption{Contour illustration for Observation~\ref{obs:aggre}.
        }
         \label{fig:aggre}
     \end{subfigure}
        \caption{Contour illustrations of the examples in Section \ref{sec:defections_hurt}.}
        \label{fig:defection}
        \vspace{-1em}
\end{figure}

\begin{definition}[Bad Regions]\label{def:bad_region}
For any given problem $\{F_m\}_{m\in[M]}$, we define a subset of the domain, denoted by $\www_{bad}(\{F_m\}_{m\in[M]})\subseteq\www$, as a ``bad region'' of this problem, if for any ICFO algorithm $\aaa$, for every initialization $w_0\in \www_{bad}$, and for any step-size $\eta_{1:R}$, the instance $\aaa(w_0, \eta_{1:R})$ outputs the model $w_R\notin S^\star$. 
\end{definition}
Note that every problem instance characterizes a bad region, but can this region ever be non-empty? \textit{Yes}.
\begin{proposition}[Existence of Bad Initialization]\label{obs:bad_region}
    There exists a problem $\{F_1, F_2\}$ with two agents in which $\www_{bad}(\{F_1, F_2\})$ is non-empty.
    Specifically, there exists a $w_0\in S_1^*\cup S_2^*$ such that for any ICFO algorithm $\cA$ with any step-size $\eta_{1:R}$, when initialized at $w_0$, $\cA(w_0,\eta_{1:R})$ will converge to a model $w_R\notin S^*$, especially $F(w_R) \geq 1/2 >0 =F(w^*)$.
\end{proposition}

\begin{proof}
Consider an example with two agents with $\www = \R^2$,  $F_1(w) = \abs{(0,1)^\top w}$ and $F_2(w) = \abs{(1,-1)^\top w}$ as illustrated in Figure~\ref{fig:example-def}.   In this example we have the optimal model $w^* = (0,0)$ and the subgradients $\nabla F_1(w) = \sign((0,1)^\top w) \cdot (0,1)$ and $\nabla F_2(w) = \sign((1,-1)^\top w)\cdot (1,-1)$. If the algorithm is initialized at $w_0 = (1,1)$, agent $2$ defects given $w_0$ and an ICFO algorithm can only update in the direction of $\nabla F_1(w) = (0,\sign((0,1)^\top w))$. It will converge to a $w_R \in \{(1,\beta)|\beta\in \R\}$ and incurs $F(w_R)\geq \frac{1}{2}$ since $F(w) \geq \frac{1}{2}$ for all $w\in \{(1,\beta)|\beta\in \R\}$. 
Note that the dotted region in Figure~\ref{fig:example-def}, which is a subset of $S_1^*\cup S_2^*\setminus S^*$, is the bad region of this example. 
If initialized at any point in the dotted region of, one agent would defect immediately and the algorithm can only make updates according to the remaining agent and leads to a $w_R\notin S^*$.
Note that the functions in this example can be smoothed, and the above observation remains valid.
\end{proof}

Clearly, no ICFO algorithm can avoid harmful defections when initialized in the bad region. But again, in practice, the algorithm is unlikely to be initialized in the target sub-level sets of one of the agents, as these sets are challenging to find. The more interesting question is: \textit{can specific algorithms avoid harmful defections when initialized in the good region, i.e., outside the bad region?} The answer is \textit{yes}.
But, we will now show that the aggregation function is vital. 
We find that any ICFO algorithm with uniform aggregation---i.e., putting equal weights on all agents' updates---cannot avoid harmful defections even when initialized in a good region.

\begin{theorem}[\textbf{\textsc{FedAvg} fails even when initialized in a good region}]\label{obs:aggre}
    For ICFO algorithm $\cA$ with uniform aggregation with any step sizes $\eta_{1:R}$ and any initialization $w_0$, there exists a problem $\{F_1, F_2\}$ with two agents for which the initialization $w_0\notin \www_{bad}(\{F_1, F_2\})$,
    such that $\cA(w_0,\eta_{1:R})$ will output a model $w_R\notin S^*$.
    \end{theorem}


\begin{proof}
We build another slightly different example from the above for this observation. As illustrated in Figure~\ref{fig:aggre}, we consider a set of problems $\{F_1(w) = \max((0,1)^\top w,0), F_2(w) = \max((1,-1)^\top w+\alpha,0)|\alpha \geq 0\}$. For all $\alpha \geq 0$ illustrated in Figure~\ref{fig:aggre}, the problem characterized by $\alpha$ is denoted by $P_\alpha$. Suppose we initialize at a $w_0$ which is not in the union of $\epsilon$-sub level sets $S_1^*\cup S_2^*$ of $P_\alpha$ for all $\alpha$. More specifically, suppose the algorithm is initialized at $w_0= (2,1)$.
This assumption is w.l.o.g. since we can always shift the set of problems.
When no agents defect, the average gradient $\nabla F(w) = (1/2,0)$ is a constant, and any ICFO algorithm can only move in the direction of $(-1,0)$ until agent $2$ defects. 
Among all choices of $\alpha$, agent $2$ will defect earliest in the problem $P_0$ (since any model good for agent $2$ for some $\alpha>0$ is also good for agent $2$ for $\alpha = 0$). 
The algorithm will follow the purple trajectory in Figure~\ref{fig:aggre}. 
Before agent $2$'s defection, the first order oracle returns the same information in all problems $\{P_\alpha|\alpha \geq 0\}$ and thus,  the algorithm's updates are identical. 
Denote by $r_0$ the round at which agent $2$ is defecting in $P_0$. Then there exists an $\alpha'$ (illustrated in the figure) s.t. $w_{r_0}$ lies in the bad region of problem $P_{\alpha'}$. It is easy to check that Observation~\ref{obs:bad_region} also holds in problem $P_{\alpha'}$.
Therefore, in problem $P_{\alpha'}$, the algorithm will output a final model $w_R\notin S^*$. Especially, we have $F(w_R) = 1/2 >0=F(w^*)$.
\end{proof}
\begin{figure}[t]
    \centering
    \includegraphics[width=0.5\textwidth]{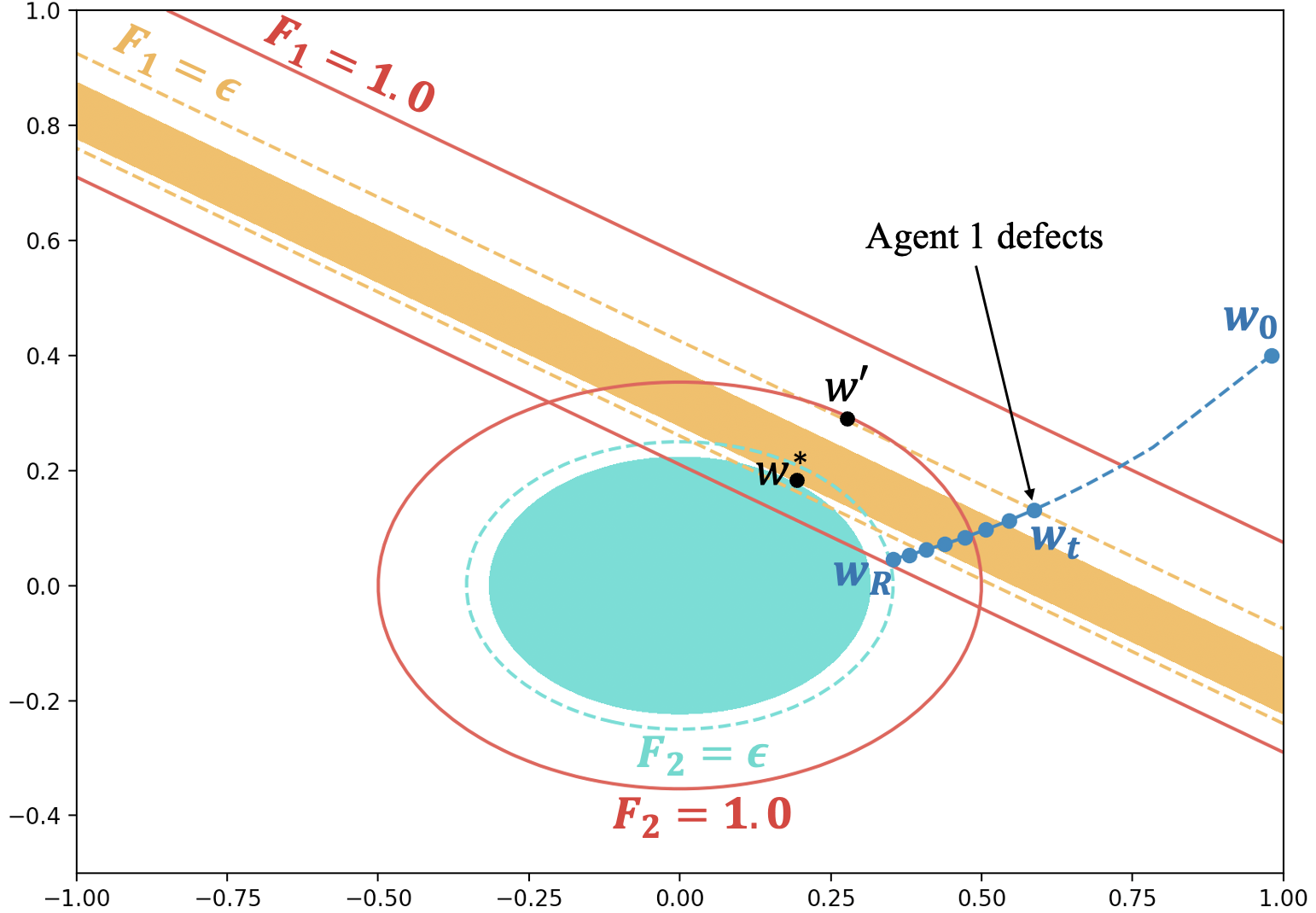}
    \caption{
Illustration of the contours of the example used to show the necessity of minimal heterogeneity (see Definition \ref{def:hetero}).}
    \label{fig:defection_example}
    \vspace{-1em}
\end{figure}

Based on the discussion so far, it is clear that if we insist on using an ICFO algorithm, we need to consider non-uniform aggregation to avoid harmful defections. One potential strategy is that when an algorithm reaches a $w$ with low $F_2(w)$ and high $F_1(w)$, we can project $\nabla F_1(w)$ into the orthogonal space of $\nabla F_2(w)$ and update in the projected direction. Figure~\ref{fig:aggre} provides a visual representation of the effectiveness of the projection technique, which is shown by the black trajectory within the context of this example. This not only showcases the benefits of projection but also serves as a motivation behind the development of our algorithm in the next section. But before we describe our algorithm, we state the final observation of this section, which would force us to constrain the class of problem instances on which we can hope to avoid defections with the above non-uniform aggregation strategy. Let us recall the definition of the orthogonal complement of a subspace.

\begin{definition}[Orthogonal Complement]
    Let $V$ be a vector space with an inner-product $\inner{\cdot}{\cdot}$. For any subspace $U$ of $V$, we define the orthogonal complement as $U^\perp := \{v\in V: \inner{v}{u} = 0,\ \forall u\in U\}$. We also define the projection operator onto a subspace $U$ by $\Pi_{U}:V\to U$. 
\end{definition}
  We use this to define a class of \textit{``good'' problems} for which we can avoid defections.
\begin{definition}[Minimal Heterogeneous Problems]\label{def:hetero}
    A problem $\{F_m:\rr^d\to \rr\}$ is called minimal heterogeneous if for all $w\in \www\setminus \www^\star$, all non-zero vectors in $\{\nabla F_m(w): m\in[M]\}$ are linearly independent. We denote by $\fff_\text{hetero}$ the class of all minimal heterogeneous problems.
\end{definition}
In most first-order algorithms, to avoid oscillation and make sure the optimization procedure converges, we set step sizes to be small. For instance, for smooth problems, the step size is set inversely proportional to the smoothness to stabilize the algorithm against the sharpest point on the optimization trajectory. With this in mind, we next observe that there exists a problem not in the class $\fff_\text{hetero}$ such that any ICFO with small step sizes cannot avoid harmful defections. Therefore, we restrict to minimal heterogeneous problems in our paper.
  
\begin{proposition}[Role of Minimal Heterogeneity]
    For any ICFO algorithm $\aaa$ with any initialization $w_0$ and any step-size sequence $\eta_{1:R}$, there exists a non-minimal heterogeneous problem with two agents $\{F_1, F_2\}\notin \fff_\text{hetero}$ for which the initialization $w_0\notin \www_{bad}(\{F_1, F_2\})$, such that there exists a constant $c\in (0,1)$ (which can depend on the problem) such that scaling step sizes with $c$ leads to $w_R\notin S^\star$, i.e., $\aaa(w_0, c\cdot \eta_{1:R})$ will output $w_R\notin S^\star$.
\end{proposition}

\begin{proof}
We provide an example in Fig~\ref{fig:defection_example}, where we plot the contour of $\{F_1,F_2\}$. Agent 1's loss function $F_1$ is piece-wise linear, while Agent 2's loss function $F_2$ is quadratic. Both functions are smoothed and truncated to have a minimum value of zero. 
The figure highlights zero-loss regions: filled orange for $F_1$ and green for $F_2$. 
Their shared optimum is $w^\star$ where $F_m(w^\star) = 0$. Assumptions \ref{ass:cvx_smth}, \ref{ass:lip}, and \ref{ass:realizable} are satisfied.
But this problem does not belong to the class of minimal heterogeneous problems as the gradients $\nabla F_1(w')$ and $\nabla F_2(w')$ are parallel at $w'$.
Dashed contour lines indicate the $\epsilon$ level set ($F_m(w) = \epsilon$), and solid lines represent the $1$ level set ($F_m(w) = 1$). 
Starting from $w_0$, any trajectory with small step sizes must pass the orange region to reach the $\epsilon$-sub level set of $F_2$. 
When the model updates to $w_t$ at time $t$, agent 1 would defect as $F_1(w_t) \leq \epsilon$. Subsequent updates follow $\nabla F_2(\cdot)$, and with a small step size, the model converges to $w_R$. This final model deviates from agent 1's $\epsilon$ level set, resulting in a poor performance with $F_1(w_R) = 1$.
\end{proof}



\section{Disincentivizing Defections through Different Aggregation Method}\label{sec:alg}
\begin{algorithm}[tbh]
\caption{\textsc{ADA-GD}}\label{alg:reweight}
\textbf{Parameters:} step size $\eta$, initialization $w_{0}$, precision requirements $\{\epsilon_1, \dots, \epsilon_M\}$, and slack $\delta$
\setlength{\abovedisplayskip}{3pt}
\setlength{\belowdisplayskip}{3pt}
\begin{algorithmic}[1]
    \FOR{$t=1,2,\ldots$}{
    \STATE Compute $(F_m(w_{t-1}), \nabla F_m(w_{t-1}))$ $\, \forall m\in[M]$\\
    \STATE \textbf{Predict defecting agents:} $D =$ \begin{small}\[ \{m\in[M]: F_m(w_{t-1}) - \eta\norm{\nabla F_m(w_{t-1})}\leq \epsilon_m + \delta\}\]\end{small}\\ 
    \STATE \textbf{Predict non-defecting agents:} $ND = [M]\setminus D$\\
    \IF{Both $D$ and $ND$ are non-empty}{
    \STATE Compute \begin{small}\[ \textstyle \nabla F_{ND}(w_{t-1}) = \sum_{m\in ND}\nabla F_m(w_{t-1}) \,\,\, \COMMENT{\textbf{Case 1}}\]\end{small}\\
    \STATE Let $P = Span\{\nabla F_n(w_{t-1}): n\in D\}^\perp$\\
    \STATE Project $\nabla F_{ND}(w_{t-1})$ and normalize \begin{small}\[\ngrad{ND}(w_{t-1}) = \frac{\Pi_P\rb{\nabla F_{ND}(w_{t-1})}}{\norm{\Pi_P\rb{\nabla F_{ND}(w_{t-1})}}}\]\end{small}\\
    \STATE Compute update $g_t =$ \begin{small}\[ -\min\cb{\norm{\Pi_P\rb{\nabla F_{ND}(w_{t-1})}}, 1}\cdot\ngrad{ND}(w_{t-1})\]\end{small}
    }
    \ENDIF
    \IF{$D$ is empty}{
    \STATE Compute update $g_t =$ \begin{small}\[ -\min\{\norm{\nabla F(w_{t-1})}, 1\}\cdot\frac{\nabla F(w_{t-1})}{\norm{\nabla F(w_{t-1})}} \, \COMMENT{\textbf{Case 2}}\]\end{small} 
    }
    \ENDIF
    \IF{$ND$ is empty}{
    \STATE \textbf{Return} $\hat{w}= w_{t-1}$  \hfill \COMMENT{\textbf{Case 3}}
    }
    \ENDIF
    \STATE $w_t = w_{t-1} + \eta g_t$
    }
    \ENDFOR
\end{algorithmic} 
\end{algorithm}

In this section, we state our algorithm Adaptive Defection-aware Aggregation for Gradient Descent (\textsc{ADA-GD}) in Algorithm \ref{alg:reweight}. At our method's core is an adaptive aggregation approach for the gradients received from agents, which disincentivizes participating agents' defection during the training. For minimal heterogeneous problems, i.e., problems in $\fff_{hetero}$ satisfying assumptions \ref{ass:cvx_smth}, \ref{ass:lip}, \ref{ass:realizable}, we show that our algorithm is legal (i.e., any denominator in Algorithm~\ref{alg:reweight} is non-zero) and converges to an approximately optimal model.
\begin{theorem}\label{thm:two_agents}
   Suppose $\{F_m\}_{m\in[M]}\in \fff_{hetero}$, satisfy Assumptions \ref{ass:cvx_smth}, \ref{ass:lip}, \ref{ass:realizable}. If we choose $w_0\notin \cup_{m\in[M]}S_m^\star$, $0 < \delta \leq \epsilon < 1$ and $\eta\leq \min\cb{\frac{\delta}{L}, \sqrt{\frac{\delta}{2H}}, \frac{1}{MH}}$, then no agents will defect and Algorithm~\ref{alg:reweight} will output $\hat w$ s.t.,
   \begin{itemize}
       \item for each \textbf{agent} $m\in[M]$, the objective $F_m(\hat w) \leq \epsilon_m + 2\delta$, and
       \item for the \textbf{server}, the proxy objective $F(\hat w) \leq \frac{1}{M}\sum_{m\in[M]} \epsilon_m + 2\delta$.
   \end{itemize}
\end{theorem}

Now we outline the high-level ideas behind Algorithm~\ref{alg:reweight}.
Intuitively, an agent is close to defection if a direction $u$ exists such that she would defect after receiving $w_{t-1}-\eta u$. Hence, the server has to carefully tune the update direction to avoid the defection of this agent. Suppose no agent has defected upon receiving $w_{t-1}$. Then, the server will receive the first-order information $\ooo_m(w_{t-1})$ from all agents and determine a direction to update $w_{t-1}$ to $w_t$. In Algorithm~\ref{alg:reweight}, the server first \textbf{predicts} which agents are close to defection through a linear approximation (see line 3) 
and calls them `defecting' agents (although these agents might not defect because of the slack $\delta$ in the precision in line 3).

\begin{figure}[tbh]
    \centering
    \includegraphics[width=0.48\textwidth]{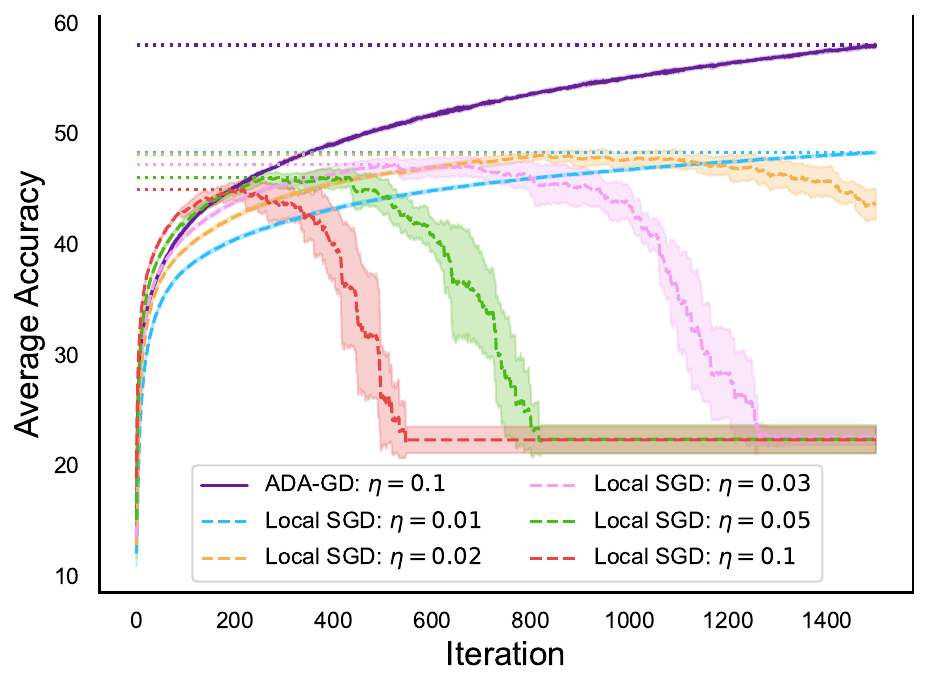}
    \caption[The performance of our method and federated averaging (Local SGD) for classification on CIFAR-10 \citep{krizhevsky2009learning} with 10 agents.]{We study the performance of our method and federated averaging (Local SGD) for classification on CIFAR-10 \citep{krizhevsky2009learning} with 10 agents. The data heterogeneity $q=0.9$ and local update step $K=5$. We observe that \textsc{ADA-GD} can avoid defection even when employing a considerably larger step size than Local SGD. As a result, this enables us to attain a significantly improved final model. In Figure \ref{fig:our_algorithm_appendix} of Appendix \ref{sec:exp}, we also present the performance of ADA-GD across different local steps. It is also noteworthy that Local SGD without defections, i.e., with the smallest step size, exhibited the highest accuracy among Local SGD curves. This observation emphasizes the harm of permanently losing data from a device. 
    All experiments report accuracy averaged across 10 runs, along with error bars for $95\%$ confidence level.}
    \label{fig:our_algorithm}
\end{figure}

If all agents are `defecting' (case 3), the server will output the current model $w_{t-1}$ as the final model (see line 15). The output model is approximately optimal since each agent has a small $F_m(w_{t-1})$ when they are `defecting'.
Suppose no agent is `defecting' (case 2). In that case, the server is certain that all agents will not defect after the update and thus will update in the steepest descending direction, i.e., the gradient of the average loss $\nabla F(w_{t-1})$ (see line 12).
If there exist both `defecting' and `non-defecting' agents (case 1), the server will aggregate the gradients from non-defecting agents, project it to the orthogonal complement of the space spanned by the gradients of the `defecting' agents, and normalize it to guarantee that `defecting' agents will not defect. Note that the projection step guarantees that the loss of `defecting' agents won't decrease while the normalization guarantees that the loss of the `non-defecting' agents does not decrease so much that they end up actually defecting. Then, the server will update the current model using the normalized projected gradients (see line 9).
By induction, we can show that no agents will defect. Furthermore, we can prove that we will make positive progress in decreasing the average loss at each round, and as a result, our algorithm will ultimately converge to an approximately optimal model. Our analysis is inherently case-based, making it difficult to replicate a simple distributed gradient descent analysis that proceeds by showing that we move in a descent direction at each iteration. Consequently, offering a convergence rate poses a significant challenge and remains an open question.

Note that Algorithm \ref{alg:reweight} fits in the class of ICFO algorithms specified in Section~\ref{sec:setup-defect}, as orthogonalization and normalization return an output in the linear span of the machines' gradients. In Figure \ref{fig:our_algorithm}, we compare \textsc{ADA-GD} against \textsc{FedAvg}, demonstrating the benefit of adaptive aggregation.

\section{Discussion}\label{sec:discussion-defect}
In this work, we initiate the study of incentives of heterogeneous agents in the multi-round federated learning. We find that defections are an unavoidable part of federated learning and can have drastic consequences for the performance/robustness of the final model. We theoretically and empirically characterize the effects of defection and demarcate between benign and harmful defections. We underline the importance of adaptive aggregation in avoiding defections and propose an algorithm \textsc{ADA-GD} with a provable guarantee and a promising empirical performance. There are several open questions and avenues for future work. 

\noindent\textbf{Convergence Rate.} Regarding theoretical analysis, we only provide an asymptotic convergence guarantee for Algorithm \ref{alg:reweight}. The non-asymptotic guarantee is an interesting open question. The main difficulty lies in that there are several phases and we analyze each phase separately. The standard technique for analyzing first-order methods cannot be applied in this case. 

\noindent\textbf{First Order Oracle.} 
We assume access to exact first-order oracles. Using stochastic oracles is common in practice, and constructing an algorithm based on stochastic oracles presents an intriguing question.

\noindent\textbf{Multiple Local Updates.} It is unclear how our Algorithm \ref{alg:reweight} can integrate these local steps. Including local steps complicates the defection model since agents might defect before communication occurs. This complication makes orthogonalization challenging in Algorithm \ref{alg:reweight}. A naive strategy is to make the defection prediction rule more stringent, while ensuring the orthogonalization happens for the aggregate local update direction on each machine.


\noindent\textbf{Approximately Realizable Setting.} In this work, we focus on the realizable setting, where there exists $w^*\in \www$ such that $F_m(w^*) = 0$ for all $m\in [M]$. However, the next natural question is: can we get similar results when $w^*$ is only approximately optimal for all agents?

\noindent\textbf{Non-convex Optimization.} We focus on the convex optimization setting. However, a natural follow-up question is how to avoid defections in non-convex federated optimization. Our theory doesn't capture this setting while we perform experiments in the non-convex setting. 

\noindent\textbf{Other Rational Behaviors.} We restrict to the setting where the server cannot save the intermediate models and wants the final model to be as good as possible. It is also interesting to consider settings where the server can save multiple intermediate models and use the best one when a new agent from the population arrives.

\chapter{Incentives in Collaborative Active Learning}\label{chap:incentives-active}
\section{Introduction}\label{sec:intro}
Active learning has emerged as a powerful paradigm in which labels of selected data points are sequentially queried from a large pool of unlabeled data, referred to as the unlabeled pool. The primary objective is to minimize labeling effort to find a classifier that exhibits low error on fresh data points from the same data source, known as generalization error. 
Typically, if the pool is large enough, a classifier that performs well on the pool can also achieve low generalization error through uniform convergence. 

Active learning has also been studied in the distributed setting, where the unlabeled pool is scattered across multiple machines (called agents), (e.g.,~\cite{shen2016distributed,aussel2020combining}). 
While active learning has demonstrated promising results, traditional approaches often operate in isolation, neglecting the potential benefits 
of collaboration among agents should they agree to collaborate. In this paper, we propose a novel framework for incentivized collaboration active learning, where agents can collaboratively explore their data pools to discover a common target function.

The motivation for collaboration in active learning stems from real-life scenarios where collaboration and collective intelligence yield improved outcomes, e.g., when agents collect data from the same distribution, and can easily end up labeling the same or very similar points. This redundancy leads to unnecessary and inefficient utilization of resources, as the labeling is often done by experts. Additionally, more data can be translated to improved accuracy, prompting agents to pool their resources and employ a more powerful model.

The incentive-driven nature of our framework aligns with the reality of collaboration in the real world. When agents are incentivized to collaborate only when their expected labeling complexity decreases, it reflects the real-life scenario where individuals are motivated to engage in cooperative endeavors if they perceive a clear benefit, such as reduced effort, faster, and better outcomes. In this work, we focus on a specific notion of incentives, where agents already have access to a baseline algorithm and they are motivated to join the collaboration if their label complexity is smaller than running the baseline algorithm on their own.

Consider, for example, the case of a new drug (e.g., Paxlovid for Covid-19\cite{Paxlovid22}, that has different efficacy on patients with different features. While individual hospitals can test the drug on their patients in an active learning fashion by executing their preferred baseline algorithm, collaborating efficiently with other hospitals, each with their own patients, often leads to a better prognosis. 

However, if the incentives of the hospitals are not maintained, i.e., the effort of some hospitals is increased, the collaboration may be compromised. By emulating this collaboration within the active learning framework, we unlock the potential of collective intelligence to enhance the learning process. 
Besides, imagine that several data labeling companies have to recover the labels of unlabeled images assigned to them. Each data labeling company would like to collaborate with other companies to recover the labels of all images while minimizing the query complexity and not increasing their burden.

Our basic model is as follows: there are $k$ agents, each with their own set of unlabeled data points, and a single hypothesis class with a prior on the hypotheses, which all agents are aware of.
We assume realizability, meaning that there exists an underlying ground truth labeling function called the target function, labeling all the data points, and that the hypothesis class encompasses such a target function. We refer readers to the discussion for more information about this assumption in the context of active learning.

The agents reach a consensus on an arbitrary baseline algorithm for pool-based active learning (e.g., the best tractable approximation algorithm).
To select whether or not to join the collaboration, the agents need to evaluate their utility from joining the collaboration. Since the goal of each individual agent (regardless of the collaboration) is to minimize their expected query complexity, the most natural cost function is the expected query complexity.
To ensure each individual benefits from their collaboration, we establish a collaboration protocol that guarantees that each agent cannot reduce their expected label complexity by running the baseline algorithm individually. This concept is referred to as individual rationality (IR).
Our objective is to design an IR collaboration protocol that minimizes the overall labeling queries.


There are cases in which collaboration is not necessarily beneficial. For example, each agent has non-zero points on a different axis and the hypothesis class contains every possible halfspace. If the prior distribution is uniform over all labelings, then no agent can reduce their label complexity by joining the collaboration.

\begin{figure}[t]
    \centering
    \scalebox{0.9}{
\begin{tikzpicture}[level distance=1.4cm,
  level 1/.style={sibling distance=4.2cm},
  level 2/.style={sibling distance=2cm}]

  \node (root) {What is the $0.5$'s label?}
    child {node (A) {What is the $0.7$'s label?}
      child {node (B) {$\ind{x\geq 0.8}$}}
      child {node (C) {$\ind{x\geq 0.6}$}}}
    child {node (D) {What is the $0.3$'s label?}
      child {node (E) {$\ind{x\geq 0.4}$}}
      child {node (F) {$\ind{x\geq 0.2}$}}};

  \path (root) -- (A) node[midway,left] {$0$};
  \path (root) -- (D) node[midway,right] {$1$};
  \path (A) -- (B) node[midway,left] {$0$};
  \path (A) -- (C) node[midway,right] {$1$};
  \path (D) -- (E) node[midway,left] {$0$};
  \path (D) -- (F) node[midway,right] {$1$};
\end{tikzpicture}}
    \caption{The query tree of binary search for thresholds.}
    \label{fig:query-tree}
\end{figure}
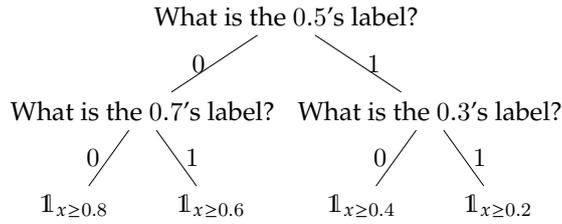

Clearly, if each agent has the same set of points, the label complexity of each agent can decrease to $1/k$ of its original label complexity if the collaboration protocol equally splits the labeling burden.
Even if agents do not share the same set of points, they can still benefit from collaboration, as we show in the next example.
\begin{example}\label{ex:treshold}
Consider the scenario with 1-dimensional thresholds $\cH =\{\ind{x\geq \alpha}|\alpha = 0.2,0.4,0.6,0.8\}$ and a uniform prior distribution over $\cH$. Suppose agent 1 has points $\{0.25,0.5,0.75\}$ and agent 2 has points $\{0.3,0.45, 0.55, 0.7\}$. 
When running binary search collaboratively, each agent only performs one labeling query, as illustrated as a search tree in Fig~\ref{fig:query-tree}. On the other hand, if they were to run binary search independently, each agent would need to query $2$ labels. Thus, collaboration can effectively reduce the label complexity for each agent by $1$.
\end{example}

 \textbf{Bayesian Assumption.} 
 The reason why we have a Bayesian assumption regarding the hypothesis class is that without it, querying all the labels to discover the target hypothesis can be inevitable, even for a simple class of linear separators in $\cR^2$ (see, e.g., Claim $1$ in~\cite{dasgupta2004analysis}).
It is worth noting that as in~\cite{dasgupta2004analysis}, we \textit{do not require the prior distribution to align with nature}. Instead, the prior distribution serves as a measure for average case analysis. Having a prior belief in our model has the following clear assumption. If the algorithm reaches a point where the remaining consistent hypotheses largely agree 
on the unlabeled data, it is reasonable to stop and output one of these remaining hypotheses~\cite{FreundSST97}. In a non-Bayesian setting, it does not make sense to operate this way. 

\textbf{Game Theory Interpretation.} 
The agreed-upon baseline algorithm induces a sort of (not private) values for agents- each agent has its (negative) individual labeling complexity as value. 
The collaboration protocol can be then interpreted as a mechanism: 
Initially, the collaboration protocol (principal) is introduced to the agents, and each agent can understand it and have confidence in the principal's commitment to implementing it faithfully. Subsequently, the agents either rely on their trust in the algorithm's IR property or have the ability to verify it autonomously. Lastly, the agents behave rationally by joining the collaboration only if it is IR.

We remark that there is an interesting parallelism between our IR collaboration algorithms and truthful mechanisms. It is well known that Vickrey–Clarke–Groves (VCG) mechanism is a truthful mechanism that maximizes social welfare, but since it is hard to compute and to approximate~\cite{Buchfuhrer10}, the optimal outcome is replaced by a sub-optimal outcome of an approximation algorithm, and the resulting mechanism is not necessarily truthful. The goal is therefore relaxed to design an efficient approximation algorithm that returns a truthful mechanism.

\textbf{Contributions and Organization.} We formalize the model in Section~\ref{sec:model-active}. In Section~\ref{sec:ir}, we demonstrate that any optimal algorithm is individually rational when the baseline is itself. This implies that optimizing for optimality ensures individual rationality for all baseline algorithms.

However, computing (or even approximating) the optimal algorithm is known to be NP-hard. To address this, we then show that the best available tractable approximation algorithm, the greedy algorithm \cite{kosaraju2002optimal,dasgupta2004analysis}, is not individually rational when the baseline is itself.
We demonstrate this by presenting an example where joining the collaboration increases the labeling complexity of an agent from $O(1)$ to $\Omega(n)$. 
In response, we introduce a general approach that can transform any arbitrary baseline algorithm into an IR collaborative algorithm. This conversion ensures that the total label complexity remains competitive with running the baseline algorithm on the entire data set. Furthermore, in Section~\ref{sec:sir} we present a scheme that converts any IR collaborative algorithm into a strict IR one, guaranteeing the label complexity is strictly lower by joining the collaboration under mild assumptions. 
When the baseline algorithm is both efficient and approximately optimal, our (strict) IR algorithms efficiently achieve label complexity that is approximately optimal.

\subsection{Related Work.}
The most related work is the recent work of \cite{xu2023fair}, which studies individual rationality in collaborative active learning in a Gaussian Process. While their notion of IR is similar to ours, we focus on query complexity in binary classification. \cite{echenique2019incentive} studied incentive compatibility in active learning, where there is a single agent that responds to a learner's query strategically.
Our work is situated at the junction of Learning in the presence of strategic behavior and active learning. 

\paragraph{Learning in the presence of strategic behavior}  encompasses a vast body of research, including~\cite{ben-porat23, Zhang22,hardt2016strategic}. 
We are particularly driven by prior research 
in this area, and how to create learning algorithms that incentivize agents to participate while maximizing the overall welfare.  For example, \textit{incentivized exploration} in Multi Arm Bandits~\cite{Kremer-JPE14,MansourSS15,MansourSSW16,MansourSW18, cohen2019optimal,Che-13,BaharST16,Bahar2019FiduciaryB,Bahar19,Immorlica19,Immorlica20,Sellke21,banihashem2023bandit,Slivkins17,slivkins2019introduction} or MDPs~\cite{simchowitz2023exploration}, where the principal recommends actions to the agents (in order to explore different alternatives), but the agents ultimately decide whether to follow the given recommendation. This raises the issue of incentives in addition to the exploration-exploitation trade-off. In particular,\cite{Baek21} study this problem in the context of fairness with a group-based regret notion. They show that regret-optimal bandit algorithms can be unfair and design a nearly optimal fair algorithm. Incentivizing agents to share their data has been studied by \cite{wei2023incentivized} in federated bandits.

\paragraph{Federated learning}  has gained popularity as a method to foster collaboration among large populations of learning agents among else for incentivizing participation and fairness purposes ~\cite{blum2021one,Lyu2020,donahue2021model,Donahue22Fair,DonahueK21,Donahue23, wang2023framework}. Our work also addresses fairness, in the sense that if a collaborative algorithm is individually rational, it is fair for all the participating agents. Another related line of research is \textit{kidney exchange}~\cite{Roth04,AshlagiR11,BlumIHPPV17kidney,BlumG21,BlumMansour20,DickersonPS19}, where the goal is to find a maximum match in a directed graph (representing transplant compatibilities between patient–donor pairs). In this problem, incentives arise in the form of individual rationallity when different hospitals have different subsets of patient–donor pairs, and will not join the collaboration if the number of pairs matched by the collaboration is lower than the number of pairs matched they could pair on their own.

\paragraph{Active learning} There are two basic models in active learning-- stream-based~\cite{FreundSST97} (where the learner has to determine immediately whether to query the label of the current instance or discard it), and pool-based, which is the basis for our model. Pool-based active learning investigates scenarios in which a learner is confronted with an array of unlabeled data points and the goal is to recover a target function by querying the labels of these points (see~\cite{Hanneke14survey} for a survey). Active learning has been studied in the context of other societal desiderata such as  fairness~\cite{ShenCW22,Camilleri23},  and safety~\cite{CamilleriWMJJ22}.

To our knowledge, no research has amalgamated these fields to explore strategic constraints in the context of active learning. This is where our work makes a valuable contribution.


\section{ Preliminaries and Model}\label{sec:model-active}
Throughout the work, we consider the binary classification problem. 
Let $\cX$ denote the input space, $\cY =\{0,1\}$ denote the label space, and $\cH\subset \cY^\cX$ denote the hypothesis class.\footnote{Results in this work can be directly extended to any active learning problem that can be formalized using a hypothesis class, e.g., multiclass classification.}
We focus on the \textit{realizable} setting in this work, namely, there exists a target hypothesis $h^*\in \cH$ correctly labels every point.
In the pool-based active learning setting \cite{Hanneke14survey}, given a collected unlabeled data set $X = \{x_1,\ldots,x_m\}$, the learning goal is to recover the labels of $X$. 
Now just suppose the pool of unlabeled data $x_1,\ldots,x_m$ is available.
The possible labelings of these points form a subset of $\{0,1\}^m$, called the effective hypothesis class, which is 
\[\hat H = \{h(X)|h\in \cH\}\,,\]
where $h(X) = (h(x_1),\ldots, h(x_m))$ is the labeling of $X$ by $h$. Note that $|\hat H|\leq 2^m$
and $|\hat H|= \cO(m^d)$ if the VC dimension of $\cH$ is $d$.

In this work, we focus on the Bayesian setting~\cite{dasgupta2004analysis}, where
the target hypothesis is chosen in advance 
from some prior distribution $\pi$ over $\hat H$. Namely, without any additional information, for any labeling $h\in \hat H$, the probability that $h$ is the correct labeling of $X$ is $\pi(h)$. Since we can eliminate any hypothesis $h$ with $\pi(h)=0$ before starting to query for labels, we assume w.l.o.g. that $\pi(h)>0$ for all effective hypotheses in $\hat H$.

We remark that assuming that the unlabeled data $X$ is collected from some distribution $D_x$, which is essentially a distribution $D$  projected onto its input space, and that this distribution $D_x$ can be accurately classified by a hypothesis in $H$ with VC dimension $d$, standard generalization guarantees apply when the prior $\pi$ over $H$ is uniform (see \cite{dasgupta2004analysis} for more details).

\textbf{Standard active learning model} In the standard pool-based active learning setting, 
a single agent owns the pool of unlabeled data $X$. 
The agent, who knows both $\hat H$ and $\pi$, can query the labels of points in $X$, and her goal is to recover the labeling of $X$ (or to find the target hypothesis) by querying as few points as possible. 

A \textit{standard query algorithm} receives as input the prior distribution $\pi$ and unlabeled data set, $X$. In each iteration  $t=1,2,\ldots$, given the history up to time $t$, $$\cF_t =((x_1,y_1),\ldots,(x_{t-1},y_{t-1}))\in (X \times \{0,1\})^{t-1},$$ it selects a point $x_t$ to query and observes its label, $y_t$.  
The algorithm stops when all the labels of $X$ are recovered. Alternatively, the algorithm stops when for every two hypotheses $h_1,h_2\in \hat H$ consistent with $\cF_t$ (meaning that $h_1(x_\tau) = h_2(x_\tau) = y_\tau$ for all $\tau=1,\ldots, t-1$), $h_1(X)=h_2(X)$.

\textbf{Collaborative active learning model.}
In the collaborative setting, we assume there is more than one agent.  
Formally, there are $k$ agents and each agent $i$ has an individual unlabeled data set $X_i$ such that they together compose the pool, i.e., $\cup_{i\in [k]}X_i = X$, and each can query points from their own set $X_i$ (but cannot query points which are not in their set).
The goal of each agent is to recover the true labeling of their own set while performing as few queries as possible. 
The collaboration protocol, also called principal, who knows $\xset$, $\hat{H}$ and $\pi$, decides which point should be queried at each iteration, and her goal is to recover all the labels of $X$ using as few queries as possible. 
We remark that since data points belong to agents, queries of any point $x\in X$ can only be performed by agents whose data set contains $x$.

The query algorithm in the collaborative setting is similar to that in the standard setting, except that the algorithm needs to coordinate among the agents and decide which agent will query each point as some data points might belong to more than one agent.
In this setting, agents can decide to join the collaboration or learn individually at the beginning of the learning. But if they join the collaboration, they commit to follow the instructions of the query algorithm.
Therefore, given a prior distribution $\pi$ over $\hat H$ and a set of agents who would join the collaboration, w.l.o.g. denoted as $\{X_1,\ldots,X_\kappa\}$ for some $\kappa\in [k]$, at time $t=1,2,\ldots$, a \textit{collaborative query algorithm} asks agent $i_t\in [\kappa]$ such that $x_t\in X_{i_t}$ to query point $x_t$, and 
observes its label, $y_t$; the algorithm stops when the labels of points in $\cup_{i\in [\kappa]}X_i$ are completely recovered.

It is straightforward to check that standard query algorithms are a special case of collaborative query algorithms when there is a single agent, i.e., $k=1$. Additionally, a standard algorithm can also be run over multiple agents by considering the union of their data $\cup_{i\in [\kappa]} X_i$ as a single agent.
Hence, we omit ``standard'' or ``collaborative'' in a query algorithm when it is clear from the context how many agents are involved.

For any collaborative algorithm $\cA$, given an input $\pi$ and any collection of unlabeled data sets $X_1,\ldots,X_\kappa\subseteq X$ of size $\kappa\geq 1$, we denote by $Q(\cA,\pi,\{X_1,\ldots,X_\kappa\}, h)$ the label complexity (number of label queries) of $\cA(\pi,\{X_1,\ldots,X_\kappa\})$ when the target hypothesis is $h$. 
For randomized algorithms, the label complexity is taken expectation over the randomness of the algorithm. 
We define the label complexity as follows.
\begin{definition}[Label complexity] 
Given any fixed unlabeled pool and effective hypothesis class $(X,\hat H)$, for any algorithm $\cA$, prior distribution $\pi$ over $\hat H$ and any collection of unlabeled data sets $X_1,\ldots,X_\kappa\subseteq X$ of size $\kappa\geq 1$, 
the label complexity of $\cA$ with $(\pi, \{X_1,\ldots,X_\kappa\})$ as input, denoted by $Q(\cA,\pi, \{X_1,\ldots,X_\kappa\})$, is the expected number of label queries when $h$ is drawn from the prior $\pi$, i.e.,
    \[Q(\cA, \pi, \{X_1,\ldots,X_\kappa\}) = \EEs{h\sim \pi}{Q(\cA, h, \{X_1,\ldots,X_\kappa\})}\,.\]
    For each agent $i\in [\kappa]$ in the collaboration, we let $Q_i(\cA, \pi, \{X_1,\ldots,X_\kappa\})$ denote the expected number of queries performed by agent $i$.
\end{definition}
For any $(\pi, \{X_1,\ldots,X_\kappa\})$, let $Q^*(\pi, \{X_1,\ldots,X_\kappa\})=\min_{\cA} Q(\cA,\pi, \{X_1,\ldots,X_\kappa\})$ denote the optimal query complexity.
An algorithm $\cA$ is said to be \textit{optimal} if $Q(\cA, \pi,\{X_1,\ldots,X_\kappa\}) = Q^*(\pi, \{X_1,\ldots,X_\kappa\})$ for any prior distribution $\pi$ and $X_1,\ldots,X_\kappa$.

\textbf{Rational agents}
We assume that agents have access to a baseline algorithm and are able to run it on their own local data. 
Agents can decide to join the collaboration or run the baseline individually at the beginning of the learning.
If they join the collaboration, they commit to follow the instructions of the query algorithm. 
Each agent is incentivized to join the collaboration if she could perform fewer label queries (assuming that all others join the collaboration) by pulling out and running the baseline $\cA$ individually. 
Formally, 
\begin{definition}[Individual rationality]
    In a collaborative learning problem with prior distribution $\pi$ and $k$ agents $\xset$, given a baseline algorithm $\cA$,
    a collaborative algorithm $\cA'$ is individually rational (IR) if 
\begin{equation}
    {Q_i(\cA',\pi, \xset)} \leq Q(\cA, \pi, \{X_i\}), \forall i\in [k]\,.\label{eq:def-incentive}
\end{equation}
\end{definition}
We say 
 $\cA'$ is \textit{strictly individually rational} (henceforth, SIR) if
\begin{equation*}
    {Q_i(\cA',\pi, \xset)} < Q(\cA, \pi, \{X_i\}), \forall i\in [k]\,.
\end{equation*}
We remark that in addition to their own sets, each agent knows all the unlabeled data sets,  $\xset$, and the prior distribution $\pi$, otherwise, they will not be able to compute $Q_i(\cA',\pi,\xset)$. The principal also has access to $\xset$ and $\pi$, and can therefore make sure these constraints are satisfied.

It is worth noting that our model can also accommodate different individual baseline algorithms. We briefly discuss this in Section~\ref{sec:discussion}.

An alternative interpretation of the problem in a game theoretic framework is as follows: each agent has a strategy space of two strategies, joining the collaboration and not.
The utility of an agent that performs $Q$ queries is $-Q$.
If the algorithm $\cA$ is IR, then the case of all agents joining the collaboration is a Nash equilibrium (since switching to not joining will not increase their utility).
If $\cA$ is SIR, then all agents joining the collaboration is a strict Nash equilibrium.

\section{Construction of IR Collaborative Algorithms}\label{sec:ir}
When agents are limited to a poor baseline algorithm, e.g., randomly selecting points to query, the principal can simply incentivize agents to collaborate by using a superior algorithm that requires fewer labeling efforts. 
We therefore start by considering optimal baseline algorithms in Section~\ref{subsec:ir-opt}.
If we are able to find an IR collaborative algorithm for an optimal baseline algorithm, $\OPT$, then it must be IR w.r.t. all baseline algorithms.
We demonstrate that, surprisingly, the optimal algorithm $\OPT$ is IR given that the baseline algorithm is $\OPT$ itself.
Since computing an optimal algorithm is known to be NP-hard, we continue by considering the best-known approximation algorithm, the greedy algorithm. 
In Section~\ref{subsec:ir-greedy}, we show that given the greedy algorithm as baseline, the collaboration protocol that runs the greedy algorithm is not IR. 
Then in Section~\ref{subsec:ir-alg}, 
we provide a general scheme that transforms any baseline algorithm into an IR algorithm while maintaining a comparable label complexity.
\subsection{Optimality Implies Universal Individual Rationality}\label{subsec:ir-opt}
Incorporating individual rationality as an additional constraint to optimality usually requires additional effort in certain settings, e.g., in online learning by \cite{blum2020advancing}. 
However, in our specific setting, optimality does not contradict the individual rationality property. That is, an optimal algorithm will not increase any agent's label complexity to benefit other agents. 
In fact, optimizing for optimality implies achieving individual rationality for all baseline algorithms. 
\begin{theorem}\label{thm:opt-ir}
For any optimal collaborative algorithm $\OPT$, we have $${Q_i(\OPT,\pi, \xset)} \leq Q(\OPT, \pi, \{X_i\})= Q^*(\pi, \{X_i\}), \forall i\in [k].$$
    Therefore, $\OPT$ is IR w.r.t. any baseline algorithm.
\end{theorem}

We prove the theorem by contradiction. If $\OPT$ is not IR for the baseline being $\OPT$, then there exists an agent $i$ such that $Q(\OPT, \pi, \{X_i\}) < Q_i(\OPT,\pi, \xset)$. In this case, we can construct a new algorithm by first running $\OPT$ over $\{X_i\}$ (to recover the labels of $X_i$)  and then running $\OPT(\pi, \xset\})$ and replacing agent $i$'s queries with the recovered the labels of $X_i$. This new algorithm incurs a strictly smaller label complexity than $\OPT$, which is a contradiction to the optimality of $\OPT$. The formal proof is deferred to Appendix~\ref{app:opt-ir}. 
Unfortunately, computing an optimal query algorithm is not just NP-hard, but also hard to approximate within
a factor of $\Omega(\log(|\hat H|))$~\cite{golovin2010near,chakaravarthy2007decision}. 
One of the most popular heuristics to find an approximated solution is greedy.
\subsection{The Greedy Algorithm is Not Individually Rational}\label{subsec:ir-greedy}
\vspace{-0.5em}

For standard Bayesian active learning, \cite{kosaraju2002optimal,dasgupta2004analysis} presented a simple greedy algorithm called generalized binary search (GBS), which chooses a point leading to the most balanced partition of the set of hypotheses consistent with the history. 
More specifically, at time step $t$, given the history $\cF_t =((x_1, i_1,y_1),\ldots,(x_{t-1},i_{t-1},y_{t-1}))$, let $\VS(\cF_t) = \{h\in \hat H|h(x_\tau) = y_\tau,\forall \tau\in [t-1]\}$ denote the set of hypotheses consistent with the history $\cF_t$ (often called the version space associated with $\cF_t$).
Given $\cF_t$ and $(\pi, \{X_1,\ldots,X_\kappa\})$ as input, GBS will query 
\[x_t = \argmax_{x\in \cup_{i\in [\kappa]}X_i} \min(\pi(\{h\in \VS(\cF_t)|h(x)=1\}), \pi(\{h\in \VS(\cF_t)|h(x)=0\}))\,\]
at time $t$. When referring to GBS as a collaborative algorithm, we complement it with an arbitrary tie-breaking rule for selecting $i_t$, as GBS itself does not specify how to choose which agent to query.
GBS is guaranteed to achieve competitive label complexity with the optimal label complexity.

\begin{lemma}[Optimality of GBS, Theorem~3 of \cite{dasgupta2004analysis}]\label{lemma:gbs}
    For any prior distribution $\pi$ over $\hat H$ and $k$ agents $\xset$, the label complexity of GBS satisfies that
    \[Q(\gbs,\pi,\xset) \leq 4Q^*(\pi,\xset) \ln(\frac{1}{\min_{h\in \hat H} \pi(h)})\,.\]
\end{lemma}
The greedy algorithm GBS not only achieves approximately optimal label complexity, but it is also computationally efficient, with a running time of $\cO(m^2|\hat H|)$. 
As GBS is the best-known efficient approximation algorithm, it is natural to think that agents would adopt GBS as a baseline. 

As we have shown that the optimal algorithm is IR w.r.t. itself, the next natural question is: \textit{Is GBS (as collaboration protocol) individually rational w.r.t. GBS itself? }

We answer this question negatively, even in the case of two agents.
Even worse, we present an example in which an agent's label complexity is $\Omega(n)$ when participating in the collaboration, but only $\cO(1)$ when not participating.

\begin{restatable}{theorem}{greedyir}\label{thm:greedyir}
For the algorithm of GBS, there exists an instance of $(X_1,X_2, \pi)$ with $|X_1|=n$, in which agent $1$ incurs a label complexity of  $Q_1(\gbs,\pi, \{X_1,X_2\}) = \Omega(n)$ when participating the collaboration and can achieve $Q(\gbs,\pi, \{X_1\}) = \cO(1)$ when not participating. %
\end{restatable}
Intuitively, at each time step, GBS only searches for an $x_t$ which leads to the most balanced partition of the version space, which does not necessarily lead to the optimal point to query. Given additional label information from the other agent, GBS possibly choose a worse point to query.
In addition, the label complexity of GBS is upper bounded by the optimal label complexity multiplied by a logarithmic factor.  It is possible that the agent achieves a smaller multiplicative factor by running GBS individually and a larger factor in the collaboration.
To prove the theorem, we construct an instance in which there exists a hypothesis with a prior probability of $1/4$, such that if GBS runs on $\{X_1,X_2\}$ and this hypothesis is the target, GBS will query almost all the points in $X_1$ (in a particular order) before returning this hypothesis. We show this part by induction. Additionally, we compute the query tree by running GBS solely on $X_1$ and use it to show that in this case, GBS has an expected query complexity of $O(1)$.
The full construction of the instance and the proof of Theorem~\ref{thm:greedyir} is deferred to Appendix~\ref{app:greedyir}.

\subsection{A Scheme of Converting Algorithms to IR Algorithms}\label{subsec:ir-alg}
\vspace{-0.5em}

Given that the greedy algorithm has been proven to be not individually rational w.r.t. itself, we raise the following question: \emph{Is it possible to develop a general scheme that can generate an IR algorithm given any baseline algorithm?} In this section, we propose such a scheme that addresses this question. 
Moreover, given a baseline algorithm $\cA$, the resulting IR algorithm can achieve a label complexity comparable to implementing the baseline algorithm over all agents, i.e., $\cA(\pi, \xset)$.
It is important to note that we aim for the label complexity to be comparable to $Q(\cA,\pi, \xset)$ rather than $\sum_{i\in [k]} Q(\cA,\pi,{X_i})$, as the latter holds true by individual rationality.
Given an efficient approximately optimal algorithm as baseline (e.g., GBS), our scheme can provide an algorithm that simultaneously exhibits individual rationality, efficiency, and approximately optimal label complexity. 

For any baseline algorithm $\cA$, we define a new algorithm $\ir(\cA)$, which runs $\cA$ as a subroutine.
Basically, we first calculate the label complexity of agent $i$ both when she is in collaboration with all the other agents, i.e., $Q_i(\cA,\pi,\xset)$,  and when she is not in collaboration, i.e., $Q(\cA,\pi,\{X_i\})$, for all $i\in [k]$. 
By doing so, we can distinguish which agents can benefit from collaboration when running $\cA$ and which cannot. 
We denote the set of agents who cannot benefit from collaboration with all others when running $\cA$ as $S = \{i|Q_i(\cA, \pi, \xset)>Q(\cA,\pi,\{X_i\})\}$. 
For those who do not benefit from the collaboration, we just run $\cA$ on their own data.
For those who benefit from collaborating  with the others together, we run $\cA$ over all agents $[k]$-- 
Only whenever $\cA(\pi,\xset)$ asks to query the label of a point belonging to some $i\in S$, since we already recovered the labels of $X_i$, we just  feed $\cA(\pi,\xset)$ with this label without actually asking agent $i$ to query.  
The detailed algorithm is described in Algorithm \ref{alg:basic2ir}.

\begin{algorithm}[H]\caption{$\ir$}\label{alg:basic2ir}
    \begin{algorithmic}[1]
    \STATE \textbf{input:} A query algorithm $\cA$, set $\xset$ and prior $\pi$ over $\hat H$
    \STATE For each $i\in[k]$, calculate $Q_i(\cA,\pi,\xset)$  and $Q(\cA, \pi, \{X_i\})$.
    \STATE Let $S \leftarrow \{i|Q_i(\cA, \pi, \xset)>Q(\cA,\pi,\{X_i\})\}$ and $X_S\leftarrow \cup_{i\in S}X_i$ // the agents who do not benefit from collaboration
    \STATE \textbf{for} {each $i\in S$} \textbf{do} $Y_i\leftarrow$ Run $\cA$ over $\{X_i\}$ // recover the labels for agent $i$
    \FOR{$t=1,\ldots$}
        \STATE $(i_t,x_t)\leftarrow$ the querying agent and the query point from $\cA(\pi, \xset)$
        \STATE \textbf{if} {$i_t\in S$} \textbf{then} Feed the label of $x_t$ from $Y_{i_t}$// we already recovered the labels of $ X_{S}$
        \STATE \textbf{else}
         Ask  agent $i_t$
        to  query the label of $x_t$         
    \ENDFOR
    \end{algorithmic}
\end{algorithm}

\begin{theorem}\label{thm:ir}
    For any baseline algorithm $\cA$, the algorithm $\ir(\cA)$ satisfies the following properties:
    \begin{itemize}[topsep=0pt,itemsep =-0.5ex,  leftmargin = 6mm]
        \item \textbf{IR property:} $\ir(\cA)$ is individually rational w.r.t. the baseline algorithm $\cA$.
        \item \textbf{Efficiency:} $\ir(\cA)$ runs in 
        $\cO(k T_{\cA, Q} + m T_{\cA,0})$ time, where $T_{\cA,0}$ is the time of computing $(i_t,x_t)$ at each time $t$ for $\cA$ and $T_{\cA, Q}$ is the maximum time of computing $Q_i(\cA,\pi,\{X_1,\ldots,X_\kappa\})$ for an  agent $i$, unlabeled data $\{X_1,\ldots,X_\kappa\}$, and algorithm $\cA$.
        \item \textbf{Label complexity:} $Q(\ir(\cA), \pi,\xset)\leq Q(\cA, \pi, \xset)$.
    \end{itemize} 
\end{theorem}

The proof follows the algorithm description immediately.
Note that when the baseline is GBS, we have $T_{\gbs,0} = \cO(m)$. We can compute $Q_i(\cA,\pi,\{X_1,\ldots,X_\kappa\})$ by simulating over all effective hypotheses $h\in \hat H$. For each $h$, we will query at most $m$ rounds.
Therefore, we have $T_{\gbs,Q} = \cO(m^2|\hat H|)$ and we can run $\ir(\gbs)$ in $\cO(k m^2 |\hat H|)$ time. 
Using GBS as the baseline, we derive the following corollary.
    
\begin{corollary}\label{crl:ir-greedy}
Given GBS as the baseline, $\ir(\gbs)$ is IR; runs in $\cO(k m^2 |\hat H|)$ time; and satisfies that $Q(\ir(\gbs), \pi, \xset)\leq 4Q^*(\pi,\xset) \ln(\frac{1}{\min_{h\in \hat H} \pi(h)})$.
\end{corollary}

\section{Converting  Algorithms to SIR Algorithms}\label{sec:sir}
In Section~\ref{sec:ir},
we provided a generic scheme for constructing an IR algorithm given any baseline algorithm.
In this section, we focus on constructing SIR algorithms given IR algorithms.
Since strict individual rationality requires that agents strictly benefit from collaboration, this is impossible without further assumptions. 
For example, consider a set of agents who only have one single independent point in their own sets 
and a prior distribution that is uniform over all labelings. 
In this case, each agent, regardless of whether she collaborates or not, has a label complexity of $1$ and cannot \emph{strictly} benefit from collaboration as the other agents cannot obtain information about her data.

Now, let us consider a notion weaker than SIR, called $i$-partially SIR, in which only agent $i$ strictly benefits from the collaboration, and any other agent $j\neq i$ does not get worse by joining the collaboration.
More formally, 
\begin{definition}[Partially SIR algorithms]
    For any baseline algorithm $\cA$, for all $i\in [k]$, an algorithm $\cO_i$ is $i$-partially SIR, if $\cO_i$ satisfies that
    \[Q_i(\cO_i, \pi, \xset) < Q(\cA,\pi, \{X_i\})\,,\]
    and 
    \[Q_{j}(\cO_i,\pi,\xset)\leq Q(\cA,\pi,\{X_j\}), \forall j\in [k]\setminus \{i\}\,.\]
\end{definition}
If we are given an $i$-partially SIR algorithm $\cO_i$ for each $i$, then we can construct a SIR algorithm by running a mixture of an IR algorithm $\cA'$ (e.g., $\ir(\cA)$ in Algorithm~\ref{alg:basic2ir}) and $\{\cO_i|i\in [k]\}$ with the label complexity a little (arbitrarily small) higher than that of $\cA'$.
\begin{lemma}\label{lmm:partsir}
    For any baseline algorithm $\cA$, given an IR algorithm $\cA'$ and partially SIR algorithms $\{\cO_i|i\in [k]\}$, for any $\epsilon>0$, let $\cA''_\epsilon$ be the algorithm of running $\cA'$ with probability $(1-\frac{\epsilon}{n})$ and running $\cO_i$ with probability $\frac{\epsilon}{kn}$.
    Then $\cA''_\epsilon$ satisfies the following properties.
    \begin{itemize}[topsep=0pt,itemsep =-0.5ex,  leftmargin = 6mm]
        \item SIR property: $\cA''_\epsilon$ is SIR with respect to the baseline algorithm $\cA$.
        \item Label complexity: $Q(\cA''_\epsilon, \pi, \xset)\leq Q(\cA', \pi, \xset) +\epsilon$.
    \end{itemize}
\end{lemma}
The proof is straightforward from the definition, and we include it in Appendix~\ref{app:part-sir} for completeness.
Since a SIR algorithm is also $i$-partially SIR for all $i\in [k]$, constructing a SIR algorithm is equivalent to constructing a set of partially SIR algorithms $\{\cO_i|i\in [k]\}$.
Therefore, the problem of constructing a SIR algorithm is reduced to constructing partially SIR algorithms $\{\cO_i|i\in [k]\}$. 

For the remainder of this section, we will present the SIR results for an optimal baseline algorithm in Section~\ref{subsec:sir-opt}, where we propose a sufficient and necessary assumption for the existence of SIR algorithms and then provide a SIR algorithm.
This algorithm is SIR w.r.t. any baseline algorithm but again, computationally inefficient.
In Section~\ref{subsec:sir-apprx}, we provide a general scheme that transforms any baseline algorithm into a SIR algorithm.

\subsection{A Universal SIR Algorithm for Any Baseline Algorithm}\label{subsec:sir-opt}
\vspace{-0.5em}
Constructing a universal SIR algorithm w.r.t. any baseline is equivalent to constructing a SIR algorithm for an optimal baseline.
For the existence of SIR algorithms given any optimal baseline algorithm, we propose the following assumption, which is sufficient and necessary. We include the proof for the necessity of this assumption in Appendix~\ref{app:nece-asp}. The sufficiency of this assumption will be verified immediately after we construct a SIR algorithm.
\begin{assumption}\label{asp:clb-help}
We assume that for any $i\in [k]$, the optimal label complexity of agent $i$ given the information regarding the labels of all other agents is strictly smaller than that without this additional information, i.e.,
    $Q^*(\pi, \{X_i\})- \EEs{h\sim \pi}{Q^*(\pi_{h,-i}, \{X_i\})} >0\,,$
    where $\pi_{h,-i}$ is the posterior distribution of $\pi$ after observing $\{(x,h(x))|x\in \cup_{j\neq i} X_{j}\}$. 
\end{assumption}

According to Lemma~\ref{lmm:partsir}, we can construct a SIR algorithm by constructing a set of partially SIR algorithms $\{\cO_i|i\in [k]\}$. 

Let $\cO_i$ be the algorithm of running an optimal algorithm $\OPT$ over $(\pi, \{X_j|j\neq i\})$ first, then given the query-label history of $\{(x, h(x))|x\in \cup_{j\neq i} X_j\}$ for some $h\in \hat H$, run $\OPT$ over $(\pi_{h,-i}, \{X_i\})$. Then it immediately follows that $\cO_i$ is $i$-partially SIR from Assumption~\ref{asp:clb-help}. Let $\OPT_\epsilon''$ denote the algorithm of of running $\OPT$ with probability $(1-\frac{\epsilon}{n})$ and running $\cO_i$ with probability $\frac{\epsilon}{kn}$ for all $i\in [k]$. By Lemma~\ref{lmm:partsir}, we have

\begin{corollary}
    Under Assumption~\ref{asp:clb-help}, for any $\epsilon>0$, $\OPT_\epsilon''$ is SIR w.r.t. $\OPT$ and satisfies $$Q(\OPT''_\epsilon, \pi, \xset)\leq Q^*(\pi, \xset) +\epsilon.$$
    In addition, $\OPT_\epsilon''$ is SIR w.r.t. any baseline algorithm $\cA$ as $$Q_i(\OPT''_\epsilon, \pi, \xset)< Q^*(\pi, \{X_i\}) \leq Q(\cA,\pi,\{X_i\}).$$
\end{corollary}

\subsection{A Scheme of Converting Algorithms to SIR Algorithms
}\label{subsec:sir-apprx}
\vspace{-0.5em}
As mentioned before, computing an optimal algorithm is NP-hard. Assumption~\ref{asp:clb-help} assumes that collaboration can strictly benefit agents when the collaboration protocol can compute the optimal algorithm given $\pi_{h,-i}$.
Hence, the assumption does not take the computational issue into consideration and thus might not be enough for the existence of an efficient SIR algorithm w.r.t. an efficient approximation algorithm like GBS.

Instead, we propose prior-independent assumption that is sufficient for the existence of efficient SIR algorithms when we are given an efficient baseline and an efficient IR algorithm w.r.t. the baseline. 
Basically, we assume that, there exists an effective hypothesis $h\in \hat H$, given the information that all other agents are labeled by $h$, the number of labelings of $X_i$ consistent with the label information is strictly smaller than the total number of labelings of $X_i$ by $\hat H$. Formally, for any $i\in [k]$, let $X_{-i} = \cup_{j\neq i} X_j$ denote the union of all agents' data except agent $i$.
Let $H(X_i)=\{h'(X_i)|h'\in \hat H\}$ denote the effective hypothesis class of $X_i$, i.e., all labelings of $X_i$.
For any $h\in \hat H$, let $H(X|h) = \{h'(X_i)|h'(X_{-i}) = h(X_{-i}),h'\in \hat H\}$ denote the subset which are consistent with all other agents being labeled by $h$.

\begin{assumption}\label{asp:clb-help-efficient}
    For all $i\in [k]$, there exists an $h\in \hat H$ s.t.  
    the number of labelings of $X_i$ consistent with $(X_{-i}, h(X_{-i}))$ is strictly smaller than the number of labelings by $\hat H$, i.e., $\abs{H(X_i|h)} < \abs{H(X_i)}$.
\end{assumption}
Intuitively, Assumption~\ref{asp:clb-help-efficient} means that for every agent $i$, there exists an hypothesis  $h$ such that when $h^*=h$, the cardinality of the set of hypotheses consistent with $(X_{-i}, h(X_{-i}))$ is strictly smaller than $|\hat H|$. We will show that this assumption is sufficient for the \textit{existence} of algorithms satisfying SIR property. Without it, it is unclear if there exist SIR algorithms. 
The assumption can be easily verified by iterating each $h\in \hat H$ (this is polynomial in  $|\hat H|$ and $m$). 

Notice that each deterministic query algorithm $\cA$ can be represented as a binary tree, $\cT_\cA$ whose internal nodes at level $t$ are queries (``what is the $x_t$'s label?''), and whose leaves are labelings as illustrated in Figure~\ref{fig:query-tree}.
Under Assumption~\ref{asp:clb-help-efficient}, we can prune the query tree of $\cA(\pi, \{X_i\})$ by removing
all subtrees whose leaves are all in $H(X_i)\setminus H(X_i|h)$.
We do not need to construct this pruned tree when we implement the algorithm. At time $t$, we just need to generate an $x_t$ from $\cA(\pi, \{X_i\})$, then check if this node should be pruned by checking if all the hypotheses $H(X_i|h)$ agree on the label of $x_t$. If this is true, it means that we have already recovered the label of $x_t$ and thus we just need to feed the label to the algorithm without actually querying $x_t$ again.
Then we can construct a $i$-partially SIR algorithm $\cO_i$ by running $\ir(\cA)$ over $(\pi, \{X_j|j\neq i\})$ to recover the labeling of $X_{-i}$ first, then running pruned version of $\cA(\pi, \{X_i\})$. Note that the implementation also works when $\cA$ is randomized.

Consider Example~\ref{ex:treshold}, where the hypothesis class $\mathcal H = \{x\geq \alpha|\alpha=0.2,0.4,0.5,0.6,0.8\}$ with a uniform prior, agent $1$ with points $X_1=\{0.25,0.5,0.75\}$ and agent $2$ with points $X_2=\{0.3,0.45, 0.55, 0.7\}$. 
When agent $1$ runs a binary search, the query tree has $0.5$ as a root, then if $0.25$ if $h^*(0.5)=1$, and $0.75$  otherwise. Now, algorithm $\cO_1$ runs a binary search on $X_2$ and obtains all the labels of the points in $X_2$. The hypothesis $h = 1(x\geq 0.5)$ holds  $|H(X_1|h)| = 1 < |\hat H(X_1)|=4$. 
When $h$ is the labeling function, $0.3$ and $0.45$ are labeled as negative, and $0.55$ and $0.7$ are labeled as positive. Then, $\cO_1$ will only need agent $1$ to query $0.5$ as the labels of $0.25$ and $0.75$ can be inferred and they are pruned in the query tree. 

\begin{lemma}\label{lemma:OiIsI-Part-SIR}
Under Assumption~\ref{asp:clb-help-efficient}, the algorithm $\cO_i$ constructed above is $i$-partially SIR and runs in time $\cO(\cT_{\ir(\cA)} + m (|\hat H| + T_{\cA,0}))$ time, where $\cT_{\ir(\cA)}$ is the running time of $\ir(\cA)$ and $T_{\cA,0}$ is the time of computing $(i_t,x_t)$ at each time $t$ for $\cA$.

\end{lemma}
The proof of Lemma~\ref{lemma:OiIsI-Part-SIR} is deferred to Appendix~\ref{app_lmm3}.

We can then construct an algorithm $\cA_\epsilon''$ by running $\ir(\cA)$ with probability with probability $(1-\frac{\epsilon}{n})$ and running $\cO_i$ constructed in the above way with probability $\frac{\epsilon}{kn}$ for all $i\in [k]$. By combining Lemmas~\ref{lmm:partsir} and \ref{lemma:OiIsI-Part-SIR}, we derive the following theorem. Then, combining it with Corollary~\ref{crl:ir-greedy}, we derive a SIR algorithm for GBS as baseline GBS.
\begin{theorem}
Under Assumption~\ref{asp:clb-help-efficient}, for any baseline algorithm $\cA$, for any $\epsilon>0$, $\cA_\epsilon''$ is SIR and satisfies $$Q(\cA''_\epsilon, \pi, \xset)\leq Q(\ir(\cA),\pi, \xset) +\epsilon.$$
In addition, Algorithm $\cA_\epsilon''$ runs in $\cO(\cT_{\ir(\cA)} + m (|\hat H| + T_{\cA,0}))$ time.
\end{theorem}

\begin{corollary}
Given GBS as the baseline, Algorithm $\gbs_\epsilon''$ is SIR; runs in $\cO(k m^2 |\hat H|)$ time; and satisfies that $$Q(\gbs_\epsilon'', \pi, \xset)\leq 4Q^*(\pi, \xset) \ln\left(\frac{1}{\min_{h\in \cH}\pi(h)}\right) +\epsilon.$$
\end{corollary}

\section{Discussion}\label{sec:discussion}
In this paper, we have initiated the study of collaboration in active learning in the presence of incentivized agents. We first show that an optimal collaborative algorithm is IR w.r.t. any baseline algorithm while approximate algorithms are not.
Then we provide meta-algorithms capable of producing IR/SIR algorithms given any baseline algorithm as input.

Our model and algorithms can also allow different agents to have different baseline algorithms---Whenever the principal plans to run an algorithm on a union of datasets, she can simply check which baseline algorithm has the lowest expected query complexity on this union, and run it. When she needs to run an algorithm on a dataset of an individual, she can simply run their baseline. This way the (S)IR is preserved.
 
There are a few problems we leave open. First, relaxing the assumption that each agent $i$ has full knowledge of $X_{-i}$ (e.g., due to privacy concerns). Second, relaxing the assumption that agents provide reliable labels. 
Third, deriving results in non-realizable settings. 
Realizability is a standard assumption in learning theory at large and particularly within active learning as highlighted in the classical machine learning theory textbook by~\cite{Shalev-Shwartz_Ben-David_2014} and the active learning theory survey by~\cite{Hanneke14survey}. 
Moreover, realizability has also been adopted in the collaborative learning setting (e.g.,~\cite{Blum17}). The reason is that without realizability, additional complications might arise in collaboration such as why collaboration would yield benefits. In general, we believe that additional assumptions would be required to relax this assumption. Forth, towards a more game theory orientation, it would be interesting to design collaborative algorithms in a setting where agents can form coalitions. 
Finally, finding a necessary and sufficient assumption(s) for the existence of efficient SIR algorithms will be an interesting direction (and we have found a sufficient one in this work).

\chapter{Learning within Games}\label{chap:games}

\section{Introduction}
Many mechanism design settings can be cast as \emph{Principal/Agent Problems}. These are Stackelberg games of incomplete information, in which the \emph{Principal} first commits to some policy, and then the \emph{Agent} chooses an action by best responding. The utility for both the Principal and the Agent can depend on the actions they each choose, as well as some underlying and unknown state of nature. The fact that the state of nature is unknown is a crucial modeling aspect of Principal Agent problems. Two canonical examples of Principal Agent problems will be instructive: a simple example of a contract theory problem (see e.g. \cite{carroll2021contract}) and of a Bayesian Persuasion Problem \citep{kamenica2011bayesian}.
\begin{enumerate}
    \item \textbf{Contract Theory}: Consider a Principal (say a university endowment office) that has capital that they would like to invest, but who does not themselves have the expertise to invest it effectively. Instead, they would like to contract with an Agent (say a hedge fund) so as to maximize their returns. The Agent will choose a strategy (say by dividing funds across a particular portfolio of investments), but the return of the strategy will be unknown at the time that they choose it---it depends on the unknown-at-the-time-of-action returns of each investment. Moreover, they may be able to choose a better strategy by investing more time, effort, and money (for example, by hiring talented fund managers away from competing hedge funds). But should they? It is in the Principal's interest that their returns (minus their fees) should be maximized, but it is in the Agent's interest that their fees (minus their costs) should be maximized. How should the Principal design the contract (i.e. a mapping from outcomes to payments to the Agent) so that their utility is maximized when the Agent best responds? 
    \item \textbf{Bayesian Persuasion}: Consider a Principal (say a pharmaceutical company) that manufactures drugs that they need to get approved by an Agent (say a regulatory agency like the FDA) before they can be sold. The drugs will have various properties which we can think of as an underlying state comprising effectiveness, safety, etc. that are initially unknown. But drug trials (that may be at least partially designed by the Principal) will be run that will provide a noisy signal about the qualities of the drug, that the Agent will use to form a belief about the state, and as a result, either approve the drug or not. It is in the Principal's interest that as many drugs as possible should be approved --- but the Agent will approve only those drugs that it believes are safe. How should the Principal design the drug trial (i.e. a stochastic mapping from state to observable signal) so that as many drugs as possible are approved when the Agent best responds? 
\end{enumerate}
The classical economic literature answers these questions in a conceptually straightforward manner: The Principal should commit to a strategy such that her payoff will be maximized after the Agent best responds. But given that the state is unknown, how will the Agent choose to best respond, and how will the Principal anticipate the Agent's choice? The classical answer is that the Principal and the Agent share a common \emph{prior distribution} on the unknown state of the world: the Agent best-responds so as to maximize his utility in expectation over this Prior, and the Principal, also being in possession of the same beliefs, anticipates this. There are some assumptions that are traditionally made about tie-breaking (that it is done in favor of the Principal) that we will interrogate, but the reader can ignore these for now. A strong general critique of the foundations of this literature asks: where does this prior belief come from, and why is it reasonable to assume it is shared? 

Recently, \cite{camara2020mechanisms} gave an elegant framework for addressing this critique head on. They study a repeated Principal Agent problem (where two long-lived parties interact with each other repeatedly) and dispense with the common prior assumption entirely. In fact, there are no distributional assumptions at all in their model: the sequence of realized states of nature can be arbitrary or even adversarially chosen. Instead, it is assumed that the Agent behaves in a way that is consistent with various efficiently obtainable online-learning desiderata, which are elaborations on the goal that they should have no \emph{swap regret} \citep{blum2007external}, and that they don't have too much ``additional information'' about the state sequence compared to the Principal (this can be formalized in various ways that we shall discuss). These are assumptions that would be satisfied were there a common prior that both Agents were optimizing under --- but can be reasonably assumed (because they can be efficiently algorithmically obtained) without this assumption. Under a collection of such behavioral assumptions --- and other assumptions on the structure of the game --- \cite{camara2020mechanisms} show that a Principal who maintains \emph{calibrated forecasts} for the unknown states of nature, and acts by treating these forecasts as if they were a common prior --- is able to guarantee themselves a strong form of \emph{policy regret}. That is, they are guaranteed to obtain utility nearly as high as they would have had they instead played any fixed policy in some benchmark class, \emph{even accounting for how the Agent would have acted under this counter-factual policy}. Moreover, it has been known since \cite{foster1998asymptotic} that it is possible to produce calibrated forecasts of an arbitrary finite dimensional state, even if the state sequence is chosen adversarially --- so the mechanism proposed by \cite{camara2020mechanisms} could in principle be implemented in their model. This makes the model of \cite{camara2020mechanisms} a compelling alternative to common prior assumptions. Nevertheless, there remain some difficulties with the mechanism they propose within this framework:
\begin{enumerate}
    \item \textbf{Computational and Statistical Complexity}: Informally speaking, a method of producing forecasts $\hat s \in \mathbb{R}^d$ of a $d$-dimensional state $s \in \mathbb{R}^d$ is \emph{calibrated} if the forecasts are unbiased, not just overall, but conditional on the forecast itself: $\E_{s, \hat s}[s | \hat s] = \hat s$, for all values of $\hat s$. When we are forecasting probability distributions over a finite collection of states of nature $\cY$, the forecasts are probability distributions represented as $|\cY|$-dimensional  vectors $\hat s \in \Delta (\cY)$. Under any reasonable discretization, there are $\Omega\left(2^{|\cY|}\right)$ many such vectors, and algorithms for maintaining calibrated forecasts in this space have both computational and statistical complexity scaling exponentially with $|\cY|$. The mechanism proposed by \cite{camara2020mechanisms} inherits these limitations: and as a result has both running time and regret bounds that suffer \emph{exponential} dependencies on the cardinality of the state space $|\cY|$. Thus these mechanisms are reasonable only for very small constant sized state spaces. 
    \item \textbf{Strong ``Alignment'' Assumptions}: Even in the classical model in which the Agent ``best responds'' to the policy of the Principal, using their prior beliefs over the state of nature, there can be ambiguity in how the Agent will act. In particular, what if their set of best responses is not a singleton set: there are multiple actions that they can take that yield the same utility for the Agent---which action will they take? This is an important detail, because even when the Agent's utilities are tied over this set, each action may yield very different utility for the Principal. The traditional assumption is that the Agent breaks ties in favor of the Principal---which although optimistic can perhaps be viewed as a mild assumption because it concerns only exact ties. However, when there is doubt or imprecision about the Agent's beliefs, this problem is exacerbated: one could assume that \emph{near} ties are broken in favor of the Principal, but it is much less reasonable to assume that the Agent will forgo small gains so as to benefit the Principal; a similar phenomenon arises with the mechanism of  \cite{camara2020mechanisms} because sequential forecasts will never be exactly, but only approximately calibrated. 
   
    \cite{camara2020mechanisms} deal with this issue by suggesting a policy and imposing strong ``alignment'' assumptions (their Assumptions 2 and 7) over this specific proposed policy, which informally require that with respect to all possible prior distributions, the difference in Agent utilities between a pair of actions is comparable to the corresponding change in Principal utilities. This has the effect of making approximate tie-breaking (almost) irrelevant for the Principal.
    However, the extent to which these alignment assumptions hold and whether this policy operates effectively across different ranges or in various scenarios is in general unclear.
    
\end{enumerate}

\subsection{Our Results}

In this paper we revisit the framework of \cite{camara2020mechanisms} and derive new mechanisms which address these issues. Our mechanisms obtain strong policy regret guarantees, but  are exponentially more efficient (in their dependence on the cardinality of the state space) in terms of both their running time and their regret bounds. Moreover, in a subset of instances (which we show includes linear contracting, that has been the exclusive focus of a large fraction of recent computational work in contract theory) our mechanisms entirely eliminate the need for alignment assumptions.

\paragraph{Computational and Statistical Efficiency---Beyond Calibration:} We show how to obtain both policy regret bounds and running time bounds that scale polynomially with the cardinality of the state space $|\cY|$, rather than exponentially (as in \cite{camara2020mechanisms}). To do this, we need to give mechanisms that do not rely on fully calibrated forecasts of the state of nature. Instead, we give mechanisms that use forecasts of the state of nature that are \emph{statistically unbiased} subject only to a polynomial number of events: informally, the events that the forecasts themselves (were they used as a common prior) would lead the Principal to propose each particular policy, and anticipate each particular action in response by the Agent. Calibration requires unbiasedness subject to exponentially many (in the cardinality of the state space $|\cY|$) events; here we require unbiasedness with respect to only quadratically many events (in the cardinality of the action space of the Principal and the Agent). Using a recent algorithm of \cite{NRRX23}, we are able to produce forecasts with these properties with running time that is polynomial in the cardinality of the state space $|\cY|$ and the action spaces of the Agent and the Principal. Under similar behavioral assumptions as \cite{camara2020mechanisms} (which strictly generalize the common prior assumption), we show that our mechanism obtains policy regret bounds that scale linearly with $|\cY|$ (again, compared to exponentially with $|\cY|$ in \cite{camara2020mechanisms}). 

\paragraph{Stable Policy Oracles---Avoiding Alignment Assumptions:} As discussed above, Alignment assumptions (their Assumptions 2 and 7) are needed in \cite{camara2020mechanisms} to address, informally, the problem of the mechanism's proposed policy inducing ``near-ties'' in the Agent's utility that nevertheless lead to very different Principal utility. In contrast, we define a policy to be \emph{stable} with respect to a state distribution $\pi$ if when compared to the Agent's best response to the policy under $\pi$, every other action \emph{either} leads to substantially lower utility for the Agent (in expectation over the distribution), or else leads to nearly the same utility for the Principal. We show that if our mechanism has the ability to construct stable policies that also lead to near optimal utility for the Principal under the Principal's current state forecast, then she can obtain strong policy regret bounds without the need for an Alignment assumption. We then turn to the task of constructing near-optimal stable policies. We show by example that this is not possible for all Principal-Agent games within the framework we consider; but show how to do it in two important special cases. The first is the \emph{linear contracting} setting---the special case of contract theory in which the contract space is restricted to be a linear function of a real valued outcome (e.g. ``The Agent receives payment equal to 10\% of the revenue of the Principal''). Linear contracts are focal within the contract theory literature because they have a variety of robustness properties (see e.g. \cite{carroll2015robustness,dutting19})---and because they are the most commonly used type of contract in practice. As a result they have been the focus of a large fraction of the recent computational work in contract theory (see our discussion in the Related Work section). The second is the Bayesian Persuasion setting when the underlying state of nature is binary: e.g. drugs that are either effective or not, or defendants that are either innocent or guilty. This captures some of the best studied Bayesian Persuasion instances.


\section{Model}
\label{sec:model-cali}

Consider a repeated Stackelberg game between a female Principal and a male Agent with policy space $\cP$, action space $\cA$, and state space $\cY$. In rounds $t\in \{1,\ldots,T\}$, the Principal selects a policy $p_t\in \cP$ and (possibly) recommends an action $r_t\in \cA$ for the Agent. After observing the policy $p_t$ and the recommendation $r_t$, the Agent takes an action $a_t\in \cA$. At the end of round $t$, a state of nature $y_t$ chosen by nature is revealed to both the Principal and the Agent.
Utility functions depend on the action, the policy and the state of nature.
We denote the Agent's utility by $U(a_t,p_t,y_t)\in [-1,1]$ and the Principal's utility by $V(a_t,p_t,y_t) \in [-1,1]$.

When there is a known (to both the Principal and the Agent) common prior $\pi\in \Delta(\cY)$ and the state of nature $y_t$ is drawn from this prior, the Principal can maximize her utility by solving for an optimal policy by backwards induction, choosing the policy that will maximize her utility after the Agent best responds by breaking ties in favor of the Principal. Formally, for any prior distribution $\pi$, if the Principal selects a policy $p$, then the Agent will best respond to $(p, \pi)$ by choosing an action in $A^*(p,\pi):= \argmax_{a\in \cA} \EEs{y\sim \pi}{U(a,p,y)}$ to maximize the Agent's utility. When there are multiple best responding actions, the traditional assumption is that the Agent will break ties by maximizing the Principal's utility, i.e., 
\begin{equation}
    \brr(p,\pi) \in \argmax_{a\in A^*(p,\pi)} \EEs{y\sim \pi}{V(a,p,y)}\,. \label{eq:brr}
\end{equation} 

The Principal, assuming that the Agent will best respond, best responds to $\pi$ by selecting policy
\begin{equation}
    \brp(\pi) \in \argmax_{p\in \cP_0} \EEs{y\sim \pi}{V(\brr(p,\pi),p,y)}\,, \label{eq:brp}
\end{equation}
where $\cP_0\subseteq \cP$ is a set of given benchmark policies.
In Eq~\eqref{eq:brr} and~\eqref{eq:brp}, we break ties arbitrarily.
Therefore, given a prior $\pi$, the Principal will choose policy $p_t = \brp(\pi)$ and (may without loss of generality) recommend that the Agent take action $r_t= \brr(\brp(\pi),\pi)$. The Agent will follow the Principal's recommendation by taking action $r_t$.

In this work, we consider a more challenging prior-free scenario where there is no common prior and the states of the world can be generated adversarially. We also will \emph{not} assume that the Agent breaks ties in favor of the Principal.  
The Agent runs a learning algorithm $\cL$, which maps the state history $y_{1:t-1}$, the action history $a_{1:t-1}$, the recommendation history $r_{1:t-1}$, the policy history $p_{1:t-1}$, and the current policy $p_t$ and recommendation $r_t$ to a distribution over actions. Formally, the Agent's action distribution at round $t$ is given by a function:
\begin{equation*}
    \cL_t: \cY^{t-1}\times \cA^{t-1}\times \cA^{t}\times \cP^t\mapsto \Delta(\cA)\,.
\end{equation*}

The Principal runs a learning algorithm (henceforth, a mechanism $\sigma$) that maps the state history $y_{1:t-1}$, the recommendation history $r_{1:t-1}$, and the policy history $p_{1:t-1}$ to a distribution over policies and recommendations. Note that the Principal's algorithm does \emph{not} depend on the action history, which is by design (and in fact it is an important modelling choice that Agent's actions need not be directly observable to the Principal). The result is that the Principal's mechanism is \emph{nonresponsive} to Agent's actions, i.e., the Principal's policy at time $t$ does not depend on the Agent's action history. 
When mechanisms are nonresponsive, non-policy regret and policy regret coincide for the Agent and so lack of ``regret'' (to be defined shortly) is an unambiguously desirable property for the Agent to have. 
Formally, the Principal's policy distribution at round $t$ is given by a function: 
\begin{equation*}
    \sigma_t: \cY^{t-1}\times \cA^{t-1}\times \cP^{t-1}\mapsto \Delta(\cP\times \cA)\,.
\end{equation*}
In this work, we consider a specific family of mechanisms in which the Principal generates a forecast of the distribution over states in each round that will satisfy certain ``unbiasedness'' conditions, to be specified shortly. These forecasts will informally play the role of the prior distribution in the Principal's decision about which policy to offer. 

Specifically, assume that the Principal has access to a forecasting algorithm (implemented by either herself or a third party), which provides a forecast $\pi_t\in \Delta(\cY)$ of (the distribution over) the state in each round $t$. By viewing $\pi_t$ as the prior, the Principal selects policy $p_t = \psi(\pi_t)$, which is determined by $\pi_t$ and recommends that the Agent play the best response $r_t = \brr(p_t,\pi_t)$ --- as if $\pi_t$ were in fact a prior. The recommendation is the best action that the Agent could play were $\pi_t$ in fact a correct prior. The Agent is under no obligation to follow this recommendation, and may not. However, in our mechanism, the forecasts will turn out to guarantee that \emph{if} the Agent follows the recommendation, then he will have strong regret guarantees with respect to his own utility function---and the behavioral assumptions we impose on the Agent will require that he satisfies these regret guarantees (whether or not he chooses to do so by following the recommendation, or satisfies these guarantees through some other means).

We only consider deterministic rules $\psi: \Delta(\cY)\mapsto \cP$, mapping forecasts $\pi_t$ to policies $p_t$ and our recommendations will always be $r_t = \brr(p_t,\pi_t)$.
The Principal-Agent interaction protocol is described as follows.

\begin{algorithm}[H]
    \caption{Principal-Agent Interaction at round $t$}
    \label{prot:interaction}
        \begin{algorithmic}[1]
            \STATE {The Principal produces or obtains a forecast $\pi_t$.}
            \STATE {The Principal chooses policy $p_t = \psi(\pi_t)$ and recommends that the Agent play action $r_t = \brr(p_t,\pi_t)$.}
            \STATE {The Principal discloses $(p_t,r_t)$ to the Agent.}
            \STATE {The Agent takes an action $a_t \sim \cL_t(y_{1:t-1}, a_{1:t-1},r_{1:t}, p_{1:t})$.}
            \STATE {The state $y_t$ is revealed to both the Principal and the Agent.}
        \end{algorithmic}
    \end{algorithm}


Mechanisms designed within this framework (the only sort we consider in this paper) are specified by  a forecasting algorithm $\cF$ and a choice rule $\psi$ mapping forecasts to polices. 
Given a forecasting algorithm $\cF$, we want a choice rule $\psi$ that guarantees that the Principal has no ``regret'' to using, relative to having counter-factually offered the best fixed policy in hindsight, which we think of as using a constant ``mechanism'' from the set $\{\sigma^{p_0}|p_0\in  \cP_0\}$. The constant mechanism $\sigma^{p_0}$ ignores the history, and consistently chooses the policy $p_0\in \cP_0$ at every round, while recommending that the Agent take action $r_t^{p_0} = a^*(p_0,\pi_t)$---i.e. his best response to $p_0$ under the current realized forecast. Note that the sequence of \emph{forecasts} is the same under both the realized and counter-factual constant mechanism. Here we will define a strong notion of policy regret --- regret to the counterfactual world in which the Principal used a fixed policy, \emph{and the Agent responded to that fixed policy, producing a different sequence of actions}. 
Formally we define the Principal's policy regret as follows.

\begin{definition}[Principal's Regret]
    For a realized sequence of states of nature $y_{1:T}$, an Agent learning algorithm $\cL$, and a realized sequence of forecasts $\pi_{1:T}$, the Principal's policy regret from having used a rule $\psi$ is defined as:
\begin{equation*}
    \textrm{PR}(\psi,\pi_{1:T}, \cL, y_{1:T}) = \max_{p_0\in \cP_0} \EEs{a_{1:T},a^{p_0}_{1:T}}{\frac{1}{T}\sum_{t=1}^T\left(V(a_t^{p_0},p_0,y_t) -V(a_t,p_t,y_t)\right)} \,,
\end{equation*}
where $p_t = \psi(\pi_t)$ is the policy selected by the rule $\psi$, $a_{1:T}$ and $a^{p_0}_{1:T}$ are the sequences of actions generated by $\cL$ when the Principal selects policies according to the proposed rule $\psi$ and the constant policy $p_0$ respectively. The expectation is taken over the randomness of the learning algorithm $\cL$.
    

Observe that the forecasts are an argument to the Principal's regret, and these are random variables because the forecasting algorithm is permitted to be randomized. For a mechanism $\sigma^\dagger = (\cF,\psi)$, we compute the Principal's regret  by taking the expectation over the random forecasts generated by $\cF$
 \begin{equation*}
    \textrm{PR}(\sigma^\dagger,\cL, y_{1:T}) = \EEs{\pi_{1:T}}{\textrm{PR}(\psi,\pi_{1:T}, \cL, y_{1:T})}\,.
\end{equation*}
 
\end{definition}

Throughout this work, we consider finite action spaces and finite state spaces.
For notational simplicity, we represent actions $a\in \cA$ and states $y\in \cY$ in their one-hot encoding vector forms.

\section{Behavioral Assumptions}\label{sec:behavior}
In the common prior setting, it is clear how to model rational Agent behavior---the standard assumption is that the Agent chooses his action so as to maximize his payoff in expectation over the prior. This assumption, of course, no longer makes sense in a prior-free setting. However, we cannot simply drop all behavioral assumptions on the Agent when moving to the prior-free setting. 
Instead, we establish more general assumptions which make sense in the prior-free setting. Our behavioral assumptions must hold in both the realized sequence of play and in several counterfactual scenarios, so that we can meaningfully measure policy regret. Taken together, the assumptions below are strictly weaker than the assumption that the Agent always best-responds to a common prior. The reader can therefore view our behavioral assumptions as a strict generalization of the definition of rational behavior in a common prior setting, which can be studied in the prior-free setting. The assumptions will also end up being strictly weaker than the assumption that the Agent follows the Principal's recommended action --- so they are easily satisfied if the Agent chooses to do this, but do not constrain the Agent to following the Principal's recommendations. We will now introduce our two key assumptions, along with intuition for how they generalize the common prior setting. 

The first assumption generalizes the `best-response' behavior of the Agent. While our Agent may not have access to a prior to best-respond to, we can still rule out some clearly suboptimal behavior. A standard prior-free rationality assumption is that the Agent should have no swap regret: i.e. for each of his actions, on the subsequence of rounds on which he played that action, he should be obtaining utility at least what he could have guaranteed by playing the best \emph{fixed} action on that subsequence. Swap regret is an efficiently obtainable guarantee, weaker than pointwise optimality under a common prior, and having lower swap regret is always desirable, since the Principal is non-responsive. Of course, in our setting, in which the Principal first commits to a policy, which defines the best response correspondence of the Agent, it makes little sense to speak of the ``best fixed action'' without first conditioning on the policy offered by the Principal. So we ask for a form of contextual swap regret that is a better fit to our setting: namely, that the Agent should have no swap regret not just overall, but on each subsequence that results from \emph{fixing} the policy and recommendation made by the Principal.  This is always desirable (since the Principal is non-responsive), and efficiently obtainable in a prior-free setting: for example, by running a copy of a no-swap-regret algorithm like \cite{blum2007external} separately for each policy/recommendation pair $(p,r)$ offered by the Principal, or by best responding to appropriately calibrated, efficiently computable forecasts as in \cite{NRRX23}.



\begin{assumption}[No Contextual Swap Regret for The  Agent]\label{asp:no-internal-reg} 
We write $h:\cP\times \cA\times \cA\mapsto \cA$ to denote a modification rule that takes as input a policy and recommended action from the Principal, as well as a played action by the Agent, and as a function of these arguments ``swaps'' the Agent's action for an alternative action.  
Given the realized sequence of states $y_{1:T}$ and the realized sequence of policies and recommendations generated by either the deployed mechanism or the constant mechanisms, we define the Agent's swap regret to be:

    \begin{equation*}        \ir(y_{1:T},p_{1:T},r_{1:T}):=\E_{a_{1:T}}\left[\max_{h:\cP\times \cA\times \cA\mapsto \cA}\frac{1}{T}\sum_{t=1}^T (U(h(p_t,r_t,a_t),p_t, y_t) - U(a_t,p_t,y_t)) \right]\,,
    \end{equation*}
    and for all $p_0\in \cP_0$,
    \begin{equation*}        
    \ir(y_{1:T},(p_0,\ldots,p_0),r^{p_0}_{1:T}):=\E_{a_{1:T}^{p_0}}\left[\max_{h:\cP\times \cA\times \cA\mapsto \cA}\frac{1}{T}\sum_{t=1}^T (U(h(p_0,r^{p_0}_t,a^{p_0}_t),p_0, y_t) - U(a^{p_0}_t,p_0,y_t)) \right]\,.
    \end{equation*}
    We assume that there exists an $\eint$ such that for all fixed policies $p_0 \in \cP_0$ we have both:
    $$\ir(y_{1:T},p_{1:T},r_{1:T}) \leq \eint \ \ \ \ \ir(y_{1:T},(p_0,\ldots,p_0),r^{p_0}_{1:T}) \leq \eint\,.$$
    
\end{assumption}

The second assumption generalizes the notion of a shared prior. One important feature of the shared prior setting is that the realized state of nature is independent of the actions chosen by both the Principal and the Agent. In an adversarial setting, we can no longer appeal to statistical independence, as there is no distribution. But we need to preclude the possibility that the Agent somehow can ``predict the future'' in ways that the Principal can't. To do this, we make a ``no secret information'' assumption that informally requires that the Agent's actions appear to be (almost) statistically independent of the states of nature in the empirical transcript in terms of the utility functions of the Principal and Agent, conditionally on the policies and recommendations chosen by the Principal. Once again, this generalizes the shared prior assumption, in which we have actual statistical independence---and in which the Principal's ``recommendation'' is always the same as the Agent's action. Even in the adversarial setting, if for example, the Agent follows the Principal's recommendations, then this assumption will always be satisfied exactly --- but it can also be satisfied in many other ways. 
For any distribution $\mu$ over actions, let $U(\mu,p,y):= \EEs{a
\sim \mu}{U(a,p,y)}$ and $V(\mu,p,y):= \EEs{a
\sim \mu}{V(a,p,y)}$ denote the expected utilities when the action is sampled from $\mu$.


\begin{assumption}[No Secret Information]\label{asp:no-correlation} 
Consider any fixed sequence of forecasts $\pi_{1:T}$.
Given the sequence of policies $p_{1:T}$ and recommendations $r_{1:T}$ generated by the deployed mechanism, for any $(p,r)\in \cP\times \cA$, for any sequence of Agent's actions $a_{1:T}$ generated by $\cL$, let $\hat \mu_{p,r} = \frac{1}{n_{p,r}}\sum_{t:(p_t,r_t)= (p,r)} a_t$, where $n_{p,r} = |\{t :(p_t,r_t)= (p,r)\}|$, denote the empirical distribution of the Agent's actions during the subsequence of rounds in which $(p_t,r_t) = (p,r)$.
Then we assume that for all $(p,r)\in \cP\times \cA$,
\begin{align*}
\frac{1}{n_{p,r}}\EEs{a_{1:T}}{\abs{\sum_{t : (p_t,r_t) = (p,r)} (U(a_t, p, y_t) -U(\hat \mu_{p,r}, p, y_t))}}\leq \cO\left(\frac{1}{\sqrt{n_{p,r}}}\right)\,,\\
\frac{1}{n_{p,r}}\EEs{a_{1:T}}{\abs{\sum_{t: (p_t,r_t) = (p,r)} (V(a_t, p, y_t) -V(\hat \mu_{p,r}, p, y_t))}}\leq \cO\left(\frac{1}{\sqrt{n_{p,r}}}\right)\,.
\end{align*}

Similarly, given the sequence of policies $(p_0,\ldots,p_0)$ and recommendations $r^{p_0}_{1:T}$ generated by constant mechanism $\sigma^{p_0}$, for any $r\in \cA$, let $\hat \mu^{p_0}_{r} = \frac{1}{n^{p_0}_{r}}\sum_{t:r^{p_0}_t=r} a_t^{p_0}$, where $n^{p_0}_{r} = |\{t : r_t^{p_0}=r\}|$, denote the empirical distribution of the Agent's actions during the period's in which the recommendation $r_t^{p_0} = r$. 
Then we assume that, for all $p_0\in \cP_0$, for all $r\in \cA$,
\begin{align*}
    \frac{1}{n^{p_0}_{r}}\EEs{a^{p_0}_{1:T}}{\abs{\sum_{t:r^{p_0}_t=r} (U(a_t^{p_0}, p_0, y_t) - U(\hat \mu^{p_0}_{r}, p_0, y_t))}}\leq \cO\left(\frac{1}{\sqrt{n^{p_0}_{r}}}\right)\,,\\
     \frac{1}{n^{p_0}_{r}}\EEs{a^{p_0}_{1:T}}{\abs{\sum_{t:r^{p_0}_t=r} (V(a_t^{p_0}, p_0, y_t) - V(\hat \mu^{p_0}_{r}, p_0, y_t))}}\leq \cO\left(\frac{1}{\sqrt{n^{p_0}_{r}}}\right)\,.
\end{align*}
\end{assumption}

While the need for Assumption~\ref{asp:no-internal-reg} is clear, the need for Assumption~\ref{asp:no-correlation} is less immediately clear. However it is indeed the case that Assumption~\ref{asp:no-internal-reg} is insufficient on its own.

\begin{restatable}[Necessity of Assumption~\ref{asp:no-correlation}]{proposition}{necessity}\label{lem:lb-easy}
There exists a simple linear contract setting where, for any Principal mechanism $\sigma$, one of the following must hold:
\begin{itemize}
    \item No learning algorithm $\cL^{*}$ can satisfy Assumption~\ref{asp:no-internal-reg} with $\eint = o(1)$ for all possible sequence of states $y_{1:T}\in \cY^T$.
    \item There exists a learning algorithm $\cL^{*}$ satisfying Assumption~\ref{asp:no-internal-reg} with $\eint = o(1)$ for all possible sequence of states $y_{1:T}\in \cY^T$ and a sequence of states $\bar y_{1:T} \in \cY^T$ for which $\sigma$ achieves non-vanishing regret, i.e., $\text{PR}(\sigma,\cL^{*},\bar y_{1:T}) = \Omega(1)$.
\end{itemize}
\end{restatable}

We will prove in Section~\ref{sec:linear-contract} that in this same setting, if $\mathcal{L}$ satisfies Assumption~\ref{asp:no-internal-reg} and~\ref{asp:no-correlation}, there does exist a Principal mechanism which guarantees vanishing policy regret against $\mathcal{L}$. Therefore, Assumption~\ref{asp:no-correlation} plays an important role in our result. 
We will further discuss the necessity of the assumption in Appendix~\ref{sec:imposs}, where we also show that this impossibility result remains true even when Assumption~\ref{asp:no-internal-reg} is paired with an additional assumption which is in the same spirit of, but strictly weaker than, Assumption~\ref{asp:no-correlation}.

\section{Games with Stable Policy Oracles}\label{sec:stable}

In this section, we present a general no-policy-regret mechanism which applies in all settings where the Agent has access to a \emph{stable policy oracle}. A stable policy oracle is informally a way of producing or adjusting a policy to ensure that the Agent has only a single approximate best response given a particular fixed prior---or else that the Principal is almost indifferent between all of the Agent's approximate best responses.  What we will show is that the existence of such an oracle obviates the need for the kinds of very strong \emph{alignment} assumptions made in \cite{camara2020mechanisms}. In Section \ref{sec:oracles} we show that we in fact can implement such ``oracles'' in two very important cases: Principal Agent problems with \emph{linear} contracts, and binary state Bayesian Persuasion games, which allows us to obtain diminishing policy regret in these settings with minimal assumptions. In Section \ref{sec:general}, we extend our analysis to the general case (where Agents might unavoidably have multiple approximate best responses that the Principal is not indifferent between) --- there we will have to make the same kind of alignment assumption that is made in \cite{camara2020mechanisms}. 


Recall that we aim to resolve \emph{two} shortcomings of \cite{camara2020mechanisms}: the exponential computational and statistical complexity of producing calibrated forecasts, as well as the necessity to make strong alignment assumptions. To resolve the first issue,  rather than having the Principal produce calibrated forecasts, we have the Principal produce forecasts that satisfy a substantially weaker condition: unbiasedness subject to polynomially many ``events'', that will be eventually determined by the Principal's choice of policy and recommendation. Recent work of \cite{NRRX23} gives an algorithm for producing $d$-dimensional forecasts that satisfy this unbiasedness condition for polynomially in $d$ many events in time that is polynomial in $d$. Hence, this condition can be obtained with running time and bias bounds that scale only polynomially (rather than exponentially) in $|\cY|$. 

To resolve the second issue, rather than using the forecast $\pi_t$ directly as a prior and choosing the policy that would exactly optimize the Principal's payoff, we choose our policy using a stable policy oracle, defined below, which finds a policy that eliminates near ties: this will remove the necessity of an alignment assumption.  


First we define our notion of conditional bias.

\begin{definition}[Conditional Bias of Forecasts]\label{def:cal}
Let $\cE$ be a collection of ``events'', each defined by a function $E:\Delta(\cY)\rightarrow \{0,1\}$. 
For any sequence of states $y_{1:T}$, any sequence of forecasts $\pi_{1:T}$, and a collection of events $\cE$, we say $\pi_{1:T}$ has bias $\alpha$ conditional on $\cE$ if for all $E\in \cE$:
\begin{equation*}
    \frac{1}{T} \norm{\sum_{t=1}^T E(\pi_t) (\pi_t -y_t)}_1 \leq \alpha(E)\,.
\end{equation*}
\end{definition}

\cite{NRRX23} show how to efficiently make predictions obtaining low conditional bias against an adversarially chosen state sequence, for any polynomially sized collection of events:

\begin{theorem}[\cite{NRRX23}]\label{thm:forecast-bias}
    For any collection of events $\cE$ that can each be evaluated in polynomial time, there is a forecasting algorithm with per-round running time polynomial in $|\cY|$ and $|\cE|$ that produces forecasts $\pi_{1:T}$ such that for any (adversarially) chosen sequence of outcomes $y_{1:T}$, the expected bias conditional on $\cE$ is bounded by:
    $$\EEs{\pi_{1:T}}{\alpha(E)} \leq O\left(\frac{|\cY|\ln(|\cY||\cE|T)}{T} + \frac{|\cY|\sqrt{\ln(|\cY||\cE|T)|\{t : E(\pi_t) = 1|\}}}{T}\right) \leq O\left(\frac{|\cY|\sqrt{\ln(|\cY||\cE|T)}}{\sqrt{T}}\right)\,.$$
\end{theorem}

Next, we formalize our notion of a ``stable policy'' and a ``stable policy oracle''. Informally, what we need to deal with is  the possibility that the Agent has a range of approximate best responses with very different payoffs for the Principal. If this is the case, then the Agent could behave very differently given seemingly unimportant changes to the Principal's mechanism, leading to high policy regret. In many settings it is possible resolve this issue by adjusting the per-round policies a small amount to ensure a unique approximate best response---or else approximate indifference for the Principal between all of the Agent's approximate best responses.  

For any given prior distribution $\pi$, we say a policy $p$ is stable if choosing any action $a$ that deviates from the optimistic best response $\brr(p,\pi)$ results in either significantly lower Agent utility or a comparable level of utility for the Principal. Informally, this will mean that the Principal's payoff can be reliably predicted given the policy, assuming only that the Agent plays an approximate best response: any approximate best response will yield approximately the same payoff for the Principal. We emphasize that we will not \emph{assume} that policies are stable, but \emph{enforce it}. More specifically, for any prior distribution $\pi$, let $V(a,p,\pi) =\EEs{y\sim \pi}{V(a,p,y)}$ and $U(a,p,\pi) = \EEs{y\sim \pi}{U(a,p,y)}$ denote the expected utilities for the Principal and the Agent when the state $y$ is drawn from $\pi$. We define stable policies as follows.
\begin{definition}[Stable Policy]
    For any $\beta, \gamma>0$ and $\pi\in \Delta(\cY)$, a policy $p$ is $(\beta,\gamma)$-stable under $\pi$ if for all $a \neq \brr(p,\pi)$ in $\cA$, we have either
    \begin{equation*}
        U(a,p,\pi)\leq U(\brr(p,\pi),p,\pi) -\beta\,, 
    \end{equation*}
    or
    \begin{equation*}
        V(a,p,\pi)\geq V(\brr(p,\pi),p,\pi) -\gamma\,. 
    \end{equation*}
\end{definition}

Classically, in the common prior setting, both the Principal and the Agent best respond to (exactly) maximize their expected utilities. As discussed, in our setting, we have relaxed this best response assumption to a low-contextual-swap-regret assumption (Assumption~\ref{asp:no-internal-reg}), which is in fact a relaxation of an \emph{approximate} best response assumption --- i.e. it is satisfied in the commmon prior setting even if Agents do not exactly best respond, but merely approximately best respond. How shall we deal with this?

The Principal's utility would be maximized if the Agent were to choose amongst his approximate best responses so as to optimize for the Principal.  Specifically, let $\cB(p,\pi,\epsilon):= \{a\in \cA| U(a,p,\pi)\geq U(\brr(p,\pi),p,\pi) -\epsilon\}$ denote the set of all $\epsilon$-best responses for the Agent and let $\brr(p,\pi,\epsilon)$ denote the utility-maximizing action for the Principal, amongst the Agent's $\epsilon$-best responses to $p$, i.e.,
$$\brr(p,\pi,\epsilon) = \argmax_{a\in \cB(p,\pi,\epsilon)} V(a,p,\pi)\,.$$
Given any $\pi$, we say that a policy $p$ is an optimal stable policy under $\pi$ if $p$ is stable and implementing $p$ will lead to utility for the Principal that is comparable with her best achievable utility---i.e. the utility that the Principal could have obtained were the Agent guaranteed to  choose amongst his $\epsilon$-approximate best responses in the way that has highest payoff for the Principal.

\begin{definition}[Optimal Stable Policy Oracle]\label{def:stabilized}
    For a prior distribution $\pi$, we say that a policy $p$ is a $(c,\epsilon,\beta,\gamma)$-optimal stable policy under $\pi$ if 
    \begin{itemize}
        \item $p$ is $(\beta,\gamma)$-stable under $\pi$;
        \item and $V(a^*(p,\pi),p,\pi)\geq V(a^*(p_0,\pi,\epsilon),p_0,\pi)-c$ for all $p_0\in \cP_0$.
    \end{itemize} 
    An optimal stable policy oracle $\cO_{c,\epsilon,\beta,\gamma}: \Delta(\cY) \mapsto \cP_\cO$, given as input any prior $\pi$, outputs a $(c,\epsilon,\beta,\gamma)$-optimal stable policy in $\cP_\cO$ under $\pi$, where $\cP_\cO\subseteq \cP$ is the set of all possible output policies by the oracle.
\end{definition}

Intuitively, when $\beta > \epsilon$, then if the Agent can be assumed to play an $\epsilon$-best response to $\pi$ this is sufficient to guarantee that when the Principal deploys an optimal stable policy, she will obtain utility comparable to the utility she could have obtained assuming that the Agent were to best respond exactly while tiebreaking in the Principal's favor (i.e.  $V(\brr(p,\pi),p,\pi)$), and that, $V(\brr(p,\pi),p,\pi)$ is larger than the the utility achieved by any benchmark policy even if the Agent could have been assumed to optimistically respond. With such an oracle we can construct the mechanism described in Algorithm~\ref{alg:general-stable}, that guarantees the Principal no policy regret. Of course, we do \emph{not} assume that the Agent $\epsilon$-best responds to the forecast $\pi_t$ at round $t$ --- but as we will show, Assumptions \ref{asp:no-internal-reg} and \ref{asp:no-correlation} will be enough to make the analysis go through.

\begin{algorithm}[H]
    \caption{Principal's choice at round $t$}
    \label{alg:general-stable}
        \begin{algorithmic}[1]
            \STATE {\textbf{Input}: 
            Forecast $\pi_t\in \Delta(\cY)$
            }
            \STATE {Call the optimal stable policy oracle $\cO_{c,\epsilon,\beta,\gamma}$ to get a policy $p_t = \cO_{c,\epsilon,\beta,\gamma}(\pi_t)$ 
            }
        \end{algorithmic}
    \end{algorithm}

Let $\pit = \argmax_{p_0\in \cP_0} V(a^*(p_0,\pip_t,\epsilon),p_0,\pip_t)$ denote the policy that the Principal would pick if the Agent optimistically best responded to $(\pit,\pi_t)$ and $\rit = a^*(\pit,\pip_t,\epsilon)$ denote the corresponding optimistic $\epsilon$-best responding action.

\begin{restatable}{theorem}{thmstable}\label{thm:stable}

Define the following collections of events:
$$\cE_1 = \{\ind{(p_t,r_t) = (p,r)}\}_{p \in \cP_\cO, r \in \cA}\,, \ \ \  \ \ \cE_2 = \{\ind{(\pit,\rit) = (p,a)}\}_{p \in \cP_0, a \in \cA}\,,$$
$$\cE_3 = \{\ind{a^*(p_0,\pi_t) = a}\}_{p_0 \in \cP_0,a\in \cA}\,.$$
Let $\cE = \cE_1 \cup \cE_2 \cup \cE_3$, the union of these events. 
Assume that the Agent's learning algorithm $\cL$ satisfies the behavioral assumptions~\ref{asp:no-internal-reg} and \ref{asp:no-correlation}.
Given access to an optimal stable policy oracle $\cO_{c,\epsilon,\beta,\gamma}$, by running the forecasting algorithm from \cite{NRRX23} for events $\cE$ and the choice rule in Algorithm~\ref{alg:general-stable}, the Principal can achieve policy regret
\begin{align*}
&\text{PR}(\sigma^\dagger, \cL,y_{1:T}) \leq\tilde \cO\left(c +\gamma +\sqrt{\frac{\abs{\cP_0}\abs{\cA}}{T}} + \frac{\eint + \abs{\cY}\sqrt{\abs{\cP_\cO}\abs{\cA}/T}}{\beta} + \frac{\eint + \abs{\cY}\sqrt{\abs{\cA}/T}}{\epsilon}\right)\,,
\end{align*}
where $\tilde \cO$ ignores logarithmic factors in $T, \abs{\cY}, \abs{\cP_\cO},\abs{\cP_0}, \abs{\cA}$.
\end{restatable}
Note that we consider a fixed benchmark policy set, a fixed action space and a fixed state space. Hence we have that $\abs{\cP_0}$, $\abs{\cA}$ and $\abs{\cY}$ are all independent of $T$.
If we can construct an optimal stable policy oracle with $c,\gamma, \frac{\eint}{\beta},\frac{\sqrt{\abs{\cP_\cO}/T}}{\beta},\frac{\eint}{\epsilon},\frac{1}{\epsilon\sqrt{T}}  = o(1)$, then we can achieve vanishing regret $\text{PR}(\sigma^\dagger,\cL,y_{1:T}) = o(1)$. 
If the Agent is running a standard no-swap-regret algorithm, e.g. \citep{blum2007external}, the Agent can obtain swap regret $\eint = \cO(\sqrt{\abs{\cP_\cO}/T})$. We note that while it appears that the regret bound is decreasing in $\beta$ and $\epsilon$, when we actually construct optimal stable policy oracles in Section \ref{sec:oracles}, $c$ will grow with $\beta$ and $\epsilon$, and so there will be a tradeoff to manage.
The proof the theorem is deferred to Section~\ref{sec:stable-proof}.

\section{Constructing Stable Policy Oracles}
\label{sec:oracles}
In this section, we instantiate the general algorithm we derived in Section \ref{sec:stable} by constructing efficient stable policy oracles for two important special cases of the general Principal-Agent setting: the \emph{linear contracting} problem and the \emph{Bayesian Persuasion} problem in which there is an unknown binary state of nature. Linear contracting in particular has been focal in the contract theory literature due to the robustness and practical ubiquity of linear contracts \cite{carroll2015robustness,dutting19} --- and much of the recent computational and learning theoretic work on contract theory has focused exclusively or primarily on linear contracts. Binary state Bayesian Persuasion is a canonical case in Bayesian Persuasion, encompassing various intriguing scenarios, such as the FDA approval example.
In the following, we will introduce these two problems and construct efficient stable policy oracles for them.

\subsection{Linear Contracts}
\label{sec:linear-contract}
 In the contract setting, there is a finite outcome space $\mathbb{O} = \{o_1,\ldots,o_m\}$ (e.g., \{success, failure\}).
A contract $p: \mathbb{O} \mapsto [0,1]$ is a mapping from outcomes to payments and the Principal commits to pay the Agent a specified amount $p(o)$ if the outcome is $o$.
The Principal provides a contract to the Agent, and the Agent then decides to take an action (e.g., working or shirking). 
The Agent's action and the state of nature (e.g., hard job or easy job) together determine the outcome through a mapping $o: \cA\times \cY\mapsto \mathbb{O}$. 
Different outcomes will lead to different outcome values.
The Agent incurs different costs by taking different actions.
Then the utility of the Principal is the difference between the the outcome value and the payment to the Agent. The utility of the Agent is the difference between the payment and the cost of taking the action.
More specifically, 
let $v: \mathbb{O}\mapsto [0,1]$ denote the value function of outcomes and $c:\cA\mapsto [0,1]$ denote the cost function for the Agent.
When the Principal offers contract $p$, the Agent takes action $a$, and the outcome is $o$, then the Principal's utility is $v(o) - p(o)$ and Agent's utility is $p(o) - c(a)$.

Our focus will be on linear contracts, a particularly simple and widespread type of contract which provides the Agent with a constant fraction of the outcome value. Linear contracts are focal in the contract theory literature in part because of their robustness properties \citep{carroll2015robustness,dutting19}.

\begin{definition}[Linear contract]
For a linear contract parameterized by $p \in [0,1]$, the Principal pays the Agent a $p$-fraction of the value, i.e., $p \cdot v(o)$ when the outcome is $o$. Hence, we use this fraction to represent the linear contract and write the policy space as $\cP = [0,1]$, the set of all parameters that can specify a linear contract. 
\end{definition}
For any linear contract $p\in \cP$, action $a\in \cA$ and state of nature $y\in \cY$, the Principal's utility is 
\begin{align*}
    V(a,p,y) = v(o(a,y)) - p\cdot v(o(a,y)) = (1-p)v(o(a,y))\,,
\end{align*}
and the Agent's utility is 
\begin{align*}
    U(a,p,y) = p\cdot v(o(a,y)) - c(a)\,.
\end{align*}
We consider a finite action space and assume that the costs are different for each action. Hence the minimum gap between the costs is positive, and we denote it by:
\begin{align*}
    \Delta_c = \min_{a_1, a_2\in \cA: a_1\neq a_2}\abs{c(a_1)-c(a_2)}>0\,.
\end{align*}
For any action $a\in \cA$ and prior $\pi$, let $$f(\pi,a) := \EEs{y \sim \pi}{v(o(a,y))} $$ denote the expected outcome value when the Agent takes action $a$ and the state of nature is drawn from the prior distribution $\pi$. Then the Principal's utility under $\pi$ can be written as 
\begin{align}
    V(a,p,\pi) = \EEs{y\sim \pi}{(1-p) \cdot v(o(a,y))} = (1-p) f(\pi,a)\,,\label{eq:expectedV}
\end{align}
and the Agent's utility can be written as 
\begin{align}
    U(a,p,\pi) =\EEs{y\sim \pi}{p \cdot v(o(a,y))}- c(a)= p f(\pi,a) - c(a)\,.\label{eq:expectedU}
\end{align}

Then we can construct an optimal stable policy oracle as follows.
Given any prior $\pi$, we initially identify the policy $p^\text{optimistic}$ that maximizes the Principal's utility assuming that the Agent optimistically approximately best responds---i.e. chooses the action amongst all of his \emph{approximate} best responses that maximizes the Principal's utility. However, $p^\text{optimistic}$ will generally be unstable, and thus the Agent may not actually  optimistically respond if we were to implement $p^\text{optimistic}$. 
The subsequent step involves stabilizing $p^\text{optimistic}$ by incrementally adjusting the contract until it becomes stable. 
It turns out that a small increase in $p^\text{optimistic}$ allows us to obtain a stable policy. Since this policy is close to $p^\text{optimistic}$, the Principal's utility remains comparable to the performance of any benchmark policy when the Agent optimistically approximately best responds---even though the stabilization means we no longer need to \emph{assume} that the Agent will optimistically best respond. Finally, recall that the regret guarantee in Theorem~\ref{thm:stable} depends on the cardinality of the output policy space. Consequently, we will have to discretize the policy space and provide a discretized stable policy.
Let $\cP_\delta = \{0, \delta, 2\delta, \ldots, \floor{\frac{1}{\delta}}\delta\}$ denote a $\delta$-cover of the linear contract space for
some $\delta =o(1)$. We construct the following optimal stable policy oracle with output space $\cP_\cO = \cP_\delta$ so that $\abs{\cP_\cO} = \floor{\frac{1}{\delta}}+1$.

\begin{algorithm}[H]
    \caption{Optimal Stable Policy Oracle for Linear Contracts}
    \label{alg:linear-oracle}
        \begin{algorithmic}[1]
            \STATE {\textbf{Parameters}: stability parameter $\beta$, discretization parameter $\delta$}
            \STATE {\textbf{Input}: prior distribution $\pi$}
            \STATE {Compute $p^\text{optimistic} = \argmax_{p\in \cP_0} \max_{a\in \cB(p,\pi,\frac{\Delta_{c}\beta}{2})} V(a,p,\pip)$}
            \STATE {\textbf{Output}: $$p(\pi) = \min\left(\left\{p\in \cP_\delta|p\geq p^\text{optimistic}, p \text{ is } \left(\frac{\Delta_c \beta}{2},0\right)\text{-stable under }\pi \right\}\cup \{1\}\right)$$}
        \end{algorithmic}
    \end{algorithm}

\begin{theorem}[Optimal Stable Policy Oracle for Linear Contracts]\label{thm:linear}
    Algorithm~\ref{alg:linear-oracle} is a $(|\cA|(\beta+\delta),\frac{\Delta_{c}\beta}{2},\frac{\Delta_{c}\beta}{2},0)$-optimal stable policy oracle with $\abs{\cP_\cO} = \cO(\frac{1}{\delta})$. By combining with Theorem~\ref{thm:stable} and setting $\beta = T^{-\frac{1}{4}}$ and $\delta = \sqrt{\beta}$, we can achieve Principal's regret: 
    $$\text{PR}(\sigma^\dagger, \cL,y_{1:T}) = \tilde\cO\left(T^{-\frac{1}{8}}\right)\,,$$
    when the Agent obtains swap regret $\eint = \cO(\sqrt{\abs{\cP_\cO}/T})$.
\end{theorem}

\begin{proof}[proof sketch]
According to the definition of optimal stable policy oracle (Definition~\ref{def:stabilized}), the proof of the theorem follows directly from Lemma~\ref{lem:stable_contract} and Lemma~\ref{lem:payoff_close}.
\begin{restatable}{lemma}{lmmlinearstable}
    \label{lmm:linear-stable}
   For any prior $\pi$, the policy $p(\pi)$ returned by Algorithm~\ref{alg:linear-oracle}, is a $(\frac{\beta \Delta_c}{2},0)$-stable policy under $\pi$ and satisfies that $p(\pi) \leq p^{\textit{optimistic}} + \abs{\cA}(\beta + \delta)$ . \label{lem:stable_contract}
\end{restatable}

\begin{restatable}{lemma}{lmmlinearopt}\label{lmm:linear-opt}
    For any prior $\pi$, the policy $p(\pi)$ returned by Algorithm~\ref{alg:linear-oracle} satisfies that
    $$V(a^*(p(\pi),\pi),p(\pi),\pi)\geq V(a^*(p_0,\pi,\frac{\beta \Delta_c}{2}),p_0,\pi) - |\cA|(\beta + \delta)$$
    for all $p_0\in \cP_0$. \label{lem:payoff_close}
\end{restatable}

Lemma~\ref{lem:stable_contract} shows that for any $\pi$, the returned  linear contract $p(\pi)$ is $(\frac{\Delta_c \beta}{2},0)$-stable and is not much larger than $p^\text{optimistic}$.
This implies that the Principal will not pay a much larger fraction of her value under $p(\pi)$ than she would under $p^\text{optimistic}$. In Lemma~\ref{lem:payoff_close}, we prove that the Principal's utility under $p(\pi)$ is comparable to her utility under any benchmark contract.

By Lemmas~\ref{lem:stable_contract} and~\ref{lem:payoff_close}, we get that, for any prior $\pi$, the policy $p(\pi)$ returned by Algorithm~\ref{alg:linear-oracle} is a $(\frac{\beta \Delta_{c}}{2},0)$-stable policy under $\pi$, and furthermore that $V(a^*(p(\pi),\pi),p(\pi),\pi)\geq V(a^*(p_0,\pi,\Delta_{c}\frac{\beta}{2}),p_0,\pi)-\delta - |\cA|\beta$ for all $p_0\in \cP_0$. Putting these together proves that Algorithm~\ref{alg:linear-oracle} is a $(\delta + \beta|\cA|,\frac{\Delta_{c}\beta}{2},\frac{\Delta_{c}\beta}{2},0)$-optimal stable policy oracle.
\end{proof}

\subsection{Bayesian Persuasion}\label{sec:bayes}
Bayesian Persuasion is another important special case of the general Principal Agent problem that is quite different from the linear contracting case. 
In Bayesian Persuasion~\citep{kamenica2011bayesian}, Sender (the Principal) wishes to persuade Receiver (the Agent), to choose a particular action: but by controlling the information structure used to communicate with Receiver, rather than by making monetary payments.
For example, a traditional example is a prosecutor (Sender) who tries to convince a judge (Receiver) that a defendant is guilty.
\subsubsection{Fundamentals of Bayesian Persuasion}
A policy in Bayesian Persuasion is a signal scheme, which consists of a signal space $\Sigma$ and a family of distributions $\{\sig(\cdot|y)\in \Delta (\Sigma)\}_{y\in \cY}$ mapping ``states of nature'' $\cY$ to ``signals'' $\Sigma$.
Sender selects and sends a signal scheme to Receiver.
After observing the signal scheme and a signal realization $\sigma\sim \sig(\cdot|y)$ as a function of the underlying state of nature $y$, Receiver selects her strategy $s$ from a strategy space $\cS$.
In other words, after observing the signal scheme, Receiver selects an action $a: \Sigma\mapsto \cS$, which maps signals to strategies.
Both Sender's utility $v(s,y)\in [0,1]$ and Receiver's utility $u(s,y)\in [0,1]$ are functions of Receiver's strategy $s\in \cS$ and the state of nature $y\in \cY$.
For any policy $p$ and any action $a$, the Principal's utility is
\begin{equation*}
    V(a,p,y)=\EEs{\sigma\sim p(\cdot|y)}{v(a(\sigma), y)}\,,
\end{equation*}
and the Agent's utility is
\begin{equation*}
    U(a,p,y)=\EEs{\sigma\sim p(\cdot|y)}{u(a(\sigma),y)}\,.
\end{equation*}

In the common prior setting, there exists a common prior distribution $\pi$ over states of nature $\cY$.
To maximize the expected utility, the Agent will form his posterior distribution conditional on the signal $\pi_\sigma= \pi(y|\sigma)$ using Bayes's rule and best respond by selecting strategy $\argmax_{s\in \cS} \EEs{y\sim \pi_{\sigma}}{u(s,y)}$.
Consider the traditional example of a  prosecutor and a judge. The state space is $\cY$ = \{Innocent, Guilty\} and the strategy space is $\cS$ = \{Convict, Acquit\}.
The judge has 0-1 utility and prefers to convict if the defendant is guilty and acquit if the defendant is innocent.
Regardless of the state, the prosecutor’s utility is 1 following a conviction and 0 following an acquittal.
Consider the case that $\pi(\text{Guilty}) = 0.3$. 
If there is no communication, the judge will  always acquits because guilt is less likely than innocence under his prior. 
However, the prosecutor can construct the following signal scheme to improve her utility.
\begin{align*}
   &p(\text{i} | \text{Innocent}) = \frac{4}{7}, \quad p(\text{g} | \text{Innocent}) = \frac{3}{7}\,,\\
   &p(\text{i} | \text{Guilty}) = 0, \quad p(\text{g} | \text{Guilty}) = 1\,.
\end{align*}
The posterior distribution of observing signal $g$ is $\pi_{g}(\text{Guilty}) = \pi_{g}(\text{Innocent}) = 0.5$ and the judge will convict when observing signal $g$. This leads the judge to convict with probability $0.6$.

A signal scheme is said to be ``straightforward'' if the signal space $\Sigma= \cS$ and Receiver's best responding strategy equals the signal realization. In other words, a straightforward signal scheme simply tells the receiver what action to take, and it is in the reciever's interest to comply. \cite{kamenica2011bayesian} shows that the optimal value can be achieved by straightforward signal schemes. 
Hence, we restrict to straightforward signal schemes in the following and let $\cP$ be the space of all straightforward signal schemes.

A common special case of Bayesian Persuasion is that both the states of nature and the strategies are real-valued, 
and Sender's preferences over Receiver's strategies do not depend on the nature state $y$. 
Hence, Sender's utility can be written as a function of Receiver's strategy, i.e.,
\begin{equation*}
    v(s, y) = v(s)\,.
\end{equation*}
We consider a simpler but very common case where the number of states of nature is $2$, i.e., $\abs{\cY} = 2$.  In the example of prosecutor, the states are \{Innocent, Guilty\}. In the context of drug trials, a drug company (the Principal) seeks approval from FDA (the Agent) for a new drug, and the states are \{Effective, Ineffective\}. We remark that in this special case, our improvement over \cite{camara2020mechanisms} is in the removal of the Alignment assumption ---  since the state space is binary, our general efficiency improvements in terms of the cardinality of the state space are not relevant.
Without loss of generality, we assume that $\cY = \{0,1\}$ and $\cS\subset [0,1]$. We consider finite discrete strategy space $\cS$.
For any $\mu\in [0,1]$, let $u(s,\mu) = \EEs{y\sim\Ber(\mu)}{u(s,y)}$ denote the expected Agent's utility of choosing $s$ when $y$ is drawn from $\Ber(\mu)$.  We will assume (without loss of generality) that every strategy is a best response for the Agent for \emph{some} prior distribution (otherwise we can remove such a strategy from $\cS$):
\begin{assumption}\label{asp-bayes:interval}
We assume that for all $s\in \cS$, there exists a $\mu\in [0,1]$ such that $u(s,\mu)>u(s',\mu)$ for all $s'\neq s$ in $\cS$. 
\end{assumption}


Since we only focus on Bernoulli distributions, when we refer to $\mu$ as a belief/prior, we are using this as shorthand for the distribution $\Ber(\mu)$.
For any $\mu\in [0,1]$, let
$S^*(\mu) = \argmax_{s\in \cS}u(s,\mu)$
denote the set of optimal strategies under prior $\mu$ and let 
$$s^*(\mu) = \argmax_{s\in S^*(\mu)}v(s)$$ 
denote the optimal strategy breaking ties by maximizing the Principal's utility.

\begin{figure}[!t]
    \centering
    \begin{subfigure}[b]{0.5\textwidth}
    \includegraphics[width = \textwidth]{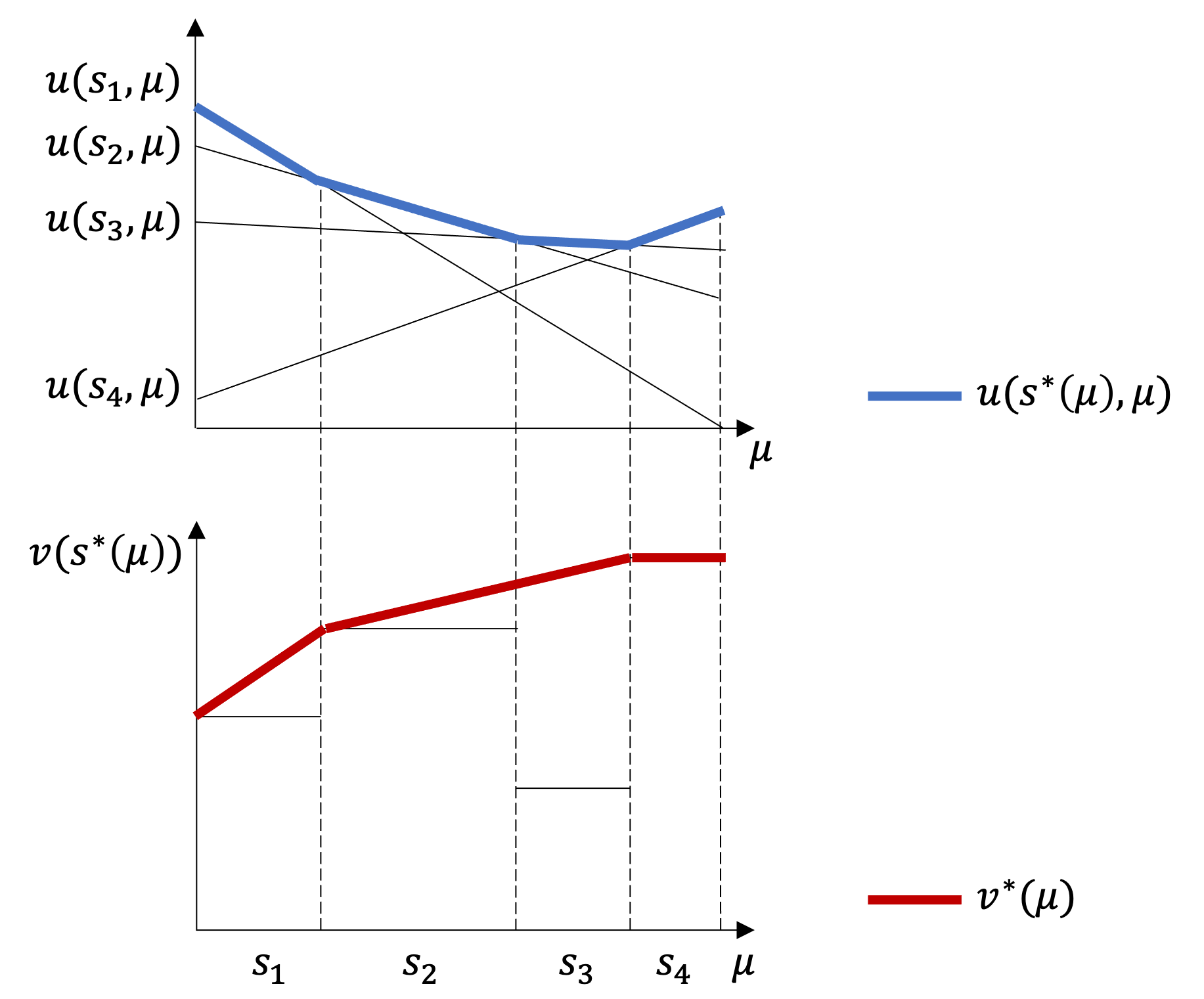}
    \caption{For any $s\in \cS$, $u(s,\mu)$ is a linear function of $\mu$. $u(s^*(\mu),\mu)$ is the maximum over all these linear functions. $v(s^*(\mu))$ is a piecewise constant function. $v^*(\mu)$ is defined in Eq~\eqref{eq:opt-bayes}.}
    \label{fig:bayes}
    \end{subfigure}
    \hfill
    \begin{subfigure}[b]{0.45\textwidth}
    \includegraphics[width = \textwidth]{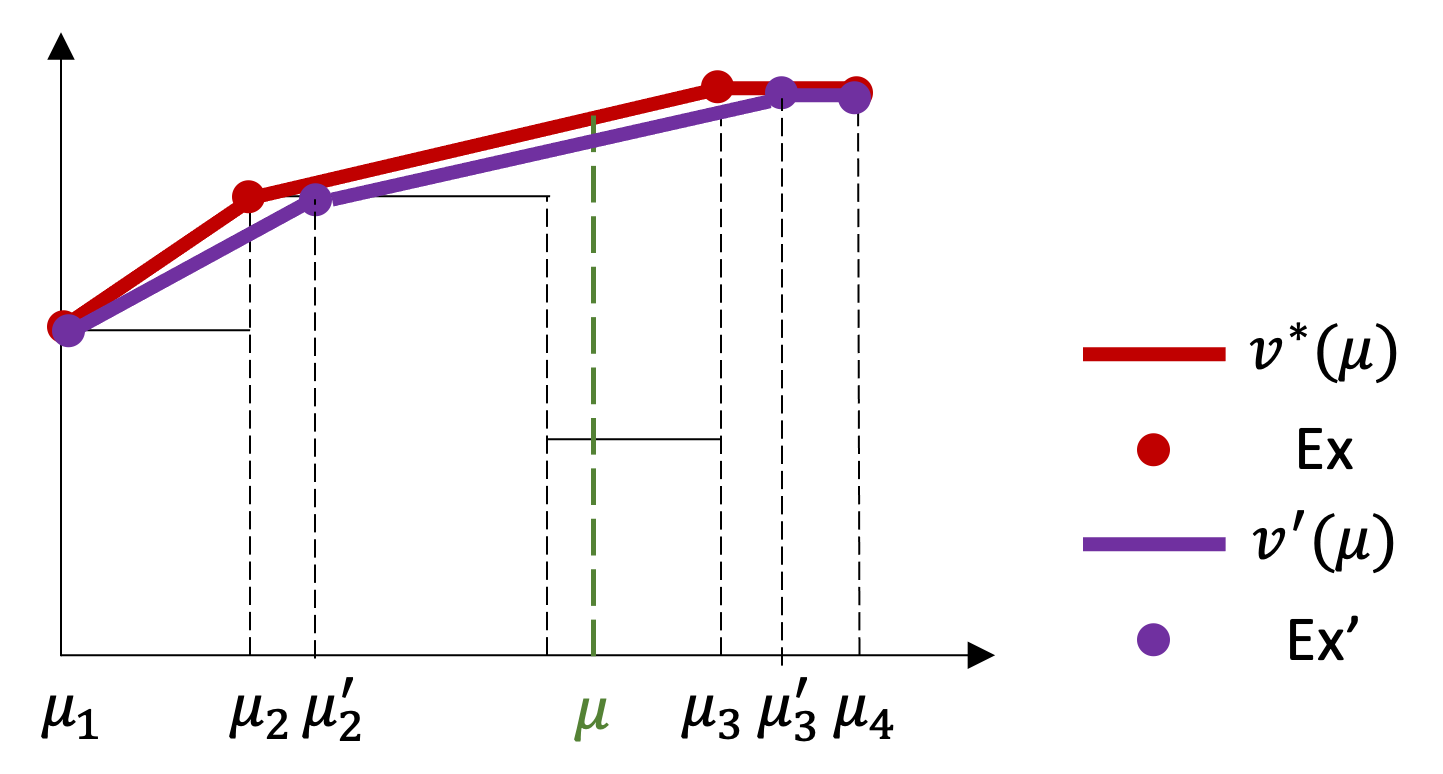}
    \caption{Illustration of $\text{Ex}, \text{Ex}', v^*(\mu), v'(\mu)$. In this figure, the optimal achievable value $v^*(\mu)$ is achieved by the convex combination of $\mu_2$ and $\mu_3$. The value achieved by our scheme $v'(\mu)$ is attained by convex combination of $\mu_2'$ and $\mu_3'$.  $v'(\mu)$ is very close to $v^*(\mu)$.}
    \label{fig:bayes2}
    \end{subfigure}
    \caption{Illustration of utilities in Bayesian Persuasion.}
\end{figure}


As depicted in Fig~\ref{fig:bayes}, for any $s\in \cS$, $u(s,\mu)$ is linear in $\mu$ with the absolute value of the slope $\abs{\partial u(s,\cdot)} \leq 1$ since the utilities are in $[0,1]$. It is easy to check that for any $\mu<\mu'\in [0,1]$, if $s$ is an optimal strategy for both $\Ber(\mu)$ and $\Ber(\mu')$, then for any $\mu''\in [\mu,\mu']$, $s$ is also an optimal strategy for $\Ber(\mu'')$.
Hence, $[0,1]$ is divided into $n$ closed intervals $(S_1,\ldots, S_{n})$ for some $n\leq |\cS|$, such that all $\mu\in S_i$ have a single shared optimal strategy, denoted by $s_i$.
\begin{restatable}{lemma}{implicationgap}\label{lmm:implication-gap-asp}
    Under Assumption~\ref{asp-bayes:interval}, we have the following observations:
    \begin{itemize}
        \item Each strategy in $\cS$ corresponds to one interval in $(S_1,\ldots, S_{n})$. In other words, we have $\cS = \{s_1, s_2,\ldots, s_n\}$ and $n=\abs{\cS}$.
        \item There exists a positive constant $C>0$ such that the length of every interval in $\{S_1,\ldots, S_{n}\}$ is lower bounded by $C$. For every interval $S_i$, for any $\mu$ inside $S_i$ (not on the edge), $s_i$ is the unique optimal strategy under prior $\mu$.
        \item There exists a positive constant $c_1>0$ such that for any two different strategies $s,s'$, the difference between the utility slopes, $\abs{\partial u(s,\cdot)-\partial u(s',\cdot)}$, is bounded below by $c_1$. 
    \end{itemize}    
\end{restatable}



Then the Agent's utility $u(s^*(\mu),\mu)$, given that he selects the optimal strategy $s^*(\mu)$, as a function of $\mu$, is taking a maximum over the set of linear functions $\{u(s,\mu)|s\in \cS\}$, as depicted in blue in Fig~\ref{fig:bayes}.
The Principal's utility $v(s^*(\mu))$ given that the Agent selects the optimal strategy $s^*(\mu)$, is a piecewise constant function since for all $\mu\in S_i$ (except for the boundary of $S_i$), $v(s^*(\mu)) = v(s_i)$.

Given any prior $\pi = \Ber(\mu)$, it is easy to see that a signal scheme induces a distribution over posteriors, $\pi(y|s)$ for all $s\in \cS$. 
The reverse is true as well: any distribution over posteriors that is consistent with our prior corresponds to a signal scheme. Given a  distribution of posteriors $\{(\tau_i,\mu_i)|i\in [n]\}$ with $\tau_i\geq 0$, $\sum_{i=1}^n \tau_i = 1$, we call the distribution Bayes-plausible if the expected posterior equals the prior, i.e., $\sum_{i}\tau_i \mu_i = \mu$.
Given a Bayes-plausible distribution of posteriors, we can recover the corresponding signal scheme  $p(s_i|y) = \frac{\tau_i\pi(y|s_i)}{\pi(y)}$ by Bayes' rule, where $s_i = s^*(\mu_i)$.
More explicitly, we have
\begin{align}
    p(s_i|y=1) = \frac{\tau_i \cdot \mu_i}{\mu}\,, \qquad p(s_i|y=0) = \frac{\tau_i \cdot (1-\mu_i)}{1-\mu}\,.\label{eq:signal}
\end{align}
Therefore, when given a prior $\mu$, selecting a signal scheme is equivalent  to selecting a Bayes-plausible distribution of posteriors. In the following, we choose a Bayes-plausible distribution of posteriors to represent a signal scheme. 

In Bayesian Persuasion, given any policy $p$ and prior $\pi=\Ber(\mu)$, the optimistic best response $a^*(p,\mu)$ for the Agent is selecting the optimal strategy $s^*(\pi(y|s))$ under the posterior $\pi(y|s)$ when observing signal $s$. Given prior $\pi=\Ber(\mu)$, the optimal achievable Principal's utility is defined as the maximum utility given that the Agent always best responds optimistically, i.e., $v^*(\mu) := \argmax_{p\in \cP}V(a^*(p,\mu), p, \mu)$.
\begin{lemma}[\cite{kamenica2011bayesian}]
    The optimal achievable Principal's utility is the concave closure of the convex hull of $(\mu,v(s^*(\mu)))$:
\begin{equation}\label{eq:opt-bayes}
    v^*(\mu) = \sup\{z|(\mu,z)\in \conv(v)\}\,,
\end{equation}
where $\conv(v)$ is the convex hull of $\{(\mu, v(s^*(\mu)))|\mu\in [0,1]\}$. 
\end{lemma}
The optimal achievable value $v^*(\mu)$ given the prior $\mu$ is depicted in red in Fig~\ref{fig:bayes}. Let 
$\textrm{Ex}= \{(\mu_{1}, v(s^*(\mu_{1}))),\ldots,(\mu_{{K}}, v(s^*(\mu_{{K}})))\}$ 
denote all extreme points of $\conv(v)$ on the concave closure, where $\mu_{1}=0$ and $\mu_{{K}} = 1$ for notation convenience.
By \cite{kamenica2011bayesian}, there exists two points in $\textrm{Ex}$ 
such that $(\mu, v^*(\mu))$ is represented as the convex combination of them. This defines a Bayes-plausible distribution of posteriors, where $\mu_{j}$ is the posterior given signal $s^*(\mu_{j})$ and the weight on $(\mu_{j}, v(s^*(\mu_{j})))$ is the probability mass assigned to $\mu_{j}$. The optimal scheme is the one which induces this distribution of posteriors. Note that all these $\mu_{j}$'s lie on the boundaries of the intervals $\{S_1,\ldots,S_n\}$.



\subsubsection{Optimal Stable Policy Oracle Construction}
Now we are ready to describe how to construct a stable policy oracle in Bayesian Persuasion based on the above optimal scheme.
The reader may notice that the above optimal scheme is not stable since each possible posterior $\mu_{j}$ lies on the edge of intervals and there will be two optimal strategies under $\mu_{j}$, which might lead to different Principal utilities. Hence, we need first to stabilize the optimal scheme. Besides, recall that the Principal's regret in Theorem~\ref{thm:stable} depends on the cardinality  $|\cP_\cO|$ of the output policy space, and so it is not enough to be able to construct near optimal stable policies --- we need to be able to construct near optimal stable policies that are always members of a small discrete set. Thus, as a second part of our construction we need to discretize the output space. We will introduce the stabilization step in the following and defer the discretization step to Appendix~\ref{app:bayes}.

\paragraph{Stabilization of the optimal scheme}
To stabilize the scheme, we need to make sure that the possible posteriors will lie inside the intervals such that each corresponds to one unique optimal Agent strategy.
For each $j\in \{2,\ldots,K-1\}$, if $s^*(\mu_j)=s_{i_j}$, 
our method will move $\mu_{j}$ into $S_{i_j}$ by $\beta$ for some $\beta>0$. Specifically, let $\mu_{j}' = \mu_{j}-\beta$ if the interval $S_{i_j}$ is below $\mu_{j}$; and $\mu_{j}' = \mu_{j}+\beta$ if the interval $S_{i_j}$ is above $\mu_{j}$. There is no need to move $\mu_1 =0$ and $\mu_K=1$ as they already correspond to a unique optimal strategy. Hence, we let $\mu_1' = \mu_1$ and $\mu_K' =\mu_K$.
As mentioned previously, the length of each interval in $\{S_1,\ldots, S_{n}\}$ is at least $C$. We set $\beta < \frac{C}{4}$ to be small enough so that $\mu_{j}'\in S_{i_j}$ and thus we have $s_{i_j}= s^*(\mu_{j}')$.
Now let $\textrm{Ex}' =\{(\mu_{1}', v(s_{i_1})),\ldots, (\mu_{K}', v(s_{i_K}))\}$ denote the modified set of extreme points and let
$$v'(\mu) = \sup\{z|(\mu,z)\in \conv(\textrm{Ex}')\}\,.$$
We illustrate $\textrm{Ex}'$ and $v'(\mu)$ in Fig~\ref{fig:bayes2}. Similar to $v^*(\mu)$, we can achieve $v'(\mu)$ by finding two points in $\textrm{Ex}'$ to represent $(\mu,v'(\mu))$ by a convex combination of them. This convex combination leads to a distribution of posteriors and thus a signal scheme. We denote this signal scheme by $p'(\mu)$.

\begin{restatable}{lemma}{lmmbayesalt}\label{lmm:bayes-alt}
    There exists a constant $c_2>0$ such that for any $\mu, \epsilon, x \in [0,1]$, $p'(\mu)$ is a $(\frac{3\beta}{C} + c_2\sqrt{\epsilon},\epsilon,x \cdot c_{1}\beta, x)$-optimal stable policy under $\mu$.  
\end{restatable}
Recall that the cardinality $|\cP_\cO|$ of the output policy space of the oracle matters (in Theorem~\ref{thm:stable}) but the output space of $p'$ could be huge. Hence we need to discretize the output space $\{p'(\mu)|\mu\in [0,1]\}$. We defer the details of discretization to Appendix~\ref{app:bayes}. The upshot of the discretization step is that together with our stabilization step, we can obtain the following theorem:
\begin{restatable}[Stable Policy Oracle for Bayesian Persuasion]{theorem}{thmbayesdiscrete}\label{thm:bayes-discrete}
     There exist positive constants $C,c_1,c_2$ such that for any $\beta\in [0,\frac{C}{4}), \epsilon,x\in [0,1]$ and any $\delta \leq \frac{\beta^2}{16}$, there exists a policy oracle $p_\delta(\cdot)$ which is
     $(\frac{3\beta}{C} + c_2 \sqrt{\epsilon} +2\sqrt{\delta},\epsilon, x \cdot c_{1}\beta/2, \max(x,\sqrt{\delta}))$-optimal stable with $|\cP_\cO|=\cO(\frac{n^2}{\delta^2})$. By combining with   Theorem~\ref{thm:stable} and setting $\epsilon=T^{-\frac{1}{5}}$, $x=\beta=\sqrt{\epsilon}$, and $\delta = \frac{\beta^2}{16}$, we can achieve Principal's regret: 
     $$\text{PR}(\sigma^\dagger, \cL,y_{1:T}) = \tilde\cO\left(T^{-\frac{1}{10}}\right)\,,$$
     when the Agent obtains swap regret $\eint = \cO(\sqrt{\abs{\cP_\cO}/T})$.
\end{restatable}

\section{The General Case}\label{sec:general}
In Section \ref{sec:stable} we solved the special case in which we have a stable policy oracle available to us, and in Section \ref{sec:oracles} we showed how to construct stable policy oracles for two important settings: linear contracting, and binary state Bayesian persuasion. 
In this section, we consider the general case, in which we cannot assume the existence of an optimal stable policy oracle. 
In Appendix~\ref{sec:imposs} we give an example of a setting in which there is no optimal stable policy  (see Proposition~\ref{lem:impossstable}) --- and so indeed, if we want to handle the general case, we need do without such oracles. 
In this case, in addition to the behavioral assumptions in Section~\ref{sec:behavior}, we propose an additional alignment assumption, following Assumption 7 of~\cite{camara2020mechanisms}.
This alignment assumption also allows us to consider a weaker ``no secret information'' assumption than Assumption~\ref{asp:no-correlation} that corresponds to assuming that the Agent's ``cross-swap-regret'' with respect to the Principal's communications (policy and recommendation) is not too negative. 

We state our result here and defer the formal definitions of the assumptions and the algorithm to Appendix~\ref{app:general}.
\begin{restatable}{theorem}{thmgeneral}
    \label{thm:general}
Recall the definition of the set of events
$$\cE_3 = \{\ind{a^*(p_0,\pi_t) = a}\}_{p_0 \in \cP_0,a\in \cA}$$
and define 
$$\cE_4 = \{p^*(\pi_t)= p, a^*(p,\pi_t) =a\}_{p\in \cP_0, a\in \cA}\,.$$
Let $\cE' = \cE_3\cup\cE_4$, the union of these events. Under Assumptions~\ref{asp:no-internal-reg} (No Contextual Swap Regret), \ref{asp:alignment} (Alignment), and \ref{asp:no-sec-info} (No Secret Information), by running the forecasting algorithm from \cite{NRRX23} for events $\cE'$ and the choice rule in Algorithm \ref{alg:general},  the Principal can achieve policy regret:
    \begin{equation*}
        \text{PR}(\sigma^\dagger, \cL,y_{1:T}) \leq \tilde \cO\left(\abs{\cY}\sqrt{\frac{\abs{\cP_0}\abs{\cA}}{T}}\right) +M_1(\eint + \eneg) + M_2.
    \end{equation*}
\end{restatable}

Recall that the forecasting algorithm of \cite{NRRX23} runs in time polynomial in $|\cY|$ and the number of events we ask for low bias on, which in this case is a set of size polynomial in the problem parameters: $|\cE'| = O(|\cP_0||\cA|)$.

\section{Comparison with \cite{camara2020mechanisms}}
We conclude with a detailed guide for the reader who wishes to compare our results to those of \cite{camara2020mechanisms}.  
The theorem given in \cite{camara2020mechanisms} to which our work is most directly comparable is their Theorem 2, which obtains vanishing policy regret while not requiring the principal to condition on aspects of the agent's learning algorithm. Their Theorem 1 is (as they write) a ``pedagogical example'' since it assumes that the Principal has oracle access to the Agent's learning algorithm, and so only applies to a single fixed learning algorithm.

Their Theorem 2 relies on their Assumptions 1 (regularity), 2 (low cost of $\epsilon$-Robustness, which is also a ``alignment''-style assumption), 4, 7 (alignment), 3 (bounded counterfactual internal regret), and 6 (lower bounded external regret). This is an analogue of our Theorem~\ref{thm:general}, which requires our Assumptions~\ref{asp:no-internal-reg} (no contextual swap regret), \ref{asp:alignment} (alignment), and \ref{asp:no-sec-info} (no negative-cross-swap regret, which we also call no secret information).
Our Assumption~\ref{asp:no-internal-reg} is an analogue of their Assumption 3, our Assumption \ref{asp:no-sec-info} is an analogue of their Assumption 6, and our Assumption~\ref{asp:alignment} is an analogue of their Assumption 7. The primary advantage of our Theorem~\ref{thm:general} compared to their Theorem 2 is that our mechanism has both running time and regret bounds that scale polynomially with the cardinality of the state space, rather than exponentially (as their theorem does).

The results in \cite{camara2020mechanisms} (both the pedagogical Theorem 1 and the main Theorem 2) also rely on their Assumption 2, which ``assumes that the optimal policy under best-case agent behavior only performs $O(\epsilon)$ better'' than under worst-case behavior. Their Assumption 2 is roughly analogous to an assumption in our language that the policy their mechanism plays is an optimal stable policy. Details aside, the primary difference here is that \cite{camara2020mechanisms} assume that the optimal robust policy played by their mechanism are optimal stable policies, whereas for us, this is not an assumption, and: we propose a general framework given access to any optimal stable policy oracle (our Theorem~\ref{thm:stable}) and explicitly show how to construct optimal stable policies for the linear contracting (our Theorem~\ref{thm:linear}, which only relies on assumptions of no contextual swap regret and no secret information) and Bayesian Persuasion settings (our Theorem~\ref{thm:bayes-discrete}). Note also that their Theorem 2 also requires Assumption 7, which assumes that the usefulness of the information to the principal is upper bounded by the usefulness of the information to the agent. We get rid this kind of alignment assumption in the linear contracting and Bayesian Persuasion settings (See Theorems~\ref{thm:linear} and \ref{thm:bayes-discrete}).

In this work, we introduce a new ``no secret information''
condition, Assumption~\ref{asp:no-correlation}, as Assumption~\ref{asp:no-sec-info} is not sufficient when we remove the alignment assumption. Hence, we propose a new version of “no secret information” (Assumption 2) and show that the combination of Assumption 1 and Assumption 2 is sufficient in linear contracting (Theorem~\ref{thm:linear}) and Bayesian Persuasion (Theorem~\ref{thm:bayes-discrete}).

\begin{restatable}[Necessity of Assumption~\ref{asp:no-correlation}, Strengthened]{proposition}{strongnecessity}\label{lem:lb-hard}
There exists a simple linear contract setting where, for any Principal mechanism $\sigma$, one of the following must hold:
\begin{itemize}
    \item No learning algorithm $\cL^{*}$ can satisfy Assumption~\ref{asp:no-internal-reg} with $\eint = o(1)$ and Assumption~\ref{asp:no-sec-info} with $\eneg = o(1)$ for all possible sequence of states $y_{1:T}\in \cY^T$.
    \item There exists a learning algorithm $\cL^{*}$ satisfying Assumption~\ref{asp:no-internal-reg} with $\eint = o(1)$ and Assumption~\ref{asp:no-sec-info} with $\eneg = o(1)$ for all possible sequence of states $y_{1:T}\in \cY^T$ and a sequence of states $\bar y_{1:T} \in \cY^T$ for which any mechanism $\sigma$ achieves non-vanishing regret for the Principal, i.e., $\text{PR}(\sigma,\cL^{*},\bar y_{1:T}) = \Omega(1)$.
\end{itemize}
\end{restatable}

\section{Discussion and Conclusion}

We have shown how to give strong \emph{policy regret} bounds for a Principal interacting with a long-lived, non-myopic Agent, in an adversarial, prior free setting. In place of common prior assumptions, we have relied on strictly weaker behavioral assumptions, in the style of \cite{camara2020mechanisms}. However, unlike \cite{camara2020mechanisms}, our mechanisms are efficient in the cardinality of the state space. Additionally, for several important special cases, including the linear contracting setting that has been focal in both the economic and computer science contract theory literature, we do not need any other assumptions (in particular avoiding the ``Alignment'' assumption of \cite{camara2020mechanisms})---which means that our setting is a strict relaxation of the common prior setting. 

In fact, our ability to avoid Alignment assumptions is not specific to linear contracting settings (or binary state Bayesian Persuasian settings) --- but is proven for any class of interactions for which we can derive algorithms implementing ``stable policy oracles''. We gave given two such examples in this paper, but surely more exist. Understanding which kinds of interactions admit stable policy oracles---and which do not---seems important to understand, towards being able to flexibly solve repeated Principal/Agent problems in an assumption minimal way.

\chapter{Eliciting User Preferences for Personalized Multi-Objective Decision Making through Comparative Feedback}\label{chap:momdp}
\section{Introduction}

Many real-world decision making problems involve optimizing over multiple  objectives. For example, when designing an investment portfolio, one’s investment strategy requires trading off maximizing
expected gain with minimizing risk.
When using Google Maps for navigation, people are concerned about various factors such as the worst and average estimated arrival time, traffic conditions, road surface conditions (e.g., whether or not it is paved), and the scenery along the way.
As the advancement of technology gives rise to personalized machine learning~\citep{mcauley2022}, in this paper, we 
design efficient algorithms for personalized multi-objective decision making. 

While prior works have concentrated on approximating the Pareto-optimal solution set\footnote{The Pareto-optimal solution set  contains a optimal personalized policy for every possible user preference.} (see \citet
{HayesRBKMRVZDHH22MORLsurvey} and \citet{roijers13} for surveys), we aim to find the optimal personalized policy for a user that reflects their unknown preferences over $k$ objectives.
Since the preferences are unknown, we need to elicit users' preferences by requesting feedback on selected policies. 
The problem of eliciting preferences has been studied in \cite{wang2022imo} using a strong query model that provides stochastic feedback on
the quality of a single policy.
In contrast, our work focuses on a more natural and intuitive query model,  comparison queries, which query the user's preference over two selected policies, e.g., `do you prefer a policy which minimizes average estimated arrival time or policy which minimizes the number of turns?'.
Our goal is to find the optimal personalized policy using as few queries as possible.
To the best of our knowledge, we are the first to 
provide algorithms with theoretical guarantees for specific personalized multi-objective decision-making via policy comparisons.


Similar to prior works on multi-objective decision making, we model the problem using a finite-horizon Markov decision process (MDP) with a $k$-dimensional reward vector, where each entry is a non-negative scalar reward representative of one of the $k$ objectives. To account for user preferences, we assume that a user is characterized by a (hidden) $k$-dimensional \textit{preference vector} with non-negative entries, and that the \textit{personalized reward} of the user for each state-action is the inner product between this preference vector and the reward vector (for this state-action pair). We also distinguish between the $k$-dimensional value of a policy, which is the expected cumulative reward when selecting actions according to the policy, and the \textit{personalized value} of a policy, which is the scalar expected cumulative personalized reward when selecting actions according to this policy.  The MDP is known to the agent and the goal is to learn an optimal policy for the personalized reward function (henceforth, the \emph{optimal personalized policy}) of a user via policy comparative feedback.

\textbf{Comparative feedback.} If people could clearly define their preferences over objectives (e.g., ``my preference vector has $3$ for the scenery objective, $2$ for the traffic objective, and $1$ for the road surface objective''), the problem would be easy---one would simply use the personalized reward function as a scalar reward function and solve for the corresponding policy. In particular, a similar problem with noisy feedback regarding the value of a single multi-objective policy (as mapping from states to actions) has been studied in \citet{wang2022imo}.
As this type of fine-grained preference feedback is difficult for users to define, especially in environments where sequential decisions are made, we restrict the agent to rely solely on comparative feedback. Comparative feedback is widely used in practice. For example, ChatGPT asks users to compare two responses to improve its performance. This approach is more intuitive for users compared to asking for numerical scores of ChatGPT responses. 


Indeed, as knowledge of the user's preference vector is sufficient to solve for their optimal personalized policy, the challenge is to learn a user's preference vector using a minimal number of easily interpretable queries.
We therefore concentrate on comparison queries. The question is then what exactly to compare? Comparing state-action pairs might not be a good option for the aforementioned tasks---what is the meaning of two single steps in a route?
Comparing single trajectories (e.g., routes in Google Maps) would not be ideal either. Consider for example two policies: one randomly generates either (personalized) GOOD or  (personalized) BAD trajectories while the other consistently generates  (personalized) MEDIOCRE trajectories. By solely comparing single trajectories without considering sets of trajectories, we cannot discern the user's preference regarding the two policies.


\textbf{Interpretable policy representation.}
Since we are interested in learning preferences via policy comparison  queries, we also suggest an alternative, more interpretable  representation of a policy. Namely, we design an algorithm that given an  explicit policy representation (as a mapping from states to distributions over actions), returns a weighted set of trajectories of size at most $k+1$, such that its expected return is identical to the value of the policy.\footnote{A return of a trajectory is the cumulative reward obtained in the trajectory. The expectation in the expected return of a weighted trajectory set is over the weights.} 
It immediately follows from our formalization that for any user, the personalized return of the weighted trajectory set and the personalized value of the policy are also identical.


In this work, we focus on answering two questions:

\vspace{-1.5mm}
 \begin{center}
    \textit{(1) How to find the optimal personalized policy by querying as few policy comparisons as possible?\\
    (2) How can we find a more interpretable representation of policies efficiently?}
\end{center}
\vspace{-1.5mm}
\textbf{Contributions.} In Section~\ref{sec:model-momdp}, we formalize the problem of eliciting user preferences and finding the optimal personalized policy via comparative feedback. 
As an alternative to an explicit policy representation, we propose a \emph{weighted trajectory set} as a more interpretable representation. 
In Section~\ref{sec:policy-level}, we provide an \textit{efficient} algorithm for finding an approximate optimal personalized policy, where the policies are given by their formal representations, thus answering (1).
In Section~\ref{sec:trajectory-level}, we design two efficient algorithms that find the weighted trajectory set representation of a policy. Combined with the algorithm in Section~\ref{sec:policy-level}, we have an algorithm for finding an approximate optimal personalized policy when policies are represented by weighted trajectory sets, thus answering (2). 
\textbf{Related Work.}
Multi-objective decision making has gained significant attention in recent years (see \citet{roijers13, HayesRBKMRVZDHH22MORLsurvey} for surveys). Prior research has explored various approaches, such as assuming linear preferences or Bayesian Settings, or finding an approximated Pareto frontier. However, incorporating comparison feedback (as was done for Multi-arm bandits or active learning, e.g., \citet{bengs2021preference} and \citet{kane2017active}) allows us a more comprehensive and systematic approach to handling different types of user preferences and provides (nearly) optimal personalized decision-making outcomes. 
We refer the reader to Appendix~\ref{sec:related_work} for additional related work.



\section{Related Work}\label{sec:related_work}
\textbf{Multi-objective sequential decision making}
There is a long history of work on multi-objective sequential decision making~\citep{roijers13}, with one key focus being the realization of efficient algorithms for approximating the Pareto front~\citep{chatterjee06,chatterjee07, Marinescu_Razak_Wilson_2017}.  Instead of finding a possibly optimal policy, we concentrate on specific user preferences and find a policy that is optimal for that specific user. Just like the ideal car for one person could be a Chevrolet Spark (small) and for another, it is a Ford Ranger (a truck). 

In the context of multi-objective RL~\citep{HayesRBKMRVZDHH22MORLsurvey}, the goal can be formulated as one of learning a policy for which the average return vector belongs to a target set (hence the term ``multi-criteria'' RL), which existing work has treated as a stochastic game~\citep{mannor01,mannor04}. Other works seek to maximize (in expectation) a scalar version of the reward that may correspond to a weighted sum of the multiple objectives~\citep{barrett08,chen19a} as we consider here, or a nonlinear function of the objectives~\citep{cheung19}. 
Multi-objective online learning has also been studied, see \citep{mannor2014approachability,blum2020online} for example. 

The parameters that define this scalarization function (e.g., the relative objective weights) are often unknown and vary with the task setting or user. In this case, preference learning~\citep{WirthRlSurvey2017} is commonly used to elicit the value of these parameters. \citet{doumpos07} describe an approach to eliciting a user's relative weighting in the context of multi-objective decision-making. \citet{bhatia20} learn preferences over multiple objectives from pairwise queries using a game-theoretic approach to identify optimal randomized policies. In the context of RL involving both scalar and vector-valued objectives, user preference queries provide an alternative to learning from demonstrations, which may be difficult for people to provide (e.g., in the case of robots with high degrees-of-freedom), or explicit reward specifications~\citep{cheng11, rothkopf2011preference, akrour2012april, wilson2012bayesian, furnkranz12, jain15, wirth2016model, christiano2017deep, WirthRlSurvey2017, ibarz18, lee21}. These works typically assume that noisy human preferences over a pair of trajectories are correlated with the difference in their
utilities 
(i.e., the reward acts as a latent term predictive of preference). Many contemporary methods estimate the latent reward by minimizing the cross-entropy loss between the reward-based predictions and the human-provided preferences (i.e., finding the reward that maximizes the likelihood of the observed preferences)~\citep{christiano2017deep, ibarz18, lee21, knox22,PacchianoDuelingRL}.

\citet{wilson2012bayesian} describe a Bayesian approach to policy learning whereby they query a user for their preference between a pair of trajectories and use these preferences to maintain a posterior distribution over the latent policy parameters. The task of choosing the most informative queries is challenging due to the continuous space of trajectories, and is generally NP-hard~\citep{ailon12}. Instead, they assume access to a distribution over trajectories that accounts for their feasibility and relevance to the target policy, and they describe two heuristic approaches to selecting trajectory queries based on this distribution. Finally, \citet{sadigh17} describe an approach to active preference-based learning in continuous state and action spaces. Integral to their work is the ability to synthesize dynamically feasible trajectory queries. \citet{biyik18} extend this approach to the batch query setting.

\textbf{Comparative feedback in other problems}
Comparative feedback has been studied in other problems in learning theory, e.g., combinatorial functions \citep{BalcanVW16}. 
One closely related problem is active ranking/learning using pairwise comparisons ~\citep{jamieson2011active,kane2017active,saha2021dueling,YonaMEG22}.
These works usually consider a given finite sample of points.
\citet{kane2017active} implies a lower bound of the number of comparisons is linear in the cardinality of the set even if the points satisfy the linear structural constraint as we assume in this work. 
In our work, the points are value vectors generated by running different policies under the same MDP and thus have a specific structure. Besides, we allow comparison of policies not in the policy set. Thus, we are able to obtain the query complexity sublinear in the number of policies.
Another related problem using comparative/preference feedback is dueling bandits that aim to learn through pairwise feedback~\citep{Ailon+14,Zoghi+14RUCB} 
(see also \citet{bengs2021preference,sui18survey} for surveys), or more generally any subsetwise feedback~\citep{Sui+17,SG18,SGwin18,Ren+18}. However, unlike dueling bandits, we consider noiseless comparative feedback.

\section{Problem Setup}\label{sec:model-momdp}

\textbf{Sequential decision model.} We consider a  Markov decision process (MDP) \emph{known}\footnote{If the MDP is unknown, one can approximate the MDP and apply the results of this work in the approximated MDP. More discussion is included in Appendix~\ref{app:discussion}.}  to the agent represented by a tuple $\langle\cS, \cA, s_0, P, R,H \rangle$, with finite state and action sets, $\cS$ and $\cA$, respectively, an initial state $s_0 \in \cS$, and finite horizon $H\in \mathbb{N}$. For example, in the Google Maps example a state is an intersection and actions are turns. The transition function $P:\cS\times \cA \mapsto \spl^\cS$ maps state-action pairs into a state probability distribution.
To model multiple objectives, the reward function $R:\cS\times \cA\mapsto [0,1]^k$ maps every state-action pair to a $k$-dimensional reward vector, where each component corresponds to one of  the $k$ objectives (e.g.,  road surface condition, worst and average estimated arrival time). 
The \textit{return} of a trajectory $\tau = (s_0,a_0,\ldots,s_{H-1},a_{H-1},s_H)$ is given by $\Phi(\tau) = \sum_{t=0}^{H-1} R(s_t,a_t)$.

A \textit{policy} $\pi$ is a mapping from  states to a distribution over actions. We denote the set of policies by $\Pi$. The \textit{value} of a policy $\pi$, denoted by $V^{\pi}$, is the expected cumulative reward obtained by executing the policy $\pi$ starting from the initial state, $s_0$. Put differently, the value of $\pi$ is $V^{\pi}= V^{\pi}(s_0)=\EEs{S_0 = s_0}{\sum_{t=0}^{H-1}  R(S_t, \pi(S_t))}\in [0,H]^k$, where $S_t$ is the random state at time step $t$ when executing $\pi$, and the expectation is over the randomness of $P$ and $\pi$. Note that $V^{\pi} = \EEs{\tau}{\Phi(\tau)}$, where $\tau =(s_0, \pi(s_0), S_1,\ldots, \pi(S_{H-1}),S_H)$ is a random trajectory generated by executing $\pi$.

We assume the existence of a ``do nothing'' action $a_0 \in \cA$, available only from the initial state $s_0$, 
that has zero reward for each objective $R(s_0,a_0)=\bZero$ and keeps the system in the initial state, i.e., $P(s_0 \mid s_0,a_0)=1$ (e.g., this action corresponds to not commuting or refusing to play in a chess game.)\footnote{{We remark that the ``do nothing'' action require domain-specific knowledge, for example in the Google maps example the ``do nothing'' will be to stay put and in the ChatGPT example the ``do nothing'' is to answer nothing.}}. We also define the (deterministic) ``do nothing'' policy $\pi_0$ that always selects action $a_0$ and has a value of $V^{\pi_0}=\bZero$. From a mathematical perspective, the assumption of ``do nothing'' ensures that $\bZero$ belongs to the value vector space $\{V^\pi \vert \pi\in \Pi\}$, which is precisely what we need.

Since the rewards are bounded between $[0,1]$, we have that $1\leq \norm{V^\pi}_2\leq \sqrt{k}H$ for every policy $\pi$. For convenience, we denote $\cv=\sqrt{k}H$. 
We denote by $d\leq k$ the rank of the space spanned by all the value vectors obtained by $\Pi$, i.e., $d:= \rank(\Span(\{V^{\pi} \vert \pi\in \Pi\}))$. 

\textbf{Linear preferences.} To incorporate personalized preferences over objectives, we assume each user is characterized by an unknown $k$-dimensional \textit{preference vector} $w^*\in \R_+^k$ with a bounded norm $1\leq \norm{w^*}_2\leq \cw$ for some  (unknown) $\cw\geq 1$. We avoid assuming that $\cw=1$ to accommodate for general linear rewards. Note that the magnitude of $w^*$ does not change the personalized optimal policy but affects the ``indistinguishability'' in the feedback model. By not normalizing $\norm{w^*}_2 =1$, we allow users to have varying levels of discernment. 
%
This preference vector encodes preferences over the multiple objectives and as a result, determines the user preferences over policies. 

Formally, for a user characterized by $w^*$,  the \textit{personalized value} of policy $\pi$ is $\inner{w^*}{V^{\pi}}\in \mathbb{R}_+$. We denote by $\pi^* := \argmax_{\pi\in \Pi} \inner{w^*}{V^{\pi}}$ and $v^* := \inner{w^*}{V^{\pi^*}}$  the optimal \textit{personalized policy} and its corresponding optimal personalized value for a user who is characterized by $w^*$. 
We remark that the ``do nothing'' policy $\pi_0$ (that always selects action $a_0$) has a value of $V^{\pi_0}=\bZero$, which implies a personalized value of $\inner{w^*}{V^{\pi_0}}=0$ for every $w^*$.
For any two policies $\pi_1$ and $\pi_2$, the user characterized by $w^*$ prefers $\pi_1$ over $\pi_2$ if  $\inner{w^*}{V^{\pi_1}-V^{\pi_2}}> 0$. 
Our goal is to find the optimal personalized policy for a given user using as few interactions with them as possible.



\textbf{Comparative feedback. }Given two policies $\pi_1,\pi_2$, the user returns $\pi_1\succ \pi_2$ whenever  
$\inner{w^*}{V^{\pi_1}-V^{\pi_2}}>\epsilon$; otherwise, the user returns ``indistinguishable'' (i.e., whenever  $\abs{\inner{w^*}{V^{\pi_1}-V^{\pi_2}}}\leq\epsilon$).
Here $\epsilon>0$ measures the precision of the comparative feedback and is small usually.
The agent can query the user about their policy preferences using two different types of policy representations:
\begin{enumerate}[nolistsep,leftmargin = *]
    \item \textit{Explicit policy representation of $\pi$}: 
    An explicit representation of policy as mapping
    ,  $\pi:\cS\rightarrow \text{Simplex}^\cA$. 
    \item \textit{Weighted trajectory set representation of $\pi$}: 
    A 
    $\kappa$-sized set of trajectory-weight pairs  $\{(p_i,\tj_i)\}_{i=1}^\kappa$ for some $\kappa\leq k+1$ such that (i) the weights $p_1,\ldots,p_\kappa$ are non-negative and sum to $1$; (ii) every  trajectory in the set is in the support\footnote{ We require the trajectories in this set to be in the support of the policy, to avoid trajectories that do not make sense, such as trajectories that ``teleport'' between different unconnected states (e.g., commuting at $3$ mph in Manhattan in one state and then at $40$ mph in New Orleans for the subsequent state).} of the policy $\pi$; and (iii) the expected return of these trajectories according to the weights is identical to the value of $\pi$, i.e., $V^{\pi} = \sum_{i=1}^{\kappa}p_i \Phi(\tj_i)$.
    Such comparison could be practical for humans. E.g., in the context of Google Maps, when the goal is to get from home to the airport, taking a specific route takes $40$ minutes $90\%$ of the time, but it can take $3$ hours in case of an accident (which happens w.p. $10\%$) vs. taking the subway which always has a duration of $1$ hour.
\end{enumerate}

In both cases, the feedback is identical and depends on the hidden precision parameter $\epsilon$.
As a result, the value of $\epsilon$ will affect the number of queries and how close the value of the resulting personalized policy is to the optimal personalized value. Alternatively, we can let the agent decide in advance on a maximal number of queries, which will affect the optimality of the returned policy.

\textbf{Technical Challenges.}
In Section~\ref{sec:policy-level}, we find an approximate optimal policy by $\cO(\log\frac{1}{\epsilon})$ queries.
To achieve this, we approach the problem in two steps. Firstly, we identify a set of linearly independent policy values, and then we estimate the preference vector $w^*$ using a linear program that incorporates comparative feedback. 
The estimation error of $w^*$ usually depends on the condition number of the linear program. 
Therefore, the main challenge we face is how to search for linear independent policy values that lead to a small condition number and providing a guarantee for this estimate.

In Section~\ref{sec:trajectory-level}, we focus on how to design \emph{efficient} algorithms to find the weighted trajectory set representation of a policy.
Initially, we employ the well-known Carathéodory's theorem, which yields an inefficient algorithm with a potentially exponential running time of $\abs{S}^H$. Our main challenge lies in developing an efficient algorithm with a running time of $\cO(\poly(H\abs{S}\abs{\cA}))$. The approach based on Carathéodory's theorem treats the return of trajectories as independent $k$-dimensional vectors, neglecting the fact that they are all generated from the same MDP. To overcome this challenge, we leverage the inherent structure of MDPs.


\section{Learning from Explicit Policies}\label{sec:policy-level}
In this section, we consider the case where the interaction with a  user is based on explicit policy comparison queries. We design an algorithm that outputs a policy being nearly optimal for this user. For multiple different users, we only need to run part of the algorithm again and again. For brevity, 
we relegate all proofs to the appendix.


If the user's preference vector $w^*$ (up to a positive scaling) is given, then one can compute the optimal policy and its personalized value efficiently, e.g., using the Finite Horizon Value Iteration algorithm. In our work, $w^*$ is unknown and we interact with the user to learn $w^*$ through comparative feedback.
Due to the structure model that is limited to meaningful  feedback only when the compared policy values differ at least $\epsilon$, the exact value of $w^*$ cannot be recovered. We proceed by providing
a high-level description of our ideas of how to estimate $w^*$. 
%
%
    
    (1) \textit{Basis policies}: We find policies $\pi_1,\ldots,\pi_d$, and their respective values,  $V^{\pi_1},\ldots,V^{\pi_d}\in[0,H]^k$, such that their  values are linearly independent and that together they span the entire space of value vectors.\footnote{Recall that $d\leq k$ is the rank of the space spanned by all the value vectors obtained by all policies.} These policies will not necessarily be personalized optimal for the current user, and instead serve only as building blocks to estimate the preference vector, $w^*$. In Section~\ref{subsec:independentvalue} we describe an algorithm that finds a set of basis policies for any given MDP.
    %
    
    (2) \textit{Basis ratios}: 
    For the basis policies, denote by $\alpha_i>0$ the ratio between the personalized value of a \textit{benchmark policy}, $\pi_1$, to the personalized value of $\pi_{i+1}$, i.e.,
    \begin{equation}\label{eq:wexact}
        \forall i\in[d-1]: \alpha_i \inner{w^*}{V^{\pi_1}}= \inner{w^*}{V^{\pi_{i+1}}}\,.
    \end{equation}
    We will estimate $\hat\alpha_i$ of $\alpha_i$ for all $i\in[d-1]$ using comparison queries.
    A detailed algorithm for estimating these ratios appears in Section~\ref{subsec:ratios}.
%
For intuition, if we obtain exact ratios $\hat{\alpha}_i=\alpha_i$ for every $i\in[d-1]$, then   we can compute the vector $\frac{w^*}{\norm{w^*}_1}$ as follows. Consider the $d-1$ equations and $d-1$ variables in Eq~\eqref{eq:wexact}. Since $d$ is the maximum number of value vectors that are linearly independent, and ${V^{\pi_{1}}},\ldots{V^{\pi_{d}}}$ form a basis, adding the equation $\norm{w}_1=1$ yields $d$ independent equations with $d$ variables, which allows us to solve for $w^*$. The details of computing an estimate of $w^*$ are described in Section~\ref{subsec:estimatew}.


\subsection{Finding a 
Basis of Policies
}\label{subsec:independentvalue}

In order to efficiently find $d$ policies with $d$ linearly independent value vectors that span the space of value vectors, one might think that selecting the $k$ policies that each optimizes one of the $k$ objectives will suffice. However, this might fail---in Appendix~\ref{app:eg-linear-dependent}, we show an instance in which these $k$ value vectors are linearly dependent even though there exist $k$ policies whose values span a space of rank $k$.
%



Moreover, our goal is to find not just any basis of policies, but a basis of policies such that (1) the personalized value of the benchmark policy $\inner{w^*}{V^{\pi_1}}$ will be large (and hence the estimation error of ratio $\alpha_i$, $\abs{\hat \alpha_i - \alpha_i}$, will be small), and (2) that the linear program generated by this basis of policies and the basis ratios will produce a good estimate of $w^*$.

\textbf{Choice of $\pi_1$.} Besides linear independence of values, another challenge is to find a basis of policies to contain a benchmark policy, $\pi_1$ (where the index $1$ is wlog) with a relatively large personalized value, $\inner{w^*}{V^{\pi_1}}$, so that $\hat \alpha_i$'s error is small (e.g., in the extreme case where $\inner{w^*}{V^{\pi_1}} = 0$, we will not be able to estimate $\alpha_i$).

For any $w\in \R^k$, we use 
$\pi^w$ denote a policy that maximizes the scalar reward $\inner{w}{R}$, i.e.,
\begin{equation}
    \pi^w = \argmax_{\pi\in \Pi} \inner{w}{V^{\pi}}\,,\label{eq:piw}
\end{equation}
and by $v^w = \inner{w}{V^{\pi^w}}$ to denote the corresponding personalized value.
Let $\be_1,\ldots,\be_k$ denote the standard basis.
To find $\pi_1$ with large personalized value $\inner{w^*}{V^{\pi_1}}$, we find policies $\pi^{\be_j}$ that maximize the $j$-th objective for every $j=1,\ldots,k$ and then query the user to compare them until we find a $\pi^{\be_j}$ with (approximately) a maximal personalized value among them. This policy will be our  benchmark policy, $\pi_1$. 
The details are described in lines~\ref{alg-line:initV1-1}--\ref{alg-line:initV1-5} of Algorithm~\ref{alg:policy-threshold-free}.

\begin{wrapfigure}[23]{L}{0.5\textwidth}
\vspace{-7mm}
    \begin{minipage}{0.5\textwidth}
    \begin{algorithm}[H]\caption{Identification of Basis Policies}\label{alg:policy-threshold-free}
    \begin{algorithmic}[1]
        \STATE initialize $\pi^{\be^*} \leftarrow \pi^{\be_1}$\label{alg-line:initV1-1} 
        \FOR{$j=2,\ldots,k$}
        \STATE compare $\pi^{\be^*}$ and $\pi^{\be_j}$
        \STATE \textbf{if} $\pi^{\be_j}\succ \pi^{\be^*}$ \textbf{then}  $\pi^{\be^*} \leftarrow \pi^{\be_j}$
        \ENDFOR
        \STATE 
        $\pi_1 \leftarrow \pi^{\be^*}$ and $u_1 \leftarrow \frac{V^{\pi^{\be^*}}}{\norm{V^{\pi^{\be^*}}}_2}$\label{alg-line:initV1-5}
        \FOR{$i=2,\ldots, k$}
            \STATE arbitrarily pick an orthonormal basis,\\ $\rho_1,\ldots, \rho_{k+1-i}$, 
             for\\ $\Span(V^{\pi_1},\ldots,V^{\pi_{i-1}})^\perp$.\label{alg-line:orthonormal-basis}
            \STATE $j_{\max} \leftarrow $\\
            $\argmax_{j\in [k+1-i]} \max(\abs{v^{\rho_j}},\abs{v^{-\rho_j}})$.\label{alg-line:largest-component}
            \IF{$\max(\abs{v^{\rho_{j_{\max}}}},\abs{v^{-\rho_{j_{\max}}}})>0$}
                \STATE $\pi_i \leftarrow\pi^{\rho_{j_{\max}}}$ \textbf{if} $\abs{v^{\rho_{j_{\max}}}}>\abs{v^{-\rho_{j_{\max}}}}$; \textbf{otherwise}  $\pi_i \leftarrow\pi^{-\rho_{j_{\max}}}$. $u_i \leftarrow \rho_{j_{\max}}$\label{alg-line: returnedu}
            \ELSE
                \STATE \textbf{return} $(\pi_1,\pi_2,\ldots)$, $(u_1,u_2,\ldots)$\label{alg-line:end}
            \ENDIF
        \ENDFOR
    \end{algorithmic}
\end{algorithm}
    \end{minipage}
\end{wrapfigure}

\textbf{Choice of $\pi_2,\ldots,\pi_d$. }After finding $\pi_1$, we next search the remaining $d-1$ polices $\pi_2,\ldots,\pi_d$ sequentially (lines \ref{alg-line:orthonormal-basis}--\ref{alg-line:end} of Algorithm~\ref{alg:policy-threshold-free}).
For $i=2,\ldots, d$, we find a direction $u_i$ such that (i) the vector $u_i$ is orthogonal to the space of current value vectors $\Span(V^{\pi_1},\ldots,V^{\pi_{i-1}})$, and (ii) there exists a policy $\pi_i$ such that $V^{\pi_i}$ has a significant component in the direction of $u_i$.
Condition (i) is used to guarantee that the policy $\pi_i$ has a value vector linearly independent of $\Span(V^{\pi_1},\ldots,V^{\pi_{i-1}})$.
Condition (ii) is used to cope with the error caused by inaccurate approximation of the ratios $\hat \alpha_i$.
Intuitively, when $\norm{\alpha_{i} V^{\pi_1}- V^{\pi_{i+1}}}_2\ll \epsilon$, the angle between $\hat\alpha_{i} V^{\pi_1}- V^{\pi_{i+1}}$ and $\alpha_{i} V^{\pi_1}- V^{\pi_{i+1}}$ could be very large, which results in an inaccurate estimate of $w^*$ in the direction of $\alpha_i V^{\pi_1}- V^{\pi_{i+1}}$.
For example, if $V^{\pi_1} = \be_1$ and $V^{\pi_i} = \be_1 + \frac{1}{w^*_i}\epsilon \be_i$ for $i=2,\ldots, k$, then $\pi_1, \pi_i$ are ``indistinguishable'' and the estimate ratio $\hat\alpha_{i-1}$ can be $1$. Then the estimate of $w^*$ by solving linear equations in Eq~\eqref{eq:wexact} is $(1,0,\ldots,0)$, which could be far from the true $w^*$.
Finding $u_i$'s in which policy values have a large component can help with this problem.

Algorithm~\ref{alg:policy-threshold-free} provides a more detailed description of this procedure.
Note that if there are $n$ different users, we will run Algorithm 1 at most $k$ times instead of $n$ times. The reason is that Algorithm 1 only utilizes preference comparisons while searching for $\pi_1$ (lines~\ref{alg-line:initV1-1}-\ref{alg-line:initV1-5}), and not for $\pi_2, \ldots, \pi_k$ (which contributes to the $k^2$ factor in computational complexity). As there are at most $k$ candidates for $\pi_1$, namely $\pi^{e_1}, \ldots, \pi^{e_k}$, we execute lines~\ref{alg-line:initV1-1}-\ref{alg-line:initV1-5} of Algorithm~\ref{alg:policy-threshold-free} for $n$ rounds and lines~\ref{alg-line:orthonormal-basis}-\ref{alg-line:end} for only $k$ rounds. 


\subsection{Computation of Basis Ratios}\label{subsec:ratios}
As we have mentioned before, comparing basis policies alone does not allow for the exact computation of the $\alpha_i$ ratios as comparing $\pi_1,\pi_i$ can only reveal which is better but not how much.
To this end, we will use the ``do nothing'' policy to approximate every ratio $\alpha_i$ up to some additive error $\abs{\hat \alpha_i - \alpha_i}$ using binary search over the parameter $\hat{\alpha}_i\in [0,C_\alpha]$ for some $C_\alpha\geq 1$ (to be determined later) and comparison queries of policy $\pi_{i+1}$ with policy $\hat{\alpha}_i\pi_1 + (1-\hat{\alpha}_i)\pi_0$ if $\hat \alpha_i\leq 1$ (or comparing $\pi_1$ and $\frac{1}{\hat \alpha_i}\pi_{i+1} + (1-\frac{1}{\hat \alpha_i})\pi_0$ instead if $\hat \alpha_i>1$).\footnote{We write $\hat{\alpha}_i\pi_1+ (1-\hat{\alpha}_i)\pi_0$ to indicate that $\pi_1$ is used with probability $\hat{\alpha}_i$, and that $\pi_0$ is used with probability $1-\hat{\alpha}_i$.}  
Notice that the personalized value of $\hat{\alpha}_i\pi_1 + (1-\hat{\alpha}_i)\pi_0$ is identical to the personalized value of $\pi_1$ multiplied by $\hat{\alpha}_i$. We stop once $\hat{\alpha}_i$ is such that  the user returns ``indistinguishable''.
    Once we stop, the two policies have roughly the same personalized value,
    \begin{align}
        \text{if }\hat \alpha_i \leq 1, \abs{\hat\alpha_i\inner{w^*}{V^{\pi_1}}-\inner{w^*}{V^{\pi_{i+1}}}}\leq \epsilon\,;
        \;\; \text{if }\hat \alpha_i > 1, \abs{\inner{w^*}{V^{\pi_1}}-\frac{1}{\hat\alpha_i}\inner{w^*}{V^{\pi_{i+1}}}}\leq \epsilon.
        \label{eq:stopcond}
    \end{align}
     Eq~\eqref{eq:wexact} combined with the above inequality implies that 
    $\abs{\hat \alpha_i - \alpha_i}\inner{w^*}{V^{\pi_1}} \leq \calpha\epsilon$. Thus, the approximation error of each ratio is bounded by $\abs{\hat \alpha_i -\alpha_i} \leq \frac{\calpha \epsilon}{\inner{w^*}{V^{\pi_1}}}$.
    To make sure the procedure will terminate, we need to set $\calpha \geq \frac{v^*}{\inner{w^*}{V^{\pi_1}}}$ since $\alpha_i$'s must lie in the interval $[0,\frac{v^*}{\inner{w^*}{V^{\pi_1}}}]$.
    Upon stopping binary search once Eq~\eqref{eq:stopcond} holds, it takes at most $\cO(d\log(\calpha \inner{w^*}{V^{\pi_1}}/\epsilon))$  comparison queries to estimate all the $\alpha_i$'s.

Due to the carefully picked $\pi_1$ in Algorithm~\ref{alg:policy-threshold-free}, we can upper bound $\frac{v^*}{\inner{w^*}{V^{\pi_1}}}$ by $2k$ and derive an upper bound for $\abs{\hat \alpha_i - \alpha_i}$ by selecting  $\calpha = 2k$.
\begin{restatable}{lemma}{lmmVinit}\label{lmm:V1}
When $\epsilon \leq \frac{v^*}{2k}$, we have $\frac{v^*}{\inner{w^*}{V^{\pi_1}}} \leq  2k$
, and 
$\abs{\hat \alpha_i - \alpha_i}\leq \frac{4k^2\epsilon}{v^*}$ for every $i\in [d]$.
\end{restatable}
In what follows we set $\calpha = 2k$. The pseudo code of the above process of estimating $\alpha_i$'s is deferred to Algorithm~\ref{alg:alpha} in Appendix~\ref{app:bin-search}. 

\subsection{Preference Approximation and Personalized  Policy}\label{subsec:estimatew}
We move on to present an algorithm that estimates $w^*$ and calculates a nearly optimal personalized policy.
Given the $\pi_i$'s returned by Algorithm~\ref{alg:policy-threshold-free} and the $\hat \alpha_i$'s returned by Algorithm~\ref{alg:alpha}, consider a matrix $\hat A \in \R^{d\times k}$ with $1$st row $V^{\pi_1\top}$ and the $i$th row  $(\hat \alpha_{i-1} V^{\pi_1} -V^{\pi_i})^\top$ for every $i=2,\ldots,d$.
%
%
%
%
Let $\hat w$ be a solution to $\hat A x = \be_1$. 
We will show that $\hat w$ is a good estimate of $w':=\frac{w^*}{\inner{w^*}{V^{\pi_1}}}$ and that $\pi^{\hat w}$ is a nearly optimal personalized policy.
In particular, when $\epsilon$ is small, we have $\abs{\inner{\hat w}{V^\pi} - \inner{w'}{ V^\pi}} = \cO(\epsilon^\frac{1}{3})$ for every policy $\pi$.
Putting this together, we derive the following theorem.
\begin{restatable}{theorem}{thmwithouttau}\label{thm:est-without-tau}
    Consider the algorithm of computing $\hat A$
    and any solution $\hat w$ to $\hat A x = \be_1$ and outputting the policy $\pi^{\hat w} = \argmax_{\pi\in \Pi} \inner{\hat w}{V^{\pi}}$, which is the optimal personalized policy for preference vector $\hat w$.
    Then the output policy $\pi^{\hat w}$ satisfying that 
    $v^*-\inner{w^*}{V^{\pi^{\hat w}}}\leq   \cO\left(\left(\sqrt{k} +1\right)^{d+ \frac{14}{3}}  \epsilon^\frac{1}{3}\right)$
     using $\cO(k\log(k/\epsilon))$ comparison queries.
\end{restatable}
%
%

\textbf{Computation Complexity} We remark that Algorithm~\ref{alg:policy-threshold-free} solves Eq~\eqref{eq:piw} for the optimal policy in scalar reward MDP at most $\cO(k^2)$ times. Using, e.g., Finite Horizon Value iteration to solve for the optimal policy takes $\cO(H|\cS|^2|\cA|)$ steps. However, while the time complexity it takes to return the optimal policy for a single user is
$\cO(k^2H|\cS|^2 |\cA|+ k\log(\frac{k }{\epsilon}))$,
considering $n$ different users rather than one results in overall time complexity of 
$\cO((k^3+n)H|\cS|^2 |\cA|+nk\log(\frac{k}{\epsilon}))$.

\textbf{Proof Technique.} 
The standard technique typically starts by deriving an upper bound on $\norm{\hat w - w^*}_2$ and then uses this bound to upper bound $\sup_\pi \abs{\inner{\hat w }{V^\pi} - \inner{w^*}{V^\pi}}$ as $\cv \norm{\hat w - w^*}_2$. However, this method fails to achieve a non-vacuous bound in case there are two basis policies that are close to each other. For instance, consider the returned basis policy values: $V^{\pi_1} =(1,0,0)$, $V^{\pi_2} =(1,1,0)$, and $V^{\pi_3} = (1, 0,\eta)$ for some $\eta>0$. When $\eta$ is extremely small, the estimate $\hat w$ becomes highly inaccurate in the direction of $(0,0,1)$, leading to a large $\norm{\hat w-w^*}_2$. Even in such cases, we can still obtain a non-vacuous guarantee since the selection of $V^{\pi_3}$ (line~\ref{alg-line: returnedu} of Algorithm~\ref{alg:policy-threshold-free}) implies that no policy in $\Pi$ exhibits a larger component in the direction of $(0,0,1)$ than $\pi_3$.

\textbf{Proof Sketch.} The analysis of Theorem~\ref{thm:est-without-tau} has two parts.
First, as mentioned in Sec~\ref{subsec:independentvalue}, when $\norm{\alpha_{i} V^{\pi_1}- V^{\pi_{i+1}}}_2\ll \epsilon$, the error of $\hat\alpha_{i+1}$ can lead to inaccurate estimate of $w^*$ in direction $\alpha_i V^{\pi_1}- V^{\pi_{i+1}}$.
Thus, we consider another estimate of $w^*$ based only on some $\pi_{i+1}$'s with a relatively large $\norm{\alpha_{i} V^{\pi_1}- V^{\pi_{i+1}}}_2$.
In particular, for any $\delta>0$, let $d_\delta := \min_{i\geq 2: \max(\abs{v^{u_i}},\abs{v^{-u_i}}) \leq \delta} i-1$.
That is to say, for $i=2,\ldots,d_\delta$, the policy $\pi_i$ satisfies $\inner{u_i}{V^{\pi_i}}>\delta$ and for any policy $\pi$, we have $\inner{u_{d_\delta+1}}{V^\pi} \leq \delta$.
Then, for any policy $\pi$ and any unit vector $\xi \in \Span(V^{\pi_1},\ldots,V^{\pi_{d_\delta}})^\perp$, we have $\inner{\xi}{V^\pi} \leq \sqrt{k} \delta$.
This is because at round ${d_\delta+1}$, we pick an orthonormal basis $\rho_1,\ldots,\rho_{k-{d_\delta}}$ of $\Span(V^{\pi_1},\ldots,V^{\pi_{d_\delta}})^\perp$  (line~\ref{alg-line:orthonormal-basis} in Algorithm~\ref{alg:policy-threshold-free}) and pick $u_{d_\delta+1}$ to be the one in which there exists a policy with the largest component as described in line~\ref{alg-line:largest-component}.
Hence, $\abs{\inner{\rho_j}{V^\pi}}\leq \delta$ for all $j\in [k-{d_\delta}]$.
Then, we have $\inner{\xi}{V^\pi}= \sum_{j=1}^{k-{d_\delta}} \inner{\xi}{\rho_j}\inner{\rho_j}{V^\pi} \leq \sqrt{k}\delta$ by Cauchy-Schwarz inequality.
Let $\hat A^{(\delta)} \in \R^{d_\delta \times k}$ be the
sub-matrix comprised of the first $d_\delta$ rows of $\hat A$.
Then we consider an alternative estimate $\hatwd =\argmin_{x:\hat A^{(\delta)} x =\be_1}\norm{x}_2$, the minimum norm solution of $x$ to $\hat A^{(\delta)} x=\be_1$. We upper bound $\sup_\pi \abs{\inner{\hat w^{(\delta)}}{V^\pi}-\inner{w' }{V^\pi}}$ in Lemma~\ref{lmm:gap-hatw-w} and $\sup_\pi \abs{\inner{\hat w }{V^\pi} - \inner{\hat w^{(\delta)}}{V^\pi}}$ in Lemma~\ref{lmm:est-without-tau}. Then we are done with the proof of Theorem~\ref{thm:est-without-tau}.

%
%
%
%
\begin{restatable}{lemma}{lmmhatww}\label{lmm:gap-hatw-w}
    If $\abs{\hat \alpha_i-\alpha_i}\leq \ealpha$ and $\alpha_i\leq \calpha$ for all $i\in [d-1]$, for every $\delta \geq 4 \calpha^{\frac{2}{3}}\cv d^{\frac{1}{3}}\ealpha^\frac{1}{3}$, 
    we have
    $\abs{\inner{\hat w ^{(\delta)}}{V^\pi} - \inner{w'}{ V^\pi}}\leq \cO( \frac{\calpha\cv^4d_\delta^{\frac{3}{2}}\norm{w'}_2^2\ealpha }{\delta^2}+\sqrt{k} \delta \norm{w'}_2)$ for all $\pi$, where $w'=\frac{w^*}{\inner{w^*}{V^{\pi_1}}}$.
\end{restatable}
Since we only remove the rows in $\hat A$ corresponding to $u_i$'s in the subspace where no policy's value has a large component, $\hat w$ and $\hat w^{(\delta)}$ are close in terms of $\sup_\pi \abs{\inner{\hat w }{V^\pi} - \inner{\hat w^{(\delta)}}{V^\pi}}$.
\begin{restatable}{lemma}{lmmwithouttau}\label{lmm:est-without-tau}
     If $\abs{\hat \alpha_i-\alpha_i}\leq \ealpha$ and $\alpha_i\leq \calpha$ for all $i\in [d-1]$, for every policy $\pi$ and every $\delta \geq 4 \calpha^{\frac{2}{3}}\cv d^{\frac{1}{3}}\ealpha^\frac{1}{3}$, we have 
   $\abs{\hat w  \cdot V^\pi - \hat w^{(\delta)} \cdot V^\pi}\leq \cO((\sqrt{k} +1)^{d-d_\delta} \calpha\epsilon^{(\delta)})\,,$
    where $\epsilon^{(\delta)} = \frac{\calpha\cv^4d_\delta^{\frac{3}{2}}\norm{w'}_2^2\ealpha }{\delta^2}+\sqrt{k} \delta \norm{w'}_2$ is the upper bound in Lemma~\ref{lmm:gap-hatw-w}.
\end{restatable}
%
%
Note that the result in Theorem~\ref{thm:est-without-tau} has a factor of $k^{\frac{d}{2}}$, which is exponential in $d$.
Usually, we consider the case where $k=\cO(1)$ is small and thus $k^d = \cO(1)$ is small. 
We get rid of the exponential dependence on $d$ by applying $\hatwd$ to estimate $w^*$ directly,
which requires us to set the value of $\delta$ beforehand. 
The following theorem follows directly by assigning the optimal value for $\delta$ in  Lemma~\ref{lmm:gap-hatw-w}.


\begin{restatable}{theorem}{thmestwithtau}\label{thm:est-with-threshold}
Consider the algorithm of computing $\hat A$
and any solution $\hatwd$ to $\hatAd x = \be_1$ for $\delta = k^\frac{5}{3}\epsilon^\frac{1}{3}$ and outputting the policy $\pi^{\hatwd} = \argmax_{\pi\in \Pi} \inner{\hatwd}{V^{\pi}}$.
    Then the policy $\pi^{\hatwd}$ satisfies that $v^*-\inner{w^*}{V^{\pi^{\hatwd}}}\leq  \cO\left(k^\frac{13}{6} \epsilon^\frac{1}{3}\right)\,.$
\end{restatable}
Notice that the algorithm in Theorem~\ref{thm:est-with-threshold} needs to set the hyperparameter $\delta$ beforehand while we don't have to set any hyperparameter in Theorem~\ref{thm:est-without-tau}. The improper value of $\delta$ could degrade the performance of the algorithm. 
But we can approximately estimate $\epsilon$ by binary searching $\eta\in [0,1]$ and comparing $\pi_1$ against the scaled version of itself $(1-\eta)  \pi_1$ until we find an $\eta$ such that the user cannot distinguish between $\pi_1$ and $(1-\eta) \pi_1$. Then we can obtain an estimate of $\epsilon$ and use the estimate to set the hyperparameter.

We remark that though we think of $k$ as a small number, it is unclear whether the dependency on $\epsilon$ in Theorems~\ref{thm:est-without-tau} and \ref{thm:est-with-threshold} is optimal.
The tight dependency on $\epsilon$ is left as an open problem. We briefly discuss a potential direction to improve this bound in Appendix~\ref{app:eps-dependence}.

\section{Learning from Representative Sets}\label{sec:trajectory-level}
In the last section, we represented policies using their explicit form as state-action mappings. However, such a representation could be challenging for users to interpret. For example, how safe is a car described by a list of $|\cS|$ states and actions such as ``turning left''? 
In this section, we design algorithms that return a more interpretable policy representation---a weighted trajectory set.

Recall the definition in Section~\ref{sec:model-momdp}, a weighted trajectory set is a small set of  trajectories from the support of the policy and corresponding weights, with the property that the expected return of the trajectories in the set (according to the weights) is \textbf{exactly} the value of the policy (henceforth, the \textit{exact value property}).\footnote{Without asking for the exact value property, one could simply return a sample of  $\cO(\log k/(\epsilon')^2)$ trajectories from the policy and uniform weights. With high probability, the expected return of every objective is $\epsilon'$-far from its expected  value. The problem is that this set does not necessarily capture rare events. For example, if the probability of a crash for any car is between $(\epsilon')^4$ and $(\epsilon')^2$, depending on the policy, users that only  care about safety (i.e., no crashes) are unlikely to observe any ``unsafe'' trajectories at all, in which case we would miss valuable feedback.}
As these sets preserve all the information regarding the multi-objective values of policies, they can be used as policy representations in policy comparison queries of Algorithm~\ref{alg:alpha} without compromising on feedback quality. Thus, using these representations obtain the same optimality guarantees regarding the returned policy in Section~\ref{sec:policy-level} (but would require extra computation time to calculate the sets).  



There are two key observations on which the  algorithms in this section are based: 

(1) Each policy $\pi$ induces a distribution over trajectories. Let $q^\pi(\tau)$ denote the probability that a trajectory $\tau$ is sampled when selecting actions according to $\pi$. The expected return of all trajectories under $q^\pi$ is identical to the value of the policy, i.e., 
    $V^\pi=\sum_\tau q^\pi(\tau)\Phi(\tau).$ 
In particular, the value of a policy is a convex combination of the returns of the trajectories in its support. 
However, we avoid using this convex combination to represent a policy since the number of trajectories in the support of a policy could be exponential in the number of states and actions.

(2) The existence of a small weighted trajectory set is implied by Carathéodory's theorem. Namely, since the value of a policy is in particular a convex combination of the returns of the trajectories in its support, Carathéodory's theorem implies that there exist $k+1$ trajectories in the support of the policy and weights for them such that a convex combination of their returns is the value of the policy. Such a $(k+1)$-sized set will be the output of our algorithms.


We can apply the idea behind Carathéodory's theorem proof to compress trajectories as follows.
For any $(k+2)$-sized set of $k$-dimensional vectors  $\{\mu_1,\ldots,\mu_{k+2}\}$, for any convex combination of them $\mu = \sum_{i=1}^{k+2}p_i \mu_i$, we can always find a $(k+1)$-sized subset such that $\mu$ can be represented as the convex combination of the subset by solving a linear equation. 
Given an input of a probability distribution $p$ over a set of $k$-dimensional vectors,  $M$, we pick $k+2$ vectors from $M$, reduce at least one of them through the above procedure. We repeat this step until we are left with at most $k+1$ vectors.
We refer to this algorithm as C4 (Compress  Convex Combination using Carathéodory's theorem).
The pseudocode is described in Algorithm~\ref{alg:Caratheodory}, which is deferred to Appendix~\ref{app:Caratheodory} due to space limit. 
The construction of the algorithm implies the following lemma immediately.
%
%

%
%
\begin{restatable}{lemma}{cara}\label{thm:cara}
    Given a set of $k$-dimensional vectors $M\subset \R^k$ and a distribution $p$ over $M$, $\text{C4}(M,p)$ outputs $M'\subset M$ with $\abs{M'}\leq k+1$ and a distribution $q \in \spl^{M'}$ satisfying that $\EEs{\mu\sim q}{\mu} = \EEs{\mu\sim p}{\mu}$ in time $\cO(\abs{M}k^3)$. 
\end{restatable}
%
%
So now we know how to compress a set of trajectories to the desired size. The main challenge is how to do it \textbf{efficiently} (in time $\cO(\poly(H\abs{S}\abs{\cA}))$). Namely, since the set of all  trajectory returns from the support of the policy could be of size $\Omega(\abs{\cS}^H)$, using it as input to C4 Algorithm is inefficient.
Instead, we will only use C4 as a subroutine when the number of trajectories is small.

We propose two efficient approaches for  finding weighted trajectory representations.
Both approaches take advantage of the property that all trajectories are generated from the same policy on the same MDP. 
First, we start with a small set of trajectories of length of $1$, expand them, compress them, and repeat until we get the set of trajectory lengths of $H$. 
The other is based on the construction of a layer graph where a policy corresponds to a flow in this graph and we show that finding representative trajectories is equivalent to flow decomposition. 

In the next subsection, we will describe the expanding and compressing approach and defer the flow decomposition based approach to Appendix~\ref{app:approach-flow-based} due to space considerations. 
We remark that the flow decomposition approach has a running time of $\cO(H^2\abs{\cS}^2 +k^3H\abs{\cS}^2 )$ (see appendix for details), which underperforms the expanding and compressing approach (see Theorem~\ref{thm:traj-compression}) whenever $|\cS|H + |\cS|k^3 = \omega(k^4 + k|\cS|)$. 
For presentation purposes, in the following, we only consider the deterministic policies. Our techniques can be easily extended to random policies.\footnote{In Section~\ref{sec:policy-level}, we consider a special type of random policy that is a mixed strategy of a deterministic policy (the output from an algorithm that solves for the optimal policy for an MDP with scalar reward) with the ``do nothing'' policy. 
For this specific random policy, we can find weighted trajectory representations for both policies and then apply Algorithm~\ref{alg:Caratheodory} to compress the representation.}


\textbf{Expanding and Compressing Approach.} 

The basic idea is to find $k+1$ trajectories of length $1$ to represent $V^\pi$ first and then increase the length of the trajectories without increasing the number of trajectories.
For policy $\pi$,
let $V^{\pi}(s,h)= \EEs{S_0 = s}{\sum_{t=0}^{h-1} R(S_t, \pi(S_t))}$ be the value of $\pi$ with initial state $S_0 = s$ and time horizon $h$.
Since we study the representation for a fixed policy $\pi$ in this section, we slightly abuse the notation and represent a trajectory by $\tau^\pi = (s,s_1,\ldots,s_H)$.
We denote the state of trajectory $\tau$ at time $t$ as $s^\tau_t =s_t$.
For a trajectory prefix $\tj = (s, s_1,\ldots,s_h)$ of $\pi$ with initial state $s$ and $h\leq H$ subsequent states, the return of $\tau$ is $\Phi(\tj) = R(s, \pi(s))+\sum_{t=1}^{h-1} R(s_t,\pi(s_t))$.
Let $J(\tau)$ be the expected return of trajectories (of length $H$) with the prefix being $\tau$, i.e.,
\[J(\tau):=\Phi(\tau) + V(s^\tau_h, H-h)\,.\]

For any $s\in \cS$, let $\tau \append s$ denote the trajectory of appending $s$ to $\tau$.
We can solve $V^\pi(s,h)$ for all $s\in \cS, h\in [H]$ by dynamic programming in time $\cO(kH\abs{\cS}^2)$.
Specifically, according to definition, we have $V^\pi(s,1) = R(s, \pi(s))$ and 
\begin{equation}
    V^\pi(s,h+1) = R(s,\pi(s)) +  \sum_{s'\in \cS} \pssp V^\pi(s',h)\,.\label{eq:stepinduction}
\end{equation}
Thus, we can represent $V^\pi$ by
\begin{align*}
    V^\pi =& R(s_0,\pi(s_0)) +\sum_{s\in \cS} P(s|s_0,\pi(s_0)) V^\pi(s,H-1) = \sum_{s\in \cS} P(s|s_0,\pi(s_0)) J(s_0,s)\,.
\end{align*}
By applying C4, we can find a set of representative trajectories of length $1$, $\Qsup{1}\subset \{(s_0,s)|s\in \cS\}$, with $\abs{\Qsup{1}}\leq k+1$ and weights $\betasup{1}\in \spl^{\Qsup{1}}$ such that
\begin{align}
    V^\pi  = \sum_{\tau\in \Qsup{1}}\betasup{1}(\tau) J(\tau)\,.\label{eq:Q1}
\end{align}
Supposing that we are given a set of trajectories $\Qsup{t}$ of length $t$ with weights $\betasup{t}$ such that $V^\pi =\sum_{\tau\in \Qsup{t}} \betasup{t}(\tau)J(\tau)$, we can first increase the length of trajectories by $1$ through Eq~\eqref{eq:stepinduction} and obtain a subset of $\{\tau \append s|\tau\in \Qsup{t},s\in \cS\}$, in which the trajectories are of length $t+1$.
Specifically, we have
\begin{equation}
    V^\pi =\sum_{\tau\in \Qsup{t}, s\in \cS} \betasup{t}(\tau) P(s \vert s^\tau_t, \pi(s^\tau_t))J(\tau\append s)\,.\label{eq:expand}
\end{equation}
Then we would like to compress the above convex combination through C4 as we want to keep track of at most $k+1$ trajectories of length $t+1$ due to the computing time.
More formally, let $J_{\Qsup{t}} := \{J(\tau\append s)|\tau\in \Qsup{t}, s\in \cS\}$ be the set of expected returns and $p_{\Qsup{t},\betasup{t}} \in \spl^{\Qsup{t}\times \cS}$ with $p_{\Qsup{t},\betasup{t}}(\tau\append s) = \betasup{t}(\tau) P(s \vert s^\tau_t, \pi(s^\tau_t))$ be the weights appearing in Eq~\eqref{eq:expand}.
Here $p_{\Qsup{t},\betasup{t}}$ defines a distribution over $J_{\Qsup{t}}$ with the probability of drawing $J(\tau\append s)$ being $p_{\Qsup{t},\betasup{t}}(\tau\append s)$.
Then we can apply C4 over $(J_{\Qsup{t}},p_{\Qsup{t},\betasup{t}})$ and compress the representative trajectories $\{\tau\append s|\tau\in \Qsup{t},s\in \cS\}$.
We start with trajectories of length $1$ and repeat the process of expanding and compressing until we get trajectories of length $H$.
The details are described in Algorithm~\ref{alg:trajcompression}.
\begin{algorithm}[H]\caption{Expanding and compressing trajectories}\label{alg:trajcompression}
\begin{algorithmic}[1]
\STATE compute $V^\pi(s,h)$ for all $s\in \cS, h\in [H]$ by dynamic programming according to Eq~\eqref{eq:stepinduction}
\STATE $\Qsup{0} = \{(s_0)\}$ and $\betasup{0}(s_0) = 1$
\FOR{$t=0,\ldots,H-1$}
\STATE $J_{\Qsup{t}} \leftarrow \{J(\tau\append s)|\tau\in \Qsup{t}, s\in \cS\}$ and $p_{\Qsup{t},\betasup{t}}(\tau\append s) \leftarrow \betasup{t}(\tau) P(s \vert s^\tau_t, \pi(s^\tau_t))$ for $\tau\in \Qsup{t}, s\in \cS$ \COMMENT{expanding step}
\STATE  $(\Jsup{t+1},\betasup{t+1}) \leftarrow \text{C4}(J_{\Qsup{t}},p_{\Qsup{t},\betasup{t}})$\label{alg-line:Qtplus1}
and $\Qsup{t+1} \leftarrow \{\tau| J(\tau)\in \Jsup{t+1}\}$\COMMENT{compressing step}
\ENDFOR
\STATE output $\Qsup{H}$ and $\betasup{H}$
\end{algorithmic}
\end{algorithm}
\vspace{-3pt}

\begin{restatable}{theorem}{trajcompress}\label{thm:traj-compression}
Algorithm~\ref{alg:trajcompression} outputs $\Qsup{H}$ and $\betasup{H}$ satisfying that $\abs{\Qsup{H}}\leq k+1$ and $\sum_{\tau\in \Qsup{H}} \betasup{H}(\tau) \Phi(\tau)=V^\pi$ in time $\cO(k^4H\abs{\cS} + kH\abs{\cS}^2)$.
\end{restatable}
The proof of Theorem~\ref{thm:traj-compression} follows immediately from the construction of the algorithm.
According to Eq~\eqref{eq:Q1}, we have
$V^\pi  = \sum_{\tau\in \Qsup{1}}\betasup{1}(\tau) J(\tau)$.
Then we can show that the output of Algorithm~\ref{alg:trajcompression} is a valid weighted trajectory set by induction on the length of representative trajectories.
C4 guarantees that $\abs{\Qsup{t}} \leq k+1$ for all $t=1,\ldots,H$, and thus, we only keep track of at most $k+1$ trajectories at each step and achieve the computation guarantee in the theorem.
Combined with Theorem~\ref{thm:est-without-tau}, we derive the following Corollary.
\begin{corollary}
Running the algorithm in Theorem~\ref{thm:est-without-tau} with weighted trajectory set representation returned by Algorithm~\ref{alg:trajcompression} gives us the same guarantee as that of Theorem~\ref{thm:est-without-tau} in time $\cO(k^2  H|\cS|^2 |\cA|+ (k^5H\abs{\cS} + k^2H\abs{\cS}^2)\log(\frac{k }{\epsilon}))$.
\end{corollary}

\section{Discussion}\label{app:discussion}
\vspace{-3pt}
In this paper, we designed efficient algorithms for learning users' preferences over multiple objectives from comparative feedback. The efficiency is expressed in both the running time and number of queries (both polynomial in $H, \abs{\cS}, \abs{\cA}, k$ and logarithmic in $1/\epsilon$).
The learned preferences of a user can then be used to reduce the problem of finding a personalized optimal policy for this user to a (finite horizon) single scalar reward MDP, a problem with a known efficient solution. 
As we have focused on minimizing the policy comparison queries, our algorithms are based on polynomial time pre-processing calculations that save valuable comparison time for users.

The results in Section~\ref{sec:policy-level} are of independent interest and can be applied to a more general learning setting, where for some unknown linear parameter $w^*$, given a set of points $X$ and access to comparison queries of any two points, the goal is to learn $\argmax_{x\in X} \inner{w^*}{x}$.
E.g., in personalized recommendations for coffee beans in terms of the coffee profile described by the coffee suppliers (body, aroma, crema, roast level,...), while users could fail to describe their optimal 
coffee beans profile, adopting the methodology in Section~\ref{sec:policy-level} can retrieve the ideal coffee beans for a user using comparisons (where the mixing with ``do nothing’’ is done by diluting the coffee with water and the optimal coffee for a given profile is the one closest to it).

When moving from the explicit representation of policies as mappings from states to actions to a more natural policy representation  as a weighted trajectory set, we then obtained the same optimality guarantees in terms of the number of queries. 
While there could be other forms of policy representations (e.g., a small subset of common states), one advantage of our weighted trajectory set representation is that it captures the essence of the policy multi-objective value in a clear manner via $\cO(k)$ trajectories and weights. The algorithms provided in  Section~\ref{sec:trajectory-level} are standalone and could also be of independent interest for explainable RL~\citep{Alharin20SurveyInt}. For example, to exemplify the multi-objective performance of generic robotic vacuum cleaners (this is beneficial if we only have e.g., $3$ of them--- we can apply the algorithms in Section~\ref{sec:trajectory-level} to generate weighted trajectory set representations and compare them directly without going through the algorithm in Section~\ref{sec:policy-level}.).

An interesting direction for future work is to relax the assumption that the MDP is known in advance. One direct way is to first learn the model (in model-based RL), then apply our algorithms in the learned MDP. The sub-optimality of the returned policy will then depend on both the estimation error of the model and the error introduced by our algorithms (which depends on the parameters in the learned model).

\chapter{Online Learning with Primary and Secondary Loss}\label{chap:primary}
\section{Introduction}

The online learning problem has been studied extensively in the literature and used increasingly in many applications including hiring, advertising and recommender systems. One classical problem in online learning is prediction with expert advice, in which a decision maker makes a sequence of $T$ decisions with access to $K$ strategies (also called ``experts''). At each time step, the decision maker observes a scalar-valued loss of each expert. The standard objective is to perform as well as the best expert in hindsight. For example, a recruiter (the decision maker) sequentially decides which job applicants to hire with the objective of minimizing errors (of hiring an unqualified applicant and rejecting a qualified one). However, this may give rise to some social concerns since the decision receiver has a different objective (getting a job) which does not receive any attention. This problem can be modeled as an online learning problem with the primary loss (for the decision maker) and secondary loss (for the decision receiver). Taking the social impact into consideration, we ask the following question:

\begin{center}
    {\em Can we achieve low regret with respect to the primary loss, while performing}\\
    {\em
not much worse than the worst expert with respect to the secondary loss?}
\end{center}

Unfortunately, we answer this question negatively. More generally, we consider a bicriteria goal of minimizing the regret to the best expert with respect to the primary loss while minimizing the regret to a linear threshold $cT$ with respect to the secondary loss for some $c$. When the value of $c$ is set to the average secondary loss of the worst expert with respect to the secondary loss, the objective reduces to no-regret for the primary loss while performing no worse than the worst expert with respect to the secondary loss. Other examples, e.g., the average secondary loss of the worst expert with respect to the secondary loss among the experts with optimal primary loss, lead to different criteria of the secondary loss. Therefore, with the notion of regret to the linear threshold, we are able to study a more general goal. Based on this goal, we pose the following two questions: 
\begin{enumerate}

\item If all experts have secondary losses no greater than $cT+o(T)$ for some $c$, can we achieve
no-regret (compete comparably to the best expert) for the primary loss while achieving secondary loss no worse than $cT+o(T)$?\label{q1}
\item If we are given some external oracles to deactivate some ``bad'' experts with unsatisfactory secondary loss, can we perform as well as each expert with respect to the primary loss during the time they are active while achieving secondary loss no worse than $cT+o(T)$?\label{q2}
\end{enumerate}

These two questions are trivial in the i.i.d. setting as we can learn the best expert with respect to the primary loss within $O(\log(T))$ rounds and then we just need to follow the best expert. In this paper, we focus on answering these two questions in the adversarial online setting.

\subsection{Contributions}

\paragraph{An impossibility result without a bounded variance assumption}
We show that without any constraints on the variance of the secondary loss, even if all experts have secondary loss no greater than $cT$, achieving no-regret with respect to the primary loss and bounding secondary loss by $cT+O(T)$ is still unachievable. This answers our motivation question that it is impossible to achieve low regret with respect to the primary loss, while performing not much worse than the worst expert with respect to the secondary loss. This result explains why minimizing one loss while bounding another
is non-trivial and applying existing algorithms for scalar-valued losses after scalarizing primary and secondary losses does not work. We propose an assumption on experts that the secondary loss of the expert during any time interval does not exceed $cT$ by $O(T^\alpha)$ for some $\alpha\in [0,1)$. 

Then we study the problem in two scenarios, a ``good'' one in which all experts satisfy this assumption and a ``bad'' one in which experts partially satisfy this assumption and we are given access to an external oracle to deactivate and reactivate experts.

\paragraph{Our results in the ``good'' scenario}
In the ``good'' scenario, we show that running an algorithm with limited switching rounds such as Follow the Lazy Leader~\citep{kalai2005efficient} and Shrinking Dartboard (SD)~\citep{geulen2010regret} can achieve both regret to the best with respect to the primary loss and regret to $cT$ with respect to the secondary loss at $O(T^{\frac{1+\alpha}{2}})$.  We also provide a lower bound of $\Omega(T^\alpha)$.

From another perspective, we relax the ``good'' scenario constraint by introducing adaptiveness to the secondary loss and constraining the variance of the secondary loss between any two switchings for any algorithm instead of that of any expert. We show that in this weaker version of ``good'' scenario, the upper bound of running switching-limited algorithms matches the lower bound at $\Theta(T^{\frac{1+\alpha}{2}})$.

\paragraph{Our results in the ``bad'' scenario}
In the ``bad'' scenario, we assume that we are given an external oracle to determine which experts to deactivate as they do not satisfy the bounded variance assumption. We study two oracles here. One oracle deactivates the experts which do not satisfy the bounded variance assumption once detecting and never reactivates them. The other one reactivates those inactive experts at fixed rounds. In this framework, we are limited to select among the active experts at each round and we adopt a more general metric, sleeping regret, to measure the performance of the primary loss. We provide algorithms for the two oracles with theoretical guarantees on the sleeping regrets with respect to the primary loss and the regret to $cT$ with respect to the secondary loss.

\subsection{Related work}

One line of closely related work is online learning with multi-objective criterion. A bicriteria setting which examines not only the regret to the best expert but also the regret to a fixed mixture of all experts is investigated by~\cite{even2008regret,kapralov2011prediction,sani2014exploiting}. The objective by~\cite{even2009online} is to learn an optimal static allocation over experts with respect to a global cost function. Another multi-objective criterion called the Pareto regret frontier studied by~\cite{koolen2013pareto} examines the regret to each expert. Different from our work, all these criteria are studied in the setting of scalar-valued losses. The problem of multiple loss functions is studied by~\cite{chernov2009prediction} under a heavy geometric restriction on loss functions. For vector losses, one fundamental concept is the Pareto front, the set of feasible points in which none can be dominated by any other point given several criteria to be optimized~\citep{hwang2012multiple,auer2016pareto}. However, the Pareto front contains unsatisfactory solutions such as the one minimizing the secondary loss, which implies that learning the Pareto front can not achieve our goal. Another classical concept is approachability, in which a learner aims at making the averaged vector loss converge to a pre-specified target set~\citep{blackwell1956analog,abernethy2011blackwell}. However, we show that our fair solution is unapproachable without additional bounded variance assumptions. Approachability to an expansion target set based on the losses in hindsight is studied by~\cite{mannor2014approachability}. However, the expansion target set is not guaranteed to be meet our criteria. Multi-objective criterion has also been studied in multi-armed bandits~\citep{turgay2018multi}.

\section{Model}\label{sec:model-primary}

We consider the adversarial online learning setting with a set of $K$ experts $\cH = \{1,\ldots,K\}$. At round $t=1,2,\ldots,T$, given an active expert set $\cH_t\subseteq \cH$, an online learner $\cA$ computes a probability distribution $p_t\in \Delta_K$ over $\cH$ with support only over $\cH_t$ and selects one expert from $p_t$. Simultaneously an adversary selects two loss vectors $\loss{1}_t,\loss{2}_t\in [0,1]^K$, where $\loss{1}_{t,h}$ and $\loss{2}_{t,h}$ are the primary and secondary losses of expert $h\in \cH$ at time $t$. Then $\cA$ observes the loss vector and incurs expected losses $\loss{i}_{t,\cA}=p_t^\top \loss{i}_t$ for $i\in \{1,2\}$. Let $\Loss{i}_{T,h} = \sum_{t=1}^T\loss{i}_{t,h}$ denote the loss of expert $h$ and $\Loss{i}_{T,\cA}=\sum_{t=1}^Tp_t^\top \loss{i}_t$ denote the loss of algorithm $\cA$ for $i\in\{1,2\}$ during the first $T$ rounds. 
We will begin by focusing on the case that the active expert set $\cH_t =\cH$.

\subsection{Regret notions}

Traditionally, the regret (to the best) is used to measure the scalar-valued loss performance of a learner, which compares the loss of the learner and the best expert in hindsight. Similar to~\cite{even2008regret}, we adopt the regret notion of $\cA$ with respect to the primary loss as
\begin{align*}
\reg{1} \triangleq  \max\left({\Loss{1}_{T,\cA}- \min_{h\in \cH} \Loss{1}_{T,h}},1\right)\:.
\end{align*} 

We introduce another metric for the secondary loss called {\em regret to $cT$} for some $c\in[0,1]$, which compares the secondary loss of the learner with a linear term $cT$, 
\begin{align*}
\reg{2}_c \triangleq \max\left(\Loss{2}_{T,\cA}- cT,1\right)\:.
\end{align*}

Sleeping experts are developed to model the problem in which not all experts are available at all times~\citep{blum1997empirical,freund1997using}. At each round, each expert $h\in \cH$ decides to be active or not and then a learner can only select among the active experts, i.e. have non-zero probability $p_{t,h}$ over the active experts. The goal is to perform as well as $h^*$ in the rounds where $h^*$ is active for all $h^*\in \cH$. We denote by $\cH_t$ the set of active experts at round $t$. The sleeping regret for the primary loss with respect to expert $h^*$ is defined as

\begin{align*}
\sreg{1}(h^*) \triangleq \max\left(\sum_{t:h^*\in\cH_t}\sum_{h\in \cH_t}p_{t,h}\loss{1}_{t,h} - \sum_{t:h^*\in\cH_t} \loss{1}_{t,h^*},1\right)\:.
\end{align*}
The sleeping regret notion we adopt here is different from the regret to the best ordering of experts in the sleeping expert setting of~\cite{kleinberg2010regret}. Since achieving the optimal regret bound in Kleinberg's setting is computationally hard~\citep{kanade2014learning}, we focus on the sleeping regret notion defined above.

\subsection{Assumptions}

Following a standard terminology, we call an adversary oblivious if her selection is independent of the learner's actions. Otherwise, we call the adversary adaptive. First, we assume that the primary loss is oblivious. This is a common assumption in the online learning literature and this assumption holds throughout the paper.

\begin{assumption}\label{asp:loss1}
The primary losses $\{\loss{1}_t\}_{t\in[T]}$ are oblivious.
\end{assumption}
For an expert $h\in\cH$, we propose a bounded variance assumption on her secondary loss: the average secondary loss for any interval does not exceed $c$ much. More formally, the assumption is described as below.
\begin{assumption}\label{asp:intv}
For some given $c, \delta,\alpha \in [0,1]$ and for all expert $h\in \cH$, for any $T_1,T_2\in [T]$ with $T_1\leq T_2$, 
\begin{align*}
\sum_{t=T_1}^{T_2}(\loss{2}_{t,h}-c)\leq \delta T^\alpha\:.
\end{align*}
\end{assumption}
We show that such a bounded variance assumption is necessary in Section~\ref{sec:neg}.
We call a scenario ``good'' if all experts satisfy assumption~\ref{asp:intv}. Otherwise, we call the scenario ``bad''. This ``good'' constraint can be relaxed by introducing adaptiveness to the secondary loss. We have a relaxed version of the ``good'' scenario in which the average secondary loss between any two switchings does not exceed $c$ much for any algorithm. More formally,

\begin{customasp}{9.2$'$}\label{asp:intv2}
For some given $c, \delta,\alpha \in [0,1]$, for any algorithm $\cA$, let $\cA_t\in \cH$ denote the selected expert at round $t$. For any expert $h\in \cH$ and $T_1\in [T]$ such that $\cA_{T_1}=h$ and $\cA_{T_1-1}\neq h$ (where $\cA_{T+1} =  T \cA_{0}=0$ for notation simplicity), we have
\begin{align*}
\sum_{\tau =T_1}^{\min_{t >T_1: \cA_t \neq h}t-1}\left(\loss{2}_{\tau,h}-c\right)\leq \delta T^\alpha\:.
\end{align*}
\end{customasp}
In the ``good'' scenario, the active expert set $\cH_t =\cH$ for all rounds and the goal is minimizing both $\reg{1}$ and $\reg{2}_c$. In the ``bad'' scenario, we consider that we are given an oracle which determines $\cH_t$ at each round and the goal is minimizing $\sreg{1}(h^*)$ for all $h^*\in \cH$ and $\reg{2}_c$.

\section{Impossibility result without any bounded variance assumption}\label{sec:neg}

In this section, we show that without any additional assumption on the secondary loss, even if all experts have secondary loss no greater than $cT$ for some $c\in [0,1]$, there exists an adversary such that any algorithm incurs $\E[{\max(\reg{1},\reg{2}_c)}]=\Omega(T)$.
\begin{theorem}\label{thm:notwork}
Given a fixed expert set $\cH$, there exists an adversary such that any algorithm will incur $\E[{\max(\reg{1}, \reg{2}_c)}]=\Omega(T)$ with $c = \max_{h\in \cH} \Loss{2}_{T,h}/T$, where the expectation is taken over the randomness of the adversary.
\end{theorem}

\begin{proof}
To prove this theorem, we construct a binary classification example as below. 

In a binary classification problem, for each sample with true label $y\in\{+,-\}$ and prediction $\hy\in\{+,-\}$, the primary loss is defined as the expected $0/1$ loss for incorrect prediction, i.e., $\EEs{y,\hy}{\II{\hy\neq y}}$ and the secondary loss is defined as the expected $0/1$ loss for false negatives, i.e.,  $\EEs{y,\hy}{\II{\hy\neq y,y=+}}$. We denote by $h(b)$ the expert predicting $-$ with probability $b$ and $+$ otherwise. Then every expert can be represented by a sequence of values of $b$. At round $t$, the true label is negative with probability $a$. We divide $T$ into two phases evenly, $\{1,\ldots,T/2\}$ and $T/2+1,\ldots,T$, in each of which the adversary generates outcomes with different values of $a$ and two experts $\cH=\{h_1,h_2\}$ have different values of $b$ in different phases. We construct two worlds with different values of $a$ and $b$ in phase $2$ and any algorithm should have the same behavior in phase $1$ of both worlds. The adversary randomly chooses one world with equal probability. The specific values of $a$ and $b$ are given in Table~\ref{tab:bin}. Let $c=1/16$.

\begin{table}[H]
\caption{The values of $a$ and $b$ in different phases for the binary classification example.}\label{tab:bin}
\centering
\begin{tabular}{c|c|c|c}
\toprule
experts\textbackslash phase&$1: a = \frac{5}{8}$& $2: a = \frac{3}{4}$ (world I)&$2: a = \frac{5}{8}$ (world II)\\
\cmidrule{1-4}
$h_1$ & $b =\frac{1}{6}$ &$b = 0$& $b =\frac{1}{6}$\\
\cmidrule{1-4}
$h_2$ & $b =0$&$b = \frac{1}{2}$ & $b =0$\\
\bottomrule
\end{tabular}
\end{table}

The loss of expert $h(b)$ is
${\loss{1}_{t,h(b)}}=(1-a)b+a(1-b)$ and ${\loss{2}_{t,h(b)}}=(1-a)b$.
In phase $1$ and phase $2$ of world II, ${\loss{1}_{t,h_1}}= 7/12$, ${\loss{2}_{t,h_1}}=1/16$, ${\loss{1}_{t,h_2}}={5}/{8}$ and ${\loss{2}_{t,h_2}}=0$. In phase $2$ of world I, ${\loss{1}_{t,h_1}}=3/4$, ${\loss{2}_{t,h_1}}=0$, ${\loss{1}_{t,h_2}}={1}/{2}$ and ${\loss{2}_{t,h_2}}=1/8$. For any $h\in \cH$, we have $\Loss{2}_{T,h}\leq T/16$.

For any algorithm which selects $h_1$ for $T_1$ (in expectation) rounds in phase $1$ and $T_2$ (in expectation) rounds in phase $2$ of world I. If $T_1\leq T/4$, then ${\reg{1}} \geq (T/2-T_1)/24\geq T/96$ in world II; else if $T_1>T/4$ and $T_2\geq T_1/4$, then ${\reg{1}} \geq T_2/4-T_1/24\geq T/192$ in world I; else ${\reg{2}_c}= T_1/16+(T/2-T_2)/8-T/16 = (T_1-2T_2)/16\geq T/128$ in world I. In any case, we have $\E[{\max(\reg{1}, \reg{2}_c)}]=\Omega(T)$.  
\end{proof}

The proof of Theorem~\ref{thm:notwork} implies that an expert with total secondary loss no greater than $cT$ but high secondary loss at the beginning will consume a lot of budget for secondary loss, which makes switching to other experts with low primary loss later costly in terms of secondary loss. The theorem answers our first question negatively, i.e., we are unable to achieve no-regret for primary loss while performing as well as the worst expert with respect to the secondary loss. 

\section{Results in the ``good'' scenario}\label{sec:good}

In this section, we consider the problem of minimizing $\max(\reg{1}, \reg{2}_c)$ with Assumption~\ref{asp:intv} or~\ref{asp:intv2}. We first provide lower bounds of $\Omega(T^\alpha)$ under Assumption~\ref{asp:intv} and of $\Omega(T^\frac{1+\alpha}{2})$ under Assumption~\ref{asp:intv2}. Then we show that applying any switching-limited algorithms such as Shrinking Dartboard (SD)~\citep{geulen2010regret} and Follow the Lazy Leader (FLL)~\citep{kalai2005efficient} can achieve $\max(\reg{1}, \reg{2}_c) = O(T^\frac{1+\alpha}{2})$ under Assumption~\ref{asp:intv} or~\ref{asp:intv2}, which matches the lower bound under Assumption~\ref{asp:intv2}.

\subsection{Lower bound}

\begin{theorem}\label{thm:lb1}
If Assumption~\ref{asp:intv} holds with some given $c,\delta,\alpha$, then there exists an adversary such that any algorithm incurs $\E[{\max(\reg{1}, \reg{2}_c)}]=\Omega(T^\alpha)$.
\end{theorem}

\begin{proof}
We construct a binary classification example to prove the lower bound. 

The losses and the experts $\cH = \{h_1,h_2\}$ are defined based on $h(b)$ in the same way as that in the proof of Theorem~\ref{thm:notwork}. We divide $T$ into $3$ phases, the first two of which have $T^\alpha$ rounds and the third has $T-2T^\alpha$ rounds. Each expert has different $b$s in different phases as shown in Table~\ref{tab:lbbin}. At each time $t$, the sample is negative with probability $3/4$. We set $c=0$.

Since $(\loss{1}_{t,h(0)},\loss{2}_{t,h(0)}) = (3/4,0)$ and $(\loss{1}_{t,h(1)},\loss{2}_{t,h(1)}) = (1/4,1/4)$, the cumulative loss for both experts are $(\Loss{1}_{T,h},\Loss{2}_{T,h}) = (3T/4-T^\alpha/2,T^\alpha/4)$. Any algorithm $\cA$ achieving $\Loss{1}_{T,h}\leq 3T/4-T^\alpha/4$ will incur $\reg{2}_c\geq T^\alpha/8$.
\end{proof}

\begin{table}[H]
\caption{The values of $b$ in different phases for the binary classification example.}\label{tab:lbbin}
\centering
\begin{tabular}{c|c|c|c}
\toprule
experts\textbackslash phase&$1: T^\alpha$ & $2:T^\alpha$&$3:T-2T^\alpha$\\
\cmidrule{1-4}
$h_1$ & $b =1$ &$b = 0$& $b =0$\\
\cmidrule{1-4}
$h_2$ & $b =0$ &$b = 1$& $b =0$ \\
\bottomrule
\end{tabular}
\end{table}

Combined with the classical lower bound of $\Omega(\sqrt{T})$ in online learning~\citep{cesa2006prediction}, $\E[\max(\reg{1}, \reg{2}_c)]=\Omega(\max(T^\alpha, \sqrt{T}))$. In the relaxed version of the ``good'' scenario, we have the following theorem.
\begin{restatable}{theorem}{restatelb}\label{thm:lb2}
If Assumption~\ref{asp:intv2} holds with some given $c,\delta,\alpha$, then there exists an adversary such that any algorithm incurs $\E[\max(\reg{1}, \reg{2}_c)]=\Omega(T^{\frac{1+\alpha}{2}})$.
\end{restatable}

\paragraph{Sketch of the proof}
Inspired by the proof of the lower bound by~\cite{altschuler2018online}, we construct an adversary such that any algorithm achieving $\reg{1}=O(T^{\frac{1+\alpha}{2}})$ has to switch for some number of times. For the secondary loss, the adversary sets $\loss{2}_{t,h}=c$ only if $h$ has been selected for more than $T^\alpha$ rounds consecutively until time $t-1$; otherwise $\loss{2}_{t,h}=c+\delta$. In this case, every switching will increase the secondary loss. Then we can show that either $\reg{1}$ or $\reg{2}_c$ is $\Omega(T^{\frac{1+\alpha}{2}})$. The complete proof can be found in Appendix~\ref{apd:lb2}.

\subsection{Algorithm}

Under Assumption~\ref{asp:intv} or~\ref{asp:intv2}, we are likely to suffer an extra $\delta T^\alpha$ secondary loss every time we switch from one expert to another. Inspired by this, we can upper bound $\max(\reg{1}, \reg{2}_c)$ by limiting the number of switching times. Given a switching-limited learner $\cL$ on scalar-valued losses, e.g., Shrinking Dartboard (SD)~\citep{geulen2010regret} and Follow the Lazy Leader (FLL)~\citep{kalai2005efficient}, our algorithm $\cA_{\SL}(\cL)$ is described as below. 

We divide the time horizon into $T^{1-\alpha}$ epochs evenly and within each epoch we select the same expert. Let $e_i = \{(i-1)T^\alpha +1,\ldots,i T^\alpha\}$ denote the $i$-th epoch and $\loss{1}_{e_i,h}=\sum_{t\in e_i} \loss{1}_{t,h}/T^{\alpha}$ denote the average primary loss of the $i$-th epoch. We apply $\cL$ over $\{\loss{1}_{e_i,h}\}_{h\in\cH}$ for $i=1,\ldots,T^{1-\alpha}$. Let $s_{\SL}(E)$ and $r_{\SL}(E)$ denote the expected number of switching times and the regret of running $\cL$ for $E$ rounds. Then we have the following theorem.

\begin{theorem}\label{thm:alg}
Under Assumption~\ref{asp:intv} or~\ref{asp:intv2}, given a switching-limited learner $\cL$, $\cA_{\SL}(\cL)$ achieves $\reg{1} \leq T^\alpha{r_{\SL}(T^{1-\alpha})}$ and $\reg{2}_c \leq \delta T^\alpha ({s_{\SL}(T^{1-\alpha})}+1)$. By adopting SD or FLL as the learner $\cL$, $\cA_{\SL}(\sd)$ and $\cA_{\SL}(\fll)$ achieve $\max(\reg{1}, \reg{2}_c) = O(\sqrt{\log(K)T^{{1+\alpha}}})$.
\end{theorem}

\begin{proof}
It is obvious that $\reg{1} \leq T^\alpha{r_{\SL}(T^{1-\alpha})}$. We denote by $S$ the random variable of the total number of switching times and $\tau_1,\ldots,\tau_S$ the time steps the algorithm switches. For notation simplicity, let $\tau_0 =1$ and $\tau_{S+1} = T+1$. Then $\reg{2}_c =\E_{\cA}[{\sum_{t=1}^T (\loss{2}_{t,\cA_t}-c)}]\leq \E_{\cA}[{\sum_{s=0}^S\sum_{t=\tau_{s}}^{\tau_{s+1}-1}(\loss{2}_{t,\cA_t}-c)}] \leq \E_{\cA}[{\sum_{s=0}^S\delta T^\alpha}] = \delta T^\alpha (s_{\SL}(T^{1-\alpha})+1)$. Both SD and FLL have ${s_{\SL}(T^{1-\alpha})}=O(\sqrt{\log(K)T^{1-\alpha}})$ and ${r_{\SL}(T^{1-\alpha})}=O(\sqrt{\log(K)T^{1-\alpha}})$~\citep{geulen2010regret,kalai2005efficient}, which completes the proof.
\end{proof}

$\cA_{\SL}(\sd)$ and $\cA_{\SL}(\fll)$ match the lower bound at $\Theta(T^\frac{1+\alpha}{2})$ under Assumption~\ref{asp:intv2}. But there is a gap between the upper bound $O(T^\frac{1+\alpha}{2})$ and the lower bound $\Omega(T^\alpha)$ under Assumption~\ref{asp:intv}, which is left as an open question. We investigate this question a little bit by answering negatively if the analysis of $\cA_{\SL}(\cL)$ can be improved to achieve $O(T^\alpha)$. We define a class of algorithms which depends only on the cumulative losses of the experts, i.e., there exists a function $g: \R^{2K}\mapsto \Delta^K$ such that $p_t = g(\Loss{1}_{t-1},\Loss{2}_{t-1})$. Many classical algorithms such as Exponential Weights~\citep{littlestone1989weighted} and Follow the Perturbed Leader~\citep{kalai2005efficient} are examples in this class. The following theorem show that any algorithm dependent only on the cumulative losses cannot achieve a better bound than $\Omega(T^{\frac{1+\alpha}{2}})$, which provides some intuition on designing algorithms for future work.  The detailed proof can be found in Appendix~\ref{apd:subopt}.

\begin{restatable}{theorem}{restatesubopt}\label{thm:subopt}
Under Assumption~\ref{asp:intv}, for any algorithm only dependent on the cumulative losses of the experts, $\E[{\max(\reg{1}, \reg{2}_c)}] = \Omega(T^\frac{1+\alpha}{2})$.
\end{restatable}

\section{Results in the ``bad'' scenario}\label{sec:bad}

In the ``bad'' scenario, some experts may have secondary losses with high variance. To compete with the best expert in the period in which it has low variance, we assume that the learner is given some fixed external oracle determining which experts to deactivate and reactivate. In this section, we consider the goal of minimizing $\sreg{1}(h^*)$ for all $h^*\in \cH$ and $\reg{2}_c$. Here we study two oracles: one deactivates the ``unsatisfactory'' expert if detecting high variance of the secondary loss and never reactivates it again; the other one deactivates the ``unsatisfactory'' expert if detecting high variance of the secondary loss and reactivates it at fixed time steps.

\subsection{The first oracle: deactivating the ``unsatisfactory'' experts}

The oracle is described as below. The active expert set is initialized to contain all experts $\cH_1 = \cH$. At time $t=1,\ldots,T$, we let $\Delta\cH_{t}=\{h\in \cH_{t}: \exists t' \leq t, \sum_{\tau = t'}^{t}(\loss{2}_{\tau,h}-c)> \delta T^\alpha\}$ denote the set of active experts which do not satisfy Assumption~\ref{asp:intv}. Then we remove these experts from the active expert set, i.e., $\cH_{t+1} = \cH_{t}\setminus \Delta\cH_{t}$. We assume that there always exist some active experts, i.e. $\cH_T\neq \emptyset$. 

One direct way is running $\cA_{\SL}(\cL)$ as a subroutine and restarting $\cA_{\SL}(\cL)$ at time $t$ if there exist experts deactivated at the end of $t-1$, i.e., $\Delta H_{t-1}\neq \emptyset$. However, restarting will lead to linear dependency on $K$ for sleeping regrets. To avoid this linear dependency, we construct pseudo primary losses for each expert such that if $h$ is active at time $t$, $\tloss{1}_{t,h}= \loss{1}_{t,h}$; otherwise, $\tloss{1}_{t,h}= 1$. The probability of selecting inactive experts degenerates due to the high pseudo losses. For those inactive experts we cannot select, we construct a mapping $f:\cH \mapsto \cH$, which maps each expert to an active expert. If $\cA_{\SL}(\cL)$ decides to select an inactive expert $h$ at time $t$, we will select $f(h)$ instead. The detailed algorithm is described in Algorithm~\ref{alg:orc1}. Although the algorithm takes $\alpha$ as an input, it is worth to mention that the algorithm only uses $\alpha$ to decide the length of each epoch. We can choose a different epoch length and derive different regret upper bounds.

\begin{algorithm}[ht] \caption{$\cA_{1}$}\label{alg:orc1}
{\begin{algorithmic}[1]
\STATE {\bfseries Input:} $T$, $\cH$, $\alpha$ and a learner $\cL$
\STATE Initialize $f(h) = h$ for all $h\in \cH$.
\STATE Start an instance $\cA_{\SL}(\cL)$.
\FOR{$t=1,\ldots,T$}
\STATE Get expert $h_t$ from $\cA_{\SL}(\cL)$. 
\STATE Select expert $f(h_t)$.
\STATE Feed $\tloss{1}_{t}$ to $\cA_{\SL}(\cL)$.
\STATE For all $h$ with $f(h)\in \Delta\cH_{t}$, set $f(h) = h_0$, where $h_0$ is any expert in $\cH_{t+1}$.
\ENDFOR
\end{algorithmic}}
\end{algorithm}

\begin{theorem}
Let $T_{h^*}$ denote the number of rounds where expert $h^*$ is active. Running Algorithm~\ref{alg:orc1} with learner $\cL$ being SD or FLL can achieve
\begin{align}
\sreg{1}(h^*) =O(\sqrt{\log(K)T_{h^*}T^\alpha})\:,\label{eq:sr1}
\end{align}
for all $h^*\in \cH$ and
\begin{align}
\reg{2}_c= O(\sqrt{\log(K)T^{1+\alpha}} + K T^\alpha)\:.\label{eq:regc1}
\end{align}
\end{theorem}

\begin{proof}
Since $\loss{1}_{m, h}\leq \tloss{1}_{m,h}$, we have
\begin{align*}
\sreg{1}(h^*) = &\left(\sum_{t=1}^{T_{h^*}}\EEs{\cA}{\loss{1}_{t,\cA_t}} - \sum_{t=1}^{T_{h^*}} \loss{1}_{t,h^*} \right)\leq \left(\sum_{t=1}^{T_{h^*}}\EEs{\cA}{\tloss{1}_{t,\cA_t}} - \sum_{t=1}^{T_{h^*}} \tloss{1}_{t,h^*} \right) 
\\=&O(\sqrt{\log(K)T_{h^*}T^\alpha})\:,
\end{align*}
where the last step uses the results in Theorem~\ref{thm:alg}. It is quite direct to have $\reg{2}_c= O(\delta T^{\alpha}(\sqrt{\log(K)T^{1-\alpha}}+K))=O(\sqrt{\log(K)T^{1+\alpha}} + K T^\alpha)$, where the first term comes from the number of switching times for running $\cA_{\SL}$ and the second term comes from an extra switching caused by deactivating one expert.
\end{proof}

For the sleeping regret for expert $h^*$, the right hand side in Eq.~\eqref{eq:sr1} is $o(T_{h^*})$ if $T_{h^*}=\omega(T^\alpha)$, which is consistent with the impossibility result without bounded variance in Section~\ref{sec:neg}. When $\alpha \geq 1/2$, the right hand side of Eq.~\eqref{eq:regc1} is dominated by $KT^\alpha$. This linear dependency on $K$ is inevitable if we want to have $\sreg{1}_{h^*}=o(T_{h^*})$ for all $h^*\in \cH$. The proof is given in Appendix~\ref{apd:linK}.

\begin{restatable}{theorem}{restatelinK}\label{thm:linK}
Let $T_{h^*}=\omega(T^\alpha)$ for all $h^*\in\cH$. There exists an adversary such that any algorithm achieving $\sreg{1}_{h^*}=o(T_{h^*})$ for all $h^*\in \cH$ will incur $\reg{2}_c = \Omega(KT^\alpha)$ for $K=O(\log(T))$.
\end{restatable}

\subsection{The second oracle: reactivating at fixed times}

Now we consider the oracle which deactivates the unsatisfactory experts once detecting and reactivate them at fixed times. The oracle is described as follows. At given $N+1$ fixed time steps $t_0=1,t_1,\ldots,t_{N}$ with $t_{n+1}-t_{n}=\Omega(T^\beta)$ for some $\beta>\alpha$ (where $t_{N+1} = T+1$ for notation simplicity), the active expert set $\cH_t$ is reset to $\cH$. At time $t=t_{n},\ldots,t_{n+1}-2$ for any $n=0,\ldots,N$, the experts $\Delta\cH_{t}=\{h\in \cH_{t}: \exists t' \text{ such that } t_n \leq t' \leq t, \sum_{\tau = t'}^{t}(\loss{2}_{\tau,h}-c)> \delta T^\alpha\}$ will be deactivated, i.e. $\cH_{t+1} = \cH_{t}\setminus \Delta\cH_{t}$. We assume that there always exists some satisfactory experts, i.e. $\cH_{t_{n}-1}\neq \emptyset$ for all $n=1,\ldots, N+1$.

Restarting Algorithm~\ref{alg:orc1} at $t=t_0,\ldots,t_{N}$ is one of the most direct methods. Let $T^{(n)}_{h^*}$ denote the number of rounds $h^*$ is active during $t=t_n,\ldots,t_{n+1}-1$ and $T_{h^*}=\sum_{n=0}^N T^{(n)}_{h^*}$ denote the total number of rounds $h^*$ is active. Then we have $\sreg{1}_{h^*} =O(\sum_{n=0}^{N} \sqrt{\log(K)T^{(n)}_{h^*}T^{\alpha}})= O(\sqrt{\log(K)T_{h^*}T^{\alpha}N})$ and $\reg{2}_c = O(\sum_{n=0}^{N}(  \sqrt{\log(K)T^\alpha (t_{n+1}-t_n)} + K\delta T^\alpha))=O(\sqrt{\log(K)T^{1+\alpha}N} +  NKT^\alpha)$. 

However, if all experts are active all times, then the upper bound of $\sreg{1}(h^*)$ for the algorithm of restarting is $O(\sqrt{\log(K)T^{1+\alpha}N}) = O(\sqrt{\log(K)T^{2+\alpha-\beta}})$, which is quite large. We consider a smarter algorithm with better sleeping regrets when $T_{h^*}$ is large. The algorithm combines the methods of constructing meta experts for time-selection functions by~\cite{blum2007external} to bound the sleeping regrets and inside each interval, we select experts based on SD~\citep{geulen2010regret} to bound the number of switching times. We run the algorithm in epochs with length $T^\alpha$ and within each epoch we play the same expert. For simplicity, we assume that the active expert set will be updated only at the beginning of each epoch, which can be easily generalized. Let $e_i = \{(i-1)T^\alpha +1,\ldots,i T^\alpha\}$ denote the $i$-th epoch and $E=\{e_i\}_{i\in[T^{1-\alpha}]}$ denote the set of epochs. We let $\loss{1}_{e,h}=\sum_{t\in e} \loss{1}_{t,h}/T^{\alpha}$ and $\loss{1}_{e,\cA}=\sum_{t\in e} \loss{1}_{t,\cA_t}/T^{\alpha}$ denote the average primary loss of expert $h$ and the algorithm. And we let $\cH_{e}$ and $\Delta \cH_{e}$ denote the active expert set at the beginning of epoch $e$ and the deactivated expert set at the end of epoch $e$. Then we define the time selection function for epoch $e$ as $I_{h^*}(e)=\1(h^* \text{ is active in epoch }e)$ for each $h^*\in \cH$. Then we construct $K$ meta experts for each time selection function. Similar to Algorithm~\ref{alg:orc1}, we adopt the same expert mapping function $f$ and using pseudo losses $\tloss{1}_{e,h} = \loss{1}_{e,h}$ if $h$ is active and $\tloss{1}_{e,h}  = 1$ if not. The detailed algorithm is shown as Algorithm~\ref{alg:orc2}. Then we have the following theorem, the detailed proof of which is provided in Appendix~\ref{apd:ub}.
\begin{restatable}{theorem}{restateub}\label{thm:ub}
Running Algorithm~\ref{alg:orc2} can achieve
\begin{align*}
\sreg{1}(h^*) = O(\sqrt{\log(K)T^{1+\alpha}} +T_{h^*}\sqrt{\log(K)T^{\alpha-1}})\:,
\end{align*}
for all $h^*\in\cH$ and
\begin{align*}
\reg{2}_c = O(\sqrt{\log(K)T^{1+\alpha}}+\log(K)T^{\alpha}N+NKT^{\alpha})\:.
\end{align*}
\end{restatable}

Algorithm~\ref{alg:orc2} achieves $o(T_{h^*})$ sleeping regrets for $h^*$ with $T_{h^*}=\omega(T^{\frac{1+\alpha}{2}})$ and outperforms restarting Algorithm~\ref{alg:orc1} when $NT_{h^*}=\omega(T)$. $\sreg{1}(h^*)$ of Algorithm~\ref{alg:orc2} is $O(\sqrt{\log(K)T_{h^*}^{1+\alpha}})$ when $T_{h^*} = \Theta(T)$, which matches the results in Theorem~\ref{thm:alg}.
\begin{algorithm}[ht] \caption{$\cA_2$}\label{alg:orc2}
{\begin{algorithmic}[1]
\STATE {\bfseries Input:} $T$, $\cH$, $\alpha$ and $\eta$
\STATE Initialize $f(h) = h$ for all $h\in \cH$.
\STATE $w_{1,h}^{h^*} = \frac{1}{K}$ for all $h\in \cH$, for all $h^*\in\cH$.
\FOR{$m=1,\ldots, T^{1-\alpha}$}
	\STATE $w_{m,h} =\sum_{h^*} \ts(e_m)w_{m,h}^{h^*}$, $W_m = \sum_{h}w_{m,h}$ and $p_{m,h} = \frac{w_{m,h}}{W_m}$.
	\STATE \textbf{if} $m\in\{(t_n-1)/T^{1-\alpha}+1\}_{n=0}^N$ \textbf{then} get $h_m$ from $p_m$. \textbf{else}
	\STATE With prob. $\frac{w_{m,h_{m-1}}}{w_{m-1,h_{m-1}}}$, get $h_m=h_{m-1}$; with prob. $1-\frac{w_{m,h_{m-1}}}{w_{m-1,h_{m-1}}}$, get $h_m$ from $p_m$.
	\STATE \textbf{end if}	
	\STATE Select expert $f(h_m)$.
	\STATE Update $w_{m+1,h}^{h^*} = w_{m,h}^{h^*}\eta^{\ts(e_m)(\tloss{1}_{e_m,h}-\eta \tloss{1}_{e_m,\cA})+1}$ for all $h,h^*\in\cH$.
	\STATE For all $h$ with $f(h)\in \Delta\cH_{e_m}$, set $f(h) = h_0$, where $h_0$ is any expert in $\cH_{e_{m+1}}$.
\ENDFOR
\end{algorithmic}}
\end{algorithm}

\section{Discussion}

We introduce the study of online learning with primary and secondary losses. We find that achieving no-regret with respect to the primary loss while performing no worse than the worst expert with respect to the secondary loss is impossible in general. We propose a bounded variance assumption over experts such that we can control secondary losses by limiting the number of switching times. Therefore, we are able to bound the regret with respect to the primary loss and the regret to $cT$ with respect to the secondary loss. Our work is only a first step in this problem and there are several open questions.

One is the optimality under Assumption~\ref{asp:intv}. As aforementioned, our bounds of $\max(\reg{1}, \reg{2}_c)$ in the ``good'' scenario are not tight and we show that any algorithm only dependent on the cumulative losses will have $\reg{1} = \Omega(T^{\frac{1+\alpha}{2}})$, which indicates that the optimal algorithm cannot only depends on the cumulative losses if the optimal bound is $o(T^{\frac{1+\alpha}{2}})$. Under Assumption~\ref{asp:intv2}, the upper bound of the algorithm of limiting switching matches the lower bound. This possibly implies that limiting switching may not be the best way to make use of the information provided by Assumption~\ref{asp:intv}.

In the ``bad'' scenario with access to the oracle which reactivates experts at fixed times, our sleeping regret bounds depend not only on $T_{h^*}$ but also on $T$, which makes the bounds meaningless when $T_{h^*}$ is small. It is unclear if we can obtain optimal sleeping regrets dependent only on $T_{h^*}$ for all $h^*\in \cH$. The algorithm of {\em Adanormalhedge} by~\cite{luo2015achieving} can achieve sleeping regret of $O(\sqrt{T_{h^*}})$ without bound on the number of switching actions. However, how to achieve sleeping regret of $o(T_{h^*})$ with limited switching cost is of independent research interest.

In the ``bad'' scenario where Assumption~\ref{asp:intv} does not hold, we assume that $c$ is pre-specified and known to the oracle. Theorem~\ref{thm:notwork} show that achieving $\max(\reg{1},\reg{2}_c)=o(T)$ with $c = \max_{h} \Loss{2}_{T,h}$ is impossible without any external oracle. How to define a setting an unknown $c$ and design a reasonable oracle in this setting is an open question.

\chapter{Robust Learning under Clean-Label Attack}\label{chap:clean-label}
\section{Introduction}
Data poisoning is an attack on machine learning algorithms where the attacker adds examples to the training set with the goal of causing the algorithm to produce a classifier that makes specific mistakes the attacker wishes to induce at test time. In this paper, we focus on clean-label attacks in which an attacker, with knowledge of the training set $S$ and the test instance $x$, injects a set of examples labeled by the target function into the training set with the intent of fooling the learner into misclassifying the test instance $x$. This type of attack is called a clean-label attack because the attacker can only add correctly-labeled examples to the training set, and it has been proposed and studied empirically by~\cite{shafahi2018poison}.

In the realizable setting when the target function belongs to the hypothesis class $\cH$, any empirical risk minimizer (ERM) will achieve error of $\tilde{O}(\frac{\text{VCdim}(\cH)}{m})$ with training set size $m$.  This means that an ERM learner will still have error rate at most $\tilde{O}(\frac{\text{VCdim}(\cH)}{m})$ even in the presence of a clean-label attack; i.e., the attacker cannot significantly increase the overall {\em error rate}.   However, an attacker could still cause the ERM learner to make {\em specific} mistakes that the attacker wishes. For example, consider an ERM learner for the hypothesis class of intervals over $[0,1]$ that predicts the positive interval of maximum length consistent with the training data, in the case that the target function labels all of $[0,1]$ negative. Then any test instance not in the training set is attackable for this ERM learner by an adversary that adds enough poison examples so that the interval that the test instance is in becomes the largest interval in the training set.  On the other hand, for any target interval, for the ERM learner that outputs the {\em smallest} consistent interval, the attackable test instances will only have probability mass $O(1/m)$ (see Example~\ref{eg:interval} for more details).  Also, notice that for the hypothesis class of threshold functions over $[0,\infty)$, any ERM learner has a small portion of attackable test instances because the disagreement region of all consistent hypotheses is small and only test instances in the disagreement region are attackable.

From these examples, we can see that given an ERM learning algorithm $\A$ and a training set $S$, the probability mass of the attackable region (the set of attackable test instances) is at least as large as the error rate of the ERM learner and no greater than the disagreement region of all consistent hypotheses, and it depends on the specific algorithm $\A$. In this paper, we study the problem of whether we can obtain a small rate of attackable test instances in binary classification.
In the process we find interesting connections to existing literature on the sample complexity of PAC learning, and complexity measures arising in that literature.  Specifically, we study this problem in the realizable setting as it is unclear how to best define ``clean-label'' in the agnostic case.



\paragraph{Related work}
Clean-label data-poisoning attacks have been studied extensively in the literature \citep{shafahi2018poison,suciu2018does}, and  \cite{shafahi2018poison} show that clean-label attacks can be very  effective on neural nets empirically. 
For example, \cite{shafahi2018poison} show that in natural image domains, given the knowledge of the training model and of the test point to be attacked, the attacker can cause the model retrained with an injection of clean-label poisoned data to misclassify the given test instance with high success rate. Moreover, the attacker is able to succeed even though the overall error rate of the trained classifier remains relatively unchanged. 

\cite{mahloujifar2017blockwise, mahloujifar2018learning, mahloujifar2019universal} study a class of clean-label poisoning attacks called $p$-tampering attacks, where the attacker can substitute each training example with a correctly labeled poison example with independent probability $p$, and its variants.  \cite{mahloujifar2019can,mahloujifar2019curse,etesami2020computational} consider a more powerful adversary that can attack training examples of its choosing (rather than chosen at random) and show that the attacker can increase the probability of failing on a particular test instance from any non-negligible probability $\Omega(1/\poly(m))$ to $\approx 1$ by replacing $\tilde{O}(\sqrt{m})$ training examples with other correctly labeled examples.  In contrast, in our setting the attacker cannot modify any of the existing training examples and can only add new ones.  In addition, we mainly focus on attacks with an unlimited budget.

Data poisoning without requiring the poisoned data to be clean has been studied extensively (see \cite{biggio2012poisoning,barreno2006can,papernot2016towards,steinhardt2017certified} for a non-exhaustive list). Robustness to data poisoning with a small portion of poison examples has been studied by~\cite{ma2019data,levine2020deep}. The concurrent work of \cite{gao2021learning} studies the instance-targeted poisoning risk (which is the probability mass of the attackable region in the classification task) by various attacker classes, which have a budget controlling the amount of training data points they can change. They mainly focus on the relationship between robust learnability and the budget.

There are other studied attacking methods, including perturbation over training examples~\citep{koh2017understanding}, perturbation over test examples~\citep{szegedy2013intriguing,goodfellow2014explaining,bubeck2019adversarial,cullina2018pac,montasser2019vc,montasser2020efficiently} and etc. Another different notion of robust learning is studied by~\cite{xu2012robustness}, where the data set is partitioned into several subsets and the goal is to ensure the losses of instances falling into the same subset are close. Another line of related work is covariate shift, where the training distribution is different from the test distribution (see~\cite{quionero2009dataset} for an extensive study).

\paragraph{Notation}
For any vectors $u,v$, we let $\norm{u}$ denote the $\ell_2$ norm of $u$ and $\theta(u,v)$ denote the angle of $u$ and $v$. We denote by $e_i\in \R^n$ the one-hot vector with the $i$-th entry being one and others being zeros. We let $\cB^n(c,r) = \{x|\norm{x-c} \leq r\}$ denote the the ball with radius $r$ centered at $c\in \R^n$ in the $n$-dimensional space and $\sph^n(c,r)$ denote the sphere of $\cB^n(c,r)$. We omit the supscript $n$ when it is clear from the context. For any $a,b\in \R$, denote $a\wedge b = \min (a,b)$ and $a\vee b = \max (a,b)$. 
We use $\ln$ to represent natural logarithms and $\log$ to represent logarithms with base $2$.
Given a data set $S=\{(x_1,y_1),\ldots,(x_m,y_m)\}$ with size $m$, for any hypothesis $h$, we let $\err_S(h) = \frac{1}{m}\sum_{i=1}^m \ind{h(x_i)\neq y_i}$ denote the empirical error of $h$ over $S$. For a data distribution $\cD$, we let $\err_\cD(h) = \EEs{(x,y)\sim \cD}{\ind{h(x)\neq y}}$ denote the error of $h$. For any $A\subseteq \cX$, we let $\cP_\cD(A) = \PPs{(x,y) \sim \cD}{x \in A}$ denote the probability mass of $A$. The subscript $\cD$ is omitted when it is clear from the context. For any data set $S$, we let $S_\cX=\{x|(x,y)\in S\}$ and for $(x,y)\sim \cD$, we let $\cD_{\cX}$ denote the marginal distribution of $x$. For a finite set of hypotheses $\cH$, we let $\Major(\cH)$ denote the majority vote of $\cH$ and for simplicity denote $\Major(\cH,x) = \Major(\cH)(x) \triangleq \ind{\sum_{h\in\cH}h(x)\geq \ceil{{\abs{H}}/{2}}}$.
\section{Problem setup and summary of results}
Let $\cX$ denote the instance space and $\cY = \{0,1\}$ denote the label space. Given a hypothesis class $\cH\subseteq \cY^\cX$, we study the realizable case where there exists a deterministic target function $h^*\in \cH$ such that the training set and the test set are realized by $h^*$. Let $D_{h^*} = \{(x,h^*(x))|x\in \cX\}$ denote the data space where every instance is labeled by $h^*$. A learning algorithm $\A$ is a map (possibly including randomization), from a labeled data set $S$ (an unordered multiset) of any size, to a hypothesis $h$, and for simplicity we denote by $\cA(S,x) = \cA(S)(x)$ the prediction of $\cA(S)$ at an instance $x$. An attacker $\Adv$ maps a target function $h^*$, a training data set $S_\trn$ and a specific test instance $x$ to a data set $\Adv(h^*,S_\trn,x)$ (a multiset) and injects $\Adv(h^*,S_\trn,x)$ into the training set with the intent of making the learning algorithm misclassify $x$. We call $\Adv$ a \emph{clean-label} attacker if $\Adv(h^*,S_\trn,x)$ is consistent with $h^*$. Then for any deterministic algorithm $\A$, we say a point $x\in\cX$ is attackable if there exists a clean-label attacker $\Adv$ such that
\begin{align*}
    \A(S_\trn\cup \Adv(h^*,S_\trn,x),x)\neq h^*(x)\,.
\end{align*}
To be clear, we are defining $S_\trn\cup \Adv(h^*,S_\trn,x)$ as an unordered multiset. 
Formally, we define clean-label attackable rate as follows.
\begin{dfn}[clean-label attackable rate]\label{dfn:rate}
  For a target function $h^*$, a training data set $S_\trn$ and a (possibly randomized) algorithm $\A$, for any distribution $\cD$ over $D_{h^*}$, the attackable rate by $\Adv$ for $(h^*, S_\trn, \A)$ is defined as
    \begin{align*}
      \atk_\cD(h^*, S_\trn,\A,\Adv) \triangleq \EEs{(x,y)\sim \cD,\cA}{\ind{\A(S_\trn\cup \Adv(h^*,S_\trn,x),x)\neq h^*(x)}}\,.  
    \end{align*}
    The clean-label attackable rate is defined by the supremum over all clean-label attackers, i.e., 
    \begin{align*}
      \atk_\cD(h^*, S_\trn,\A) \triangleq \sup_\Adv \atk_\cD(h^*, S_\trn,\A,\Adv)\,.  
    \end{align*}
\end{dfn}
Then we define our learning problem as follows.
\begin{dfn}[$(\epsilon,\delta)$-robust learnability]\label{dfn:learnablity}
  For any $\epsilon,\delta \in (0,1)$, the sample complexity of $(\epsilon,\delta)$-robust learning of $\cH$, denoted by $\cM_{\rbst}(\epsilon,\delta)$, is defined as the smallest $m\in \NN$ for which there exists an algorithm $\A$ such that for every target function $h^*\in\cH$ and data distribution over $D_{h^*}$, with probability at least $1-\delta$ over $S_\trn\sim \cD^m$,
  \[\atk_\cD(h^*, S_\trn,\A)\leq \epsilon\,.\]
  If no such $m$ exists, define $\cM_{\rbst}(\epsilon,\delta)=\infty$. We say that $\cH$ is $(\epsilon,\delta)$-robust learnable if $\forall \epsilon,\delta \in (0,1)$, $\cM_{\rbst}(\epsilon,\delta)$ is finite.
\end{dfn}
It is direct to see that the error of $\A(S_\trn)$ is the attackable rate by attacker $\Adv_0$ which injects an empty set to the training set, i.e., $\Adv_0(\cdot)=\emptyset$. Therefore, for any algorithm $\A$, we have 
\[
  \atk_\cD(h^*, S_\trn,\A)\geq \atk_\cD(h^*, S_\trn,\A,\Adv_0) = \err_\cD(\A(S_\trn))\,,
  \]
which indicates any hypothesis class that is not PAC learnable is not robust learnable. For any deterministic $\A$, let us define $\ATK(h^*, S_\trn,\A,\Adv) \triangleq \{x\in \cX| \A(S_\trn\cup \Adv(h^*,S_\trn,x),x)\neq h^*(x)\}$ the attackable region by $\Adv$. For any ERM learner and any clean-label attacker $\Adv$, we have $\ATK(h^*, S_\trn,\ERM,\Adv)\subseteq \DIS(\VS_{\cH,S_\trn})$,
where $\VS_{\cH,S_\trn}$ is the version space of $S_\trn$, i.e., the set of all hypotheses in $\cH$ that classify $S_\trn$ correctly and $\DIS(\VS_{\cH,S_\trn}) = \{x|\exists h,h'\in \VS_{\cH,S_\trn}, h(x)\neq h'(x)\}$ is the disagreement region of the version space. Therefore, we have
\[
  \inf_\A\atk_\cD(h^*, S_\trn,\A) \leq \cP_{\cD}(\DIS(\VS_{\cH,S_\trn}))\,.
\]
However, large $\cP_{\cD}(\DIS(\VS_{\cH,S_\trn}))$ does not always result in large clean-label attackable rate. Below is an example showing the gap between them.


\begin{eg}[Interval over ${[0,1]}$]\label{eg:interval}
  The hypothesis class $\cH=\{\ind{(a,b)}: 0\leq a\leq b\leq 1\}\cup \{\ind{[a,b]}: 0\leq a\leq b\leq 1\}$ contains all intervals on $[0,1]$. We consider the following two learners.
  \begin{itemize}
    \setlength{\itemsep}{0pt}
    \setlength{\parsep}{0pt}
    \setlength{\parskip}{0pt}
    \item $\A_1(S)$: return $\ind{\emptyset}$ (the empty interval) if there are no positive examples in $S$ and return the consistent positive closed interval with minimum length otherwise.
    \item $\A_2(S)$: return the consistent positive open interval with maximum length.
    \end{itemize}
  Both are ERM learners for $\cH$. For any $h^* \in \cH$,  let the data distribution $\cD$ be a distribution on $D_{h^*}$ and $S_\trn\sim \cD^m$ for any $m>0$, then $\A_1$'s attackable rate is $\atk_\cD(h^*, S_\trn,\A_1)=\err_\cD(\A_1(S_\trn)) = \tilde{O}(1/m)$. However, consider algorithm $\A_2$ with $h^* = \ind{\emptyset}$. For $S_\trn = \{(x_1,y_1),\ldots,(x_m,y_m)\}$, w.l.o.g. assume that $x_1\leq \ldots\leq x_m$ and let $x_0=0$, $x_{m+1}=1$ for notation simplicity. Then for any $x\in (x_i,x_{i+1})$, the attacker can add enough poison data points to intervals $\{(x_j,x_{j+1})\}_{j\neq i}$ to make $(x_i,x_{i+1})$ be the interval with the maximum length. Therefore, so long as $\cD$ has no point masses, $\A_2$'s attackable rate is $\atk_\cD(h^*, S_\trn,\A_2) = \cP_{\cD}(\DIS(\VS_{\cH,S_\trn}))=1$.
\end{eg}

\paragraph{Main results}
We summarize the main contributions of this work.
\begin{itemize}
  \item In Section~\ref{sec:examples}, we present results on robust learnability under assumptions based on some known structural complexity measures, e.g., VC dimension $d=1$, hollow star number $k_o=\infty$, etc. In addition, we show that all robust algorithms can achieve optimal dependence on $\epsilon$ in their PAC sample complexity.
  
  \item In Section~\ref{sec:linear}, we show that the $n$-dimensional linear hypothesis class with $n\geq 2$ is not $(\epsilon,\delta)$-robust learnable. Then we study the linear problem in the case where the data distribution $\cD$ has margin $\gamma>0$. We propose one algorithm with sample complexity $O({n}{({2}/{\gamma})^n}\log(2/\gamma))$ and show that the optimal sample complexity is $ e^{\Omega(n)}$. We propose another algorithm in $2$-dimensional space with sample complexity $O(\log({1}/{\gamma})\log\log({1}/{\gamma}))$. We also show that even in the case where $\gamma$ is large and the attacker is only allowed to inject one poison example into the training set, SVM requires at least $ e^{\Omega(n)}$ samples to achieve low attackable rate.
 
  \item In Section~\ref{sec:finite}, we show that for any hypothesis class $\cH$ with VC dimension $d$, when the attacker is restricted to inject at most $t$ poison examples, $\cH$ is robust learnable with sample complexity $\widetilde{O}(\frac{dt}{\epsilon})$. We also show that there exists a hypothesis class with VC dimension $d$ such that any algorithm requires $\Omega(\frac{dt}{\epsilon})$ samples to achieve $\epsilon$ attackable rate.

\end{itemize}

\section{Connections to some known complexity measures and PAC learning}\label{sec:examples}
In this section, we analyze the robust learnability of hypothesis classes defined by a variety of known structural complexity measures. For some of these, we show they have the good property that there exists an algorithm such that adding clean-label points can only change the predictions on misclassified test instances and thus, the algorithm can achieve $\atk(h^*,S_\trn,\cA)\leq \err(\A(S_\trn))$. For some other structure, we prove that there will be a large attackable rate for any consistent proper learner. We also show the connection to optimal PAC learning in Section~\ref{sec:opt-pac}.

\subsection{Connections to some known complexity measures}
\paragraph{Hypothesis classes with VC dimension $d=1$ are $(\epsilon,\delta)$-robust learnable.} 
First, w.l.o.g., assume that for every $x\neq x'\in\cX$, there exists $h\in\cH$ such that $h(x)\neq h(x')$ (otherwise, operate over the appropriate equivalence classes). Then we adopt the partial ordering $\leq^\cH_f$ for any $f\in\cH$ over $\cX$ proposed by~\cite{ben20152} defined as follows.
\begin{dfn}[partial ordering $\leq^\cH_f$]
  For any $f\in\cH$, 
  \begin{align*}
    \leq^\cH_f \eqdef \{(x,x')|\forall h\in\cH, h(x')\neq f(x')\Rightarrow h(x)\neq f(x)\}\,.
  \end{align*}
\end{dfn} 
By Lemma~5 of~\cite{ben20152}, $\leq^\cH_f$ for $d=1$ is a tree ordering. Due to this structural property of hypothesis classes with VC dimension $d=1$, there is an algorithm 
originally proposed by \cite{ben20152} (Algorithm~\ref{alg:vc1} in Appendix~\ref{appx:vcd1}) such that adding clean-label poison points can only narrow down the error region (the set of misclassified instances). 
Roughly, the algorithm finds a maximal (by $\leq^\cH_f$) point $x'$ in the data such that 
$h^*(x') \neq f(x')$, and outputs the classifier 
labeling all $x \leq^\cH_f x'$ as $1-f(x)$ 
and the rest as $f(x)$.
We show that this algorithm can robustly learn $\cH$ using $m$ samples, where
  \[
    m= \frac{2\ln(1/\delta)}{\epsilon}\,.
  \]
The detailed algorithm and proof are given in Appendix~\ref{appx:vcd1}.
\paragraph{Intersection-closed hypothesis classes are $(\epsilon,\delta)$-robust learnable.}
A hypothesis class $\cH$ is called intersection-closed if the collection of sets $\{\{x|h(x)=1\}|h\in \cH\}$ is closed under intersections, i.e., $\forall h,h'\in\cH$, the classifier $x\mapsto \ind{h(x)=h'(x)=1}$ is also contained in $\cH$. For intersection-closed hypothesis classes, there is a general learning rule, called the Closure algorithm~\citep{helmbold:90,auer:07}. For given data $S$, the algorithm outputs $\hat{h} = \ind{\{x|\forall h\in\VS_{\cH,S}, h(x)=1\}}$. Since $\hat{h}(x)=1$ implies $h^*(x)=1$, and since adding clean-label poison points will only increase the region being predicted as positive,  we have $\atk(h^*, S_\trn,\clsr) = \err(\clsr(S_\trn))$. Then by Theorem~5 of~\cite{hanneke2016refined}, for any intersection-closed hypothesis class $\cH$ with VC dimension $d$, the Closure algorithm can robustly learn $\cH$ using $m$ samples, where
\[
  m= \frac{1}{\epsilon}(21d+16\ln(3/\delta))\,.
  \]

\paragraph{Unions of intervals are $(\epsilon,\delta)$-robust learnable.}
Let $\cH_k = \cup_{k'\leq k}\{\ind{\cup_{i=1}^{k'} (a_i,b_i)}|0\leq a_i< b_i\leq 1,\forall i\in [k']\}$ denote the union of at most $k$ positive open intervals for any $k\geq 1$. This hypothesis class is a generalization of Example~\ref{eg:interval}. There is a robust learning rule: output $\ind{\emptyset}$ if there is no positive sample and otherwise, output the consistent union of minimum number of closed intervals, each of which has minimum length. More specifically, given input (poisoned) data $S= \{(x_1,y_1),\ldots,(x_{m'},y_{m'})\}$ with $x_1\leq x_2\leq \ldots\leq x_{m'}$ w.l.o.g., for notation simplicity, let $y_0 =y_{m'+1}=0$. Then the algorithm $\A$ outputs $\hat{h}=\ind{X}$ where $X = \cup \{[x_i,x_j]|\forall i\leq l \leq j\in[m'], y_{i-1}=y_{j+1}=0, y_i=y_l=y_j=1\}$. The algorithm $\cA$ can robustly learn union of intervals $\cH_k$ using $m$ samples, where
\[
  m= O\!\left(\frac{1}{\epsilon}(k\log(1/\epsilon)+\log(1/\delta))\right)\,.
  \]
The detailed proof can be found in Appendix~\ref{appx:unionintvls}.
\paragraph{Hypothesis classes with finite star number are $(\epsilon,\delta)$-robust learnable.}
The star number, proposed by~\cite{hanneke2015minimax}, can measure the disagreement region of the version space.
\begin{dfn}[star number]
  The star number $\mathfrak{s}$ is the largest integer $s$ such that there exist distinct points $x_1,\ldots,x_s \in \cX$ and classifiers $h_0,\ldots,h_s$ with the property that $\forall i\in [s]$, $\DIS(\{h_0,h_i\})\cap \{x_1,\ldots,x_s\} = \{x_i\}$; if no such largest integer exists, define $\mathfrak{s}=\infty$.
\end{dfn}
By Theorem~10 of~\cite{hanneke2016refined}, for any $\cH$ with star number $\mathfrak{s}$, with probability at least $1-\delta$ over $S_\trn\sim \cD^m$, $\cP(\DIS(\VS_{\cH,S_\trn}))\leq\epsilon$ where
\[
  m= \frac{1}{\epsilon}(21\mathfrak{s}+16\ln(3/\delta))\,.
\]
As aforementioned, $\ATK(h^*, S_\trn,\ERM,\Adv)\subseteq \DIS(\VS_{\cH,S_\trn})$ for any clean-label attacker $\Adv$ and thus any ERM can robustly learn $\cH$ using $m$ samples.
\paragraph{Hypothesis classes with infinite hollow star number are not consistently properly $(\epsilon,\delta)$-robust learnable.}
The hollow star number, proposed by~\cite{bousquet2020proper}, characterizes proper learnability. For any set $S= \{(x_1,y_1),\ldots,(x_k,y_k)\}$, $S^{i} =\{(x_1,y_1'),\ldots,(x_k,y_k')\}$ is said to be a neighbor of $S$ if $y_i'\neq y_i$ and $y_j'=y_j$ for all $j\neq i$, for any $i\in[k]$.

\begin{dfn}[hollow star number]
  The hollow star number $k_o$ is the largest integer $k$ such that there is a set $S = \{(x_1,y_1),\ldots,(x_k,y_k)\}$ (called the hollow star set) which is not realizable by $\cH$, however every set $S'$ which is a neighbor of $S$ is realizable by $\cH$. If no such largest $k$ exists, define $k_o=\infty$.
\end{dfn}
For any hypothesis class $\cH$ with hollow star number $k_o$, for any consistent proper learner $\A$, there exists a target function $h^*$ and a data distribution $\cD$ such that if $m\leq \floor{(k_o-1)/2}$, then the expected attackable rate
\[
  \EEs{S_\trn\sim \cD^m}{\atk(h^*,S_\trn,\A)}\geq 1/4\,,
\]
which implies $\PPs{S_\trn}{\atk(h^*,S_\trn,\A)>1/8}\geq 1/7$ by Markov's inequality. The construction of the target function, the data distribution and the attacker is as described below. 
Consider a hollow star set $S$ as above, with size $k$. By definition, there exists a set of hypotheses $\{h_1,\ldots,h_{k}\}\subseteq \cH$ such that each neighbor $S^{i}$ is realized by $h_i$ for any $i\in[k]$. Consider the target function being $h_{i^*}$ where $i^*$ is drawn uniformly at random from $[k]$ and the marginal data distribution is a uniform distribution over $\{x_i|i\in[k]\setminus \{i^*\}\}$.  For any $\floor{(k-1)/{2}}$ i.i.d. samples from the data distribution, there are at least $k-\floor{(k-1)/{2}}$ instances in $S$ not sampled. To attack an unseen instance $x_i$, the attacker adds all examples in $S$ except $x_i, x_{i^*}$. Then any algorithm cannot tell whether $h_{i^*}$ or $h_i$ is the true target and any consistent proper learner will misclassify $\{x_i,x_{i^*}\}$ with probability $1/2$. 

For hypothesis classes with $k_o=\infty$, there is a sequence of hollow star sets with increasing sizes $\{k_i\}_{i=1}^\infty$. Therefore, any hypothesis class with $k_o=\infty$ is not consistently properly robust learnable. The detailed proof is included in Appendix~\ref{appx:hollow}.

\subsection{All robust learners are optimal PAC learners}\label{sec:opt-pac}

There is an interesting connection between algorithms robust to clean-label poisoning attacks
and the classic literature on the sample complexity of PAC learning.  Specifically, we can show that 
\emph{any} learning algorithm that is robust to clean-label poisoning attacks necessarily obtains
the optimal dependence on $\epsilon$ in its 
PAC sample complexity: that is, $O(1/\epsilon)$.
This is a very strong property, and not many such learning algorithms are known, as most learning 
algorithms have at least an extra $\log(1/\epsilon)$ factor in their sample complexity (see e.g., 
\citealp*{haussler:94,auer:07,hanneke:thesis,hanneke:16a,hanneke2016refined,darnstadt:15,bousquet2020proper}).  
Thus, this property can be very informative regarding what types of learning algorithms one should consider 
when attempting to achieve robustness to clean-label poisoning attacks.  This claim is 
formalized in the following result.  
Its proof is presented in Appendix~\ref{appx:opt-pac}.

\begin{thm}
\label{thm:opt-pac}
Fix any hypothesis class $\cH$.
Let $\A$ be a deterministic learning algorithm that always outputs a deterministic hypothesis.
Suppose there exists a non-negative 
sequence $R(m) \to 0$ such that, 
$\forall m \in \mathbb{N}$, 
for every target function $h^* \in \cH$ 
and every distribution $\cD$ over $D_{h^*}$, 
for $S_\trn \sim \cD^m$, 
with probability at least $1/2$, 
$\atk_{\cD}(h^*,S_\trn,\A) \leq R(m)$.
Then there exists an ($R$-dependent) finite constant $c_{R}$ such that,
for every $\delta \in (0,1)$, 
$m \in \mathbb{N}$, $h^* \in \cH$, and every distribution $\cD$ over $D_{h^*}$, 
for $S_\trn \sim \cD^m$, 
with probability at least $1-\delta$, 
$\atk_{\cD}(h^*,S_\trn,\A) \leq \frac{c_{R}}{m}\log\frac{2}{\delta}$.
\end{thm}

An immediate implication of this result 
(together with Markov's inequality) 
is that any deterministic $\A$ outputting 
deterministic predictors, if 
$\E_{S_\trn \sim \cD^m}[ \atk_{\cD}(h^*,S_\trn,\A) ] \leq R(m)/2 \to 0$
for all $h^* \in \cH$ and $\cD$ on $D_{h^*}$, 
then for any $h^* \in \cH$, 
$\cD$ on $D_{h^*}$, $\delta \in (0,1)$, 
$\err_{\cD}(\A(S_\trn)) \leq \frac{c_{R}}{m}\log\frac{2}{\delta}$ 
with probability at least $1-\delta$.
As mentioned, this is a strong 
requirement of the learning algorithm $\A$; 
for instance, for many classes $\cH$, many 
ERM learning rules would have an extra 
$\log(m)$ factor \citep*{hanneke2016refined}.
This also establishes a further connection 
to the hollow star number, which in some cases strengthens the result mentioned above 
(and detailed in Appendix~\ref{appx:examples}).
Specifically, \citet*{bousquet2020proper} have 
shown that when $k_o = \infty$, 
for any fixed $\delta$ sufficiently small, 
any proper learning algorithm has, 
for some infinite sequence of $m$ values, 
that $\exists h^* \in \cH$ and $\cD$ on $D_{h^*}$ for which, 
with probability greater than $\delta$, 
$\err_{\cD}(\A(S_\trn)) \geq \frac{c \log(m)}{m}$
for a numerical constant $c$.
Together with Theorem~\ref{thm:opt-pac}, 
this implies that for such classes, any 
deterministic proper learning algorithm cannot have a 
sequence $R(m) \to 0$ as in the above theorem.
Formally, using the fact that $\atk_{\cD}(h^*,S,\A)$ 
is non-increasing in $S$ (see the proof of 
Theorem~\ref{thm:opt-pac}), 
we arrive at the following corollary, 
which removes the ``consistency'' 
requirement from the result for  
classes with $k_o=\infty$ stated above, 
but adds a requirement of being deterministic.

\begin{crl}
\label{cor:proper-not-robust}
If $k_o = \infty$, then 
for any deterministic \emph{proper} learning algorithm $\A$ 
that always outputs a deterministic hypothesis,
there exists a constant $c > 0$ such that, 
for every $m \in \nats$, 
$\exists h^* \in \cH$ and distribution 
$\cD$ on $D_{h^*}$ 
such that 
$\E_{S_\trn \sim \cD^m}[ \atk_{\cD}(h^*,S_\trn,\A) ] > c$.
\end{crl}
\section{Linear hypothesis class}\label{sec:linear}
In this section, we first show that $n$-dimensional linear classifiers $\cH = \{\ind{\inner{w}{x}+b\geq 0}|(w,b)\in \R^{n+1}\}$ with $n\geq 2$ are not robust learnable. Then we study a restrictive case where the support of the data distribution has a positive margin to the boundary. We introduce two robust learners and prove a robust learning sample complexity lower bound. In addition, we also show the vulnerability of SVM. 

\subsection{Linear hypothesis class is not robust learnable}
In this section, we show that the class of linear hypotheses is not robust learnable.
\begin{thm}\label{thm:lblinear}
    For $n\geq 2$, the class of linear hypotheses is not robust learnable.
\end{thm}

\paragraph{Proof sketch}
We present the proof idea in the case of $n=3$ here and for simplicity, we allow the decision boundary to be either positive or negative. The construction details of limiting the boundary to be positive and the construction for $n=2$ are deferred to Appendix~\ref{appx:lb_linear}.

Consider the case where $\cX  = \sph^3(\bZero,1)$ is the sphere of the $3$-dimensional unit ball centered at the origin and the target function is uniformly randomly chosen from all linear classifiers with the decision boundary at distance $1/2$ from the origin, and the boundary labeled different from $\bZero$, i.e., $h^*\sim \Unif(\cH^*)$, where $\cH^* = \{\ind{\inner{w}{x}-\frac{1}{2}\geq 0}|\norm{w}=1\}\cup \{1-\ind{\inner{w}{x}-\frac{1}{2}\geq 0}|\norm{w}=1\}$. W.l.o.g., suppose $h^* = \ind{\inner{w^*}{x}-\frac{1}{2}\geq 0}$. The data distribution is the uniform distribution over the intersection of the decision boundary and the sphere, i.e., $\cD_\cX = \Unif(C_{w^*})$, where $C_{w^*} = \{x|\inner{w^*}{x}-\frac{1}{2}= 0\}\cap \sph^3(\bZero,1)$. Then all training data come from the circle $C_{w^*}$ and are labeled positive. 

Given training data $S_\trn\sim \cD^m$ and a test point $x_0 \in C_{w^*}$ (not in $S_{\trn,\cX}$), the attacker constructs a fake circle $C_{w'}$ tangent to $C_{w^*}$ at point $x_0$, i.e., $C_{w'}= \{x|\inner{w'}{x}-\frac{1}{2}= 0\}\cap \sph^3(\bZero,1)$ where $w' = 2\inner{x_0}{w^*}x_0 -w^*$. Then the attacker adds $m$ i.i.d. samples from the uniform distribution over $C_{w'}$ and labels them negative. Any algorithm cannot tell which circle is the true circle and which one of $\{\ind{\inner{w^*}{x}-\frac{1}{2}\geq 0},1-\ind{\inner{w'}{x}-\frac{1}{2}\geq 0} \}$ is the true target. Hence, any algorithm will misclassify $x_0$ with probability $1/2$.

\subsection{Linear hypothesis class is robust learnable under distribution with margin}\label{subsec:linearmrg}
In this section, we discuss linear classifiers in the case where the distribution has a positive margin. Specifically, considering the instance space $\cX\subseteq\cB^n(\bZero,1)$, we limit the data distribution $\cD$ to satisfy that $\forall (x,y)\in \supp(\cD), (2y-1)(\inner{w^*}{x}+b^*)\geq \gamma\norm{w^*}/2$ for some margin $\gamma\in(0,2]$ and target function $h^*(x) = \ind{\inner{w^*}{x}+b^*\geq 0}$.

\vspace{-3pt}
\subsubsection{A learner for arbitrary \texorpdfstring{$n>0$}{n>0}}

The learner $\cA$ fixes a $\gamma/2$-covering $V$ of $\cX$, i.e., $\forall x\in \cX,\exists v\in V, x\in \cB(v,\gamma/2)$, where $\abs{V}\leq (2/\gamma)^n$. It is easy to check that such a $V$ always exists. Then given input data $S$, the learner outputs a classifier: for $x\in\cB(v,\gamma/2)$, if $\exists(x',y')\in S$ s.t. $x'\in \cB(v,\gamma/2)$, predicting $h(x) = y'$; otherwise, predicting randomly. Note that $\Adv$ does not necessarily need to be restricted to such margin.

\begin{thm}
  The algorithm can robustly learn linear classifiers with margin $\gamma$ using $m$ samples where
  \[
    m=\frac{(2/\gamma)^n}{\epsilon}\left(n\ln \frac{2}{\gamma} +\ln \frac{1}{\delta}\right) \,.
  \]
\end{thm}
\begin{proof}
First, for every $v\in V$, at least one of $y \in \{0,1\}$ has $\cD( x \in \cB(v,\gamma/2) : h^*(x)=y ) = 0$. Then with probability at least $1-\abs{V}(1-\epsilon/\abs{V})^m$ over $S_\trn\sim \cD^m$, for every ball $\cB(v,\gamma/2)$ with probability mass at least $\epsilon/\abs{V}$, there exists $(x',y')\in S_\trn$ such that $x'\in \cB(v,\gamma/2)$. Let $m={\abs{V}\ln(\abs{V}/\delta)}/{\epsilon}$, we have with probability at least $1-\delta$, $\atk(h^*, S_\trn, \A)\leq \epsilon$.
\end{proof}
\vspace{-20pt}
\subsubsection{A learner for \texorpdfstring{$n=2$}{n=2}}
In the $2$-dimensional case, the hypothesis class can be represented as $\cH = \{h_{\beta,b}|\beta\in[0,2\pi), b\in[-2,2]\}$ where $h_{\beta,b} = \ind{(\cos \beta,\sin \beta)\cdot x + b \geq 0}$. When there is no ambiguity, we use $(\beta,b)$ to represent $h_{\beta,b}$. The target is $h^* =h_{\beta^*,b^*}$. Then we propose a robust algorithm based on binary-search for the target direction $\beta^*$ as shown in Algorithm~\ref{alg:linear2d}.
\begin{algorithm}[t]\caption{Robust algorithm for $2$-dimensional linear classifiers}\label{alg:linear2d}
  \begin{algorithmic}[1]
    \STATE \textbf{input}: data $S$
    \STATE \textbf{initialize} $l\la 0$, $h\la 2\pi$ and $\beta\la \frac{l+h}{2}$
    \STATE \textbf{if} $\exists b\in [-2,2]$ s.t. $(0,b)$ is consistent \textbf{then} output $(0,b)$
    \WHILE{$\nexists b\in [-2,2]$ s.t. $(\beta,b)$ is consistent with $S$}
    \STATE \textbf{if} $\exists \beta\in (l,\frac{l+h}{2})$ s.t. $\exists b\in [-2,2]$, $(\beta,b)$ is consistent with $S$ \textbf{then} let $h\la \frac{l+h}{2}$, $\beta \la \frac{l+h}{2}$
    \STATE \textbf{else} let $l\la \frac{l+h}{2}$, $\beta \la \frac{l+h}{2}$
    \ENDWHILE
    \STATE \textbf{return} $(\beta,b)$ with any consistent $b$
  \end{algorithmic}
\end{algorithm}

\begin{thm}\label{thm:lin2d}
  For any data distribution $\cD$, let $f(\epsilon'')=\max \{s\geq 0|\cP(\{x|(\cos \beta^*,\sin \beta^*)\cdot x + b^*\in [-s,0]\})\leq \epsilon'', \cP(\{x|(\cos \beta^*,\sin \beta^*)\cdot x + b^*\in [0,s]\})\leq \epsilon''\}$ for $\epsilon''\in[0,1]$ denote the maximum distance between the boundary and two parallel lines (on positive side and negative side respectively) such that the probability between the boundary and either line is no greater than $\epsilon''$. With probability at least $1-\delta$ over $S_\trn\sim \cD^m$, Algorithm~\ref{alg:linear2d} achieves
  \[
    \atk(h^*, S_\trn, \cA)  \leq \log(\frac{32}{f(\epsilon'')\wedge 2})2\epsilon' + 2\epsilon'' \,,
  \]
  for any $\epsilon''\in[0,1]$ using $m$ samples where
  \[
  m= \frac{24}{\epsilon'} \log \frac{13}{\epsilon'}+\frac{4}{\epsilon'} \log \frac{2}{\delta}\,.
  \]
\end{thm}
\paragraph{Proof sketch} First, by uniform convergence bound in PAC learning~\citep{blumer1989learnability}, when $m \geq \frac{24}{\epsilon'} \log \frac{13}{\epsilon'}+\frac{4}{\epsilon'} \log \frac{2}{\delta}$, every linear classifier consistent with $S_\trn$ has error no greater than $\epsilon'$. For any fixed $\beta$, the probability mass of union of error region of all $(\beta,b)$ consistent with the training data is bounded by $2\epsilon'$. Then given a target $(\beta^*,b^*)$, the binary-search path of $\beta$ is unique and adding clean-label poison examples will only change the depth of search.  When $h-l< \arctan(f(\epsilon'')/2)$, the attackable rate caused by deeper search is at most $2\epsilon''$. Combining these results together proves the theorem. The formal proof of Theorem~\ref{thm:lin2d} is included in Appendix~\ref{appx:lin2d}.

\begin{thm}\label{thm:lin2dmargin}
  For any $\gamma \in(0,2]$, Algorithm~\ref{alg:linear2d} can $(\epsilon,\delta)$-robustly learn $2$-dimensional linear classifiers with margin $\gamma$ using $m$ samples where
  \[
    m=\frac{48\log(64/\gamma)}{\epsilon}  \log \frac{26\log(64/\gamma)}{\epsilon}+\frac{8\log(64/\gamma)}{\epsilon} \log\frac 2\delta\,.\]
\end{thm}
Theorem~\ref{thm:lin2dmargin} is the immediate result of Theorem~\ref{thm:lin2d} as $f(0)=\gamma/2$. 

\subsubsection{SVM requires \texorpdfstring{${e^{\Omega(n)}}/{\epsilon}$}{Omega(exp(n)/epsilon)} samples against one-point attacker}
SVM is a well-known optimal PAC learner for linear hypothesis class~\citep{bousquet2020proper}. In this section, we show that even in the case where $\gamma\geq 1/8$ and the attacker is limited to add at most one poison point, SVM requires $e^{\Omega(n)}/\epsilon$ samples to achieve $\epsilon$ attackable rate.
\begin{thm}\label{thm:linSVM}
    For $n$-dim linear hypothesis class, for any $\epsilon< 1/16$, there exists a target $h^*\in\cH$ and a distribution $\cD$ over $D_{h^*}$ with margin $\gamma = 1/8$ such that $\EEs{S_\trn\sim \cD^m}{\atk_\cD(h^*, S_\trn,\SVM)}> \epsilon$ when the sample size $m< \frac{e^{n/128}}{768\epsilon}\vee \frac{1}{8\epsilon}$.
\end{thm}
\paragraph{Proof sketch}
Consider the case where $\cX = \{x\in\R^3|\norm{x}=1, \inner{x}{e_1}\geq0\}\cup \{-e_1\}$ is the union of a half sphere and a point $-e_1$. The target function is $h^* = \ind{\inner{w^*}{x}\geq -\gamma/2}$ with $w^* = e_1$ and margin $\gamma= 1/8$. Note that $h^*$ labels all points on the half sphere positive and $-e_1$ negative. Then we define the data distribution $\cD_\cX$ by putting probability mass $1-8\epsilon$ on $-e_1$ and putting probability mass $8\epsilon$ uniformly on the half sphere.

Then we draw training set $S_\trn\sim \cD^m$ and a test point $(x_0,y_0)\sim \cD$. Condition on that $x_0$ is on the half sphere, with high probability, $\inner{x_0}{w^*}\leq 1/8$. Then we define two base vectors $v_1 = w^*$ and $v_2 = \frac{x_0 - \inner{x_0}{w_*}w^*}{\norm{x_0 - \inner{x_0}{w_*}w^*}}$ in the $2$-dimensional space defined by $w^*$ and $x_0$. With high probability over the choice of $S_\trn$, for all positive training examples $x$ on the half sphere, we have $\inner{x}{v_1}\leq 1/8$ and $\inner{x}{v_2}\leq 1/8$. Then the attacker injects a poison point at $-\gamma v_1+\sqrt{1-\gamma^2}v_2$, which is closer to $x_0$ than all the positive samples in $S_\trn$. Since the poison point is classified as negative by the target function, SVM will misclassify $x_0$ as negative. The detailed proof can be found in Appendix~\ref{appx:linSVM}.

\subsubsection{Lower bound}
Here we show that robust learning of linear hypothesis class under distribution with margin $\gamma>0$ requires sample complexity 
${e^{\Omega(n)}}/{\epsilon}$.
\begin{thm}\label{thm:lblinmrg}
For $n$-dimensional linear hypothesis class with $n> 256$, for any $\epsilon\leq 1/16$ and for any algorithm $\cA$, there exists a target function $h^*\in\cH$ and a distribution $\cD$ over $D_{h^*}$ with margin $\gamma = 1/8$ such that $\EEs{S_\trn\sim \cD^m}{\atk_\cD(h^*, S_\trn,\A)}> \epsilon$ when the sample size $m\leq \frac{e^\frac{n-1}{128}}{192\epsilon}$. For convenience, here we relax the instance space by allowing $\cX\subseteq B^n(\bZero, 9/8)$. 
\end{thm}
The construction of the target function and the data distribution is similar to that in the proof of Theorem~\ref{thm:linSVM}. To attack a test instance $x_0$, the attacker adds the reflection points of all training points through the hyperplane defined by $x_0$ and $w^*$ such that any algorithm will misclassify $x_0$ with probability $1/2$. The detailed proof is included in Appendix~\ref{appx:lb_linmrg}.

\section{Results for finite-point attackers}\label{sec:finite}
In this section, instead of considering the case where the attacker can add a set of poison examples of arbitrary size, we study a restrictive case where the attacker is allowed to add at most $t$ poison examples for some $t<\infty$, i.e., $\abs{\Adv(h^*,S_\trn,x_0)}\leq t$ for any $h^*,S_\trn,x_0$. Following Definition~\ref{dfn:rate} and \ref{dfn:learnablity}, we define $t$-point clean-label attackable rate and $(t,\epsilon,\delta)$-robust learnability as follows. 
\begin{dfn}[$t$-point clean-label attackable rate]
For a target function $h^*$, a training data set $S_\trn$ and a (possibly randomized) algorithm $\A$, for any distribution $\cD$ over $D_{h^*}$, the $t$-point clean-label attackable rate is
\begin{align*}
    \atk_\cD(t, h^*, S_\trn,\A) \triangleq \sup_\Adv \atk_\cD(h^*, S_\trn,\A,\Adv)\text{ s.t. } \abs{\Adv(h^*,S_\trn,x)}\leq t,\forall x\in\cX\,.
\end{align*}
\end{dfn}
\begin{dfn}[$(t,\epsilon,\delta)$-robust learnability]
    A hypothesis class $\cH$ is $(t,\epsilon,\delta)$-robust learnable if there exists a learning algorithm $\A$ such that $\forall \epsilon,\delta\in (0,1)$, $\exists m(t,\epsilon,\delta)\in \NN$ such that $\forall h^*\in\cH, \forall \cD$ over $D_{h^*}$, with probability at least $1-\delta$ over $S_\trn\sim \cD^m$,
    \[\atk_\cD(t,h^*, S_\trn,\A)\leq \epsilon\,.\]

\end{dfn}
\subsection{\texorpdfstring{Algorithms robust to $t$}{t}-point attacker}
Robustness to a small number of poison examples has been studied by~\cite{ma2019data,levine2020deep}. \cite{ma2019data} show that differentially-private learners are naturally resistant to data poisoning when the attacker can only inject a small number of poison examples. \cite{levine2020deep} propose an algorithm called Deep Partition Aggregation (DPA), which partitions the training set into multiple sets by a deterministic hash function, trains base classifiers over each partition and then returns the majority vote of base classifiers. They show that for any instance $x$, the prediction on $x$ is unchanged if the number of votes of the output exceeds half of the number of the total votes by $t$. But the attackable rate of DPA is not guaranteed. Here we propose several algorithms similar to DPA but with guarantees on the attackable rate. In Algorithm~\ref{alg:impropert}, we provide a protocol converting any given ERM learner $\cL$ to a learner with small $t$-point clean-label attackable rate. 
\begin{algorithm}[t] \caption{A robust protocol for $t$-point attacker}\label{alg:impropert}
    {\begin{algorithmic}[1]
        \STATE \textbf{input}: A proper ERM learner $\cL$, data $S$
        \STATE divide $S$ into $10t+1$ blocks $\{S^{(1)},S^{(2)},\dots,S^{(10t+1)}\}$ with size $\floor{\frac{\abs{S}}{10t+1}}$ randomly without replacement (throw away the remaining $\abs{S}-(10t+1) \floor{\frac{\abs{S}}{10t+1}}$ points)
        \STATE \textbf{return} $\Major(\cH')$ where $\cH'=\{\cL(S^{(i)})|i\in[10t+1]\}$
    \end{algorithmic}}
\end{algorithm}

\begin{thm}\label{thm:timproper}
    For any hypothesis class $\cH$ with VC dimension $d$ with any proper ERM learner $\cL$, Algorithm~\ref{alg:impropert} can $(t,\epsilon,\delta)$-robustly learn $\cH$ using $m$ samples where
    \[
        m=O\left(\frac{dt}{\epsilon}\log\frac{dt}{\epsilon}+ \frac{d}{\epsilon} \log \frac{1}{\delta}  \right)\,.
    \]
\end{thm}
\paragraph{Proof sketch} For every misclassified point $x_0\in\cX$, there are at least $5t+1$ classifiers among $\{\cL(S^{(i)})\}_{i=1}^{10t+1}$ misclassifying $x_0$. Since there are at most $t$ blocks containing poison data, there are at least $4t+1$ non-contaminated classifiers (output by blocks without poison data) misclassifying $x_0$. Then $t$-point clean-label attackable rate is bounded by bounding the error of one non-contaminated classifier. The detailed proof is provided in Appendix~\ref{appx:improper_finite}. 

{\vskip 2mm}As we can see, Algorithm~\ref{alg:impropert} is improper even if $\cL$ is proper. Inspired by the projection number and the projection operator defined by~\cite{bousquet2020proper}, we propose a proper robust learner in Algorithm~\ref{alg:propertproj}. First, let us introduce the definitions of the projection number and the projection operator as follows. For a finite (multiset) $\cH'\subseteq \cH$, for $l\geq 2$, define the set $\cX_{\cH',l}\subseteq \cX$ of all the points $x$ on which less than $\frac{1}{l}$-fraction of all classifiers in $\cH'$ disagree with the majority. That is,
\begin{align*}
    \cX_{\cH',l} = \left\{x\in \cX: \sum_{h\in\cH'} \ind{h(x)\neq \Major(\cH',x)} < \frac{|\cH'|}{l}  \right\}.
\end{align*}
\vspace{-5pt}
\begin{dfn}[projection number and projection operator]
The projection number of $\cH$, denoted by $k_p$, is the smallest integer $k\geq 2$ such that, for any finite multiset $\cH'\subseteq\cH$ there exists $h\in \cH$ that agrees with $\Major(\cH')$ on the entire set $\cX_{\cH',k}$. If no such integer $k$ exists, define $k_p=\infty$. If $k_p<\infty$, the projection operator $\Proj_\cH: \cH'\mapsto \cH$ is a deterministic map from $\cH'$ to $\cH$ such that $\Proj_\cH(\cH',x)=\Major(\cH',x),\forall x\in \cX_{\cH',k_p}$.
\end{dfn}
\begin{algorithm}[t]
\caption{A proper robust learner for $t$-point attacker given projection number $k_p$}\label{alg:propertproj}
{\begin{algorithmic}[1]
    \STATE \textbf{input}: A proper $\ERM$ learner $\cL$, data $S$
    \STATE Divide the data $S$ into $10k_pt + 1$ sets $\{S^{(1)},S^{(2)},\dots,S^{(10k_pt+1)}\}$ with size $\floor{\frac{\abs{S}}{10 k_p t+1}}$ randomly without replacement (throw away the remaining $\abs{S}-(10k_pt+1) \floor{\frac{\abs{S}}{10k_pt+1}}$ points)
    \STATE \textbf{return} $\hat{h} = \Proj_\cH(\cH')$, where $\cH'=\{ h_i = \cL(S^{(i)})|i\in [10k_pt+1] \}$
\end{algorithmic}}
\end{algorithm}

\begin{thm}\label{thm:tpropproj}
For any hypothesis class $\cH$ with VC dimension $d$ and projection number $k_p$, with any proper $\ERM$ learner $\cL$, Algorithm~\ref{alg:propertproj} can $(t,\epsilon,\delta)$-robustly learn $\cH$ using $m$ samples where
\[
m=O\left(\frac{k_p^2dt}{\epsilon}\log\frac{k_pdt}{\epsilon} +  \frac{k_pd}{\epsilon}\ln \frac{1}{\delta} \right)\,.
\]
\end{thm}
The proof adopts the same idea as the proof of Theorem~\ref{thm:timproper} and is included in Appendix~\ref{appx:proper_finite}. For the hypothesis class with infinite projection number, we can obtain a proper learner in a similar way: randomly selecting $\floor{{\epsilon\abs{S}}/{3t}}$ samples with replacement from input data set $S$ and run ERM over the selected data. We show that this algorithm can $(t,\epsilon,\delta)$-robustly learn $\cH$ using $O(\frac{dt}{\epsilon^2} \log \frac{d}{\epsilon}+\frac{d}{\epsilon}\log \frac{1}{\delta})$ samples. The details of the algorithm and the analysis can be found in Appendix~\ref{appx:proper_finite}.

\subsection{Lower bound}
\begin{thm}\label{thm:lbfnt}
    For any $d\geq 1$ and $\epsilon\leq \frac{3}{8}$, there exists a hypothesis class $\cH$ with VC dimension $5d$ such that for any algorithm $\A$, there exists a target function $h^*\in \cH$ and a data distribution $\cD$ on $D_{h^*}$, such that $\EEs{S_\trn\sim \cD^m}{\atk_\cD(t,h^*,S_\trn,\A)}> \epsilon$ when the sample size $m< \frac{3td}{64\epsilon}$.
\end{thm}
\paragraph{Proof sketch}
Consider $d$ disjoint spheres in $\mathbb{R}^3$ and the target function is chosen by randomly selecting a circle on each sphere. Then label each circle differently from the rest of the sphere the circle lies on. Specifically, we flip $d$ independent fair coins, one for each circle to decide whether the circle is labeled positive or negative. The data distribution puts probability mass $\frac{t}{8m}$ uniformly on each circle and $1-\frac{td}{8m}$ probability mass on an irrelevant point (not on any of the spheres). Then we can show that with constant probability, every unseen point on each circle can be attacked by an attacker similar to the one in the proof sketch of Theorem~\ref{thm:lblinear}. The detailed proof is included in Appendix~\ref{appx:lb_finite}.
\begin{rmk}
Actually, our algorithms above even work for $t$-point \emph{unclean}-label attackers (where the poison data are not necessarily labeled by the target function) as well, which indicates that cleanness of poison examples does not make the problem fundamentally easier in the worst-case over classes 
of a given VC dimension, in the $t$-point 
attack case (although it can potentially 
make a difference for particular algorithms 
or particular classes $\cH$).
\end{rmk}



\section{Discussion and future directions}
In this paper, we show the impossibility of robust learning in the presence of clean-label attacks for some hypothesis classes with bounded VC dimension, e.g., the class of linear separators, and the robust learnability of some hypothesis classes characterized by known complexity measures, e.g., finite star number. There are several interesting open questions.
\vspace{-5pt}
\begin{itemize}
    \setlength{\itemsep}{1pt}
    \setlength{\parsep}{0pt}
    \setlength{\parskip}{0pt}
    \item The first question is what are necessary and sufficient conditions for $(\epsilon,\delta)$-robust learnability. Finite star number is a sufficient but not necessary condition. Here is an example where the instance space is $\cX=\NN$ and the hypothesis class is $\cH=\{\ind{i}|i\in \NN\}\cup \{0\}$. The star number of $\cH$ is $\mathfrak{s}=\infty$, but $\cH$ is robust learnable since $\vcd(\cH)=1$. One intriguing possible complexity measure is the largest number $k$ such that there is a set of distinct points $S =\{x_1,\ldots,x_k\}\in \cX^k$ and classifiers $\{h_0,\ldots,h_k\}$, where for any $i\in[k]$, there exists an involutory function $f_i:\cX\mapsto \cX$ (i.e., $f_i(f_i(x)) = x ,\forall x\in\cX$) such that $f_i(x_i) = x_i$ and $\DIS(\{h_0,h_i\})\cap \{S\cup f_i(S)\} = \{x_i\}$.
    
    \item For proper robust learning, we prove that any hypothesis class with infinite hollow star number $k_o=\infty$ is neither consistently properly $(\epsilon,\delta)$-robust learnable nor deterministically properly $(\epsilon,\delta)$-robust learnable. Compared with the fact that $k_o=\infty$ only brings an extra $\Omega(\log({1}/{\epsilon}))$ in the optimal PAC sample complexity~\citep{bousquet2020proper}, we see that the hollow star number has a dramatically larger impact on proper robust learnability. On the other hand, finite hollow star number does not suffice for robust learnability (e.g., \citealp{bousquet2020proper}, show linear classifiers on $\mathbb{R}^n$ have $k_o = n+2$), and it is unclear what is the necessary and sufficient condition for proper robust learnability.
    
    \item For linear classifiers with margin $\gamma>0$, the lower bound of the sample complexity presented in Section~\ref{sec:linear} ignores the dependence on $\gamma$. For the two learners introduced in Section~\ref{sec:linear}, the one using the covering set has sample complexity of $O(n(2/\gamma)^n\log(1/\gamma))$ and the other one designed for the $2$-dimension has sample complexity of $O(\log(1/\gamma)\log\log(1/\gamma))$. There is a huge gap between the lower bound and the upper bound and thus far, the optimal dependence on $\gamma$ remains unclear.
    
    \item For finite-point attacks, we construct a hypothesis class such that the $t$-point clean-label attackable rate is $\Omega(\frac{t}{m})$ in the proof of Theorem~\ref{thm:lbfnt} and Algorithm~\ref{alg:impropert} achieves $O(\frac{t\log(m)}{m})$ attackable rate. It is unclear to us for what kind of hypothesis class, there is an algorithm able to achieve $o(\frac{t}{m})$ attackable rate. At the same time, we are curious about its connection to $(\epsilon,\delta)$-robust learnablility. Notice that in all the proofs of the negative results in this paper, the attacker we construct never injects more than $m$ poison examples. This triggers the following suspicion: are infinite-point attackers strictly more powerful than $m$-point attackers? Specifically, we have the following conjecture.
    \begin{conj}[infinite to finite]
    For any hypothesis class $\cH$, for every target function $h^*\in\cH$, data distribution $\cD$ over $D_{h^*}$, there exist a pair of constants $c,c'>0$ such that for any $m>0$, any training data $S_\trn\in D_{h^*}^m$ and any algorithm $\cA$, $\atk_\cD(h^*, S_\trn,\A)\geq c$ iff  $\atk_\cD(m,h^*, S_\trn,\A)\geq c'$.
    \end{conj}
    Assuming that this conjecture holds, hence for any hypothesis class $\cH$, if there exists an algorithm $\A$ able to $(t,\epsilon,\delta)$-robustly learn $\cH$ with attackable rate $o(\frac{t}{m})$, then $\cH$ is $(m,\epsilon,\delta)$-robust learnable and thus, $(\epsilon,\delta)$-robust learnable.
     
    \item Another open question is whether abstention helps. Considering the case where the algorithm is allowed to abstain on $\epsilon$-fraction of inputs if the algorithm detects abnormality. That is to say, the algorithm outputs a selective classifier $(\hat h, \CR(S_\trn))$, where the prediction hypothesis $\hat h$ is a map from $\cX$ to $\cY$ and $\CR(S_\trn)\subseteq \cX$ is the confidence region of the prediction. The algorithm predicts $\A(S_\trn,x) = \hat h(x)$ if $x\in \CR(S_\trn)$ and $\A(S_\trn,x) = \perp$ if $x\notin \CR(S_\trn)$, where $\perp$ means the algorithm predicts ``I don't know''. Then for any deterministic algorithm, we say a test instance $x\in \cX$ is attackable if there is a clean-label attacker such that $x$ is predicted incorrectly as well as $x$ is in the confidence region, i.e.,
    \[\A(S_\trn\cup \Adv(h^*,S_\trn,x),x)\neq h^*(x) \qquad\&\qquad x\in \CR(S_\trn\cup \Adv(h^*,S_\trn,x))\,.\]
    We define the event $\cE(h^*,S_\trn,\cA,\Adv,x,y) = \{\A(S_\trn\cup \Adv(h^*,S_\trn,x),x)\neq h^*(x) \cap x\in \CR(S_\trn\cup \Adv(h^*,S_\trn,x))\}$ and then define the selective attackable rate as
\[\sup\nolimits_\Adv \EEs{(x,y)\sim \cD,\A}{\ind{\cE(h^*,S_\trn,\cA,\Adv,x,y)}}.\]
We are curious about the sample complexity required to achieve $\epsilon$ selective attackable rate while keeping the probability mass of the confidence region $\PPs{(x,y)\sim \cD}{x\in \CR(S)}\leq \epsilon$ for any input $S\supseteq S_\trn$. 
\end{itemize}

\chapter{Learning under Transformation Invariances}\label{chap:transformation}
\section{Introduction}
Transformation invariances are present in many real-world learning problems.
That is, given a certain set of transformations, the label of an instance is preserved under any transformation from the set.
Image classification is often invariant to rotation/flip/color translation.
Syntax parsing is invariant to exchange of noun phrases in a sentence.
Such invariances are often built into the learning process by two ways.
One is designing new architectures in neural networks to learn a transformation invariant feature, which is usually task-specific and challenging.
A more universally applicable and easier way is data augmentation (DA)\footnote{Throughout the paper, we refer to ERM over the augmented data by DA.}, that is, adding the transformed data into the training set and training a model with the augmented data.
Although DA performs well empirically, it is unclear whether and when DA ``helps''. 
In this paper, we focus on answering two questions:
\begin{center}
    \textit{How does data augmentation perform theoretically? 
    \\What is the optimal algorithm in terms of sample complexity under transformation invariances?}
\end{center}

We formalize the problem of binary classification under transformation invariances in the PAC model.
Given instance space $\cX$, label space $\cY=\{0,1\}$, and hypothesis class $\cH$,
we consider the following three settings according to different levels of realizability.
\begin{itemize}
    \item[(i)] Invariantly realizable setting: There exists a hypothesis $h^*\in \cH$ such that $h^*$ can correctly classify not only the natural data (drawn from the data distribution) but also the transformed data. For example, considering the transformation of rotating images where all natural images are upright, the hypothesis $h^*$ can correctly classify every upright image (natural data) and their rotations (transformed data).
    \item[(ii)] Relaxed realizable setting: There exists a hypothesis $h^*\in \cH$ such $h^*$ has zero error over the support of the data distribution (and therefore will correctly classify the transformed data that lies in the support of the data distribution), but $h^*$ may not correctly classify transformed data that lies outside the support of the natural data distribution. 
    For example, there exists an $h^*$ classifying all small rotations that lie in the support of the distribution correctly, but misclassifying upside-down cars.
    \item[(iii)] Agnostic setting: Every hypothesis in $\cH$ might not fit the natural data.
\end{itemize}
In most of this work, we consider the case where the set of transformations forms a group (e.g., all rotations and all color translations), which is a classic setting studied in literature~\citep[e.g.,][]{cohen2016group, bloem2020probabilistic, Chen2020}.
Some algorithms and analyses in this work also apply to non-group transformations (e.g., croppings).

\textbf{Main contributions} 
First, we show that DA outperforms vanilla ERM but is sub-optimal in setting (i) above.
We then introduce a complexity measure (see Definition~\ref{def:vco}) that characterizes the optimal sample complexity of learning in setting (i), 
and we give an optimal (up to log-factors) algorithm in this setting based on 1-inclusion-graph predictors.
Second, we characterize the complexity of learning in setting (ii) when the learner only receives the augmented data (without specifying which are natural).
Such a characterization provides us with a sufficient condition under which DA "hurts".
Third, we introduce a complexity measure (see Definition~\ref{def:vcao}) that characterizes the optimal sample complexity of learning in settings (ii) and (iii) above, and we give optimal algorithms for these settings. 
Finally, we also provide adaptive learning algorithms that interpolate between settings (i) and (ii), i.e., when $h^*$ is {\em partially} invariant. 
We want to emphasize that our complexity measures take into account the complexity of both the hypothesis class $\cH$ and the set of transformations being considered.
The results are formally summarized in Section~\ref{sec:results}.

\textbf{Related work}
Theoretical guarantees of DA has received a lot of attention recently. 
\cite{Chen2020,Lyle2020} study theoretical guarantees of DA under the assumption of ``equality'' in distribution, i.e., for any transformation in the transformation group,
the data distribution of the transformed data is approximately the same as that of the natural data (e.g., the upside-down variations of images happen at the same probability as the original upright images).
Under this assumption, they show that DA reduces variance and induces better generalization error upper bounds.
Our work does not make such an assumption.
\cite{Dao2019} models augmentation as a Markov process and shows that for kernel linear classifiers, 
DA can be approximated by
first-order feature averaging and second-order variance regularization components.
The concurrent work by~\cite{shen2022data} studies the benefit of DA when training a two layer convolutional neural
network in a specific multi-view model, showing that DA can alter the relative importance of various features. 
There is a line of theoretical study on the invariance gain in different models.
For example, \cite{elesedy2021linear} study the linear model and
\cite{elesedy2021kernel,mei2021learning, bietti2021sample} study the non-parametric regression.
The concurrent work by~\cite{elesedy2022group} also studies PAC learning under transformation invariances but only provides an upper bound on the sample complexity, while our work provides a complete characterization of learning under this model with optimal algorithms.
There is a parallel line of theoretical study on architecture design (e.g., \cite{wood1996representation,ravanbakhsh2017equivariance,kondor2018generalization,bloem2020probabilistic}).

Learning under transformation invariances has also been studied a lot empirically. 
Here we briefly mention a few results.
DA has been applied as standard method in modern deep learning, e.g., in Alexnet~\citep{krizhevsky2012imagenet}.
\cite{gontijo2020affinity} proposes two measures, affinity and diversity, to quantify the performance of the existing DA methods.
\cite{fawzi2016adaptive,cubuk2018autoaugment,chatzipantazis2021learning} study how to automatically search for improved data
augmentation policies.
For architecture design, one celebrated example is convolutions~\citep{fukushima1982neocognitron,lecun1989backpropagation}, which are translation equivariant.
See \cite{cohen2016group, dieleman2016exploiting, worrall2017harmonic} for more different architectures invariant or equivariant to different symmetries.


\textbf{Notation}
For any $n\in \NN$, let $\be_1,\be_2,\ldots$ denote the standard basis vectors in $\R^n$.
For any set $\cV$ and any $\bv\in \cV^n$, let $\bv_{-i}=(v_1,\ldots,v_{i-1},v_{i+1},\ldots,v_n)\in \cV^{n-1}$ denote the remaining part of $\bv$ after removing the $i$-th entry and $(v',\bv_{-i})=(v_1,\ldots,v_{i-1},v',v_{i+1},\ldots,v_n)\in \cV^n$ denote the vector after replacing $i$-th entry of $\bv$ with $v'\in \cV$.
Let $\oplus$ denote the bitwise XOR operator. 
For any $h\in \cY^\cX$ and $X=\{x_1,\ldots,x_n\}\subset \cX$, denote $h_{|X}= (h(x_1),\ldots,h(x_n))$ the restriction of $h$ on $X$.
A data set or a sample is a multiset of $\cX\times \cY$.
For any sample $S$, let $S_\cX = \{x|(x, y) \in S\}$ (with multiplicity) and for any distribution $\cD$ over $\cX\times \cY$, for $(x,y)\sim \cD$, let $\cD_\cX$ denote
the marginal distribution of $x$.
For any data distribution $\cD$ and any hypothesis $h$, the expected error $\err_{\cD}(h) := \Pr_{(x,y)\sim \cD}(h(x)\neq y)$.
Denote $\err(h)=\err_{\cD}(h)$ when $\cD$ is clear from the context.
For any sample $S$ of finite size,
$\err_{S}(h) := \frac{1}{\abs{S}}\sum_{(x,y)\in S} \ind{h(x)\neq y}$.
For any sample $S$ of possibly of infinite size, we say $\err_{S}(h) = 0$ if $h(x) = y$ for all $(x,y)\in S$.

\section{Problem setup}\label{sec:setup}
We study binary classification under transformation invariances. We denote by $\cX$ the instance space, $\cY=\{0,1\}$ the label space and $\cH$ the hypothesis class.

\textbf{Group transformations} 
We consider a group $\cG$ of transformations acting on the instance space through a mapping $\alpha: \cG \times \cX \mapsto \cX$, which
is compatible with the group operation.
For convenience, we write $\alpha(g,x) =gx$ for $g\in \cG$ and $x\in \cX$.
For example, consider $\cG = \{e, g_1,g_2, g_3\}$ where $e$ is the identify function and $g_i$ is rotation by $90 i$ degrees. Given an image $x$, $e x = x$ is the original image and $g_1 x$ is the image rotated by $90$ degrees.
The \emph{orbit} of any $x\in \cX$ is the subset of $\cX$ that can be obtained by acting an element in $\cG$ on $x$, $\cG x := \{gx|g\in \cG\}$. 
Note that since $\cG$ is a group, for any $x'\in \cG x$, we have $\cG x' = \cG x$. 
Thus we can divide the instance space $\cX$ into a collection of separated orbits, which does not depend on the data distribution.
Given a (natural) data set $S\subset \cX\times \cY$, we call $\cG S := \{(gx,y)|(x,y)\in S,g\in \cG\}$ the \emph{augmented data set}.

\textbf{Transformation invariant hypotheses and distributions}
To model transformation invariance, we assume that the true labels are invariant over the orbits of natural data.
Formally, for any transformation group $\cG$ and $X\subset\cX$, we say a hypothesis $h$ is \emph{$(\cG,X)$-invariant} if 
\[h(gx) = h(x),\forall g\in \cG, x\in X\,.\]
That is to say, for every $x\in X$, $h$ predicts every instance in the orbit of $x$ the same as $x$.
For any marginal distribution $\cD_\cX$ over $\cX$, we say a hypothesis $h$ is \emph{$(\cG,\cD_\cX)$-invariant} if 
$h(gx) = h(x)$ for all $g\in \cG$, for all $x \in \supp(\cD_\cX)$, i.e., $\Pr_{x\sim \cD_\cX}(\exists x'\in \cG x : h(x')\neq h(x)) = 0$.
We say a distribution $\cD$ over $\cX\times \cY$ is \emph{$\cG$-invariant} if there exists a $(\cG,\cD_\cX)$-invariant hypothesis $f^*$ (possibly not in $\cH$) with $\err_\cD(f^*) = 0$.
We assume that the data distribution is $\cG$-invariant throughout the paper.

\textbf{Realizability of hypothesis class} We consider three settings according to the different levels of realizability of $\cH$: (i) invariantly realizable setting, where there exists a $(\cG, \cD_\cX)$-invariant hypothesis $h^*\in \cH$ with $\err_\cD(h^*) = 0$; (ii) relaxed realizable setting, where there exists a (not necessarily $(\cG, \cD_\cX)$-invariant) hypothesis $h^*\in \cH$ with $\err_\cD(h^*) = 0$; and (iii) agnostic setting, where there might not exist a hypothesis in $\cH$ with zero error.
To understand the difference among the three settings, here is an example.

\begin{example}
    Consider $\cX = \{\pm 1,\pm 2\}$, $\cG=\{e,-e\}$ being the group generated by flipping the sign (i.e., $\cG x = \{x,-x\}$), and the data distribution $\cD$ being the uniform distribution over $\{(1,0),(2,0)\}$.
    If $\cH = \{ h(\cdot) = 0\}$ contains only the all-zero function, then it is in setting (i) as $h(\cdot) = 0$ is $(\cG,\cD_\cX)$-invariant and $\err_\cD(h) = 0$; If $\cH = \{\ind{x<0}\}$ contains only the hypothesis predicting $\{-1, -2\}$ as $1$ and $\{1,2\}$ as $0$, then it is in setting (ii) as $\ind{x<0}$ is not $(\cG,\cD_\cX)$-invariant but $\err_\cD(\ind{x<0}) = 0$; If $\cH = \{\ind{x>0}\}$, it is in setting (iii) as no hypothesis in $\cH$ has zero error.
\end{example}
The following definitions formalize the notion of PAC learning in the three settings.

\begin{definition}[Invariantly realizable PAC learnability]\label{def:inv-pac}
    For any $\epsilon,\delta \in (0,1)$, the sample complexity of invariantly realizable $(\epsilon,\delta)$-PAC learning of $\cH$ with respect to transformation group $\cG$, 
    denoted $\cM_\inv(\epsilon,\delta;\cH,\cG)$, 
    is defined as the smallest $m\in \NN$ for which there exists a learning rule $\cA$ such that, 
    for every $\cG$-invariant data distribution $\cD$ over $\cX\times \cY$ where there exists a \emph{$(\cG, \cD_\cX)$-invariant} predictor $h^*\in \cH$ with zero error, 
    $\err_{\cD}(h^*)=0$, with probability at least $1-\delta$ over $S\sim \cD^m$,
    \[\err_{\cD}(\cA(S))\leq \epsilon\,.\]
    If no such $m$ exists, define $\cM_\inv(\epsilon,\delta;\cH,\cG)=\infty$. 
    We say that $\cH$ is PAC learnable in the invariantly realizable setting with respect to transformation group $\cG$ if $\forall \epsilon,\delta\in (0,1)$, $\cM_\inv(\epsilon,\delta;\cH,\cG)$ is finite.
    For any algorithm $\cA$, denote by $\cM_{\inv}(\epsilon,\delta;\cH,\cG,\cA)$ the sample complexity of $\cA$. 
\end{definition}

\begin{definition}[Relaxed realizable PAC learnability]\label{def:re-pac}
    For any $\epsilon,\delta \in (0,1)$, the sample complexity of relaxed realizable $(\epsilon,\delta)$-PAC learning of $\cH$ with respect to transformation group $\cG$, 
    denoted $\cM_\re(\epsilon,\delta;\cH,\cG)$, 
    is defined as the smallest $m\in \NN$ for which there exists a learning rule $\cA$ such that, 
    for every $\cG$-invariant data distribution $\cD$ over $\cX\times \cY$ where there exists a predictor $h^*\in \cH$ with zero error, 
    $\err_{\cD}(h^*)=0$, with probability at least $1-\delta$ over $S\sim \cD^m$,
    \[\err_{\cD}(\cA(S))\leq \epsilon\,.\]
    If no such $m$ exists, define $\cM_\re(\epsilon,\delta;\cH,\cG)=\infty$. 
    We say that $\cH$ is PAC learnable in the relaxed realizable setting with respect to transformation group $\cG$ if $\forall \epsilon,\delta\in (0,1)$, $\cM_\re(\epsilon,\delta;\cH,\cG)$ is finite.
    For any algorithm $\cA$, denote by $\cM_{\re}(\epsilon,\delta;\cH,\cG,\cA)$ the sample complexity of $\cA$.
\end{definition}

\begin{definition}[Agnostic PAC learnability]\label{def:ag-pac}
    For any $\epsilon,\delta \in (0,1)$, the sample complexity of agnostic $(\epsilon,\delta)$-PAC learning of $\cH$ with respect to transformation group $\cG$, 
    denoted $\cM_\ag(\epsilon,\delta;\cH,\cG)$, 
    is defined as the smallest $m\in \NN$ for which there exists a learning rule $\cA$ such that, 
    for every $\cG$-invariant data distribution $\cD$ over $\cX\times \cY$, with probability at least $1-\delta$ over $S\sim \cD^m$,
    \[\err_{\cD}(\cA(S))\leq \inf_{h\in \cH} \err_\cD(h)+\epsilon\,.\]
    If no such $m$ exists, define $\cM_\ag(\epsilon,\delta;\cH,\cG)=\infty$. 
    We say that $\cH$ is PAC learnable in the agnostic setting with respect to transformation group $\cG$ if $\forall \epsilon,\delta\in (0,1)$, $\cM_\ag(\epsilon,\delta;\cH,\cG)$ is finite.
\end{definition}

\textbf{Data augmentation} One main goal of this work is to analyze the sample complexity of data augmentation.
When we talk of data augmentation (DA) as an algorithm, it actually means \emph{ERM over the augmented data}.
Specifically, given a fixed loss function $\cL$ mapping a data set and a hypothesis to $[0,1]$, and a training set $S_\trn$, DA outputs an $h\in \cH$ such that $h(x)=y$ for all $(x,y)\in \cG S_\trn$ if there exists one; 
outputs a hypothesis $h\in \cH$ with the minimal loss $\cL(\cG S_\trn, h)$ otherwise.
When we say DA without specifying the loss function, it means DA w.r.t. an arbitrary loss function, which can be defined based on any probability measure on the transformation group.

To characterize sample complexities, we define two measures as follows.

\begin{definition}[VC dimension of orbits]
\label{def:vco}
    The VC dimension of orbits, denoted $\vco(\cH, \cG)$, is defined as the largest integer $k$ for which there exists a set 
    $X = \{x_1,\ldots,x_k\}\subset \cX$ such that their orbits are pairwise disjoint, i.e., $\cG x_i \cap \cG x_j =\emptyset, \forall i,j\in [k]$ and every labeling of $X$ is realized by a \emph{$(\cG, X)$-invariant} hypothesis in $\cH$, i.e., $\forall y \in \{0,1\}^k$, there exists a $(\cG, X)$-invariant hypothesis $h\in \cH$ s.t. $h(x_i)=y_i,\forall i\in[k]$.
\end{definition}

\begin{definition}[VC dimension across orbits]
\label{def:vcao}
    The VC dimension across orbits, denoted $\vcao(\cH, \cG)$, is defined as the largest integer $k$ for which there exists a set $X = \{x_1,\ldots,x_k\}\subset \cX$ such that their orbits are pairwise disjoint, i.e., $\cG x_i \cap \cG x_j =\emptyset, \forall i,j\in [k]$ and every labeling of $X$ is realized by a hypothesis in $\cH$, i.e., $\forall y \in \{0,1\}^k$, there exists a hypothesis $h\in \cH$ s.t. $h(x_i)=y_i,\forall i\in[k]$.
\end{definition}
Let $\vcd(\cH)$ denote the VC dimension of $\cH$. 
By definition, it is direct to check that $\vco(\cH,\cG)\leq \vcao(\cH,\cG)\leq \vcd(\cH)$. 
For any $\cH,\cG$ with $\vco(\cH,\cG)=d$,
we can supplement $\cH$ to a new hypothesis class $\cH'$ such that $\vco(\cH',\cG)$ is still $d$ while $\vcao(\cH',\cG)$ is as large as the total number of orbits with at least two instances, i.e., $\vcao(\cH',\cG)=\abs{\{\cG x|\abs{\cG x}\geq 2, x\in \cX\}}$.
This can be done by supplementing $\cH$ with all hypotheses predicting $\cG x$  with two different labels for all $x$ with $\abs{\cG x}\geq 2$.
Besides, for any $\cH$ with $\vcd(\cH)=d$, we can construct a transformation group $\cG$ to make all instances lie in one single orbit, which makes $\vcao(\cH,\cG)\leq 1$.
Hence the gap among the three measures can be arbitrarily large.
Here are a few examples for better understanding of the gaps.

\begin{example}\label{eg:vco0}
    Consider $\cX = \{\pm 1, \pm 2,\ldots, \pm 2d\}$ for some $d>0$ , $\cH= \{\ind{x\in A}|A\subset [2d] \text{ and } \abs{A} = d\}$ being the set of all hypotheses labeling exact $d$ elements from $[2d]$ as $1$ and $\cG = \{e,-e\}$ being the group generated by flipping the sign.
    Then we have $\vco(\cH,\cG) = 0$ since for any $i\in [2d]$, there is no $(\cG, \{i\})$-invariant hypothesis that can label $i$ as $1$ (which is due to the fact that $-i$ is labeled as $0$ by any hypothesis in $\cH$).
    It is direct to check that $\vcao(\cH,\cG) = \vcd(\cH) = d$.
\end{example}

\begin{example}
    Consider $\cX = \{x\in \R^2|\norm{x}_2 = 1\}$ being a circle, the hypothesis class $\cH = \{0,1\}^{\cX}$ being all labeling functions and $\cG$ being all rotations (thus $\forall x, \cG x = \cX$).
    Then we have $\vco(\cH,\cG)=\vcao(\cH,\cG)= 1$ as there is only one orbit and $\vcd(\cH) = \infty$. 
\end{example}
\begin{example}
     Consider the natural data being $k$ upright images and the transformation set $\cG$ is rotation by $0, 360/n, 2\cdot 360/n, \ldots, (n-1)\cdot 360/n$ degrees for some integer n. For an expressive hypothesis class $\cH$ (e.g., neural networks) that can shatter all rotated versions of these images, we have $\vcd(\cH)=nk$ and $\vco(\cH,\cG) = \vcao(\cH,\cG) = k$.
     For a hypothesis class $\cH'$ composed of all hypotheses labeling all upright images and their upside-down variations differently, we have $\vcd(\cH') = (n-1)k$, $\vcao(\cH',\cG) = k$ and $\vco(\cH',\cG) = 0$.
\end{example}


\section{Main results}\label{sec:results}
We next present and discuss our main results.
\begin{itemize}[leftmargin=*]
    \item Invariantly realizable setting (Definition~\ref{def:inv-pac})
    \begin{itemize}[leftmargin=*]
        \item \textbf{DA ``helps'' but is not optimal. The sample complexity of DA is characterized by $\vcao(\cH,\cG)$.} For any $\cH,\cG$, DA can learn $\cH$ with sample complexity $\tilde O(\frac{\vcao(\cH,\cG)}{\epsilon}+\frac{1}{\epsilon}\log \frac 1 \delta)$, where $\tilde O$ ignores log-factors of $\frac{\vcao(\cH,\cG)}{\epsilon}$ (Theorem~\ref{thm:inv-da-ub}). 
        For all $d> 0$, there exists $\cH, \cG$ with $\vcao(\cH, \cG)=d$ and $\vco(\cH,\cG) = 0$ such that DA needs $\Omega(\frac{d}{\epsilon})$ samples (Theorem~\ref{thm:inv-da-lb}).
        \item
        \textbf{The optimal sample complexity is characterized by $\vco(\cH,\cG)$.} 
        For any $\cH,\cG$, we have $\Omega(\frac{\vco(\cH,\cG)}{\epsilon} +\frac{1}{\epsilon}\log\frac{1}{\delta})\leq \cM_\inv(\epsilon,\delta;\cH,\cG) \leq O(\frac{\vco(\cH,\cG)}{\epsilon}\log \frac{1}{\delta})$ (Theorem~\ref{thm:inv-opt}).
        We propose an algorithm achieving this upper bound based on 1-inclusion graphs, which does not distinguish between the original and transformed data.
        It is worth noting that the algorithm takes the invariance over the test point into account, which provides some theoretical justification for test-time adaptation such as~\cite{wang2020tent}.
    \end{itemize}

    \item Relaxed realizable setting (Definition~\ref{def:re-pac})
    \begin{itemize}[leftmargin=*]
        \item \textbf{DA can ``hurt''.} DA belongs to the family of algorithms not distinguishing the original data from the transformed data.
        We show that the optimal sample complexity of this family is characterized by $\mu(\cH,\cG)$ (see Definition~\ref{def:mu}) (Theorem~\ref{thm:re-da}), which can be arbitrarily larger than $\vcd(\cH)$.
        This implies that for any $\cH,\cG$ with $\mu(\cH,\cG)>\vcd(\cH)$, 
        the sample complexity of DA is higher than that of ERM.

        \item \textbf{The optimal sample complexity is characterized by $\vcao(\cH,\cG)$.}
        For any $\cH,\cG$, we have $\Omega(\frac{\vcao(\cH,\cG)}{\epsilon} + \frac{\log(1/\delta)}{\epsilon})\leq \cM_\re(\epsilon,\delta;\cH,\cG) \leq \tilde O(\frac{\vcao(\cH,\cG)}{\epsilon}+\frac{1}{\epsilon}\log \frac 1 \delta)$ (Theorem~\ref{thm:re-opt}).
        We propose two algorithms achieving similar upper bounds, with one based on ERM and one based on 1-inclusion graphs.
        Both algorithms have to distinguish between the original and the transformed data.
        \item \textbf{An adaptive algorithm interpolates between two settings.} We present an algorithm that adapts to different levels of invariance of the target function $h^*$, which achieves $\tilde O(\frac{\vcao(\cH,\cG)}{\epsilon})$ sample complexity in the relaxed realizable setting and $\tilde O(\frac{\vco(\cH,\cG)}{\epsilon})$ sample complexity in the invariantly realizable setting without knowing it (Theorem~\ref{thm:re-eta-unknown-ub} in Appendix). 
    \end{itemize}
    \item Agnostic setting (Definition~\ref{def:ag-pac})
    \begin{itemize}[leftmargin=*]
    \item \textbf{The optimal sample complexity is characterized by $\vcao(\cH,\cG)$.}
    For any $\cH,\cG$, $$\cM_\ag(\epsilon,\delta;\cH,\cG) = O\left(\frac{\vcao(\cH,\cG)}{\epsilon^2}\log^2\left(\frac{\vcao(\cH,\cG)}{\epsilon}\right)+\frac{1}{\epsilon^2}\log(\frac 1 \delta)\right)$$ (see Theorem~\ref{thm:ag-opt}).
    Since $\cM_\ag(\epsilon,\delta;\cH,\cG)  \geq \cM_\re(\epsilon,\delta;\cH,\cG)$, $\vcao(\cH,\cG)$ characterizes the optimal sample complexity.
\end{itemize}
\end{itemize}
\section{Invariantly Realizable setting}\label{sec:inv}
In this section, we discuss the results in the invariantly realizable setting (see Definition~\ref{def:inv-pac}).

\subsection{DA ``helps'' but is not optimal}
We show that in the invariantly realizable setting, DA indeed ``helps'' to improve the sample complexity from $\tilde O(\frac{\vcd(\cH)}{\epsilon})$ (the sample complexity of ERM in standard PAC learning) to $\tilde O(\frac{\vcao(\cH,\cG)}{\epsilon})$.
First, we have the following upper bound on the sample complexity of DA.
\begin{theorem}\label{thm:inv-da-ub}
    For any $\cH,\cG$ with $\vcao(\cH,\cG)<\infty$, DA satisfies that $\cM_\inv(\epsilon,\delta;\cH,\cG,\da) = O( \frac{\vcao(\cH,\cG)}{\epsilon}\log^3\frac{\vcao(\cH,\cG)}{\epsilon}+\frac{1}{\epsilon}\log \frac 1 \delta)$.
\end{theorem}
Intuitively, for a set of instances in one orbit that can be labeled by $\cH$ in multiple ways, we only need to observe one instance from this orbit to learn the labels of all the instances by applying DA. Thus, DA helps to improve the accuracy.
The detailed proof is deferred to Appendix~\ref{app:inv-da-ub}.
However, DA does not fully exploit the transformation invariances as it only utilizes the invariances of the training set.
Hence, DA does not perform optimally in presence of the transformation invariances.
In fact, besides DA, all proper learners (i.e., learners outputting a hypothesis in $\cH$) have the same problem.

\begin{theorem}\label{thm:inv-da-lb}
    For any $d>0$, there exists a hypothesis class $\cH_d$ and a group $\cG_d$ with $\vcao(\cH_d,\cG_d)=d$ and $\vco(\cH_d,\cG_d) = 0$ such that $\cM_\inv(\epsilon,\frac{1}{9};\cH_d,\cG_d,\cA)=\Omega( \frac{d}{\epsilon})$ for any proper learner $\cA$, including DA and standard ERM.
\end{theorem}
The theorem shows that DA is sub-optimal as we will show that the optimal sample complexity is characterized by $\vco(\cH,\cG)$ in Theorem~\ref{thm:inv-opt}. 
We provide an idea of the construction here and defer the detailed proof to Appendix~\ref{app:inv-da-lb}.
Consider the $\cX$, $\cH$ and $\cG$ in Example~\ref{eg:vco0}.
Pick the target function $\ind{x\in A}$ uniformly at random from $\cH$ and let the data distribution only put probability mass on points in $[2d]\setminus A$, the orbits of which are labeled as $0$ by the target function.
Then any proper learner must predict $d$ unobserved examples of $[2d]$ as $1$, which leads to high error if the learner observes fewer than $d/2$ examples.
Theorem~\ref{thm:inv-da-lb} also implies that for any hypothesis class including $\cH$ as a subset,
there exists a DA learner (i.e., a proper learner fitting the augmented data) whose sample complexity is $\Omega(\frac{d}{\epsilon})$.

Theorem~\ref{thm:inv-da-ub} shows that the sample complexity of DA is $\tilde O(\frac{\vcao(\cH,\cG)}{\epsilon})$, better than that of ERM in standard PAC learning, $\tilde O(\frac{\vcd(\cH)}{\epsilon})$.
This is insufficient to show that DA outperforms ERM as it might be possible that ERM can also achieve better sample complexity in presence of transformation invariances.
To illustrate that DA indeed "helps", we show that any algorithm without exploiting the transformation invariances still requires sample complexity of $\Omega(\frac{\vcd(\cH)}{\epsilon})$.
\begin{theorem}\label{thm:inv-da-help}
    For any $\cH$, there exists a group $\cG$ with $\vcao(\cH,\cG) \leq 5$ s.t. $\cM_\inv(\epsilon,\delta;\cH,\cG,\cA)=\Omega(\frac{\vcd(\cH)}{\epsilon} +\frac{1}{\epsilon}\log \frac{1}{\delta})$ for any algorithm $\cA$ not given any information about $\cG$ (e.g., ERM).
\end{theorem}
The basic idea is that, given a set of $k$ instances that can be shattered by $\cH$ for some $k>0$,
$\cG$ is uniformly at random picked from a set of $2^k$ groups, each of which partitions the set into two orbits in a different way.
If given $\cG$, the algorithm only need to observe one instance in each orbit to learn the labels of all $k$ instances.
If not, the algorithm can only randomly guess the label of every unobserved instance.
The detailed construction is included in Appendix~\ref{app:inv-da-help}.

\subsection{The optimal algorithm}\label{subsec:inv-opt}

We show that the optimal sample complexity is characterized by $\vco(\cH,\cG)$.
\begin{theorem}\label{thm:inv-opt}
    For any $\cH, \cG$ with $\vco(\cH,\cG)<\infty$, we have $\Omega(\frac{\vco(\cH,\cG)}{\epsilon} +\frac{1}{\epsilon}\log\frac{1}{\delta})\leq \cM_{\inv}(\epsilon, \delta;\cH,\cG)\leq O(\frac{\vco(\cH,\cG)}{\epsilon}\log \frac{1}{\delta})$.
\end{theorem}
Our algorithm is based on the 1-inclusion-graph predictor by~\cite{haussler1994predicting}.
Given hypothesis class $H$ and instance space $X=\{x_1,\ldots,x_t\}$, the classical 1-inclusion-graph consists of vertices $\{h_{|X}|h\in H\}$, which are labelings of $X$ realized by $H$, and two vertices are connected by an edge if and only if they differ at the labeling of exactly a single $x_i\in X$.
\cite{haussler1994predicting} shows that the edges can be oriented such that each vertex has in-degree at most $\vcd(H)$.
This orientation can be translated to a prediction rule.
Specifically, for any $i\in [t]$, given the labels of all instances in $X$ except $x_i$, if there are two hypotheses $h,h'\in H$ such that their labelings are consistent with the labels of $X\setminus\{x_i\}$ and different at $x_i$, then $h_{|X},h'_{|X}$ are two vertices in the graph and we predict the label of $x_i$ as the edge between $h_{|X},h'_{|X}$ is oriented against.
The average leave-one-out-error is upper bounded by $\frac{\vcd(H)}{t}$.
\begin{lemma}[Theorem~2.3 of \cite{haussler1994predicting}]\label{lmm:1-inclusion}
For any hypothesis class $H$ and instance space $X$ with $\vcd(H)<\infty$, there is a function $Q: (X\times \cY)^*\times X\mapsto \cY$ such that, for any $t\in \NN$ and sample $\{(x_1,y_1),\ldots,(x_t,y_t)\}$ that is realizable w.r.t. $H$,
\begin{align}
    \frac{1}{t!}\sum_{\sigma\in \Sym(t) } \ind{Q(\{x_{\sigma(i)},y_{\sigma(i)}\}_{i\in [t-1]},x_{\sigma(t)})\neq y_{\sigma(t)}}\leq \frac{\vcd(H)}{t}\,,\label{eq:1-inclusion}
\end{align}
where $\Sym(t)$ denotes the symmetric group on $[t]$. The function $Q$ can be constructed by a 1-inclusion-graph predictor.
\end{lemma}
Denote by $Q_{H,X}$ the function guaranteed by Eq~\eqref{eq:1-inclusion} for hypothesis class $H$ and instance space $X$.
For any $t\in \NN$ and $S = \{(x_1,y_1),\ldots,(x_t,y_t)\}$, 
let $X_S$ denote the set of different elements in $S_\cX$.
Define $\cH(X_S):=\{h_{|X_S}|h\in \cH \text{ is } (\cG,X_S) \text{-invariant}\}$ being the set of all possible $(\cG,X_S)$-invariant labelings of $X_S$.
We then define our algorithm $\cA(S)$ by letting $\cA(S)(x) = Q_{\cH(X_S\cup \{x\}), X_{S}\cup \{x\}}(S, x)$ if $\cH(X_S\cup \{x\})\neq \emptyset$ and predicting arbitrarily if $\cH(X_S\cup \{x\})= \emptyset$.
That is to say, $\cA(S)$ needs to construct a function $Q$ for every test example. 
Given any test example, this 1-inclusion-graph-based algorithm takes into account whether the prediction can be invariant over the whole orbit of the test example and thus benefits from the invariance of test examples.
This can provide some theoretical justification for test-time adaptation such as~\cite{wang2020tent}.
By definition, we have $\vcd(\cH(X_S\cup \{x\}))\leq \vco(\cH,\cG)$ for all $S$ and $x$.
Then the expected error of $\cA$ can be bounded by $\vco(\cH,\cG)$ through Lemma~\ref{lmm:1-inclusion}.
We defer the details and the proof of Theorem~\ref{thm:inv-opt} to Appendix~\ref{app:inv-opt}.
Note that the results of Theorem~\ref{thm:inv-opt} also apply to non-group transformations\footnote{In this case, we only assume that $\cG$ contains the identity element.}.

\section{Relaxed realizable setting}\label{sec:re}
In this section, we discuss the results in the relaxed realizable setting (see Definition~\ref{def:re-pac}).
As we can see, DA belongs to the family of algorithms not distinguishing between the original and transformed data. 
In Section~\ref{subsec:dahurts}, we provide a tight characterization $\mu(\cH,\cG)$ (Definition~\ref{def:mu}) on the sample complexity of this family algorithms.
This implies that when $\mu(\cH,\cG)>\vcd(\cH)$, there exists a distribution s.t. DA performs worse than ERM.
We then show that there exists $\cH,\cG$ such that $\mu(\cH,\cG)>\vcd(\cH)$ and the gap can be arbitrarily large.
In Section~\ref{subsec:re-opt}, we provide two optimal algorithms, both of which have to distinguish between the original and transformed data.

\subsection{DA can even ``hurt''}\label{subsec:dahurts}
In the invariantly realizable setting, the optimal algorithm based on 1-inclusion graphs does not need to distinguish between the original and transformed data since $\cH(X_S\cup \{x\})$ in the algorithm is fully determined by the augmented data.
However, in the relaxed realizable setting, distinguishing between the original and transformed data is crucial.
In the following, we will provide a characterization of the sample complexity of algorithms not distinguishing between the original and transformed data, including DA.
Such a characterization induces a sufficient condition when DA ``hurts''.

Let $\cM_\da (\epsilon,\delta;\cH,\cG)$ be the smallest integer $m$ for which there exists a learning rule $\cA$ such that for every $\cG$-invariant data distribution $\cD$, with probability at least $1-\delta$ over $S_\trn\sim \cD^m$, $\err_\cD(\cA(\cG S_\trn))\leq \epsilon$.
The quantity $\cM_\da (\epsilon,\delta;\cH,\cG)$ is the optimal sample complexity achievable if algorithms can only access the augmented data without knowing the original training set.
In standard PAC learning, the optimal sample complexity can be characterized by the maximum density of any subgraph of the 1-inclusion graphs, which is actually equal to the VC dimension~\citep{haussler1994predicting, daniely2014optimal}.
Analogously, we characterize $\cM_\da (\epsilon,\delta;\cH,\cG)$ based on a variant of 1-inclusion graphs, which is constructed as follows.

In the 1-inclusion graph for standard PAC learning~\citep{haussler1994predicting}, given any sequence of instances $\bx = (x_1,\ldots,x_t)$, the vertices are labelings of $\vec x$ and two vertices are connected iff. they are different at only one instance in $\vec x$ and this instance appears once.
In our setting, the input is a multiset of labeled orbits and an unlabeled test instance, hence the vertices are pairs of labelings of orbits and unlabeled instances.
Specifically, for any $t\in \NN$,
given a multi-set of orbits $\bphi = \{\phi_1,\ldots,\phi_t\}$ of some unknown original data,
a labeling $\bff\in \cY^t$ is possible iff. there exists a sequence of instances $\bx = (x_1,\ldots,x_t) \in \prod_{i=1}^t \phi_i$ and a hypothesis $h\in \cH$ such that $h_{|\bx} = \bff$ and that instances in the same orbit are labeled the same,
i.e., $(\cG x_i\times \{1-f_i\})\cap (\{(x_j,f_j)\}_{j\in [t]})=\emptyset$ for all $i\in [t]$.
We denote the set of all possible labelings of $\bphi$ by
\begin{small}
\begin{equation}\label{eq:def-Pi}
    \Pi_\cH(\bphi) := \{\bff\in \cY^t| \exists h\in \cH, \exists \bx\in \prod_{i=1}^t \phi_i, h_{|\bx} = \bff \text{ and }\cup_{i\in [t]}(\cG x_i\times \{1-f_i\})\cap \{(x_j,f_j)|j\in [t]\}=\emptyset\}\,.
\end{equation}
\end{small}
Denote the set of all such sequences $\bx \in \prod_{i=1}^t \phi_i$ of instances that can be labeled as $\bff$ by
\[\cU_{\bff}(\bphi) := \{\bx\in  \prod_{i=1}^t \phi_i|\exists h\in \cH, h_{|\bx} = \bff \text{ and }\cup_{i\in [t]}(\cG x_i\times \{1-f_i\})\cap \{(x_j,f_j)|j\in [t]\}=\emptyset\}\,.\]
Denote the set of all pairs of labeling and its corresponding instance sequence by
\begin{equation}\label{eq:def-B}
    B(\cH,\cG,\bphi):=\cup_{\bff\in \Pi_\cH(\bphi)}\{\bff\}\times \cU_{\bff}(\bphi)\,.
\end{equation}
For any $(\bff,\vec x)\in B(\cH,\cG,\bphi)$, $\vec x$ is a candidate of original data and $\bff$ is a candidate of labeling of $\vec x$.
Now we define a graph $G_{\cH,\cG}(\bphi)$, 
where the vertices are all pairs of labeling $\bff \in \Pi_\cH(\bphi)$ and an element in a instance sequence corresponding to $\bff$. Formally, the vertex set is
\[V = \{(\bff,x_i)|(\bff, \bx)\in B(\cH,\cG,\bphi), i\in [t]\}\,.\]
For every two vertices $(\bff,x)$ and $(\bg,z)$, they are connected if and only if (i) $x=z$; (ii) there exists $j\in[t]$ such that $x\in \phi_j$, $f_i =g_i, \forall i\neq j$ and $f_j \neq g_j$; 
and (iii) $\phi_j$ only appear once in $\bphi$. 
Each edge can be represented by $e=\{\bff,\bg,x\}$ and we denote $E$ the edge set. 
If an edge $e=\{\bff,\bg,x\}$ exists, the edge could be recovered given only $\bff, x$ or $\bg,x$, and thus, we also denote by $e(\bff,x) =e(\bg,x) =\{\bff,\bg,x\}$.
Any algorithm accessing only the augmented data corresponds to an orientation of edges in the graph we constructed, which leads to the following definition.
    
\begin{definition}\label{def:mu}
    Let $w: E\times \Pi_\cH(\bphi)\mapsto [0,1]$ be a mapping such that for every $e=\{\bff,\bg,x\}, w(e,\bff)+w(e,\bg)=1$ and $w(e,\bh)=0$ if $\bh\notin e$ and let $W$ be the set of all such mappings.
    Note that $w$ actually defines a randomized orientation of each edge in graph $G_{\cH,\cG}(\bphi)$: the edge $e=\{\bff,\bg,x\}$ is oriented towards vertex $(\bff,x)$ with probability $w(e,\bff)$.
    For any $(\bff,\bx)\in B(\cH,\cG,\bphi)$, it corresponds to a cluster of vertices $\{(\bff, x_i)|i\in [t]\}$ in $G_{\cH,\cG}(\bphi)$ and $\sum_{i\in [t]: \exists e\in E, \{\bff,x_i\}\subset e} w(e(\bff,x_i),\bff)$ is the expected in-degree of the cluster. 
    Let $\Delta(B(\cH,\cG,\bphi))$ denote the set of all distributions over $B(\cH,\cG,\bphi)$. 
    For any $P\in \Delta(B(\cH,\cG,\bphi))$, we define 
    \begin{equation}\label{eq:mu-P}
        \mu(\cH,\cG,\bphi,P) :=  \min_{w\in W}\EEs{(\bff,\bx)\sim P}{ \sum_{i\in [t]: \exists e\in E, \{\bff,x_i\}\subset e} w(e(\bff,x_i),\bff)}\,.
    \end{equation}
    By taking the supremum over $P$, we define
    $\mu(\cH,\cG,\bphi) := \sup_{P\in \Delta(B(\cH,\cG,\bphi))} \mu(\cH,\cG,\bphi,P)$. By taking supremum over $\bphi$, we define $\mu(\cH,\cG,t) := \sup_{\bphi:\abs{\bphi}=t} \mu(\cH,\cG,\bphi)$
    and
    \begin{equation}
        \mu(\cH,\cG) := \sup_{t\in \NN} \mu(\cH,\cG,t)\,.
    \end{equation}
\end{definition}
\begin{theorem}\label{thm:re-da}
    For any $\cH,\cG$, $\cM_{\da}(\epsilon,\delta;\cH,\cG)$ satisfies the following bounds:
    \begin{itemize}[leftmargin = *]
        \item For all $t\geq 2$ with $\mu(\cH,\cG,t)<\infty$, $\cM_{\da}(\epsilon,\frac{\mu(\cH,\cG,t)}{16(t-1)};\cH,\cG)=\Omega( \frac{\mu(\cH,\cG,t)}{\epsilon})$.
        This implies that if $\mu(\cH,\cG)<\infty$, there exists a constant $c$ dependent on $\cH,\cG$ s.t. $\cM_{\da}(\epsilon,c;\cH,\cG)=\Omega( \frac{\mu(\cH,\cG)}{\epsilon})$.
        \item For all $t$ with $\frac{1}{6}\leq \frac{\mu(\cH,\cG,t)}{t}\leq \frac{1}{3}$, $\cM_{\da}(\epsilon,\delta;\cH,\cG)=O( \frac{\mu(\cH,\cG,t)}{\epsilon} \log^2\frac{\mu(\cH,\cG,t)}{\epsilon} + \frac{1}{\epsilon}\log \frac{1}{\delta})$ .
        \item If $\mu(\cH,\cG)<\infty$, $\cM_{\da}(\epsilon,\delta;\cH,\cG)=O( \frac{\mu(\cH,\cG)\log(1/\delta)}{\epsilon})$.
    \end{itemize}
\end{theorem}
Theorem~\ref{thm:re-da} implies that when $\mu( \cH,\cG)>\vcd(\cH)$, there exists a distribution such that 
any algorithm not differentiating between the original and transformed data performs worse than simply applying ERM over the original data.
We defer the proof of Theorem~\ref{thm:re-da} to Appendix~\ref{app:re-da}.
As we can see, the definition of $\mu(\cH,\cG)$ is not intuitive and it might be difficult to calculate $\mu(\cH,\cG)$ as well as to further determine when $\mu(\cH,\cG)>\vcd(\cH)$. 
We introduce a new dimension as follows, which lower bounds $\mu(\cH,\cG)$ and is easier to calculate.
\begin{definition}[VC Dimension of orbits generated by $\cH$]\label{def:dim}
    The VC dimension of orbits generated by $\cH$, denoted $\dim(\cH, \cG)$, 
    is defined as the largest integer $k$ for which there exists a set $X=\{x_1,\ldots,x_k\}\subset \cX$ such that (i) their orbits $\bphi = \{\phi_1,\ldots,\phi_k\}$ are pairwise disjoint, (ii) $\Pi_\cH(\bphi) = \cY^k$ (defined in Eq~\eqref{eq:def-Pi}) and (iii) there exists a set $B=\{(\bff, \bx_\bff)\}_{\bff\in \cY^k}\subset B(\cH,\cG,\bphi)$ (defined in Eq~\eqref{eq:def-B}) such that $\bff \oplus \bg = \be_i$ implies $x_{\bff,i} = x_{\bg,i}$ for all $i\in [k], \bff,\bg\in \cY^k$.
\end{definition}
\begin{theorem}\label{thm:re-mu-dim}
    For any $\cH,\cG$, $\mu(\cH,\cG)\geq {\dim(\cH,\cG)}/{2}$.
\end{theorem}
The proof is included in Appendix~\ref{app:re-mu-dim}.
Through this dimension, 
we claim that the gap between $\mu(\cH,\cG)$ and $\vcd(\cH)$ can be arbitrarily large.
In the following, we give an example of $\cH,\cG$ with $\dim(\cH,\cG) \gg \vcd(\cH)$, which cannot be learned by DA but can be easily learned by ERM.
\begin{example}
    For any $d>0$, let $\cX = \{\pm 1\}\times \{\be_i|i\in [2d]\}\subset \R^{2d+1}$, $\cH = \{x_1> 0, x_1\leq 0\}$ and $\cG = \{I_{2d+1}, I_{2d+1}-2\diag(\be_1)\}$ (i.e., the cyclic group generated by flipping the sign of $x_1$).
    It is easy to check that $\vcd(\cH) = 1$.
    Let $X = \{(1,\be_1), (1,\be_2),\ldots,(1,\be_{2d})\}$ and then the orbits generated from $X$ are $\bphi = \{\{(-1,\be_i),(1,\be_i)\}|i\in [2d]\}\}$.
    For every labeling $\bff\in \cY^{2d}$, if $\sum_{i\in [2d]}f_i$ is odd, let $\bx_\bff = ((2f_i-1,\be_i))_{i=1}^{2d}$; if $\sum_{i\in [2d]}f_i$ is even, let $\bx_f = ((1-2f_i,\be_i))_{i=1}^{2d}$.
    It is direct to check that $(\bff,\bx_\bff)\in B(\cH,\cG,\bphi)$ for all $\bff\in \cY^{2d}$.
    Then for all $i\in [2d]$, $\bff\oplus \bg=\be_i$ implies $x_{\bff,i}=x_{\bg,i}$.
    Hence, $\dim(\cH,\cG) = 2d$, where $d$ can be an arbitrary positive integer.
    According to Theorem~\ref{thm:re-mu-dim}, we have $\mu(\cH,\cG)\geq d$.
\end{example}
The above example can be interpreted in a vision scenario. Let's consider an example of classifying land birds versus water birds.
The natural data is $2d$ images of land birds with land background and water birds with water background. The transformation set is composed of keeping the current background and changing the background from land (water) to water (land). 
Consider simple hypotheses depending on backgrounds only.
Specifically, $\cH = \{h_1,h_2\}$ with $h_1$ predicting all images with water background as water birds and $h_2$ predicting all images with water background as land birds.
Let the data distribution be the uniform distribution over all the original images.
Then given any training data, $h_1$ and $h_2$ have the same empirical loss on the augmented training data.
Thus, for any unobserved image, DA will make a mistake with constant probability. Hence DA requires at least $\Omega(d)$ sample complexity.
It is direct to check that standard ERM only needs one labeled instance to achieve zero error.

\textbf{Open question: }
It is unclear whether $\mu(\cH,\cG)$ is upper bounded by $\dim(\cH,\cG)$.
If true, then we can tightly characterize $\cM_\da (\epsilon,\delta;\cH,\cG)$ by $\dim(\cH,\cG)$.

\subsection{The optimal algorithms}\label{subsec:re-opt}
Different from the invariantly realizable setting, the optimal sample complexity in the relaxed realizable setting is characterized by $\vcao(\cH,\cG)$.
The optimal (up to log-factors) sample complexity can be achieved by another variant of 1-inclusion-graph predictor.
Besides, we propose an ERM-based algorithm, called ERM-INV (see Appendix~\ref{app:re-opt} for details), achieving the similar guarantee.

\begin{theorem}\label{thm:re-opt}
    For any $\cH,\cG$ with $\vcao(\cH,\cG)<\infty$, we have $\Omega(\frac{\vcao(\cH,\cG)}{\epsilon} + \frac{\log(1/\delta)}{\epsilon})\leq \cM_{\re}(\epsilon,\delta;\cH,\cG) \leq O\left(\min\left( \frac{\vcao(\cH,\cG)}{\epsilon}\log^3\frac{\vcao(\cH,\cG)}{\epsilon}+\frac{1}{\epsilon}\log \frac 1 \delta, \frac{\vcao(\cH,\cG)}{\epsilon}\log \frac{1}{\delta}\right)\right)$.
\end{theorem}
We defer the details of algorithms and the proof of Theorem~\ref{thm:re-opt} to Appendix~\ref{app:re-opt}.
Usually, ERM-INV is more efficient than the 1-inclusion-graph predictor.
But the 1-inclusion-graph predictor as well as the lower bound can apply to non-group transformations.
Another advantage of 1-inclusion-graph predictor is allowing us to design an adaptive framework which automatically adjusts to different levels of invariance of $h^*$.
Specifically, for any hypothesis $h\in \cH$, we say $h$ is $(1-\eta)$-invariant over the distribution $\cD_{\cX}$ for some $\eta\in [0,1]$ if $\PPs{x\sim \cD_{\cX}}{\exists x'\in \cG x, h(x')\neq h(x)}= \eta$.
When $\eta(h^*)=0$, it degenerates into the invariantly realizable setting, which implies that we can achieve better bounds when $\eta(h^*)$ is smaller.
We propose an adaptive algorithm with sample complexity dependent on $\eta(h^*)$ and the details are included in Appendix~\ref{app:unified-re}.

\section{Agnostic setting}
In the agnostic setting (Definition~\ref{def:ag-pac}), $\inf_{h\in \cH} \err(h)$ is possibly non-zero.
Different from the agnostic setting in the standard PAC learning allowing probabilistic labels, our problem is limited to deterministic labels 
because we assume that the data distribution is $\cG$-invariant, i.e., there exists a $(\cG,\cD_\cX)$-invariant hypothesis $f^*$ (possibly not in $\cH$) with $\err_\cD(f^*) = 0$.

\begin{theorem}\label{thm:ag-opt}
    The sample complexity in the agnostic setting satisfies:
    \begin{itemize}[leftmargin = *]
        \item For all $d>0$, there exists $\cH,\cG$ with $\vcao(\cH,\cG)=d$, $\cM_\ag(\epsilon,1/64;\cH,\cG)= \Omega(\frac{d}{\epsilon^2})$.
        \item For any $\cH,\cG$ with $\vcao(\cH,\cG)<\infty$, $\cM_\ag(\epsilon,\delta;\cH,\cG)= \tilde O\left(\frac{\vcao(\cH,\cG)}{\epsilon^2}+\frac{1}{\epsilon^2}\log(\frac 1 \delta)\right)$.
    \end{itemize}
\end{theorem}
For upper bound, we show that ERM-INV achieves sample complexity $\tilde O(\frac{\vcao(\cH,\cG)}{\epsilon^2})$.
There is another way of achieving similar upper bound based on applying the reduction-to-realizable technique of \cite{david2016supervised}.
Note that a direct combination of any reduction-to-realizable technique and any optimal algorithm in relaxed realizable setting does not work in our agnostic setting.
This is because the relaxed realizable setting requires not only realizability, but also invariance in the support of the data distribution.
For example, the reduction method of \cite{hopkins2021realizable} needs to run a realizable algorithm over a set labeled by each $h\in \cH$, which might label two instances in the same orbit differently and make the realizable algorithm not well-defined.
When combining the reduction method of \cite{david2016supervised} and the 1-inclusion-graph-type algorithm, the similar problem also exists but can be fixed by predicting arbitrarily when the invariance property is not satisfied.
For lower bound, According to \cite{ben2014sample}, the sample complexity of agnostic PAC learning under deterministic labels is not fully determined by the VC dimension.
Following the construction by \cite{ben2014sample}, we provide an analogous lower bound in our setting.
The algorithm details and the proofs are deferred to Appendix~\ref{app:ag-opt}.
Analogous to the realizable setting, we provide one algorithm adapting to different levels of invariance of the optimal hypothesis in $\cH$ in Appendix~\ref{app:unified-ag}.
Similar to the results in the realizable settings, the lower bound and the 1-inclusion-graph predictor in the agnostic setting also apply to non-group transformations.

\section{Discussion}
\textbf{Definition of invariance under probabilistic labels} 
In this work, we model invariance by assuming that the data distribution is $\cG$-invariant, which restricts the labels to be deterministic.
It is unclear what ``invariance under probabilistic labels'' means.
One option is assuming that the distribution of the labels is invariant over the orbits, $\Pr(y|g x)=\Pr(y|x)$ for all $g\in \cG, x\in \cX$.
However, such a condition may not characterize invariance in real-world scenarios due to classes having different underlying distributions.
For example, given a fuzzy image with probability $0.5$ being a car and $0.5$ being a tree, it is uncertain if the chance of this image being a car is still $0.5$ after rotation.

\textbf{The performance of DA under non-group transformations}
Most results of DA and ERM-type algorithms only hold when the transformation set is a group.
If we regard adversarial training as a special type of data augmentation through a ball around the natural data, then the transformation set is not a group.
The appropriate way to formulate theoretical guarantees for DA under arbitrary transformations is still an open question.

\chapter{Future Directions}
I am interested in developing theoretical foundations for challenges in machine learning that are not well-modeled by existing theory.  These include:

\paragraph{Imperfect Social Behaviors}
Applying concepts and methods from algorithmic game theory to comprehend interactions within large interactive systems, often necessitates an accurate model of human behavior. For instance, in strategic classification, the existing study requires individuals to accurately perceive the implemented classifier and take the exact action maximizing their utility (which is the benefit of receiving a positive prediction minus the cost of manipulation). This might not always hold in reality, as the implemented classifier may not be disclosed and the individuals might only be able to obtain an approximate optimal action due to computational limitations.
In incentivized collaboration within single-round federated learning, it is often assumed that self-interested data collectors will optimize their contributions. However, in reality, it is not clear if collectors' utility (which is the benefit of receiving an accurate model minus the cost of data contribution) can be expressed as a function of their contribution levels, let alone optimized.
For instance, what is the accuracy of the output model given each data contributor's contribution level? 
Additionally, optimizing contributions relies on information about other data collectors, which might not reflect real-world scenarios accurately. Hence, it presents a question: \textit{what happens if humans can't execute perfect social behavior due to restricted information and computational resources?}
I'm enthusiastic about delving into this challenge of modeling imperfect social behaviors to enhance the trustworthiness of machine learning systems operating in imperfect social contexts.

\vspace{-0.9em}

\paragraph{Restrictive Adversarial Attacks and Detectability} 
Much like the limitations faced by clean-label attackers--restricted to inserting accurately labeled points--and label-invariant transformation attackers--constrained to using label-invariant transformations--many real-world attackers have specific boundaries within which they operate. 
For example, attackers providing data to influence financial or legal decision-making might be criminally liable if they fabricate data, but safe if they are just selective in which data they provide.  This prompts the following question:
\textit{What restrictions of an attacker enable the development of robustness, and how can we devise a robust algorithm?} 
On the other hand, regarding detectability, the evolution of machine learning allows attackers to create fake data--such as counterfeit images--challenging to differentiate from authentic data and capable of deceiving humans. This raises another query: \textit{What attributes of an attacker make it hard to detect, and how do we detect the fake data generated by attackers?}

\vspace{-0.9em}

\paragraph{Societal Dimensions in Machine Learning}
Beyond strategic behaviors and adversarial attacks, I am interested in investigating other societal issues within machine learning, such as fairness and privacy.
Moreover, the surge of machine learning has posed numerous new challenges to existing theories and problems. The widespread adoption of AI interfaces by individuals and organizations is reshaping human behavior, unveiling new societal concerns, such as the copyright of AI-generated content. It also alters fundamental assumptions in classical theories. 
For instance, conventional game theory often assumes rational agents will act by best responding, whereas machine learning introduces a new modeling approach--agents employing learning algorithms to optimize their utilities.
I am enthusiastic about delving into these varied topics in my future research.

\appendix
\chapter{Strategic Classification for Finite Hypothesis Class}\label{app:strategic-finite}
\section{Technical Lemmas}
\subsection{Boosting expected guarantee to high probability guarantee}\label{subsec:boost}
Consider any (possibly randomized) PAC learning algorithm $\cA$ in strategic setting, which can output a predictor $\cA(S)$ after $T$ steps of interaction with i.i.d. agents $S\sim \cD^T$ s.t. $\EE{\err(\cA(S))}\leq \epsilon$, where the expectation is taken over both the randomness of $S$ and the randomness of algorithm.
One standard way in classic PAC learning of boosting the expected loss guarantee to high probability loss guarantee is: running $\cA$ on new data $S$ and verifying the loss of $\cA(S)$ on a validation data set; if the validation loss is low, outputting the current $\cA(S)$, and repeating this process otherwise.

We will adopt this method to boost the confidence as well. The only difference in our strategic setting is that we can not re-use validation data set as we are only allowed to interact with the data through the interaction protocol. Our boosting scheme is described in the following.
\begin{itemize}
    \item For round $r = 1, \ldots, R,$
    \begin{itemize}
        \item Run $\cA$ for $T$ steps of interactions to obtain a predictor $h_r$.
        \item Apply $h_r$ for the following $m_0$ rounds to obtain the empirical strategic loss on $m_0$, denoted as $\hat l_r =\frac{1}{m_0}\sum_{t=t_r+1}^{t_r+m_0} \lossstr( h_r,(x_t,r_t,y_t))$, where $t_r+1$ is the starting time of these $m_0$ rounds.
        \item Break and output $h_r$ if $\hat l_r\leq 4\epsilon$.
    \end{itemize}
    \item If for all $r\in [R]$, $\hat l_r > 4\epsilon$, output an arbitrary hypothesis.
\end{itemize}
\begin{lemma}\label{lmm:boost}
Given an algorithm $\cA$, which can output a predictor $\cA(S)$ after $T$ steps of interaction with i.i.d. agents $S\sim \cD^T$ s.t. the expected loss satisfies $\EE{\err(\cA(S))}\leq \epsilon$. Let $h_\cA$ denote the output of the above boosting scheme given algorithm $\cA$ as input. By setting $R= \log \frac{2}{\delta}$ and $m_0= \frac{3\log(4R/\delta)}{2\epsilon}$, we have $\err(h_\cA) \leq 8\epsilon$ with probability $1-\delta$. The total sample size is $R(T+m_0) =\cO(\log(\frac{1}{\delta})(T +
    \frac{\log(1/\delta)}{\epsilon}))$. 
\end{lemma}
\begin{proof}
For all $r=1,\ldots,R$, we have $\EE{\err(h_r)} \leq \epsilon$.
By Markov's inequality, we have
\begin{align*}
    \Pr(\err(h_r) >2\epsilon)\leq \frac{1}{2}\,.
\end{align*}
For any fixed $h_r$, if $\err(h_r)\geq 8\epsilon$, we will have $\hat l_r \leq 4\epsilon$ with probability $\leq e^{-m_0\epsilon}$; if $\err(h_r)\leq 2\epsilon$, we will have $\hat l_r \leq 4\epsilon$ with probability $\geq 1- e^{-2m_0\epsilon/3}$ by Chernoff bound.

Let $E$ denote the event of \{$\exists r\in [R], \err(h_r)\leq 2\epsilon$\} and $F$ denote the event of \{$\hat l_r > 4\epsilon$ for all $r\in [R]$\}.
When $F$ does not hold, our boosting will output $h_r$ for some $r\in [R]$.
\begin{align*}
    &\Pr(\err(h_\cA)>8\epsilon) \\
    \leq &  \Pr(E,\neg F)\Pr(\err(h_\cA)>8\epsilon | E, \neg F) + \Pr(E, F)+\Pr(\neg E) \\
    \leq & \sum_{r=1}^R \Pr(h_\cA = h_r,\err(h_r)>8\epsilon|E,\neg F)
    + \Pr(E, F)+\Pr(\neg E)\\
    \leq & Re^{-m_0\epsilon} + e^{-2m_0\epsilon/3} + \frac{1}{2^R}\\
   \leq  & \delta\,,
\end{align*}
by setting $R= \log \frac{2}{\delta}$ and $m_0= \frac{3\log(4R/\delta)}{2\epsilon}$.
\end{proof}

\subsection{Converting mistake bound to PAC bound}\label{app:mistake2pac}
In any setting of $(C,F)$, if there is an algorithm $\cA$ that can achieve the mistake bound of $B$, then we can convert $\cA$ to a conservative algorithm by not updating at correct rounds. The new algorithm can still achieve mistake bound of $B$ as $\cA$ still sees a legal sequence of examples.
Given any conservative online algorithm, we can convert it to a PAC learning algorithm using the standard longest survivor technique~\citep{10008965845}.
\begin{lemma}\label{lmm:mistake2pac}
    In any setting of $(C,F)$, given any conservative algorithm $\cA$ with mistake bound $B$, let algorithm $\cA'$ run $\cA$ and output the first $f_t$ which survives over $\frac{1}{\epsilon}\log(\frac{B}{\delta})$ examples.
    $\cA'$ can achieve sample complexity of 
    $\cO(\frac{B}{\epsilon}\log(\frac{B}{\delta}))$.
\end{lemma}
    \begin{proof}[Proof of Lemma~\ref{lmm:mistake2pac}]
        When the sample size $m\geq \frac{B}{\epsilon}\log(\frac{B}{\delta})$, the algorithm $\cA$ will produce at most $B$ different hypotheses and there must exist one surviving for $\frac{1}{\epsilon}\log(\frac{B}{\delta})$ rounds since $\cA$ is a conservative algorithm with at most $B$ mistakes.
        Let $h_1,\ldots,h_B$ denote these hypotheses and let $t_1,\ldots,t_B$ denote the time step they are produced.
        Then we have
        \begin{align*}
            &\Pr(f_\out = h_i \text{ and } \err(h_i) >\epsilon) = \EE{\Pr(f_\out = h_i \text{ and } \err(h_i) >\epsilon|t_i, z_{1:t_i-1})} \\
            <& \EE{(1-\epsilon)^{\frac{1}{\epsilon}\log(\frac{B}{\delta})}} = \frac{\delta}{B}\,.
        \end{align*}
        By union bound, we have
        \begin{align*}
            \Pr(\err(f_\out)>\epsilon) \leq \sum_{i=1}^B \Pr_{z_{1:T}}(f_\out = h_i \text{ and } \err(h_i) >\epsilon)<\delta.
        \end{align*}  
        We are done.
    \end{proof}

\subsection{Smooth the distribution}
\begin{lemma}\label{lmm:smooth}
    For any two data distribution $\cD_1$ and $\cD_2$, let $\cD_3 = (1-p)\cD_1+ p\cD_2$ be the mixture of them.
    For any setting of $(C,F)$ and any algorithm, let $\bP_{\cD}$ be the dynamics of $(C(x_1), f_1, y_1,\hat y_1, F(x_1,\Delta_1),\ldots, C(x_T), f_T, y_T,\hat y_T, F(x_T,\Delta_T))$ under the data distribution $\cD$.
    Then for any event $A$, we have $\abs{\bP_{\cD_3}(A) - \bP_{\cD_1}(A)}\leq 2pT$.
\end{lemma}
\begin{proof}
    Let $B$ denote the event of all $(x_t,u_t,y_t)_{t=1}^T$ being sampled from $\cD_1$. 
    Then $\bP_{\cD_3}(\neg B) \leq  pT$.
    Then 
    \begin{align*}
        \bP_{\cD_3}(A) &= \bP_{\cD_3}(A|B)\bP_{\cD_3}(B)+ \bP_{\cD_3}(A|\neg B)\bP_{\cD_3}(\neg B) \\
        & = \bP_{\cD_1}(A)\bP_{\cD_3}(B)+ \bP_{\cD_3}(A|\neg B)\bP_{\cD_3}(\neg B)\\
        & = \bP_{\cD_1}(A)(1-\bP_{\cD_3}(\neg B))+ \bP_{\cD_3}(A|\neg B)\bP_{\cD_3}(\neg B)\,.
    \end{align*}
    By re-arranging terms, we have
    \begin{align*}
        \abs{\bP_{\cD_1}(A) - \bP_{\cD_3}(A)} = \abs{\bP_{\cD_1}(A)\bP_{\cD_3}(\neg B)- \bP_{\cD_3}(A|\neg B)\bP_{\cD_3}(\neg B)}\leq 2pT\,.
    \end{align*}
\end{proof}

\section{Proof of Theorem~\ref{thm:halving}}\label{app:halving}

\begin{proof}
When a mistake occurs, there are two cases.
\begin{itemize}
    \item If $f_t$ misclassifies a true positive example $(x_t,r_t, +1)$ by negative, we know that $d(x_t, f_t)> r_t$ while the target hypothesis $h^*$ must satisfy that $d(x_t, h^*)\leq r_t$. Then any $h\in \VS$ with $d(x_t,h)\geq d(x_t,f_t)$ cannot be $h^*$ and are eliminated. Since $d(x_t,f_t)$ is the median of $\{d(x_t,h)|h\in \VS\}$, we can eliminate half of the version space.
    \item If $f_t$ misclassifies a true negative example $(x_t,r_t, -1)$ by positive, we know that $d(x_t, f_t)\leq r_t$ while the target hypothesis $h^*$ must satisfy that $d(x_t, h^*)> r_t$. Then any $h\in \VS$ with $d(x_t,h)\leq d(x_t,f_t)$ cannot be $h^*$ and are eliminated. Since $d(x_t,f_t)$ is the median of $\{d(x_t,h)|h\in \VS\}$, we can eliminate half of the version space.
\end{itemize}
Each mistake reduces the version space by half and thus, the algorithm of Strategic Halving suffers at most $\log_2(\abs{\cH})$ mistakes.
\end{proof}
\section{Proof of Theorem~\ref{thm:x-first-pac}}\label{app:x_firtst_pac}
\begin{proof}
In online learning setting, an algorithm is conservative if it updates it's current predictor only when making a mistake.
It is straightforward to check that Strategic Halving is conservative. Combined with the technique of converting mistake bound to PAC bound in Lemma~\ref{lmm:mistake2pac}, we prove Theorem~\ref{thm:x-first-pac}.
\end{proof}
\section{Proof of Theorem~\ref{thm:x-end-online}}\label{app:x-end-online}
\begin{proof}
Consider the feature space $\cX = \{\bZero, \be_1,\ldots,\be_n, 0.9\be_1,\ldots,0.9\be_n\}$, where $\be_i$'s are standard basis vectors in $\R^n$ and metric
$d(x,x') = \norm{x-x'}_2$ for all $x,x'\in \cX$.
Let the hypothesis class be a set of singletons over $\{\be_i|i\in [n]\}$, i.e., $\cH =\{2\ind{\{\be_i\}}-1|i\in [n]\}$.
We divide all possible hypotheses (not necessarily in $\cH$) into three categories:
\begin{itemize}
    \item The hypothesis $2\ind{\emptyset}-1$, which predicts all negative.
    \item For each $x\in \{\bZero, 0.9\be_1,\ldots,0.9\be_n\}$, let
    $F_{x,+}$ denote the class of hypotheses $h$ predicting $x$ as positive.   
    \item For each $i\in [n]$, let $F_i$ denote the class of hypotheses $h$ satisfying $h(x)=-1$ for all $x\in \{\bZero, 0.9\be_1,\ldots,0.9\be_n\}$ and $ h(\be_i) = +1$. And let $F_* = \cup_{i\in [n]} F_i$ denote the union of them.
\end{itemize}
Note that all hypotheses over $\cX$ fall into one of the three categories.

Now we consider a set of adversaries $E_1,\ldots,E_n$, such that the target function in the adversarial environment $E_i$ is $2\ind{\{\be_i\}}-1$.
We allow the learners to be randomized and thus, at round $t$, the learner draws an $f_t$ from a distribution $D(f_t)$ over hypotheses. 
The adversary, who only knows the distribution $D(f_t)$ but not the realization $f_t$, picks an agent $(x_t,r_t,y_t)$ in the following way. 
\begin{itemize}
    \item  Case 1: If there exists $x\in \{\bZero, 0.9\be_1,\ldots,0.9\be_n\}$ such that $\Pr_{f_t\sim D(f_t)}(f_t\in F_{x,+}) \geq c$ for some $c>0$,  then for all $j\in [n]$, the adversary $E_j$ picks $(x_t,r_t,y_t) = (x,0,-1)$. 
    Let $B^t_{1,x}$ denote the event of $f_t\in F_{x,+}$.
    \begin{itemize}
        \item In this case, the learner will make a mistake with probability $c$. Since for all $h\in \cH$, $h(\Delta(x,h,0)) = h(x) = -1$, they are all consistent with $(x,0,-1)$.
    \end{itemize}
    \item Case 2: If $\Pr_{f_t\sim D(f_t)}(f_t = 2\ind{\emptyset}-1)\geq c$, then for all $j\in [n]$, the adversary $E_j$ picks $(x_t,r_t,y_t) = (\bZero,1,+1)$. Let $B^t_2$ denote the event of $f_t = 2\ind{\emptyset}-1$.
    \begin{itemize}
        \item In this case, with probability $c$, the learner will sample a $f_t = 2\ind{\emptyset}-1$ and misclassify $(\bZero,1,+1)$.
    Since for all $h\in \cH$, $h(\Delta(\bZero,h,1)) = +1$, they are all consistent with $(\bZero,1,+1)$.
    \end{itemize}
    \item Case 3: If the above two cases do not hold, let $i_t = \argmax_{i\in [n]} \Pr(f_t(\be_i) =1|f_t\in F_*)$, $x_t = 0.9\be_{i_t}$.
    For radius and label, different adversaries set them differently. Adversary $E_{i_t}$ will set $(r_t,y_t) = (0,-1)$ while other $E_j$ for $j\neq i_t$ will set $(r_t,y_t) = (0.1, -1)$. Since Cases 1 and 2 do not hold, we have $\Pr_{f_t\sim D(f_t)}(f_t\in F_*)\geq 1-(n+2)c$. Let $B^t_3$ denote the event of $f_t\in F_*$ and $B^t_{3,i}$ denote the event of $f_t\in F_i$.
    \begin{itemize}
        \item[(a)] With probability $\Pr(B^t_{3,i_t})\geq \frac{1}{n}\Pr(B^t_3) \geq \frac{1-(n+2)c}{n}$, the learner samples a $f_t \in F_{i_t}$, and thus misclassifies $(0.9\be_{i_t},0.1,-1)$ in $E_j$ for $j\neq i_t$ but correctly classifies $(0.9\be_{i_t},0,-1)$.
        In this case, the learner observes the same feedback in all $E_j$ for $j\neq i_t$ and identifies the target function $2\ind{\{\be_{i_t}\}}-1$ in $E_{i_t}$.
        \item[(b)] If the learner samples a $f_t$ with $f_t(\be_{i_t}) =f_t(0.9\be_{i_t}) = -1$, then the learner observes $x_t = 0.9\be_{i_t}$, $y_t = -1$ and $\hat y_t =-1$ in all $E_j$ for $j\in [n]$. Therefore the learner cannot distinguish between adversaries in this case.
        \item[(c)] If the learner samples a $f_t$ with $f_t(0.9\be_{i_t}) = +1$, then the learner observes $x_t = 0.9\be_{i_t}$, $y_t = -1$ and $\hat y_t =+1$ in all $E_j$ for $j\in [n]$.
        Again, since the feedback are identical in all $E_j$ and the learner cannot distinguish between adversaries in this case.
    \end{itemize}
    \end{itemize}
For any learning algorithm $\cA$, his predictions are identical in all of  adversarial environments $\{E_j|j\in [n]\}$ before he makes a mistake in Case 3(a) in one environment $E_{i_t}$.
His predictions in the following rounds are identical in all of  adversarial environments $\{E_j|j\in [n]\}\setminus \{E_{i_t}\}$ before he makes another mistake in Case 3(a).
Suppose that we run $\cA$ in all adversarial environment of $\{E_j|j\in [n]\}$ simultaneously.
Note that once we make a mistake, the mistake must occur simultaneously in at least $n-1$ environments.
Specifically, if we make a mistake in Case 1, 2 or 3(c), such a mistake simultaneously occur in all $n$ environments.
If we make a mistake in Case 3(a), such a mistake simultaneously occur in all $n$ environments 
except $E_{i_t}$.
Since we will make a mistake with probability at least $\min(c, \frac{1-(n+2)c}{n})$ at each round, there exists one environment in $\{E_j|j\in [n]\}$ in which $\cA$ will make $n-1$ mistakes.

Now we lower bound the number of mistakes dependent on $T$.
Let $t_1,t_2,\ldots$ denote the time steps in which we makes a mistake. Let $t_0=0$ for convenience.
Now we prove that
\begin{align*}
    &\Pr(t_i > t_{i-1} + k|t_{i-1}) = \prod_{\tau = t_{i-1}+1}^{t_{i-1} + k} \Pr(\text{we don't make a mistake in round } \tau)\\
    \leq & \prod_{\tau = t_{i-1}+1}^{t_{i-1} + k} (\1{\text{Case 3 at round }\tau} (1- \frac{1-(n+2)c}{n}) + \1{\text{Case 1 or 2 at round }\tau} (1- c))\\
    \leq &
    (1-\min(\frac{1-(n+2)c}{n}, c))^k \leq (1-\frac{1}{2(n+2)})^k\,,
\end{align*}
by setting $c = \frac{1}{2(n+2)}$.
Then by letting $k = 2(n+2)\ln(n/\delta)$, we have $$\Pr(t_i > t_{i-1} + k|t_{i-1}) \leq \delta/n\,.$$
For any $T$,
\begin{align*}
    &\Pr(\text{\# of mistakes} <\min(\frac{T}{k+1}, n-1))\\
    =\leq & \Pr(\exists i\in [n-1], t_i-t_{i-1}> k) \\
    \leq& \sum_{i=1}^{n-1} \Pr(t_i-t_{i-1}> k)\leq \delta\,.
\end{align*}
Therefore, we have proved that for any $T$, with probability at least $1-\delta$, we will make at least $\min(\frac{T}{2(n+2)\ln(n/\delta)+1}, n-1)$ mistakes.
\end{proof}

\section{Proof of Theorem~\ref{thm:mw}}\label{app:mw}
\begin{algorithm}[H]\caption{MWMR (Multiplicative Weights on Mistake Rounds)}\label{alg:mw}
    \begin{algorithmic}[1]
    \STATE Initialize the version space $\VS=\cH$.
    \FOR{t=1,\ldots,T}
    \STATE Pick one hypotheses $f_t$ from $\VS$ uniformly at random.
    \IF{$\hat y_t \neq y_t$ and $y_t = +$}
    \STATE $\VS\leftarrow \VS\setminus \{h\in \VS|d(x_t,h)\geq d(x_t,f_t)\}$.
    \ELSIF{$\hat y_t \neq y_t$ and $y_t = -$}
    \STATE $\VS\leftarrow \VS \setminus \{h\in \VS|d(x_t,h)\leq d(x_t,f_t)\}$.
    \ENDIF
    \ENDFOR
    \end{algorithmic}
\end{algorithm}
\begin{proof}
    First, when the algorithm makes a mistake at round $t$, he can at least eliminate $f_t$. Therefore, the total number of mistakes will be upper bounded by $\abs{\cH}-1$.
    
    Let $p_t$ denote the fraction of hypotheses misclassifying $x_t$.
    We say a hypothesis $h$ is inconsistent with $(x_t,f_t, y_t, \hat y_t)$ iff $(d(x_t,h)\geq d(x_t,f_t)\wedge \hat y_t = - \wedge y_t = +)$ or  $(d(x_t,h)\leq d(x_t,f_t)\wedge \hat y_t = + \wedge y_t = -)$.
    Then we define the following events.
    \begin{itemize}
        \item $E_t$ denotes the event that MWMR makes a mistake at round $t$. We have $\Pr(E_t) = p_t$.
        \item $B_t$ denotes the event that at least $\frac{p_t}{2}$ fraction of hypotheses are inconsistent with $(x_t,f_t, y_t, \hat y_t)$. We have $\Pr(B_t|E_t)\geq \frac{1}{2}$.
    \end{itemize}
    Let $n =\abs{\cH}$ denote the cardinality of hypothesis class and $n_t$ denote the number of hypotheses in $\VS$ after round $t$.
    Then we have 
    \[1\leq n_T = n\cdot \prod_{t=1}^T (1- \1{E_t}\1{B_t}\frac{p_t}{2})\,.\]
    By taking logarithm of both sides, we have
    \begin{align*}
        0\leq \ln(n_T) = \ln(n) +\sum_{t=1}^T \ln(1- \1{E_t}\1{B_t}\frac{p_t}{2})\leq \ln(n) -\sum_{t=1}^T \1{E_t}\1{B_t}\frac{p_t}{2}\,,
    \end{align*}
    where the last inequality adopts $\ln(1-x)\leq -x$ for $x\in [0,1)$.
    Then by taking expectation of both sides,
    we have
    \begin{equation*}
        0\leq \ln(n) - \sum_{t=1}^T \Pr(E_t \wedge B_t)\frac{p_t}{2}\,.
    \end{equation*}
    Since $\Pr(E_t) = p_t$ and $\Pr(B_t|E_t)\geq \frac{1}{2}$, then we have
    \begin{align*}
        \frac{1}{4}\sum_{t=1}^T p_t^2\leq \ln(n)\,.
    \end{align*}
    Then we have the expected number of mistakes $\EE{\cM_\MW(T)}$ as
    \begin{equation*}
        \EE{\cM_\MW(T)} = \sum_{t=1}^T p_t \leq \sqrt{\sum_{t=1}^T p_t^2 } \cdot \sqrt{T}\leq \sqrt{4\ln(n)T}\,,
    \end{equation*}
    where the first inequality applies Cauchy-Schwarz inequality.
\end{proof}
\section{Proof of Theorem~\ref{thm:x-end-pac-proper}}\label{app:x-end-pac-proper}
\begin{proof}
\textbf{Construction of $\cQ,\cH$ and a set of realizable distributions}
\begin{itemize}
    \item Let feature space $\cX = \{\bZero, \be_1,\ldots, \be_n\}\cup X_0$, where $X_0 = \{\frac{\sigma(0,1,\ldots,n-1)}{z}|\sigma\in \cS_n\}$ with $z = \frac{\sqrt{1^2+\ldots+(n-1)^2}}{\alpha}$ for some small $\alpha=0.1$.
    Here $\cS_n$ is the set of all permutations over $n$ elements.
    So $X_0$ is the set of points whose coordinates are a permutation of $\{0,1/z,\ldots,(n-1)/z\}$ and all points in $X_0$  have the  $\ell_2$ norm equal to $\alpha$.
    Define a metric $d$ by letting $d(x_1,x_2) = \norm{x_1-x_2}_2$ for all $x_1,x_2\in \cX$.
    Then for any $x\in X_0$ and $i\in [n]$, $d(x,\be_i) = \norm{x-\be_i}_2 = \sqrt{(x_i-1)^2 +\sum_{j\neq i} x_j^2} = \sqrt{1+\sum_{j=1}^n x_j^2 - 2x_i} = \sqrt{1+\alpha^2-2x_i}$.
    Note that we consider space $(\cX,d)$ rather than $(\R^n, \norm{\cdot}_2)$.
    \item Let the hypothesis class be a set of singletons over $\{\be_i|i\in [n]\}$, i.e., $\cH = \{2\ind{\{\be_i\}}-1|i\in [n]\}$.
    \item We now define a collection of distributions $\{\cD_i|i\in [n]\}$ in which $\cD_i$ is realized by $2\ind{\{\be_i\}}-1$.
    For any $i\in [n]$, $\cD_i$ puts probability mass $1-3n\epsilon$ on $(\bZero, 0, -1)$.
    For the remaining $3n\epsilon$ probability mass, $\cD_i$ picks $x$ uniformly at random from $X_0$ and label it as positive.
    If $x_{i} =0$, set radius $r(x) = r_u := \sqrt{1+\alpha^2}$; otherwise, set radius $r(x) = r_l := \sqrt{1+\alpha^2 -2\cdot \frac{1}{z})}$.
    Hence, $X_0$ are all labeled as positive.
    For $j\neq i$, $h_j = 2\ind{\{\be_j\}}-1$ labels $\{x\in X_0|x_j = 0\}$ negative since $r(x) = r_l$ and $d(x,h_j) = r_u> r(x)$.
    Therefore, $\err(h_j) = \frac{1}{n} \cdot 3n\epsilon = 3\epsilon$.
    To output $f_\out \in \cH$, we must identify the true target function.
\end{itemize}

\textbf{Information gain from different choices of $f_t$} 
Let $h^* = 2\ind{\{\be_{i^*}\}}-1$ denote the target function.
Since $(\bZero, 0, -1)$ is realized by all hypotheses, we can only gain information about the target function when $x_t\in X_0$.
For any $x_t \in X_0$, if $d(x_t,f_t)\leq r_l$ or $d(x_t,f_t)> r_u$, we cannot learn anything about the target function. 
In particular, if $d(x_t,f_t)\leq r_l$, the learner will observe $x_t\sim \Unif(X_0)$, $y_t = +1$, $\hat y_t = +1$ in all $\{\cD_i|i\in [n]\}$.
If $d(x_t,f_t)> r_u$, the learner will observe $x_t\sim \Unif(X_0)$, $y_t = +1$, $\hat y_t = -1$ in all $\{\cD_i|i\in [n]\}$.
Therefore, we cannot obtain any information about the target function.

Now for any $x_t \in X_0$, with the $i_t$-th coordinate being $0$, we enumerate the distance between $x$ and $x'$ for all $x'\in \cX$.
\begin{itemize}
    \item For all $x'\in X_0$, $d(x,x')\leq \norm{x}+\norm{x'}\leq 2\alpha < r_l$;
    \item For all $j\neq i_t$, $d(x,\be_j) = \sqrt{1+\alpha^2 - 2 x_j}\leq r_l$;
    \item $d(x,\be_{i_t}) = r_u$;
    \item $d(x,\bZero) =\alpha <r_l$.
\end{itemize}
Only $f_t = 2\ind{\{\be_{i_t}\}}-1$ satisfies that $r_l<d(x_t,f_t)\leq r_u$ and thus, we can only obtain information when $f_t = 2\ind{\{\be_{i_t}\}}-1$.
And the only information we learn is whether $i_t = i^*$ because if $i_t\neq i^*$, no matter which $i^*$ is, our observation is identical. If $i_t\neq i^*$, we can eliminate $2\ind{\{\be_{i_t}\}}-1$.

\textbf{Sample size analysis}
For any algorithm $\cA$, his predictions are identical in all environments $\{\cD_i|i\in [n]\}$ before a round $t$ in which $f_t = 2\ind{\{\be_{i_t}\}}-1$. Then either he learns $i_t$ in $\cD_{i_t}$ or he eliminates $2\ind{\{\be_{i_t}\}}-1$ and continues to perform the same in the other environments $\{\cD_i|i\neq i_t\}$.
Suppose that we run $\cA$ in all stochastic environments $\{\cD_i|i\in [n]\}$ simultaneously. When we identify $i_t$ in environment $\cD_{i_t}$, we terminate $\cA$ in $\cD_{i_t}$. Consider a good algorithm $\cA$ which can identify $i$ in $\cD_i$ with probability $\frac{7}{8}$ after $T$ rounds of interaction for each $i\in [n]$, that is,
\begin{align}
    \Pr_{\cD_i,\cA}(i_\out \neq i)\leq \frac{1}{8},\forall i\in[n]\,.\label{eq:good4all}
\end{align}
Therefore, we have
\begin{align}
    \sum_{i\in [n]}\Pr_{\cD_i,\cA}(i_\out \neq i)\leq \frac{n}{8}\,.\label{eq:good4sum}
\end{align}


Let $n_T$ denote the number of environments that have been terminated by the end of round $T$.
Let $B_t$ denote the event of $x_t$ being in $X_0$ and $C_t$ denote the event of $f_t=2\ind{\{\be_{i_t}\}}-1$.
Then we have $\Pr(B_t) = 3n\epsilon$ and $\Pr(C_t|B_t) = \frac{1}{n}$, and thus $\Pr(B_t \wedge C_t) = 3n\epsilon \cdot \frac{1}{n}$.
Since at each round, we can eliminate one environment only when $B_t\wedge C_t$ is true, then we have
\begin{align*}
    \EE{n_T} \leq \EE{\sum_{t=1}^T\1{B_t\wedge C_t}} = T\cdot 3n\epsilon \cdot \frac{1}{n} = 3\epsilon T\,.
\end{align*}
Therefore, by setting $T = \frac{\floor{\frac{n}{2}}-1}{6\epsilon}$ and Markov's inequality, we have
\begin{align*}
    \Pr(n_T\geq \floor{\frac{n}{2}}-1)\leq \frac{3\epsilon T}{\floor{\frac{n}{2}}-1} = \frac{1}{2}\,.
\end{align*}
When there are $\ceil{\frac{n}{2}}+1$ environments remaining, the algorithm has to pick one $i_\out$, which fails in at least $\ceil{\frac{n}{2}}$ of the environments.
Then we have
\begin{align*}
    \sum_{i\in [n]}\Pr_{\cD_i,\cA}(i_\out \neq i)\geq \ceil{\frac{n}{2}}\Pr(n_T\leq \floor{\frac{n}{2}}-1)\geq \frac{n}{4}\,,
\end{align*}
which conflicts with Eq~\eqref{eq:good4sum}.
Therefore, for any algorithm $\cA$, to achieve Eq~\eqref{eq:good4all}, it requires $T\geq \frac{\floor{\frac{n}{2}}-1}{6\epsilon}$.
\end{proof}
\section{Proof of Theorem~\ref{thm:x_end_pac}}\label{app:x_end_pac}
Given Lemma~\ref{lmm:alg-exp}, we can upper bound the expected strategic loss, then we can boost the confidence of the algorithm through the scheme in Section~\ref{subsec:boost}.
Theorem~\ref{thm:x_end_pac} follows by combining Lemma~\ref{lmm:alg-exp} and Lemma~\ref{lmm:boost}.
Now we only need to prove Lemma~\ref{lmm:alg-exp}.
\begin{proof}[Proof of Lemma~\ref{lmm:alg-exp}]
    For any set of hypotheses $H$, for every $z = (x,r,y)$, we define 
    \begin{align*}
        \kappa_p(H, z) := \begin{cases}
            \abs{\{h\in H|h(\Delta(x,h,r)) = -\}} & \text{if } y = +\,,\\
            0 & \text{otherwise.}
        \end{cases}
    \end{align*}
    So $\kappa_p(H, z)$ is the number of hypotheses mislabeling $z$ for positive $z$'s and $0$ for negative $z$'s.
    Similarly, we define $\kappa_n$ as follows,
    \begin{align*}
        \kappa_n(H, z) := \begin{cases}
            \abs{\{h\in H|h(\Delta(x,h,r)) = +\}} & \text{if } y = -\,,\\
            0 & \text{otherwise.}
        \end{cases}
    \end{align*}
    So $\kappa_n(H, z)$ is the number of hypotheses mislabeling $z$ for negative $z$'s and $0$ for positive $z$'s.

In the following, we divide the proof into two parts. First, recall that in Algorithm~\ref{alg:end-iid-ball}, the output is constructed by randomly sampling two hypotheses with replacement and taking the union of them. We represent the loss of such a random predictor using $\kappa_p(H, z)$ and $\kappa_n(H, z)$ defined above.
Then we show that whenever the algorithm makes a mistake, with some probability, we can reduce $\frac{\kappa_p(\VS_{t-1}, z_t)}{2}$ or $\frac{\kappa_n(\VS_{t-1}, z_t)}{2}$ hypotheses and utilize this to provide a guarantee on the loss of the final output.

\paragraph{Upper bounds on the strategic loss} 
For any hypothesis $h$, let $\fpr(h)$ and $\fnr(h)$ denote the false positive rate and false negative rate of $h$ respectively.
Let $p_+$ denote the probability of drawing a positive sample from $\cD$, i.e., $\Pr_{(x,r,y)\sim \cD}(y=+)$ and $p_-$ denote the probability of drawing a negative sample from $\cD$.
Let $\cD_+$ and $\cD_-$ denote the data distribution conditional on that the label is positive and that the label is negative respectively.
Given any set of hypotheses $H$, we define a random predictor $R2(H) = h_1 \vee h_2$ with $h_1, h_2$ randomly picked from $H$ with replacement.
    For a true positive $z$, $R2(H)$ will misclassify it with probability $\frac{\kappa_p(H,z)^2}{\abs{H}^2}$.
    Then we can find that the false negative rate of $R2(H)$ is
    \begin{align*}
        \fnr(R2(H)) = \EEs{z=(x,r,+)\sim \cD_+}{\Pr(R2(H)(x) = -)} = \EEs{z=(x,r,+)\sim \cD_+}{\frac{\kappa_p(H,z)^2}{\abs{H}^2}}\,.
    \end{align*}
    Similarly, for a true negative $z$, $R2(H)$ will misclassify it with probability $1- (1-\frac{\kappa_n(H,z)}{\abs{H}})^2 \leq \frac{2\kappa_n(H,z)}{\abs{H}}$.
    Then the false positive rate of $R2(H)$ is
    \begin{align*}
        \fpr(R2(H)) = \EEs{z=(x,r,-)\sim \cD_-}{\Pr(R2(H)(x) = +)} \leq \EEs{z=(x,r,-)\sim \cD_+}{\frac{2\kappa_n(H,z)}{\abs{H}}}\,.
    \end{align*}
    Hence the loss of $R2(H)$ is
    \begin{align}
        \err(R2(H)) &\leq p_+\EEs{z\sim \cD_+}{\frac{\kappa_p(H,z)^2}{\abs{H}^2}} + p_- \EEs{z\sim \cD_+}{\frac{2\kappa_n(H,z)}{\abs{H}}}\nonumber\\
        &= \EEs{z\sim \cD}{\frac{\kappa_p(H,z)^2}{\abs{H}^2} + 2\frac{\kappa_n(H,z)}{\abs{H}}}\,,\label{eq:loss-rep}
    \end{align}
    where the last equality holds since $\kappa_p(H,z) = 0$ for true negatives and $\kappa_n(H,z) = 0$ for true positives.

    \paragraph{Loss analysis} In each round, the data $z_t = (x_t,r_t,y_t)$ is sampled from $\cD$.
    When the label $y_t$ is positive, if the drawn $f_t$ satisfying that 1) $f_t(\Delta(x_t,f_t,r_t)) = -$ and 2) $d(x_t,f_t)\leq \textrm{median}(\{d(x_t,h)|h\in \VS_{t-1}, h(\Delta(x_t,h,r_t))=-\})$, then we are able to remove 
    $\frac{\kappa_p(\VS_{t-1},z_t)}{2}$ hypotheses from the version space.
    Let $E_{p,t}$ denote the event of $f_t$ satisfying the conditions 1) and 2).
    With probability $\frac{1}{\floor{\log_2(n_t)}}$, we sample $k_t=1$.
    Then we sample an $f_t\sim \Unif(\VS_{t-1})$.
    With probability $\frac{\kappa_p(\VS_{t-1},z_t)}{2n_t}$, the sampled $f_t$ satisfies the two conditions.
    So we have
    \begin{equation}
        \Pr(E_{p,t}|z_t,\VS_{t-1})\geq \frac{1}{\log_2(n_t)}\frac{\kappa_p(\VS_{t-1},z_t)}{2n_t}\,.\label{eq:eventp}
    \end{equation}
     

    The case of $y_t$ being negative is similar to the positive case.
    Let $E_{n,t}$ denote the event of $f_t$ satisfying that 1) $f_t(\Delta(x_t,f_t,r_t)) = +$ and 2) $d(x_t,f_t)\geq \textrm{median}(\{d(x_t,h)|h\in \VS_{t-1}, h(\Delta(x_t,h,r_t))=+\})$.
    If $\kappa_n(\VS_{t-1},z_t)\geq \frac{n_t}{2}$, then with probability $\frac{1}{\floor{\log_2(n_t)}}$, we sample $k_t = 1$.
    Then with probability greater than $\frac{1}{4}$ we will sample an $f_t$ satisfying that 1) $f_t(\Delta(x_t,f_t,r_t)) = +$ and 2) $d(x_t,f_t)\geq \textrm{median}(\{d(x_t,h)|h\in \VS_{t-1}, h(\Delta(x_t,h,r_t))=+\})$.
    If $\kappa_n(\VS_{t-1},z_t)< \frac{n_t}{2}$, then
    with probability $\frac{1}{\floor{\log_2(n_t)}}$, we sampled a $k_t$ satisfying 
    \[\frac{n_t}{4\kappa_n(\VS_{t-1},z_t)} <k_t \leq \frac{n_t}{2\kappa_n(\VS_{t-1},z_t)}\,.\]
    Then we randomly sample $k_t$ hypotheses and the expected number of sampled hypotheses which mislabel $z_t$ is $k_t\cdot \frac{\kappa_n(\VS_{t-1},z_t)}{n_t} \in (\frac{1}{4},\frac{1}{2}]$.
    Let $g_t$ (given the above fixed $k_t$) denote the number of sampled hypotheses which mislabel $x_t$ and we have $\EE{g_t} \in (\frac{1}{4},\frac{1}{2}]$.
    When $g_t>0$, $f_t$ will misclassify $z_t$ by positive.
    We have
    \[\Pr(g_t = 0) = (1-\frac{\kappa_n(\VS_{t-1},z_t)}{n_t})^{k_t} < (1-\frac{\kappa_n(\VS_{t-1},z_t)}{n_t})^{\frac{n_t}{4\kappa_n(\VS_{t-1},z_t)}}\leq e^{-1/4}\leq 0.78\]
    and by Markov's inequality, we have
    \[\Pr(g_t \geq 3) \leq \frac{\EE{g_t}}{3} \leq \frac{1}{6}\leq 0.17\,.\]
    Thus $\Pr(g_t \in \{1,2\}) \geq 0.05$.
    Conditional on $g_t$ is either $1$ or $2$, with probability $\geq \frac{1}{4}$, all of these $g_t$ hypotheses $h'$ satisfies $d(x_t,h')\geq \textrm{median}(\{d(x_t,h)|h\in \VS_{t-1}, h(\Delta(x_t,h,r_t))=+\})$, which implies that $d(x_t,f_t)\geq \textrm{median}(\{d(x_t,h)|h\in \VS_{t-1}, h(\Delta(x_t,h,r_t))=+\})$.
    Therefore, we have
    \begin{equation}
        \Pr(E_{n,t}|z_t,,\VS_{t-1})\geq \frac{1}{80\log_2(n_t)}\,.\label{eq:eventn}
    \end{equation}
    
    Let $v_t$ denote the fraction of hypotheses we eliminated at round $t$, i.e., $v_t = 1 -\frac{n_{t+1}}{n_t}$ . Then we have
    \begin{equation}
        v_t\geq \1{E_{p,t}}\frac{\kappa_p(\VS_{t-1},z_t)}{2n_t} + \1{E_{n,t}}\frac{\kappa_n(\VS_{t-1},z_t)}{2n_t}\,.\label{eq:removefrac}
    \end{equation}
    Since $n_{t+1} = n_t(1-v_t)$, we have
    \[1 \leq n_{T+1} = n \prod_{t=1}^T (1-v_t)\,.\]
    By taking logarithm of both sides, we have
    \begin{align*}
        0 \leq \ln n_{T+1} = \ln n +\sum_{t=1}^T \ln(1-v_t)\leq \ln n -\sum_{t=1}^T v_t\,,
    \end{align*}
    where we use $\ln(1-x)\leq -x$ for $x\in [0,1)$ in the last inequality.
    By re-arranging terms, we have
    \[\sum_{t=1}^T v_t \leq \ln n\,.\]
    Combined with Eq~\eqref{eq:removefrac}, we have
    \begin{align*}
        \sum_{t=1}^T \1{E_{p,t}}\frac{\kappa_p(\VS_{t-1},z_t)}{2n_t} + \1{E_{n,t}}\frac{\kappa_n(\VS_{t-1},z_t)}{2n_t}\leq \ln n\,.
    \end{align*}
    By taking expectation w.r.t. the randomness of $f_{1:T}$ and dataset $S = z_{1:T}$ on both sides, we have
    \[\sum_{t=1}^T\EEs{f_{1:T},z_{1:T}}{\1{E_{p,t}}\frac{\kappa_p(\VS_{t-1},z_t)}{2n_t} + \1{E_{n,t}}\frac{\kappa_n(\VS_{t-1},z_t)}{2n_t}} \leq \ln n\,.\]
    Since the $t$-th term does not depend on $f_{t+1:T},z_{t+1:T}$ and $\VS_{t-1}$ is determined by $z_{1:t-1}$ and $f_{1:t-1}$, the $t$-th term becomes
    \begin{align}
        &\EEs{f_{1:t},z_{1:t}}{\1{E_{p,t}}\frac{\kappa_p(\VS_{t-1},z_t)}{2n_t} + \1{E_{n,t}}\frac{\kappa_n(\VS_{t-1},z_t)}{2n_t}} \nonumber\\
    = &\EEs{f_{1:t-1},z_{1:t}}{\EEs{f_t}{\1{E_{p,t}}\frac{\kappa_p(\VS_{t-1},z_t)}{2n_t} + \1{E_{n,t}}\frac{\kappa_n(\VS_{t-1},z_t)}{2n_t}| f_{1:t-1},z_{1:t}}}\nonumber\\
    =& \EEs{f_{1:t-1},z_{1:t}}{\EEs{f_t}{\1{E_{p,t}}|f_{1:t-1},z_{1:t}}\frac{\kappa_p(\VS_{t-1},z_t)}{2n_t} + \EEs{f_t}{\1{E_{n,t}}|f_{1:t-1},z_{1:t}}\frac{\kappa_n(\VS_{t-1},z_t)}{2n_t}}\label{eq:vs-ndep-f}\\
    \geq &\EEs{f_{1:t-1},z_{1:t}}{\frac{1}{\log_2(n_t)}\frac{\kappa_p^2(\VS_{t-1},z_t)}{4n_t^2} + \frac{1}{80\log_2(n_t)}\frac{\kappa_n(\VS_{t-1},z_t)}{2n_t}}\label{eq:prob-good}\,,
    \end{align}
    where Eq~\eqref{eq:vs-ndep-f} holds due to that $\VS_{t-1}$ is determined by $f_{1:t-1}, z_{1:t-1}$ and does not depend on $f_t$ and Eq~\eqref{eq:prob-good} holds since $\Pr_{f_t}(E_{p,t}|f_{1:t-1},z_{1:t}) = \Pr_{f_t}(E_{p,t}|\VS_{t-1},z_t) \geq \frac{1}{\log_2(n_t)}\frac{\kappa_p(\VS_{t-1},z_t)}{2n_t}$ by Eq~\eqref{eq:eventp} and $\Pr_{f_t}(E_{n,t}|f_{1:t-1},z_{1:t})=\Pr_{f_t}(E_{n,t}|\VS_{t-1},z_t) \geq \frac{1}{80\log_2(n_t)}$ by Eq~\eqref{eq:eventn}.
    Thus, we have
    \[\sum_{t=1}^T \EEs{f_{1:t-1},z_{1:t}}{\frac{1}{\log_2(n_t)}\frac{\kappa_p^2(\VS_{t-1},z_t)}{4n_t^2} + \frac{1}{80\log_2(n_t)}\frac{\kappa_n(\VS_{t-1},z_t)}{2n_t}} \leq \ln n\,.\]
    Since $z_t \sim \cD$ and $z_t$ is independent of $z_{1:t-1}$ and $f_{1:t-1}$, thus, 
    we have the $t$-th term on the LHS being
    \begin{align*}
        &\EEs{f_{1:t-1},z_{1:t}}{\frac{1}{\log_2(n_t)}\frac{\kappa_p^2(\VS_{t-1},z_t)}{4n_t^2} + \frac{1}{80\log_2(n_t)}\frac{\kappa_n(\VS_{t-1},z_t)}{2n_t} }\\
    =& \EEs{f_{1:t-1},z_{1:t-1}}{\EEs{z_t \sim \cD}{\frac{1}{\log_2(n_t)}\frac{\kappa_p^2(\VS_{t-1},z_t)}{4n_t^2} + \frac{1}{80\log_2(n_t)}\frac{\kappa_n(\VS_{t-1},z_t)}{2n_t}}}\\
    \geq& \frac{1}{320\log_2(n)}\EEs{f_{1:t-1},z_{1:t-1}}{\EEs{z\sim \cD}{\frac{\kappa_p^2(\VS_{t-1},z)}{n_t^2} + \frac{2\kappa_n(\VS_{t-1},z)}{n_t}}}\\
    \geq& \frac{1}{320\log_2(n)}\EEs{f_{1:t-1},z_{1:t-1}}{\err(R2(\VS_{t-1}))}\,,
    \end{align*}
    where the last inequality adopts Eq~\eqref{eq:loss-rep}.
    By summing them up and re-arranging terms,
    we have
    \[\EEs{f_{1:T},z_{1:T}}{\frac{1}{T}\sum_{t=1}^T \err(R2(\VS_{t-1}))}=\frac{1}{T}\sum_{t=1}^T \EEs{f_{1:t-1},z_{1:t-1}}{\err(R2(\VS_{t-1}))} \leq \frac{320\log_2(n)\ln(n)}{T}\,.\]
    For the output of Algorithm~\ref{alg:end-iid-ball}, which randomly picks $\tau$ from $[T]$, randomly samples $h_1, h_2$ from $\VS_{\tau-1}$ with replacement and outputs $h_1\vee h_2$, the expected loss is
    \begin{align*}
        \EE{\err(\cA(S))}= &\EEs{S,f_{1:T}}{\frac{1}{T}\sum_{t=1}^T \EEs{h_1,h_2\sim \Unif(\VS_{t-1})}{\err(h_1\vee h_2)}} \\
        =& \EEs{S,f_{1:T}}{\frac{1}{T}\sum_{t=1}^T \err(R2(\VS_{t-1}))}\\
        \leq& \frac{320\log_2(n)\ln(n)}{T} \leq \epsilon\,,
    \end{align*}
    when $T\geq \frac{320\log_2(n)\ln(n)}{\epsilon}$.
\end{proof}  
\paragraph{Post proof discussion of Lemma~\ref{lmm:alg-exp}}
\begin{itemize}
    \item Upon first inspection, readers might perceive a resemblance between the proof of the loss analysis section and the standard proof of converting regret bound to error bound.This standard proof converts a regret guarantee on $f_{1:T}$ to an error guarantee of $\frac{1}{T}\sum_{t=1}^T f_t$.
    However, in this proof, the predictor employed in each round is $f_t$, while the output is an average over $R2(\VS_{t-1})$ for all $t\in [T]$. Our algorithm does not provide a regret guarantee on $f_{1:T}$.
    \item Please note that our analysis exhibits asymmetry regarding losses on true positives and true negatives. 
    Specifically, the probability of identifying and reducing half of the misclassifying hypotheses on true positives, denoted as $\Pr(E_{p,t}|z_t,\VS_{t-1})$ (Eq~\eqref{eq:eventp}), is lower than the corresponding probability for true negatives, $\Pr(E_{n,t}|z_t,\VS_{t-1})$ (Eq~\eqref{eq:eventn}). This discrepancy arises due to the different levels of difficulty in detecting misclassifying hypotheses.
    For example, if there is exactly one hypothesis $h$ misclassifying a true positive $z_t=(x_t,r_t,y_t)$, it is very hard to detect this $h$.
    We must select an $f_t$ satisfying that $d(x_t,f_t)> d(x_t,h')$ for all $h'\in \cH\setminus \{h\}$ (hence $f_t$ will make a mistake), and that $d(x_t,f_t)\leq d(x_t,h)$ (so that we will know $h$ misclassifies $z_t$). Algorithm~\ref{alg:end-iid-ball} controls the distance $d(x_t,f_t)$ through $k_t$, which is the number of hypotheses in the union. In this case, we can only detect $h$ when $k_t =1$ and $f_t = h$, which occurs with probability $\frac{1}{n_t \log(n_t)}$.
    
    However, if there is exactly one hypothesis $h$ misclassifying a true negative $z_t=(x_t,r_t,y_t)$, we have that $d(x_t,h) = \min_{h'\in \cH} d(x_t,h')$.
    Then by setting $f_t = \vee_{h\in \cH} h$, which will makes a mistake and tells us $h$ is a misclassifying hypothesis. Our algorithm will pick such an $f_t$ with probability $\frac{1}{\log(n_t)}$.
\end{itemize}
\section{Proof of Theorem~\ref{thm:delta-csv}}\label{app:delta-csv}
\begin{proof} 
We will prove Theorem~\ref{thm:delta-csv} by constructing an instance of $\cQ$ and $\cH$ and showing that for any conservative learning algorithm, there exists a realizable data distribution s.t. achieving $\epsilon$ loss requires at least $\tilde \Omega(\frac{\abs{\cH}}{\epsilon})$ samples.
\paragraph{Construction of $\cQ$, $\cH$ and a set of realizable distributions} 
    \begin{itemize}
        \item Let the input metric space $(\cX,d)$ be constructed in the following way. Consider the feature space $\cX = \{\be_1,\ldots, \be_n\}\cup X_0$,  where $X_0 = \{\frac{\sigma(0,1,\ldots,n-1)}{z}|\sigma\in \cS_n\}$ with $z = \frac{\sqrt{1^2+\ldots+(n-1)^2}}{\alpha}$ for some small $\alpha=0.1$. Here $\cS_n$ is the set of all permutations over $n$ elements. 
        So $X_0$ is the set of points whose coordinates are a permutation of $\{0,1/z,\ldots,(n-1)/z\}$ and all points in $X_0$  have the  $\ell_2$ norm equal to $\alpha$.
        We define the metric $d$ by restricting $\ell_2$ distance to $\cX$, i.e., $d(x_1,x_2) = \norm{x_1-x_2}_2$ for all $x_1,x_2\in \cX$. 
        Then we have that for any $x\in X_0$ and $i\in [n]$, the distance between $x$ and $\be_i$ is 
        $$d(x,\be_i) = \norm{x-\be_i}_2 = \sqrt{(x_i-1)^2 +\sum_{j\neq i} x_j^2} = \sqrt{1+\sum_{j=1}^n x_j^2 - 2x_i} = \sqrt{1+\alpha^2-2x_i}\,,$$
        which is greater than $\sqrt{1+\alpha^2-2\alpha}>0.8>2\alpha$.
        For any two points $x,x'\in X_0$, $d(x,x')\leq 2\alpha$ by triangle inequality.
        
        \item Let the hypothesis class be a set of singletons over $\{\be_i|i\in [n]\}$, i.e., $\cH = \{2\ind{\{\be_i\}}-1|i\in [n]\}$.
        \item We now define a collection of distributions $\{\cD_i|i\in[n]\}$ in which $\cD_i$ is realized by $2\ind{\{\be_i\}}-1$. 
        For any $i\in [n]$, we define $\cD_i$ in the following way.
        Let the marginal distribution $\cD_\cX$ over $\cX$ be uniform over $X_0$.
        For any $x$, the label $y$ is $+$ with probability $1-6\epsilon$ and $-$ with probability $6\epsilon$, i.e., $\cD(y|x)= \Rad(1-6\epsilon)$.
        Note that the marginal distribution $\cD_{\cX\times\cY} = \Unif(X_0)\times \Rad(1-6\epsilon)$ is identical for any distribution in $\{\cD_i|i\in[n]\}$ and does not depend on $i$.
        
        If the label is positive $y=+$, then let the radius $r = 2$. If the label is negative $y=-$, then let $r= \sqrt{1+\alpha^2-2(x_{i} +\frac{1}{z})}$, which guarantees that $x$ can be manipulated to $\be_j$ iff $d(x,\be_j)< d(x,\be_{i})$ for all $j\in [n]$. 
        Since $x_i\leq \alpha$ and $\frac{1}{z}\leq \alpha$, we have $\sqrt{1+\alpha^2-2(x_{i} +\frac{1}{z})} \geq \sqrt{1-4\alpha}>2\alpha$. Therefore, for both positive and negative examples, we have radius $r$ strictly greater than $2\alpha$ in both cases.
    \end{itemize}

    \paragraph{Randomization and improperness of the output $f_\out$ do not help}
    Note that algorithms are allowed to output a randomized $f_\out$ and to output $f_\out\notin \cH$.
    We will show that randomization and improperness of $f_\out$ don't make the problem easier.
    That is, supposing that the data distribution is $\cD_{i^*}$ for some $i^*\in [n]$, finding a (possibly randomized and improper) $f_\out$ is not easier than identifying $i^*$.
    Since our feature space $\cX$ is finite, we can enumerate all hypotheses not equal to $2\ind{\{\be_{i^*}\}}-1$ and calculate their strategic population loss as follows.
    \begin{itemize}
        \item $2\ind{\emptyset}-1$ predicts all negative and thus
    $\err(2\ind{\emptyset}-1) = 1-6\epsilon$;
    \item For any $a\subset \cX$ s.t. $a\cap X_0\neq \emptyset$, 
    $2\ind{a}-1$ will predict any point drawn from $\cD_{i^*}$ as positive (since all points have radius greater than $2\alpha$ and the distance between any two points in $X_0$ is smaller than $2\alpha$) and thus
    $\err(2\ind{a}-1) = 6\epsilon$;
    \item For any $a\subset \{\be_1,\ldots,\be_n\}$ satisfying that $\exists i\neq i^*$, $\be_i\in a$, we have  $\err(2\ind{a}-1)\geq 3\epsilon$. This is due to that when $y=-$, $x$ is chosen from $\Unif(X_0)$ and the probability of $d(x,\be_i)<d(x,\be_{i^*})$ is $\frac{1}{2}$.
    When $d(x,\be_i)<d(x,\be_{i^*})$, $2\ind{a}-1$ will predict $x$ as positive.
    \end{itemize}
    Under distribution $\cD_{i^*}$, if we are able to find a (possibly randomized) $f_\out$ with strategic loss of $\err(f_\out)\leq \epsilon$, then we have $\err(f_\out) = \EEs{h\sim f_\out}{\err(h)}\geq \Pr_{h\sim f_\out}(h\neq 2\ind{\{\be_{i^*}\}}-1) \cdot 3\epsilon$.
    Thus, $\Pr_{h\sim f_\out}(h= 2\ind{\{\be_{i^*}\}}-1)\geq \frac{2}{3}$.
    Hence, if we are able to find a (possibly randomized) $f_\out$ with $\epsilon$ error, then we are able to identify $i^*$ by checking which realization of $f_\out$ has probability greater than $\frac{2}{3}$.
    In the following, we will focus on the sample complexity to identify $i^*$.
    Let $i_\out$ denote the algorithm's answer to question ``what is $i^*$?''.
    \paragraph{Conservative algorithms}
    When running a conservative algorithm, the rule of choosing $f_t$ at round $t$ and choosing the final output $f_\out$ does not depend on the correct rounds, i.e. $\{\tau \in [T]|\hat y_\tau = y_\tau\}$.
    Let's define
    \begin{align}
        \Delta_t' = \begin{cases}
            \Delta_t & \text{ if } \hat y_t \neq  y_t\\
            \perp & \text{ if } \hat y_t = y_t\,,
        \end{cases}\label{eq:deltaprime}
    \end{align}
    where $\perp$ is just a symbol representing ``no information''.
    Then for any conservative algorithm, the selected predictor $f_t$ is determined by $(f_\tau,\hat y_\tau, y_\tau, \Delta'_\tau)$ for $\tau<t$ and the final output $f_\out$ is determined by $(f_t,\hat y_t, y_t, \Delta'_t)_{t=1}^T$.
    From now on, we consider $\Delta_t'$ as the feedback in the learning process of a conservative algorithm since it make no difference from running the same algorithm with feedback $\Delta_t$.

    \paragraph{Smooth the data distribution} 
    For technical reasons (appearing later in the analysis), we don't want to analyze distribution $\{\cD_i|i\in [n]\}$ directly as the probability of $\Delta_t = \be_{i}$ is $0$ when $f_t(\be_{i}) = +1$ under distribution $\cD_i$.
    Instead, we consider the mixture of $\cD_i$ and another distribution $\cD_i''$, which is identical to $\cD_i$ except that $r(x) = d(x,\be_i)$ when $y=-$.
    More specifically, let $\cD_i' = (1-p) \cD_i + p  \cD_i''$ with some extremely small $p$, where 
    $\cD_i''$'s marginal distribution over $\cX\times \cY$ is still $\Unif(X_0)\times \Rad(1-6\epsilon)$; the radius is $r = 2$
    when $y=+$, ; and the radius is $r= d(x,\be_i)$ when $y=-$.
    For any data distribution $\cD$, let $\bP_\cD$ be the dynamics of $(f_1, y_1,\hat y_1,\Delta'_1,\ldots, f_T, y_T,\hat y_T,\Delta'_T)$ under $\cD$.
    According to Lemma~\ref{lmm:smooth}, by setting $p=\frac{\epsilon}{16n^2}$, when $T\leq \frac{n}{\epsilon}$, with high probability we never sample from $\cD_i''$ and have that for any $i,j\in [n]$
    \begin{equation}
        \abs{\bP_{\cD_{i}}(i_\out = j)-\bP_{\cD_{i}'}(i_\out = j)}\leq \frac{1}{8}\,.\label{eq:smooth-res-csv}
    \end{equation}

    From now on, we only consider distribution $\cD_i'$ instead of $\cD_i$. The readers might have the question that why not using $\cD_i'$ for construction directly. This is because $\cD_i'$ does not satisfy realizability and no hypothesis has zero loss under $\cD_i'$.

\paragraph{Information gain from different choices of $f_t$} In each round of interaction, the learner picks a predictor $f_t$, which can be out of $\cH$.
    Here we enumerate all choices of $f_t$.
    \begin{itemize}
        \item $f_t(\cdot) = 2\ind{\emptyset}-1$ predicts all points in $\cX$ by negative. 
        No matter what $i^*$ is, we will observe $(\Delta_t = x_t, y_t)\sim \Unif(X_0)\times \Rad(1-6\epsilon)$ and $\hat y_t = -$. 
        They are identically distributed for all $i^*\in [n]$, and thus, $\Delta_t'$ is also identically distributed. We cannot tell any information of $i^*$ from this round.
        \item $f_t=2\ind{a_t}-1$ for some $a_t\subset \cX$ s.t. $a\cap X_0\neq \emptyset$. 
        Then $\Delta_t = \Delta(x_t,f_t,r_t) = \Delta(x_t,f_t,2\alpha)$ since $r_t> 2\alpha$ and $d(x_t,f_t)\leq 2\alpha$, $\hat y_t = +$, $y_t\sim \Rad(1-6\epsilon)$. None of these depends on $i^*$ and again, the distribution of $(\hat y_t, y_t, \Delta'_t)$ is identical for all $i^*$ and we cannot tell any information of $i^*$ from this round.
        \item $f_t = 2\ind{a_t}-1$ for some non-empty $a_t\subset \{\be_1,\ldots,\be_n\}$. For rounds with $y_t = +$,  we have $\hat y_t = +$ and $\Delta_t = \Delta(x_t,f_t,2)$, which still not depend on $i^*$.
        Thus we cannot learn any information about $i^*$.
        But we can learn when $y_t=-$.
        For rounds with $y_t=-$, if $\Delta_t \in a_t$, then we could observe  $\hat y_t = +$
        and $\Delta_t' = \Delta_t$, which at least tells that $2\ind{\{\Delta_t\}}-1$ is not the target function (with high probability); if $\Delta_t \notin a_t$, then $\hat y_t = -$ and we observe $\Delta_t' = \perp$.
    \end{itemize}
    Therefore, we only need to focus on the rounds with $f_t = 2\ind{a_t}-1$ for some non-empty $a_t\subset \{\be_1,\ldots,\be_n\}$ and $y_t =-$.
    It is worth noting that drawing an example $x$ from $X_0$ uniformly, it is equivalent to uniformly drawing a permutation of $\cH$ such that the distances between $x$ and $h$ over all $h\in \cH$ are permuted according to it. Then $\Delta_t = \be_j$ iff $\be_j\in a_t$, $d(x, \be_j)\leq d(x,\be_{i^*})$ and $d(x,\be_j)\leq d(x,\be_l)$ for all $\be_l\in a_t$.  
    Let $k_t = \abs{a_t}$ denote the cardinality of $a_t$.
    In such rounds, under distribution $\cD_{i^*}'$, the distribution of $\Delta_t'$ are described as follows.
    \begin{enumerate}
        \item The case of $\be_{i^*} \in a_t$: For all $j\in a_t\setminus\{i^*\}$, with probability $\frac{1}{k_t}$, $d(x_t,\be_j) = \min_{\be_l\in a_t} d(x_t,\be_l)$ and thus, $\Delta_t' =\Delta_t = \be_j$ and $\hat y_t = +$ (mistake round). With probability $\frac{1}{k_t}$, we have $d(x_t,\be_{i^*}) = \min_{\be_l\in a_t} d(x_t,\be_l)$. If the example is drawn from $\cD_{i^*}$, we have $\Delta_t = x_t$ and $y_t = -$ (correct round), thus $\Delta_t' = \perp$. If the example is drawn from $\cD_{i^*}''$, we have we have $\Delta_t' = \Delta_t = \be_{i^*}$ and $y_t = +$ (mistake round). Therefore, according to the definition of $\Delta_t'$ (Eq~\eqref{eq:deltaprime}), we have
        \begin{align*}
            \Delta'_t = \begin{cases}
                \be_j & \text{w.p. } \frac{1}{k_t} \text{ for } \be_j \in a_t, j\neq i^*\\
                \be_{i^*} & \text{w.p. } \frac{1}{k_t}p \\
                \perp & \text{w.p. } \frac{1}{k_t}(1-p)\,.
            \end{cases}
        \end{align*}
        We denote this distribution by $P_\in(a_t,i^*)$.
        \item The case of $\be_{i^*} \notin a_t$: For all $j\in a_t$, with probability $\frac{1}{k_t+1}$,  then $d(x_t,\be_j) = \min_{\be_l\in a_t\cup \{\be_{i^*}\}} d(x_t,\be_l)$ and 
        thus, $\Delta_t = \be_j$ and $\hat y_t = +$ (mistake round). With probability $\frac{1}{k_t+1}$, we have $d(x,\be_{i^*}) < \min_{\be_l\in a_t} d(x_t,\be_l)$ and thus, $\Delta_t = x_t$, $\hat y_t = -$ (correct round), and $\Delta_t' =\perp$. Therefore, the distribution of $\Delta_t'$ is
        \begin{align*}
            \Delta'_t = \begin{cases}
                \be_j & \text{w.p. } \frac{1}{k_t+1} \text{ for } \be_j \in a_t\\
                \perp & \text{w.p. } \frac{1}{k_t+1}\,.
            \end{cases}
        \end{align*}
        We denote this distribution by $P_{\notin}(a_t)$.
    \end{enumerate}

    To measure the information obtained from $\Delta_t'$, we will utilize the KL divergence of the distribution of $\Delta_t'$ under the data distribution $\cD_{i^*}$ from that under a benchmark distribution.
    Let $\bar \cD = \frac{1}{n}\sum_{i\in n}\cD_i'$ denote the average distribution. 
    The process of sampling from $\bar \cD$ is equivalent to sampling $i^*$ uniformly at random from $[n]$ first and drawing a sample from $\cD_{i^*}$.
    Then under $\bar \cD$, for any $\be_j \in a_t$, we have
    \begin{align*}
        \Pr(\Delta'_t = \be_j) =& \Pr(i^*=j)\Pr(\Delta'_t = \be_j|i^*=j) +\Pr(i^*\in a_t\setminus\{j\})\Pr(\Delta'_t = \be_j|i^*\in a_t\setminus\{j\})\\
        &+ \Pr(i^*\notin a_t)\Pr(\Delta'_t = \be_j|i^*\notin a_t)\\=& \frac{1}{n} \cdot \frac{p}{k_t} +\frac{k_t-1}{n}\cdot \frac{1}{k_t} + \frac{n-k_t}{n}\cdot \frac{1}{k_t+1} = \frac{nk_t-1 +p(k_t+1)}{nk_t(k_t+1)}\,,
    \end{align*}
    and
    \begin{align*}
        \Pr(\Delta'_t = \perp) &= \Pr(i^*\in a_t)\Pr(\Delta'_t = \perp|i^*\in a_t)+ \Pr(i^*\notin a_t)\Pr(\Delta'_t = \perp|i^*\notin a_t)\\&=\frac{k_t}{n}\cdot \frac{1-p}{k_t} + \frac{n-k_t}{n}\cdot \frac{1}{k_t+1} = \frac{n+1 - p(k_t+1)}{n(k_t+1)}\,.
    \end{align*}
    Thus, the distribution of $\Delta'_t$ under $\bar\cD$ is
    \begin{align*}
            \Delta_t' = \begin{cases}
                \be_j & \text{w.p. } \frac{nk_t-1+p(k_t+1)}{nk_t(k_t+1)}\text{ for } \be_j \in a_t\\
                \perp & \text{w.p. } \frac{n+1-p(k_t+1)}{n(k_t+1)}\,.
            \end{cases}
    \end{align*}
    We denote this distribution by $\bar P(a_t)$.
    Next we will compute the KL divergences of $P_{\in}(a_t,i^*)$ and $P_{\notin}(a_t)$ from $\bar P(a_t)$. We will use the inequality $\log(1+x)\leq x$ for $x\geq 0$ in the following calculation. For any $i^*$ s.t. $\be_{i^*}\in a_t$, we have 
    \begin{align}
        &\KL{\bar P(a_t)}{P_{\in}(a_t,i^*)} \nonumber\\
        =& (k_t-1) \frac{nk_t-1+p(k_t+1)}{nk_t(k_t+1)}\log(\frac{nk_t-1+p(k_t+1)}{nk_t(k_t+1)}k_t) \nonumber\\
        &+\frac{nk_t-1+p(k_t+1)}{nk_t(k_t+1)}\log(\frac{nk_t-1+p(k_t+1)}{nk_t(k_t+1)}\cdot \frac{k_t}{p}) \nonumber\\
        &+ \frac{n+1-p(k_t+1)}{n(k_t+1)}\log(\frac{n+1-p(k_t+1)}{n(k_t+1)}\cdot \frac{k_t}{1-p}) \nonumber\\
        \leq & 0 + \frac{1}{k_t+1}\log(\frac{1}{p}) + \frac{2p}{k_t+1}
        =\frac{1}{k_t+1}\log(\frac{1}{p}) + \frac{2p}{k_t+1}\,,\label{eq:kl-in-csv}
        \end{align}
        and
        \begin{align}
            &\KL{\bar P(a_t)}{P_{\notin}(a_t)}\nonumber\\
            = & k_t \frac{nk_t-1+p(k_t+1)}{nk_t(k_t+1)}\log(\frac{nk_t-1+p(k_t+1)}{nk_t(k_t+1)}(k_t+1)) \nonumber\\
            &+ \frac{n+1-p(k_t+1)}{n(k_t+1)}\log (\frac{n+1-p(k_t+1)}{n(k_t+1)}(k_t+1))\nonumber \\
            \leq & 0+ \frac{n+1}{n^2(k_t+1)} =
            \frac{n+1}{n^2(k_t+1)}\label{eq:kl-out-csv}\,.
        \end{align}
\paragraph{Lower bound of the information} 
We utilize the information theoretical framework of proving lower bounds for linear bandits (Theorem 11 by \cite{rajaraman2023beyond}) here.
For notation simplicity, for all $i\in [n]$, let $\bP_i$ denote the dynamics of $(f_1,\Delta'_1, y_1,\hat y_1,\ldots, f_T,\Delta'_T, y_T,\hat y_T)$ under $\cD_i'$ and $\bar \bP$ denote the dynamics under $\bar \cD$.
Let $B_t$ denote the event of $\{f_t = 2\ind{a_t}-1 \text{ for some non-empty } a_t\subset \{\be_1,\ldots,\be_n\}\}$. 
As discussed before, for any $a_t$, conditional on $\neg B_t$ or $y_t =+1$,  $(\Delta_t', y_t,\hat y_t)$ are identical in all $\{\cD_i'|i\in [n]\}$, and therefore, also identical in $\bar \cD$.
We can only obtain information at rounds when $B_t \wedge (y_t =-1)$ occurs.
In such rounds, we know that $f_t$ is fully determined by history (possibly with external randomness , which does not depend on data distribution), $y_t =-1$ and $\hat y_t$ is fully determined by $\Delta_t'$ ($\hat y_t = +1$ iff. $\Delta_t' \in a_t$).

Therefore, conditional the history $H_{t-1} = (f_1,\Delta_1', y_1,\hat y_1,\ldots, f_{t-1},\Delta_{t-1}', y_{t-1},\hat y_{t-1})$ before time $t$, we have
\begin{align}
    &\KL{\bar \bP(f_{t},\Delta_{t}', y_{t},\hat y_{t}|H_{t-1})}{\bP_i( f_{t},\Delta_{t}', y_{t},\hat y_{t}|H_{t-1})} \nonumber\\
    =&\bar \bP(B_t  \wedge (y_t =-1)) \KL{\bar \bP(\Delta_t'|H_{t-1},B_t  \wedge (y_t =-1))}{\bP_i(\Delta_t'|H_{t-1},B_t  \wedge (y_t =-1))} \nonumber\\
    =& 6\epsilon \bar \bP(B_t) \KL{\bar \bP(\Delta_t'|H_{t-1},B_t  \wedge (y_t =-1))}{\bP_i(\Delta_t'|H_{t-1},B_t  \wedge (y_t =-1))}\label{eq:kl-delta-csv}\,,
\end{align}
where the last equality holds due to that $y_t\sim \Rad(1-6\epsilon)$ and does not depend on $B_t$.

For any algorithm that can successfully identify $i$ under the data distribution $\cD_i$ with probability $\frac{3}{4}$ for all $i\in [n]$, then $\bP_{\cD_i}(i_\out = i)\geq \frac{3}{4}$ and $\bP_{\cD_j}(i_\out = i)\leq \frac{1}{4}$ for all $j\neq i$.
Recall that $\cD_i$ and $\cD_i'$ are very close when the mixture parameter $p$ is small. Combining with Eq~\eqref{eq:smooth-res-csv}, we have 
\begin{align*}
    &\abs{\bP_{i}(i_\out = i) - \bP_{j}(i_\out = i)}\\
    \geq& \abs{\bP_{\cD_i}(i_\out = i) - \bP_{\cD_j}(i_\out = i)} - \abs{\bP_{\cD_i}(i_\out = i)- \bP_{i}(i_\out = i)}- \abs{\bP_{\cD_j}(i_\out = i)- \bP_{j}(i_\out = i)} \\
    \geq&\frac{1}{2}- \frac{1}{4} = \frac{1}{4}\,.
\end{align*}

Then we have the total variation distance between $\bP_{i}$ and $\bP_{j}$
\begin{align}
       \TV(\bP_{i},\bP_{j})\geq \abs{\bP_{i}(i_\out = i) - \bP_{j}(i_\out = i)}\geq \frac{1}{4}\,.\label{eq:tv-lb-csv}
\end{align}


Then we have
    \begin{align*}
    &\EEs{i\sim \Unif([n])}{\TV^2(\bP_{i},\bP_{(i+1) \text{ mod } n})} \leq 4 \EEs{i\sim \Unif([n])}{\TV^2(\bP_{i},\bar \bP )}\\
    \leq& 2\EEs{i}{\KL{\bar \bP}{\bP_{i}}} \tag{Pinsker's ineq}\\
    =& 2\EEs{i}{\sum_{t=1}^T \KL{\bar \bP(f_t,\Delta'_t, y_t,\hat y_t|H_{t-1})}{\bP_i(f_t,\Delta'_t, y_t,\hat y_t|H_{t-1})}}\tag{Chain rule}\\
    =& 12\epsilon \EEs{i}{\sum_{t=1}^T \bar \bP(B_t) \KL{\bar \bP(\Delta_t'|H_{t-1},B_t  \wedge (y_t =-1))}{\bP_i(\Delta_t'|H_{t-1},B_t  \wedge (y_t =-1))}}\tag{Apply Eq~\eqref{eq:kl-delta-csv}}\\
     =& \frac{12\epsilon}{n} \sum_{t=1}^T \bar \bP(B_t)\sum_{i=1}^n \KL{\bar \bP(\Delta_t'|H_{t-1},B_t  \wedge (y_t =-1))}{\bP_i(\Delta_t'|H_{t-1},B_t  \wedge (y_t =-1))}\\
     =& \frac{12\epsilon}{n}  \EEs{f_{1:T}\sim \bar \bP}{\sum_{t=1}^T\true{B_t}\left(\sum_{i:i\in a_t} \KL{\bar P(a_t)}{P_{\in}(a_t,i)}
     + \sum_{i:i\notin a_t}\KL{\bar P(a_t)}{P_{\notin}(a_t)}\right)}\\
     \leq & \frac{12\epsilon}{n} \sum_{t=1}^T \EEs{f_{1:T}\sim \bar \bP}{\sum_{i:i\in a_t} \left(\frac{1}{k_t+1}\log(\frac{1}{p}) + \frac{2p}{k_t+1}
     \right) + \sum_{i:i\notin a_t}\frac{n+1}{n^2(k_t+1)}}\tag{Apply Eq~\eqref{eq:kl-in-csv},\eqref{eq:kl-out-csv}}\\
     \leq & \frac{12\epsilon}{n} \sum_{t=1}^T (\log(\frac{1}{p})+ 2p + 1)\\
     \leq & \frac{12T\epsilon(\log(16n^2/\epsilon) +2)}{n}\,.
\end{align*}
Combining with Eq~\eqref{eq:tv-lb-csv}, we have that there exists a universal constant $c$ such that $T\geq \frac{cn}{\epsilon (\log(n/\epsilon) +1 )}$.
\end{proof}
\section{Proof of Theorem~\ref{thm:x-delta-never}}\label{app:x-delta-never}
\begin{proof} We will prove Theorem~\ref{thm:x-delta-never} by constructing an instance of $\cQ$ and $\cH$ and then reduce it to a linear stochastic bandit problem.
\paragraph{Construction of $\cQ, \cH$ and a set of realizable distributions}
\begin{itemize}
    \item Consider the input metric space in the shape of a star, where $\cX=\{0,1,\ldots,n\}$ and the distance function of $d(0,i) = 1$ and $d(i,j) =2$ for all $i\neq j\in [n]$.
    \item Let the hypothesis class be a set of singletons over $[n]$, i.e., $\cH =\{2\ind{\{i\}}-1|i\in [n]\}$.
    \item We define a collection of distributions $\{\cD_i|i\in [n]\}$ in which $\cD_i$ is realized by $2\ind{\{i\}}-1$. The data distribution $\cD_{i}$ put $1-3(n-1)\epsilon$ on $(0,1,+)$ and $3\epsilon$ on $(i,1,-)$ for all $i\neq i^*$. Hence, note that all distributions 
    in $\{\cD_i|i\in [n]\}$ share the same distribution support $\{(0,1,+)\}\cup\{(i,1,-)|i\in [n]\}$, but have different weights.
\end{itemize}

\paragraph{Randomization and improperness of the output $f_\out$ do not help.}
Note that algorithms are allowed to output a randomized $f_\out$ and to output $f_\out\notin \cH$.
We will show that randomization and improperness of $f_\out$ don't make the problem easier.
Supposing that the data distribution is $\cD_{i^*}$ for some $i^*\in [n]$, finding a (possibly randomized and improper) $f_\out$ is not easier than identifying $i^*$.
Since our feature space $\cX$ is finite, we can enumerate all hypotheses not equal to $2\ind{\{i^*\}}-1$ and calculate their strategic population loss as follows.
The hypothesis $2\ind{\emptyset}-1$ will predict all by negative and thus $\err(2\ind{\emptyset}-1) = 1-3(n-1)\epsilon$.
For any hypothesis predicting $0$ by positive, it will predict all points in the distribution support by positive and thus incurs strategic loss $3(n-1)\epsilon$.
For any hypothesis predicting $0$ by negative and some $i\neq i^*$ by positive, then it will misclassify $(i,1,-)$ and incur strategic loss $3\epsilon$.
Therefore, for any hypothesis $h\neq 2\ind{\{i^*\}}-1$, we have $\err_{\cD_{i^*}}(h)\geq 3\epsilon$. 

Similar to the proof of Theorem~\ref{thm:delta-csv}, under distribution $\cD_{i^*}$, if we are able to find a (possibly randomized) $f_\out$ with strategic loss $\err(f_\out)\leq \epsilon$.
Then $\Pr_{h\sim f_\out}(h=2\ind{\{i^*\}}-1) \geq \frac{2}{3}$.
We can identify $i^*$ by checking which realization of $f_\out$ has probability greater than $\frac{2}{3}$.
In the following, we will focus on the sample complexity to identify the target function $2\ind{\{i^*\}}-1$ or simply $i^*$.
Let $i_\out$ denote the algorithm's answer to question of ``what is $i^*$?''.

\textbf{Smooth the data distribution}
For technical reasons (appearing later in the analysis), we don't want to analyze distribution $\{\cD_i|i\in [n]\}$ directly as the probability of $(i,1,-)$ is $0$ under distribution $\cD_i$.
Instead, for each $i\in [n]$, let $\cD_i' = (1-p)\cD_i + p \cD_i''$ be the mixture of $\cD_i$ and $\cD_i''$ for some small $p$, where $\cD_i'' = (1-3(n-1)\epsilon)\ind{\{(0,1,+)\}} + 3(n-1)\epsilon \ind{\{(i,1,-)\}}$.
%
Specifically,
\begin{align*}
    \cD_i'(z) = \begin{cases}
        1-3(n-1)\epsilon & \text{for }z = (0,1,+)\\
        3(1-p)\epsilon & \text{for } z = (j,1,-),\forall j\neq i\\
        3(n-1)p\epsilon & \text{for } z = (i,1,-)
    \end{cases}
\end{align*}
For any data distribution $\cD$, let $\bP_\cD$ be the dynamics of $(f_1, y_1,\hat y_1,\ldots, f_T, y_T,\hat y_T)$ under $\cD$.
    According to Lemma~\ref{lmm:smooth}, by setting $p=\frac{\epsilon}{16n^2}$, when $T\leq \frac{n}{\epsilon}$, we have that for any $i,j\in [n]$
    \begin{equation}
        \abs{\bP_{\cD_{i}}(i_\out = j)-\bP_{\cD_{i}'}(i_\out = j)}\leq \frac{1}{8}\,.\label{eq:smooth-res-never}
    \end{equation}
    From now on, we only consider distribution $\cD_i'$ instead of $\cD_i$.
The readers might have the question that why not using $\cD_i'$ for construction directly. This is because no hypothesis has zero loss under $\cD_i'$, and thus $\cD_i'$ does not satisfy realizability requirement.

\paragraph{Information gain from different choices of $f_t$}
Note that in each round, the learner picks a $f_t$ and then only observes $\hat y_t$ and $y_t$. 
Here we enumerate choices of $f_t$ as follows.
\begin{enumerate}
    \item $f_t = 2\ind{\emptyset}-1$ predicts all points in $\cX$ by negative. No matter what $i^*$ is, we observe $\hat y_t = -$ and $y_t = 2\1{x_t = 0}-1$.
    Hence $(\hat y_t,y_t)$ are identically distributed for all $i^*\in [n]$, and thus, we cannot learn anything about $i^*$ from this round.
    \item $f_t$ predicts $0$ by positive. Then no matter what $i^*$ is, we have $\hat y_t = +$ and $y_t = \1{x_t = 0}$. Thus again, we cannot learn anything about $i^*$.
    \item $f_t = 2\ind{a_t}-1$ for some non-empty $a_t\subset [n]$. For rounds with $x_t =0$, we have $\hat y_t = y_t = +$ no matter what $i^*$ is and thus, we cannot learn anything about $i^*$. For rounds with $y_t=-$, i.e., $x_t\neq 0$, we will observe $\hat y_t = f_t(\Delta(x_t,f_t, 1)) = \1{x_t\in a_t}$. 
\end{enumerate}
Hence, we can only extract information with the third type of $f_t$ at rounds with $x_t\neq 0$.

\paragraph{Reduction to stochastic linear bandits}
In rounds with  $f_t = 2\ind{a_t}-1$ for some non-empty $a_t\subset [n]$ and $x_t\neq 0$, our problem is identical to a stochastic linear bandit problem.
Let us state our problem as Problem~\ref{prob:ours} and a linear bandit problem as Problem~\ref{prob:bandit}. Let $A = \{0,1\}^n\setminus \{\bZero\}$.
\begin{problem}\label{prob:ours}
The environment picks an $i^*\in [n]$. At each round $t$, the environment picks $x_t \in \{\be_i|i\in [n]\}$ with $P(i) = \frac{1-p}{n-1}$ for $i\neq i^*$ and $P(i^*) = p$ and the learner picks an $a_t\in A$ (where we use a $n$-bit string to represent $a_t$ and $a_{t,i} = 1$ means that $a_t$ predicts $i$ by positive). Then the learner observes $\hat y_t = \1{a_t^\top x_t >0}$ (where we use $0$ to represent nagative label).
\end{problem}

\begin{problem}\label{prob:bandit}
The environment picks a linear parameter $w^* \in \{w^i|i\in [n]\}$ with $w^i = \frac{1-p}{n-1}\bOne-(\frac{1-p}{n-1} -p)\be_i$.
The arm set is $A$. For each arm $a\in A$, the reward is i.i.d. from the following distribution:
\begin{align}
    r_w(a) = \begin{cases}
    -1, \text{ w.p. } w^\top a\,,\\
    0\,.
    \end{cases}\label{eq:dist-reward}
\end{align}
If the linear parameter $w^* = w^{i^*}$, the optimal arm is $\be_{i^*}$.

\end{problem}

\begin{claim}
For any $\delta>0$, for any algorithm $\cA$ that identify $i^*$ correctly with probability $1-\delta$ within $T$ rounds for any $i^*\in [n]$ in Problem~\ref{prob:ours}, we can construct another algorithm $\cA'$ can also identify the optimal arm in any environment with probability $1-\delta$ within $T$ rounds in Problem~\ref{prob:bandit}.
\end{claim}
This claim follows directly from the problem descriptions. Given any algorithm $\cA$ for Problem~\ref{prob:ours}, we can construct another algorithm $\cA'$ which simulates $\cA$.
At round $t$, if $\cA$ selects predictor $a_t$, then $\cA'$ picks arm the same as $a_t$.
Then $\cA'$ observes a reward $r_{w^{i^*}}(a_t)$, which is $-1$ w.p. $w^{i^*\top} a_t$ and feed $-r_{w^{i^*}}(a_t)$ to $\cA$. Since $\hat y_t$ in Problem~\ref{prob:ours} is $1$ w.p. $\sum_{i=1}^n a_{t,i}P(i) = w^{i^*\top}a_t$, it is distributed identically as $-r_{w^{i^*}}(a_t)$. Since $\cA$ will be able to identify $i^*$ w.p. $1-\delta$ in $T$ rounds, $\cA'$ just need to output $\be_{i^*}$ as the optimal arm.

Then any lower bound on $T$ for Problem \ref{prob:bandit} also lower bounds Problem \ref{prob:ours}. Hence, we adopt the information theoretical framework of proving lower bounds for linear bandits (Theorem 11 by \cite{rajaraman2023beyond}) to prove a lower bound for our problem. In fact, we also apply this framework to prove the lower bounds in other settings of this work, including Theorem~\ref{thm:delta-csv} and Theorem~\ref{thm:non-ball}.

\paragraph{Lower bound of the information}

For notation simplicity, for all $i\in [n]$, let $\bP_i$ denote the dynamics of $(f_1, y_1,\hat y_1,\ldots, f_T, y_T,\hat y_T)$ under $\cD_i'$ and and $\bar \bP$ denote the dynamics under $\bar \cD = \frac{1}{n}\cD_{i}'$.
Let $B_t$ denote the event of $\{f_t = 2\ind{a_t}-1 \text{ for some non-empty } a_t\subset [n]\}$.
As discussed before, for any $a_t$, conditional on $\neg B_t$ or $y_t =+1$,  $(x_t, y_t,\hat y_t)$ are identical in all $\{\cD_i'|i\in [n]\}$, and therefore, also identical in $\bar \cD$.
We can only obtain information at rounds when $B_t \wedge y_t =-1$ occurs.
In such rounds, $f_t$ is fully determined by history (possibly with external randomness , which does not depend on data distribution), $y_t =-1$ and $\hat y_t = -r_w(a_t)$ with $r_w(a_t)$ sampled from the distribution defined in Eq~\eqref{eq:dist-reward}.

For any algorithm that can successfully identify $i$ under the data distribution $\cD_i$ with probability $\frac{3}{4}$ for all $i\in [n]$, then $\bP_{\cD_i}(i_\out = i)\geq \frac{3}{4}$ and $\bP_{\cD_j}(i_\out = i)\leq \frac{1}{4}$ for all $j\neq i$.
Recall that $\cD_i$ and $\cD_i'$ are very close when the mixture parameter $p$ is small. Combining with Eq~\eqref{eq:smooth-res-never}, we have 
\begin{align}
    &\abs{\bP_{i}(i_\out = i) - \bP_{j}(i_\out = i)}\nonumber\\
    \geq& \abs{\bP_{\cD_i}(i_\out = i) - \bP_{\cD_j}(i_\out = i)} - \abs{\bP_{\cD_i}(i_\out = i)- \bP_{i}(i_\out = i)}- \abs{\bP_{\cD_j}(i_\out = i)- \bP_{j}(i_\out = i)} \nonumber\\
    \geq&\frac{1}{2}- \frac{1}{4} = \frac{1}{4}\,.\label{eq:tv-lb-never}
\end{align}


Let $\bar w = \frac{1}{n}\bOne$.
Let $\kl{q}{q'}$ denote the KL divergence from $\Ber(q)$ to $\Ber(q')$. Let $H_{t-1} = (f_1,y_1,\hat y_1,\ldots,f_{t-1},y_{t-1},\hat y_{t-1})$ denote the history up to time $t-1$.
Then we have
\begin{align}
    &\EEs{i\sim \Unif([n])}{\TV^2(\bP_i,\bP_{{i+1} \text{ mod } n})} \leq 4 \EEs{i\sim \Unif([n])}{\TV^2(\bP_i,\bar \bP)}\nonumber\\
    \leq& 2\EEs{i}{\KL{\bar \bP}{\bP_i}} \tag{Pinsker's ineq}\nonumber\\
    =& 2\EEs{i}{\sum_{t=1}^T \KL{\bar \bP(f_t, y_t,\hat y_t|H_{t-1})}{\bP_i(f_t, y_t,\hat y_t|H_{t-1})}}\tag{Chain rule}\nonumber\\
    =& 2\EEs{i}{\sum_{t=1}^T \bar \bP(B_t\wedge y_t=-1)\EEs{a_{1:T}\sim \bar \bP}{ \KL{\Ber(\inner{\bar w}{a_t})}{\Ber(\inner{w^i}{a_t})}}}\nonumber\\
    =& 6(n-1)\epsilon\EEs{i}{\sum_{t=1}^T \bar \bP(B_t)\EEs{a_{1:T}\sim \bar \bP}{ \KL{\Ber(\inner{\bar w}{a_t})}{\Ber(\inner{w^i}{a_t})}}}\nonumber\\
     =& \frac{6(n-1)\epsilon}{n} \sum_{t=1}^T \EEs{a_{1:T}\sim \bar \bP}{\sum_{i=1}^n \KL{\Ber(\inner{\bar w}{a_t})}{\Ber(\inner{w^i}{a_t})}}\nonumber\\
     =& \frac{6(n-1)\epsilon}{n} \sum_{t=1}^T \EEs{a_{1:T}\sim \bar \bP}{\sum_{i:i\in a_t}\kl{\frac{k_t}{n}}{ \frac{(k_t-1)(1-p)}{n-1}+ p} 
     + \sum_{i:i\notin a_t}\kl{\frac{k_t}{n}}{ \frac{k_t(1-p)}{n-1}}}\nonumber\\
     =& \frac{6(n-1)\epsilon}{n} \sum_{t=1}^T \EEs{a_{1:T}\sim \bar \bP}{k_t \kl{\frac{k_t}{n}}{ \frac{(k_t-1)(1-p)}{n-1}+ p} 
     + (n-k_t)\kl{\frac{k_t}{n}}{ \frac{k_t(1-p)}{n-1}}}\label{eq:info-ub-never}
\end{align}
If $k_t = 1$, then
\begin{align*}
    k_t\cdot \kl{\frac{k_t}{n}}{ \frac{(k_t-1)(1-p)}{n-1}+ p} = \kl{\frac{1}{n}}{ p}\leq \frac{1}{n} \log(\frac{1}{p})\,,
\end{align*}
and 
\begin{align*}
    (n-k_t)\cdot\kl{\frac{k_t}{n}}{ \frac{k_t(1-p)}{n-1}}= (n-1)\cdot\kl{\frac{1}{n}}{\frac{1-p}{n-1}}\leq \frac{1}{(1-p)n(n-2)}\,,
\end{align*}
where the ineq holds due to $\kl{q}{q'}\leq \frac{(q-q')^2}{q'(1-q')}$.
If $k_t = n-1$, it is symmetric to the case of $k_t = 1$. We have
\begin{align*}
    &k_t\cdot \kl{\frac{k_t}{n}}{ \frac{(k_t-1)(1-p)}{n-1}+ p} = (n-1)\kl{\frac{n-1}{n}}{\frac{n-2}{n-1} +\frac{1}{n-1}p}=
    (n-1)\kl{\frac{1}{n}}{\frac{1-p}{n-1}}\\
    \leq& \frac{1}{(1-p)n(n-2)}\,,
\end{align*}
and 
\begin{align*}
    (n-k_t)\cdot\kl{\frac{k_t}{n}}{ \frac{k_t(1-p)}{n-1}}= \kl{\frac{n-1}{n}}{1-p}= 
     \kl{\frac{1}{n}}{ p}\leq \frac{1}{n} \log(\frac{1}{p})\,.   
\end{align*}
If $1<k_t<n-1$, then
\begin{align*}
    k_t\cdot\kl{\frac{k_t}{n}}{ \frac{(k_t-1)(1-p)}{n-1}+ p} = &k_t\cdot\kl{\frac{k_t}{n}}{ \frac{k_t-1}{n-1} +\frac{n-k_t}{n-1}p}\stackrel{(a)}{\leq} 
    k_t\cdot\kl{\frac{k_t}{n}}{ \frac{k_t-1}{n-1}}\\
    \stackrel{(b)}{\leq}&
    k_t\cdot\frac{(\frac{k_t}{n}-\frac{k_t-1}{n-1} )^2}{\frac{k_t-1}{n-1}(1-\frac{k_t-1}{n-1})}= k_t\cdot\frac{n-k_t}{n^2(k_t-1)}\leq \frac{k_t\cdot}{n(k_t-1)}\leq \frac{2}{n}\,,
\end{align*}
where inequality (a) holds due to that $\frac{k_t-1}{n-1} +\frac{n-k_t}{n-1}p \leq \frac{k_t}{n}$ and $\kl{q}{q'}$ is monotonically decreasing in $q'$ when $q'\leq q$ and inequality (b) adopts $\kl{q}{q'}\leq \frac{(q-q')^2}{q'(1-q')}$,
and
\begin{align*}
    (n-k_t)\cdot\kl{\frac{k_t}{n}}{ \frac{k_t(1-p)}{n-1}}\leq (n-k_t)\cdot\kl{\frac{k_t}{n}}{ \frac{k_t}{n-1}}\leq \frac{k_t(n-k_t)}{n^2(n-1-k_t)}\leq \frac{2k_t}{n^2}\,,
\end{align*}
where the first inequality hold due to that $\frac{k_t(1-p)}{n-1} \geq \frac{k_t}{n}$, and $\kl{q}{q'}$ is monotonically increasing in $q'$ when $q'\geq q$ and the second inequality adopts $\kl{q}{q'}\leq \frac{(q-q')^2}{q'(1-q')}$.
Therefore, we have
\begin{equation*}
    \text{Eq~\eqref{eq:info-ub-never}} \leq \frac{6(n-1)\epsilon}{n} \sum_{t=1}^T \EEs{a_{1:T}\sim \bar \bP}{\frac{2}{n}\log(\frac{1}{p})} \leq \frac{12\epsilon T\log(1/p)}{n} \,.
\end{equation*}
Combining with Eq~\eqref{eq:tv-lb-never}, we have that there exists a universal constant $c$ such that $T\geq \frac{cn}{\epsilon (\log(n/\epsilon) +1 )}$.
\end{proof}
\section{Proof of Theorem~\ref{thm:non-ball}}\label{app:non-ball}

\begin{proof}
We will prove Theorem~\ref{thm:non-ball} by constructing an instance of $\cQ$ and $\cH$ and showing that for any learning algorithm, there exists a realizable data distribution s.t. achieving $\epsilon$ loss requires at least $\tilde \Omega(\frac{\abs{\cH}}{\epsilon})$ samples.
\paragraph{Construction of $\cQ$, $\cH$ and a set of realizable distributions} 
    \begin{itemize}
        \item Let feature vector space $\cX = \{0,1,\ldots,n\}$ and let the space of feature-manipulation set pairs $\cQ = \{(0,\{0\}\cup s)|s\subset [n]\}$.
        That is to say, every agent has the same original feature vector $x =0$ but has different manipulation ability according to $s$.
        \item Let the hypothesis class be a set of singletons over $[n]$, i.e., $\cH = \{2\ind{\{i\}}-1|i\in [n]\}$.
        \item We now define a collection of distributions $\{\cD_i|i\in [n]\}$ in which $\cD_i$ is realized by $2\ind{\{i\}}-1$.
        For any $i\in [n]$, let $\cD_{i}$ put probability mass $1-6\epsilon$ on $(0,\cX,+1)$ and $6\epsilon$ uniformly over $\{(0, \{0\}\cup s_{\sigma, i}, -1)|\sigma \in \cS_n\}$, where $\cS_n$ is the set of all permutations over $n$ elements and $s_{\sigma, i}:= \{j|\sigma^{-1}(j)< \sigma^{-1}(i)\}$ is the set of elements appearing before $i$ in the permutation $(\sigma(1),\ldots,\sigma(n))$.
        In other words, with probability $1-6\epsilon$, we will sample $(0,\cX,+1)$ and with $\epsilon$, we will randomly draw a permutation $\sigma\sim \Unif(\cS_n)$ and return $(0, \{0\}\cup s_{\sigma, i}, -1)$. 
        The data distribution $\cD_{i}$ is realized by $2\ind{\{i\}}-1$ since for negative examples $(0, \{0\}\cup s_{\sigma, i}, -1)$, we have $i\notin s$ and for positive examples $(0,\cX,+1)$, we have $i\in \cX$.
    \end{itemize}

    \paragraph{Randomization and improperness of the output $f_\out$ do not help}
    Note that algorithms are allowed to output a randomized $f_\out$ and to output $f_\out\notin \cH$.
    We will show that randomization and improperness of $f_\out$ don't make the problem easier.
    That is, supposing that the data distribution is $\cD_{i^*}$ for some $i^*\in [n]$, finding a (possibly randomized and improper) $f_\out$ is not easier than identifying $i^*$.
    Since our feature space $\cX$ is finite, we can enumerate all hypotheses not equal to $2\ind{\{i^*\}}-1$ and calculate their strategic population loss as follows.
    \begin{itemize}
        \item $2\ind{\emptyset}-1$ predicts all points in $\cX$ by negative and thus
    $\err(2\ind{\emptyset}-1) = 1-6\epsilon$;
    \item For any $a\subset \cX$ s.t. $0\in a$, $2\ind{a}-1$ will predict $0$ as positive and thus will predict any point drawn from $\cD_{i^*}$ as positive.
    Hence
    $\err(2\ind{a}-1) = 6\epsilon$;
    \item For any $a\subset [n]$ s.t. $\exists i\neq i^*$, $i\in a$, we have  $\err(2\ind{a}-1)\geq 3\epsilon$. This is due to that when $y=-1$, the probability of drawing a permutation $\sigma$ with $\sigma^{-1}(i)<\sigma^{-1}(i^*)$ is $\frac{1}{2}$. In this case, we have $i\in s_{\sigma,i^*}$ and the prediction of $2\ind{a}-1$ is $+1$.
    \end{itemize}
    Under distribution $\cD_{i^*}$, if we are able to find a (possibly randomized) $f_\out$ with strategic loss $\err(f_\out)\leq \epsilon$, then we have $\err(f_\out) = \EEs{h\sim f_\out}{\err(h)}\geq \Pr_{h\sim f_\out}(h\neq 2\ind{\{i^*\}}-1) \cdot 3\epsilon$.
    Thus, $\Pr_{h\sim f_\out}(h= 2\ind{\{i^*\}}-1)\geq \frac{2}{3}$ and then, we can identify $i^*$ by checking which realization of $f_\out$ has probability greater than $\frac{2}{3}$.
    In the following, we will focus on the sample complexity to identify the target function $2\ind{\{i^*\}}-1$ or simply $i^*$.
    Let $i_\out$ denote the algorithm's answer to question of ``what is $i^*$?''.

\paragraph{Smoothing the data distribution} 
    For technical reasons (appearing later in the analysis), we don't want to analyze distribution $\{\cD_i|i\in [n]\}$ directly as the probability of $\Delta_t = i^*$ is $0$ when $f_t(i^*) = +1$.
    Instead, we consider the mixture of $\cD_i$ and another distribution $\cD_i''$ to make the probability of $\Delta_t = i^*$ be a small positive number.
    More specifically, let $\cD_i' = (1-p) \cD_i + p  \cD_i''$, where $\cD_i''$
    is defined by drawing $(0,\cX, +1)$ with probability $1-6\epsilon$ and $(0,\{0,i\}, -1)$ with probability $6\epsilon$. When $p$ is extremely small, we will never sample from $\cD_i''$ when time horizon $T$ is not too large and therefore, the algorithm behaves the same under $\cD_i'$ and $\cD_i$.
    For any data distribution $\cD$, let $\bP_\cD$ be the dynamics of $(x_1,f_1,\Delta_1, y_1,\hat y_1,\ldots, x_T, f_T,\Delta_T, y_T,\hat y_T)$ under $\cD$.
    According to Lemma~\ref{lmm:smooth}, by setting $p=\frac{\epsilon}{16n^2}$, when $T\leq \frac{n}{\epsilon}$, we have that for any $i,j\in [n]$
    \begin{equation}
        \abs{\bP_{\cD_{i}}(i_\out = j)-\bP_{\cD_{i}'}(i_\out = j)}\leq \frac{1}{8}\,.\label{eq:smooth-res}
    \end{equation}
   
    From now on, we only consider distribution $\cD_i'$ instead of $\cD_i$. The readers might have the question that why not using $\cD_i'$ for construction directly. This is because no hypothesis has zero loss under $\cD_i'$, and thus $\cD_i'$ does not satisfy realizability requirement.

    \paragraph{Information gain from different choices of $f_t$} In each round of interaction, the learner picks a predictor $f_t$, which can be out of $\cH$.
    Suppose that the target function is $2\ind{\{i^*\}}-1$ .
    Here we enumerate all choices of $f_t$ and discuss how much we can learn from each choice.
    \begin{itemize}
        \item $f_t = 2\ind{\emptyset}-1$ predicts all points in $\cX$ by negative. 
        No matter what $i^*$ is, we will observe $\Delta_t = x_t = 0$, $y_t\sim \Rad(1-6\epsilon)$, $\hat y_t = -1$. They are identically distributed for any $i^*\in [n]$ and thus we cannot tell any information of $i^*$ from this round.

        \item $f_t=2\ind{a_t}-1$ for some $a_t\subset \cX$ s.t. $0\in a_t$. 
        Then no matter what $i^*$ is, we will observe $\Delta_t = x_t = 0$, $y_t\sim \Rad(1-6\epsilon)$, $\hat y_t = +1$.  Again, we cannot tell any information of $i^*$ from this round.

        \item $f_t = 2\ind{a_t}-1$ for some some non-empty $a_t\subset [n]$. 
        For rounds with $y_t = +1$,  we have $x_t =0, \hat y_t = +1$ and $\Delta_t = \Delta(0, f_t, \cX)\sim \Unif(a_t)$,
        which still do not depend on $i^*$.
        For rounds with $y_t=-1$, if the drawn example $(0,\{0\}\cup s, -1)$ satisfies that $s\cap a_t\neq \emptyset$, the we would observe $\Delta_t \in a_t$ and $\hat y_t = +1$.
        At least we could tell that $\ind{\{\Delta_t\}}$ is not the target function.
       Otherwise, we would observe $\Delta_t=x_t=0$ and $\hat y_t = -1$.
    \end{itemize}
    Therefore, we can only gain some information about $i^*$ at rounds in which $f_t = 2\ind{a_t}-1$ for some non-empty $a_t\subset [n]$ and $y_t =-1$.
    In such rounds, under distribution $\cD_{i^*}'$, the distribution of $\Delta_t$ is described as follows.
    Let $k_t = \abs{a_t}$ denote the cardinality of $a_t$.
    Recall that agent $(0,\{0\}\cup s, -1)$ breaks ties randomly when choosing $\Delta_t$ if there are multiple elements in $a_t\cap s$.
    Here are two cases: $i^* \in a_t$ and $i^* \notin a_t$.
    \begin{enumerate}
        \item The case of $i^* \in a_t$: With probability $p$, we are sampling from $\cD_{i^*}''$ and then $\Delta_t = i^*$.
        With probability $1-p$, we are sampling from $\cD_{i^*}$. 
        Conditional on this, with probability $\frac{1}{k_t}$, we sample an agent $(0,\{0\}\cup s_{\sigma, i^*}, -1)$ with the permutation $\sigma$ satisfying that $\sigma^{-1}(i^*)<\sigma^{-1}(j)$ for all $j\in a_t\setminus \{i^*\}$ and thus, $\Delta_t = 0$.
        With probability $1-\frac{1}{k_t}$, there exists $j\in a_t\setminus \{i^*\}$ s.t. $\sigma^{-1}(j)<\sigma^{-1}(i^*)$ and $\Delta_t \neq 0$. Since all $j\in a_t\setminus \{i^*\}$ are symmetric, we have $\Pr(\Delta_t = j) = (1-p)(1-\frac{1}{k_t}) \cdot \frac{1}{k_t-1} = \frac{1-p}{k_t}$.
        Hence, the distribution of $\Delta_t$ is
        \begin{align*}
            \Delta_t = \begin{cases}
                j & \text{w.p. } \frac{1-p}{k_t} \text{ for } j \in a_t, j\neq i^*\\
                i^* & \text{w.p. } p \\
               0 & \text{w.p. } \frac{1-p}{k_t}\,.
            \end{cases}
        \end{align*}
        We denote this distribution by $P_\in(a_t, i^*)$.
        \item The case of $i^*\notin a_t$: With probability $p$, we are sampling from $\cD_{i^*}''$, we have $\Delta_t=x_t=0$.
        With probability $1-p$, we are sampling from $\cD_{i^*}$.
        Conditional on this, with probability of $\frac{1}{k_t+1}$, $\sigma^{-1}(i^*)<\sigma^{-1}(j)$ for all $j\in a_t$ and thus, $\Delta_t =x_t= 0$.
        With probability $1-\frac{1}{k_t+1}$
        there exists $j\in a_t$ s.t. $\sigma^{-1}(j)<\sigma^{-1}(i^*)$ and $\Delta_t \in a_t$. Since all $j\in a_t$ are symmetric, we have $\Pr(\Delta_t = j) = (1-p)(1-\frac{1}{k_t+1}) \cdot \frac{1}{k_t} = \frac{1-p}{k_t+1}$.
        Hence the distribution of $\Delta_t$ is
        \begin{align*}
            \Delta_t = \begin{cases}
                j & \text{w.p. } \frac{1-p}{k_t+1} \text{ for } j \in a_t\\
                0 & \text{w.p. } p+ \frac{1-p}{k_t+1}\,.
            \end{cases}
        \end{align*}
        We denote this distribution by $P_{\notin}(a_t)$.
    \end{enumerate}
To measure the information obtained from $\Delta_t$, we will use the KL divergence of the distribution of $\Delta_t$
 under the data distribution $\cD_{i^*}'$ from that under a benchmark data distribution.
We use the average distribution over $\{\cD_i'|i\in [n]\}$, which is denoted by $\bar \cD = \frac{1}{n}\sum_{i\in n}\cD_i'$.
The sampling process is equivalent to drawing $i^*\sim \Unif([n])$ first and then sampling from $\cD'_{i^*}$.
Under $\bar \cD$, for any $j\in a_t$, we have
    \begin{align*}
        \Pr(\Delta_t = j) 
        =& \Pr(i^*\in a_t\setminus\{j\})\Pr(\Delta_t = j|i^*\in a_t\setminus\{j\})+ \Pr(i^*=j)\Pr(\Delta_t = j|i^*=j) \\
        &+ \Pr(i^*\notin a_t)\Pr(\Delta_t = \be_j|i^*\notin a_t) \\&=\frac{k_t-1}{n}\cdot \frac{1-p}{k_t}+\frac{1}{n}\cdot p + \frac{n-k_t}{n}\cdot \frac{1-p}{k_t+1} = \frac{(nk_t-1)(1-p)}{nk_t(k_t+1)} +\frac{p}{n}\,,
    \end{align*}
    and
    \begin{align*}
        \Pr(\Delta_t = 0) &= \Pr(i^*\in a_t)\Pr(\Delta_t = 0|i^*\in a_t)+ \Pr(i^*\notin a_t)\Pr(\Delta_t = 0|i^*\notin a_t)\\&=\frac{k_t}{n}\cdot \frac{1-p}{k_t} + \frac{n-k_t}{n}\cdot (p+ \frac{1-p}{k_t+1}) = \frac{(n+1)(1-p)}{n(k_t+1)} + \frac{(n-k_t)p}{n}\,.
    \end{align*}
    Thus, the distribution of $\Delta_t$ under $\bar\cD$ is
    \begin{align*}
            \Delta_t = \begin{cases}
                j & \text{w.p. } \frac{(nk_t-1)(1-p)}{nk_t(k_t+1)} +\frac{p}{n}\text{ for } j \in a_t\\
                0 & \text{w.p. } \frac{(n+1)(1-p)}{n(k_t+1)} + \frac{(n-k_t)p}{n}\,.
            \end{cases}
    \end{align*}
    We denote this distribution by $\bar P(a_t)$.
    Next we will compute the KL divergence of $P_{\notin}(a_t)$ and $P_{\in}(a_t)$ from $\bar P(a_t)$.
    Since $p =\frac{\epsilon}{16n^2}\leq \frac{1}{16n^2}$, we have $\frac{(nk_t-1)(1-p)}{nk_t(k_t+1)} +\frac{p}{n} \leq \frac{1-p}{k_t+1}$ and $\frac{(n+1)(1-p)}{n(k_t+1)} + \frac{(n-k_t)p}{n} \leq \frac{1}{k_t}+p$.
    We will also use $\log(1+x)\leq x$ for $x\geq 0$ in the following calculation.
    For any $i^*\in a_t$, we have
    \begin{align}
        &\KL{\bar P(a_t)}{P_{\in}(a_t,i^*)} \nonumber\\
        =&(k_t-1)\left(\frac{(nk_t-1)(1-p)}{nk_t(k_t+1)} +\frac{p}{n}\right) \log\left((\frac{(nk_t-1)(1-p)}{nk_t(k_t+1)} +\frac{p}{n}) \cdot \frac{k_t}{1-p}\right) \nonumber\\
        &+ \left(\frac{(nk_t-1)(1-p)}{nk_t(k_t+1)} +\frac{p}{n}\right) \log\left((\frac{(nk_t-1)(1-p)}{nk_t(k_t+1)} +\frac{p}{n}) \cdot \frac{1}{p}\right) \nonumber\\
        &+\left( \frac{(n+1)(1-p)}{n(k_t+1)} + \frac{(n-k_t)p}{n}\right) \log \left(\left( \frac{(n+1)(1-p)}{n(k_t+1)} + \frac{(n-k_t)p}{n}\right) \cdot \frac{k_t}{1-p}\right)\nonumber\\
        \leq & (k_t-1)\left(\frac{(nk_t-1)(1-p)}{nk_t(k_t+1)} +\frac{p}{n}\right)\log(\frac{1-p}{k_t+1}\cdot \frac{k_t}{1-p}) + \frac{1-p}{k_t+1} \log(1\cdot\frac{1}{p}) \nonumber\\
        &+ (\frac{1}{k_t}+p)\cdot \log\left(1+pk_t\right)\nonumber\\
        \leq & 0 + \frac{1}{k_t+1}\log(\frac{1}{p}) + \frac{2}{k_t}\cdot p k_t =\frac{1}{k_t+1}\log(\frac{1}{p}) + 2p 
        \,.\label{eq:kl-in}
        \end{align}
        For $P_{\notin}(a_t)$, we have
        \begin{align}
            &\KL{\bar P(a_t)}{P_{\notin}(a_t)}\nonumber\\
            =&k_t\left(\frac{(nk_t-1)(1-p)}{nk_t(k_t+1)} +\frac{p}{n}\right) \log\left((\frac{(nk_t-1)(1-p)}{nk_t(k_t+1)} +\frac{p}{n}) \cdot \frac{k_t+1}{1-p}\right) \nonumber\\
        &+\left( \frac{(n+1)(1-p)}{n(k_t+1)} + \frac{(n-k_t)p}{n}\right) \log \left(\left( \frac{(n+1)(1-p)}{n(k_t+1)} + \frac{(n-k_t)p}{n}\right) \cdot \frac{1}{p+ \frac{1-p}{k_t+1}}\right)\nonumber\\
        \leq & k_t \left(\frac{(nk_t-1)(1-p)}{nk_t(k_t+1)} +\frac{p}{n}\right)\log(\frac{1-p}{k_t+1}\cdot \frac{k_t+1}{1-p}) \nonumber\\
        &+ (\frac{1}{k_t}+p)\log \left(\left( \frac{(n+1)(1-p)}{n(k_t+1)} + \frac{(n-k_t)p}{n}\right) \cdot \frac{1}{p+ \frac{1-p}{k_t+1}}\right)\nonumber\\
        = & 0+ (\frac{1}{k_t}+p) \log(1+ \frac{1-p(k_t^2+k_t+1)}{n(1+k_tp)})\nonumber
        \\
        \leq &(\frac{1}{k_t}+p) \frac{1}{n(1 + k_tp)}
        = \frac{1}{nk_t}\label{eq:kl-out}\,.
        \end{align}

\paragraph{Lower bound of the information} Now we adopt the similar framework used in the proofs of Theorem~\ref{thm:delta-csv} and \ref{thm:x-delta-never}.
For notation simplicity, for all $i\in [n]$, let $\bP_i$ denote the dynamics of $(x_1,f_1,\Delta_1, y_1,\hat y_1,\ldots, x_T, f_T,\Delta_T, y_T,\hat y_T)$ under $\cD_i'$ and and $\bar \bP$ denote the dynamics under $\bar \cD$.
Let $B_t$ denote the event of $\{f_t = 2\ind{a_t}-1 \text{ for some non-empty } a_t\subset [n]\}$.
As discussed before, for any $a_t$, conditional on $\neg B_t$ or $y_t =+1$,  $(x_t, \Delta_t, y_t,\hat y_t)$ are identical in all $\{\cD_i'|i\in [n]\}$, and therefore, also identical in $\bar \cD$.
We can only obtain information at rounds when $B_t \wedge (y_t =-1)$ occurs.
In such rounds, we know that $x_t$ is always $0$, $f_t$ is fully determined by history (possibly with external randomness , which does not depend on data distribution), $y_t =-1$ and $\hat y_t$ is fully determined by $\Delta_t$ ($\hat y_t = +1$ iff. $\Delta_t \neq 0$).

Therefore, conditional the history $H_{t-1} = (x_1,f_1,\Delta_1, y_1,\hat y_1,\ldots, x_{t-1}, f_{t-1},\Delta_{t-1}, y_{t-1},\hat y_{t-1})$ before time $t$, we have
\begin{align}
    &\KL{\bar \bP(x_{t}, f_{t},\Delta_{t}, y_{t},\hat y_{t}|H_{t-1})}{\bP_i(x_{t}, f_{t},\Delta_{t}, y_{t},\hat y_{t}|H_{t-1})} \nonumber\\
    =&\bar \bP(B_t  \wedge y_t =-1) \KL{\bar \bP(\Delta_t|H_{t-1},B_t  \wedge y_t =-1)}{\bP_i(\Delta_t|H_{t-1},B_t  \wedge y_t =-1)} \nonumber\\
    =& 6\epsilon \bar \bP(B_t) \KL{\bar \bP(\Delta_t|H_{t-1},B_t  \wedge y_t =-1)}{\bP_i(\Delta_t|H_{t-1},B_t  \wedge y_t =-1)}\label{eq:kl-delta}\,,
\end{align}
where the last equality holds due to that $y_t\sim \Rad(1-6\epsilon)$ and does not depend on $B_t$.

For any algorithm that can successfully identify $i$ under the data distribution $\cD_i$ with probability $\frac{3}{4}$ for all $i\in [n]$, then $\bP_{\cD_i}(i_\out = i)\geq \frac{3}{4}$ and $\bP_{\cD_j}(i_\out = i)\leq \frac{1}{4}$ for all $j\neq i$.
Recall that $\cD_i$ and $\cD_i'$ are very close when the mixture parameter $p$ is small. Combining with Eq~\eqref{eq:smooth-res}, we have 
\begin{align*}
    &\abs{\bP_{i}(i_\out = i) - \bP_{j}(i_\out = i)}\\
    \geq& \abs{\bP_{\cD_i}(i_\out = i) - \bP_{\cD_j}(i_\out = i)} - \abs{\bP_{\cD_i}(i_\out = i)- \bP_{i}(i_\out = i)}- \abs{\bP_{\cD_j}(i_\out = i)- \bP_{j}(i_\out = i)} \\
    \geq&\frac{1}{2}- \frac{1}{4} = \frac{1}{4}\,.
\end{align*}

Then we have the total variation distance between $\bP_{i}$ and $\bP_{j}$
\begin{align}
       \TV(\bP_{i},\bP_{j})\geq \abs{\bP_{i}(i_\out = i) - \bP_{j}(i_\out = i)}\geq \frac{1}{4}\,.\label{eq:tv-lb}
\end{align}

Then we have
    \begin{align*}
    &\EEs{i\sim \Unif([n])}{\TV^2(\bP_{i},\bP_{(i+1)\text{ mod } n})} \leq 4 \EEs{i\sim \Unif([n])}{\TV^2(\bP_{i},\bar \bP )}\\
    \leq& 2\EEs{i}{\KL{\bar \bP}{\bP_{i}}} \tag{Pinsker's ineq}\\
    =& 2\EEs{i}{\sum_{t=1}^T \KL{\bar \bP(x_{t}, f_{t},\Delta_{t}, y_{t},\hat y_{t}|H_{t-1})}{\bP_i(x_{t}, f_{t},\Delta_{t}, y_{t},\hat y_{t}|H_{t-1})} }\tag{Chain rule}\\
    \leq & 12\epsilon\EEs{i}{\sum_{t=1}^T \bar \bP(B_t) \KL{\bar \bP(\Delta_t|H_{t-1},B_t  \wedge y_t =-1)}{\bP_i(\Delta_t|H_{t-1},B_t  \wedge y_t =-1)}}\tag{Apply Eq~\eqref{eq:kl-delta}}\\
     \leq& \frac{12\epsilon}{n} \sum_{t=1}^T \bar \bP(B_t)\sum_{i=1}^n \KL{\bar \bP(\Delta_t|H_{t-1},B_t  \wedge y_t =-1)}{\bP_i(\Delta_t|H_{t-1},B_t  \wedge y_t =-1)}\\
     =& \frac{12\epsilon}{n}  \EEs{a_{1:T}\sim \bar \bP}{\sum_{t=1}^T\true{B_t}\left(\sum_{i:i\in a_t} \KL{\bar P(a_t)}{P_{\in}(a_t)}
     + \sum_{i:i\notin a_t}\KL{\bar P(a_t)}{P_{\notin}(a_t)}\right)}\\
     \leq & \frac{12\epsilon}{n}  \EEs{a_{1:T}\sim \bar \bP}{\sum_{t:\true{B_t} =1}\left(\sum_{i:i\in a_t} \left(\frac{1}{k_t+1}\log(\frac{1}{p}) + 2p
     \right) + \sum_{i:i\notin a_t}\frac{1}{n k_t}\right)}\tag{Apply Eq~\eqref{eq:kl-in},\eqref{eq:kl-out}}\\
     \leq & \frac{12\epsilon}{n} \sum_{t=1}^T (\log(\frac{1}{p})+ 2np + 1)\\
     \leq & \frac{12T\epsilon(\log(16n^2/\epsilon) + 2)}{n}\,.
\end{align*}
Combining with Eq~\eqref{eq:tv-lb}, we have that there exists a universal constant $c$ such that $T\geq \frac{cn}{\epsilon (\log(n/\epsilon) +1 )}$. 
\end{proof}

\chapter{Strategic Classification for Infinite Hypothesis Class}\label{app:strategic-infinite}

\section{Learning Complexity- Background}\label{app:background}
Here we recall the definitions of the Vapnik–Chervonenkis dimension  ($\vcd$)~\citep{vapnik:74}, and Littlestone dimension ($\ld$) \citep{littlestone1988learning}. 

\begin{definition}[VC-dimension]\label{def:vc}
Let $\mathcal{\mathcal{H}} \subseteq \cY^\mathcal{X}$ be a hypothesis class. A subset $S = \{x_1, ..., x_{|S|}\} \subseteq \mathcal{X}$ is shattered by $\mathcal{H}$ if:
$\left| \left\{
\left(h(x_1), ..., h(x_{|S|})\right) : h \in \mathcal{H}
\right\} \right|
= 2^{|S|}$.
The VC-dimension of $\mathcal{H}$, denoted $VCdim(\mathcal{H})$, is the maximal cardinality of a subset $S \subseteq \mathcal{X}$ shattered by $\mathcal{H}$.
\end{definition}

\begin{definition}[$\cH$-shattered tree]
    Let $\cH \subseteq \cY^\cX$ be a hypothesis class and let $d\in\mathbb N$. A sequence $(x_1,\ldots,x_{2^d-1}) \in \X^{2^d-1}$ is an $\cH$\textup{-shattered tree} of depth $d$ if, for every labeling $(y_1,\ldots,y_d)\in\cY^d$, there exists $h\in\cH$ such that for all $i \in [d]$ we have that $h(x_{j_i}) = y_i$, where $j_i = 2^{i-1} + \sum_{k=1}^{i-1} y_k 2^{i-1-k}$. 
\end{definition}

\begin{definition}[Littlestone dimension] 
The \textup{Littlestone dimension} of a hypothesis class $\cH$ is the maximal
depth of a tree shattered by $\cH$. 
\end{definition}

\section{Fully Informative Setting}\label{app:fullInfo}
In this setting, the manipulation graph $G^\star$ is known and the learner observes $x_t$ before implementing $h_t$. We remark that though the learner has knowledge of the entire graph $G^\star$ in this model, the algorithms in this section only require access to the out-neighborhood $N_{G^\star}(x_t)$ of $x_t$.
\subsection{PAC Learning}
In the PAC setting, no matter what learner $h_t$ implemented in each round, we obtain $T$ i.i.d. examples $S = ((x_1, y_1),\ldots, (x_T, y_T))$ after $T$ rounds of interaction.
Then, by running ERM over $\cH$ w.r.t. the strategic loss, we can obtain a hypothesis $\hat h$ by minimizing the empirical strategic loss,
\begin{align*}
\eqERM
\end{align*}
Since $\lstr_{G^\star}(h,(x,y))$ (see Def~\ref{def:lstr}) only depends on the graph $G^\star$ through the neighborhood $N_{G^\star}(x)$, we can optimize Eq~\eqref{eq:erm} given only $\{N_{G^\star}(x_t)|t\in [T]\}$ instead of the entire graph $G^\star$.
Readers familiar with VC theory might notice that the sample complexity can be bounded by the VC dimension of the $(G^\star,\cH)$-induced labeling class $\bar \cH_{G^\star}$, $\vcd(\bar \cH_{G^\star})$.  This is correct and has been shown in~\cite{sundaram2021pac}. However, since our goal is to connect the PAC learnability in the strategic setting and standard setting, we provide sample complexity guarantees dependent on the VC dimension of $\cH$. In fact, our sample complexity results are proved by providing upper and lower bounds of $\vcd(\bar \cH_{G^\star})$ using $\vcd(\cH)$.

\vcdinduced*
\begin{proof}
    For any $x\in \cX$, if two hypotheses $h, h'$ label its closed neighborhood $N_{G^\star}[x] = \{x\}\cup N_{G^\star}(x)$ in the same way, i.e., $h(z)=h'(z), \forall z\in N_{G^\star}[x]$, then the labeling of $x$ induced by $(G^\star,h)$ and $(G^\star,h')$ are the same, i.e., $\bar{h}_{G^\star}(x) = \bar{h'}_{G^\star}(x)$. 
Hence, for any $n$ points $X = \{x_1,x_2,\ldots,x_n\}\subset \cX$, the labeling of $X$ induced by $(G^\star,h)$ is determined by the prediction of $h$ at their closed neighborhoods, $h(\cup_{x\in X}N_{G^\star}[x])$. 
Since $\abs{\cup_{x\in X}N_{G^\star}[x]}\leq n(k+1)$, there are at most $O((nk)^d)$ different ways of implementation since $\vcd(\cH) = d$ by Sauer-Shelah-Perels Lemma.
If $n$ points can be shattered by $\bar \cH_{G^\star}$, then the number of their realized induced labeling ways is $2^n$, which should be bounded by the number of possible implementations of their neighborhoods, which is upper bounded by $O((nk)^d)$.
Hence, $2^n\leq O((nk)^d)$ and thus, we have $n\leq O(d\log(kd))$. Hence, we have $\vcd(\bar \cH_{G^\star}) \leq O(d\log(kd))$.

For the lower bound, let us start with the simplest case of $d=1$. Consider $(k+1)\log k$ nodes $\{x_{i,j}|i = 1,\ldots,\log k, j=0,\ldots, k\}$, where $x_{i,0}$ is connected to $\{x_{i,j}|j=1,\ldots, k\}$ for all $i\in [\log k]$. 
For $j=1,\ldots,k$, let $\text{bin}(j)$ denote the $\log k$-bit-binary representation of $j$. 
Let hypothesis $h_j$ be the hypothesis that labels $x_{i,j}$ by $\text{bin}(j)_i$ and all the other nodes by $0$.
Let $\cH = \{h_j|j\in [k]\}$.
Note that $\vcd(\cH) = 1$. This is because for any two points $x_{i,j}$ and $x_{i',j'}$, if $j\neq j'$, no hypothesis will label them as positive simultaneously; if $j=j'$, they can only be labeled as $(\text{bin}(j)_i,\text{bin}(j)_{i'})$ by $h_j$ and $(0,0)$ by all the other hypotheses.
However, $\bar \cH_{G^\star}$ can shatter $\{x_{1,0},x_{2,0},\ldots,x_{\log k, 0}\}$ since $h_j$'s labeling over  $\{x_{1,0},x_{2,0},\ldots,x_{\log k, 0}\}$ is $\text{bin}(j)$.
We can extend the example to the case of $\vcd{\cH} = d$ by making $d$ independent copies of $(\mathcal X,\cH, G^\star)$.
\end{proof}

Following directly Theorem~\ref{thm:vcd-induced} through the tools from VC theory~\citep{vapnik:71,vapnik:74}, we prove the following sample complexity upper and lower bounds.

\begin{restatable}{theorem}{thmFiPacRel}\label{thm:fi-pac}
    For any hypothesis class $\cH$ with $\vcd(\cH)=d$, graph $G^\star$ with maximum out-degree $k$, any data distribution $\cD$ and any $\epsilon,\delta\in (0,1)$, with probability at least $1-\delta$ over $S\sim \cD^T$ where $T = O(\frac{d\log(kd)+\log(1/\delta)}{\epsilon^2})$, the output $\hat h$ satisfies $\Lstr_{G^\star,\cD}(\hat h)\leq \min_{h^\star\in \cH}\Lstr_{G^\star,\cD}(h^\star)+ \epsilon$.

    In the realizable case, the output $\hat h$ satisfies $\Lstr_{G^\star,\cD}(\hat h)\leq \epsilon$ when $T = O(\frac{d\log(kd)\log(1/\epsilon)+\log(1/\delta)}{\epsilon})$.
\end{restatable}




\begin{restatable}{theorem}{fiPacLb}\label{thm:fi-pac-lb}
    For any $k,d>0$, there exists a graph $G^\star$ with maximum out-degree $k$ and a hypothesis class $\cH$ with VC dimension $d$ such that for any algorithm, there exists a distribution $\cD$ for which  achieving $\Lstr_{G^\star,\cD}(\hat h)\leq \min_{h^\star\in \cH}\Lstr_{G^\star,\cD}(h^\star)+ \epsilon$ with probability $1-\delta$ requires sample complexity $\Omega(\frac{d\log k + \log(1/\delta)}{\epsilon^2})$. In the realizable case, it requires sample complexity $\Omega(\frac{d\log k + \log(1/\delta)}{\epsilon})$.
\end{restatable}



\subsection{Online Learning}
Recall that in the online setting, the agents are generated by an adversary, and the learner’s goal is to minimize
the Stackelberg regret (which is the mistake bound in the realizable case). 
Similar to Algorithm~\ref{alg:reduction2online-pmf}, we will show that any learning algorithm in the standard online setting can be converted to an algorithm in the strategic online setting, with the mistake bound increased by a logarithmic multiplicative factor of $k$. The main difference from Algorithm~\ref{alg:reduction2online-pmf} is that now we know $x_t$ before implementing the hypothesis, we do not need to reduce the weight of predicting positive by a factor $1/k$ as we did in line~\ref{alg-line:k-fraction-weight} of Algorithm~\ref{alg:reduction2online-pmf}.

\begin{algorithm}[t]\caption{Red2Online-FI: Reduction to online learning 
 in the fully informative setting}\label{alg:reduction2online-fi}
    \begin{algorithmic}[1]
        \STATE \textbf{Input: } a standard online learning algorithm $\cA$
        \STATE \textbf{Initialization: } let the expert set $E = \{\cA\}$ and weight $w_{\cA}$ =1
        \FOR{$t=1,2,\ldots$}
        \STATE \underline{Prediction}: after observing $x_t$
        \IF{$\sum_{A\in E: \exists x\in N_{G^\star}[x_t], A(x) = 1}w_{A} \geq {\sum_{A\in E} w_{A}}/{2}$} \STATE let $h_t(x_t) =1$ and $h_t(x)=0$ for all $x\neq x_t$
        \ELSE 
        \STATE let $h_t =\ind{\emptyset}$ be the all-negative function
        \ENDIF
        \STATE \underline{Update}:  when we make a mistake
        \IF{$y_t = 0$}
        \STATE for all $A\in E$ satisfying $\exists x\in N_{G^\star}[x_t], A(x) = 1$, feed $A$ with $(x,0)$ and update $w_{A} \leftarrow \frac{1}{2} w_{A}$
        \ELSIF{$y_t = 1$}
        \STATE for all $A \in E$ satisfying $\forall x\in N_{G^\star}[x_t], A(x) =0$
        \STATE for all $x\in N_{G^\star}[x_t]$, by feeding $(x,1)$ to $A$, we can obtain a new expert $A(x,1)$
        \STATE remove $A$ from $E$ and add $\{A(x,1)|x\in N_{G^\star}[x_t]\}$ to $E$ with weights $w_{
        A(x,1)} = \frac{w_{A}}{2\abs{N_{G^\star}[x_t]}}$
        \ENDIF
        \ENDFOR
    \end{algorithmic}
\end{algorithm}
\begin{restatable}{theorem}{fiOnline}\label{thm:fi-online}
For any hypothesis class $\cH$, graph $G^\star$ with maximum out-degree $k$, and a standard online learning algorithm with mistake bound $M$ for $\cH$, for any realizable sequence, we can achieve mistake bound of $O(M \log k)$ by Algorithm~\ref{alg:reduction2online-fi}.
By letting the SOA algorithm by~\cite{littlestone1988learning} as the input standard online algorithm $\cA$,  Red2Online-FI(SOA)
makes at most $O(\ld(\cH)\log k)$ mistakes.
\end{restatable}
\begin{proof}
Let $\cA$ be an online learning algorithm with mistake bound $M$.
Let $W_t$ denote the total weight of experts at the beginning of round $t$. When we make a mistake at round $t$, there are two cases.
\begin{itemize}
    \item False positive: It means that the induced labeling of $x_t$ is $1$ but the true label is $y_t = 0$. By our prediction rule, we have $\sum_{\cA_H\in E: \exists x\in N_{G^\star}[x_t], \cA_H(x) = 1}w_{\cA_H} \geq \frac{W_t}{2}$, and thus, we will reduce the total weight by at least $\frac{W_t}{4}$. Hence, we have $W_{t+1}\leq (1-\frac{1}{4})W_t$.
    \item False negative: It means that the induced labeling of $x_t$ is $0$, but the true label is $y_t = 1$. This implies that our learner $h_t$ labeled the entire neighborhood of $x_t$ with $0$, but $h^\star$ labels some point in the neighborhood by $1$.
By our prediction rule, we have $\sum_{\cA_H\in E:\forall x\in N_{G^\star}[x_t]\cA_H(x) = 0}w_{\cA_H} \geq \frac{W_t}{2}$.
We split $\cA_H$ into $\{\cA_{H_{x,+1}}|x\in N_{G^\star}[x_t]\}$ and split the weight $w_{\cA_H}$ equally and multiply by $\frac{1}{2}$. Thus, we have $W_{t+1} \leq (1-\frac{1}{4})W_t$.
\end{itemize}
Hence, whenever we make a mistake the total weight is reduced by a multiplicative factor of $\frac{1}{4}$. Let $N$ denote the total number of mistakes by Red2Online-FI($\cA$). Then we have total weight $\leq (1-1/4)^N$.

On the other hand, note that there is always an expert, whose hypothesis class contains $h^\star$, being fed with examples $(\pi_{G^\star,h^\star}(x_t),y_t)$.
At each mistake round $t$, if the expert (containing $h^\star$) makes a false positive mistake, its weight is reduced by half.
If the expert made a false negative mistake, it is split into a few experts, one of which is fed by $(\pi_{G^\star,h^\star}(x_t),y_t)$. This specific new expert will contain $h^\star$ and the weight is reduced by at most $\frac{1}{2(k+1)}$.
Thus, since this expert will make at most $M$ mistakes, the weight of such an expert will be at least $(\frac{1}{2(k+1)})^M$.

Thus we have $(1-1/4)^N \geq  1/(2(k+1))^M$, which yields the bound $N\leq 4M\ln(2(k+1))$.
\end{proof}

Next, we derive a lower bound for the number of mistakes. 
\begin{restatable}{theorem}{fiOnlineLb}\label{thm:fi-online-lb}
    For any $k,d>0$, there exists a graph $G^\star$ with maximum out-degree $k$ and a hypothesis class $\cH$ with Littlestone dimension $\ld(\cH) = d$ such that for any algorithm, there exists a realizable sequence for which the algorithm will make at least $d\log k$ mistakes.
\end{restatable}

\begin{proof}
    Consider the same hypothesis class $\cH$ and the graph $G^\star$ as Theorem~\ref{thm:fi-pac-lb}. Again, let us start with the simplest case of $d=1$. Consider $(k+1)\log k$ nodes $\{x_{i,j}|i = 1,\ldots,\log k, j=0,\ldots, k\}$, where $x_{i,0}$ is connected to $\{x_{i,j}|j=1,\ldots, k\}$ for all $i\in [\log k]$. 
For $j=1,\ldots,k$, let $\text{bin}(j)$ denote the $\log k$-bit-binary representation of $j$. 
Let hypothesis $h_j$ labels $x_{i,j}$ by $\text{bin}(j)_i$ and all the other nodes by $0$.
Let $\cH = \{h_j|j\in [k]\}$. The Littlestone dimension $\ld(\cH) = 1$ because there is an easy online learning in the standard setting by predicting the all-zero function until we observe a positive point $x_{i,j}$. Then we know $h_j$ is the target hypothesis.
    
At round $t$, the adversary picks $x_{t,0}$. For any deterministic learning algorithm:
\begin{itemize}
    \item If the learner predicts any point in $\{x_{t,j}|j=0,1,\ldots,k\}$ by positive, then the prediction is positive. We know half of the hypotheses predict negative. Hence the adversary forces a mistake by letting $y_t = 0$ and reduces half of the hypotheses.
    \item If the learner predicts all-zero over $\{x_{t,j}|j=0,1,\ldots,k\}$, the adversary forces a mistake by letting $y_t = 1$ and reduces half of the hypotheses.
\end{itemize}
We can extend the example to the case of $\ld{\cH} = d$ by making $d$ independent copies of $(X,\cH, G^\star)$. Then we prove the lower bound for all deterministic algorithms.
We can then extend the lower bound to randomized algorithms by the technique of \cite{FilmusHMM23}.
\end{proof}
\section{Post-Manipulation Feedback Setting}\label{app:postMani}
In this setting, the underlying graph $G^\star$ is known, but the original feature $x_t$ is not observable before the learner selects the learner. Instead, either the original feature $x_t$ or the manipulated feature $v_t$ is revealed afterward.
For PAC learning, since we can obtain an i.i.d. sample $(x_t,N_{G^\star}(x_t),y_t)$ by implementing $h_t = \1{\cX}$, we can still run ERM and obtain $\hat h$ by optimizing Eq~\eqref{eq:erm}. Hence, the positive PAC learning result (Theorem~\ref{thm:fi-pac}) in the fully informative setting should still apply here. Since the post-manipulation feedback setting is harder than the fully informative setting, the negative result (Theorem~\ref{thm:fi-pac-lb}) also holds. However, this is not true for online learning.

\subsection{Online Learning}
We have stated the algorithms and results in Section~\ref{sec:highlight-online}. 
Here we only provide the proofs for the theorems.

\fuOnline*
\begin{proof}
Given any online learning algorithm $\cA$ with mistake bound $M$.
Let $W_t$ denote the total weight of experts at the beginning of round $t$. When we make a mistake at round $t$, there are two cases.
\begin{itemize}
    \item False positive: It means that the induced labeling of $x_t$ is $1$ but the true label is $y_t = 0$. Since the true label is $0$, the neighbor $v_t$ we observe is labeled $0$ by the target hypothesis $h^\star$. In this case, we proceed by updating all experts that predict $v_t$ with $1$ with the example $(v_t,0)$ and we also half their weights. Since $h_t(v_t) = 1$, we have $\sum_{\cA_H \in E: \cA_H(x) = 1}w_{\cA_H} \geq \frac{W_t}{2(k+1)}$. Therefore, we have $W_{t+1}\leq W_t(1- \frac{1}{4(k+1)})$.

    \item False negative: It means that the induced labeling of $x_t$ is $0$, but the true label is $y_t = 1$. This implies that our learner $h_t$ labeled the entire neighborhood of $x_t = v_t$ with $0$, but $h^\star$ labels some point in the neighborhood by $1$.
By our prediction rule, we have 
\begin{align*}
    \sum_{\cA_H\in E:\forall x\in N_{G^\star}[x_t]\cA_H(x) = 0}w_{\cA_H}
    =&  W_t - \sum_{x\in N_{G^\star}[x_t]} \sum_{\cA_H\in E: \cA_H(x) = 1} w_{\cA_H}\\
    \geq&  (1- (k+1)\frac{1}{2(k+1)})W_t \\
    =&\frac{1}{2}W_t
\end{align*}
We split $\cA_H$ into $\{\cA_{H_{x,+1}}|x\in N_{G^\star}[x_t]\}$ and split the weight $w_{\cA_H}$ equally and multiply by $\frac{1}{2}$. Thus, we have $W_{t+1} \leq (1-\frac{1}{4})W_t$.
\end{itemize}

Hence, whenever we make a mistake the total weight is reduced by a multiplicative factor of $\frac{1}{4(k+1)}$. Let $N$ denote the total number of mistakes by Red2Online($\cA$)-FI. Then we have total weight $\leq (1-\frac{1}{4(k+1)})^N$.

On the other hand, note that there is always an expert, whose hypothesis class contains $h^\star$, being fed with examples $(\pi_{G^\star,h^\star}(x_t),y_t)$.
At each mistake round $t$, if the expert (containing $h^\star$) makes a false positive mistake, its weight is reduced by half.
If the expert made a false negative mistake, it is split into a few experts, one of which is fed by $(\pi_{G^\star,h^\star}(x_t),y_t)$. This specific new expert will contain $h^\star$ and the weight is reduced by at most $\frac{1}{2(k+1)}$.
Thus, since this expert will make at most $M$ mistakes, the weight of such an expert will be at least $(\frac{1}{2(k+1)})^M$.

Thus we have $(1-\frac{1}{4(k+1)})^N \geq  1/(2(k+1))^M$, which yields the bound $N\leq 4kM\ln(2(k+1))$.
\end{proof}

\fuOnlineLb*        
\begin{proof}
    Consider a star graph $G = (X,E)$ where $x_0$ is the center and $x_1,\ldots,x_k$ are the leaves. 
    The hypothesis class is singletons over leaves, i.e., $H = \{h_i|i\in [k]\}$ with $h_i = \1{x_i}$. Similar to the proof of Theorem~4.6 in  \cite{ahmadi2023fundamental}, any deterministic algorithm will make at least $k-1$ mistakes.
    At round $t$:
\begin{itemize}
    \item If the learner predicts all points in $X$ as $0$, the adversary picks the agent $x_{0}$. The agent will not move, the induced labeling is $0$, and the true label is $1$ (no matter which singleton is the target hypothesis). We do not learn anything about the target hypothesis.
    \item If the learner predicts $x_{0}$ as positive, then the adversary picks the agent  $x_{0}$. The agent does not move, the induced labeling is $1$, and the true label is $0$ (no matter which singleton is the target hypothesis). Again, we learn nothing about the target hypothesis.
    \item If the learner predicts any node $x_{i}$ with $i$ as positive, the adversary picks the agent $x_{i}$ and label it by $0$. The learner's induced labeling of $x_i$ is $1$, and the true label is $0$. Only one hypothesis $\1{X_i}$ is inconsistent and eliminated.
\end{itemize}
Hence, for any deterministic algorithm, there exists a realizable sequence such that the algorithm will make at least $k-1$ mistakes.

    Now consider $d$ independent copies of the star graphs and singletons. More specifically, consider $d(k+1)$ nodes $\{x_{i,j}|i = 1,\ldots,d, j=0,\ldots, k\}$, where $x_{i,0}$ is connected to $\{x_{i,j}|j=1,\ldots, k\}$ for all $i\in [\log k]$. So $\{x_{i,j}|j=0,\ldots, k\}$ compose a star graph.
    For each hypothesis in the hypothesis class $\cH$, it labels exactly one node in $\{x_{i,j}|j=1,\ldots, k\}$ as positive for all $i\in [d]$.
    Hence, $\cH$ has Littlestone dimension $\ld(\cH) = d$.
    Since every copy of the star graph-singletons is independent, we can show that any deterministic algorithm will make at least $d(k-1)$ mistakes.
\end{proof}

\section{Unknown Manipulation Graph Setting}\label{app:unknownGraph}
In this setting, the underlying graph $G^\star$ is unknown. Instead, the learner is given the prior knowledge of a finite graph class $\cG$. When $G^\star$ is undisclosed, we cannot compute $v_t$ given $x_t$. Consequently, the scenarios involving the observation timing of the features encompass the following: observing $x_t$ beforehand followed by $v_t$ afterward, observing $(x_t,v_t)$ afterward, and observing either $x_t$ or $v_t$ afterward, arranged in order of increasing difficulty. In this section, we provide results for the easiest case of observing $x_t$ beforehand followed by $v_t$ afterward. At the end, we will provide a negative result in the second easiest setting of observing $(x_t,v_t)$ afterward.

Since we not only have a hypothesis class $\cH$ but also a graph class $\cG$, we formally define realizability based on both the hypothesis class and the graph class as follows.

\begin{definition}[$(\cG,\cH)$-Realizability]\label{def:rel-graph}
A sequence of agents $(x_1,y_1),\ldots,(x_T,y_T)$ is  $(\cG,\cH)-$realizable if there exists a graph $G\in \cG$ such that the neighborhood of $x_t$ in $G$ is identical to that in $G^\star$, i.e., $N_{G}(x_t) = N_{G^\star}(x_t)$ and there exists a perfect hypothesis $h^\star\in \cH$ satisfying $\lstr_{G^\star}(h,(x_t,y_t)) =0$ for all $t=1,\ldots, T$.

For any data distribution $\cD$, we say that $\cD$ is $(\cG,\cH)$-realizable if there exists a graph $G\in \cG$ such that $\PPs{(x,y)\sim \cD}{N_{G}(x)\neq N_{G^\star}(x)} =0$ and there exists a perfect hypothesis  $h^\star\in \cH$ s.t. $\Lstr_{G^\star,\cD}(h^\star)= 0$.
\end{definition}
    
\subsection{PAC Learning}
\paragraph{Realizable PAC Learning} 
We have described our algorithm in Section~\ref{sec:highlight-ug-pac}. Here, we restate the algorithm and the theorems with the proofs.

\underline{Prediction:} At each round $t$, after observing $x_t$, we implement the hypothesis $h(x)=\1{x\neq x_t}$, which labels all every point as positive except $x_t$. Then we can obtain a manipulated feature vector $v_t\sim \Unif(N_{G^\star}(x))$. 

\underline{Output:} 
Let $W$ denote the all graph-hypothesis pairs of $(G,h)$ satisfying that
\begin{align}
    \sum_{t=1}^T \1{v_t\notin N_G(x_t)} = 0 \,,\label{eq:lossG-app} \\
    \sum_{t=1}^T \1{\bar{h}_G(x_t)\neq y_t} = 0\,,\label{eq:lossGH-app}
\end{align}
where Eq~\eqref{eq:lossG-app} guarantees that every observed feature $v_t$ is a neighbor of $x_t$ in $G$ and Eq~\eqref{eq:lossGH-app} guarantees that $h$ has zero empirical strategic loss when the graph is $G$.

Let 
$$(\hat G, \hat h) = \argmin_{(G,h)\in W} \sum_{t=1}^T \abs{N_G(x_t)}$$
be the graph-hypothesis pair such that the graph has a \textbf{minimal empirical degree}. Finally, we output $\hat h$.

\ugPacRel*
\begin{proof}
    Since the manipulation is local (i.e., agents can only manipulate to their neighbors), if the neighborhood of $x$ is the same in two different graphs $G_1, G_2$, i.e., $N_{G_1}(x) = N_{G_2}(x)$, then for any implementation $h$, the induced labeling of $x$ is the same $\bar{h}_{G_1}(x) = \bar{h}_{G_2}(x)$.
Therefore, the strategic loss of $\hat h$ can be written as
\begin{align}
\Lstr_{G^\star,\cD}(\hat h)= &
    \PPs{(x,y)\sim \cD}{\bar{\hat h}_{G^\star}(x)\neq y}\nonumber\\
    \leq& \PPs{(x,y)\sim \cD}{N_{G^\star}(x)\neq N_{\hat G}(x)} +\PPs{(x,y)\sim \cD}{\bar{\hat h}_{\hat G}(x)\neq y \wedge N_{G^\star}(x)= N_{\hat G}(x) }\nonumber\\
    \leq& \PPs{(x,y)\sim \cD}{N_{G^\star}(x)\neq N_{\hat G}(x)} +\PPs{(x,y)\sim \cD}{\bar{\hat h}_{\hat G}(x)\neq y}\,.\label{eq:loss}
\end{align}
We bound the second term by uniform convergence since its empirical estimate is zero (see Eq~\eqref{eq:lossGH-app}) in Lemma~\ref{lmm:pair-term} and the first term in Lemma~\ref{lmm:neighbor-term}.
\end{proof}

\begin{lemma}\label{lmm:pair-term}
    With probability at least $1-\delta$ over $(x_1,y_1),\ldots,(x_T,y_T)\sim \cD^T$, for all $(h,G)$ satisfying $\frac{1}{T}\sum_{t=1}^T \1{\bar{h}_{G}(x_t)\neq y_t}=0$, we have 
    $$\Pr_{(x,y)\sim \cD}(\bar{h}_{G}(x)\neq y) \leq \epsilon\,,$$
    when $T= O(\frac{(d \log(kd) + \log|\cG|)\log(1/\epsilon) + \log(1/\delta)}{\epsilon})$.
    
    Hence, we have
    $$\Pr_{(x,y)\sim \cD}(\bar{\hat h}_{\hat G}(x)\neq y) \leq \epsilon\,.$$
\end{lemma}
\begin{proof}
Let $\ell_{\text{pair}}(h,G) := \Pr_{(x,y)\sim \cD}(\bar{h}_{G}(x)\neq y)$ denote the loss of the hypothesis-graph pair and $\hat \ell_{\text{pair}}(h,G) = \frac{1}{T}\sum_{t=1}^T \1{\bar{h}_{G}(x_t)\neq y_t}$ denote the corresponding empirical loss.
To prove the lemma, it suffices to bound the VC dimension of the composite function class $\cH\circ \cG := \{\bar{h}_{G}|h\in \cH, G\in \cG\}$ and then apply the tool of VC theory.

Since for any $n$ points, the number of labeling ways by $\cH\circ \cG$ is upper bounded by the sum of the number of labeling ways by $\bar \cH_G$ over all graphs $G\in \cG$. As discussed in the proof of Theorem~\ref{thm:vcd-induced}, for any fixed $G$ with maximum out-degree $k$, the number of labeling ways by $\bar \cH_G$ is at most $O((nk)^d)$. Therefore, the total number of labeling ways over $n$ points by $\cH\circ \cG$ is at most $O(|\cG|(nk)^d)$. Hence, we have $\vcd(\cH\circ \cG) = O(d \log(kd) + \log(|\cG|))$.
Therefore, by VC theory, we have
\begin{align*}
    \Pr(\ell_{\text{pair}}(\hat h,\hat G) >\epsilon)\leq \Pr(\exists (h,G)\in \cH\times \cG \text{ s.t. }\ell_{\text{pair}}(h, G) >\epsilon, \hat \ell_{\text{pair}}(h, G)=0)\leq \delta
\end{align*}
when $T= O(\frac{(d \log(kd) + \log|\cG|)\log(1/\epsilon) + \log(1/\delta)}{\epsilon})$.

\end{proof}
\begin{lemma}\label{lmm:neighbor-term}
    With probability at least $1-\delta$ over $x_{1:T},v_{1:T}$, for all $G$ satisfying $\frac{1}{T}\sum_{t=1}^T\1{N_{G^\star}(x_t)\neq N_{G}(x_t)} = 0$, we have 
    $$\Pr_{(x,y)\sim \cD}(N_{G^\star}(x)\neq N_{G}(x)) \leq \epsilon\,,$$
when $T\geq \frac{8k\log(|\cG|/\delta)}{\epsilon}$. 

Hence, we have    
    $$\Pr_{(x,y)\sim \cD}(N_{G^\star}(x)\neq N_{\hat G}(x)) \leq \epsilon\,.$$
\end{lemma}

\begin{proof}[of Lemma~\ref{lmm:neighbor-term}]
Let $\lgraph (G) = \Pr_{(x,y)\sim \cD}(N_{G^\star}(x)\neq N_{G}(x))$ denote the loss of a graph $G$, which is the 0/1 loss of neighborhood prediction.
Let $\hatlgraph (G) = \frac{1}{T}\sum_{t=1}^T\1{N_{G^\star}(x_t)\neq N_{G}(x_t)}$ denote the corresponding empirical loss.
It is hard to get an unbiased estimate of the loss  $\lgraph (G)$ since we cannot observe the neighborhood of any sampled $x$. 
However, it is easy for us to observe a $v \in N_{G^\star}(x)$ and remove all inconsistent $G$. Then the challenge is: how can we figure out the case of a strictly larger graph? This corresponds to the case that $N_{G^\star}(x)$ is a strict subset of $N_{G}(x)$. We deal with this case by letting $\hat G$ be the ``smallest'' consistent graph.
\begin{claim}\label{cla:neighborhood-loss}
Suppose that $G^\star$ and all graphs $G\in \cG$ have maximum out-degree at most $k$. For any $G\in \cG$ with the minimal empirical degrees, i.e., $\sum_{t=1}^T\abs{N_{G}(x_t)} - \abs{N_{G^\star}(x_t)} \leq 0$, we have 
    $$\hatlgraph (G)\leq \frac{2k}{T}\sum_{t=1}^T \Pr_{v\sim \Unif(N_{G^\star}(x_t)) }( v\notin N_{G}(x_t))\,.$$
\end{claim}
\begin{proof}
We decompose
\begin{align}
    \1{N_{G^\star}(x)\neq N_{G}(x)}
    \leq \abs{N_{G}(x)\setminus N_{G^\star}(x)} + \abs{N_{G^\star}(x)\setminus N_{G}(x)}\,.\label{eq:decompose}
\end{align}

    Since $\sum_{t=1}^T\abs{N_{G}(x_t)} \leq \sum_{t=1}^T \abs{N_{G^\star}(x_t)}$, we have 
\begin{align}
    \sum_{t=1}^T\abs{N_{G}(x_t)\setminus N_{G^\star}(x_t)} \leq \sum_{t=1}^T\abs{N_{G^\star}(x_t)\setminus N_{G}(x_t)}\,,\label{eq:mindegree}
\end{align}
by subtracting $\sum_{t=1}^T |N_{G}(x_t)\cap N_{G^\star}(x_t)|$ on both sides.

By combining Eqs~\eqref{eq:decompose} and \eqref{eq:mindegree}, we have
\begin{align*}
    \hatlgraph (G) =&\frac{1}{T}\sum_{t=1}^T \1{N_{G^\star}(x_t)\neq N_{G}(x_t)}\\
    \leq &  \frac{2}{T}\sum_{t=1}^T\abs{N_{G^\star}(x_t)\setminus N_{G}(x_t)}\\
    = & \frac{2}{T}\sum_{t=1}^T\sum_{v\in N_{G^\star}(x_t)}\1{v\notin N_{G}(x_t)}\\
    = & \frac{2}{T}\sum_{t=1}^T \abs{N_{G^\star}(x_t)}\Pr_{v\sim \Unif(N_{G^\star}(x_t)) }( v\notin N_{G}(x_t))\\
    \leq & \frac{2}{T}\sum_{t=1}^T k\Pr_{v\sim \Unif(N_{G^\star}(x_t)) }( v\notin N_{G}(x_t))
    \,.
\end{align*}
\end{proof}
Now we have connected $\hatlgraph (G)$ with $\frac{1}{T}\sum_{t=1}^T \Pr_{v\sim \Unif(N_{G^\star}(x_t)) }( v\notin N_{G}(x_t))$, where the latter one is estimable. 
Since $\hat G$ is a consistent graph, we have
$$\frac{1}{T}\sum_{t=1}^T \1{ v_t\notin N_{\hat G}(x_t))} =0\,,$$
which is an empirical estimate of $\frac{1}{T}\sum_{t=1}^T \Pr_{v\sim \Unif(N_{G^\star}(x_t)) }( v\notin N_{\hat G}(x_t))$.
Then by showing that this loss is small, we can show that  $\hatlgraph (\hat G)$ is small.
\begin{claim}\label{cla:empirical-neighbor}
    Suppose that $G^\star$ and all graphs $G\in \cG$ have maximum out-degree at most $k$ and $\cD$ is $(\cG,\cH)$-realizable. For any fixed sampled sequence of $x_{1:T}$, with probability at least $1-\delta$ over $v_{1:T}$ (where $v_t$ is sampled from $\Unif(N_{G^\star}(x_t))$), we have 
    $$\hatlgraph (\hat G)\leq\epsilon\,,$$
    when $T\geq \frac{14k\log(|\cG|/\delta)}{3\epsilon}$.
\end{claim}
\begin{proof}
According to Claim~\ref{cla:neighborhood-loss}, we only need to upper bound $\frac{1}{T}\sum_{t=1}^T \Pr_{v\sim \Unif(N_{G^\star}(x_t)) }( v\notin N_{\hat G}(x_t))$ by $\frac{\epsilon}{2k}$.
For any graph $G$, by empirical Bernstein bounds~(Theorem 11 of \cite{maurer2009empirical}), with probability at least $1-\delta$ over $v_{1:T}$ we have
\begin{align*}
    \frac{1}{T}\sum_{t=1}^T \Pr_{v\sim \Unif(N_{G^\star}(x_t)) }( v\notin N_{G}(x_t))\leq &\frac{1}{T}\sum_{t=1}^T \1{v_t\notin N_{G}(x_t)} \\
    &+ \sqrt{\frac{2V_{G,T}(v_{1:T})\log(2/\delta)}{T}} +\frac{7\log(1/\delta)}{3T}\,,
\end{align*}
where $V_{G,T}(v_{1:T}) := \frac{1}{T(T-1)} \sum_{t,\tau=1}^T \frac{(\1{v_t\notin N_{G}(x_t)}-\1{v_\tau\notin N_{G}(x_\tau)} )^2}{2}$ is the sample variance.
Since $\hat G$ is consistent, we have $\frac{1}{T}\sum_{t=1}^T \1{v_t\notin N_{\hat G}(x_t)} =0$ and $V_{\hat G,T}(v_{1:T}) =0$.
Then by union bound over all $G\in \cG$, we have that with probability at least $1-\delta$,
\begin{align*}
    \frac{1}{T}\sum_{t=1}^T \Pr_{v\sim \Unif(N_{G^\star}(x_t)) }( v\notin N_{\hat G}(x_t))\leq \frac{7\log(|\cG|/\delta)}{3T}\,.
\end{align*}
\end{proof}

Now we have that for any fixed sampled sequence of $x_{1:T}$, w.p. at least $1-\delta$ over $v_{1:T}$,  $$\hatlgraph (\hat G)\leq\epsilon.$$ We will apply concentration inequality over $x_{1:T}$ to show that $ \lgraph (\hat G)$ is small and then finish the proof of Lemma~\ref{lmm:neighbor-term}.
\begin{align*}
    &\Pr_{x_{1:T},v_{1:T}}(\lgraph (\hat G) >2\epsilon) \\
    \leq & \Pr_{x_{1:T},v_{1:T}}(\lgraph (\hat G) >2\epsilon, \hatlgraph (\hat G)\leq\epsilon) + \Pr_{x_{1:T},v_{1:T}}(\hatlgraph (\hat G)>\epsilon) \\
    \leq& \Pr_{x_{1:T}}(\exists G\in \cG, \lgraph (G) >2\epsilon,\hatlgraph ( G)\leq\epsilon) + \delta \tag{Claim~\ref{cla:empirical-neighbor}}\\
    \leq & 2\delta\,,
\end{align*}
where the last inquality holds by Chernoff bounds when $T\geq \frac{8}{\epsilon} (\log(|\cG|/\delta))$. Hence, combined with Claim~\ref{cla:empirical-neighbor}, we need $T \geq O(\frac{k\log(|\cG|/\delta)}{\epsilon})$ overall. 

\end{proof}

\thmUgPacRelLb*
\begin{proof}
    Consider the input space of one node $o$ and $n(n+1)$ nodes $\{x_{ij}|i=0,\ldots,n, j=1,\ldots,n\}$, which are partitioned into $n$ subsets $X_0 = \{x_{01}, x_{02},...x_{0n}\}, X_1 = \{x_{11}, x_{12},...x_{1n}\},..., X_n$. The hypothesis will label one set in $\{X_1,\ldots,X_n\}$ by positive and the hypothesis class is $H = \{\1{X_i}|i\in [n]\}$. The target function is $\1{X_{i^*}}$. This class is analogous to singletons if we view each group as a composite node. However, since the degree of the manipulation graph is limited to $1$, we split one node into $n$ copies.
    The marginal data distribution put probability mass $1-2\epsilon$ on the irrelevant node $o$ and the remaining $2\epsilon$ uniformly over $X_0$.

    Let the graph class $\cG$ be the set of all graphs which connect $x_{0i}$ to at most one node in $\{ x_{1i},\ldots,x_{ni}\}$ for all $i$. So the cardinality of the graph class $\cG$ is $|\cG| = (n+1)^n$.
    
    When the sample size $T$ is smaller than $\frac{n}{8\epsilon}$, we can sample at most $n/2$ examples from $X_0$ with constant probability by Chernoff bounds. 
    Then there are at least $n/2$ examples in $X_0$ that have not been sampled.

    W.l.o.g., let $x_{01},\ldots,x_{0\frac{n}{2}}$ denote the sampled examples. The graph $G$ does not put any edge on these sampled examples. So all these examples are labeled as negative no matter what $i^*$ is. 
    Then looking at the output $\hat h$ at any unseen example $x_{0j}$ in $X_0$.
    \begin{itemize}
    \item If $\hat h$ predicts all points in $\{x_{0j}, x_{1j},\ldots,x_{nj}\}$ as $0$, we add an edge between $x_{0j}$ and $x_{i^*j}$ to $G$. So $\hat h$ misclassfy $x_{0j}$ under manipulation graph $G$.
    \item If $\hat h$ predicts $x_{0j}$ as positive, then we do not add any edge on $x_{0t}$ in $G$.  So $\hat h$ will classify $x_{0j}$ as $1$ but the target hypothesis will label $x_{0j}$ as $0$.
    \item If $\hat h$ predicts any node $x_{ij}$ with $i\neq i^*\in [n]$ as positive, we add an edge between $x_{0j}$ and $x_{ij}$. So $\hat h$ will classify $x_{0j}$ as $1$ but the target hypothesis will label $x_{0j}$ as $0$.
    \item If $\hat h$ predicts exactly $x_{i^*j}$ in $\{x_{0j}, x_{1j},\ldots,x_{nj}\}$ as positive. Then since we can arbitrarily pick $i^*$ at the beginning, there must exist at least one $i^*$ such that at most this case will not happen.
\end{itemize}
Therefore, $\hat h$ misclassfy every unseen point in $X_0$ under graph $G$ and $\Lstr_{G,\cD}(\hat h)\geq \epsilon$.
\end{proof}

\paragraph{Agnostic PAC Learning} 
Now, we explore the agnostic setting where there may be no perfect graph in $\cG$, no perfect hypothesis $h^\star\in \cH$, or possibly neither. In the following, we first define the loss of the optimal loss of $\cH$ and the optimal loss of $\cG$ and aim to find a predictor with a comparable loss.

\begin{definition}[Optimal loss of $\cH$]
Let $\Delta_\cH$ denote the optimal strategic loss achievable by hypotheses in $\cH$. That is to say,  $$\Delta_\cH := \inf_{h\in \cH} \Lstr_{G^\star,\cD}(h)\,.$$
\end{definition}

\begin{definition}[Optimal loss of $\cG$]\label{def:opt-loss-g}
Let $\Delta_\cG$ denote the graph loss (0/1 loss of neighborhood) of the optimal graph $G^\dagger\in \cG$. 
That is to say,  $$\Delta_\cG := \min_{G^\dagger\in \cG} \lgraph (G)= \min_{G^\dagger\in \cG} \Pr_{(x,y)\sim \cD}(N_{G^\star}(x)\neq N_{G}(x))\,.$$
\end{definition}


We start introducing our algorithm by providing the following lemma, which states that if we are given an approximately good graph, then by minimizing the strategic loss under this approximate graph, we can find an approximately good hypothesis.
\begin{lemma} \label{lmm:given-app-graph}
    Given a graph $G$ with the loss of neighborhood being $\lgraph (G)=\alpha$, for any $h$ being $\epsilon$-approximate optimal under manipulation graph $G$ is the true graph, i.e., $h$ satisfies
    $\Lstr_{G,\cD}(h)\leq \inf_{h^\star\in \cH}\Lstr_{G,\cD}(h^\star) +\epsilon$,
    it must satisfy
    $$\Lstr_{G^\star,\cD}(h) \leq 2\alpha + \Delta_\cH +\epsilon\,.$$
\end{lemma}
\begin{proof}
By definition, we have
    \begin{align*}
        \Lstr_{G^\star,\cD}(h)  =&\PPs{(x,y)\sim \cD}{\bar {h}_{G^\star}(x)\neq y} \\
        = &\PPs{(x,y)\sim \cD}{\bar {h}_{G}(x)\neq y \wedge N_{G^\star}(x) = N_{G}(x)} + \PPs{(x,y)\sim \cD}{\bar {h}_{G^\star}(x)\neq y \wedge N_{G^\star}(x) \neq N_{G}(x)}\\
        \leq &\PPs{(x,y)\sim \cD}{\bar {h}_{G}(x)\neq y} + \PPs{(x,y)\sim \cD}{N_{G^\star}(x) \neq N_{G}(x)}\\
        \leq &\PPs{(x,y)\sim \cD}{\bar {h^\star}_{G}(x)\neq y} +\epsilon + \alpha\tag{$\epsilon$-approximate optimality of $h$}\\
        = &\PPs{(x,y)\sim \cD}{\bar {h^\star}_{G^\star}(x)\neq y \wedge N_{G^\star}(x) = N_{G}(x)} + \PPs{(x,y)\sim \cD}{N_{G^\star}(x) \neq N_{G}(x)} +\epsilon+ \alpha\\
        \leq &\PPs{(x,y)\sim \cD}{\bar {h^\star}_{G^\star}(x)\neq y} +\epsilon+ 2\alpha =\Delta_\cH +\epsilon+ 2\alpha\,.
    \end{align*}
\end{proof}

Then the remaining question is: \textbf{How to find a good approximate graph?} As discussed in the previous section, the graph loss $\lgraph(\cdot)$ is not estimable. Instead, we construct a proxy loss that is not only close to the graph loss but also estimable. 
We consider the following alternative loss function as a proxy:
$$\lproxy(G) = 2\mathbb{E}_{x}[P_{v\sim N_{G^\star}(x)}(v\notin N_G(x))] + \frac{1}{k}\mathbb{E}[|N_G(x)|] - \frac{1}{k}\mathbb{E}[|N_{G^\star}(x)|]\,,$$
where $k$ is the degree of graph $G^\star$. Note that the first two terms are estimable and the third term is a constant. Hence, we can minimize this proxy loss.

\lmmproxyLoss*
\begin{proof}
Let $d_G = \mathbb{E}[|N_G(x)|] - \mathbb{E}[|N_{G^\star}(x)|]$ denote the difference between the expected degree of $G$ and that of $G^\star$.
We have 
\begin{equation}
    d_G = \mathbb{E}_{(x,y)\sim D}[|N_G(x)|] - \mathbb{E}_{(x,y)\sim D}[|N_{G^\star}(x)|] = \mathbb{E}[|N_G(x)\setminus N_{G^\star}(x)|] - \mathbb{E}[|N_{G^\star}(x)\setminus N_{G}(x)|]\,.\label{eq:dg}
\end{equation}
 

Then, we have 
\begin{align*}
    \lproxy(G) =& 2\mathbb{E}_{x}[\frac{|N_{G^\star}(x)\setminus N_{G}(x)|}{|N_{G^\star}(x)|}] +\frac{1}{k} d_G \\
    \geq& 2\cdot \frac{1}{k}\mathbb{E}_{x}[|N_{G^\star}(x)\setminus N_{G}(x)|] +\frac{1}{k} d_G \\
    =& \frac{1}{k} (\mathbb{E}[|N_G(x)\setminus N_{G^\star}(x)| + |N_{G^\star}(x)\setminus N_{G}(x)|])\tag{Applying Eq~\eqref{eq:dg}}\\
    \geq & \frac{1}{k} \lgraph (G)\,.
\end{align*}

On the other hand, we have
\begin{align*}
    \lproxy(G) =& 2\mathbb{E}_{x}[\frac{|N_{G^\star}(x)\setminus N_{G}(x)|}{|N_{G^\star}(x)|}] +\frac{1}{k} d_G \\ 
    \leq & 2\EEs{x}{\1{N_{G}(x) \neq N_{G^\star}(x)}} +\EEs{x}{\1{N_{G}(x) \neq N_{G^\star}(x)}}\\
    =& 3 \lgraph(G)\,.
\end{align*}
\end{proof}

Given a sequence of $S = ((x_1,v_1),\ldots,(x_T,v_T))$, we define the empirical proxy loss over the sequence $S$ as 
$$\hatlproxy (G, S) = \frac{2}{T}\sum_{t=1}^T\1{v_t \notin N_G(x_t)} + \frac{1}{kT}\sum_{t=1}^T |N_G(x_t)|- \frac{1}{k}\EEs{(x,y)\sim D}{|N_{G^\star}(x)|}$$
where the first two terms are the empirical estimates of the first two terms of $\lproxy(G)$ and the last term is not dependent on $G$, which is a constant.

Similar to the realizable setting, by implementing the hypothesis $h_t=\1{x\neq x_t}$, which labels every point as positive except $x_t$, we observe the manipulated feature vector $v_t\sim \Unif(N_{G^\star}(x))$. 

Given two samples, $$S_1 = ((x_1,v_1,y_1),\ldots, (x_{T_1},v_{T_1},y_{T_1}))$$
and $$S_2=((x_1',v_1',y_1'),\ldots, (x_{T_2}',v_{T_2}',y_{T_2}')),$$ we use $S_1$ to learn an approximate good graph and $S_2$ to learn the hypothesis.

Let $\hat G$ be the graph minimizing the empirical proxy loss, i.e., $$\hat G = \argmin_{G\in \cG} \hatlproxy(G, S_1)= \argmin_{G\in \cG} \frac{2}{T}\sum_{t=1}^{T_1}\1{v_t \notin N_G(x_t)} + \frac{1}{kT}\sum_{t=1}^{T_1} |N_G(x_t)|\,.$$
Then let $\hat h$ be the ERM implementation assuming that $\hat G$ is the true graph, i.e.,
$$\hat h =\argmin_{h\in \cH} \frac{1}{T_2}\sum_{t=1}^{T_2} \1{\bar h_{\hat G} (x_t') \neq y_t'}\,.$$
Next, we bound the strategic loss of $\hat h$ using $\epsilon,\Delta_\cG$ and $\Delta_\cH$.
\begin{theorem}\label{thm:ug-pac-agn}
For any hypothesis class $\cH$ with $\vcd(\cH)=d$, the underlying true graph $G^\star$ with maximum out-degree at most $k$, finite graph class $\cG$ in which all graphs have maximum out-degree at most $k$, any data distribution, and any $\epsilon,\delta\in (0,1)$, with probability at least $1-\delta$ over $S_1\sim \cD^{T_1}, S_2\sim \cD^{T_2}$ and $v_{1:T_1}$ where $T_1=O( \frac{k^2\log(|\cG|/\delta)}{\epsilon^2})$ and $T_2= O( \frac{d\log(kd) + \log(1/\delta)}{\epsilon^2})$
, the output $\hat h$ satisfies
\begin{equation*}
        \Lstr_{G^\star,\cD}(\hat h)\leq 6k\Delta_\cG  +\Delta_\cH + \epsilon\,.
\end{equation*}
\end{theorem}
\begin{proof}
    We first prove that $\lgraph(\hat G) \leq 3k \Delta_\cG + 2\epsilon$.
    
    By Hoeffding bounds and union bound, with probability at least $1-\delta/2$ over $S_1$, for all $G\in \cG$,
    \begin{equation}
        |\hatlproxy(G, S_1) -\lproxy(G)| \leq \epsilon_1\,,
    \end{equation}
    when $T_1= O(\frac{\log(|\cG|/\delta)}{\epsilon_1^2})$.
    
    Hence, we have
    \begin{align*}
        \lgraph(\hat G)
        \leq & k \lproxy(\hat G)\tag{Applying Lemma~\ref{lmm:proxy-loss}}\\
        \leq & k\hatlproxy(\hat G, S_1) +k \epsilon_1\\
        \leq & k\hatlproxy(G^\dagger, S_1) + k \epsilon_1\\
        \leq & k\lproxy(G^\dagger) +2k \epsilon_1\\
        \leq & 3k \lgraph(G^\dagger) + 2k \epsilon_1\tag{Applying Lemma~\ref{lmm:proxy-loss}}\\
        = & 3k \Delta_\cG + 2k \epsilon_1\,.\tag{Def~\ref{def:opt-loss-g}}
    \end{align*}

    Then by VC theory and uniform convergence, with probability at least $1-\delta/2$ over $S_2$, for all $h\in \cH$, 
    \begin{equation*}
        |\hatlstr_{\hat G}(h) - \Lstr_{\hat G,\cD}(h)| \leq \epsilon_2\,,
    \end{equation*}
    when $T_2=O( \frac{d\log(kd) + \log(1/\delta)}{\epsilon_2^2})$.
    
    Therefore $\hat h$ is an approximately good implementation if $\hat G$ is the true graph, i.e.,
    \begin{equation*}
        \Lstr_{\hat G,\cD}(\hat h)\leq \min_{h\in \cH}\Lstr_{\hat G,\cD}(h) + 2\epsilon_2\,.
    \end{equation*}
    Then by applying Lemma~\ref{lmm:given-app-graph}, we get
    \begin{equation*}
        \Lstr_{G^\star,\cD}(\hat h)\leq 6k\Delta_\cG  +\Delta_\cH +4k\epsilon_1+ 2\epsilon_2\,. 
    \end{equation*}  
    We complete the proof by plugging in $\epsilon_1 = \frac{\epsilon}{8k}$ and $\epsilon_2 = \frac{\epsilon}{4}$.
\end{proof}


\subsection{Application of the Graph Learning Algorithms in Multi-Label Learning}
Our graph learning algorithms have the potential to be of independent interest in various other learning problems, including multi-label learning. To illustrate, let us consider scenarios like the recommender system, where we aim to recommend movies to users. In such cases, each user typically has multiple favorite movies, and our objective is to learn this set of favorite movies.
Similarly, in the context of object detection in computer vision, every image often contains multiple objects. Here, our learning goal is to accurately identify and output the set of all objects present in a given image.

This multi-label learning problem can be effectively modeled as a bipartite graph denoted as $G^\star = (\cX, \cY, \cE^\star)$, where $\cX$ represents the input space (in the context of the recommender system, it represents users), $\cY$ is the label space (in this case, it signifies movies), and $\cE^\star$ is a set of edges. In this graph, the presence of an edge $(x, y) \in \cE^\star$ implies that label $y$ is associated with input $x$ (e.g., user $x$ liking movie $y$). Our primary goal here is to learn this graph, i.e., the edges $\cE^\star$. More formally, given a marginal data distribution $\cD_\cX$, our goal is to find a graph $\hat G$ with minimum neighborhood prediction loss $\lgraph (G)= \Pr_{x\sim \cD_\cX}(N_{G}(x)\neq N_{G^\star}(x))$. Suppose we are given a graph class $\cG$, then our goal is to find a graph $G^\dagger$ such that 
\[G^\dagger = \argmin_{G\in \cG}\lgraph (G).\]

However, in real-world scenarios, the recommender system cannot sample a user along with her favorite movie set.
Instead, at each time $t$, the recommender recommends a set of movies $h_t$ to the user, and the user randomly clicks on one of the movies in $h_t$ that she likes (i.e., $v_t\in N_{G^\star}(x_t)\cap h_t$). Here we abuse the notation a bit by letting $h_t$ represent the set of positive points labeled by this hypothesis.
This setting perfectly fits into our unknown graph setting.

Therefore, in the realizable setting where $\lgraph (G^\dagger) =0$, we can find $\hat G$ such that $\hat G$ is satisfying $\lgraph (\hat G) \leq \epsilon$ given 
$O(\frac{8k\log(|\cG|/\delta)}{\epsilon})$ examples according to Lemma~\ref{lmm:neighbor-term} .
In the agnostic setting, we can find $\hat G$ satisfying $\lgraph (\hat G) \leq 3k\lgraph (G^\dagger) + \epsilon$ given $O(\frac{k^2 \log(|\cG|/\delta)}{\epsilon^2})$ examples according to Theorem~\ref{thm:ug-pac-agn}.

\subsection{Online Learning}
In the unknown graph setting, we do not observe $N_{G^\star}(x_t)$ or $N_{G^\star}(v_t)$ anymore.
We then design an algorithm by running an instance of Algorithm~\ref{alg:reduction2online-pmf} over the majority vote of the graphs.
\begin{algorithm}[t]\caption{Online Algorithm in the Unknown Graph Setting}\label{alg:ug-online}
    \begin{algorithmic}[1]
        \STATE initialize an instance $\cA=$ Red2Online-PMF (SOA, $2k$) 
        \STATE let $\cG_1 = \cG$
        \FOR{$t=1,2,\ldots$}
        \STATE \underline{Prediction}: after observing $x_t$, for every node $x$
        \IF{$(x_t,x)$ are connected in at most half of the graphs in $\cG_t$}
        \STATE $h_t(x) = 1$\label{alg-line:type1}
        \ELSE
        \STATE let the prediction at $x$ follow $\cA$, i.e., $h_t(x) = \cA(x)$ \label{alg-line:type2}
        \ENDIF
        \STATE \underline{Update}:  when we make a mistake at the observed node $v_t$:
        \IF{$(x_t,v_t)$ are connected in at most half of the graphs in $\cG_t$}
        \STATE update $\cG_{t+1} = \{G\in \cG_t| (x_t,v_t) \text{ are connected in }G\}$ to be the set of consistent graphs
        \ELSE
        \STATE feed $\cA$ with $(v_t, \tilde N(v_t), \hat y_t, y_t)$ with
       $\tilde N(v_t) = \{x| \abs{\{G\in \cG_t|(v_t,x) \text{ are connected in } G\}} \geq |\cG_t|/2\}$, which is the set of vertices which are an out-neighbor of $v_t$ in more than half of the graphs in $\cG_t$
        \STATE $\cG_{t+1} = \cG_t$
        \ENDIF
        \ENDFOR
    \end{algorithmic}
\end{algorithm}

\ugOnline*
\begin{proof}
Let $W_t$ denote the total weight of experts at the beginning of round $t$ in the algorithm instance $\cA$. 
When we make a mistake at round $t$, there are two cases.
\begin{itemize}
    \item Type 1: $(x_t,v_t)$ are connected in at most half of the graphs in $\cG_t$. In this case, we can remove half of the graphs in $\cG_t$, i.e., $\abs{\cG_{t+1}}\leq \frac{\abs{\cG_t}}{2}$. The total number of mistakes in this case will be $N_1\leq \log(\abs{\cG})$.

    \item Type 2: $(x_t,v_t)$ are connected in more than half of the graphs in $\cG_t$. In this case, our prediction rule applies $\cA$ to predict at $v_t$. Then there are two cases (as we did in the proof of Theorem~\ref{thm:fu-online})
    \begin{itemize}
        \item False positive: meaning that we predicted the neighbor $v_t$ of $x_t$ by $1$ but the correct prediction is $0$. This means that the neighbor $v_t$ we saw should also be labeled as $0$ by the target implementation $h^\star$. In this case, since Algorithm~\ref{alg:reduction2online-pmf} does not use the neighborhood to update, and it does not matter what $\tilde N_{G^\star}(v_t)$ is.
    \item False negative: meaning we predicted $x_t$ with $0$ but the correct prediction is $1$. In this case, $v_t=x_t$, the entire neighborhood $N_{G^\star}(x_t)$ is labeled as $0$ by $h_t$, and $h^\star$ labeled some point in $N_{G^\star}(x_t)$ by $1$.
    Since only nodes in $\tilde N(v_t)$ are labeled as $0$ by our algorithm (as stated in line~\ref{alg-line:type1}, all nodes not in $\tilde N(v_t)$ are labeled as $1$ by $h_t$),  the true neighborhood $N_{G^\star}(v_t)$ must be a subset of $\tilde N(v_t)$. Since all graphs in $\cG$ has maximum out-degree at most $k$, we have $|\tilde N(v_t)|\leq \frac{k\cdot |\cG_t|}{|\cG_t|/2} = 2k$.  
    \end{itemize}
    Then, by repeating the same analysis of Theorem~\ref{thm:fu-online} (since the analysis only relies on the fact that the observed $N_{G^\star}(v_t)$ satisfy $|N_{G^\star}(v_t)|\leq k$), we will make at most $N_2\leq O(k\log k \cdot \ld(\cH))$ type 2 mistakes.

\end{itemize}
Therefore, we will make at most $N_1+N_2 \leq  O(\log\abs{\cG}+k\log k \cdot \ld(\cH))$ mistakes.
\end{proof}

\ugOnlineLb*
\begin{proof}
The example is very similar to the one in the proof of Theorem~\ref{thm:ug-pac-rel-lb}.
    Consider the input space of $n(n+1)$ nodes, which are partitioned into $n+1$ subsets $X_0 =\{x_{01}, x_{02},...x_{0n}\}, X_1 = \{x_{11}, x_{12},...x_{1n}\},..., X_n$. The hypothesis will label one set in $\{X_1,\ldots,X_n\}$ by positive and the hypothesis class is $H = \{\1{X_i}|i\in [n]\}$. The target function is $\1{X_{i^*}}$. This class is analogous to singletons if we view each group as a composite node. However, since the degree of the manipulation graph is limited to $1$, we split one node into $n$ copies.
    
    The agent will always pick an agent in $X_0$ and the true label is positive only when it is connected to $X_{i^*}$. To ensure that the degree of every node is at most $1$, only one node in each set will only be used in round $t$, i.e., $\{x_{0t}, x_{1t},\ldots,x_{nt}\}$.

At round $t$:
\begin{itemize}
    \item If the learner predicts all points in $\{x_{0t}, x_{1t},\ldots,x_{nt}\}$ as $0$, the adversary picks the agent $x_{0t}$ and adds an edge between $x_{0t}$ and $x_{i^*t}$. The agent does not move, the learner predicts $0$, and the true label is $1$ no matter what $i^*$ is. We do not learn anything about the target function.
    \item If the learner predicts $x_{0t}$ as positive, then the adversary picks the agent  $x_{0t}$  and does not add any edge on  $x_{0t}$. The agent does not move, the learner predicts $1$, and the true label is $0$. We learn nothing.
    \item If the learner predicts any node $x_{it}$ with $i\neq i^*\in [n]$ as positive, the adversary picks the agent  $x_{it}$  and adds no edge  on $x_{0t}$. The agent will stay at $x_{it}$, the learner predicts $1$, and the true label is $0$. But we can only eliminate one hypothesis $\1{X_i}$.
\end{itemize}
Hence there must exist an $i^*$ the algorithm will make at least $n-1$ mistakes.
Since all possible graphs will add at most one edge between $x_{0t}$ and $\{x_{1t},\ldots,x_{nt}\}$, the graph class $\cG$ has at most $(n+1)^n$ graphs.
\end{proof}
    
Next, we restate Proposition~\ref{prop:online-v-lb}. 
\uglessinfo*
\begin{proof}
    Consider $n+2$ nodes, $A,B$ and $C_1, C_2,\ldots, C_n$. 
    The graph class $\cG$ has $n$ graphs in the form of $A-B-C_i$ for $i\in [n]$. 
    The hypothesis class is singletons over $C_1,...,C_n$. 
    So we have $n$ graphs with degree $2$ and $n$ hypotheses. 
Suppose that the true graph is $A-B-C_{i^*}$ and the target function is $\1{C_{i^*}}$. Hence only the agents $B$ and $C_{i^*}$ are positive.

Then at each round:
\begin{itemize}
    \item If the learner is all-negative, the adversary picks agent $(B,1)$. The learner makes a mistake at 
    $x_t = v_t =B$, and learn nothing.
    \item If the learner implements $h_t$ satisfying $h_t(B)= 1$ (or $h_t(A)= 1$), then the adversary picks agent $(A,-1)$. The learner makes a mistake at $v_t = B$ (or $v_t = A$) and learns nothing.
    \item If the learner implements $h_t$ predicting any $C_i$ with $i \neq i^*$ by positive, then the adversary picks agent $(C_i, -1)$. The learner makes a mistake at $v_t = x_t = C_i$ and can only eliminate one hypothesis $\1{C_{i}}$ (and graph).
\end{itemize}
Hence, there must exist an $i^*$ such that the learner makes at least $n-1$ mistakes.
\end{proof}

\section{From Realizable Online Learning to Agnostic Online Learning}\label{app:online-agn}
Since we cannot achieve sublinear regret even in the standard online learning by deterministic algorithms, we have to consider randomized algorithms in agnostic online learning as well.
Following the model of ``randomized algorithms'' (instead of ``fractional classifiers'') by \cite{ahmadi2023fundamental},
for any randomized learner $p$ (a distribution over hypotheses), the strategic behavior depends on the realization of the learner. Then the strategic loss of $p$ is 
\begin{equation*}
    \lstr_G (p,(x,y)) = \EEs{h \sim p}{ \lstr_G(h,(x,y))}\,.
\end{equation*}

When we know $x_t$ and $N_{G^\star}(x_t)$, we actually have full information feedback (i.e., we can compute the prediction of every hypothesis ourselves). 
Hence, we can apply the realizable-to-agnostic technique by constructing a cover of $\cH$ with $T^M$ experts, where $M$ is the mistake bound in the realizable setting, and then running multiplicative weights over these experts.

For the construction of the experts, we apply the same method by~\cite{ben2009agnostic}. The only minor difference is that the learner in our setting needs to produce a hypothesis at each round to induce some labeling instead of deciding a labeling directly. Given the hypothesis $\tilde h_t$ generated by a realizable algorithm $\cA_{\text{REL}}$, to flip the prediction induced by $\tilde h_t$, we can change the entire neighborhood of $x_t$ to negative if $\tilde h_t$'s prediction is positive, and predict $x_t$ by positive if $\tilde h_t$'s prediction is negative. If we do not know $(x_t,N_{G^\star}(x_t))$, then we do not even know how to flip the prediction.

\begin{algorithm}[t]\caption{Construction of expert $\text{Exp}(i_1,\ldots,i_L)$ }\label{alg:expert-cover}
    \begin{algorithmic}[1]
    \STATE \textbf{Input: } an algorithm in the realizable setting $\cA_{\text{REL}}$ and indices $i_1,\ldots,i_L$
    \FOR{$t=1,\ldots, T$}{
    \STATE receive $x_t$
    \STATE let $\tilde h_t$ denote the hypothesis produced by $\cA_{\text{REL}}$, i.e.,\\ $\tilde h_t = \cA_{\text{REL}}((x_1,N_{G^\star}(x_1),\tilde y_1), \ldots,(x_{t-1},N_{G^\star}(x_{t-1}),\tilde y_{t-1}))$
    \IF{$t\in \{i_1,\ldots,i_L\}$}{
    \IF{$\tilde h_t$ labels the whole $N_{G^\star}(x_t)$ by negative}{
    \STATE define $h_t = \tilde h_t$ except $h_t(x_t) = 1$ and implement $h_t$
    \STATE let $\tilde y_t = 1$
    }
    \ELSE
    \STATE define $h_t$ s.t. $h_t$ labels the entire neighborhood $N_{G^\star}(x_t)$ by negative and implement $h_t$
    \STATE let $\tilde y_t = 0$
    \ENDIF
    }
    \ELSE   
    \STATE implement $\tilde h_t$ and let $\tilde y_t = \bar{\tilde h_t}_{G^\star}(x_t)$  
    \ENDIF
    }
    \ENDFOR
    \end{algorithmic}
\end{algorithm}
\begin{lemma}
    Let $\cA_{\text{REL}}$ be an algorithm that makes at most $M$ mistakes in the realizable setting. Let $x_1,\ldots,x_T$ be any sequence of instances. For any $h\in \cH$, there exists $L\leq M$ and indices $i_1,\ldots,i_L$, such that running $\text{Exp}(i_1,\ldots,i_L)$ (Algorithm~\ref{alg:expert-cover}) on the sequence $((x_1,N_{G^\star}(x_1)), \ldots, (x_T,N_{G^\star}(x_T)))$ generate the same prediction as implementing $h$ on each round, i.e., $\bar h_{tG^\star}(x_t) = \bar h_{G^\star}(x_t)$ for all $t$.
\end{lemma}
\begin{proof}
    Fix $h\in \cH$ and the sequence $((x_1,N_{G^\star}(x_1)), \ldots, (x_T,N_{G^\star}(x_T)))$. Then consider running $\cA_{\text{REL}}$ over $((x_1,N_{G^\star}(x_1), \bar h(x_1)), \ldots, (x_T,N_{G^\star}(x_T),\bar h(x_T)))$ and $\cA_{\text{REL}}$ makes $L\leq M$ mistakes. Define $\{i_1,\ldots,i_L\}$ to be the set of rounds $\cA_{\text{REL}}$ makes mistakes. Since the prediction of $\cA_{\text{REL}}$ is only different from $\bar h_{G^\star}$ at rounds in $\{i_1,\ldots,i_L\}$, the predictions when implementing $\text{Exp}(i_1,\ldots,i_L)$ are the same as implementing $h$.
\end{proof}
We finish with a result that holds directly by following~\cite{ben2009agnostic}.
\begin{proposition}
    \label{res:agn-online}
By running multiplicative weights over the experts, we can achieve regret
\begin{align*}
    \regret(T)\leq O(\sqrt{k\log k\cdot \ld(\cH) T\log(T)})\,.
\end{align*} 
\end{proposition}
It is unclear to us how to design algorithms in post-manipulation feedback and unknown graph settings.


\chapter{Incentives in Single-Round Federated Learning}\label{app:incentives-single}

\section{Real-Valued Strategies} \label{app:integral}

In the examples of random coverage and general PAC learning, it is common to consider integral values of $\theta_i$. For a real-valued $\theta_i$, we consider one natural interpretation: randomized rounding over $\floor{\theta_i}$ and $\ceil{\theta_i}$. More specifically, let agent $i$ randomly draw an integral value $m_i\sim \sigma(\theta_i)$, where $\sigma(\theta_i)=\floor{\theta_i} + \Ber(\theta_i-\floor{\theta_i})$, and then uses $m_i$ as her strategy. Then the utility function is defined by taking expectation over $\bfm = (m_1,\ldots,m_k)$. That is,
\[
u_i(\btheta) = \EEs{\bfm}{1 - \frac 12 \sum_{x\in \cX} q_{ix} \prod_{j=1}^k \left(1-q_{jx} \right)^{m_j}}\,.
\]
Similarly, we define the utility function in general PAC learning as
\[
u_i(\btheta) = 1 -\EEs{\bfm}{\EEs{\{S_j\sim \cD_j^{m_j}\}_{j\in[k]}}{\err_{\cD_i}(h_{S})}}\,.
\]
Note that these definitions work for integral-valued $\theta_i$ as well.

\section{Calculation of Well-behaved Property}\label{app:wellbehave}
\paragraph{Linear Utilities.} 
The linear utilities are well-behaved over any $\bigtimes_{i=1}^k [0,C_i]\subseteq \Theta$. Agent $i$'s utility increases at a constant rate $\partial \theta_i(\btheta)/\partial \theta_i=W_{ii} = 1$ when the agent increases its strategy unilaterally and increases at rate $\partial \theta_i(\btheta)/\partial \theta_j=W_{ij}\leq 1$ when agent $j$ increases its strategy unilaterally.

\paragraph{Random Coverage.}
For any $\bigtimes_{i=1}^k [0,C_i+1]\subseteq \Theta$, if $u_i(C_i+1,\bC_{-1})-u_i(\bC)$ is bounded away from $0$ for all $i$, then the utilities are well-behaved over $\bigtimes_{i=1}^k [0,C_i]$, where $\bC =(C_1,\ldots,C_k)$. At a high level, the smallest impact that an additional sample by agent $i$ has on $u_i$ is when $\btheta \rightarrow \bC$. This impact is at least $ u_i(C_i + 1, \bC_{-i})-u_i(\bC) > 0$.
On the other hand, 
$\partial u_i(\btheta)/\partial \theta_j$ is bounded above, because the marginal impact of any one sample on $u_i$ is largest when no agent has yet taken a sample.

First, by direct calculation, we have that for any non-integral $\theta_j$,
\begin{align*}
    \frac{\partial u_i(\btheta)}{\partial \theta_j} 
    =& -\frac 12 \frac{\partial \sum_{x\in\cX} q_{ix} \prod_{l=1}^k \EE{(1-q_{lx})^{m_l}}}{\partial \theta_j}\\
    =& -\frac 12 \frac{\partial \sum_{x\in\cX} q_{ix} \prod_{l\neq j} \EE{(1-q_{lx})^{m_l}}\left((\theta_j -\floor{\theta_j})(1-q_{jx})^{\floor{\theta_j}+1} +(1+\floor{\theta_j}-\theta_j)(1-q_{jx})^{\floor{\theta_j}}\right)}{\partial \theta_j}\\
    =& -\frac 12 \sum_{x\in\cX} q_{ix} \prod_{l\neq j} \EE{(1-q_{lx})^{m_l}}\left((1-q_{jx})^{\floor{\theta_j}+1}-(1-q_{jx})^{\floor{\theta_j}}\right)\\
    =& u_i({\floor{\theta_j}+1},\btheta_{-j}) - u_i({\floor{\theta_j}},\btheta_{-j}) \,.
\end{align*}
For integral-value $\theta_j$, when we increase $\theta_j$ by a small amount $\epsilon\in (0,1)$, $\alpha=\floor{\theta_j+\epsilon} = \floor{\theta_j}$ does not change. Then we have
\begin{align*}
    \frac{\partial_+ u_i(\btheta)}{\partial \theta_j}=& -\frac 12 \frac{\partial_+ \sum_{x\in\cX} q_{ix} \prod_{l\neq j} \EE{(1-q_{lx})^{m_l}}\left((\theta_j -\alpha)(1-q_{jx})^{\alpha+1} +(1+\alpha-\theta_j)(1-q_{jx})^{\alpha}\right)}{\partial \theta_j}\\ 
    =& u_i({\alpha+1},\btheta_{-j}) - u_i(\alpha,\btheta_{-j})\\
    =& u_i({{\theta_j}+1},\btheta_{-j}) - u_i(\btheta) \,.
\end{align*}
When we decrease $\theta_j$ by $\epsilon$,  $\alpha=\floor{\theta_j-\epsilon} = \floor{\theta_j-1}$. Then for all $x\in [\theta_j-1,\theta_j]$, we can represent 
\[u_i(x,\btheta_{-1}) =1-\frac 12 {\sum_{x\in\cX} q_{ix} \prod_{l\neq j} \EE{(1-q_{lx})^{m_l}}\left((x -\alpha)(1-q_{jx})^{\alpha+1} +(1+\alpha-x)(1-q_{jx})^{\alpha}\right)}\,.\]
Thus we have
\begin{align*}
    \frac{\partial_- u_i(\btheta)}{\partial \theta_j}=& -\frac 12 \frac{\partial_- \sum_{x\in\cX} q_{ix} \prod_{l\neq j} \EE{(1-q_{lx})^{m_l}}\left((\theta_j -\alpha)(1-q_{jx})^{\alpha+1} +(1+\alpha-\theta_j)(1-q_{jx})^{\alpha}\right)}{\partial \theta_j}\\ 
    =& u_i({\alpha+1},\btheta_{-j}) - u_i(\alpha,\btheta_{-j})\\
    =& u_i(\btheta)-u_i({{\theta_j}-1},\btheta_{-j})\,.
\end{align*}
Then we argue that for any $\btheta\in \bigtimes_{i=1}^k [0,C_i+1]$, any $t\in \NN\cap [0,C_i+1]$, $u_i(t+1,\btheta_{-j})-u_i(t,\btheta_{-j})= \frac 12 {\sum_{x\in\cX} q_{ix} \prod_{l\neq j} \EE{(1-q_{lx})^{m_l}}q_{jx}(1-q_{jx})^{t}}$ is non-increasing with respect to $t$ and with respect to $\theta_l$ for any $l\neq j$. 

Combining the computing results on sub-gradients and the monotonicity of $u_i(t+1,\btheta_{-j})-u_i(t,\btheta_{-j})$, we know that
\[\frac{\partial u_i(\btheta)}{\partial \theta_i} \geq u_i(C_i+1,\btheta_{-1}) - u_i(C_i,\btheta_{-1})\geq u_i(C_{i}+1,\bC_{-1})-u_i(\bC)\,,\]
and
\[\frac{\partial u_i(\btheta)}{\partial \theta_j}\leq u_i(1,\btheta_{-j}) - u_i(0,\btheta_{-j})\leq u_i(1,\bZero_{-j}) - u_i(0,\bZero_{-j})\leq \frac 12 \sum_{x\in\cX} q_{ix}q_{jx}\leq \frac 12\,.\]

\paragraph{General PAC Learning.} 

In the previous two examples, the utilities are well-behaved over any bounded convex set. However, this might not be true in the general PAC learning case. For example, recall the example in the proof of Theorem~\ref{thm:eqintexist} and let us extend the strategy space $\Theta$ from $\{0,1\}^3$
to $[0,1]^3$ by the randomized rounding method as aforementioned, i.e., 
\[u_i(\btheta) = \frac 12(1+\theta_i+\theta_{i\ominus 1} -\theta_i\theta_{i\ominus 1})\,.\]
Then the utility function is ill-behaved over $[0,1]^3$ since ${\partial u_i(\btheta)}/{\partial \theta_i}  = 0$ when $\theta_{i\ominus 1}=1$. However, it is easy to check that for any $C\in [0,1)$, the utility function is well-behaved over $[0,C]^3$.

\section{Proof of Lemma~\ref{lmm:fixedpt}}\label{app:fixedpt}

\rstfixedpt*

\begin{proof}
The celebrated Brouwer fixed-point theorem states that any continuous function on a compact and convex subset of $\R^k$ has a fixed point. First note that $\bFlocal$ is a well-defined map from $\bigtimes_{i=1}^k [0, \vartheta_i]$ to $\bigtimes_{i=1}^k [0, \vartheta_i]$, which is a convex and compact subset of $\R^k$. All that is left to show is that $\bFlocal$ is a continuous function over $\bigtimes_{i=1}^k [0, \vartheta_i]$.

At a high level, $f$ is continuous because in well-behaved utility functions a small change in other agents' contributions affect the utility of agent $i$ only by a small amount, so a small adjustment to agent $i$'s contribution will be sufficient to meet his constraint. More formally, we show that for any
$\bdelta\in \R^k$ with $\norm{\bdelta}_1\leq 1$, $\lim_{\epsilon \rightarrow 0} \lvert f_i(\btheta)  - f_i(\btheta +\epsilon\bdelta) \rvert =0$.
Define $\btheta' = \btheta +\epsilon\bdelta$, $x = f_i(\btheta)$, and $x'= f_i(\btheta')$.
For every $i$, we have
    \begin{align*}
        u_i\left( x'+\frac{c^i_1\epsilon}{c^i_2},\btheta_{-i}\right) 
        \geq u_i\left( x',\btheta_{-i}\right) + c^i_1\epsilon
        \geq u_i\left( x',\btheta_{-i}\right) + c^i_1\epsilon \|\bdelta_{-i} \|_1
        \geq u_i(x',\btheta_{-i}+\epsilon \bdelta_{-i})
        \geq \mu_i\,,
    \end{align*}
    where the first and third transitions are by the definition of well-behaved functions, and the last transition is by the definition of $\btheta'$ and $x'$.
    This shows that $x\leq x'+\frac{c^i_1\epsilon}{c^i_2}$.
Similarly,
    \begin{align*}
        u_i\left( x+\frac{c^i_1\epsilon}{c^i_2},(\btheta+\epsilon\bdelta)_{-i} \right)     
        \geq u_i(x,(\btheta+\epsilon\bdelta)_{-i})+c^i_1\epsilon
        \geq u_i(x,(\btheta+\epsilon\bdelta)_{-i})+c^i_1\epsilon\norm{-\bdelta_{-i}}_1
        \geq u_i(x,\btheta_{-i})
        \geq \mu_i\,,
    \end{align*}
    which indicates that $x+\frac{c^i_1\epsilon}{c^i_2}\geq x'$. Hence, we have $\abs{x-x'}\leq \frac{c^i_1\epsilon}{c^i_2}$. Therefore, $\bFlocal$ is continuous over $\bigtimes_{i=1}^k [0, \vartheta_i]$.

The proof follows by applying the Brouwer Fixed-Point Theorem.
\end{proof}
\section{More General Construction for Theorem~\ref{thm:eqintexist}}\label{app:general-consruction}

We extend the simple example in Section~\ref{sec:proofofeqexist} into a more general one. 

Consider the domain $\cX = \{0,\ldots,6d-1\}$ for any $d>1$ and the label space $\cY = \{0, 1\}$. We consider agents $\{0,1,2\}$ with distributions $\cD_0,\cD_1,\cD_2$ over $\cX\times \cY$. Similar to the example in Section~\ref{sec:proofofeqexist}, we give a probabilistic construction for  $\cD_0,\cD_1,\cD_2$. Take independent random variables $\bZ_0, \bZ_1, \bZ_2$ that are each uniform over $\{0,1\}^d$. For each $i\in\{0,1,2\}$, distribution $\cD_i$ is a uniform distribution over instance-label pairs $\{((2i+ z_{i,j})d + j, z_{i\ominus 1,j})\}_{j=0}^{d-1}$.
In other words, the marginal distribution of $\cD_i$ is a uniform distribution over $\cX_i=\{x_1,\ldots,x_d\}$ where $x_j$ is equally likely to be $2id + j$ or $(2i+1)d+j$ and independent of other $x_l$ for $l\neq j$. Moreover, the labels of points in distribution $\cD_{i\oplus 1}$ are decided according to the marginal distribution of $\cD_{i}$: if the support of the marginal distribution of $\cD_{i}$ contains $2id+j$, then the points $2(i\oplus 1)d+j$ and $(2(i\oplus 1) +1)d+j$ are both labeled $0$, and if the support of the marginal distribution of $\cD_{i}$ contains $(2i+1)d+j$, then the points $2(i\oplus 1)d+j$ and $(2(i\oplus 1) +1)d+j$ are both labeled $1$.

Consider the optimal classifier conditioned on the event where agent $i$ takes samples $\{((2i + z_{i,j})d+j, z_{i\ominus 1,j})\}_{j\in J_i}$ from $\cD_i$ for all $i$. This reveals $z_{i,j}$ and $z_{i\ominus 1,j}$ for all $j\in J_i$. 
Therefore, the optimal classifier conditioned on this event classifies $(2i + z_{i,j})d+j$ for each $j\in J_i\cup J_{i\ominus 1}$ correctly and misclassifies $(2i + z_{i,j})d+j$ for each $j\notin J_i\cup J_{i\ominus 1}$ with probability $1/2$.

Now we formally define the strategy space and the utility functions that corresponding to this setting. Let $\Theta = \NN^3$ to be the set of strategies in which each agent can take any integral number of samples. Let $u_i(\btheta)$ be the expected accuracy of the optimal classifier given the samples taken at random under $\btheta$. As a consequence of the above analysis,
\begin{align*}
    u_i(\btheta) = 1 -\frac{1}{2d} \sum_{j=0}^{d-1}\left(1-\frac{1}{d}\right)^{\theta_i+\theta_{i\ominus 1}} = 1- \frac{1}{2}\left(1-\frac{1}{d}\right)^{\theta_i+\theta_{i\ominus 1}}\,.
\end{align*}
Then let $\bmu = \mu \bOne$ for any $\mu\in (1/2,1)$ such that $m(\mu):=\ceil{\frac{\log(2(1-\mu))}{\log(1-1/d)}}$ is an odd number. It is easy to find such a $\mu$: arbitrarily pick a $\mu'\in (1/2,1)$; if $m(\mu')$ is odd, let $\mu=\mu'$; otherwise let $\mu=(1-1/d)\mu'+1/d$ such that $m(\mu)=m(\mu')+1$.

Agent $i$'s constraint is satisfied when $\theta_i+\theta_{i\ominus 1}\geq m(\mu)$ and is not satisfied when $\theta_i+\theta_{i\ominus 1}\leq m(\mu)-1$. If $\theta_i+\theta_{i\ominus 1}\geq m(\mu)+1$, agent $i$ can unilaterally decrease her strategy by $1$ and still meet her constraint. Therefore, we have
\[\theta_i +\theta_{i\ominus 1}=m(\mu),\forall i=0,1,2\,.\]
This results in $\theta_0=\theta_1=\theta_2$, which is impossible as $m(\mu)$ is odd and $\theta_i$ is integral for all $i$. Hence, no stable equilibrium over $\Theta = \NN^3$ exists.

\section{Proof of Theorem~\ref{thm:flowergap}}\label{app:flowergap}

\rstflowergap*
\begin{proof}
Consider a family of sets each of size $b = k-1$ demonstrated in Figure~\ref{fig:flower}, where there is one \emph{core agent} that owns $b$ central points and $k-1$ \emph{petal agents} whose sets intersect with that of the core agent.
More formally, let agent $0$ be the core agent whose distribution is uniform over the points $\cX_0 = \{1, \dots, b\}$.
Partition $\cX_0 = \{1, \dots, b\}$ to $\sqrt{b}$ equally sized groups of $\sqrt{b}$ instances $\cX^{1}_0, \dots, \cX^{\sqrt{b}}_0$.
Similarly, partition the $b = k-1$ agents to $\sqrt{b}$ equally sized groups of $\sqrt{b}$ agents $I_{1}, \dots, I_{\sqrt{b}}$.
Each $i \in I_j$ has uniform distribution over the set $\cX_i = \cX^j_0\cup \mathcal{O}_i$, where $\mathcal{O}_i$ is a set of $b-\sqrt{b}$ points that are unique to $i$. The strategy space is $\Theta=\R_+^k$.
\begin{figure}[ht]
    \centering
    \includegraphics[width=0.6\textwidth]{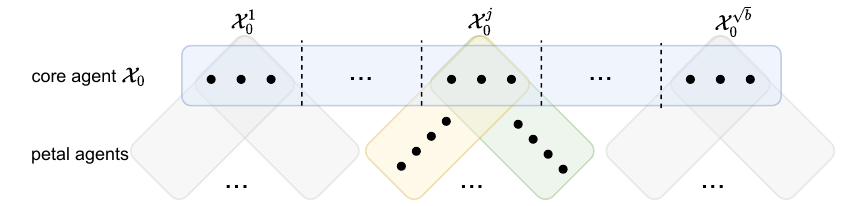}
    \caption{The illustration of the core agent and the petal agents}
    \label{fig:flower}
\end{figure}
Then we consider two learning settings: a) random coverage example and b) linear utility example.

\paragraph{Random Coverage.} Let $m_i\sim \sigma(\theta_i)$ denote the realized integral strategy of agent $i$ for all $i$. For any $I_j$, let $M_j = \sum_{i\in I_j} m_i$ be the total number of samples taken by agents in $I_j$. Then for $i\in I_j$, 
\[
u_i(\btheta)=  1  - \frac{1}{2} \EEs{\bfm}{\frac{1}{\sqrt{b}}\left(1-\frac 1b \right)^{m_0 + M_j} + \left(1-\frac{1}{\sqrt{b}} \right) \left( 1 - \frac 1b \right)^{m_i}}%
\]
and
\[
u_0(\btheta)= 1-\frac{1}{2\sqrt{b}} \EEs{\bfm}{\sum_{j=1}^{\sqrt{b}} \left(1-\frac 1 b\right)^{m_0 + M_j}}\,.
\]
Let $\mu_i=\frac 12 + \frac{1}{2b}$ for all agent $i$.
Note that our choice of $\mu_i$ and distributions implies that the constraint of agent $i$ is met when in expectation at least \emph{one} of the instances in their support is observed by some agent. We use this fact to describe the high level properties of each of the solution concepts.

\underline{The optimal collaborative solution:} Consider the strategy in which the core agent takes 
$O(\sqrt{k})$ samples and all other agents take $0$ samples. This is a feasible solution, because in expectation each $I_j$ receives one of these samples. Therefore, the number of samples in the optimal collaborative solution is at most $O\big( \sqrt{k}\big)$. Specifically, consider the solution in which the core takes $\theta_0 = \ceil{\frac{\ln(1-1/\sqrt{b})}{\ln(1-1/b)}}$ samples and all other agents take $0$ samples.  Let $\btheta^\optlocal$ denote the socially optimal solution. By direct calculation, it is not hard to check that this is a feasible solution and that $\bOne^\top \btheta^\optlocal \leq \ceil{\frac{\ln(1-1/\sqrt{b})}{\ln(1-1/b)}}=O(\sqrt{k})$. 

\underline{The Optimal envy-free solution:} By the symmetry of the utility functions for all $i\in I_j$ and for all $j\in \{1, \dots, \sqrt{b}\}$, any envy-free solution must satisfy $\theta_i=\theta$ for some $\theta$ and all $i\in [b]$. This is not hard to check. First, for two petal agents in the same group, i.e., $i,l\in I_j$, and any feasible solution with $\theta_i>\theta_l$, then
\[u_i(\btheta^{(i,l)}) = u_l(\btheta)\geq \mu\,,\]
which indicates that agent $i$ envies agent $l$. Therefore, for any envy-free solution $\theta_i=\theta_l$ for any $i,l\in I_j$. Then for any feasible solution in which any two agents in the same group have the same number of samples, if $\theta_i>\theta_l$ for any $i\in I_j$ and any $l\in I_p$ with $j\neq p$,
\[u_i(\btheta^{(i,l)})\geq u_l(\btheta) \geq \mu\,,\]
which indicates that agent $i$ envies agent $l$.

Furthermore, in any envy-free feasible solution the $0$-th agent's number of sample can be no larger than any other agent. If $\theta_0>\theta$, considering $m_0\sim \sigma(\theta_0)$ and $m\sim \sigma(\theta)$, we have
\[u_0(\btheta^{(0,i)}) = 1-\frac{1}{2}\EEs{\bfm}{\left(1-\frac 1b\right)^{m_0+M_j} +\sum_{p\in [\sqrt{b}]:p\neq j} (1-\frac 1b)^{m_i+M_p}}\geq u_i(\btheta)\geq \mu\,. \]

Let $\btheta^\ef$ represent the optimal envy-free solution. If $\theta_i=\theta>1$ for all $i\in[b]$, we have $\bOne^\top \btheta^\ef =\Omega(k)$. If $\theta\leq 1$, there exists a constant $C>0$ such that for an large enough $b$,
\begin{align}
    u_i(\btheta^{\ef}) =& 1  - \frac 12 \EEs{\bfm}{\frac{1}{\sqrt{b}}\left(1-\frac 1b \right)^{m_0 + \sqrt{b}m} + \left(1-\frac{1}{\sqrt{b}} \right) \left( 1 - \frac 1b \right)^{m}}\nonumber\\
    \leq& 1-\frac{1}{2\sqrt{b}}\left(1-\frac 1b \right)^{\theta_0 + \sqrt{b}\theta} - \frac 12\left(1-\frac{1}{\sqrt{b}} \right) \left( 1 - \frac 1b \right)^{\theta}\label{eq:usingjensen}\\
    \leq& 1-\frac{1}{2\sqrt{b}}\left(1-\frac 1b \right)^{(1 + \sqrt{b})\theta} - \frac 12\left(1-\frac{1}{\sqrt{b}} \right) \left( 1 - \frac 1b \right)^{\theta}\label{eq:usingdef}\\
    \leq & 1-\frac{1}{2\sqrt{b}}e^{-\ln(4) (1 + \sqrt{b})\theta/b} - \frac 12\left(1-\frac{1}{\sqrt{b}} \right) e^{-\ln(4) \theta/b}\nonumber\\
    \leq& 1-\frac{1}{2\sqrt{b}}\left(1-\frac{C(1 + \sqrt{b})\theta}{b}\right) - \frac 12\left(1-\frac{1}{\sqrt{b}} \right) \left(1-\frac{C \theta}{b}\right)\nonumber\\
    \leq& \frac 12 + \frac{3C\theta}{2b}\nonumber\,,
\end{align}
where Eq.~\eqref{eq:usingjensen} adopts Jensen's inequality and Eq.~\eqref{eq:usingdef} uses the property that $\theta_0\leq \theta$. Then since $u_i(\btheta^{\ef})\geq \mu$, we have $\theta\geq \frac{1}{3C}$. Hence, $\bOne^\top \btheta^\ef =\Omega(k)$ and the Price of Fairness is at least $\Omega (\sqrt{k})$.

\underline{The Optimal stable equilibrium:}
First, by the symmetry of the utility functions for all $i\in I_j$ and for all $j\in [\sqrt{b}]$, any stable equilibrium must satisfy $\theta_i=\theta$ for some $\theta$ and all $i\in [b]$. This is not hard to check. For two petal agents $i$ and $l$ in the same group, for any stable feasible solution, if $\theta_i>\theta_l\geq 0$, then $u_i(\btheta)>u_l(\btheta)\geq \mu$, which results in $\theta_i =0$. This is a contradiction. Now for a stable feasible solution in which any two agents in the same group have the same number of samples, if $\theta_i>\theta_l$ for any $i,l$ in different groups, $u_i(\btheta)>u_l(\btheta)\geq \mu$ and thus, $\theta_i=0$. This is a contradiction. Hence, all petal agents have $\theta_i=\theta$ for all $i\in [b]$.

Furthermore, since in any stable equilibrium with $\theta_i=\theta$ for all $i\in [b]$, $u_0(\btheta)>u_i(\btheta)$, agent $0$ must take $0$ samples in any stable equilibrium. Let $\btheta^\eq$ represent the optimal stable equilibrium. Following the similar computation to the case of envy-free solution, if $\theta\leq 1$, we have 
\begin{align*}
    u_i(\btheta^\eq) =& 1  -  \EEs{\bfm}{\frac{1}{2\sqrt{b}}\left(1-\frac 1b \right)^{m_0 + \sqrt{b}m} + \frac 12\left(1-\frac{1}{\sqrt{b}} \right) \left( 1 - \frac 1b \right)^{m}}\nonumber\\
    \leq& 1-\frac{1}{2\sqrt{b}}\left(1-\frac 1b \right)^{\sqrt{b}\theta} - \frac 12\left(1-\frac{1}{\sqrt{b}} \right) \left( 1 - \frac 1b \right)^{\theta}\\
    \leq& \frac 12 + \frac{3C\theta}{2b}\,.
\end{align*}
Therefore, $\theta \in \Omega(1)$, $\bOne^\top \btheta^\eq=\Omega(k)$ and the Price of Stability is at least $\Omega (\sqrt{k})$.

\paragraph{Linear Utilities.} In this flower structure, for any $i\in I_j$,
\[u_i(\btheta) = \theta_i + \frac{1}{\sqrt{b}}(\theta_0+\sum_{l\in I_j:l\neq i}\theta_l) \]
and
\[u_0(\btheta) = \theta_0 + \frac{1}{\sqrt{b}} \sum_{i=1}^{b}\theta_i\,.\]
Let $\bmu = \bOne$. Here the choice of $\bmu$ implies that the constraint of agent $i$ is met when in expectation at least $one$ time, there is an instance being discovered. Similar to the random coverage example, we have the following results.

\underline{The optimal collaborative solution:} There is one feasible solution in which the core agent takes 
$\sqrt{b}$ samples and all other agents take $0$ samples. This is a feasible solution because the core can help every other agent with effort $\frac{1}{\sqrt{b}}$. Let $\btheta^\optlocal$ denote the socially optimal solution and we have $\bOne^\top \btheta^\optlocal \leq \sqrt{b} = O(\sqrt{k})$.

\underline{The optimal envy-free solution:} By the symmetry of the utility functions, similar to the random coverage case, any envy-free solution must satisfy $\theta_i=\theta$ for some $\theta$ and all $i\in [b]$. 

Furthermore, in any envy-free feasible solution we must have $\theta_0 \leq \theta$ since $ u_0(\btheta^{(0,i)})\geq u_i(\btheta) \geq 1$. In other words, in any envy-free solution the $0$-th agent's number of sample can be no larger than any other agent, and all other agents take the same number of samples. Let $\btheta^\ef$ denote the optimal envy-free solution. We have
\[1 \leq u_i(\btheta^\ef)\leq \theta + \frac{1}{\sqrt{b}}(\theta+\sum_{l\in I_j:l\neq i}\theta)=2\theta\,,\]
which indicates that $\theta\geq 1/2$. Therefore, $\bOne^\top \btheta^\ef \geq \frac{b}{2}$ and the Price of Fairness is at least $\Omega (\sqrt{k})$.

\underline{The optimal stable equilibrium:}
By the symmetry of the utility functions, similar to the random coverage case, any stable equilibrium must satisfy $\theta_i=\theta$ for some $\theta$ and all $i\in [b]$. Then $u_0(\btheta)=\theta_0+\sqrt{b}\theta$ and $u_i(\btheta) = (2-\frac{1}{\sqrt{b}})\theta + \frac{1}{\sqrt{b}}\theta_0 < u_0(\btheta)$ for $b\geq 2$. Therefore, agent $0$ must take $0$ samples in any stable equilibrium. Then for optimal stable equilibrium $\btheta^\eq$, it is not hard to find that
\[1 \leq u_i(\btheta^\eq)\leq \theta + \frac{\sqrt{b}-1}{\sqrt{b}}\theta\,,\]
which indicates that $\theta \geq \frac{1}{2}$. Therefore, $\bOne^\top \btheta^\eq \geq \frac{b}{2}$ and the Price of Stability is at least $\Omega (\sqrt{k})$.
\end{proof}

\section{Proofs of Theorem~\ref{thm:lineareqopt} and Corollary~\ref{cor:lineareqopt}}\label{app:lineareqopt}

To prove Theorem~\ref{thm:lineareqopt} and Corollary~\ref{cor:lineareqopt}, we first introduce the following three lemmas.

\begin{lemma}\label{lmm:linearpridual}
For any optimal stable equilibrium $\btheta^\eq$ for linear utilities $u_i(\btheta) = W_i^\top \btheta$ and $\mu_i=\mu$ for $i\in[k]$, $\bar{\btheta}^\eq$ is a socially optimal solution for the set of agents $i\in [k]\setminus I_{\btheta^\eq}$, i.e., $\bar{\btheta}^\eq$ is an optimal solution to the following problem.
\begin{equation}\label{eq:linearrm}
\begin{array}{ll}
\min_{\bx} &\bOne^\top \bx \\
\st &\bar{W} \bx\geq \mu \bOne\\
&\bx \geq \bZero\,.
\end{array}
\end{equation}
\end{lemma}

\begin{proof}
The dual problem of Equation~\eqref{eq:linearrm} is 
\begin{equation*}
\begin{array}{ll}
\max_{\by} &\mu \bOne^\top \by \\
\st &\bar{W} \by\leq \bOne\\
&\by \geq \bZero\,,
\end{array}
\end{equation*}
which is equivalent to
\begin{equation}\label{eq:lineardual}
\begin{array}{ll}
\max_\by &\bOne^\top \by \\
\st &\bar{W} \by\leq \mu \bOne\\
&\by \geq \bZero\,.
\end{array}
\end{equation}
Due to the definition of stable equilibrium, for agent $i\in [k]\setminus I_{\btheta^\eq}$, we have $\theta^\eq_i\neq 0$ and thus, $\bar{W} _i^\top \bar{\btheta}^\eq = W_i^\top \btheta^\eq =\mu$. Therefore, $\bar\btheta^\eq$ is a feasible solution to both the primal problem~\eqref{eq:linearrm} and the dual problem~\eqref{eq:lineardual}. This proves that  $\bar\btheta^\eq$ is an optimal solution to Equation~\eqref{eq:linearrm}.
\end{proof}

\begin{lemma}\label{lmm:linearallmu}
   If $\bar\btheta$ is an optimal solution to Equation~\eqref{eq:linearrm}, then $\bar{W}\bar\btheta = \mu\bOne$.
\end{lemma}
\begin{proof}
As proved in Lemma~\ref{lmm:linearpridual}, $\bar{\btheta}^\eq$ is an optimal solution to Equation~\eqref{eq:linearrm} with $\bar W \bar\btheta^\eq = \mu\bOne$. Assume that there exists another optimal solution $\bar\btheta$ such that $\bar W \bar\btheta = \mu\bOne+ \bv$ with $\bv\geq \bZero$. Let $T^* = \bOne^\top\bar\btheta^\eq = \bOne^\top\bar\btheta$ denote the optimal value of Equation~\eqref{eq:linearrm}. Then we have
\[
\bar\btheta^\top \bar W \bar\btheta^\eq=\bar\btheta^\top \mu \bOne = \mu T^*\,,
\]
and 
\[
\bar\btheta^{\eq\top} \bar W \bar\btheta =\bar\btheta^{\eq\top} (\mu \bOne +v)= \mu T^* + \bar\btheta^{\eq\top} \bv\,.
\]
Hence, $\bar\btheta^{\eq\top} \bv=0$. Since $\bar\btheta^{\eq}>\bZero$, then $\bv=\bZero$ and $\bar{W}\bar\btheta = \mu\bOne$.
\end{proof}

Without loss of generality, we let
\[
    W=\begin{bmatrix}\bar W & B\\ B^\top & C\end{bmatrix}\,,
\]
and let $d= k-\abs{I_{\btheta^\eq}}$ denote the dimension of $\bar W$.

\begin{lemma}\label{lmm:linearnull}
   If $\bar\btheta$ is an optimal solution to Equation~\eqref{eq:linearrm}, then we have $B^\top (\bar \btheta^\eq - \bar \btheta) = \bZero$.
\end{lemma}
\begin{proof}
If $\bar W$ is a full-rank matrix, then the optimal solution to Equation~\eqref{eq:linearrm} is unique and thus, $\bar\btheta=\bar\btheta^{\eq}$. 

If $\bar W$ is not a full-rank matrix, we assume that $\bar\btheta \neq \bar\btheta^\eq$. Let $\bv_1,\bv_2,\ldots,\bv_d$ denote the eigenvectors of $\bar W$ with eigenvalues $\lambda_1\geq \lambda_2\geq\ldots\geq \lambda_d$. Since $\bar W$ is not a full-rank matrix, let $d'$ denote the number of zero eigenvalues and we have $\lambda_{d-d'+1} = \ldots \lambda_d =0$. We let $\bb_i$ denote the $i$-th column of $B$ and $c_i = C_{ii}\in[0,1]$.

For any $i\in [k-d]$, let $(\bx,y\be_i)$ with any $\bx\in \R^d,y\in \R$ denote a $k$-dimensional vector with the first $d$ entries being $x$, the $d+i$-th entry being y and all others being $0$s. Since $W$ is PSD, we have
\begin{align*}
    (\bx,y\be_i)^\top W (\bx,y\be_i) = \bx^\top \bar W \bx + 2y\bb_i^\top \bx + c_i y^2 \geq 0\,.
\end{align*}
For any $j=d-d'+1, \ldots, d$, let $\bx = \bv_j$ and $y=-\bb_i^\top \bv_j$, then we have 
\[
    (2-c_i)(\bb_i^\top \bv_j)^2\leq \bv_j^\top \bar W \bv_j = 0\,,
\]
and thus $\bb_i^\top \bv_j=0$ for all $j=d-d'+1, \ldots, d$. By Lemma~\ref{lmm:linearallmu}, we know that $\bar W (\bar \btheta^\eq - \bar \btheta)=\bZero$. Hence $\bar \btheta^\eq - \bar \btheta$ lie in the null space of $\bar W$, i.e., there exists $\balpha\neq \bZero\in \R^{d'}$ such that $\bar \btheta^\eq - \bar \btheta = \sum_{i=1}^{d'}\alpha_i \bv_{d+1-i}$. Then $\bb_i^\top (\bar \btheta^\eq - \bar \btheta) =  \sum_{i=1}^{d'}\alpha_i \bb_i^\top \bv_{d+1-i} = 0$.
\end{proof}

Now we are ready to prove Theorem~\ref{thm:lineareqopt}.
\rstthmlinearopt*
\begin{proof}
  Lemma~\ref{lmm:linearpridual} proves the first part of the theorem. For the second part of the theorem, we prove it by using Lemma~\ref{lmm:linearnull}. For $i\in I_{\btheta^\eq}$, $W_i^\top \tilde{\btheta} = \bar W_i \bar\btheta = \mu$. For $i\in [k]\setminus I_{\btheta^\eq}$, by Lemma~\ref{lmm:linearnull} we have $W_i^\top \tilde{\btheta} = \bb_i^\top \bar\btheta = \bb_i^\top \bar\btheta^\eq = W_i^\top \btheta^\eq \geq \mu$. Therefore, $\tilde{\btheta}$ is a stable equilibrium. Combined with that $\bOne^\top \tilde{\btheta} =\bOne^\top \bar{\btheta} =\bOne^\top \bar{\btheta}^\eq =\bOne^\top {\btheta}^\eq$,  $\tilde{\btheta}$ is an optimal stable equilibrium for agents $[k]$.
\end{proof}

\rstcrllinearopt*
Corollary~\ref{cor:lineareqopt} is a direct result of Theorem~\ref{thm:lineareqopt}.

\section{Proof of Theorem~\ref{thm:lineareqef}}\label{app:lineareqef}
\rstlineareqef*
\begin{proof}
Note that only agents with non-zero number of samples can envy others.
Assume on the contrary that there is agent $i$ with  $\theta^\eq_i>0$ that envies another agent $j$. By the definition of a stable equilibrium, we have that 
$W_i^\top \btheta^\eq =\mu_i$. Let ${\btheta}^{(i,j)}$ represent the strategy with $i$ and $j$'s contributions swapped. Then, 
\[ u_i({\btheta}^{(i,j)})= u_i (\btheta) + (\theta_i - \theta_j)(W_{ij} - W_{ii}) < u_i (\btheta)  = \mu_i,
\]
where the second transition is by $\theta_i> \theta_j$ and $W_{ii} >W_{ij}$. This shows that no agent can have envy in an equilibrium.
\end{proof}
\section{Structure of Equilibria in Random Coverage}\label{app:ncvxef}
In Section~\ref{sec:linearcvx}, we show that the optimal stable equilibrium can be computed by a convex program in the linear case. However, this is not true in random coverage. In the following, we provide an example in which the utility function is non-concave and the the stable feasible set is non-convex. In addition, we provide another example in which the envy-free feasible set is non-convex.

\subsection{Proof of Theorem~\ref{thm:cvrgeqnoncvx}}
\rstthmcvrgeqnoncvx*
\begin{proof}
Let us consider an example where there are $2$ agents and both are with a uniform distribution over the instance space $\cX=\{0,1\}$. Then for any $i\in [2]$, agent $i$'s utility function is
\[u_i(\btheta) = 1-\frac{1}{2}\EEs{\bfm}{\left(\frac{1}{2}\right)^{m_1+m_2}}\,.\]
By direct computation, we have $u_i(\be_1)= u_i(\be_2) = 1-\frac{1}{2}\cdot \frac{1}{2} = \frac{3}{4}$. For $\alpha\in (0,1)$, 
\begin{align*}
    u_i(\alpha \be_1 +(1-\alpha) \be_2) = 1-\frac{1}{2}\left(\alpha\cdot \frac{1}{2} +(1-\alpha) \cdot 1\right)\left((1-\alpha)\cdot \frac{1}{2} + \alpha\cdot 1\right) = \frac 34 - \frac{\alpha(1-\alpha)}{8}\,,
\end{align*}
which is smaller than $\alpha u_i(\be_1) +(1-\alpha) u_i(\be_2)$. Therefore, the utilities in this example are non-concave. 

Let $\mu_i = \frac{3}{4}$ for $i=1,2$. Then, $\be_1$ and $\be_2$ are stable equilibria as no agent has incentive to decrease her number of samples. However, since $\alpha \be_1 +(1-\alpha) \be_2$ is not a feasible solution, the stable feasible set is non-convex.
\end{proof}

\subsection{Proof of Theorem~\ref{thm:cvrgefnoncvx}}
\rstthmcvrgefnoncvx*
\begin{proof}
Now we consider another example showing that the envy-free feasible set is non-convex. Considering the complete graph on $4$ vertices and let each edge correspond to one agent. As illustrated in Figure~\ref{fig:efnoncvx}, we put one point in the middle of every edge and one point on every vertex and let each agent's distribution be a uniform distribution over $\cX_i$, which is the $3$ points on agent $i$'s edge. 
\begin{figure}[t]
    \centering
    \includegraphics[width=0.3\textwidth]{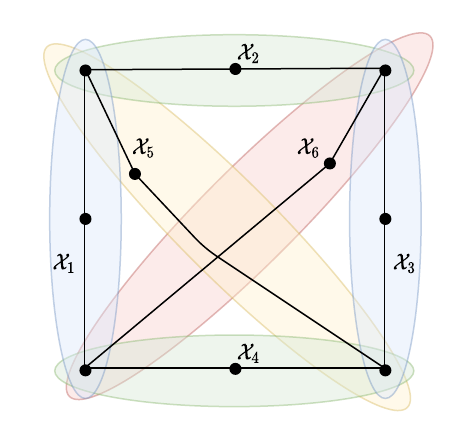}
    \caption{Illustration of the example.}
    \label{fig:efnoncvx}
\end{figure}

Then agent $i$ utility function is
\[
u_i(\btheta) = 1-\frac{1}{6}\EEs{\bfm}{\sum_{x\in \cX_i} (\frac{2}{3})^{n_x}}\,,
\]
where $n_x = \sum_{j:x\in \cX_j} m_j$.
Let $\mu_i=0.6$ for all $i$. Then we consider a solution: pick any perfect matching on this complete graph and then let $\theta_i=1$ if edge $i$ is in this matching and $\theta_i=0$ otherwise. Such a $\btheta$ is an envy-free solution. In this solution, for agent $i$ with $\theta_i=1$, any point $x\in\cX_i$ has $n_x=1$ and the utility is 
\[u_i(\btheta) = 1-\frac{1}{6}\left(3\cdot \frac{2}{3}\right)\geq 0.6\,;\]
for agent $i$ with $\theta_i=0$, two points in $\cX_i$ has $n_x=1$ and one point (in the middle of the edge) has $n_x=0$, and the utility is
\[u_i(\btheta) = 1-\frac 16 \left(2\cdot\frac{2}{3} + 1\right)=\frac{11}{18}\geq 0.6\,.\]
If $\theta_i=1$ and agent $i$ envies another agent $j$ with $\theta_j=0$, agent $i$'s utility after swapping $\theta_i$ and $\theta_j$ is
\[u_i(\btheta^{(i,j)}) = 1-\frac 16\left(\frac{2}{3} + 2\right)=\frac 59<0.6\,.\]
Therefore, this is an envy-free solution.

Then let $\btheta = \be_1 +\be_3$ and $\btheta' = \be_2 +\be_4$. Both are envy-free solutions. Now we show that $\btheta'' = 0.9\btheta+0.1\btheta'$ is not envy-free.
First we show that the agent $2$ meets her constraint in solution  $\btheta''$.
\[u_2(\btheta'') = 1-\frac{1}{6} \left(2\cdot\left(0.09\cdot \left(\frac{2}{3}\right)^2 + 0.82\cdot \frac{2}{3}+0.09\right) + \left(0.1\cdot \frac{2}{3} + 0.9\right)\right)\geq 0.6\,.\]
Now we show that agent $2$ can still meet her constraint after swapping with agent $6$. After swapping $\theta''_2$ and $\theta''_6$, agent $2$'s utility is
\[u_2(\btheta''^{(2,6)}) = 1-\frac{1}{6} \left(\left(0.09\cdot \left(\frac{2}{3}\right)^2 + 0.82\cdot \frac{2}{3}+0.09\right) +1+ \left(0.9\cdot \frac{2}{3} + 0.1\right)\right)\geq 0.6\,.\]
Therefore, $\btheta ''$ is not envy-free and the envy-free feasible set in this example is non-convex.
\end{proof}

\section{Experimental}\label{experimental_section}

\subsection{Dataset}

We use the balanced split of the EMNIST, which is meant to be the broadest split of the EMNIST dataset \citep{DBLP:journals/corr/CohenATS17}. The task consists of classifying English letters and whether they are capitalized or lowercase. Some letters which are similar in their upper and lower case forms, such as C and P, are merged, resulting in just 47 distinct classes. From this dataset, we randomly sample 60,000 points for training and validating the federated learning algorithms. We then take a disjoint sample of an additional 30,000 points to pre-train the model that we will later fine-tune via federation. To select hyperparameters for this model (which we will also use for the federated algorithms), we take the remaining 31,600 points as a validation set. We use top-1 accuracy as the performance metric.

\begin{table}[h]
\centering

\begin{tabular}{|cc|}
\toprule
\textbf{Dataset}                                      & \textbf{Number of Points} \\
\midrule
Potential Training and Validation for Agents & 60,000           \\
Pre-Training                                 & 30,000           \\
Pre-Training Validation                      & 41,600         \\
\bottomrule
\end{tabular}
\end{table}

\subsection{Learning model}
\paragraph{Model} We use a straightforward four-layer neural network with two convolutional layers and two fully-connected layers. We optimize the model with Adam \citep{kingma15} and use Dropout \citep{srivastava2014dropout} for regularization. Architecture details and an implementation can be found via \href{https://github.com/rlphilli/Collaborative-Incentives}{Collaborative-Incentives on Github}. As stated previously, we pre-train the model for $40$ epochs to an accuracy of approximately $55\%$.

\begin{table}[h]
\centering
\begin{tabular}{|ccccc|}
\toprule
\textbf{Algorithm}           & \textbf{Batch Size Per Agent}   & \textbf{Learning Rate} & \textbf{Threshold Accuracy} & \textbf{Local Batches}  \\
\midrule
Individual Learning & 256          & 0.002          & N/A\%   &  N/A\%           \\
FedAvg              & 64           & 0.002          & N/A\%     & 1          \\
MW-FED              & 64 (Average) & 0.002          & 70\%    & 1   \\
\bottomrule        
\end{tabular}
\end{table}

We select hyperparameters using a randomized search on the pre-training validation set. The grid for this search consists of logarithmically-weighted learning rates between $1e-06$ and $1e-02$ and batch sizes of $4$, $8$, $64$, $128$, $256$, and $512$ all together sampled $40$ times. Parameters selected for the individual learning sampling are equivalently translated to the federated learning algorithms.

\begin{algorithm}[H]\caption{\textsc{FedAvg} (simplified to sample all populations each iteration) Let $\eta$ be the learning rate, $m$ be the minibatch size, $B$ be the number of local batches, $k$ be the number of clients, $X_i$ be the set of points for agent $i$, and $\ell$ the loss function }\label{alg:FED}
\begin{algorithmic}[1]
\STATE initialize server weights $\beta_{serv}$ and client weights $\beta_0 \dots \beta_k$
\FOR{each round t=1,2 \dots T }
\FOR{each client $i\in k$}
\STATE $\beta_i \leftarrow{\beta_{serv}}$
\FOR{each local batch $j$ from $1, 2, \dots B$}
\STATE \textbf{sample} $m$ points $x$ from $X_i$
\STATE $\beta_i \leftarrow \beta_i - \eta \nabla \ell \left(\beta_i ; x\right)$
\ENDFOR
\ENDFOR
\STATE $\beta_{serv} \leftarrow  \frac{1}{B \cdot k} \sum_{i=1}^k \beta_i$
\ENDFOR
\STATE return $\beta_{serv}$
\end{algorithmic}
\end{algorithm}

\begin{algorithm}[H]\caption{\textsc{MW-FED} Let $\eta$ be the learning rate, $m$ be the \textit{average} minibatch size, $B$ be the average number of local batches, $k$ be the number of clients, $c$ be the multiplicative factor, $X_i^{train}$ and $X_i^{val}$ be the sets of training and validation points, respectively, for agent $i$, $\epsilon_i$ the desired maximum loss for agent $i$, and $\ell$ the loss function }\label{alg:MW}
\begin{algorithmic}[1]
\STATE initialize server weights $\beta_{serv}$ and client weights $\beta_0 \dots \beta_k$
\STATE initialize contribution-weights $w_1, w_2, \dots w_k = \frac{1}{k}$
\FOR{each round t=1,2 \dots T }
\FOR{each client $i\in k$}
\STATE $\beta_i \leftarrow{\beta_{serv}}$
\STATE $m_i \leftarrow  m \cdot B \cdot \frac{k\cdot w_i}{\sum w}$
\FOR{each local batch $j$ from $1, 2, \dots \floor{\frac{m_i}{m}}$}
\STATE \textbf{sample} $m$ points $x$ from $X_i^{train}$
\STATE $\beta_i \leftarrow \beta_i - \eta \nabla \ell \left(\beta_i ; x\right)$
\ENDFOR
\ENDFOR
\STATE $\beta_{serv} \leftarrow  \frac{1}{ \sum_{i=1}^{k} \floor{\frac{m_i}{m}}} \cdot  \sum_{i=1}^k \beta_i \cdot \floor{\frac{m_i}{m}}  $
\FOR{each client $i\in k$}
    \IF{$\ell\left(\beta_{serv} ; X_i^{val}\right) \geq \epsilon$}
    \STATE $w_i \leftarrow c \cdot w_i$                                 
    \ENDIF 
\ENDFOR
\ENDFOR
\STATE return $\beta_{serv}$
\end{algorithmic}
\end{algorithm}

\subsection{Encouraging heterogeneity across agent datasets}
To encourage heterogeneity between the different agents, we run a series of sampling trials to determine which training points lead to convergence on a holdout data set most quickly. Specifically, over $10,000$ trials we randomly sample the potential agent training set for 1000 points. Then, we train a newly instantiated instance of our network on this data with a batch size of 16 until it reaches a cross-entropy loss of 0.5. For each trial we record the number of iterations it takes for the model to reach 60\% accuracy. At the end of the trials we find the average number of iterations for trials that each point was involved in. The range of these values is from 235 to 670 batches. The mean is 286 and the standard deviation is 21.6 iterations. We then generate agents using mixtures of samples from the top 10\% and bottom 10\% of difficult examples in terms of time to reach the threshold. 

This is an imperfect proxy for difficulty, but we found it useful for producing observable heterogeneity in our chosen samples. We considered other proxies for data value and uncertainty such as output entropy for a sample on the pre-trained model, but found that, in many cases, these samples did not do as much to create differences in how quickly a model trained.

For the main experiment of this section, we create 4 different mixtures : one distribution of 100\% difficult samples, a mixture of 90\% difficult samples, a mixture of 90\% easier samples, and one distribution of 100\% easy samples. As opposed to individual devices, these mixtures might be considered as four different populations with similar, but not identical, objectives. One-hundred averaged training runs for each of these 4 distributions can be found in Figure \ref{fig:solo}.\footnote{Note that, as the batch size differs, these iteration counts can not be directly compared with other statistics in this section.} This figure also shows that they are, in fact, distinct from one another over many repetitions of their training regimes.

\begin{figure}[ht]
    \centering
 \includegraphics[width=.4\textwidth]{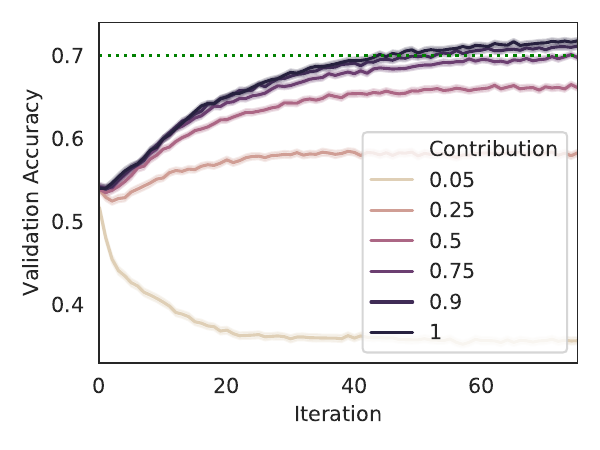}
    \caption[Individual (non-federated) learning averaged over all four agents with different individual contribution levels. ]{Individual (non-federated) learning averaged over all four agents with different individual contribution levels. At the size of each agent's training dataset (1600), using half or fewer of an agent's unique data points will generally not lead to success.}
    \label{fig:circsolo}
\end{figure}

\paragraph{Non-federated defection is not enough} An important note is that distributions that are often happy while making large defections in the federated settings in Figure \ref{fig:Doublebar} are not generally happy with much less data. Figure \ref{fig:circsolo} shows the averaged learning trajectories over agents who, in the non-federated setting, only use a fraction of their data. In this setting, agents can reduce their contributions by very little if they still hope to be successful. %

\subsection{Connections with algorithms in prior work}
Algorithm \ref{alg:MW} mirrors the multiplicative weights-based solutions that \citet{blum2017collaborative, chen:tight2018, nguyen2018improved} use in the learning-theoretic setting. Specifically, the algorithms in the above prescribe learning in rounds. Each round involves sampling from a weighted mixture of distributions, testing the performance of the learned model on each distribution, and up-weighting those that have not yet reached their performance threshold for the following rounds. %

Section \ref{sec:linearcvx} shows that, in the linear setting, we can use a convex program to find a minimum-cost equilibrium. As previously stated, ensuring there are no $0$-contributors means that we can simply use \ref{eq:LP} to find an equilibrium. Packing LPs such as this are frequently solved using similar multiplicative-weights based strategies \citep{10.2307/3690406, arora_multiplicative_nodate}.

\subsection{Source code}
Code is available \href{https://anonymous.4open.science/r/b3084725-612c-4c6a-8716-61dd0326ec8d/}{in an anonymized repo here.}

\subsection{Computing infrastructure}
The experiments in this work were run using a NVIDIA V100 Tensor Core GPU.
\chapter{Incentives in Multi-Round Federated Learning}\label{app:incentives-multi}

\section{Missing Details from Section \ref{sec:intro}}
\begin{figure}[h]
     \centering
     \begin{subfigure}[b]{0.32\textwidth}
         \centering
         \includegraphics[width=\textwidth]{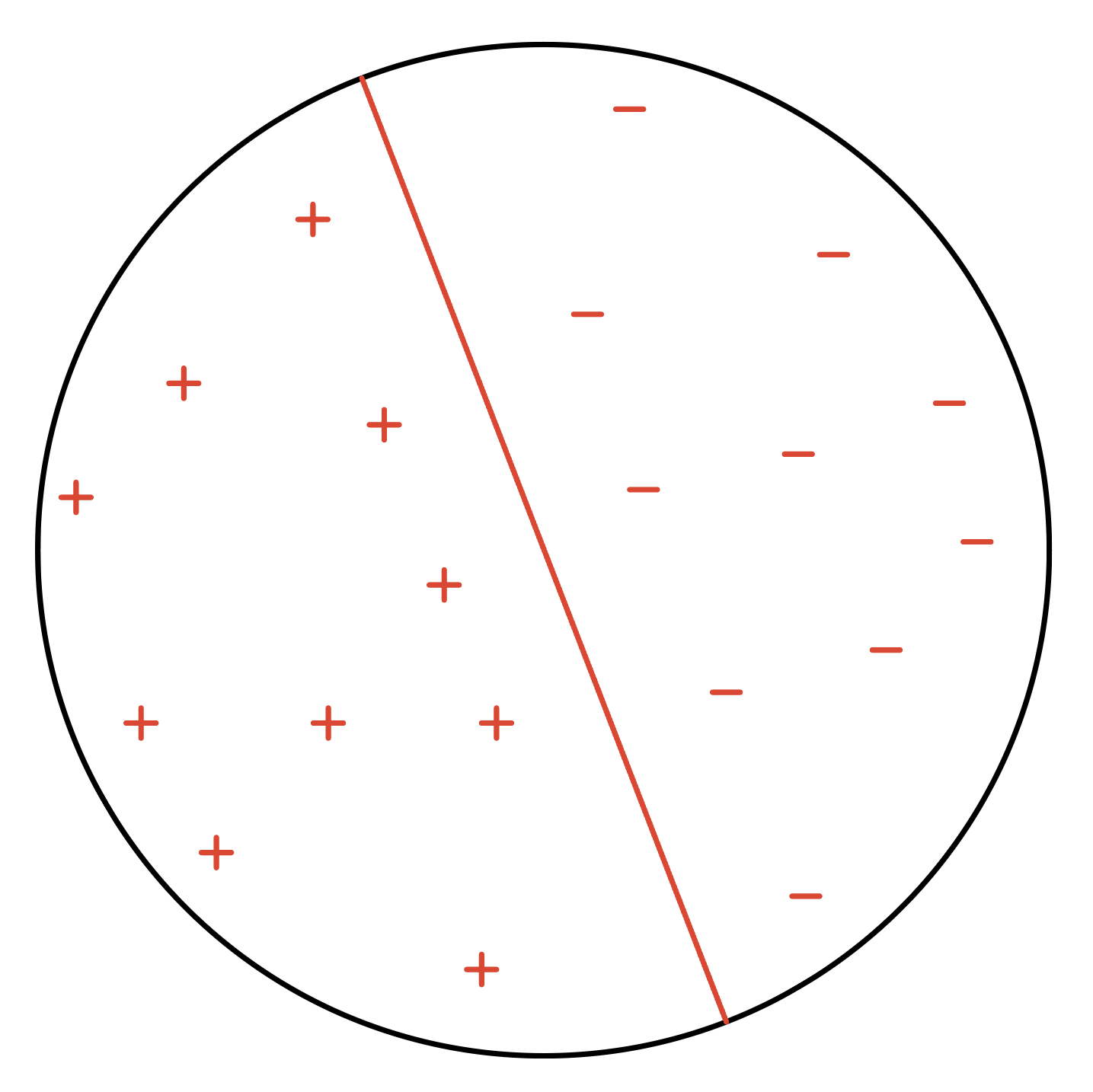}
         \caption{$\ddd_{red}$}
         \label{fig:red}
     \end{subfigure}
     \hfill
     \begin{subfigure}[b]{0.32\textwidth}
         \centering
         \includegraphics[width=\textwidth]{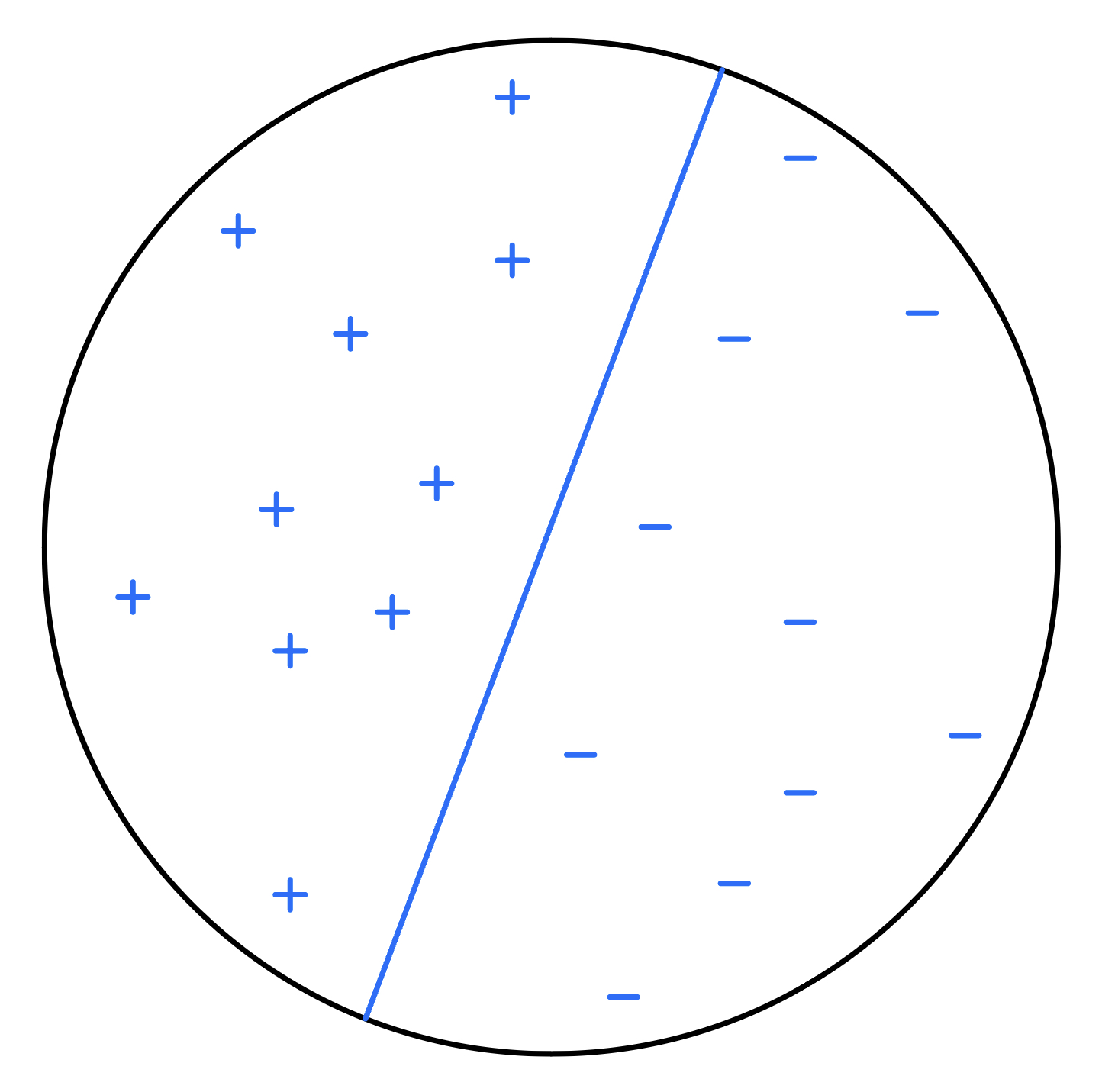}
         \caption{$\ddd_{blue}$}
         \label{fig:blue}
     \end{subfigure}
     \hfill
     \begin{subfigure}[b]{0.32\textwidth}
         \centering
         \includegraphics[width=\textwidth]{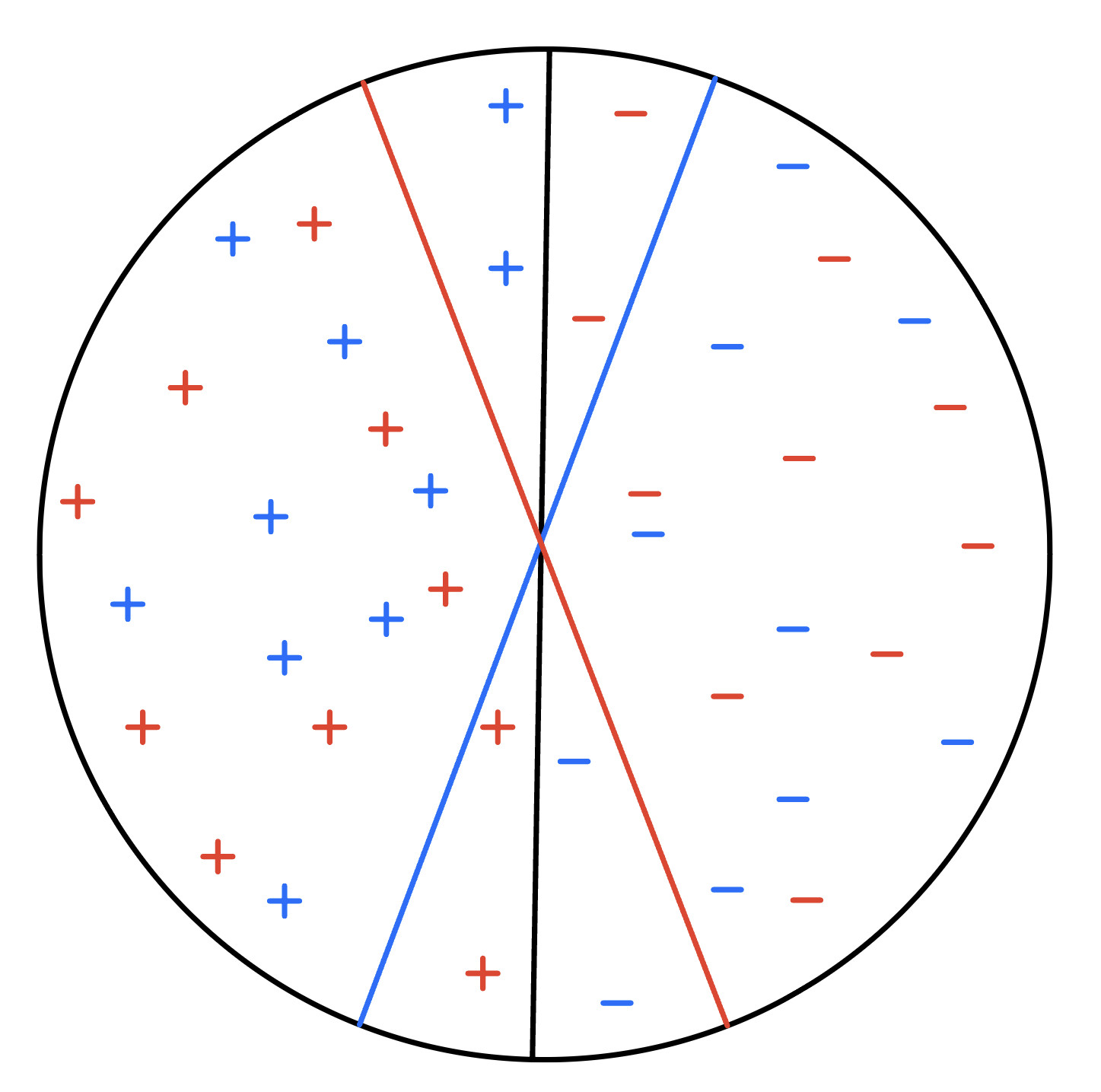}
         \caption{$\ddd_{red}\cup \ddd_{blue}$}
         \label{fig:combined}
     \end{subfigure}
        \caption[An example of two agents in which performance degrades due to defections.]{ 
        Consider two agents \{red, blue\} with distributions $\ddd_{red}$ and $\ddd_{blue}$ on the ball in $\R^2$. Figures \ref{fig:red} and \ref{fig:blue} depict these distributions, where the number of $+$ and $-$ represent point masses in the density function of each distribution. Note that each distribution has multiple zero-error linear classifiers, but we depict the max-margin classifiers in plots (a) and (b). Figure \ref{fig:combined} shows the combined data from these agents and the best separator (black), classifying the combined data perfectly. However, the red (blue) separator on the blue (red) data has a higher error rate of $20$ percent. Thus, if either agent defects during training, any algorithm converging to the max-margin separator for the remaining agent will incur an average $10$ percent error.
        }
        \label{fig:defection_cartoon}
\end{figure}

\section{Other Related Work}\label{app:related}
Designing incentive mechanisms for federated learning has received much attention recently, but as illustrated in this paper it is a hard problem. The main difficulty lies in an \textit{``information asymmetry"} \citep{kang2019incentive}. The server does not know the available computation resources and the data sizes on the devices for training. Furthermore, it doesn't know the local data quality of a device and can't estimate it using common metrics such as data Shapley \citep{ghorbani2019data} as it doesn't have access to raw data. As a result, the server may incur a high cost when incentivizing the devices to encourage truthfulness or avoid defections. Or worse, it might not even be able to incentivize desired behavior. In this final section, we survey recent advances in solving this problem. We divide the survey into four parts,
\begin{enumerate}
    \item In section \ref{sec:valuation}, we survey papers that tackle the problem of \textit{agent selection} based on several factors, but primarily the value of the data the agent can provide. This is an important step because for cross-device FL to work, the agents defining problem \eqref{eq:relaxed} should (i) represent the meta distribution $\ppp$ well, (ii) have enough data samples/computational power to converge to a solution for \eqref{eq:relaxed}.
    \item In section \ref{sec:collection}, we survey the papers that assume the pool of agents is fixed, but the agents are self-serving and want to contribute the least amount of data possible. This is because data collection and privacy costs are involved with sharing data. The works in this category also tackle the "free-rider" problem \citep{karimireddy2022mechanisms} so a handful of devices don't collect most of the data at equilibrium. 
    \item In section \ref{sec:csfl}, we survey papers relevant to a simpler problem called \textit{cross-silo FL} \citep{kairouz2019advances} where $\ppp$ is supported on a finite set of machines, and as a result, we can directly consider problem \eqref{eq:relaxed}. Defection is not a problem in this setting, and we are more interested in incentivizing the machines to contribute higher-quality model updates.
    \item Finally, in section \ref{sec:cdfl}, we survey the very limited work in the general problem we are interested in solving, i.e., avoiding defections in cross-device FL.
\end{enumerate}
There is another orthogonal line of research on fairness in FL (see \cite{ezzeldin2021fairfed} and the references within, for example). 

\subsection{Data Valuation and Agent Selection}\label{sec:valuation}

There are two main series of works that are most relevant. The first one used contract theory to study the design of incentive mechanisms under a principal-agent model. 
   
\cite{kang2019incentive} proposed a contract theory-based incentive mechanism for federated learning in mobile networks. They considered the principal-agent model that there are $M$ types of agents, where the agent's local data quality defines the type. The server does not know the type of each device and only has a prior distribution over all agent types. It aims to design an incentive-compatible mechanism that incentivizes every agent to behave truthfully while maximizing the server's expected reward.  \cite{tian2021contract} extended this model by considering the agent's willingness to participate as part of the type. As a result, they proposed a two-dimensional contract model for the incentive mechanism design. \cite{cong2020vcg} considered a similar model where the data quality and cost define the agent's type. The server decides payment and data acceptance rates based on the reported agent type. The authors proposed implementing the payment and data acceptance rate functions with neural networks and showed they satisfy proper properties such as incentive compatibility.  Another related work is by \cite{zeng2020fmore}, who proposed a multi-dimension auction mechanism with multiple winners for agent selection. The authors presented a first-price auction mechanism and analyzed the server's equilibrium state. 

Besides applying contract theory to incentivize agents' truthful behavior, there has also been another series of work that focuses on designing reputation-based agent selection schemes to select high-quality agents. Specifically, \cite{kang2019incentive} proposed a model that works as follows: the server computes the reputation of every agent and selects agents for the federated learning based on the reputation scores, and the reputation of every agent is updated by a management system after finishing an FL task, where a consortium block-chain implements the reputation management system. \cite{zhang2021incentive} extended the reputation-based agent selection scheme with a reverse auction mechanism to select agents by combining their bids and comprehensive reputations. \cite{xu2020reputation} proposed a different reputation mechanism that computes the reputation of every agent based on the correlation between their report and the average report. Each agent will be rewarded according to their reputation, and those with low reputations will be removed. 

Finally, it is also worth mentioning two other related works. \cite{richardson2020budget} designed another data valuation method. They collect an independent validation data set and value data through an influence function, defined as the loss decreases over the validation data set when the data is left out. And \cite{hu2020trading} cast the problem as a Stackelberg game and choose a subset of agents, which they claim are most likely to provide high-quality data. They provide extensive experiments to test their idea. 

\subsection{Incentivizing Agent Participation and Data Collection}\label{sec:collection}

Usually, the utility of participating agents is a function of model accuracy, computation/sample collection cost, and additional constant cost, e.g., communication cost and payment to the server. In a simple case where the model accuracy is a function of the total number of input samples, \cite{karimireddy2022mechanisms, zhang2022enabling} show that when the server gives the same global model to every agent like federated averaging \citep{mcmahan2016communication} does, a Nash Equilibrium exists for individual sample contributions. But, at this Nash Equilibrium, agents with high sampling costs will be ``free-riders" and will not contribute any data, while agents with low sampling costs will contribute most of the total data.

To alleviate this situation, \cite{karimireddy2022mechanisms} propose a mechanism to incentivize agents to contribute more by customizing the final model's accuracy for every agent, which means sending different models to different agents.
They show that their mechanism is ``data-maximizing" in the face of rational agents. \cite{sim2020collaborative} study a similar setting as \cite{karimireddy2022mechanisms} and aim to provide model-based rewards. Specifically, the authors use information gain as a metric for data valuation and show their reward scheme satisfies some desirable properties. \cite{zhang2022enabling} also study an infinitely repeated game, where the utility is the discounted cumulative utility, and propose a cooperative strategy to achieve the minimum number of free riders.

\cite{zhan2020learning} formulate the problem as a Stackelberg game, where the server decides the payment amount first, and agents decide the amount of data they are willing to contribute. The server's utility is the model accuracy minus the payment, while the agents' utility is the payment (proportional to the contributing data size) minus the computation cost. In this setting, the agents do not care about accuracy, which differs from our case.

In a bit of orthogonal work, \cite{cho2022federate} propose changing objective \eqref{eq:relaxed} itself to maximize the number of agents that benefit from collaboration. They provide some preliminary theoretical guarantees for a simple mean estimation problem. While in our setting, we assume the server can get the model from agents. \cite{liu2020incentives} study the problem of eliciting truthful reports of the learned model from the agents by designing proper scoring rules. Specifically, the authors consider two settings where the server has a ground truth verification data set or only has access to features. The authors demonstrate the connections between this question and proper scoring rule and peer predictions (i.e., information elicitation without verification) and test the performance with real-world data sets.

\subsection{Maximizing Data Quality in Cross-silo FL}\label{sec:csfl}

\cite{kairouz2019advances} demarcate between two prominent federated learning paradigms: cross-silo and cross-device. We have already discussed an example of cross-device Fl: training on mobile devices. Cross-silo FL captures the traditional training in data centers or between big organizations with similar interests. One example is a collaboration between medical institutions to improve their models without leaking sensitive patient information \cite{bergen2012genome}. In cross-silo FL, usually, the agents initiate the FL process and pay the central server for global aggregation. As a result, defection is not an issue in cross-silo FL because, ultimately, the goal is to develop better individual models. Unfortunately, there can still be ``free-riding" behavior in the cross-silo setting \citep{zhang2022enabling, richardson2020budget} as the devices have incentives to contribute less to maximize their own benefit. Therefore, maximizing the data quality is one of the main problems in cross-silo FL.

\cite{xu2021gradient} propose a heuristic Shapley score based on the gradient information from each agent. The score is calculated after communication by comparing the alignment of an agent's gradient with the aggregate gradient. Then the agents with a high score are provided an un-tarnished version of the aggregated gradient, while the agents with a lower score only get a noisy version of the gradient. This incentivizes devices to provide higher-quality gradient updates to get a final model close to the model of the server. There are, unfortunately, no guarantees showing this won't hurt the optimization of objective \eqref{eq:relaxed}, or it at least provably maximizes data quality in any sense. A similar idea has also been used in \cite{shi2022fedfaim}. \cite{zheng2021fl} also propose an auction mechanism modeled using a neural network that decides the appropriate perturbation rule for agents' gradients and an aggregation rule that helps recover a good final guarantee despite this perturbation. \cite{richardson2020budget} take a different approach, and instead of perturbing the model updates, they make budget-bound monetary payments to devices. Their metric is like the leave-one-out metric in data valuation but for model updates. Finally, \cite{tang2021incentive} formulate a social welfare maximization problem for cross-silo FL and propose an incentive mechanism with preliminary theoretical guarantees.

\subsection{Towards Avoiding Defections in Cross-device FL}\label{sec:cdfl}
There doesn't exist any theoretical work for our proposed problem, i.e., avoiding agent defection while optimizing problem \eqref{eq:relaxed} using an iterative algorithm with several communication rounds. There are, however, some empirical insights in other works.     

The most relevant work is about \textsc{MW-FED} algorithm \citep{blum2021one}, an algorithm that fits into the intermittent communication model and explicitly slows down the progress of devices closer to their target accuracies. Specifically, the algorithm asks the devices to report their target accuracies at the beginning of training. Then after each communication round, the devices report their validation accuracy on the current model. The server then uses a multiplicative weight update rule to devise a sample load for each device for that communication round. Intuitively, devices closer to their target accuracies get a lighter sample load and vice-versa. Practically this ensures that the devices are all satisfied at roughly the same point, thus avoiding any incentives to defect. While \textsc{MW-FED} hasn't been analyzed in the context of federated learning, it is well-known that it has an optimal sample complexity for optimizing distributed learning problems where the goal is to come up with a single best model for all agents, much like problem \eqref{eq:relaxed}.

\section{Alternative Models of Rational Agent Behaviour}\label{app:rational}
This paper introduced a framework for studying and avoiding agent defections in federated learning. In addition to the rational model discussed in the paper, there are several other behaviors that rational agents might depict in practice. We discussed some of these behaviors in this section to highlight directions for future work on this important topic. 

\subsection{Incentive Compatibility and Truthfulness}
One crucial assumption we make in the paper is that the server knows the desired precision of each agent, i.e., $\{\epsilon_m\}_{m\in[M]}$ apriori. This is crucial to the design of our algorithm as we use these values to predict when an agent could defect by putting it in the set $D$ in Algorithm \ref{alg:reweight}. However, in practice, the agent might not want to reveal this information so that they can further benefit from the information asymmetry. 
For instance, in the absence of an incentive-compatible mechanism, the agents are incentivized to pretend that they require a precision $\epsilon_m'<< \epsilon_m$, so they can fool Algorithm \ref{alg:reweight} from ever putting them in the set $D$, and defect as early as possible. 
As a result, designing incentive-compatible mechanisms to elicit truthful information from agents is an intriguing future direction. 

\subsection{Underconfident Clients}
Throughout the paper, we assume that the agents can compute their objective function when they want to make the decision to leave or stay in the collaboration. However, in practice, agents might only have access to a finite sample from their data distribution, i.e., each agent $m$ samples $S_m\sim \ddd_m^{\otimes N_m}$ and then computes gradients of the following objective, $$\hat F_m(w) = \frac{1}{N_m}\sum_{z\in S_m}f(w; z).$$ This, of course, affects changes the server's ability to accurately obtain a proxy objective for $\tilde F(w) = \ee_{m\sim \ppp, z\in \ddd_m}\sb{f(w;z)}$ but more importantly the clients are likely to stay in the collaboration even if they reach their target precision $\epsilon_m$ on $\hat F_m$. This is because the agents only have a proxy objective for $F_m$, and unless $N_m=\Omega\rb{1/\epsilon_m^2}$, the agents can not be certain of attaining their required precision based on $\hat F_m$. In fact, in practice, this is a likely motivator for the agents to stay in the collaboration and benefit from the sample of other agents who might have a similar data distribution to the agent. For instance, if the algorithm eventually converges to a model in the set $S^\star$, then the agent is certain to enjoy a good model. This uncertainty about the gap between an agent's optimization and generalization error makes it under-confident about when to defect. If the client is using some generalization upper bound to such as $\hat F_m(w) + \frac{c}{\sqrt{N_m}}$, to get a better handle on its population loss, then in Algorithm \ref{alg:reweight} we can also use $\epsilon_m + \frac{c}{\sqrt{N_m}}$ to detect potentially defecting agents. The incentive compatibility reported in the previous section is further exacerbated in this section as the agents might choose to under-report the number of samples $N_m$ they actually have access to. It is an important future question to understand how the server can use this under-confidence in the agents in designing a better algorithm to disincentivize defections.  

\subsection{Using Agents' Group Information}
In several circumstances, the server knows the agents' group identity, which it can use to detect as well as disincentivize defections. For instance, assume the sampling of the agent $m\sim \ppp$ can be decomposed in two steps: first, sample a group $l$ with probability $p_l$ and then sample $m\sim G_l$, i.e., $$\tilde F(w) = \ee_{m\sim \ppp, z\sim \ddd_m}\sb{f(w;z)} = \ee_{l\sim \ggg, m\sim \ggg_l, z\sim \ddd_m}\sb{f(w;z)}.$$ 
We still assume that each agent has a precision $\epsilon_m$, which is not determined by the group identity. Then the server can predict defecting agents, even without access to their objective values. For instance, assuming the distributions within each group are related to each other much more than they are to distributions related to other groups. Then when an agent $m$ from a particular group defects, the server can speculate that the other agents from the group have also attained a precision $\epsilon_m$. In fact, this model is much more reasonable for cross-device federated learning, where group information is easily available, and the server is fine with some agents defecting because there are other agents from the group who can provide information about the group distribution. Having said that, the server needs to be able to control the influence of each group towards the final learned model and does not want defections to be overrepresented from a single group. This presents an interesting question for future study.

\subsection{Beta Users and the Public Model}
Digital service providers often have beta users with more updated models than the ones available to the general public. These beta models are then updated using the data from the beta users who participate in the training process. Intermittently, the public models are updated to these beta models, and then all users get access to the updated model. At this point, the beta users can choose to defect because even if they do, and revoke the beta model, they can get access to a public model that they are satisfied with. The server's task is harder in this setting because, unlike the problem considered in this paper, if beta users leave when they see a good beta model, they are sure they will eventually get a good public model. Usually, beta users are motivated by other reasons, such as getting the newest features most quickly, but if that is not enough of an incentive, the server has to find a means to incentivize and retain agent participation. This is another unexplored but prevalent problem in distributed learning.

\section{Missing Details from Section \ref{sec:setup-defect}}\label{app:setup}
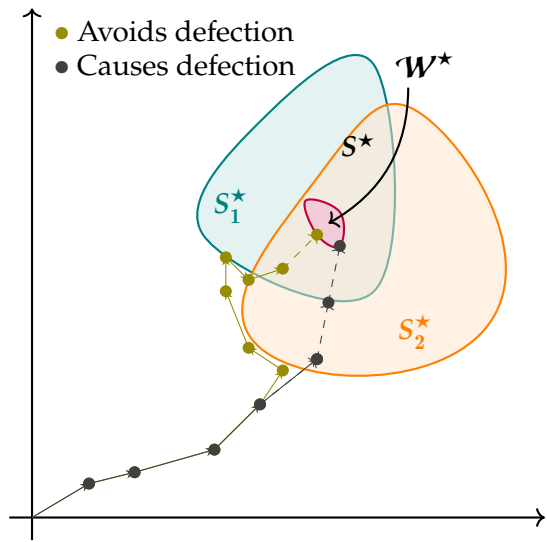
\begin{figure*}
    \centering
    \begin{tikzpicture}[scale=1.5]
  \fill[lime!0] (-0.2,-0.2) rectangle (4.7,4.7);

  \draw[->, thick] (-0.2,0) -- (4.5,0);
  \draw[->, thick] (0,-0.2) -- (0,4.5);

  \filldraw[fill=teal!20, fill opacity=0.5, draw=teal, thick] plot [smooth cycle, tension=0.7] coordinates {(1.5,2.5) (3,2) (3,4) (2,3.5)};
  \node[teal, below right] at (1.5,3) {$\boldsymbol{S_1^\star}$};

  \filldraw[fill=orange!20, fill opacity=0.5, draw=orange, thick] plot [smooth cycle, tension=1] coordinates {(2,1.5) (4,1.7) (3.5,3.5) (2.5,3)};
  \node[orange, above left] at (3.6,1.4) {$\boldsymbol{S_2^\star}$};

    \node[black, above left] at (3.1,3.1) {$\boldsymbol{S^\star}$};

  \filldraw[fill=purple!20, fill opacity=1.0, draw=purple, thick] plot [smooth cycle, tension=0.7] coordinates {(2.5,2.5) (2.7,2.4) (2.7,2.7) (2.4,2.8)};

  \foreach \x/\y/\u/\v in {0/0/0.5/0.3, 0.5/0.3/0.9/0.4, 0.9/0.4/1.6/0.6, 1.6/0.6/2.0/1.0, 2.0/1.0/2.2/1.3, 2.2/1.3/1.9/1.5, 1.9/1.5/1.7/2.0, 1.7/2.0/1.7/2.3, 1.7/2.3/1.9/2.1, 1.9/2.1/2.2/2.2}{
    \draw[->, >=stealth, color=olive] (\x,\y) -- (\u,\v);
    \fill[color=olive] (\u,\v) circle (1.5pt);
  }

  \foreach \x/\y/\u/\v in {2.2/2.2/2.5/2.5 }{
    \draw[->, dashed, color=olive] (\x,\y) -- (\u,\v);
    \fill[color=olive] (\u,\v) circle (1.5pt);
    }

  \foreach \x/\y/\u/\v in {0/0/0.5/0.3, 0.5/0.3/0.9/0.4, 0.9/0.4/1.6/0.6, 1.6/0.6/2.0/1.0, 2.0/1.0/2.5/1.4}{
    \draw[->, >=stealth, color=darkgray] (\x,\y) -- (\u,\v);
    \fill[color=darkgray] (\u,\v) circle (1.5pt);
  }
  \foreach \x/\y/\u/\v in {2.0/1.0/2.5/1.4, 2.5/1.4/2.6/1.9, 2.6/1.9/2.7/2.4}{
    \draw[->, dashed, color=darkgray] (\x,\y) -- (\u,\v);
    \fill[color=darkgray] (\u,\v) circle (1.5pt);
  }

  \node[align=left, anchor=west] at (3.1,4.0) {$\boldsymbol{\mathcal{W}^\star}$};
  \draw[->, thick] (3.3,3.8) to [bend left=30] (2.6,2.6);

  \node[align=left, anchor=west] at (0.1,4.3) {\textcolor{olive}{$\boldsymbol{\bullet}$} Avoids defection};
  \node[align=left, anchor=west] at (0.1,4.0) {\textcolor{darkgray}{$\boldsymbol{\bullet}$} Causes defection};
\end{tikzpicture}
    \caption[An illustration of two different algorithmic behaviors and agent defections.]{An illustration of two different algorithmic behaviors and agent defections. The $\epsilon$-sublevel sets of two agents are denoted by $S_1^\star$ and $S_2^\star$, and the intersection of their optima is denoted by $\www^\star$. We note that the olive and gray algorithms are convergent, i.e., if the agents were irrational, both algorithms would converge to $\www^\star$ as indicated by the dashed trajectories. For rational agents, both algorithms will cause defections, but the defection with the olive algorithm is benign/not harmful, as the algorithm already converges to the target intersection $S^\star = S_1^\star\cap S_2^\star$ before any agent defects. On the other hand, the gray algorithm causes a harmful defection by causing agent two to defect by entering the set $S_2^\star$ before entering the set $S^\star$. 
    }
    \label{fig:intersection}
\end{figure*}

We include the pseudo-code for FedAvg in Algorithm~\ref{alg:fedavg} and an illustration of two different algorithmic behaviors and agent defections in Figure~\ref{fig:intersection} in this section.

\begin{algorithm}[h]  
\caption{\textsc{FedAvg} w/o any defections}\label{alg:fedavg}
\textbf{Input:} step size $\eta$\\
\textbf{Initialize} $w_0^m=w_0=0$ on all agents $m\in[M]$\\

 \begin{algorithmic}[1]
     \FOR{$r\in\{1, \dots, R\}$}{
		      \FOR{$m \in [M]$ \textbf{in parallel}}{
		        \STATE Set $w_{r,1}^m = w_{r-1}$\\
                    \FOR{$k\in\{1, \dots, K\}$}{
                        \STATE Sample $z^m_{r,k}\sim \ddd_m$\\          
                        \STATE Compute stochastic gradient at $w_{r,k}^m$, $g_{r,k}^m \gets \nabla f(w_{r,k}^m; z_{r,k}^m)$\\  
                        \STATE Update $w_{r, k+1}^{m} \gets w_{r,k}^m - \frac{\eta}{K_r^m} g_{r,k}^m$
                        }
                    \ENDFOR
                    \STATE \textcolor{blue}{\textbf{Communicate to server:}} $w_r^m \gets w_{r, K_r^m+1}^m = w_{r-1} - \frac{\eta}{K_r^m}\sum_{k\in[K_r^m]}g_{r,k}^m$ 

                }
                \ENDFOR
		     \STATE \textcolor{blue}{\textbf{Communicate to agents:}} $w_r \gets \frac{1}{M}\sum_{m\in[M]}\cdot w_r^m$ 
            }
            \ENDFOR
 \end{algorithmic}
 \textbf{Output:} Return $w_{R}$
\end{algorithm}

\section{Missing Details from Section \ref{sec:alg}}\label{app:alg}
\subsection{Proof of Theorem \ref{thm:two_agents}}
\begin{proof}
We will show that algorithm \ref{alg:reweight} with a small enough step-size sequence,
\begin{enumerate}
    \item is well-defined, i.e., at time $t$, if we are in case  1, we have $\norm{\Pi_P\rb{\nabla F_{ND}(w_{t-1})}}\neq 0$; if we are in case 2, we have $\norm{\nabla F(w_{t-1})}\neq 0$. 
    \item will not cause any agent to defect, and
    \item will terminate and output a model $\hat{w}$ which is approximately optimal.
\end{enumerate}
For the first property, we have the following lemma.
\begin{lemma}[Updates are well-defined]\label{lem:legal}
    Under the same conditions of Theorem \ref{thm:two_agents}. Suppose the algorithm is in case 1 or case 2 at any time step $t$. If we are in case  1, we have $\norm{\Pi_P\rb{\nabla F_{ND}(w_{t-1})}}\neq 0$; if we are in case 2, we have $\norm{\nabla F(w_{t-1})}\neq 0$.
\end{lemma}
 
Now to see the second property, note that defections can only happen if (i) the algorithm runs into case 1 or 2, (ii) it makes the corresponding update, and (iii) the agent chooses to defect after seeing the updated model.
\begin{lemma}[No agent will defect in case 1 and case 2]\label{lem:case1a_2a_defect}
    Under the same conditions as in Theorem \ref{thm:two_agents}. Suppose the algorithm is in case 1 or case 2 at any time step $t$, and no agent has defected up to time step $t$, i.e., after receiving model $w_{t-1}$. If $\eta \leq  \sqrt{\delta/(2H)}$, no defection will occur once the update is made. 
\end{lemma}
The reason that no agent defects in case 1 is that we create our update direction in case 1 so that for all agents in $D$, the update is orthogonal to their current gradient, and thus they don't reduce their objective value. And for all agents in $ND$, they do make progress, but we control the step size so that they don't make \textit{``too much progress"}. Thus, no agent defects in case 1. Similarly, in case 2 we avoid defections by ensuring the step size is small enough. 

Finally, we show that the algorithm will terminate and the returned model is good.
\begin{lemma}[The algorithm will terminate]\label{lem:termination}
    Under the same conditions of Theorem \ref{thm:two_agents}, assuming $\eta \leq \frac{1}{MH}$ Algorithm~\ref{alg:reweight} terminates.
\end{lemma}
To prove that the algorithm terminates, we first show that the algorithm makes non-zero progress on the average objective every time it is in case 1 and case 2 (which is shown in lemma \ref{lem:case1a_2a_progress}), which implies that the average loss will converge. Then we prove that the average loss can only converge to zero, and thus, we will certainly get into case 3.
\begin{lemma}[The returned model is good]\label{lem:case1b_2b_3}
    Under the same conditions as in Theorem \ref{thm:two_agents}. Suppose Algorithm~\ref{alg:reweight} terminates in case 3 at time step $t$, and no agent has defected up to time step $t$. If $\eta \leq \min\cb{\sqrt{2\delta/H}, \delta/L}$, then the algorithm will output $\hat w$ that satisfies $F(\hat w) \leq \epsilon + 2\delta$.
\end{lemma}

Thus, by combining the three properties, we can conclude that the algorithm outputs a good model and avoids defections. This finishes the proof.  
\end{proof}

\subsection{Proof of Lemma~\ref{lem:legal}}
First, consider \textbf{case 2}. Suppose that $\norm{\nabla F(w_{t-1})}= 0$. If there exists $n\in [M]$ s.t. $F_n(w_{t-1}) > F_n(w^*)=0$, then we have
\begin{align*}
     F(w^\star) &\geq F(w_{t-1}),\\ 
    &= \frac{1}{M}\sum_{m\in[M]}F_m(w_{t-1}),\\
    &\geq \frac{1}{M}\sum_{m\neq n}F_m(w^\star) + \frac{F_n(w_{t-1})}{M},\\
    &> \frac{1}{M}\sum_{m\neq n}F_m(w^\star) + \frac{F_n(w^\star)}{M},\\
    &= F(w^\star),
    \end{align*}
    which is a contradiction. Hence we have $F_n(w_{t-1})=0$ for all $n\in [M]$, which contradicts with the condition that $D$ is empty due to line 3 of Algorithm~\ref{alg:reweight}.

Next, consider \textbf{case 1}. We first introduce the following lemma.
\begin{lemma}\label{lem:ass_hetero}
    Suppose $\{F_m\}\in \fff_{hetero}$. For all $A\subset [M]$ and $B = [M] \setminus A$, if $F_A(\cdot):= \sum_{m\in A}F_m(\cdot)$, then for all $w\in \www$, such that $\nabla F_A(w)\neq 0$, $\nabla F_A(w)\notin Span\{\nabla F_m(w): m\in B\}$.
\end{lemma}
\begin{proof}
    Consider some point $w\in\www\setminus \www^\star$ and let the non-zero gradients on the agents be linearly independent at that point. If possible, let the above property be violated, then we have at least one $A\subseteq [M]$ such that $\nabla F_A(w) \in Span\{\nabla F_m(w): m\in B\}$ and $\nabla F_A\neq 0$. In particular there are coefficients $\{\gamma_n\in \rr\}_{n\in B}$ (not all zero) such that,  
    \begin{align*}
        &\sum_{m\in A}\nabla F_m(w) = \sum_{n\in B}\gamma_n \nabla F_n(w),\\
        &\Leftrightarrow \sum_{m\in A}\nabla F_m(w)  + \sum_{n\in B}(-\gamma_n) \nabla F_n(w) = 0,
    \end{align*}
    This implies that the gradients are linearly dependent (note that not all gradients can be zero as then $w$ would be in $\www^\star$), which is a contradiction.

By combining definition \ref{def:hetero} of $\fff_{hetero}$ and lemma~\ref{lem:ass_hetero}, we know that $\norm{\Pi_P\rb{\nabla F_{ND}(w_{t-1})}}= 0$ implies that $\norm{\nabla F_{ND}(w_{t-1})}= 0$.  This implies that $F_{ND}(w_{t-1}) = F_{ND}(w^\star)=0$ for $w^\star\in \www^\star$, which contradicts with the definition of ND (see line 4 of Algorithm~\ref{alg:reweight}). This concludes the proof of Lemma \ref{lem:legal} showing the updates are well-defined in both cases 1 and 2. 
\end{proof}

\subsection{Proof of Lemma \ref{lem:case1a_2a_defect}}

We begin with introducing the following lemma.
\begin{lemma}[Predicting defections with single first-order oracle call]\label{lem:defection_detect}
    Under the same conditions of Theorem \ref{thm:two_agents}, at any time step $t$ assuming $\eta \leq \sqrt{2\delta/H}$,
    \begin{itemize}
        \item if agent $m\in D$ then $F_m(w_{t-1} - \eta\ngrad{m}(w_{t-1}))\leq \epsilon_m + 2\delta$, and
        \item if agent $m\in ND$ then $F_m(w_{t-1}~-~\eta\ngrad{m}(w_{t-1}))~>~\epsilon_m +\delta$,
    \end{itemize}
    where $\ngrad{m}(w_{t-1}) = \frac{\nabla F_{m}(w_{t-1})}{\norm{\nabla F_{m}(w_{t-1})}}$ is the normalized gradient.
\end{lemma}
\begin{proof}
    Let's say we are at time step $t$. Assume $m\in D$ and note that the smoothness of function $F_m$ implies that,
    \begin{align*}
        F_m(w_{t-1} - \eta_t\ngrad{m}(w_{t-1})) &\leq F_m(w_{t-1}) + \inner{\nabla F_m(w_{t-1})}{-\eta_t\ngrad{m}(w_{t-1})} + \frac{H}{2}\norm{\eta_t\ngrad{m}(w_{t-1})}^2,\\ 
        &= F_m(w_{t-1}) -\eta_t\norm{\nabla F_m(w_{t-1})} + \frac{H\eta_t^2}{2},\\
        &\leq^{(m\in D)} \epsilon_m + \delta + \frac{H\eta_t^2}{2},\\
        &\leq \epsilon_m + 2\delta,
    \end{align*}
    where we used that $\eta_t \leq \sqrt{\frac{2\delta}{H}}$. Now assume $m\in ND$ and note using convexity of $F_m$ that,
    \begin{align*}
        F_m(w_{t-1} - \eta_t\ngrad{m}(w_{t-1})) &\geq F_m(w_{t-1}) + \inner{\nabla F_m(w_{t-1})}{-\eta_t\ngrad{m}(w_{t-1})},\\ 
        &= F_m(w_{t-1}) -\eta_t\norm{\nabla F_m(w_{t-1})},\\
        &>^{(m\in ND)} \epsilon_m +\delta.
    \end{align*}
    This proves the lemma.
\end{proof}

Now we are ready to prove lemma~\ref{lem:case1a_2a_defect}.
First, assume we are in case 1 at time $t$. Let's first show that we don't make any agent $m\in D$ defect,
\begin{align*}
    F_m(w_t) &\geq F_{m}(w_{t-1}) + \inner{\nabla F_{m}(w_{t-1})}{g_t},\\
    &= F_{m}(w_{t-1}),\\
    &> \epsilon_m,
\end{align*}
as $g_t\perp \nabla F_{m}(w_{t-1})$ for all $m\in D$ by design and no agent defected up to time $t$, i.e., after receiving model $w_{t-1}$. For any non-defecting agent $m\in ND$ we have,
    \begin{align*}
        &F_{m}(w_{t-1} + \eta_t g_t)-F_{m}(w_{t-1} - \eta_t \ngrad{m}(w_{t-1})) \\
        &\geq^{\text{(convexity)}}\inner{\nabla F_{m}(w_{t-1} - \eta_t \ngrad{m}(w_{t-1}))}{\eta_t(g_t +\ngrad{m}(w_{t-1}))},\\
        &= \inner{\nabla F_{m}(w_{t-1} - \eta_t \ngrad{m}(w_{t-1})) -\nabla F_{m}(w_{t-1}) + \nabla F_{m}(w_{t-1})}{\eta_t(g_t +\ngrad{m}(w_{t-1}))},\\
        &\geq^{\text{(C.S. inequality)}} \inner{\nabla F_{m}(w_{t-1} )}{\eta_t(g_t +\ngrad{m}(w_{t-1}))}\\
        &\qquad - \norm{\nabla F_{m}(w_{t-1}) - \nabla F_{m}(w_{t-1} - \eta_t \ngrad{m}(w_{t-1}))}\cdot \norm{\eta_t(g_t +\ngrad{m}(w_{t-1}))},\\
        &\geq^{\text{(Ass. \ref{ass:cvx_smth})}} \inner{\nabla F_{m}(w_{t-1} )}{\eta_t(g_t +\ngrad{m}(w_{t-1}))} - \eta_t^2 H\norm{\ngrad{m}(w_{t-1})}\cdot \norm{g_t +\ngrad{m}(w_{t-1})},\\
        &\geq^{\text{(normalized gradients)}} \inner{\nabla F_{m}(w_{t-1} )}{\eta_t(g_t +\ngrad{m}(w_{t-1}))} - 2\eta_t^2H,\\
        &= \eta_t\norm{\nabla F_{m}(w_{t-1} )}\rb{1 + \inner{\ngrad{m}(w_{t-1})}{g_t}} - 2\eta_t^2H,\\
        &\geq - 2\eta_t^2H,
    \end{align*}
    Re-arranging this gives the following,
    \begin{align*}
        F_{m}(w_{t-1} + \eta_t g_t) &\geq F_{m}(w_{t-1} - \eta_t \ngrad{m}(w_{t-1})) - 2\eta_t^2H\\
        &>^{(\text{lemma \ref{lem:defection_detect}})} \epsilon_m + \delta - 2\eta_t^2H,\\
        &\geq \epsilon_m, 
    \end{align*}
    where we assume that $\eta_t \leq \sqrt{\frac{\delta}{2H}}$.  

    Now assume instead we are in case 2 at time $t$. For agent $m\in[M]$ we have,
    \begin{align*}
        &F_{m}(w_{t-1} + \eta_t g_t)-F_{m}(w_{t-1} - \eta_t \ngrad{m}(w_{t-1})) \\
        &\geq^{\text{(convexity)}}\inner{\nabla F_{m}(w_{t-1} - \eta_t \ngrad{m}(w_{t-1}))}{\eta_t(g_t +\ngrad{m}(w_{t-1}))},\\
        &= \inner{\nabla F_{m}(w_{t-1} - \eta_t \ngrad{m}(w_{t-1})) -\nabla F_{m}(w_{t-1}) + \nabla F_{m}(w_{t-1})}{\eta_t(g_t +\ngrad{m}(w_{t-1}))},\\
        &\geq^{\text{(C.S. inequality)}} \inner{\nabla F_{m}(w_{t-1} )}{\eta_t(g_t +\ngrad{m}(w_{t-1}))}\\
        &\qquad - \norm{\nabla F_{m}(w_{t-1}) - \nabla F_{m}(w_{t-1} - \eta_t \ngrad{m}(w_{t-1}))}\cdot \norm{\eta_t(g_t +\ngrad{m}(w_{t-1}))},\\
        &\geq^{\text{(Ass. \ref{ass:cvx_smth})}} \inner{\nabla F_{m}(w_{t-1} )}{\eta_t(g_t +\ngrad{m}(w_{t-1}))} - \eta_t^2 H\norm{\ngrad{m}(w_{t-1})}\cdot \norm{g_t +\ngrad{m}(w_{t-1})},\\
        &\geq \inner{\nabla F_{m}(w_{t-1} )}{\eta_t(g_t +\ngrad{m}(w_{t-1}))} - 2\eta_t^2H,\\
        &= \eta_t\norm{\nabla F_{m}(w_{t-1})}\rb{1 + \inner{\ngrad{m}(w_{t-1})}{g_t}} - 2\eta_t^2H,\\
        &\geq - 2\eta_t^2H.
    \end{align*}
    Choosing $\eta_t \leq \sqrt{\frac{\delta}{2H}}$ ensures that, 
    \begin{align*}
        F_{m}(w_{t}) &>^{\text{(lemma \ref{lem:defection_detect})}} \epsilon_m + \delta - 2\eta_t^2H,\\
        &> \epsilon_m.
    \end{align*}
    and thus agent $m$ doesn't defect.

\subsection{Proof of Lemma~\ref{lem:termination}}

We first introduce the following lemma.
\begin{lemma}[Progress in case 1 and 2]\label{lem:case1a_2a_progress}
    Under the same conditions as in Theorem \ref{thm:two_agents}. Suppose the algorithm is in case 1 or 2 at any time step $t$, and no agent has defected up to time step $t$. If $\eta \leq 1/(MH)$, then 
    $$F(w_t) < F(w_{t-1}) -\frac{\eta_t}{2}\min(\norm{\nabla F(w_{t-1})}^2,1)$$ in case 2 and $$F(w_t) < F(w_{t-1}) -\frac{\eta_t}{2M}\min(\norm{\Pi_P\rb{\nabla  F_{ND}(w_{t-1})}}^2,1)$$ in case 1, where we recall that $P = Span\{\nabla F_n(w_{t-1}): n\in D\}^\perp$.
\end{lemma}
\begin{proof}
We first assume we make an update in \textbf{case 1}. First, using the smoothness assumption and then using the fact that $g_t$ is orthogonal to the gradients of all the agents in $D$, we get
    \begin{align*}
        F(w_t) &= F(w_{t-1}+\eta g_t),\\
        &\leq^{\text{(ass. \ref{ass:cvx_smth})}} F(w_{t-1}) + \eta \inner{\frac{\sum_{m\in D}\nabla F_{m}(w_{t-1}) + \nabla F_{ND}(w_{t-1})}{M}}{g_t} + \frac{H\eta^2}{2} \norm{g_t}^2,\\
        &\leq F(w_{t-1}) +\frac{\eta}{M} \inner{\nabla F_{ND}(w_{t-1})}{g_t} +\frac{H\eta^2}{2} \norm{g_t}^2,\\
    \end{align*}
    Now we will consider two cases. In the first case, assume $\norm{\Pi_P\rb{\nabla  F_{ND}(w_{t-1})}}< 1$. Since we can write $\nabla F_{ND}(w_{t-1}) = \Pi_P\rb{\nabla  F_{ND}(w_{t-1})} + \Pi_{P^\perp}\rb{\nabla  F_{ND}(w_{t-1})}$ we get that,
    \begin{align}
        F(w_t)&\leq F(w_{t-1}) -\frac{\eta}{M} \inner{\nabla F_{ND}(w_{t-1})}{\Pi_P\rb{\nabla F_{ND}(w_{t-1})}} +\frac{H\eta^2}{2}\norm{\Pi_P\rb{\nabla  F_{ND}(w_{t-1})}}^2,\nonumber\\
        &= F(w_{t-1}) -\frac{\eta}{M}\norm{\Pi_P\rb{\nabla  F_{ND}(w_{t-1})}}^2 +\frac{H\eta^2}{2} \norm{\Pi_P\rb{\nabla  F_{ND}(w_{t-1})}}^2,\nonumber\\
        &\leq F(w_{t-1}) -\frac{\eta}{2M}\norm{\Pi_P\rb{\nabla  F_{ND}(w_{t-1})}}^2,\label{eq:progress-case-1}
    \end{align}
    where we assume $\eta \leq \frac{1}{M H}$. Since in case 1, $\norm{\Pi_P\rb{\nabla F_{ND}(w_{t-1}))}}\neq 0$ (recall definition \ref{def:hetero}), we will provably make progress on the average objective. In the second case, assume $\norm{\Pi_P\rb{\nabla F_{ND}(w_{t-1})}}\geq 1$. Then noting that $\inner{\nabla F_{ND}(w_{t-1})}{g_t} = -\norm{\Pi_P\rb{\nabla F_m(w_{t-1})}}$ and $\norm{g_t}=1$ we will get that,                        
    \begin{align*}
        F(w_t)&\leq F(w_{t-1}) -\frac{\eta_t}{M} \norm{\Pi_P\rb{\nabla F_m(w_{t-1})}} + \frac{H\eta_t^2}{2},\\
        &\leq F(w_{t-1}) -\frac{\eta_t}{M} +\frac{H\eta_t^2}{2},\\
        &\leq F(w_{t-1}) -\frac{\eta_t}{2M},
    \end{align*}
    where we assume $\eta_t \leq \frac{1}{M H}$. This finishes the proof.

    Next, we assume we make an update in \textbf{case 2}. Note that using smoothness,
    \begin{align*}
        F(w_{t}) &= F(w_{t-1} + \eta_t g_t ),\\
        &\leq F(w_{t-1}) +\eta_t \inner{\nabla F(w_{t-1})}{g_t} + \frac{H\eta_t^2}{2}\norm{g_t}^2.
    \end{align*}
    Let's first assume $\norm{\nabla F(w_{t-1})}\geq 1$. Using the definition of $g_t$ we get,
    \begin{align*}
        F(w_{t}) &\leq F(w_{t-1}) -\eta_t \inner{\nabla F(w_{t-1})}{\frac{\nabla F(w_{t-1})}{\norm{\nabla F(w_{t-1})}}} + \frac{H\eta_t^2}{2}\norm{g_t}^2,\\
        &= F(w_{t-1}) -\eta_t \norm{\nabla F(w_{t-1})} + \frac{H\eta_t^2}{2},\\
        &\leq F(w_{t-1}) -\eta_t + \frac{H\eta_t^2}{2},\\
        &\leq F(w_{t-1}) -\frac{\eta_t}{2},  
    \end{align*}
    where we assume $\eta_t \leq \frac{1}{H}$. Now let's consider the case when $\norm{\nabla F(w_{t-1})}< 1$.
    \begin{align}
        F(w_{t}) &\leq F(w_{t-1}) -\eta_t \inner{\nabla F(w_{t-1})}{\nabla F(w_{t-1})} + \frac{H\eta_t^2}{2}\norm{\nabla F(w_{t-1})}^2,\nonumber\\
        &= F(w_{t-1}) -\eta_t \norm{\nabla F(w_{t-1})}^2 + \frac{H\eta_t^2}{2}\norm{\nabla F(w_{t-1})}^2,\nonumber\\
        &\leq F(w_{t-1}) -\frac{\eta_t}{2}\norm{\nabla F(w_{t-1})}^2, \label{eq:progress-case-2} 
    \end{align}
    where we assume $\eta_t \leq \frac{1}{H}$. Note that $\norm{\nabla F(w_{t-1})}\neq 0$ in this case, which means the algorithm makes non-zero progress on the average objective.
\end{proof}
We are ready to prove Lemma \ref{lem:termination}. Let's assume that the algorithm doesn't terminate, i.e., it does not enter case 3. Then Lemma \ref{lem:case1a_2a_defect} implies that for all time steps $t$, $w_{t-1}\notin S_m^\star$ for all $m\in[M]$ as no machine defects in cases 1 and 2. This implies that for all time steps $t$, $\norm{\nabla F_m(w_{t-1})}>0$ for all $m\in[M]$. Along with Lemma~\ref{lem:case1a_2a_progress}, this means that the algorithm makes non-zero progress on the average loss every time it is in case 1 and case 2. Thus, the sequence $F(w_t)$ is strictly monotonic decreasing. Since we know that the average function $F$ is bounded from below by $F(w^\star)=0$, the monotone convergence theorem implies that $F(w_t)$ must converge. 

If $F(w_t)$ converges to $0$, then the iterate must enter the set $S^\star$. However, for this to happen, the algorithm must enter case 3, which is a contradiction. Now suppose the average loss sequence $F(w_t)$ will converge to $F(w^*)+ v =v$ for some $v>0$. This would, for any sub-sequence, $F(w_{i_t})$ converges to $v$. We not the following about two sub-sequences in particular
\begin{enumerate}
    \item Consider the sub-sequence where the algorithm is in case 2. Recall that smoothness of $F$ implies that $\norm{\nabla F(w_{t-1})}^2\leq 2H\rb{F(w_{t-1})-\min_{w^\star} F(w^\star)}$. Therefore, at some time step $t_0$ in the sub-sequence $\norm{\nabla F(w_{t_0})}^2$ must be smaller than $1$. After this time step, we will progress scaling with $\|\nabla F(w_{t-1})\|_2^2$. Since the sub-sequence converges, this implies that $\|\nabla F(w_{t-1})\|_2$ must also converge to zero eventually. Thus applying convexity of $F$ we get that $$F(w_{t-1}) \leq F(w^*)+ \nabla F(w_{t-1}) ^\top (w_{t-1} - w^\star) \to F(w^\star) = 0.$$ This is a contradiction, as this would imply an agent must defect or the algorithm must enter case 3! 
    \item For any subset $S$ of $[M]$, for the sub-sequence, where the algorithm is in case 1, and the predicted non-defecting set is $ND=S$, we will eventually make progress scaling with $\|\Pi_P\left(\nabla F_{ND}(w_{t-1})\right)\|_2^2$. It implies that $\|\Pi_P\left(\nabla F_{ND}(w_{t-1})\right)\|_2^2$ would converge to zero for $t$ in this sub-sequence. There are two possible cases:
\begin{itemize}
    \item Let $\|\nabla F_{ND}(w_{t-1})\|_2^2$ also converge to zero for this sub-sequence. In this case again applying convexity of $F_{ND}$ we get that, $$F_{ND}(w_{t-1}) \leq F_{ND}(w^\star)+ \nabla F_{ND}(w_{t-1}) ^\top (w_{t-1} - w^\star) \to F(w^\star) = 0$$ converges to zero. This is again a contradiction, as no agents can defect, or the algorithm can enter case 3.
    \item Let $\|\nabla F_{ND}(w_{t-1})\|_2^2$ not converge to zero for this sub-sequence. Then this would violate the minimal heterogeneity condition stated in definition \ref{def:hetero}, as everywhere outside the set $\mathcal{W}^\star$, $\nabla F_{ND}(w_{t-1})$ must have a non-zero component in $P$.
\end{itemize}
\end{enumerate}
Therefore, the algorithm can not converge to a point with a function value $F(w^*)+ v =v$ and $v>0$. This implies that the initial assumption that the algorithm doesn't terminate is not feasible. We are done with the proof of Lemma \ref{lem:termination}.


\subsection{Proof of Lemma \ref{lem:case1b_2b_3}}

 Let's say that Algorithm~\ref{alg:reweight} terminates in Case 3 at time $t$ then we have for all $m \in D = [M]$,
\begin{align*}
    \epsilon_m + \delta &\geq^{(\text{Case 3})} F_{m}(w_{t-1})- \eta_t \nabla F_{m}(w_{t-1}),\\ 
    &\geq^{(Ass. \ref{ass:lip})} F_{m}(w_{t-1}) -\eta_tL.
\end{align*}
Assuming $\eta_t \leq \frac{\delta}{L}$ we get that for all $m\in[M]$,
\begin{align*}
    F_m(w_{t-1})  &\leq \epsilon _m+ 2\delta,
\end{align*}
which proves the claim.

\begin{figure}[tbh]
    \centering
    \includegraphics[width=\textwidth]{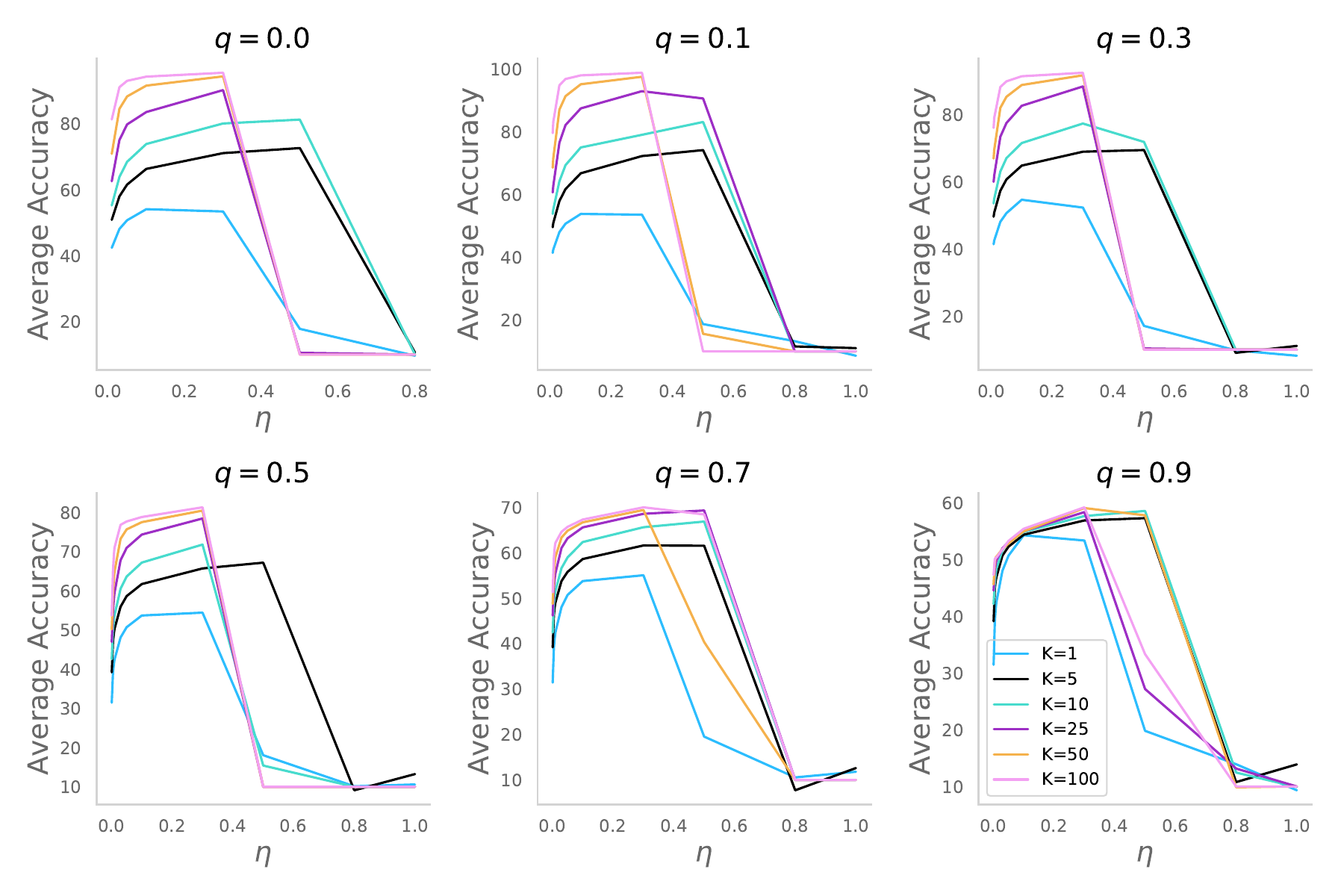}
    \caption{Fine-tuning the step-size $\eta$ for different data heterogeneity $q$ (across different plots) and the number of local update steps $K$ (different curves in each plot). The required precision $\epsilon =0$ during the fine-tuning phase.}
    \label{fig:fine_tune}
\end{figure}

\section{More Details on Experiments}\label{sec:exp}

\paragraph{Generating Data with Heterogeneity $q$.}  Denote the dataset $\ddd = \{\ddd_1,\cdots, \ddd_n\}$ where $n$ is the number of devices. To create a dataset with heterogeneity $q \in [0,1]$ for every device, we first pre-process the dataset of every device such that $|\ddd_1|=\cdots=|\ddd_n|$. Then for every device $i$, we let that device keep $(1-q)\cdot \ddd_i$  samples from their own dataset and generate a union dataset $\hat{\ddd}$ with the remaining samples from all devices, i.e. $\hat{\ddd} = q \cdot \ddd_1 \cup \cdots \cup q \cdot \ddd_n$. We use $q \cdot \ddd_i$ to denote a random split of $q$ portion from the dataset $\ddd_i$. Finally, the data with heterogeneity $q$ for every device $i$ is generated by 
\[
\hat{\ddd}_i = (1-q)\cdot \ddd_i \cup \frac{1}{n} \cdot \hat{\ddd}.
\]

\begin{figure}[tbh]
    \centering
    \includegraphics[width=0.5\textwidth]{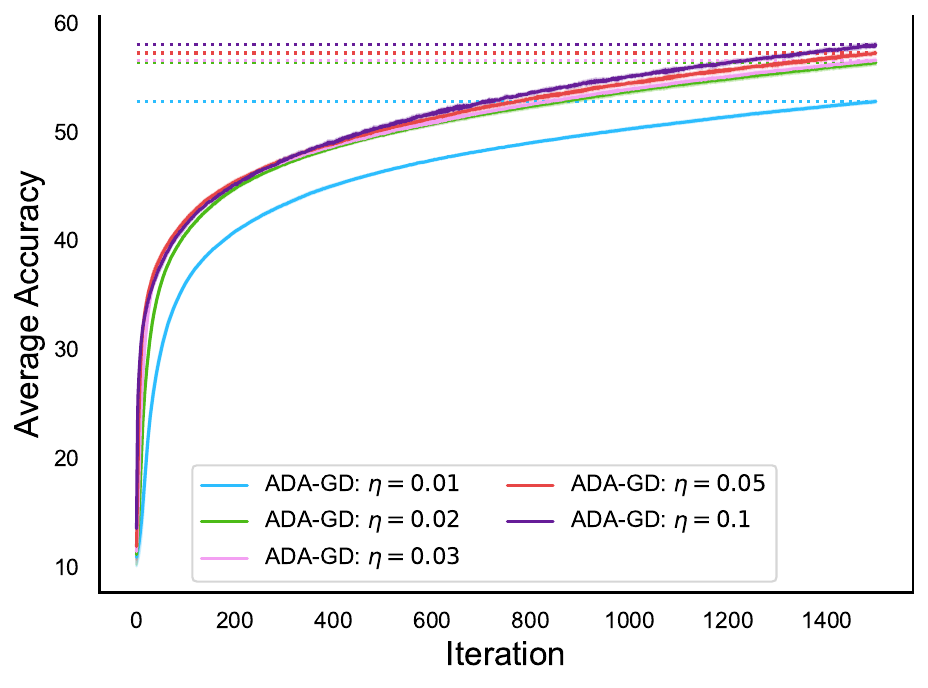}
    \caption{Additional findings on the performance of our method (ADA-GD) under varying step sizes. All experiments report accuracy averaged across 10 runs, along with error bars for $95\%$ confidence level.}
    \label{fig:our_algorithm_appendix}
\end{figure}

\begin{figure}[tbh]
    \centering
    \includegraphics[width=0.75\textwidth]{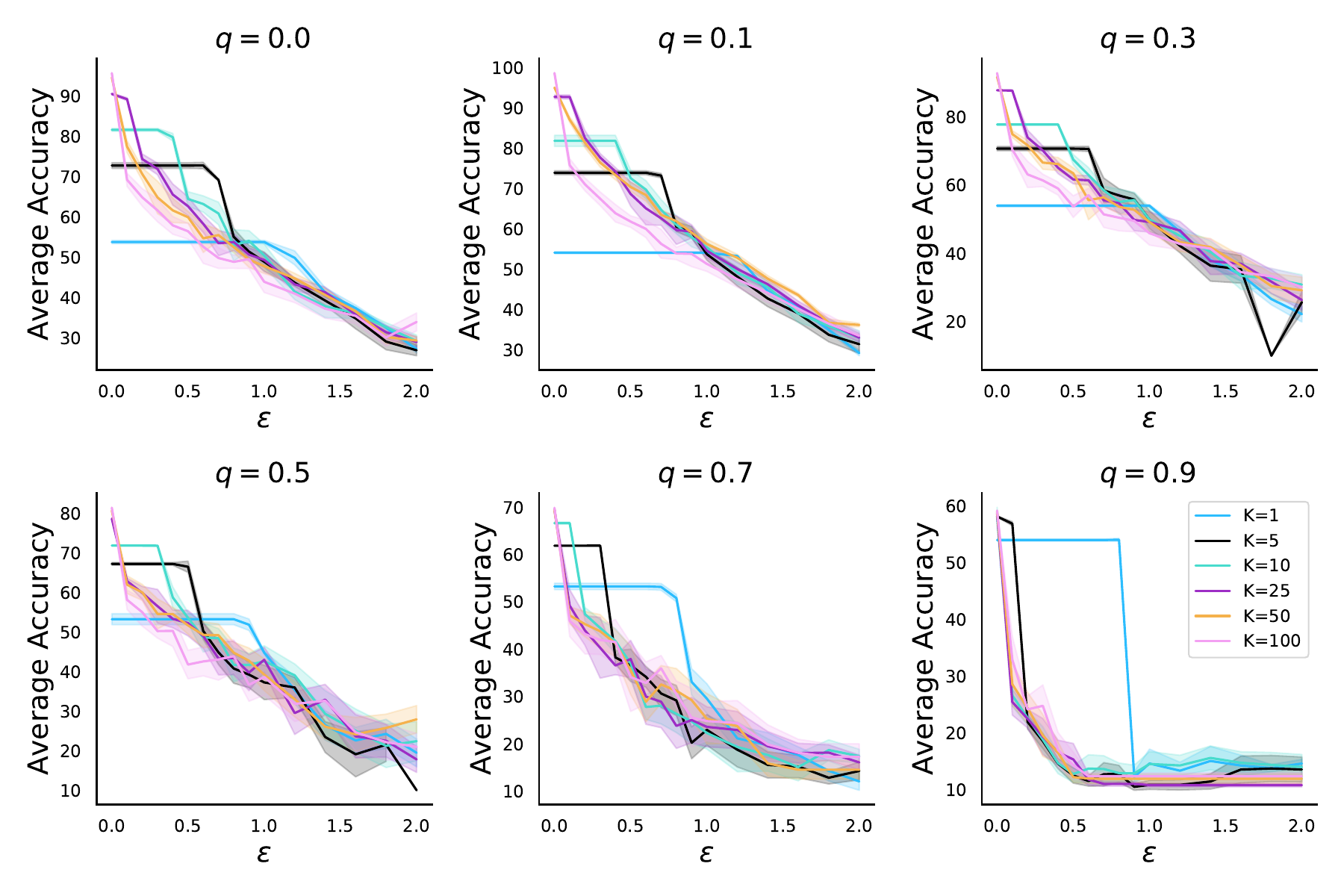}
    \caption{Additional findings on the effect of defection on average accuracy}
    \label{fig:defection_train_full}
\end{figure}

\begin{figure}[tbh]
    \centering
    \includegraphics[width=0.75\textwidth]{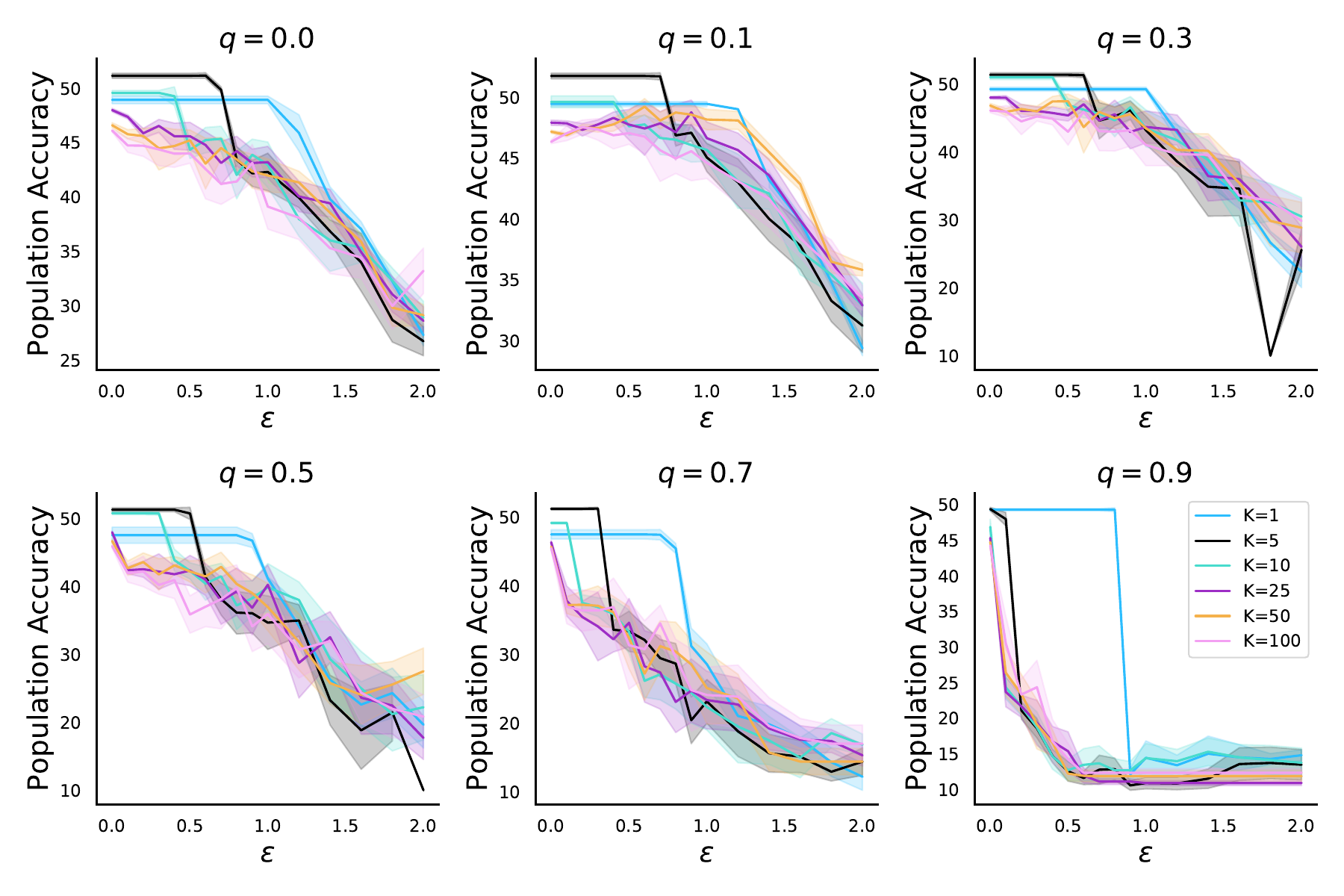}
    \caption{Additional findings on the effect of defection on population accuracy}
    \label{fig:defection_test_full}
\end{figure}

\begin{figure}[tbh]
    \centering
    \includegraphics[width=0.75\textwidth]{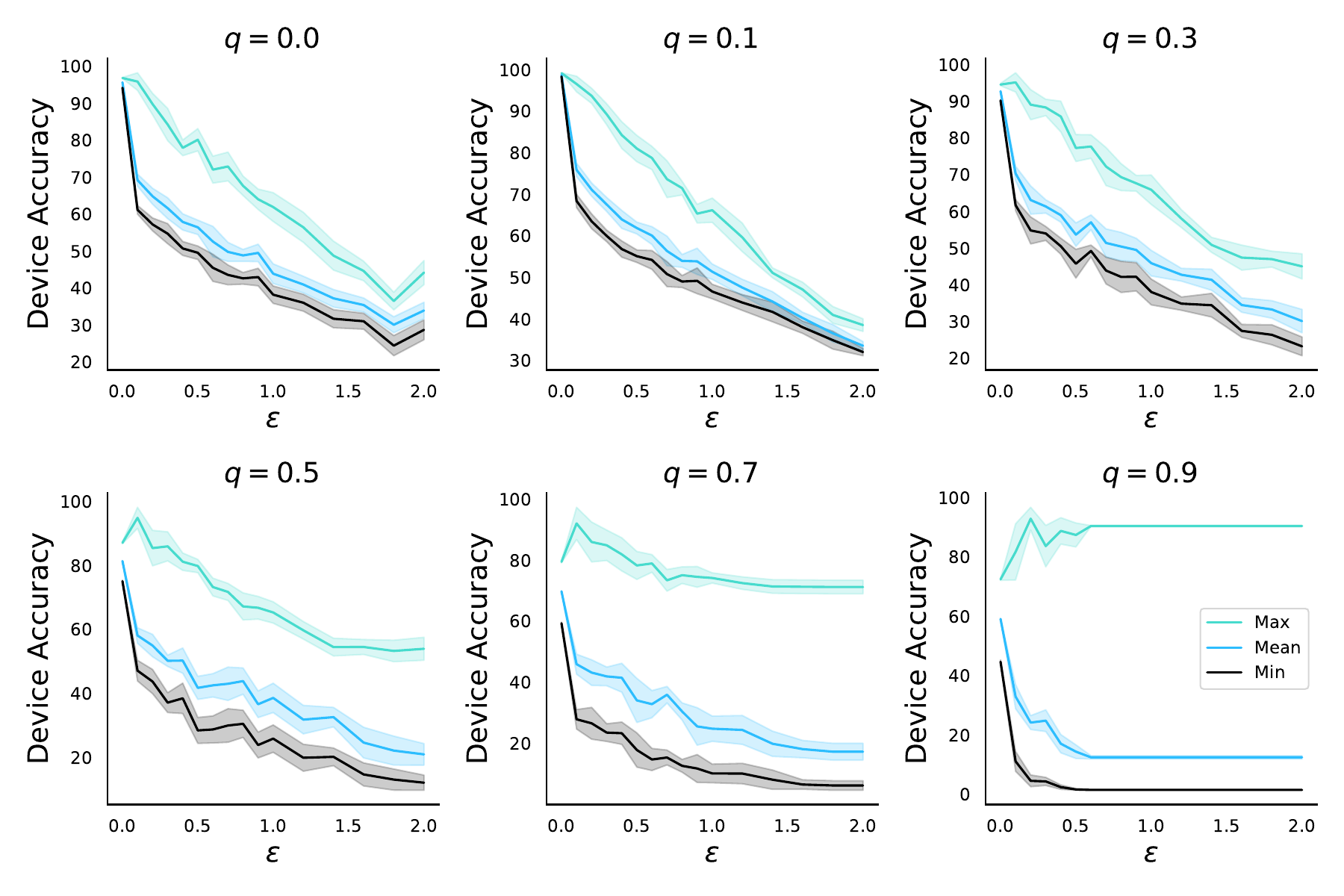}
    \caption{Additional findings on the effect of defection on the min, mean, and max device accuracies}
    \label{fig:defection_worker_full}
\end{figure}

\paragraph{Additional Experimental Results.} We present our fine-tuning process for finding the step size for different settings in Figure \ref{fig:fine_tune}. We also present the performance of our method (ADA-GD) under varying step sizes in Figure \ref{fig:our_algorithm_appendix}. In Figures \ref{fig:defection_train_full} ---\ref{fig:defection_worker_full}, we also present additional findings with more variations in data heterogeneity $q$ and the number of local update steps $K$ besides the results we presented in Figure \ref{fig:defection_worker} of the main paper. 

\chapter{Incentives in Collaborative Active Learning}\label{app:incentives-active}
\section{Proof of Theorem~\ref{thm:opt-ir}}\label{app:opt-ir}
\begin{proof}
    For any randomized optimal algorithm $\OPT$, every realization of the internal randomness of $\OPT$ must have the same query complexity and be optimal.
    Otherwise, there exists a realization with query complexity smaller than $\OPT$, which conflicts with that $\OPT$ is optimal. 
    Therefore, it suffices to prove the theorem for deterministic optimal collaborative algorithms.
    
    We will prove the theorem for deterministic algorithms by contradiction.
    Suppose that there exists a deterministic optimal collaborative algorithm $\OPT$ that is not individually rational w.r.t. running itself as baseline. Hence, there exists an agent $i\in [k]$ such that $Q_{i}(\OPT,\pi, \xset) > Q(\OPT, \pi, \{X_i\})$.   
    In this case, we can construct another algorithm $\cA'$ with smaller label complexity, which will contradict  the optimality of $\OPT$.
    
    The basic idea of $\cA'$ is to run $\OPT$ over $(\pi,\{X_i\})$ first and to recover the labels of $X_i$. 
    Then, $\cA'$ simulates $\OPT(\pi, \xset)$ and asks $\OPT(\pi, \xset)$ what point to query. But whenever $\OPT(\pi, \xset)$ asks to query the label of some point in $X_i$, since we already know the labeling of $X_i$, we can just feed $\OPT(\pi, \xset)$ with these labels without actually asking agent $i$ to query them. 
    
    Thus, the label complexity of $\cA'$ is
    \begin{align*}
        Q(\cA',\pi,\xset) &= Q(\OPT, \pi, \{X_i\}) + \sum_{j:j\neq i}Q_j(\OPT, \pi, \xset)\\
        &< Q_{i}(\OPT,\pi, \xset) + \sum_{j:j\neq i}Q_j(\OPT, \pi, \xset) \\
        &= Q(\OPT, \pi, \xset)= Q^*(\pi, \xset)\,,
    \end{align*}
    where the first inequality holds due to that $\OPT$ is not IR and the last equality holds since $\OPT$ is optimal.
    Since $Q^*(\pi, \xset)\leq Q(\cA',\pi,\xset)$ by definition, there is a contradiction.
\end{proof}
\section{Proof of Theorem~\ref{thm:greedyir}}\label{app:greedyir}
\begin{proof}
    The construction is inspired by~\cite{dasgupta2004analysis}.
    Consider $k=2$ and let the unlabeled data set of agent $1$ be 
    \[X_1 = \{(0,1,0),(0,2,0), (0,0,1), (0,0,2),\ldots, (0,0,n)\}\]
    for some $n\in \NN_+$.
    
    Let  the unlabeled data set of 
    agent $2$ be
    \[X_2 = \{(1,0,0)\}\,.\]
    Let the unlabeled pool $X = X_1\cup X_2$.
    Let $h_{i,j,l}$ denote the hypothesis which labels $(i,0,0),(0,j,0),(0,0,l)$ as $1$ and the rest as $0$.
    
    Let the hypothesis class be $\cH = \{h_{i,j,l}|i\in \{0,1\}, j\in [2], l\in [n]\}$.
    
    Let the prior distribution $\pi_0 = \pi$ be defined as follows:
    \begin{equation*}
        \begin{cases}
            \pi(h_{0,0,0}) = \frac{1}{4}&\\
            \pi(h_{0,j,l}) = \frac{1}{4 \cdot 3^l}& \text{ for } j=1,2, l=1,\ldots,n-1\\
            \pi(h_{0,j,n}) = \frac{1}{8\cdot 3^{n-1}}& \text{ for } j=1,2\\
            \pi(h_{1,1,l}) = \frac{1}{3^l}& \text{ for } l=1,\ldots,n-1\\
            \pi(h_{1,1,n}) = \frac{1}{2\cdot 3^{n-1}}\,.&
        \end{cases}
    \end{equation*}

    Now we show that the label complexity of agent $1$ in the collaboration is $Q_1(\gbs, \pi, \{X_1,X_2\}) = \Omega(n)$. While the label complexity of running GBS itself is $Q(\gbs,\pi, \{X_1\}) = \cO(1)$.

    \paragraph{Label complexity of agent $1$ in the collaboration} Let $\VS$ denote the version space. And for any point $x$, let $\VS_{x}^+ = \{h\in \VS|h(x) =1\}$ denote the subset of the version space which labels $x$ by $1$. Similarly, let $\VS_{x}^- = \{h\in \VS|h(x) =0\}$.

    Now let us consider the length of the 
    path in the query tree when the target hypothesis is $h_{0,0,0}$.
    
    A-priori (before starting to query), for point $(1,0,0)$, we have 
    \begin{equation*}
        \pi(\VS_{(1,0,0)}^+) = \sum_{l=1}^n \pi(h_{1,1,l}) = \frac{1}{3} +\frac{1}{3^2} +\ldots +\frac{1}{3^{n-1}} +\frac{1}{2\cdot 3^{n-1}} =\frac{1}{2}\,.
    \end{equation*}   
    For point $(0,1,0)$, we have
    \[\pi(\VS_{(0,1,0)}^+) = \sum_{l=1}^n (\pi(h_{0,1,l}) + \pi(h_{1,1,l}))= \sum_{l=1}^n \pi(h_{0,1,l}) + \pi(\VS_{(1,0,0)}^+) >\frac{1}{2}\,.\]
    For point $(0,2,0)$, we have
    \[\pi(\VS_{(0,2,0)}^+) = \sum_{l=1}^n \pi(h_{0,2,l}) = \frac{1}{4\cdot 3} +\frac{1}{4\cdot 3^2} +\ldots +\frac{1}{4\cdot 3^{n-1}} +\frac{1}{8\cdot 3^{n-1}} =\frac{1}{8}\,.\]
    For other points $(0,0,l)$ for $l\in [n-1]$, we have $\pi(\VS_{(0,0,l)}^+) = \frac{1}{4\cdot 3^l} + \frac{1}{3^l} = \frac{5}{4\cdot 3^l}<\frac{1}{2}$ and for $(0,0,n)$, we have $\pi(\VS_{(0,0,n)}^+)<\pi(\VS_{(0,0,n-1)}^+)<\frac{1}{2}$.
    
    Therefore, the algorithm $\gbs(\pi,\{X_1,X_2\})$ will query $(1,0,0)$ at time $1$.
    Suppose the label of $(1,0,0)$ is $0$ since we consider the path corresponding to $h_{0,0,0}$ as the target hypothesis.
    
    Now we show that $\gbs(\pi,\{X_1,X_2\})$ will query points $(0,0,1), (0,0,2),\ldots, (0,0,n)$ sequentially by induction.
    
    At time $1$, the version space is $\VS = \{h_{0,0,0}\}\cup \{h_{i,j,l}\in \hat H|i=0\}$. We list $\pi(h_{0,j,l})$ for $j\in \{1,2\}$ and $l\in [n]$ in Table~\ref{tab:pi-0jl} for illustration.
    \begin{table}[H]
        \centering
        \begin{tabular}{c|c|c}
             & $(0,1,0)$ & $(0,2,0)$\\\hline
            $(0,0,1)$ & $\frac{1}{4 \cdot 3}$ & $\frac{1}{4 \cdot 3}$\\\hline
            $(0,0,2)$ &$\frac{1}{4 \cdot 3^2}$ & $\frac{1}{4 \cdot 3^2}$ \\\hline
            $\cdots$ & $\cdots$ & $\cdots$ \\\hline
            $(0,0,n)$ & $\frac{1}{8 \cdot 3^{n-1}}$ & $\frac{1}{8 \cdot 3^{n-1}}$\\\hline
        \end{tabular}
        \caption{Table of $\pi(h_{0,j,l})$ for $j\in \{1,2\}$ and $l\in [n]$.}
        \label{tab:pi-0jl}
    \end{table}
    Then we can compute that 
    \[\pi(S_{(0,1,0)}^+) = \pi(S_{(0,2,0)}^+)=\frac{1}{4\cdot 3} +\frac{1}{4\cdot 3^2} +\ldots +\frac{1}{4\cdot 3^{n-1}} +\frac{1}{8\cdot 3^{n-1}} =\frac{1}{8}\,,\]
    \[\pi(S_{(0,0,1)}^+) = \frac{1}{6}> \pi(S_{(0,0,l)}^+)\,,\]
    \[\pi(S_{(0,0,l)}^+) \leq \pi(S_{(0,0,2)}^+) = \frac{1}{18}\,,\]
    for all $l\geq 2$. 
    
    Thus, the algorithm $\gbs(\pi,\{X_1,X_2\})$ will choose $(0,0,1)$ at time $2$. 
    
    Suppose that at time $t=2,3,\ldots,l$, $\gbs(\pi,\{X_1,X_2\})$ has picked $(0,0,1),\ldots, (0,0,l-1)$ and all are labeled $0$.
    
    Now we show that $\gbs(\pi,\{X_1,X_2\})$ will pick $(0,0,l)$ at time $t= l+1$.
    The version space at the beginning of time $l+1$ is
    $\VS = \{h_{0,0,0}\}\cup \{h_{i,j,p}\in \hat H|p\geq l\}$.
    We can compute that 
    \[\pi(S_{(0,0,l)}^+) = \frac{1}{2\cdot 3^l}> \pi(S_{(0,0,p)}^+)\] 
    for all $p>l$, and that 
    \[
    \pi(S_{(0,1,0)}^+) = \pi(S_{(0,2,0)}^+)= \frac{1}{4 \cdot 3^l} + \frac{1}{4 \cdot 3^{l+1}} + \ldots+\frac{1}{4 \cdot 3^{n-1}} + \frac{1}{8 \cdot 3^{n-1}} = \frac{1}{8 \cdot 3^{l-1}}\,.
    \]
    Hence, $\gbs(\pi,\{X_1,X_2\})$ will pick $(0,0,l)$.
    
    Therefore, we proved that when the target hypothesis is $h_{0,0,0}$, $\gbs(\pi,\{X_1,X_2\})$ will query \\
    $(1,0,0),(0,0,1), (0,0,2),\ldots, (0,0,n)$ sequentially. 
    
    Thus, we have that $Q_1(\gbs,\pi,\{X_1,X_2\},h_{0,0,0}) = n + 1$, and 
    $Q_1(\gbs,\pi,\{X_1,X_2\}) \geq \frac{n+1}{4}$ as $\pi(h_{0,0,0}) =\frac{1}{4}$.

\paragraph{Label complexity of agent $1$ when she runs the (GBS) baseline individually}
Now we show that $Q(\gbs,\pi,\{X_1\}) = \cO(1)$. Since $X_1$ does not contain $(1,0,0)$, both $h_{0,j,l}$ and $h_{1,j,l}$ label $X_1$ identically. Every effective hypothesis over $X_1$ can be written as $h_{*,j,l}$ with $\pi(h_{*,j,l}) = \pi(h_{0,j,l})+\pi(h_{1,j,l})$, which is listed in Table~\ref{tab:pi-xjl}.
       \begin{table}[H]
        \centering
        \begin{tabular}{c|c|c}
             & $(0,1,0)$ & $(0,2,0)$\\\hline
            $(0,0,1)$ & $\frac{1}{4 \cdot 3} + \frac{1}{3}$ & $\frac{1}{4 \cdot 3}$\\\hline
            $(0,0,2)$ &$\frac{1}{4 \cdot 3^2}+ \frac{1}{3^2}$ & $\frac{1}{4 \cdot 3^2}$ \\\hline
            $\cdots$ & $\cdots$ & $\cdots$ \\\hline
            $(0,0,n)$ & $\frac{1}{8 \cdot 3^{n-1}} +\frac{1}{2 \cdot 3^{n-1}}$ & $\frac{1}{8 \cdot 3^{n-1}}$\\\hline
        \end{tabular}
        \caption{Table of $\pi(h_{*,j,l})$ for $j\in \{1,2\}$ and $l\in [n]$.}
        \label{tab:pi-xjl}
    \end{table}  
    Notice that if we know that the label of $(0,0,l)$ is positive for some $l$, then the version space has at most $2$ effective hypotheses, $h_{*,1,l}$ and $h_{*,2,l}$.
    In this case, the algorithm needs at most $2$ more queries.
    
    At time $t=1$, we have
    \[\pi(S_{(0,0,1)}^+) = \frac{1}{4 \cdot 3} + \frac{1}{3} + \frac{1}{4 \cdot 3} =\frac{1}{2}\,,\] 
    \[\pi(S_{(0,0,l)}^+) < \pi(S_{(0,0,1)}^+)\,,\forall l\geq 2\,,\]
    \[\pi(S_{(0,2,0)}^+) = \frac{1}{4\cdot 3} +\frac{1}{4\cdot 3^2} +\ldots +\frac{1}{4\cdot 3^{n-1}} +\frac{1}{8\cdot 3^{n-1}} = \frac{1}{8}\,,\]
    \[\pi(S_{(0,1,0)}^+) = \pi(S_{(0,2,0)}^+) \cdot 5 =\frac{5}{8}\,.\]
    Therefore, $\gbs(\pi,\{X_1\})$ will query $(0,0,1)$ at $t=1$.
    
    We complete the proof by exhaustion. 
    If $(0,0,1)$ is labeled as $1$, then the algorithm needs at most two more queries as aforementioned.
    
    If $(0,0,1)$ is labeled as $0$, then $h_{*,1,1}$ and $h_{*,2,1}$ will be removed from the version space and $\gbs(\pi,\{X_1\})$ will query $(0,1,0)$ at $t=2$ then.
    
    If the label is $1$, the version space is reduced to $\{h_{*,1,l}|l=2,\ldots,n\}$ and $\gbs(\pi,\{X_1\})$ will query $(0,0,2),(0,0,3),\ldots$ sequentially until receiving a positive label.
    
    If the label of $(0,1,0)$ is $0$, $\gbs(\pi,\{X_1\})$ will query $(0,2,0)$ at time $t=3$.
    If the label of $(0,2,0)$ is $1$, then it is similar to the case of  $(0,1,0)$ being labeled $1$ and the algorithm will query $(0,0,2),(0,0,3),\ldots$ sequentially.
    
    If the label of $(0,2,0)$ is $0$, we know the target hypothesis is $h_{0,0,0}$ and we are done.
    
    Hence we have $Q(\gbs,\pi,\{X_1\}) \leq \sum_{l=1}^n (\pi(h_{*,1,l}) + \pi(h_{*,2,l})) \cdot (3+ l) + \pi(h_{0,0,0})\cdot 3 = \sum_{l=1}^{n-1} \frac{1}{2\cdot 3^{l-1}}  \cdot (3+ l) + \frac{1}{4}\cdot 3 = \cO(1)$.
\end{proof}
\section{Proof of Lemma~\ref{lmm:partsir}}\label{app:part-sir}
\begin{proof}
\textbf{SIR property: } Since $\cA'$ and $\{\cO_i|i\in [k]\}$ are IR and agent $i$ can strictly benefit from $\cO_i$, 
we have $Q_i(\cA_\epsilon'', \pi, \xset) < Q(\cA,\pi, \{X_i\})$ for all $i\in [k]$.

\textbf{Label complexity: } The label complexity of $\cA_\epsilon''$ is
\begin{align*}
    Q(\cA_\epsilon'',\pi, \xset) = & (1-\frac{\epsilon}{n})Q(\cA',\pi,\xset) + \frac{\epsilon}{kn} \sum_{i=1}^k Q(\cO_i,\pi,\xset)\\
    \leq& (1-\frac{\epsilon}{n})Q(\cA',\pi,\xset) + \epsilon\,.
\end{align*}
Then we are done.
\end{proof}
\section{Proof of necessity of Assumption~\ref{asp:clb-help}}\label{app:nece-asp}
\begin{proof}[Proof of necessity]
Suppose that there exists an SIR algorithm $\cA''$ when the baseline algorithm is optimal. We therefore have 
$Q_i(\cA'',\pi,\xset) < Q^*(\pi,\{X_i\})$ by definition.  
We claim that $\cA''$ must satisfy $Q_i(\cA'',\pi,\xset) \geq \EEs{h\sim \pi}{Q^*(\pi_{h,-i}, \{X_i\})}$.
This is because we can construct another algorithm $\cB$ by running $\cA''$ over all other agents except agent $i$, i.e., running $\cA''$ over $(\pi, X_{-i})$ with $X_{-i}= \{X_j|j\neq i\}$ first to recover the labels of all other agents $X_{-i}$.
Then, $\cB$ simulates $\cA''$ over $(\pi, \xset)$ without actually querying any point in $X_{-i}$ (similarly to Algorithm~\ref{alg:basic2ir}).
In this case, the label complexities of agent $i$ are identical for algorithms $\cB$ and $\cA''$, i.e., $Q_i(\cB,\pi,\xset)= Q_i(\cA'',\pi,\xset)$. 
Since $Q^*(\pi_{h,-i}, \{X_i\})$ is the optimal label complexity of agent $i$ given the label information of $X_{-i}$, we have $Q_i(\cB,\pi,\xset) \geq Q^*(\pi_{h,-i}, \{X_i\})$.
Therefore, we have $Q^*(\pi,\{X_i\}) > Q_i(\cA'',\pi,\xset)\geq Q^*(\pi_{h,-i}, \{X_i\})$.
\end{proof}
\section{Proof of Lemma~\ref{lemma:OiIsI-Part-SIR}}\label{app_lmm3}
\begin{proof}
    First, note that $\cO_i$ is IR as $\ir(\cA)$ is IR, and pruning the query tree does not increase label complexity.
    For any $i\in [k]$, suppose that there exists an hypothesis $h\in \hat H$ s.t. $\abs{H(X_i|h)} < \abs{H(X_i)}$.
    Then in the query tree of $\cA(\pi, \{X_i\})$, either all leaves are inconsistent with $h(X_{-i})$ or there exists one internal node $v$ who has exactly one subtree with all leaves inconsistent with $h(X_{-i})$.
    This node $v$ as well as the corresponding subtree are pruned in $\cO_i$ and thus the leaves in the other subtree rooted at $v$  have their depth reduced by at least $1$. Now, there exists an hypothesis $h''\in \hat H$ such that $h''(X_i)\in H(X_i|h)$ and $h''(X_i)$ is in the other subtree.
    Since $h''(X_{-i}) = h(X_{-i})$, when the underlying hypothesis is $h''$, the pruned tree given $h''(X_{-i})$ is the same as that given $ h(X_{-i})$. Hence, we have $Q_i(\cO_i,\pi,\xset, h'') \leq Q_i(\cA,\pi,\xset,h'')-1$.
    
    Then we have 
    \begin{align*}
        &Q_i(\cO_i,\pi,\xset) =  \EEs{h\sim \pi}{Q_i(\cO_i,\pi,\xset, h)}
        \\
        =& \pi(h'')Q_i(\cO_i,\pi,\xset, h'') + (1-\pi(h''))\EEs{h\sim \pi|h \neq h''}{Q_i(\cO_i,\pi,\xset, h)}\\
        \leq & \pi(h'')(Q_i(\cA,\pi,\xset, h'')-1) + (1-\pi(h''))\EEs{h\sim \pi|h\neq h''}{Q_i(\cA,\pi,\xset, h)}\\
        <& Q_i(\cA,\pi,\xset)\,,
    \end{align*}
where the last inequality holds due to $\pi(h'')>0$ since  w.l.o.g., we assumed $\pi(h)>0$ for all $h\in \hat H$ in Section~\ref{sec:model-active}.
\end{proof}
\chapter{Learning within Games}\label{app:games}
\section{Table of Notation}
\begin{table}[H]
    \begin{tabular}{p{4cm}p{10.5cm}}
    \toprule
        \centering\textbf{Symbol} & \textbf{Description}\\
    \midrule
        \centering$\cP$& Policy space.\\
        \centering$\cA$& Action space.\\
        \centering$\cP_0 \subset \cP$ & Benchmark policy set.\\
        \centering$p \in \cP$& Principal's policy.\\
        \centering $a \in \cA$ & Agent's action.\\
        \centering $\mu \in \Delta (\cA)$ & Distribution over Agent's actions.\\
        \centering$r \in \cA$ & Principal's recommended action for the Agent.\\
         \centering$y \in \cY$ & State of nature.\\
        \centering$\hat{y}$ & Empirical distribution over states of nature over a particular subsequence. \\
        \centering$\pi_t\in \Delta(\cY)$ & forecast at time $t$.\\
        \centering $V(a,p,y)$ & Principal's utility.\\
        \centering $U(a,p,y)$ & Agent's utility.\\
        \centering$\brp(\pi)$ & Principal best response assuming a shared prior $\pi$. \\
        \centering$\brr(p,\pi)$ & Agent best response assuming a prior $\pi$, breaking ties in favor of the Principal's utility. \\
        \centering $\cB(p,\pi,\epsilon)$ & the set of all $\epsilon$-best responses for the Agent.\\
        \centering $\brr(p,\pi,\epsilon)$ & the utility-maximizing action for the Principal amongst the Agent's $\epsilon$-best responses to $p$.\\
        \centering$\alpha$ & conditional bias parameter. \\
        \centering$\eint$ & swap regret upper bound. \\
        \centering$\eneg$ & negative regret upper bound. \\
    \bottomrule
    \end{tabular}
    \caption{Summary of game-theoretic notation used in this article.}
    \label{tab:notation}
\end{table}
\section{Additional Related Work}
The foundations of principal agent problems and contract theory (in the standard setting with common priors) date back to \cite{holmstrom1979moral} and \cite{grossman1992analysis}. This literature is far too large to survey --- we refer the reader to \cite{bolton2004contract} for a textbook introduction, and here focus on only the most relevant work. 

Optimal contracts under a common prior assumption can be very complicated, and do not reflect structure seen in real world contracts. This criticism goes back to at least \cite{holmstrom1987aggregation}, who show a dynamic setting in which optimal contracts are linear. Recently, linear contracts have become an object of intense study, with work showing that they are optimal in various worst-case settings. In the classical common prior setting, \cite{carroll2015robustness} shows that linear contracts are minimax optimal for a Principal who knows \emph{some} but not \emph{all} of the Agent's actions. Similarly, \cite{dutting19} shows that if the Principal only knows the costs and expected rewards for each Agent action, then linear contracts are minimax optimal over the set of all reward distributions with the given expectation. \cite{dutting2022combinatorial} extends this robustness result to a combinatorial setting. \cite{dutting19} also show linear contracts are bounded approximations to optimal contracts, where the approximation factor can be bounded in terms of various quantities (e.g. the number of agent actions, or the ratio of the largest to smallest reward, or the ratio of the largest to smallest cost, etc). 
\cite{castiglioni2021bayesian} studies linear contracts in Bayesian settings (when the Principal knows a distribution over types from which the Agent's type is drawn) and studies how well linear contracts can approximate optimal contracts. In this setting, optimal contracts can be computationally hard to construct, and show that linear contracts obtain optimal approximations amongst tractable contracts. 

There is also a more recent tradition of studying sequential (repeated) principle agent games. \cite{ho2014adaptive} study online contract design by approaching it as a bandit problem in which an unknown distribution over myopic agents arrive and respond to an offered Principal contract by optimizing their expected utility with respect to a known prior. \cite{cohen2022learning} extend this to the case in which the Agent has bounded risk aversion. 
\cite{zhu2022sample} revisit this problem and characterize the sample complexity of online contract design in general (with nearly matching upper and lower bounds) and for the special case of linear contracts (with exactly matching upper and lower bounds). In contrast to this line of work, our Agent is not myopic --- a primary challenge is that we need to manage their long-term incentives --- and we make no distributional assumptions at all, either about the actual realizations nor about agent beliefs. 

\cite{chassang2013calibrated} studies a repeated interaction between a Principal and a long-lived Agent, with a focus on the \emph{limited liability} problem. As discussed, linear contracts have many attractive robustness properties, but can require negative payments from the Agent, which are difficult to implement. A limited liability contract, in contrast, never requires negative payments. Using a Blackwell-approachability argument, \cite{chassang2013calibrated} shows how to repeatedly contract with a single Agent (or instead to use a free outside option) so that the aggregate payments made to the agent is the same as they would have been under a linear contract, but negative payments are never required, and the Principal has no regret to either always contracting with the agent or always using the outside option.

The Bayesian Persuasian problem was introduced by \cite{kamenica2011bayesian} and has been studied from a computational perspective since \cite{dughmi2016algorithmic}. It has been applied to various problems, including incentivizing exploration in multiarmed bandit problems \citep{cohen2019optimal,sellke2021price,mansour2022bayesian}. A recent literature has studied sequential Bayesian Persuasian problems. \cite{zu2021learning} and \cite{bernasconi2022sequential} study a sequential Bayesian Persuasian problem in which the Principal does \emph{not initially} know the underlying distribution on the state space, and needs to learn it while acting in the game. \cite{wu2022sequential} study a sequential problem in which a Principal repeatedly interacts with myopic agents, using tools from reinforcement learning. \cite{gan2022bayesian} study a sequential Bayesian Persuasian problem in which the state evolves according to a Markov Decision Process, and show that for a myopic agent, the optimal signalling scheme can be computed efficiently, but that it is computationally hard for a non-myopic agent. \cite{bernasconi2023optimal} study regret bounds for a Principal in a sequential Bayesian Persuasian problem facing a sequence of myopic Agents, whose utility functions can be chosen by an adversary.

There is a substantial body of work on learning in repeated Stackelberg games (both in general and in various special cases like security games, strategic classification, and dynamic pricing) in settings in which the Agent has complete information and the Principal needs to learn about the Agent's preferences (see e.g. \citep{blum2014learning,Balcan2015CommitmentWR,roth2016watch,dong2018strategic,chen2020learning,roth2020multidimensional}). In these works, the Agent is myopic and optimizes for their one-round payoff. \cite{haghtalab2022learning} consider a non-myopic agent who discounts the future, and give no-regret learning rules for the Principal that take advantage of the fact that for a future-discounting agent, mechanisms that are slow to incorporate learned information will induce near-myopic behavior. The regret bounds in \cite{haghtalab2022learning} tend to infinity as the Agent becomes more patient. 
\cite{collina2023efficient} derive optimal commitment algorithms for complete-information Stackelberg games when the follower is maximizing their total payoff in expectation. In contrast to these works, we (and \cite{camara2020mechanisms} before us) operate in a setting without distributions (or assumed distributions that Agents can be said to optimize over) and give policy regret bounds contingent on the Agent satisfying behavioral assumptions defined by regret bounds.  This is  similar in spirit to \cite{deng2019strategizing}, which considers playing a repeated game against an agent playing a no-swap regret algorithm and shows that the optimal strategy is to play the single-shot Stackelberg equilibrium at each round. \cite{haghtalab2023calibrated} show that the same is true if an agent is best-responding to a calibrated predictor for the Principal's actions --- and accomplish this also by using a form of ``stable'' policies as we do. 

There is a long tradition of using ``no-regret'' assumptions as relaxations of classical assumptions that players in a game either best respond to beliefs or play a Nash equilibrium --- for example, when proving price of anarchy bounds \citep{blum2008regret,roughgarden2015intrinsic,lykouris2016learning}, when doing econometric inference \citep{nekipelov2015econometrics}, or when designing optimal pricing rules \citep{braverman2018selling,cai2023selling}, as well as work focused on how to play games against no-regret learning agents \citep{deng2019strategizing,mansour2022strategizing,kolumbus2022and,brown2023learning}.

Finally, the use of calibrated forecasts in decision-making settings dates back to \cite{foster1999regret}, who showed that agents best-responding to calibrated forecasts of their payoffs have no internal (equivalently swap) regret. Similarly \cite{kakade2008deterministic} and \cite{foster2018smooth} connect a determinstic ``smooth'' version of calibration to Nash equilibrium. A recent literature on ``multicalibration'' \citep{hebert2018multicalibration} has investigated various refinements of calibration; this has developed into a large literature and we refer the reader to \cite{RothNotes} for an introductory overview. Work on ``omniprediction''  \citep{GopalanKRSW22,gopalan2023loss,GHK23,gopalan2023characterizing,GJRR23} uses multicalibration to provide guarantees for a variety of 1-dimensional downstream decision making problems. Decision calibration \citep{zhao2021calibrating} (in the batch setting) aims to calibrate predictions to the best-response correspondence of a downstream decision maker. The tools we use, developed by \cite{NRRX23} arise from this literature. 
\section{Impossiblity Results}
\label{sec:imposs}

Throughout this paper, we have given policy regret bounds for the Principal under a variety of kinds of assumptions: behavioral assumptions for the Agent, and either alignment assumptions on the interaction, or else assumed access to a way of constructing optimal stable policies. In this Section we interrogate the necessity of those assumptions.

\subsection*{Stable Policies Do Not Always Exist}
We avoided alignment assumptions by showing how to construct optimal ``stable'' policies in two important special cases: linear contracting settings, and binary state Bayesian persuasion settings. Might we be able to avoid alignment assumptions in full generality this way? Unfortunately not. The lemma below implies that it is sometimes not possible to construct a $(c, \epsilon, \beta, \gamma)$-stable policy oracle such that Theorem~\ref{thm:stable} guarantees vanishing policy regret. The counterexample involves a simple two-policy, two-action contract setting in which the Principal can get high regret to either of their policies, depending on the tiebreaking rule of the Agent.

\begin{restatable}{proposition}{nostable}\label{lem:impossstable}
There exists a Principal/Agent problem 
in which for all priors $\pi$ and for all $c \leq \frac{1}{4}$, $\epsilon \geq 0$, $\gamma \leq \frac{1}{2}$ and $\beta > 0$, there is no $(c,\epsilon,\beta, \gamma)$-optimal stable policy under $\pi$.
\end{restatable}

An implication of this is that it is not possible to extend our ``stable policy oracle'' approach to capture the entire scope of the Principal/Agent problem we study in this paper.

\subsection*{A No-Secret-Information Assumption is Necessary}

\necessity*

Recall that in Section \ref{sec:behavior} we introduced two behavioral assumptions: A no contextual-swap-regret assumption (Assumption \ref{asp:no-internal-reg}), as well as a ``no-secret-information'' assumption (Assumption \ref{asp:no-correlation}). Assumption \ref{asp:no-internal-reg} was straightforwardly motivated as the ``rationality'' assumption in our model, but it was less clear that Assumption \ref{asp:no-correlation}---which informally asked that the Agent's actions be un-correlated with the states, conditional on the Principal's actions---was necessary. In this Section we establish the necessity of Assumption \ref{asp:no-correlation}.

This proposition can be interpreted as follows: against any Principal mechanism, either there is an Agent learning algorithm that achieves vanishing Contextual Swap Regret and ensures the Principal high regret, or it is impossible for any Agent learning algorithm to achieve vanishing Contextual Swap Regret. This second case is a degenerate case and could only occur if the Principal mechanism is allowed to output $\Omega(T)$ different policies, leading to an unfairly fine-grained context for the Agent to compete against. In this case, the no contextual-swap-regret assumption will rule out all learning algorithms.

One might ask whether Assumption~\ref{asp:no-correlation} is unnecessarily strong for this task; in other words, it might be possible to prove a positive result when the Agent is constrained by Assumption~\ref{asp:no-internal-reg} and a weakened version of Assumption~\ref{asp:no-correlation}. To address this, we also prove that if the Agent is allowed to play any algorithm satisfying Assumption~\ref{asp:no-internal-reg}
and Assumption~\ref{asp:no-sec-info} (introduced in Section~\ref{sec:general}), which is similar to but weaker than Assumption~\ref{asp:no-correlation}, he can ensure the Principal high regret. 

Intuitively, Assumption~\ref{asp:no-correlation} asks for the Agent's actions to not be statistically correlated with the state of nature, while Assumption~\ref{asp:no-sec-info} asks for the Agent to not perform much better than the best fixed mapping from (policy, recommendation) to actions. We show in Lemma~\ref{lmm:weakerasp} that Assumption~\ref{asp:no-sec-info} is weaker than Assumption~\ref{asp:no-correlation}. However, it still asks for something quite strong from the Agent: when combined, Assumptions~\ref{asp:no-internal-reg} and~\ref{asp:no-sec-info} bound the performance of the Agent from above and below. This might seem to suggest that the Agent cannot do much other than play a standard no-regret algorithm.

However, we show that even when satisfying Assumptions~\ref{asp:no-internal-reg} and~\ref{asp:no-sec-info}, an Agent can leverage extra information he has to ensure that the Principal attains high regret. In a simple linear contract setting, we construct an Agent algorithm $\mathcal{L}$ which either plays a simple no-regret algorithm, or uses knowledge of the states of nature to play a sequence that gets him the same utility and ensures the Principal larger utility. Depending on the Principal's actions and the states of nature, $\mathcal{L}$ selects which sub-algorithm to run. We show that for every Principal mechanism, there must be some state of nature sequence under which $\mathcal{L}$ picks the worst option for the Principal, leading to non-vanishing policy regret.

For this additional result to hold, we only need there to exist some Agent learning algorithm which not only gets vanishing Contextual Swap Regret, but also gets vanishing negative regret. Many well-known no-regret algorithms are known to have this guarantee \citep{gofer16}.

\strongnecessity*
We show our impossibility result in a linear contract setting, the same setting we show positive results for in Section~\ref{sec:linear-contract} when the Agent is further constrained by Assumption~\ref{asp:no-correlation}. Therefore, when keeping all else fixed, we prove that Assumption~\ref{asp:no-correlation} makes the difference between a tractable and intractable setting. Note that this does not imply that a Principal can never achieve vanishing regret without Assumption~\ref{asp:no-correlation}. Indeed in Section~\ref{sec:general} we show that Assumption~\ref{asp:no-sec-info} (which is weaker than Assumption~\ref{asp:no-correlation}) suffices if it is paired with an \emph{Alignment} assumption (Assumption \ref{asp:alignment}). However, Alignment assumptions are different in character to our behavioral assumptions: they constrain the sequence of states of nature, and simply rule out the kinds of examples we use in proving our lower bound statements. Thus we can also view this proposition as demonstrating the necessity of the Alignment condition in general.


\section{Proofs from Section~\ref{sec:stable}}\label{sec:stable-proof}

\thmstable*

Let $E_{1,p,r}$ denote the event of $\ind{(p_t,r_t) = (p,r)}$ for all $(p,r)$, $E_{2,p,a}$ denote the event of $\ind{(\pit,\rit) = (p,a)}$ for all $(p,a)$ and $E_{3,p_0,a}$ denote the event of $\ind{a^*(p_0,\pi_t) = a}$ for all $a$.
Let $\cE_{3,p_0} = \{\ind{a^*(p_0,\pi_t) = a}\}_{a\in \cA}$.
Let $\alpha(\cE_1) = \sum_{E\in \cE_1} \alpha(E)$, $\alpha(\cE_2) = \sum_{E\in \cE_2} \alpha(E)$ and $\alpha(\cE_{3,p_0}) = \sum_{E\in \cE_{3,p_0}} \alpha(E)$.
We introduce the following generalized version of Theorem~\ref{thm:stable}.
\begin{restatable}{theorem}{thmstableapp} \label{thm:stable-appendix}
Assume that the Agent's learning algorithm $\cL$ satisfies the behavioral assumptions~\ref{asp:no-internal-reg} and \ref{asp:no-correlation} and 
that the forecasts $\pip_{1:T}$  have conditional bias $\alpha$ conditional on the events $\cE$. 
Given access to an optimal stable policy oracle $\cO_{c,\epsilon,\beta,\gamma}$, by running Algorithm~\ref{alg:general-stable}, which uses $\cO_{c,\epsilon,\beta,\gamma}$ as the choice rule, the Principal can achieve policy regret
\begin{align*}
    &\text{PR}(\cO_{c,\epsilon,\beta,\gamma}, \pi_{1:T}, \cL,y_{1:T}) \\
    = & c+ 3\alpha(\cE_1) + 2\alpha(\cE_2) + \max_{p_0\in \cP_0}\alpha(\cE_{3,p_0}) + \gamma + \frac{\eint +\cO(\sqrt{\abs{\cP_\cO}\abs{\cA}/T})+2\alpha(\cE_1)  }{\beta} \\
    & + \frac{\eint + \cO(\sqrt{\abs{\cA}/T})+2\max_{p_0\in \cP_0}\alpha(\cE_{3,p_0}) }{\epsilon}\,.
\end{align*}
    
\end{restatable}

\begin{proof}[Proof of Theorem~\ref{thm:stable}]
    By Theorem~\ref{thm:forecast-bias}, we have
\begin{align*}
    \EEs{\pi_{1:T}}{\alpha(E)}\leq O\left(\frac{|\cY|\ln(|\cY||\cE|T)}{T} + \frac{|\cY|\sqrt{\ln(|\cY||\cE|T)|\{t : E(\pi_t) = 1|\}}}{T}\right) \,.
\end{align*}
Hence, we have:
\begin{align*}
    &\EEs{\pi_{1:T}}{\alpha(\cE_1)} \leq \cO(\frac{|\cY|\ln(|\cY|(|\cP_{\cO}| +|\cP_0|)\abs{\cA}T)}{T} +\abs{\cY} \sqrt{\frac{\ln(|\cY|(|\cP_{\cO}| +|\cP_0|)\abs{\cA}T)\abs{\cP_\cO}\abs{\cA}}{T}})\,,
    \\
    & \EEs{\pi_{1:T}}{\alpha(\cE_2)} \leq  \cO(\frac{|\cY|\ln(|\cY|(|\cP_{\cO}| +|\cP_0|)\abs{\cA}T)}{T}+\abs{\cY}\sqrt{\frac{\ln(|\cY|(|\cP_{\cO}| +|\cP_0|)\abs{\cA}T)\abs{\cP_0}\abs{\cA}}{T}})\,,\\
   & \EEs{\pi_{1:T}}{\alpha(\cE_{3,p_0})} \leq  \cO(\frac{|\cY|\ln(|\cY|(|\cP_{\cO}| +|\cP_0|)\abs{\cA}T)}{T}+\abs{\cY}\sqrt{\frac{\ln(|\cY|(|\cP_{\cO}| +|\cP_0|)\abs{\cA}T)\abs{\cA}}{T}})\,.
\end{align*}
By taking expectation over $\pi_{1:T}$ and plugging these values into Theorem~\ref{thm:stable-appendix}, we have
\begin{align*}
    &\text{PR}(\sigma^\dagger, \cL,y_{1:T}) \leq\tilde \cO\left(c +\gamma +\sqrt{\abs{\cP_0}\abs{\cA}/T} + \frac{\eint + \abs{\cY}\sqrt{\abs{\cP_\cO}\abs{\cA}/T}}{\beta} + \frac{\eint + \abs{\cY}\sqrt{\abs{\cA}/T}}{\epsilon}\right)\,.
\end{align*}
Hence we are done with proof of Theorem~\ref{thm:stable}.
\end{proof}

\subsection*{Proof of Theorem~\ref{thm:stable-appendix} }
\thmstableapp*
\begin{proof}
    For any sequence of states $y_{1:T}$ and sequence of forecasts $\pi_{1:T}$, and any constant policy $p_0\in \cP_0$, for any realized sequence of actions $a_{1:T}$ and $a^{p_0}_{1:T}$,
    we can decompose the (realized) regret compared with constant mechanism $\sigma^{p_0}$ as 
\begin{align*}
    &\frac{1}{T}\sum_{t=1}^T\left(V(a_t^{p_0},{p_0},y_t)- V(a_t,p_t ,y_t)\right)\\
    =& \frac{1}{T}\left(\underbrace{\sum_{t=1}^T(V(\rit,\pit,y_t)- V(r_t,p_t,y_t) )}_{(a)} + \underbrace{\sum_{t=1}^T(V(r_t,p_t,y_t)-V(a_t,p_t ,y_t))}_{(b)}\right.\\
    &\left.+\underbrace{\sum_{t=1}^T(V(a_t^{p_0},{p_0},y_t)- V(\rit,\pit,y_t))}_{(c)} \right)
\end{align*}
\begin{enumerate}
    \item We bound term (a) using the fact that $p_t$ is a $(c,\epsilon,\beta,\gamma)$-optimal stable policy under $\pi_t$.
    According to the definition of stable policy oracle (Definition~\ref{def:stabilized}), we have $V(r_t, p_t,\pip_t)\geq V(\rit,\pit,\pip_t) - c$. Then since $\pip_{1:T}$ has $\alpha$ bias conditional on $(p_t, r_t)$ and $(\pit,\rit)$, we have
    \begin{align*}
        \frac{1}{T}\sum_{t=1}^T V(r_t,p_t, y_t) \geq & \frac{1}{T}\sum_{t=1}^T V(r_t,p_t, \pip_t) - \alpha(\cE_1)\tag{$\cE_1$-bias}\\
        \geq &\frac{1}{T}\sum_{t=1}^TV(\rit,\pit,\pip_t)  -c-\alpha(\cE_1)\tag{stabilization}\\
        \geq &\frac{1}{T}\sum_{t=1}^TV(\rit,\pit,y_t)  -c-\alpha(\cE_2)-\alpha(\cE_1)\,.\tag{$\cE_2$-bias}
    \end{align*}
    Therefore, we have
    \begin{equation*}
        \text{Term (a)} \leq (c+\alpha(\cE_2)+\alpha(\cE_1))T\,.
    \end{equation*}

    \item We bound term (c) using the fact that $V(\rit,\pit,\pi_t)$ is the optimal optimistic achievable utility of the Principal.

For constant mechanism $\sigma^{p_0}$, let $t\in (r)$ denote $t: r^{p_0}_t = r$. Let $n^{p_0}_r = \sum_{t\in (r)}1 $ denote the number of rounds in which $r$ is recommended. 
Let 
\begin{align*}
    b^{p_0}_r = \frac{1}{n^{p_0}_r}\max \left(\abs{\sum_{t:r_t=r} U(a_t^{p_0}, p, y_t) - U(\hat \mu^{p_0}_{r}, p, y_t)}, \abs{\sum_{t:r_t=r} V(a_t^{p_0}, p, y_t) - V(\hat \mu^{p_0}_{r}, p, y_t)} \right)\,.
\end{align*}
By Assumption~\ref{asp:no-correlation}, we have $\EEs{\cL}{b^{p_0}_r} = \cO(\frac{1}{\sqrt{n^{p_0}_r}})$. 
Let $\hat \mu^{p_0}_r = \frac{1}{n^{p_0}_r}\sum_{t\in (r)} a^{p_0}_t$ denote the empirical distribution of Agent's action in the subsequence where $r$ is the recommendation.
Let 
\begin{equation*}
    \ir^{p_0}_{r} = \max_{h:\cA\mapsto \cA}\sum_{t\in (r)} \left(U(h(a^{p_0}_t),p_0, y_t) - U(a^{p_0}_t,{p_0},y_t)\right) 
\end{equation*}
denote the swap regret in this subsequence and let $\ir^{p_0} = \sum_{r\in \cA}\ir^{p_0}_{r}$ denote the swap regret for $a^{p_0}_{1:T}$.
Then we have
    \begin{align}
        &\sum_{t\in (r)}U(\hat \mu^{p_0}_r,p_0,\pip_t) \nonumber\\
        \geq &\sum_{t\in (r)}U(\hat \mu^{p_0}_r,p_0,y_t) - \alpha(E_{3,p_0,r})  T \tag{$\cE_3$-bias} \nonumber\\
        \geq &\sum_{t\in (r)}U(a_t^{p_0},p_0,y_t) -n^{p_0}_r b^{p_0}_r - \alpha(E_{3,p_0,r}) T \tag{no secret info} \nonumber\\
        \geq &\sum_{t\in (r)}U( r,p_0,y_t)-\ir^{p_0}_{r}-n^{p_0}_r b^{p_0}_r-\alpha(E_{3,p_0,r})  T\tag{definition of $\ir^{p_0}_{r}$}\nonumber\\
        \geq &\sum_{t\in (r)}U(r,{p_0},\pip_t)-\ir^{p_0}_{r}-n^{p_0}_r b^{p_0}_r -2\alpha(E_{3,p_0,r}) T\label{eq:u-gap}\,, 
    \end{align}
    where the last inequality again uses our bound on $\cE_3$-bias. 
    For a random action $a\sim \hat \mu^{p_0}_r$, let $F_t$ denote the event that $U(a,{p_0},\pip_t)< U(r,{p_0},\pip_t)-\epsilon$. We have
    \begin{align*}
    &\sum_{t\in (r)}U(\hat \mu^{p_0}_r,p_0,\pip_t)\\
        =&\sum_{t\in (r)}\left(\Pr_{a\sim \hat \mu^{p_0}_r}(F_t) \EEc{U(a,{p_0},\pip_t)}{F_t} +\Pr_{a\sim \hat \mu^{p_0}_r}(\neg F_t) \EEc{U(a,{p_0},\pip_t)}{\neg F_t}\right)\\
        \leq & \sum_{t\in (r)}\left(\Pr_{a\sim \hat \mu^{p_0}_r}(F_t) (U(r,{p_0},\pip_t)-\epsilon) +\Pr_{a\sim \hat \mu^{p_0}_r}(\neg F_t) U(r,{p_0},\pip_t)\right)\,.
    \end{align*}
    By combining with Eq~\eqref{eq:u-gap}, we have
    \begin{align}
        \sum_{t\in (r)}\Pr_{a\sim \hat \mu^{p_0}_r}(F_t)\leq \frac{\ir^{p_0}_{r}+n^{p_0}_r b^{p_0}_r +2\alpha(E_{3,p_0,r}) T}{\epsilon}\,.\label{eq:sum-prob}
    \end{align}
    We also have:
    \begin{align}
    V(\hat \mu^{p_0}_r,p_0,\pip_t) \leq& 
        \Pr_{a\sim \hat \mu^{p_0}_r}(\neg F_t)\max_{\tilde r\in \cB( p_0,\pip_t,\epsilon)}V(\tilde r,p_0,\pip_t) + \Pr_{a\sim \hat \mu^{p_0}_r}(F_t)\nonumber\\
        \leq& \max_{\tilde r\in \cB( p_0,\pip_t,\epsilon)}V(\tilde r,p_0,\pip_t) + \Pr_{a\sim \hat \mu^{p_0}_r}(F_t)\,.\label{eq:v-max}
    \end{align}
    By combining Eqs~\eqref{eq:sum-prob} and \eqref{eq:v-max}, we have
    \begin{align}
        \sum_{t\in (r)}\max_{\tilde r\in \cB( p_0,\pip_t,\epsilon)}V(\tilde r,p_0,\pip_t) \geq \sum_{t\in (r)}V(\hat \mu^{p_0}_r,p_0,\pip_t) - \frac{\ir^{p_0}_{r}+n^{p_0}_r b^{p_0}_r +2\alpha(E_{3,p_0,r}) T}{\epsilon}\,.\label{eq:rounds-not-in-eps-ball}
    \end{align}

    Then we have
    \begin{align*}
        &\sum_{t=1}^T V(\rit,\pit,y_t) \\
        \geq & \sum_{t=1}^T V(\rit,\pit,\pip_t)- \alpha(\cE_2) T \tag{$\cE_2$-bias}\\
        = & \sum_{t=1}^T \max_{\tilde p\in \cP}\max_{\tilde r\in \cB(\tilde p,\pip_t,\epsilon)}V(\tilde r,\tilde p,\pip_t)- \alpha(\cE_2) T\tag{definition of $(\pit,\rit)$}\\
        \geq & \sum_{t=1}^T \max_{\tilde r\in \cB( p_0,\pip_t,\epsilon)}V(\tilde r,p_0,\pip_t)- \alpha(\cE_2) T\\
        = & \sum_{r\in \cA}\sum_{t\in (r)}\max_{\tilde r\in \cB( p_0,\pip_t,\epsilon)}V(\tilde r,p_0,\pip_t)- \alpha(\cE_2) T\\
        \geq & \sum_{r\in \cA}\left(\sum_{t\in (r)}V(\hat \mu^{p_0}_r,p_0,\pip_t)- \frac{\ir^{p_0}_{r}+n^{p_0}_r b^{p_0}_r +2\alpha(E_{3,p_0,r}) T}{\epsilon}\right)- \alpha(\cE_2) T\tag{applying Eq~\eqref{eq:rounds-not-in-eps-ball}}\\
        \geq & \sum_{r\in \cA}\sum_{t\in (r)}V(\hat \mu^{p_0}_r,p_0,y_t)- \frac{\ir^{p_0} + \sum_{r\in \cA}n^{p_0}_r b^{p_0}_r +2\alpha(\cE_{3,p_0}) T}{\epsilon}-\alpha(\cE_{3,p_0}) T- \alpha(\cE_2) T\tag{$\cE_3$-bias}\\
        \geq &  \sum_{t} V(a_t^{p_0},p_0,y_t) - \sum_{r\in \cA}n^{p_0}_r b^{p_0}_r- \frac{\ir^{p_0} + \sum_{r\in \cA}n^{p_0}_r b^{p_0}_r +2\alpha(\cE_{3,p_0}) T}{\epsilon} -  (\alpha(\cE_{3,p_0}) + \alpha(\cE_2)) T\,.\tag{no secret info}
    \end{align*}
    Hence, we have
    \begin{align*}
        \text{Term (c)}=& \sum_{t=1}^T V(a_t^{p_0},{p_0},y_t)- V(\rit,\pit,y_t) \\
        \leq& \sum_{r\in \cA}n^{p_0}_r b^{p_0}_r+ \frac{\ir^{p_0} + \sum_{r\in \cA}n^{p_0}_r b^{p_0}_r +2\alpha(\cE_{3,p_0}) T}{\epsilon} +  (\alpha(\cE_{3,p_0}) + \alpha(\cE_2)) T\,.
    \end{align*}
    By taking expectation over the randomness of the Agent's learning algorithm $\cL$, we have
    \begin{equation*}
        \EEs{\cL}{\text{Term (c)}}\leq \left( \cO(\sqrt{\abs{\cA}/T})+ \frac{\eint + \cO(\sqrt{\abs{\cA}/T})+2\alpha(\cE_{3,p_0}) }{\epsilon} +  \alpha(\cE_{3,p_0}) + \alpha(\cE_2)\right)T\,.
    \end{equation*}
    \item We bound term (b) by proving that the number of rounds in which the Agent does not follow the recommendation $r_t$ is small using the fact that that $p_t$ is $(\beta,\gamma)$-stable under $\pi_t$.

    For proposed mechanism, let $t\in (p,r)$ denote $t:(p_t,r_t) = (p,r)$. Let $n_{p,r} = \sum_{t=1}^T\ind{t\in (p,r)}$ denote the number of rounds in which $(p_t,r_t) = (p,r)$.
    \begin{align*}
    b_{p,r} = \frac{1}{n_{p,r}}\max \left(\abs{\sum_{t\in (p,r)} U(a_t, p, y_t) -U(\hat \mu_{p,r}, p, y_t)}, \abs{\sum_{t\in (p,r)} V(a_t, p, y_t) -V(\hat \mu_{p,r}, p, y_t)} \right)\,.
\end{align*}
By Assumption~\ref{asp:no-correlation}, we have $\EEs{\cL}{b_{p,r}} = \cO(\frac{1}{\sqrt{n_{p,r}}})$. 
    Let $\hat \mu_{p,r} = \frac{1}{n_{p,r}}\sum_{t\in (p,r)} a_t
    $ denote the empirical distribution of the actions on this subsequence.
    Let $\hat y_{p,r} = \frac{1}{n_{p,r}}\sum_{t\in (p,r)} y_t$ denote the empirical distribution of states in these rounds and $\pip_{p,r} = \frac{1}{n_{p,r}}\sum_{t\in (p,r)} \pip_t$ denote the empirical distribution of the forecasts.
    Let
    \begin{equation*}
        \ir_{p,r} = \max_{h:\cA\mapsto \cA}\sum_{t\in (p,r)} \left(U(h(a_t),p_t, y_t) - U(a_t,p_t,y_t)\right)
    \end{equation*}
    denote the swap regret for the Agent over the subsequence in which $(p_t,r_t) = (p,r)$
    and let $\ir = \sum_{(p,r)\in \cP_\cO\times \cA} \ir_{p,r}$ denote the total swap regret (for the action sequence $a_{1:T}$).

In the rounds in which $(p_t,r_t) = (p,r)$, similar to Eq~\eqref{eq:u-gap}, we have
    \begin{align*}
        &\sum_{t\in (p,r)}U(\hat \mu_{p,r},p,\pip_t) \\
        \geq &\sum_{t\in (p,r)}U(\hat \mu_{p,r},p,y_t) - \alpha(E_{1,p,r})  T \tag{$\cE_1$-bias} \nonumber\\
        \geq &\sum_{t\in (p,r)}U(a_t,p,y_t) -n_{p,r}b_{p,r} - \alpha(E_{1,p,r})T \tag{no secret info} \nonumber\\
        \geq &\sum_{t\in (p,r)}U( r,p,y_t)-\ir_{p,r}-n_{p,r}b_{p,r}  -\alpha(E_{1,p,r})  T\tag{definition of $\ir^{p_0}_{r}$}\nonumber\\
        \geq &\sum_{t\in (p,r)}U( r,p,\pi_t)-\ir_{p,r}-n_{p,r}b_{p,r}  -2\alpha(E_{1,p,r})  T\tag{$\cE_1$-bias}\,. 
    \end{align*}
    Since $p$ is $(\beta,\gamma)$-stable under $\pi_t$ for all $t\in (p,r)$, we have $U(a,p,\pip_t)\leq U(r,p,\pip_t)-\beta$ or $V(a,p,\pip_t)\geq V(r,p,\pip_t)-\gamma$ for all $a\neq r$ in $\cA$.
    Let $\rho_{p,r,t} = \Pr_{a\sim \hat \mu_{p,r}}(U(a,p,\pip_t)\leq U(r,p,\pip_t)-\beta)$ denote the probability of $U(a,p,\pip_t)\leq U(r,p,\pip_t)-\beta$ for $a\sim \hat \mu_{p,r}$.
    By combining with $U(a,p,\pip_t)\leq U(r,p,\pip_t)$ for all $a\in \cA$, we have 
    \begin{equation}
        \sum_{t\in (p,r)}\rho_{p,r,t} \leq \frac{\ir_{p,r}+n_{p,r}b_{p,r}  +2\alpha(E_{1,p,r})  T}{\beta}\,.\label{eq:rho}
    \end{equation}
    Therefore, we have
    \begin{align*}
        \text{Term (b)} =& \sum_{t=1}^T (V(r_t,p_t,y_t)-V(a_t,p_t ,y_t))\\
        \leq&\sum_{(p,r)\in \cP_\cO\times \cA} \sum_{t\in (p,r)} (V(r,p,y_t)-V(\hat \mu_{p,r},p ,y_t)) + \sum_{(p,r)\in \cP_\cO\times \cA} n_{p,r}b_{p,r}\tag{no secret info}\\
        \leq &\sum_{(p,r)\in \cP_\cO\times \cA} \sum_{t\in (p,r)} V(r,p,\pi_t)-V(\hat \mu_{p,r},p ,\pi_t) + 2\alpha(\cE_1) T +\sum_{(p,r)\in \cP_\cO\times \cA} n_{p,r}b_{p,r}\tag{$\cE_1$-bias}\\
        \leq& \gamma T + \sum_{(p,r)\in \cP_\cO\times \cA}\sum_{t\in (p,r)}\rho_{p,r,t} + 2\alpha(\cE_1) T+\sum_{(p,r)\in \cP_\cO\times \cA} n_{p,r}b_{p,r}\tag{stability of $p$}\\
        \leq & \gamma T + \frac{\ir +\sum_{(p,r)\in \cP_\cO\times \cA}n_{p,r}b_{p,r}  +2\alpha(\cE_1)  T}{\beta}+ 2\alpha(\cE_1) T+\sum_{(p,r)\in \cP_\cO\times \cA} n_{p,r}b_{p,r}\,.\tag{Apply Eq~\eqref{eq:rho}}
    \end{align*}  
    Hence, by taking the expectation over the randomness of the Agent's algorithm $\cL$, we have
    \begin{align*}
        \EEs{\cL}{\text{Term (b)}} \leq \left(\gamma + \frac{\eint +\cO(\sqrt{\abs{\cP_\cO}\abs{\cA}/T})+2\alpha(\cE_1)  }{\beta}+ 2\alpha(\cE_1)+\cO(\sqrt{\abs{\cP_\cO}\abs{\cA}/T})\right)T\,.
    \end{align*}
\end{enumerate}
Now we have the Principal's regret upper bounded by 
\begin{align*}
    &\textrm{PR}(\cO_{c,\epsilon,\beta,\gamma},\pi_{1:T}, \cL, y_{1:T})\\
    \leq & c+\alpha(\cE_2)+\alpha(\cE_1) 
    \\& \gamma + \frac{\eint +\cO(\sqrt{\abs{\cP_\cO}\abs{\cA}/T})+2\alpha(\cE_1)  }{\beta}+ 2\alpha(\cE_1)+\cO(\sqrt{\abs{\cP_\cO}\abs{\cA}/T})
    \\ & \cO(\sqrt{\abs{\cA}/T})+ \frac{\eint + \cO(\sqrt{\abs{\cA}/T})+2\max_{p_0\in \cP_0}\alpha(\cE_{3,p_0}) }{\epsilon} +  \max_{p_0\in \cP_0}\alpha(\cE_{3,p_0}) + \alpha(\cE_2)
    \\
    = & c+ 3\alpha(\cE_1) + 2\alpha(\cE_2) + \max_{p_0\in \cP_0}\alpha(\cE_{3,p_0}) + \gamma + \frac{\eint +\cO(\sqrt{\abs{\cP_\cO}\abs{\cA}/T})+2\alpha(\cE_1)  }{\beta} \\
    & + \frac{\eint + \cO(\sqrt{\abs{\cA}/T})+2\max_{p_0\in \cP_0}\alpha(\cE_{3,p_0}) }{\epsilon} \,.
\end{align*}
Since $\abs{\cP_0}$ and $\abs{\cA}$ are $\Theta(1)$, we have
\begin{align*}
    &\textrm{PR}(\cO_{c,\epsilon,\beta,\gamma},\pi_{1:T}, \cL, y_{1:T}) \\
    = &\cO(c+ \abs{\cP_\cO}\alpha + \gamma +\frac{\eint +\sqrt{\abs{\cP_\cO}/T}+\abs{\cP_\cO}\alpha }{\beta} + \frac{\eint + \sqrt{1/T}+\alpha }{\epsilon} +\sqrt{\abs{\cP_\cO}/T})\,.
\end{align*}
\end{proof}
\section{Proofs from Section \ref{sec:linear-contract}}

\lmmlinearstable*
\begin{proof}[Proof of Lemma~\ref{lmm:linear-stable}]
The intuition for this stability result is that, for any policy returned, either the Agent has a unique best response that gets him a payoff $\frac{\beta \Delta_{c}}{2}$ higher than all other actions, or the Principal is completely indifferent between what actions the Agent selects. We first show that there must be a policy with such a unique best response in the interval $[p^\text{optimistic}, p^\text{optimistic} + \abs{\cA}(\beta + \delta)]$, as long as this interval lies fully within the linear contract policy space of $[0,1]$, i.e., $p^\text{optimistic} + \abs{\cA}(\beta + \delta)\leq 1$. To do this, we take advantage of the fact that for a fixed $\pi$, there are a bounded number of policies which induce ties between actions (Lemma~\ref{lem:ties}), and for all policies far enough away from these policies, the Agent actions are well-separated (Lemma~\ref{lem:gap}). When $p^\text{optimistic}$ is larger than $1 - |\cA|(\beta + \delta)$, we no longer have this guarantee--however, if the Principal does not return a $(\frac{\beta \Delta_c}{2},0)$-stable policy in this case, she will return $p(\pi)= 1$, which is still close to $p^\text{optimistic}$, and furthermore gets the Principal a payoff of $0$ regardless of what action the Agent takes, leading her to be indifferent to the Agent's action.

\begin{restatable}{lemma}{lemties}\label{lem:ties}
For any $\pi$, there are at most $|\cA|-1$ linear contracts resulting in more than one best response for the Agent, i.e.: 
$$\abs{\{p\in \cP|\cB(p,\pi,0)| >1\}}\leq |\cA|-1\,.$$ 
\end{restatable}

\begin{restatable}{lemma}{lemgap}\label{lem:gap}

For any prior $\pi$ and any $\bar p\in [0,1]$, if $a^*$ is an Agent's best response to both $(\bar p - \beta, \pi)$, and  $(\bar p + \beta, \pi)$, then $U(a^*,\bar p,\pi) \geq U(a,\bar p,\pi) + \Delta_c \cdot \beta$, for all actions $a \neq a^*$.  
\end{restatable}
Now we start formally proving Lemma~\ref{lmm:linear-stable}. There are two cases:
\begin{itemize}
\item $p^\text{optimistic} \leq 1 - |\cA|(\beta + \delta)$. Then, let us consider the policies in the range $[p^\text{optimistic}, p^\text{optimistic} + \abs{\cA}(\beta + \delta)]$ for which the Agent has more than one optimal response. Call this set $s$. By Lemma~\ref{lem:ties}, we have $|s| \leq |\cA|-1$. Note that, by the definition of $s$, for any given $i\in [|s|]$, all policies $p \in (s_{i}, s_{i+1})$ (where $s_i$ is the $i$-th smallest element in $s$) must lead to a unique best response action for the Agent, and must lead to the same best response as each other by the continuity of the Agent's utility with respect to the Principal policy.

Now, let's augment $s$ with the endpoints of the interval by letting $s' = \{p^\text{optimistic}\} \cup s \cup \{p^\text{optimistic} + \abs{\cA}(\beta\ + \delta)\}$.
For any $i\in [|s'|]$, let $s'_i$ denote the $i$-th smallest element in $s'$.
We will lower bound the largest gap between any two neighboring policies in $s'$.
\begin{align*}
 \argmax_{i \in [|s'|-1]}(s'_{i+1}-s'_{i}) & \geq \frac{s'_{|s'|} - s'_{1}}{|s'|-1} = \frac{|\cA|(\beta + \delta)}{|s'|-1} \geq  \frac{|\cA|(\beta + \delta)}{|s|+1}\geq \beta + \delta \,,
\end{align*}
where the last inequality applies Lemma~\ref{lem:ties}.

Hence, there exists an $i\in [|s'|-1]$ such that $s'_{i+1}-s'_{i} \geq \beta + \delta$. Now, consider any policy $p \in [s'_{i} + \frac{\beta}{2}, s'_{i+1} - \frac{\beta}{2}]$. By Lemma~\ref{lem:gap}, we have $U(a^{*}(p,\pi),p,\pi) \geq U(a,p,\pi) + \frac{\Delta_c \beta}{2}$ for all $a \neq a^{*}(p,\pi)$. Therefore, every policy in this range is $(\frac{\Delta_c \beta}{2},0)$-stable under $\pi$. As this range is of size at least $\delta$, there must be at least one policy $p \in \cP_{\delta}$ in the range $[p^\text{optimistic}, p^\text{optimistic} + \abs{\cA}(\beta + \delta)]$ that is $(\frac{\Delta_c \beta}{2},0)$-stable under $\pi$. By the definition of the algorithm, the returned $p(\pi)$ is $(\frac{\Delta_c \beta}{2},0)$-stable under $\pi$ and is in the range $[p^\text{optimistic}, p^\text{optimistic} + \abs{\cA}\beta + \delta]$.

\item $p^\text{optimistic} \geq 1 - |\cA|(\beta + \delta)$. Then the returned policy must be in the range $[p^\text{optimistic}, p^\text{optimistic} + \abs{\cA}(\beta + \delta)]$. If some $p(\pi) < 1$ is returned, by the definition of the algorithm, it will be $(\frac{\beta \Delta_{c}}{2},0)$-stable. Otherwise, the algorithm returns $p(\pi) = 1$, and we have that 
\begin{align*}
V(a^{*}(p(\pi),\pi),p(\pi),\pi)  
& = (1-p(\pi)) \cdot f(a^{*}(p(\pi),\pi),\pi) = 0  \leq V(a,p(\pi),\pi)\,,
\end{align*}
for any $a \in \A$. Thus, in this case we have that $ V(a,p(\pi),\pi)\geq V(\brr(p(\pi),\pi),p(\pi),\pi) - 0$, and thus the policy is also $(\frac{\beta \Delta_{c}}{2},0)$-stable. Furthermore, in this case the returned $p(\pi)$ is also in the range $[p^\text{optimistic}, p^\text{optimistic} + \abs{\cA}(\beta + \delta)]$.
\end{itemize}
This completes the proof of Lemma~\ref{lmm:linear-stable}.
\end{proof}

\lmmlinearopt*
\begin{proof}[Proof of Lemma~\ref{lmm:linear-opt}]
Now we move on to prove Lemma~\ref{lmm:linear-opt}.
For this part, we must upper bound the difference between the Principal's utility under the policy $p=p(\pi)$ returned by Algorithm~\ref{alg:linear-oracle} and her utility under the best benchmark policy $p_{0}$. 
To do this, 
we compare the utility of the Principal under $p^\text{optimistic}$ to her utility under $p$, taking advantage of the fact that $p$ is not much larger than $p^\text{optimistic}$. We crucially make use of the monotone relationship between $p$ and $f(\pi, a^{*}(p,\pi,\epsilon))$ for linear contracts (Lemma~\ref{lem:monotonicity}).

\begin{restatable}{lemma}{lemmonotonicity}
    \label{lem:monotonicity}
For any two linear contracts $p_{1} \geq p_{2}$, $$\max_{a \in \cB(p_{1},\pi,\epsilon)}f(\pi,a) \geq \max_{a \in \cB(p_{2},\pi,\epsilon)}f(\pi,a)$$ for all $\pi$ and all $\epsilon \geq 0$.
\end{restatable}

We consider two cases: $p(\pi)<1$ and $p(\pi)=1$.
\begin{itemize}
    \item $p(\pi)< 1$. Since $p(\pi)$ is $(\frac{\Delta_{c}\beta}{2},0)$-stable according to Lemma~\ref{lmm:linear-stable}, then for all $a \neq a^{*}(p,\pi)$, either $ U(a,p,\pi) \leq  U(a^{*}(p,\pi),p,\pi) - \frac{\Delta_c \beta}{2}$  or $V(a,p(\pi),\pi) = V(a^{*}(p(\pi),\pi),p(\pi),\pi)$. 
    For all $a$ with $V(a,p(\pi),\pi) = V(a^{*}(p(\pi),\pi),p(\pi),\pi)$, we have $f(\pi,a) = \frac{V(a,p(\pi),\pi)}{1-p(\pi)} = f(\pi,a^*(p(\pi),\pi))$. Therefore, we have 
    \begin{equation}\label{eq:maxball}
        \max_{a \in \mathcal{B}(p(\pi),\pi,\frac{\Delta_{c}\beta}{2})}f(\pi,a) = f(\pi,a^{*}(p(\pi),\pi))\,.
    \end{equation}
\begin{align*}
 &\max_{p_0 \in \cP_{0}}V(a^*(p_0,\pi,\frac{\Delta_{c}\beta}{2}), p_0,\pi) \\
 = &  V(a^*(p^\text{optimistic},\pi,\frac{\Delta_{c}\beta}{2}),p^\text{optimistic},\pi) \tag{ Definition of $p^\text{optimistic}$} \\
= & \max_{a \in \mathcal{B}(p^\text{optimistic},\pi,\frac{\Delta_{c}\beta}{2})}V(a,p^\text{optimistic},\pi)  \\ 
= & (1 - p^\text{optimistic})\max_{a \in \mathcal{B}(p^\text{optimistic},\pi,\frac{\Delta_{c}\beta}{2})}f(\pi,a)\tag{Applying Eq~\eqref{eq:expectedV}}\\ 
\leq & (1 - p^\text{optimistic})\max_{a \in \mathcal{B}(p(\pi),\pi,\frac{\Delta_{c}\beta}{2})}f(\pi,a) \tag{Applying Lemma~\ref{lem:monotonicity}, as $p(\pi) \geq p^\text{optimistic}$}\\ 
= & (1 - p^\text{optimistic})f(\pi,a^{*}(p(\pi),\pi)) \tag{Applying Eq~\eqref{eq:maxball}}\\
\leq & (1 - p(\pi) + |\cA|(\beta + \delta))f(\pi,a^{*}(p(\pi),\pi))) \tag{Applying the gap condition in Lemma~\ref{lem:stable_contract}}\\
= & V(a^*(p,\pi),p,\pi) +|\cA|(\beta + \delta) \cdot f(a^{*}(p,\pi),p,\pi)  \\
\leq & V(a^*(p,\pi),p,\pi) + |\cA|(\beta + \delta)\,.
\end{align*}
\item $p(\pi) = 1$. In this case, we have
$p^\text{optimistic}\geq 1-|\cA|(\beta+\delta)$.
Then we have
\begin{align*}
    &\max_{p_0 \in \cP_{0}}V(a^*(p_0,\pi,\frac{\Delta_{c}\beta}{2}), p_0,\pi)\\
    =&V(a^*(p^\text{optimistic},\pi,\frac{\Delta_{c}\beta}{2}),p^\text{optimistic},\pi) \\
    \leq & 1-p^\text{optimistic}\\
    \leq &|\cA|(\beta+\delta) \leq V(a^*(p,\pi),p,\pi) + |\cA|(\beta+\delta)
\end{align*}
\end{itemize}

This completes the proof of Lemma 2.
\end{proof}

\begin{lemma}
For any $\pi$, and for any Agent actions $a_{1}$ and $a_{2}$ s.t. $a_{1} \neq a_{2}$, there is a unique linear contract $p$ such that $$U(a_{1},p,\pi) = U(a_{2},p,\pi)$$
\label{lem:tie}
\end{lemma}

\begin{proof}
In order for two Agent actions to give the same payoff, we need a $p$ such that

\begin{align*}
& pf(\pi, a_{1}) - c(a_{1}) = pf(\pi, a_{2}) - c(a_{2})  \\
& p = \frac{c(a_{1}) - c(a_{2})}{f(\pi, a_{1}) - f(\pi, a_{2})}
\end{align*}

If $f(\pi, a_{1}) - f(\pi, a_{2}) \neq 0$ this expression is well defined and has a unique solution, and therefore there can be at most one $p$ for which this is true. If $f(\pi, a_{1}) = f(\pi, a_{2})$, then 
\begin{align*}
& pf(\pi, a_{1}) - c(a_{1}) = pf(\pi, a_{1}) - c(a_{2})  \\
& \Leftrightarrow c_{a_{1}} = c_{a_{2}}
\end{align*}

This is a contradiction, as we assume all costs are separated by $\Delta_{c} \geq 0$. Therefore this expression must be well defined and have a unique solution.
\end{proof}

\lemties*

\begin{proof}
To show this, we will first show that for any Agent action $a^{*}$, there are at most $2$ policies for which $a^{*}$ a non-unique best response. To see this, let's consider the smallest linear contract $p_{1}$ such that $a$ is a best response. Let us also consider the largest linear contract $p_{2}$ such that $a$ is a best response. We will show that for all $p$ such that $p_{1} < p < p_{2}$, $a$ is a unique best response. 

As $a^{*}$ is a best response to $p_{1}$, we have

\begin{align*}
& U(a^{*},p_{1},\pi) = \max_{a \in \cA}U(a,p_{1},\pi) \\
& \Leftrightarrow p_{1} f(\pi, a^{*}) - c(a^{*}) = \max_{a \in \cA}(p_{1} f(\pi, a) - c(a) )
\end{align*}

Similarly,

\begin{align*}
& p_{2} f(\pi, a^{*}) - c(a^{*}) = \max_{a \in \cA}(p_{2} f(\pi, a) - c(a) )
\end{align*}

Combining these, we get that, for any $x \in [0,1]$: 

\begin{align*}
& (x p_{1} + (1-x) p_{2})\cdot f(\pi, a^{*}) - c(a^{*}) = x \cdot \max_{a \in \cA}(p_{1} f(\pi, a) - c(a) ) + (1-x) \cdot \max_{a \in \cA}(p_{2} f(\pi, a) - c(a) )
\end{align*}

Now, consider any action $\bar a \neq a^{*}$, evaluated on the linear contract defined by  $(x p_{1} + (1-x) p_{2})$. Assume for contradiction that $\bar a$ is optimal on this contract. Then we have that 

\begin{align*}
& (x p_{1} + (1-x) p_{2})\cdot f(\pi, \bar a) - c(\bar a) = x(p_{1}f(\pi, \bar a) -  c(\bar a)) + (1-x)(p_{2}f(\pi, \bar a) -  c(\bar a))\\
& \geq x \cdot \max_{a \in \cA}(p_{1} f(\pi, a) - c(a) ) + (1-x) \cdot \max_{a \in \cA}(p_{2} f(\pi, a) - c(a) ) \\
\end{align*}

Therefore, it must be the case that $\bar a$ is optimal at $p_{1}$ and $p_{2}$. So $a^{*}$ and $\bar a$ have the same Agent utility at $2$ different contracts. But this is a contradiction of Lemma~\ref{lem:tie}. Therefore any action can be non-uniquely optimal at at most $2$ contracts, the smallest contract at which it is optimal and the largest contract at which it is optimal. At $p = 0$, the optimal action must be the cheapest action, which by our assumption is unique. Therefore there is at least one action that is uniquely optimal at its smallest optimal contract and can only be non-uniquely optimal at $1$ contract. So the total number of contracts with multiple optimal actions is at most
$$\frac{2(|\cA| - 1) + 1}{2}$$
The largest integer value this could be is $|\cA| - 1$, completing our proof.

\end{proof}

\lemgap*

\begin{proof}
Consider any action $a \neq a^{*}$, and the linear contract $\hat p$ such that $U(a, \hat p, \pi) = U(a^{*}, \hat p, \pi)$. Then,

\begin{align*}
& \hat p f(\pi, a) - c(a) = \hat p f(\pi, a^{*}) - c(a^{*}) \\
& \Rightarrow  \hat p (f(\pi, a) - f(\pi, a^{*})) =  c(a)- c(a^{*}) \\
& \Rightarrow  |\hat p (f(\pi, a) - f(\pi, a^{*}))| =  |c(a)- c(a^{*})| \geq \Delta_c \\
& \Rightarrow  |f(\pi, a) - f(\pi, a^{*})| \geq  \Delta_c \\
\end{align*}

At linear contract $\bar p$, the payoff of $a$ is 

\begin{align*}
& U(a, \bar p, \pi) = \bar p f(\pi, a) - c(a) \\
& = (\bar p - \hat p) f(\pi, a) + \hat p f(\pi, a) - c(a) \\
& = (\bar p - \hat p) f(\pi, a) + \hat p f(\pi, a^{*}) - c(a^{*}) = (\bar p - \hat p) f(\pi, a) + U(a^{*},\hat p, \pi) \tag{By the definition of $\hat p$} \\
\end{align*}

Furthermore, we know that 

\begin{align*}
& U(a^{*},\bar p, \pi) = \bar p f(\pi, a^*) - c(a^*) \\
& = (\bar p - \hat p) f(\pi, a^*) + \hat p f(\pi, a^{*}) - c(a^{*}) = (\bar p - \hat p) f(\pi, a^*) + U(a^*, \hat p, \pi)
\end{align*}

Combining these, we get that 

\begin{align*}
& U(a^{*},\bar p, \pi) - U(a, \bar p, \pi) = (\bar p - \hat p) (f(\pi, a^*) - f(\pi, a)) \\
&= |(\bar p - \hat p)| \cdot |(f(\pi, a^*) - f(\pi, a))| \tag{As $a^*$ is optimal at $\bar p$, and thus this difference cannot be negative}\\
& \geq \beta \cdot \Delta_c \\
\end{align*}

\end{proof}

\lemmonotonicity*

\begin{proof}
Let $a_{1} = \max_{a \in \cB(p_{1},\pi,0)}f(\pi,a)$, let $a_{1,\epsilon} = \max_{a \in \cB(p_{1},\pi,\epsilon)}f(\pi,a)$ and let \\$a_{2,\epsilon} = \max_{a \in \cB(p_{2},\pi,\epsilon)}f(\pi,a)$. Note that $a_{1}$ is the Agent's exact best response action under $p_{1}$ which is best for the Principal, while $a_{1,\epsilon}$ and $a_{2,\epsilon}$ are the Agent's $\epsilon$-approximate best response actions which are best for the Principal, under their respective policies. This, we can restate our lemma as proving that for any two linear contracts $p_{1}$, $p_{2}$ s.t. $p_{1} \geq p_{2}$, $f(\pi, a_{1,\epsilon}) \geq f(\pi, a_{2,\epsilon})$.


Assume for contradiction that this is not the case, and $f(\pi, a_{1,\epsilon}) < f(\pi, a_{2,\epsilon})$. Then it must be that $a_{2,\epsilon} \notin \cB(p_{1},\pi,\epsilon)$, as otherwise we would have that 

\begin{align*}
& f(\pi, a_{2,\epsilon}) > f(\pi, a_{1,\epsilon})  \\
& \geq f(\pi, a_{2,\epsilon}) \tag{By the fact that $a_{2,\epsilon} \in \cB(p_{1},\pi,\epsilon)$ and $a_{1,\epsilon}$ is optimal over all $\cB(p_{1},\pi,\epsilon)$} \\
\end{align*}
This is a contradiction. 

As $a_{2,\epsilon} \notin \cB(p_{1},\pi,\epsilon)$, $a_{2,\epsilon}$ is not an $\epsilon$-approximate best response to $p_{1}$. So we have that





 

\begin{align*}
&p_{1}f(\pi,a_{1}) - c(a_{1}) > p_{1}f(\pi,a_{2,\epsilon}) - c(a_{2,\epsilon}) + \epsilon \\
& \Leftrightarrow  c(a_{2,\epsilon}) - c(a_{1}) - \epsilon > p_{1}f(\pi,a_{2,\epsilon}) - f(\pi,a_{1}))  
\end{align*}

Furthermore, as $a_{2,\epsilon}$ is an $\epsilon$-approximate best response under $p_{2}$, we have that 

\begin{align*}
&p_{2}f(\pi,a_{1}) - c(a_{1}) \leq p_{2}f(\pi,a_{2,\epsilon}) - c(a_{2,\epsilon}) + \epsilon \\
& \Leftrightarrow c(a_{2,\epsilon}) - c(a_{1}) - \epsilon \leq p_{2}(f(\pi,a_{2,\epsilon}) - f(\pi,a_{1})) \\
\end{align*}

Finally, we note that 

\begin{align*}
& f(\pi,a_{2,\epsilon}) - f(\pi,a_{1}) \geq f(\pi,a_{2,\epsilon}) - f(\pi,a_{1,\epsilon}) \tag{As $a_{1,\epsilon}$ is maximizing over a larger set} \\
& > 0 \tag{By our assumption}
\end{align*}

Putting these together, we get that 

\begin{align*}
& p_{2}(f(\pi,a_{2,\epsilon}) - f(\pi,a_{1})) > p_{1}(f(\pi,a_{2,\epsilon}) - f(\pi,a_{1}))  \\
& \Rightarrow p_{2} > p_{1} \\
\end{align*}

We have derived a contradiction, completing our proof.
\end{proof}

\section{More Details and Proofs from Section~\ref{sec:bayes}}\label{app:bayes}
\subsection*{Discretization details}\label{app:bayes-discrete}
Recall the explicit representation of the signal scheme in Eq~\eqref{eq:signal}. Note that each signal scheme selected under our construction of $p'$ selects two strategies in $\cS$ and each distribution $p'(\cdot|y)$ is supported only on these two strategies.
Now we want to discretize $p'(\cdot|y)$.
For some discretization precision $\delta\ll \beta$ with $\frac{1}{\delta}\in \NN_+$, let $\varphi_{i,j,k_0,k_1}$ for $i,j\in [n], k_0,k_1\in \{0,1,\ldots, \frac{1}{\delta}\}$  represent the signal scheme with 
\[\varphi(s_i|y=1) = k_0\delta \,, \qquad \varphi(s_i|y=0) = k_1\delta\,.\]
Then we let $\cP_\delta = \{\varphi_{i,j,k_0,k_1}| i,j\in [n], k_0,k_1\in  \{0,1,\ldots, \frac{1}{\delta}\}\}$  denote the set of all such signal schemes. We have $\abs{\cP_\delta} = \cO(\frac{n^2}{\delta^2})$. We will return the signal scheme $p_\delta(\mu) \in P_\delta$ closest to $p'(\mu)$. 
Recall that our definition of $p'(\mu)$ induces a convex combination of two points in $\text{Ex}'$, saying $\mu = \tau \cdot \mu_{k}' + (1-\tau) \cdot \mu'_{l}$. Then the explicit form of $p'(\mu)$ is
    \begin{align*}
         p(s_{i_{k}}|y=1) = \frac{\tau \cdot \mu_{k}'}{\mu}\,, \quad &p(s_{i_{l}}|y=1) = \frac{(1-\tau) \cdot \mu_{l}'}{\mu}\\
          p(s_{i_{k}}|y=0) = \frac{\tau \cdot (1-\mu_{k}')}{1-\mu}\,, \quad &p(s_{i_{l}}|y=0) = \frac{(1-\tau) \cdot (1-\mu_{l}')}{1-\mu}\,.
    \end{align*}
    By rounding these two probabilities, we obtain a discretized signal scheme $p_{\delta}(\mu)$ with
    \begin{align*}
         p_\delta(s_{i_{k}}|y=1) = \delta \cdot \argmin_{k\in \{0,\ldots,1/\delta\}} \abs{k\delta -p(s_{i_{k}}|y=1) }\,, \\
         p_\delta(s_{i_{k}}|y=0) =\delta \cdot \argmin_{k\in\{0,\ldots,1/\delta\}} \abs{k\delta -p(s_{i_{k}}|y=0) }\,.
    \end{align*}


\subsection*{Proofs}
For any signal $p$ and any prior distribution $\pi = \Ber(\mu)$, let $\{(\tau_i, \Ber(\mu_i))\}_{i\in [n]}$ denote the induced distribution of posteriors where $\tau_i = \sum_{y\in \cY}p(s_i|y)\pi(y)$ is the probability of the signal being $s_i$ and $\Ber(\mu_i)$ is the posterior distribution $\pi(y|s_i)$ of $y$ given the signal $s_i$. 
Then the expected Principal's utility is
\begin{equation*}
    V(a,p,\mu):=\EEs{y\sim \Ber(\mu)}{V(a,p,y)}=\EEs{y\sim \Ber(\mu)}{\EEs{s\sim p(\cdot|y)}{v(a(s),y)}} = \sum_{i\in [n]}\tau_i v(a(s_i))\,,
\end{equation*}
and the expected Agent's utility is 
\begin{equation*}
    U(a,p,\mu):=\EEs{y\sim \Ber(\mu)}{U(a,p,y)}=\EEs{y\sim \Ber(\mu)}{\EEs{s\sim p(\cdot|y)}{u(a(s),y)}} = \sum_{i\in [n]}\tau_i u(a(s_i),\mu_i)\,.
\end{equation*}
Hence, the best response $a^*(p,\mu)$ is defined by letting $a^*(p,\mu)(s_i) = s^*(\mu_i)$ and an action $a$ is an $\epsilon$-best response if $\sum_{i\in[n]} \tau_i u(a(s_i), \mu_i) \geq \sum_{i\in [n]} \tau_i u(s^*(\mu_i), \mu_i)- \epsilon$. Then we first introduce the following lemma to prove our results in Bayesian Persuasion.

\begin{lemma}\label{lmm:bayes-stable}
    For any $x\in [0,1]$, for any $\mu\in [0,1]$, a signal scheme $p$, which induces distribution of posteriors as $(\tau, w_i), ((1-\tau),w_j)$ with $ w_i\in S_i$ and $w_j\in S_j$, is $(x\cdot \eta, x)$-stable under $\mu$ for any $\eta$ with $[w_i-\eta,w_i +\eta]\subset S_i$ and $[w_j-\eta,w_j +\eta]\subset S_j$.
\end{lemma}
    
\begin{proof}
By Assumption \ref{asp-bayes:interval}, each interval has a length of at least $C$. Then for any $i\in n$ and any $\eta < \frac{C}{2}$, let $S_i^\eta$ denote the interval $[\min(S_i)+\eta,\max(S_i)-\eta]$ by removing $\eta$ top values and $\eta$ bottom values from the interval $S_i$. Then for all $\mu\in S_i^\eta$, we have
\begin{align*}
    u(s_j, \mu) \leq u(s_i, \mu) - c_1 \eta\,,
\end{align*}
for all $j\neq i$. 
This directly follows from Assumption \ref{asp-bayes:interval}. As mentioned before, by Assumption \ref{asp-bayes:interval}, there is some minimum difference $c_{1}$ between the utility slopes $\partial u(s,\cdot)$ of any two strategies. Hence for any $\mu$ which is $\eta$-far away from an interval edge, we can see that the Agent utility of every strategy $s_j$ other than the optimal strategy $s_i$ at $\mu$ is at least $c_{1}\cdot \eta$ lower. 
Hence, taking any strategy other than $s_i$ after seeing signal $s_i$ would achieve a utility at least $c_1 \eta$ lower under $w_i$.


If the action $a$ taken by the Agent plays a non-optimal strategy to both $s_i$ and $s_j$, it leads to an expected loss for the Agent of $\geq \tau \cdot c_{1} \eta + (1-\tau) \cdot c_{1} \eta = c_{1}\eta$. More formally, $U(a, p, \mu)\leq U(a^*(p,\mu), p,\mu) - c_1 \eta$. Thus, for any $x\in [0,1]$, we have
\begin{align*}
    U(a, p, \mu)\leq U(a^*(p,\mu),p,\mu) - x\cdot c_1 \eta\,.
\end{align*}

Now consider action $a$ playing one optimal response and one non-optimal response. W.l.o.g., assume that $a(s_i) \neq s_i$ and $a(s_j) = s_j$.
Then we have
\begin{align*}
    U(a,p,\mu) &\leq U(a^*(p,\mu), p,\mu) -\tau c_1 \eta\,,\\
    V(a,p,\mu) &\geq V(a^*(p,\mu), p,\mu) - \tau\,.
\end{align*}
Hence, for any $x\in [0,1]$, if $\tau\leq x$, we have
\begin{align*}
    V(a,p,\mu) \geq V(a^*(p,\mu), p,\mu) - x\,.
\end{align*}
If $\tau>x$, we have
\begin{align*}
    U(a,p,\mu) \leq U(a^*(p,\mu), p,\mu) -x\cdot c_1 \eta\,.
\end{align*}
By combining the two cases, we have proved the lemma.
\end{proof}

\subsubsection{Proof of Lemma~\ref{lmm:bayes-alt}}
\lmmbayesalt*
Before the proof, we first introduce the following lemma.
\begin{lemma}\label{lmm:stabilized}
For any $\mu\in [0,1]$,  we have
    \begin{equation*}
        V(a^*(p'(\mu),\mu),p'(\mu),\mu) \geq V(a^*(p,\mu,\epsilon),p,\mu) -\frac{3\beta}{C} - c_2\sqrt{\epsilon}\,,
    \end{equation*}
    for all $p\in \cP$.
\end{lemma}



\begin{proof}
The proof is decomposed to two parts.
\begin{itemize}
    \item $V(a^*(p'(\mu),\mu),p',\mu)\geq v^*(\mu)- \frac{3\beta}{C}$. (Lemma~\ref{lmm:tech3})
    \item There exists a constant $c_2$ such that $V(a^*(p,\mu,\epsilon),p,\mu) \leq v^*(\mu)+ c_2\sqrt{\epsilon}$ for all $p\in \cP$. (Lemma~\ref{lmm:tech4})
\end{itemize}
By combining these two parts, we prove Lemma~\ref{lmm:stabilized}.

\begin{lemma}\label{lmm:tech3}
For any $\mu\in [0,1]$, we have $V(a^*(p',\mu),p',\mu)\geq v^*(\mu)- \frac{3\beta}{C}$ where $p'=p'(\mu)$.
\end{lemma}
\begin{proof}[Proof of Lemma~\ref{lmm:tech3}]
Recall that the method of finding the optimal achievable Principal's utility by \cite{kamenica2011bayesian}, we have $(\mu,v^*(\mu)) =\tau (\mu_{i_j}, v(s_{i_j})) + (1-\tau)(\mu_{i_{j+1}}, v(s_{i_{j+1}}))$.
Now considering our signal scheme $p'$, there are two cases.

\paragraph{Case 1} The prior $\mu$ lies in $[\mu_{i_j}',\mu_{i_{j+1}}']$ with $\mu = \tau' \mu_{i_j}'+ (1-\tau')\mu_{i_{j+1}}'$.
Recalling our definition of $p'$ (where we find the optimal convex combination of points in $\textrm{Ex}'$), we must have 
$$V(a^*(p',\mu),p',\mu) \geq \tau' v(s_{i_j})+ (1-\tau')v(s_{i_{j+1}})\,.$$
Since $\mu = \tau' \mu_{i_j}'+ (1-\tau')\mu_{i_{j+1}}'$ and $\mu = \tau \mu_{i_j}+ (1-\tau)\mu_{i_{j+1}}$, we have 
\begin{equation*}
    \tau (\mu_{i_{j+1}}-\mu_{i_j}) - \tau' (\mu_{i_{j+1}}'-\mu_{i_j}') = \mu_{i_{j+1}}- \mu_{i_{j+1}}'\,.
\end{equation*}
According to the definition of $\mu'$s, we have
\begin{align*}
     \mu_{i_{j+1}}-\mu_{i_j} + 2\beta \leq \mu_{i_{j+1}}'-\mu_{i_j}' \leq \mu_{i_{j+1}}-\mu_{i_j} + 2\beta \,.
\end{align*}
Therefore, we have
\begin{align*}
    \abs{\tau-\tau'} \leq \frac{\abs{\mu_{i_{j+1}}- \mu_{i_{j+1}}'} + \tau' \cdot 2\beta}{\mu_{i_{j+1}}-\mu_{i_j}}\,.
\end{align*}
According to Assumption \ref{asp-bayes:interval} and definition of $\mu'$s, we have
\begin{align*}
    &\mu_{i_{j+1}}-\mu_{i_j} \geq C\,,\\
    &\\
    &\abs{\mu_{i_{j+1}}- \mu_{i_{j+1}}'}\leq\beta\,.
\end{align*}
Hence, we have $\abs{\tau-\tau'}\leq \frac{3\beta}{C}$. Thus, we have
\begin{align*}
    V(a^*(p',\mu),p',\mu) \geq &\tau' v(s_{i_j})+ (1-\tau')v(s_{i_{j+1}}) \geq  \tau v(s_{i_j})+ (1-\tau)v(s_{i_{j+1}}) -\frac{3\beta}{C} \\
    = &v^*(\mu)-\frac{3\beta}{C}\,. 
\end{align*}

\paragraph{Case 2 }
The prior $\mu$ does not lie in $[\mu_{i_j}',\mu_{i_{j+1}}']$.
Since $\mu \in [\mu_{i_j}, \mu_{i_{j+1}}]$, we have $\mu$ lies in either $[\mu_{i_j},\mu_{i_j}')$ or $(\mu_{i_{j+1}}', \mu_{i_{j+1}}]$. 
W.l.o.g., suppose that $\mu$ lies in $[\mu_{i_j},\mu_{i_j}')$. Then we have $\abs{\mu - \mu_{i_j}} \leq \beta$ and $\abs{\mu - \mu_{i_j}'} \leq \beta$.
Hence we have $\tau \geq 1 - \frac{\beta}{C}$ and 
\begin{align*}
    v^*(\mu)\leq v(s_{i_j}) +\frac{\beta}{C}\,.
\end{align*}
Since $\beta < \frac{C}{4}$, we could find a $\tau'\in [0,1]$ s.t. $\mu = (1-\tau')\mu_{i_{j-1}}'+\tau' \mu_{i_j}'$. Similarly, we have $\tau' \geq 1 - \frac{\beta}{C}$ and thus
\begin{align*}
    V(a^*(p',\mu),p',\mu) \geq (1-\tau')v(s_{i_{j-1}})+\tau' v(s_{i_{j}}) \geq v(s_{i_j}) - \frac{\beta}{C}\,.
\end{align*}
Hence, we have $V(a^*(p',\mu),p',\mu) \geq v^*(\mu) -\frac{2\beta}{C}$.
\end{proof}
\begin{lemma}\label{lmm:tech4}
    There exists a constant $c_2$ such that $V(a^*(p,\mu,\epsilon),p,\mu) \leq v^*(\mu)+ c_2\sqrt{\epsilon}$ for all $p\in \cP$.
\end{lemma}
For any $p\in \cP$, let $\{(\tau_i,\Ber(w_i))\}_{i\in[n]}$ denote the distribution of posteriors induced by policy $p$ and prior $\mu$.
Let $a$ be any $\epsilon$-best response to $(p,\mu)$. Then to prove the lemma, we need to show that there exists a constant $c_2$ such that $V(a,p,\mu) \leq v^*(\mu)+ c_2\sqrt{\epsilon}$ for all $p$.
We introduce lemmas~\ref{lmm:tech1} and \ref{lmm:tech2} to prove Lemma~\ref{lmm:tech4}.

\begin{lemma}\label{lmm:tech1}
        For any $\alpha\in [0, c_1 \cdot C)$, if a strategy $s$ is an $\alpha$-approximate optimal strategy to $\mu$, i.e., $u(s,\mu) =  u(s^*(\mu),\mu)-\alpha$, then there exists $\mu'\in [\mu-\frac{\alpha}{c_1},\mu+\frac{\alpha}{c_1}]$ s.t. $s \in s^*(\mu')$.
    \end{lemma}
    \begin{proof}
    If $\alpha = 0$, then let $\mu' = \mu$. Now we consider the case of $\alpha>0$.
    We first show that if $s_i$ is a best response to $\mu$ and $s_{i+1}$ is not for some $i\in [n]$, then $s_{i+2}$ cannot be an $\alpha$-approximate optimal strategy to $\mu$. This is because $u(s_i,\mu) -u(s_{i+2},\mu)\geq c_1\cdot C$.
    Therefore, if a strategy $s$ is an $\alpha$-approximate optimal strategy to $\mu$, then $s$ can only be $s_{i-1}$ or $s_{i+1}$.    
    W.l.o.g., suppose that $s=s_{i+1}$. 
    Let $\mu' = S_i\cap S_{i+1}$ be the boundary value s.t. both $s_i$ and $s_{i+1}$ are best response to $\mu'$. Then we have $\alpha \geq c_1 \abs{\mu'-\mu}$.         
    \end{proof}
    \begin{lemma}\label{lmm:tech2}
        If an action $a$ is an $\epsilon$-best response to $(p,\mu)$, i.e., $\sum_{i\in[n]} \tau_i u(a(s_i), w_i) \geq \sum_{i\in [n]} \tau_i u(s_i, w_i)- \epsilon$ with $s_i \in s^*(w_i)$, then we can find a set of $\{w_i'|i\in [n]\}$ such that $a(s_i)\in s^*(w_i')$ and $\sum_{i\in [n]}\tau_i \abs{w_i - w_i'}\leq \frac{\epsilon}{c_1}(1+\frac{1}{C})$.
    \end{lemma}
    \begin{proof}
        For each $i\in [n]$, if $u(a(s_i),w_i)\geq u(s_i,w_i) - c_1\cdot C$, then we can find $w_i'$ in the way introduced in Lemma~\ref{lmm:tech1}. Let $A=\{i|u(a(s_i),w_i)\geq u(s_i,w_i) - c_1\cdot C\}$ denote the corresponding subset of $i$'s.
        According to Lemma~\ref{lmm:tech1}, for all $i\in A$, we have
        \begin{align*}
            \abs{w_i - w_i'} \leq \frac{u(s_i,w_i)-u(a(s_i),w_i)}{c_1}\,.
        \end{align*}
        For $i\notin A$, we just arbitrarily pick an $w_i'$ s.t. $a(s_i)$ is an optimal strategy under $w_i'$, i.e, $a(s_i) \in s^*(w_i')$. 
        Then we have $u(s_i,w_i)-u(a(s_i),w_i)> c_1\cdot C$ for all $i\notin A$, and thus
        \begin{equation*}
            \epsilon\geq \sum_{i\notin A} \tau_i \left(u(s_i,w_i)-u(a(s_i),w_i)\right)\geq c_1\cdot C \sum_{i\notin A} \tau_i \,.
        \end{equation*}
        Therefore, we have $\sum_{i\notin A} \tau_i\leq \frac{\epsilon}{c_1 C}$.
        Then we have
        \begin{equation*}
            \sum_{i\in [n]}\tau_i \abs{w_i - w_i'} \leq \frac{1}{c_1}\sum_{i\in A} \tau_i ( u(s_i,w_i)-u(a(s_i),w_i)) + \sum_{i\notin A} \tau_i\leq \frac{\epsilon}{c_1}(1+\frac{1}{C}).
        \end{equation*}
    \end{proof}
    \begin{proof}[Proof of Lemma~\ref{lmm:tech4}]
    Recall that $\{(\tau_i,\Ber(w_i))\}_{i\in[n]}$ is the distribution of posteriors induced by signal scheme $p$ and prior $\mu$ and $a$ is an $\epsilon$-best response to $(p,\mu)$. Now we want to construct another signal scheme $\varphi$ such that $V(a, p, \mu) \leq V(a^*(\varphi, \mu),\varphi,\mu) +c_2 \sqrt{\epsilon}$. 
    Since $v^*(\mu)\geq V(a^*(\varphi, \mu),\varphi,\mu)$ due to that $v^*(\mu)$ is the optimal achievable value when the Agent best respond, we prove Lemma~\ref{lmm:tech4}.

    Our goal is to apply the construction in Lemma~\ref{lmm:tech2}, and construct a distribution of posteriors with support $\{w_i'|i\in [n]\}$. 
    Since $\sum_{i\in [n]} \tau_i w_i' \neq \mu$, we need to find an alternative set of weights $\tau_i'$'s such that $\sum_{i\in [n]} \tau_i' w_i' = \mu$. 
    According to the construction in Lemma~\ref{lmm:tech1}, for those $i$ with $w_i'\neq w_i$, $w'_i$ must lie in $[C,1-C]$ since $w_i'$ always lie on the boundary of two intervals. Let $B = \{i|w_i'\neq w_i\}$.
    Let $q = \sum_{i\in [n]}\tau_i (w_i'-w_i)$. We have
\begin{equation*}
    \sum_{i\in B} \tau_i w_i' = \mu' + q \,,
\end{equation*}
with $\mu'=\mu-\sum_{i\notin B} \tau_i w_i$. According to Lemma~\ref{lmm:tech2}, we have $q\leq \frac{\epsilon}{c_1}(1+\frac{1}{C})$.  Let $\tau_B = \sum_{i\in B}\tau_i$ denote the probability mass of $i\in B$. Then there are three cases.

\begin{itemize}
    \item $\mu'< \frac{\epsilon}{c_1C}(1+\frac{1}{C})$. In this case, we move all probability mass of $\tau_B$ to $w_{n+1}' = \frac{\mu'}{\tau_B}$, which must lie in $[0,1]$ as $\mu' = \sum_{t\in B} \tau_i w_i$. That is to say, let $\tau_i'=0$ for all $i\in B$, $\tau_i'=\tau_i$ for all $i\notin B$ and $\tau_{n+1}'=\tau_B$ for $w_{n+1}' = \mu'$. 
    Then we have $\sum_{i=1}^{n+1}\tau_i' w_i' = \mu' + \sum_{i\notin B} \tau_i w_i = \mu$.
    Thus, $\{(\tau_i',w_i')|i=1,\ldots,n+1\}$ is a Bayesian-plausible distribution of posteriors with $s^*(w_i') = a(s_i)$ for all $i\in [n]$.
    \item $q>0$. we let $\tau_i' = \frac{\mu'\tau_i}{\mu'+q}$ for $i\in B$, $\tau_i' = \tau_i$ for $i\notin B$, and the remaining probability mass $\tau_{n+1}' = 1-\sum_{i\in [n]}\tau_i'$ on $w_{n+1}'=0$. Then we have  $\sum_{i=0}^n \tau_i' w_i' = \mu$ and thus, $\{(\tau_i',w_i')|i=1,\ldots,n+1\}$ is a Bayesian-plausible distribution of posteriors with $s^*(w_i') = a(s_i)$ for all $i\in [n]$.
    \item $q<0$ and $\mu'\geq \frac{\epsilon}{c_1C}(1+\frac{1}{C})$.
Then let $\tau_i' = \frac{(\tau_B-\mu') \tau_i}{\tau_B-\mu'-q}$ for $i\in B$, $\tau_i' = \tau_i$ for $i\notin B$ and the remaining probability mass $\tau_{n+1}' = 1-\sum_{i}\tau_i'$ on $w_{n+1}'=1$.
Note that $\tau_B\geq \frac{1}{1-C}(\mu'+q) \geq \mu'$ where the first inequality holds due to $w_i'\leq 1-C$ for all $i\in B$ and the second inequality holds due to $\mu'\geq \frac{\epsilon}{c_1C}(1+\frac{1}{C})\geq \frac{|q|}{C}$. Thus we have $\tau_i'\geq 0$ and $\tau_i$'s define a legal distribution.
Then we have
\begin{align*}
    \sum_{i\in [n+1]}\tau_i' w_i' &= \frac{(\tau_B-\mu')}{\tau_B-\mu'-q}\sum_{i\in B}\tau_i w_i' + \sum_{i\notin B}\tau_i w_i + (1-\frac{(\tau_B-\mu')}{\tau_B-\mu'-q})\tau_B\\
    & = \frac{(\tau_B-\mu')}{\tau_B-\mu'-q}(\mu'+q) + \sum_{i\notin B}\tau_i w_i + (1-\frac{(\tau_B-\mu')}{\tau_B-\mu'-q})\tau_B\\
    & = \mu'+ \sum_{i\notin B}\tau_i w_i =\mu\,.
\end{align*}
Hence, $\{(\tau_i',w_i')|i=1,\ldots,n+1\}$ is a Bayesian-plausible distribution of posteriors with $s^*(w_i') = a(s_i)$ for all $i\in [n]$. 
\end{itemize}
Since $w_i'\in [C,1-C]$ for all $i\in B$, we have $\mu'+q \in [\tau_B C, \tau_B (1-C)]$.
Then in the first case, we have
\begin{align*}
     V(a, p, \mu) = &\sum_{i\in [n]}\tau_i v(a(s_i)) =\sum_{i\notin B}\tau_i v(a(s_i)) + \tau_B v(s^*(w_{n+1}'))+ \sum_{i\in B}\tau_i (v(a(s_i)) - v(s^*(w_{n+1}'))) \\
    \leq& \sum_{i\in [n+1]} \tau_i' v(s^*(w_i')) + \tau_B \leq V(a^*(\varphi, \mu),\varphi,\mu) +\frac{2\epsilon}{c_1C^2}(1+\frac{1}{C})\,,
\end{align*}
where the last inequality holds due to $\tau_B \leq \frac{\mu'+q}{C}$.

Since $\mu'+q \in [\tau_B C, \tau_B (1-C)]$, in both of the second case and the third case, we have $\tau_i \leq (1+\frac{\abs{q}}{C\tau_B -\abs{q}})\tau_i'$ for all $i\in B$.
Then we have
\begin{align*}
    V(a, p, \mu) = &\sum_{i\in [n]}\tau_i v(a(s_i)) \leq \sum_{i\in B}(1+\frac{\abs{q}}{C\tau_B -\abs{q}})\tau_i' v(a(s_i)) + \sum_{i\notin B}\tau_i' v(a(s_i))\\
    \leq& \sum_{i\in [n]} \tau_i' v(s^*(w_i')) + \frac{\abs{q}}{C\tau_B-\abs{q}} = V(a^*(\varphi, \mu),\varphi,\mu) +\frac{\abs{q}}{C\tau_B -\abs{q}}\,,
\end{align*}
and
\begin{align*}
    V(a, p, \mu) \leq \tau_B + \sum_{i\notin B} \tau_i v(a(s_i))= \tau_B + \sum_{i\notin B} \tau_i' v(s^*(w_i')) \leq V(a^*(\varphi, \mu),\varphi,\mu) +\tau_B\,.
\end{align*}
Since $\min(\tau_B, \frac{\abs{q}}{C\tau_B -\abs{q}}) \leq \sqrt{\frac{\abs{q}}{C}} + \frac{\abs{q}}{C}$, 
by combining these two inequalities together, we have
\begin{align*}
    V(a, p, \mu)\leq V(a^*(\varphi, \mu),\varphi,\mu) + \sqrt{\frac{\abs{q}}{C}} + \frac{\abs{q}}{C} \leq V(a^*(\varphi, \mu),\varphi,\mu) +2\sqrt{\frac{\epsilon}{c_1 C} (1+\frac{1}{C})} \,.
\end{align*}
when $\epsilon$ is small.

    \end{proof}

\end{proof}

\begin{proof}[Proof of Lemma~\ref{lmm:bayes-alt}]
    According to our definition of $p'(\mu)$, it induces a convex combination of two points in $\text{Ex}'$, saying $\mu = \tau \cdot \mu'_{i_{k}} + (1-\tau) \cdot \mu'_{i_{l}}$. 
Recall that all $\mu$ values associated with points in $\text{Ex}$ must be on the boundary between two intervals. 
Furthermore, by construction, any point in $\text{Ex}'$ have $\mu'$ values which are exactly $\beta$ different from some $\mu$ in $\text{Ex}$.
Hence $\mu'_{i_{k}}$ and $\mu'_{i_{l}}$ will be at least $\beta$-far from the edge of any interval. 
Therefore, 
Lemma~\ref{lmm:bayes-stable} implies that $p'(\mu)$ is a $(x \cdot c_{1}\beta,x)$-stable policy under $\mu$.
By combining with Lemma~\ref{lmm:stabilized}, we prove Lemma~\ref{lmm:bayes-alt}.
\end{proof}

\subsubsection{Proof of Theorem~\ref{thm:bayes-discrete}}
\thmbayesdiscrete*
\begin{proof}[Proof of Theorem~\ref{thm:bayes-discrete}]
    Recalling our definition of $p'(\mu)$, it induces a convex combination of two points in $\text{Ex}'$, saying $\mu = \tau \cdot \mu'_{i_{k}} + (1-\tau) \cdot \mu'_{i_{l}}$. Then the explicit form of $p'(\mu)$ is
    \begin{align*}
         p(s_{i_{k}}|y=1) = \frac{\tau \cdot \mu_{k}'}{\mu}\,, \quad p(s_{i_{k}}|y=0) = \frac{\tau \cdot (1-\mu_{k}')}{1-\mu}
    \end{align*}
    By rounding these two probabilities, we obtain a discretized signal scheme $p_{\delta}(\mu)$ with
    \begin{align*}
         p_\delta(s_{i_{k}}|y=1) = \delta \cdot \argmin_{k\in \{0,\ldots,1/\delta\}} \abs{k\delta -p(s_{i_{k}}|y=1) }\,, \quad p_\delta(s_{i_{k}}|y=0) =\delta \cdot \argmin_{k\in\{0,\ldots,1/\delta\}} \abs{k\delta -p(s_{i_{k}}|y=0) }\,.
    \end{align*}
    Let $\delta_1 = p_\delta(s_{i_{k}}|y=1) - p'(s_{i_{k}}|y=1) $ and $\delta_0 = p_\delta(s_{i_{k}}|y=0) - p'(s_{i_{k}}|y=0)$ denote the discretization errors with $\abs{\delta_0},\abs{\delta_1}<\delta$. We have the new distribution of posteriors $(\tau_\delta, \mu_{\delta,i_k}), (1-\tau_\delta, \mu_{\delta,i_l})$
    with
    \begin{align*}
     \tau_\delta &=p_\delta(s_{i_k}|y=1)\mu +p_\delta(s_{i_k}|y=0)(1-\mu) =  (\frac{\tau \cdot \mu_{i_{k}}'}{\mu} + \delta_1)\mu + (\frac{\tau \cdot (1-\mu_{i_{k}}')}{1-\mu} +\delta_0)(1-\mu) \\
     &=\tau + \delta_1 \mu + \delta_0 (1-\mu)\,,\\
        \mu_{\delta,i_k} &= \pi(y|s_{i,k}) = \frac{p_\delta(s_{i_k}|y=1)\mu}{\tau_\delta} = \frac{(\frac{\tau \cdot \mu_{i_{k}}'}{\mu} + \delta_1)\mu}{\tau + \delta_1 \mu + \delta_0 (1-\mu)}=\mu_{i_{k}}' + \frac{\delta_1\mu(1-\mu_{i_{k}}') - \delta_0(1-\mu)\mu_{i_{k}}'}{\tau + \delta_1 \mu + \delta_0 (1-\mu)}\,,\\
       \mu_{\delta,i_l} &= \frac{\mu - \tau_\delta \mu_{\delta,i_k}}{1-\tau_\delta} \,.
    \end{align*}
    Thus, we have $\abs{\tau_\delta - \tau} \leq \delta$ and $\abs{\mu_{\delta,i_k}-\mu_{i_{k}}'} \leq \frac{\delta}{|\tau -\delta|}$. Due to the symmetry, we have $\abs{\mu_{\delta,i_l}-\mu_{i_{l}}'} \leq \frac{\delta}{|(1-\tau) -\delta|}$.
    Then we consider two cases based on the value of $\tau$.
    \begin{itemize}
        \item $\tau <\sqrt{\delta}$ or $\tau > 1-\sqrt{\delta}$. W.l.o.g., we assume that $\tau > 1-\sqrt{\delta}$. Then we can show that $\abs{\mu_{\delta,i_k}-\mu_{i_{k}}'} \leq 2\delta$. Then $p_\delta$ is $((1-\sqrt{\delta})c_1(\beta-2\delta),\sqrt{\delta})$-stable. Since $\mu_{\delta,i_k}$ is at least $\beta-2\delta$ way from the edge and if the Agent chooses the strategy $a(s^*(\mu_{\delta,i_k}))$ is not $s^*(\mu_{\delta,i_k})$ itself given  the signal $s^*(\mu_{\delta,i_k})$, then 
        \begin{align*}
            U(a,p_\delta,\mu) \leq U(a^* (p_\delta,\mu),p_\delta,\mu) - \tau c_1(\beta-2\delta)\leq U(a^*(p_\delta,\mu),p_\delta,\mu) - (1-\sqrt{\delta})c_1(\beta-2\delta)\,.
        \end{align*}
        If the Agent follows the signal $a(s^*(\mu_{\delta,i_k})) = s^*(\mu_{\delta,i_k})$, then 
        \begin{align*}
            V(a,p_\delta,\mu) \geq V(a^*(p_\delta,\mu),p_\delta,\mu) - (1-\tau)\geq V(a^*,p_\delta,\mu) - \sqrt{\delta}\,.
        \end{align*}
        And also, since $\abs{\mu_{\delta,i_k}-\mu_{i_{k}}'} \leq 2\delta$, we have $s^*(\mu_{\delta,i_k}) = s^*(\mu_{i_{k}}')$. Thus, we have
        \begin{align*}
            &V(a^*(p_\delta,\mu),p_\delta,\mu)\\
            = &\tau_\delta v(s^*(\mu_{\delta,i_k})) +(1-\tau_\delta) v(s^*(\mu_{\delta,i_l})) \\
            =& \tau_\delta v(s^*(\mu_{i_{k}}')) +(1-\tau_\delta) v(s^*(\mu_{\delta,i_l}))\\
            \geq& \tau_\delta v(s^*(\mu_{i_{k}}'))\\
            \geq& (\tau-\delta) v(s^*(\mu_{i_{k}}')) \\
            \geq&  V(a^*(p'(\mu),\mu),p'(\mu),\mu) - (1-\tau) -\delta\\ 
            \geq& V(a^*(p'(\mu),\mu),p'(\mu),\mu) -2\sqrt{\delta}
        \end{align*}
        \item $\tau \in [\sqrt{\delta}, 1-\sqrt{\delta}]$. Then both $\abs{\mu_{\delta,i_k}-\mu_{i_{k}}'} \leq 2\sqrt{\delta}$ and $\abs{\mu_{\delta,i_l}-\mu_{i_{l}}'} \leq 2\sqrt{\delta}$. Let $\sqrt{\delta} < \frac{\beta}{4}$. 
        Then by Lemma~\ref{lmm:bayes-stable}, we have that $p_\delta$ is $(x\cdot c_1 \beta/2, x)$-stable for any $x\in [0,1]$.
        Since $s^*(\mu_{\delta,i_k}) = s^*(\mu_{i_{k}}')$ and $s^*(\mu_{\delta,i_l}) = s^*(\mu_{i_{l}}')$, we have
        \begin{align*}
            V(a^*,p_\delta,\mu)=V(a^*,p'(\mu),\mu).
        \end{align*}  
    \end{itemize}
    Hence, $p_\delta(\mu)$ is a $(\frac{3\beta}{C} + c_2 \sqrt{\epsilon} +2\sqrt{\delta},\epsilon, x \cdot c_{1}\beta/2, \max(x,\sqrt{\delta}))$-optimal stable policy under $\mu$ for any $x\in [0,1]$.
\end{proof}

\subsubsection{Proof of Lemma~\ref{lmm:implication-gap-asp}}
\implicationgap*
\begin{proof}
    This is because for each strategy $s\in \cS$, there exists $\mu_s\in [0,1]$ and $c_S>0$ such that $u(s,\mu_s) \geq \mu(s',\mu_s) +C_s$ for all $s'\neq s$ in $\cS$. Hence for all $\mu \in (\mu_s - \frac{C_s}{2}, \mu_s + \frac{C_s}{2})$, we have $u(s,\mu) \geq u(s,\mu_s)-\frac{C_s}{2} 
\geq \mu(s',\mu_s) +\frac{C_s}{2} \geq u(s',\mu)$, where the first and the last inequalities follow from the fact that  $u(s,\cdot)$ is a linear function with $\abs{\partial u(s,\cdot)} \leq 1$.
Let $C =\min_{s\in \cS} C_s$ denote the minimal width of the intervals in $\{S_1,\ldots, S_{n}\}$ observe that since each $C_s > 0$, $C>0$. 

Note that Assumption~\ref{asp-bayes:interval} also implies that, for any two different strategies $s,s'$, the slopes of $u(s,\cdot)$ and $u(s',\cdot)$, denoted by $\partial u(s,\cdot)$ and $\partial u(s',\cdot)$, are different. Otherwise, one of the strategies is dominated by the other one and cannot be strictly optimal at any prior $\mu$, which conflicts with Assumption~\ref{asp-bayes:interval}. 
\end{proof}
\section{Details and Proofs from Section~\ref{sec:general}}\label{app:general}
\subsection*{Details of Assumptions and Algorithms}
In Section \ref{sec:stable} we solved the special case in which we have a stable policy oracle available to us, and in Section \ref{sec:oracles} we showed how to construct stable policy oracles for two important settings: linear contracting, and binary state Bayesian persuasion. 
In this section, we consider the general case, in which we cannot assume the existence of an optimal stable policy oracle. 
In Section~\ref{sec:imposs} we give an example of a setting in which there is no optimal stable policy  (see Proposition~\ref{lem:impossstable}) --- and so indeed, if we want to handle the general case, we need do without such oracles. 
In this case, in addition to the behavioral assumptions in Section~\ref{sec:behavior}, we propose an additional alignment assumption, following~\cite{camara2020mechanisms}.
To build intuition for the Alignment assumption, recall that the Principal provides recommendations $r_t$ to the Agent which are the Agent's best response \emph{under the prior corresponding to the Principal's forecast}. We can view the Principal's recommendation as a reflection of what she expects the Agent to do.  The Agent is under no obligation to follow these recommendations however, and will instead play some action $a_t$. In hindsight, we can consider the optimal policy for the Agent mapping the Principal's chosen policies and recommendations to actions for the Agent. We can view this as the benchmark that the Principal expects the Agent to do well with respect to. Alternately, we could consider a richer set of ``swap'' policies that map the Principal's chosen policies and recommendations \emph{and} the Agent's chosen actions to new actions.  The Agent will do well according to this set of swap benchmark policies because of our low swap regret assumption.  This counterfactual ``swap'' set of policies is only richer than the Principal's expectation for the Agent (as it takes as input more information), and so leads to utility for the Agent that is only greater: We call this difference the ``Gap''. The Alignment assumption says that the difference in Principal utility when the Agent plays actions $a_1,\ldots,a_T$ rather than recommendations $r_1,\ldots,r_T$ is upper bounded as a function of the Gap. Or in other words, the only reason that the Principal's utility can substantially suffer given what the Agent plays, compared to what the Principal's expectation was, is if the Gap was large. Said another way, the Principal's utility may well suffer compared to her expectation because the Agent deviates in ways that are beneficial to himself --- but the Agent will not ``frivolously'' deviate in ways that are harmful to the Principal without being helpful to the Agent. In this sense we can view the Alignment assumption as a moral analogue of the traditional assumption that the Agent breaks ties in favor of the Principal.

There is a subtle distinction between our assumption and the one employed in \cite{camara2020mechanisms}: they apply this alignment assumption to the utilities of the stage game for any prior $\pi$ and any $\epsilon$-best response action, whereas we make a similar assumption concerning the sequence of states $y_{1:T}$ for a specific learning algorithm $\cL$ employed by the Agent. Thus it can be that our alignment is satisfied even if the alignment assumption in \cite{camara2020mechanisms} is not.

\begin{assumption}[Alignment]\label{asp:alignment}
For mechanism $\sigma$, let $p^\sigma_{1:T}$ and $r^\sigma_{1:T}$ denote the sequences of realized policies and recommendations and let  $a^\sigma_{1:T}$ denote a realized sequence of actions selected by the Agent's learning algorithm $\cL$. We define the gap of the Agent's utilities to be the difference between the optimal achievable utility when the Agent can adopt any modification rule taking (policy, recommendation, action) as input and the optimal achievable utility when the Agent can adopt any modification rule taking (policy, recommendation) as input. More formally, $\ugap(y_{1:T},p^\sigma_{1:T},r^\sigma_{1:T},a^\sigma_{1:T})$ is defined as
\[\frac{1}{T}\max_{h:\cP_0\times\cA\times \cA\mapsto \cA} \min_{h':\cP_0\times\cA \mapsto \cA}\sum_{t=1}^T (U(h(p^\sigma_t,r^\sigma_t, a^\sigma_t),p^\sigma_t, y_t) -U(h'(p^\sigma_t,r^\sigma_t),p^\sigma_t, y_t))\,.\]
Then we assume that the sequence of states of nature $y_{1:T}$ satisfies that there exists an $M_1=\cO(1)$ and $M_2 =o(1)$ for which, under the proposed mechanism,
\begin{align*}
    &\frac{1}{T}\sum_{t=1}^T (V(r_t,p_t, y_t) -V(a_t,p_t, y_t)) \leq M_1\cdot \ugap(y_{1:T},p_{1:T},r_{1:T},a_{1:T}) + M_2\,, 
\end{align*}
and under any constant mechanism $\sigma^{p_0}$,
\begin{align*}
   \frac{1}{T}\sum_{t=1}^T (V(a_t^{p_0},p_0,y_t)-V(r^{p_0}_t,p_0,y_t))\leq M_1\cdot \ugap(y_{1:T},(p_0,\ldots,p_0),r^{p_0}_{1:T},a^{p_0}_{1:T}) + M_2\,.
    \end{align*}
\end{assumption}


Again, as discussed in Section~\ref{sec:behavior}, behavioral assumptions are still necessary. We maintain the no contextual swap regret assumption and a less restrictive version of the no secret information assumption. 
    

We consider a weaker ``no secret information'' assumption than Assumption~\ref{asp:no-correlation} that corresponds to assuming that the Agent's ``cross-swap-regret'' with respect to the Principal's communications (policy and recommendation) is not too negative. Intuitively, cross swap regret compares the Agent's utility to a benchmark that lets the Agent choose an action using an arbitrary mapping from the Principal's policies and recommendations to actions. Having very negative cross swap regret means that the Agent is performing substantially better than is possible using the information contained in the Principal's communications. We assume that this is not the case.  
\begin{assumption}[No Negative Cross-Swap-Regret]\label{asp:no-sec-info}
Fix any realized sequence of states $y_{1:T}$.
The Agent's corresponding negative \emph{cross-swap-regret} given the sequence of policy-recommendation pairs  $(p_{1:T}^\sigma,r^\sigma_{1:T})$ is defined to be: 
$$\nr(y_{1:T},p_{1:T}^\sigma,r^\sigma_{1:T}):=\frac{1}{T}\EEs{a_{1:T}^\sigma}{\sum_{t=1}^T U(a_t^\sigma,p_t^\sigma,y_t)- \max_{h:\cP_0 \times \cA\mapsto \cA} \sum_{t=1}^T U(h(p_t^\sigma, r_t^\sigma),p_t^\sigma,y_t)}\,.$$
We assume that the Agent's negative cross swap regret is bounded by $\eneg$ for both the realized sequence of policies and recommendations generated by the Principal's mechanism, as well as counterfactually for any constant mechanism:

    \begin{equation*}
        \nr(y_{1:T},p_{1:T},r_{1:T})\leq\eneg\,,
    \end{equation*}
    and for all $p_0\in \cP_0$,
    \begin{equation*}
        \nr(y_{1:T},(p_0,\ldots,p_0),r^{p_0}_{1:T}) \leq\eneg\,.
    \end{equation*}
\end{assumption}

The no negative-cross-swap-regret assumption can be viewed as a ``no-secret-information'' assumption. But it seems to have a different character than the no-secret-information assumption we made in previous sections (Assumption~\ref{asp:no-correlation}). Recall that Assumption~\ref{asp:no-correlation} informally asked that the Agent's actions should appear to be statistically independent of the state of nature, conditional on the policy and recommendation offered by the Principal. We note, however, that Assumption \ref{asp:no-sec-info} is strictly weaker than Assumption~\ref{asp:no-correlation}:

\begin{restatable}{lemma}{lmmweakerasp}\label{lmm:weakerasp}
    Assumption~\ref{asp:no-sec-info} is weaker than Assumption~\ref{asp:no-correlation}. More specifically, Assumption~\ref{asp:no-correlation} implies Assumption~\ref{asp:no-sec-info} with $\eneg= \cO(\sqrt{\abs{\cP'}\abs{\cA}/T})$, where $\cP'$ is the set of all possible output policies by the proposed mechanism.
\end{restatable}

\begin{remark}
We also note a more intuitive and direct way to model the idea of ``no secret information'': to assume that the Agent cannot consistently outperform the Principal's recommendation, i.e., 
    \begin{equation}
        \frac{1}{T}\sum_{t=1}^T U(a_t,p_t,y_t) - \frac{1}{T}\sum_{t=1}^T U(r_t,p_t,y_t) \leq\eneg\,.\label{eq:no-better-than-guess}
    \end{equation}
    This is also a stronger assumption than Assumption~\ref{asp:no-sec-info}. 
    If the Agent can't consistently outperform the Principal's  recommendation (Eq~\eqref{eq:no-better-than-guess}), then Assumption~\ref{asp:no-sec-info} holds.
\end{remark}

Under this new set of assumptions, the Principal only needs to select the policy that would be optimal each round in the common prior setting, treating the forecast $\pi_t$ as the common prior (Algorithm~\ref{alg:general}). 

\begin{algorithm}[H]
    \caption{Principal's choice at round $t$}
    \label{alg:general}
        \begin{algorithmic}[1]
            \STATE {\textbf{Input}: Forecast $\pi_t\in \Delta(\cY)$}
            \STATE {Select policy $p_t = p^*(\pi_t)\in \argmax_{p\in \cP_0} \EEs{y\sim \pi_t}{V(\brr(p,\pi_t),p,y)}$
            }
        \end{algorithmic}
    \end{algorithm}

\thmgeneral*
\begin{proof}
The proof of Theorem \ref{thm:general} decomposes into two lemmas. The first lemma bounds the loss of the Principal when the Agent behaves in a very simple manner: he simply follows the recommendation of the Principal at every round. In this case, we can bound the regret of the Principal  by the conditional bias of the Principal's predictions:

\begin{restatable}[Regret is Low if Agent Follows Recommendations]
    {lemma}{lmmfollowrecommendation}\label{lmm:regret-follow-recommendation}Recall the definition of events
$$\cE_3 = \{\ind{a^*(p_0,\pi_t) = a}\}_{p_0 \in \cP_0,a\in \cA}\,,\ \ \  \ \  \cE_4 = \{p^*(\pi_t)= p, a^*(p,\pi_t) =a\}_{p\in \cP_0, a\in \cA}\,.$$
Let $\cE' = \cE_3\cup\cE_4$, the union of these events. If the Principal runs the forecasting algorithm from \cite{NRRX23} for events $\cE'$ and the choice rule in Algorithm \ref{alg:general}, and the Agent follows the Principal's recommendations, then we have:

    \begin{align*}
        \EEs{\pi_{1:T}}{\max_{p_0\in \cP} \frac{1}{T}\sum_{t=1}^T \left(V(r_t^{p_0}, {p_0}, y_t)-V(r_t, p_t, y_t)\right)} \leq  \tilde \cO\left(\abs{\cY}\sqrt{\frac{\abs{\cP_0}\abs{\cA}}{T}}\right) \,,
    \end{align*}
    where $r_t^{p_0} = \brr({p_0},\pip_t)$ and $r_t = \brr(p_t,\pip_t)$ are recommendations under constant mechanism $\sigma^{p_0}$ and the proposed mechanism respectively.
\end{restatable}

The next lemma compares the Principal's cumulative utility under the Agent's actual behavior, compared to the utility he would have obtained had the Agent simply followed the Principal's recommendations. It states that under our behavioral assumptions on the Agent, these two quantities are similar, for both the mechanism run by the Principal and for any constant benchmark mechanism. Specifically, the utility obtained by the Principal under the run mechanism cannot be much smaller than the utility she would have obtained had the Agent followed her recommendations --- and for the constant benchmark mechanisms, the utility obtained by the Principal cannot be much \emph{larger} than the utility she would have obtained had the Agent followed her recommendations. Here ``much smaller'' and ``much larger'' are controlled by the parameters $\eint$ and $\eneg$ in the behavioral assumptions.

\begin{restatable}[Principal's Utility is Close to Agent Following Recommendations]{lemma}{lmmutilitydiff}\label{lmm:utility-difference}
    For any sequence of states of nature $y_{1:T}$ and sequnece of forecast $\pi_{1:T}$, under Assumptions \ref{asp:no-internal-reg}, \ref{asp:alignment} and \ref{asp:no-sec-info}, we have
     \begin{align*}
        \EEs{a_{1:T}}{\frac{1}{T}\sum_{t=1}^T  V(a_t,p_t, y_t)}\geq \frac{1}{T}\sum_{t=1}^T V(r_t,p_t, y_t) - M_1(\eint + \eneg) - M_2
    \end{align*}
     and for all $p_0\in \cP_0$,
    \begin{align*}
    \EEs{a^{p_0}_{1:T}}{\frac{1}{T}\sum_{t=1}^T V(a_t^{p_0},p_0,y_t)}\leq \frac{1}{T}\sum_{t=1}^T V(r^{p_0}_t,p_0,y_t) + M_1(\eint + \eneg) + M_2\,.
    \end{align*}
    
\end{restatable}
Together, these two lemmas combine to give the Theorem. 
\end{proof}

\subsection*{Proof of Lemma~\ref{lmm:weakerasp}}
\lmmweakerasp*
\begin{proof}
    When Assumption~\ref{asp:no-correlation} holds, we have
    $$\frac{1}{n_{p,r}}\EEs{a_{1:T}}{\abs{\sum_{t\in (p,r)} U(a_t, p, y_t) -U(\hat \mu_{p,r}, p, y_t)}}\leq \cO\left(\frac{1}{\sqrt{n_{p,r}}}\right)\,,$$
    and
    $$\frac{1}{n^{p_0}_{r}}\EEs{a^{p_0}_{1:T}}{\abs{\sum_{t:r_t=r} U(a_t^{p_0}, p, y_t) - U(\hat \mu^{p_0}_{r}, p, y_t)}}\leq \cO\left(\frac{1}{\sqrt{n^{p_0}_{r}}}\right)\,.$$
    This directly implies the following.
    \begin{align*}
    &\nr(y_{1:T},p_{1:T}^\sigma,r^\sigma_{1:T}) \\
    = & \frac{1}{T}\EEs{a_{1:T}}{\sum_{t=1}^T U(a_t,p_t,y_t) - \max_{h:\cP_0 \times \cA\mapsto \cA} \sum_{t=1}^T U(h(p_t, r_t),p_t,y_t)}\\
       \leq  &\frac{1}{T}\EEs{a_{1:T}}{\sum_{(p,r)\in \cP_\cO\times \cA}\left(\sum_{t\in (p,r)} U(\hat \mu_{p,r}, p, y_t) - \max_{a
       \in \cA} \sum_{t\in (p,r)}U(a, p, y_t)\right)} + \cO(\sqrt{\abs{\cP'}\abs{\cA}/T})\\
        \leq & \cO(\sqrt{\abs{\cP'}\abs{\cA}/T})\,.
    \end{align*}
    Similarly, we have $\nr(y_{1:T},(p_0,\ldots,p_0),r^{p_0}_{1:T})\leq \cO(\sqrt{\abs{\cA}/T})$\,.
\end{proof}
\subsection*{Proof of Lemma~\ref{lmm:regret-follow-recommendation}}
\lmmfollowrecommendation*
\begin{proof}[Proof of Lemma~\ref{lmm:regret-follow-recommendation}]
For proposed mechanism $\sigma^\dagger$, let $t\in (p,r)$ denote $t:(p_t,r_t) = (p,r)$. Let $n_{p,r} = \sum_{t=1}^T\ind{t\in (p,r)}$ denote the number of rounds in which $(p_t,r_t) = (p,r)$. 
Let $\hat y_{p,r} = \frac{1}{n_{p,r}}\sum_{t\in (p,r)} y_t$ denote the empirical distribution of states in these rounds and $\pip_{p,r} = \frac{1}{n_{p,r}}\sum_{t\in (p,r)} \pip_t$ denote the empirical distribution of the forecasts.
For constant mechanism $\sigma^{p_0}$, let $t\in (r)$ denote $t: r^{p_0}_t = r$. 
Let $\cE_{3,p_0} = \{\ind{a^*(p_0,\pi_t) = a}\}_{a\in \cA}$.
Let $\alpha(\cE_{3,p_0}) = \sum_{E\in \cE_{3,p_0}} \alpha(E)$ and $\alpha(\cE_4) = \sum_{E\in \cE_4} \alpha(E)$.
For any $p_0\in \cP_0$, we have
    \begin{align*}
        &\sum_{t=1}^T V(r_t,p_t,y_t) \\
        =& \sum_{(p,r)\in \cP_0\times \cA} \sum_{t\in (p,r)} V(r,p,y_t)\\
        =& \sum_{(p,r)\in \cP_0 \times \cA} n_{p,r} V(r,p,\hat y_{p,r}) \\
        \geq & \sum_{(p,r)\in \cP_0 \times \cA} n_{p,r} V(r,p,\pip_{p,r}) -\alpha(\cE_4)T\tag{$\cE_4$-bias}\\
        = & \sum_{(p,r)\in \cP_0 \times \cA} \sum_{t\in(p,r)} V(\brr(p,\pip_t),p,\pip_t) -\alpha(\cE_4)T 
        \tag{since $r_t= \brr(p,\pip_t)$}\\
        =& \sum_{(p,r)\in \cP_0 \times \cA} \sum_{t\in(p,r)} \max_{p'\in\cP_0}V(\brr(p',\pip_t),p',\pip_t) -\alpha(\cE_4)T \tag{since $p_t = p^*(\pi_t)$}\\
        \geq& \sum_{t=1}^T V(\brr(p_0,\pip_t),p_0,\pip_t)-\alpha(\cE_4)T\\
        =& \sum_{r}\sum_{t\in (r)} V(r,p_0,\pip_t) -\alpha(\cE_4)T\\
        \geq & \sum_{r}\sum_{t \in (r)} V(r,p_0,y_t) - \alpha(\cE_{3,p_0})T -\alpha(\cE_4)T\tag{$\cE_3$-bias}\\
        =& \sum_{t=1}^T V(r^{p_0}_t, p_0,y_t)-\alpha(\cE_{3,p_0})T -\alpha(\cE_4)T\,.
    \end{align*}
    According to Theorem~\ref{thm:forecast-bias}, we have $\alpha(\cE_{3,p_0}) = \tilde \cO(\abs{\cY}\sqrt{\abs{\cA}/T})$ and $\alpha(\cE_{4}) = \tilde \cO(\abs{\cY}\sqrt{\abs{\cP_0}\abs{\cA}/T})$. Then we are done with the proof.
\end{proof}

\subsection*{Proof of Lemma~\ref{lmm:utility-difference}}
\lmmutilitydiff*
\begin{proof}[Proof of Lemma~\ref{lmm:utility-difference}]
For the proposed mechanism $\sigma^\dagger$, let 
    $$\ir^\dagger_{p,r} = \max_{h:\cA\mapsto \cA}\sum_{t\in (p,r)} \left(U(h(a_t),p_t, y_t) - U(a_t,p_t,y_t)\right)$$
    denote the contextual swap regret for the Agent over the subsequence in which $(p_t,r_t) = (p,r)$.
    Similarly, for the fixed mechanism $\sigma^{p_0}$, let 
    $$\ir^{p_0}_{r} = \max_{h:\cA\mapsto \cA}\sum_{t\in (r)} \left(U(h(a^{p_0}_t),p_0, y_t) - U(a^{p_0}_t,{p_0},y_t)\right)$$
    denote the contextual swap regret for the Agent over the subsequence in which $r^{p_0}_t =  r$.
    
    Similarly, let 
    $$\nr^\dagger_{p,r} = \sum_{t\in (p,r)} U(a_t,p_t, y_t) - \max_{a\in \cA} \sum_{t\in (p,r)} U(a,p_t,y_t)$$
    and  
    $$\nr^{p_0}_{r} = \sum_{t\in (r)} U(a^{p_0}_t,p_0, y_t) -\max_{a\in \cA} \sum_{t\in (r)} U(a,{p_0},y_t)$$
    denote the negative cross swap regrets for the Agent over the subsequence in which $(p_t,r_t) = (p,r)$ under the proposed mechanism $\sigma^\dagger$ and the subsequence in which $r^{p_0}_t =  r$ under the constant mechanism $\sigma^{p_0}$ respectively. For proposed mechanism $\sigma^\dagger$, let $t\in (p,r,a)$ denote $t:(p_t,r_t,a_t) = (p,r,a)$.
    For constant mechanism $\sigma^{p_0}$, let $t\in (r,a)$ denote $t: (r^{p_0}_t, a^{p_0}_t) = (r,a)$.
    We have
    \begin{align*}
    &\ugap(y_{1:T},p^\sigma_{1:T},r^\sigma_{1:T},a^\sigma_{1:T})\\
    =&\max_{h:\cP_0\times\cA\times \cA\mapsto \cA} \min_{h':\cP_0\times\cA \mapsto \cA}\sum_{t=1}^T (U(h(p_t,r_t, a_t),p_t, y_t) -U(h'(p_t,r_t),p_t, y_t))\\
    = & \sum_{(p,r)\in \cP_0\times \cA}\left( \max_{h:\cA\mapsto \cA}\sum_{t\in (p,r)}(U(h(a_t),p, y_t) -U(a_t,p, y_t)) + \min_{r'\in \cA} \sum_{t\in (p,r)}(U(a_t,p, y_t) - U(r',p, y_t) )\right)\\
    =&\sum_{(p,r)\in \cP_0\times \cA} \ir^\dagger_{p,r} + \nr^\dagger_{p,r}\,.
    \end{align*}
    Similarly, for constant mechanism $\sigma^{p_0}$, we have
    $$\ugap(y_{1:T},(p_0,\ldots,p_0),r^{p_0}_{1:T},a^{p_0}_{1:T}) =\sum_{r\in \cA}\ir^{p_0}_{r} +\nr^{p_0}_{r}\,.$$
    According to Assumption~\ref{asp:alignment}, we have
    $$\frac{1}{T}\sum_{t=1}^T (V(r_t,p_t, y_t) -V(a_t,p_t, y_t)) \leq M_1\cdot \ugap(y_{1:T},p_{1:T},r_{1:T},a_{1:T}) + M_2\,. $$
    Therefore,
    \begin{align*}
        \EEs{a_{1:T}}{\frac{1}{T}\sum_{t=1}^T  V(a_t,p_t, y_t)}\geq & \frac{1}{T}\sum_{t=1}^T V(r_t,p_t, y_t)- M_1\cdot \EEs{a_{1:T}}{\ugap(y_{1:T},p^\sigma_{1:T},r^\sigma_{1:T},a^\sigma_{1:T})} - M_2\\
        \geq & \frac{1}{T}\sum_{t=1}^T V(r_t,p_t, y_t) - M_1(\eint + \eneg) - M_2\,,
    \end{align*}
    and
    \begin{align*}
    \EEs{a^{p_0}_{1:T}}{\frac{1}{T}\sum_{t=1}^T V(a_t^{p_0},p_0,y_t)}\leq &\frac{1}{T}\sum_{t=1}^T V(r^{p_0}_t,p_0,y_t) + M_1\cdot \EEs{a^{p_0}_{1:T}}{\ugap(y_{1:T},(p_0,\ldots,p_0),r^{p_0}_{1:T},a^{p_0}_{1:T})} + M_2\\
        \leq & \frac{1}{T}\sum_{t=1}^T V(r^{p_0}_t,p_0,y_t) + M_1(\eint + \eneg) + M_2\,.
    \end{align*}
\end{proof}

\section{Proofs from Section~\ref{sec:imposs}}

\nostable*

\begin{proof}
Consider the following contract setting: there are two actions the Agent can take, $a_{1}$ and $a_{2}$. $a_{1}$ gives the Principal a value of $1$, and $a_{2}$ gives her a value of $2$. The cost of $a_{1}$ for the Agent is $\frac{1}{4}$, and the cost of $a_{2}$ is $\frac{1}{2}$. The Principal's contract space has only two linear contracts, $p_{1} = \frac{1}{4}$ and $p_{2} = \frac{1}{2}$. Thus, $p_{1}$ equally incentivizes $a_{1}$ and $a_{2}$, while $p_{2}$ strictly incentivizes $a_{2}$. 

Intuitively, we will show that $p_{1}$ is not stable, as the Agent could tiebreak in favor of $a_{1}$ instead of $a_{2}$ and significantly decrease the Principal's payoff. Furthermore, $p_{2}$ is not optimal, as if the Agent were tiebreaking in favor of $a_{2}$, the Principal would have rather played $p_{1}$. We formalize this below.

Note that the payoffs for the Principal and Agent are independent of the state of nature, and thus of the prior $\pi$. Furthermore, $a^{*}(p_{1},\pi) = a_{2}$, and $a^{*}(p_{1},\pi) = a_{2}$, $\forall \pi$. Let us first assume for contradiction that $p_{1}$ is a $(\beta,\gamma)$-stable optimal policy where $\gamma = o(1)$ and $\beta > 0$. This means that either

$$U(a,p_{1},\pi) \leq U(a^{*}(p_{1},\pi),p_{1},\pi) - \beta$$
or
$$V(a,p_{1},\pi) \geq V(a^{*}(p_{1},\pi),p_{1},\pi) - \gamma$$

For the first condition, we get that

\begin{align*}
& U(a_{1},p_{1},\pi) \leq U(a_{2},p_{1},\pi) - \beta \\
& \Rightarrow p_{1}f(a_{1}) - c(a_{1}) \leq p_{2}f(a_{1}) - c(a_{2}) - \beta \\ 
& \Rightarrow \frac{1}{4} - \frac{1}{4} \leq \frac{1}{2} - \frac{1}{2} - \beta\\ 
& \Rightarrow \beta \leq 0
\end{align*}

This derives a contradiction, so the second condition must be satisfied.

For the second condition, we get that 

\begin{align*}
& V(a_{1},p_{1},\pi) \geq V(a_{2},p_{1},\pi) - \gamma \\
& \Rightarrow (1 - \frac{1}{4})\cdot 1 \geq (1 - \frac{1}{4}) \cdot 2 - \gamma \\
& \Rightarrow \gamma \geq \frac{3}{4} 
\end{align*}

 This also derives a contradiction. Therefore neither condition is satisfied, so $p_{1}$ is not a $(c,\epsilon,\beta,\gamma)$-stable optimal policy for any $\gamma = o(1)$ and $\beta > 0$.

Next, consider $p_{2}$. Let us assume for contradiction that $p_{2}$ is a $(c,\epsilon,\beta,\gamma)$-stable optimal policy where $c = o(1)$ and $\epsilon = 0$. 

Then, 
\begin{align*}
& V(a^*(p_{2},\pi),p_{2},\pi)\geq V(a^*(p_1,\pi,0),p_1,\pi)-c \\
& \Rightarrow V(a_{2},p_{2},\pi)\geq V(a_{2},p_1,\pi)-c \\
& \Rightarrow 1 \geq \frac{3}{2} - c \\
& \Rightarrow c \geq \frac{1}{2} 
\end{align*} 
This derives a contradiction.

As neither $p_{1}$ nor $p_{2}$ are $(c,\epsilon,\beta,\gamma)$-stable optimal policies for $c = o(1)$, $\epsilon \geq 0$, $\beta > 0$ and $\gamma = o(1)$, this completes our proof.
\end{proof}

\necessity*

\strongnecessity*

We will prove these propositions in conjunction. Our proof assumes the existence of and makes use of the learning algorithm $\mathcal{L}^{*}$, and we derive results for both propositions, depending on which guarantees $\mathcal{L}^{*}$ has.

\begin{proof}

Consider a repeated linear contracting problem with two states of nature, $M$ and $H$, and let the realized state sequence be $y_{1:T}$. The Agent's per-round action space is $\mathcal{A} = \{work, shirk\}$. The Principal's per-round policy space is discretized according to $\cP_\delta = \{0, \delta, 2\delta, \ldots, \floor{\frac{1}{\delta}}\delta\}$, the set of all $\delta$-discretized linear contracts. We assume $\delta$ is such that $0.5, 0.6 \in \cP_\delta$. If the state of nature in a given round is $M$, the task will be completed if and only if the Agent plays $work$. If the state of nature is $H$, the task will not be completed regardless. The Principal gets payoff $2$ if the task is completed. It costs the Agent $0$ to shirk and $1$ to work.




For any mechanism $\sigma$, we will construct an algorithm $\mathcal{L}$ for the Agent that gives the Principal high regret. Unlike standard learning algorithms, $\mathcal{L}$ has access to the entire state sequence. Towards defining this algorithm, we will first define two simpler algorithms that will be used as a subroutines which use knowledge of $y_{2:T}$. We will call these algorithms $a^{*}$ and $b^{*}$.  

$a^{*}$ plays $work$ if $y_{t} = M$ and $shirk$ if $y_{t} = H$. 

$b^{*}$ plays $work$ if $y_{t} = M$ and plays $shirk$ w.p. $\frac{4}{5}$ and $work$ w.p. $\frac{1}{5}$ if $y_{t} = H$. 

Furthermore, let us pick an algorithm which always achieves sublinear Contextual Swap Regret for all states of nature sequences against $\sigma$, and call it $noreg$. We know that $noreg$ must exist, by our assumption that some $\cL^{*}$ exists. If there is a learning algorithms in this setting which achieve sublinear negative regret for all sequences against $\sigma$, we will pick such an algorithm. For some $y_{1:T}$, let $m_{y,t}$ be the number of medium states seen in the first $t$ rounds. Let $balanced_{t}=true$ if, on round $t$, $|m_{m,t} - m_{h,t}| \leq \sqrt{12T\ln\left(2\left(1+\log_{2}\left(T\right)\right)^{2}\right)}$. Furthermore, let $balanced_{all}$ be the event that $balanced_{t} = true$ for all $t \leq T$. Intuitively, this condition checks whether the history of nature states is roughly balanced between $M$ and $H$ at each  round.

We are finally ready to define $\mathcal{L}$. This algorithm uses $a^{*}$, $b^{*}$, $noreg$ and $balanced_{t}$ to exploit knowledge about the states of nature fully, but does so deliberately imperfectly so as not to incur negative regret.

In $\mathcal{L}$, if the very first state of nature of $y$ is $M$, then $\mathcal{L}$ plays $a^{*}$ until the Principal ever plays a contract which is not $(0.5,r_{t}=work)$, and then it plays $noreg$ for the rest of the game. If the very first state of nature of $y$ is $H$, then it plays $b^{*}$ until the Principal ever plays a contract which is not $(0.6,r_{t}=work)$, and then it plays $noreg$ for the rest of the game. Furthermore, if the state sequence ever invalidates the balanced condition, the algorithm immediately begins playing $noreg$ for the rest of the game.


\begin{algorithm}
\caption{$\mathcal{L}$}\label{alg:L1}
\begin{algorithmic}
\STATE $t \gets 1$\;
\STATE Play $shirk$ on the first round;
\STATE Observe $y_{1}$\;
\STATE $t \gets 2$
\IF{$y_{1} = M$}
\STATE { $\bar{p} \gets 0.5$, 
$alg \gets a^{*}$\;}
\ELSE
\STATE {$\bar p \gets 0.6$, 
$alg \gets b^{*}$\;}
\ENDIF

\WHILE{$(t \leq T)$, $(p_{t} = \bar p)$, $(r_{t} = work)$ and $balanced_{t}$}
{
\STATE
{Play according to $alg$}\;
{$t \gets t+1$}\;
}
\ENDWHILE
\WHILE{$(t \leq T)$} 
{\STATE 
{Play $noreg$ with the entire history of play in mind}\;
{$t \gets t+1$}\;
}
\ENDWHILE
\end{algorithmic}
\end{algorithm}

The intuition is as follows: if the number of $M$ and $H$ states is approximately equal, the Principal gets a higher payoff when the Agent plays according to $a^{*}$ or $b^{*}$ than when he plays according to $noreg$. But the Agent himself is roughly indifferent between these algorithms. Therefore if the Principal's mechanism causes $noreg$ to be played when $a^{*}$ or $b^{*}$ could have been played, the Principal will have non-vanishing policy regret. Of course, if the number of $M$ and $H$ states is not approximately equal, there is no guarantee on the performance of $a^{*}$ or $b^{*}$. However, if this is ever the case, $\cL$ will switch to playing $noreg$ to ensure that it continues to satisfy the assumptions on its performance.


We prove that $\cL$ ensures the Principal high regret in Lemma~\ref{lem:nonvanishing}. To do this, we introduce a distribution $y^{*}$ which is i.i.d. between $M$ and $H$ in each round. We use the fact that, in expectation over $y^{*}$, the Principal payoff under $noreg$ is $o(T)$, and the Principal payoff when the Agent is playing either $a^{*}$ or $b^{*}$ is $\Omega(T)$ (Lemma~\ref{lem:payoffUBextended}). This implies that there is at least one sequence under which this difference is realized, or in other words, there is a sequence where the Principal has significant regret when $noreg$ is played rather than $a^{*}$ or $b^{*}$. The final piece we need is that such a sequence exists where $balanced_{all}$ is satisfied, in order that the Agent is actually playing $a^{*}$ or $b^{*}$. Because the probability of a sequence from $y^{*}$ not satisfying the balanced condition approaches $0$ with $T$ (Lemma~\ref{lem:iteratedlog}), we can show that such a sequence must exist.

Next, we turn to proving that $\cL$ has vanishing Contextual Swap regret in Lemma~\ref{lem:contextualnegative}. Towards this, use the fact that if $balanced_{all}$ is true and the Principal is playing in a way that causes the Agent to play $a^{*}$ or $b^{*}$, $a^{*}$ and $b^{*}$ have bounded swap regret (Lemma~\ref{lem:boundreg}). We use this with the fact that $\cL$ switches to playing $noreg$ when either $balanced_{all}$ is not true or the Principal misbehaves to show that $\cL$ always has vanishing swap regret. Combining Lemmas~\ref{lem:nonvanishing} and~\ref{lem:contextualnegative} completes the proof of Proposition~\ref{lem:lb-easy}.

Finally, in the case where $noreg$ also has bounded negative regret, we show in Lemma~\ref{lem:negative} that $\cL$ has bounded negative regret as well, completing the proof of Proposition~\ref{lem:lb-hard}.
\end{proof}

\begin{lemma}
For any Principal mechanism, there is a sequence of states of nature such that $\mathcal{L}$ will ensure the Principal non-vanishing policy regret.
\label{lem:nonvanishing}
\end{lemma}

\begin{proof}
Consider any Principal mechanism $\sigma$. On round $t = 1$, before observing any information about the nature states, the mechanism must provide the first policy. There are two cases:
\begin{itemize}
\item The mechanism provides the contract $0.5$ and the recommendation $work$ $w.p. \leq \frac{1}{2}$. Then, we will evaluate the expected regret of $\sigma$ over the distribution of nature states which begin with $y_{1} = M$ and then are distributed according to $y^{*}_{2:T}$. In the first round, with probability at least $\frac{1}{2}$, the Agent immediately begins playing $noreg$. Alternately, if the Principal had played $(0.5,work)$ in the first round (and throughout the entire game), the Agent would have played $a^{*}$. We can compute the regret of the Principal to this alternate policy sequence, in expectation over $y^{*}_{2:T}$. 

\begin{align*}
& \mathbb{E}_{y_{2:T}^{*},\mathcal{L},\sigma}[\sum_{t = 1}^{T}V(a_{t}^{\sigma},(0.5,work),y_{t})] - \mathbb{E}_{y_{2:T}^{*},\mathcal{L},\sigma}[\sum_{t = 1}^{T}V(a_{t}^{\sigma},p_{t}^{\sigma},y_{t})] \\
& = \mathbb{P}(balanced_{all})\mathbb{E}_{y_{2:T}^{*},\mathcal{L}}[\sum_{t = 1}^{T}V(\mathcal{L},(0.5,work),y_{t})|balanced_{all}]\\ & + \mathbb{P}(\neg balanced_{all})\mathbb{E}_{y_{2:T}^{*},\mathcal{L},\sigma}[\sum_{t = 1}^{T}V(a_{t}^{\sigma},(0.5,work),y_{t})|\neg balanced_{all}]  - \mathbb{E}_{y_{2:T}^{*},\mathcal{L},\sigma}[\sum_{t = 1}^{T}V(a_{t}^{\sigma},p_{t}^{\sigma},y_{t})] \\
& \geq \frac{3}{4}\mathbb{E}_{y_{2:T}^{*},\mathcal{L},\sigma}[\sum_{t = 1}^{T}V(a_{t}^{\sigma},(0.5,work),y_{t})|balanced_{all}] - \mathbb{E}_{y_{2:T}^{*},\mathcal{L},\sigma}[\sum_{t = 1}^{T}V(a_{t}^{\sigma},p_{t}^{\sigma},y_{t})] \tag{By Lemma~\ref{lem:iteratedlog}} \\
& \geq \frac{3}{4} \mathbb{E}_{y_{2:T}^{*},a^{*},\sigma}[\sum_{t = 1}^{T}V(a_{t}^{\sigma},(0.5,work),y_{t})|balanced_{all}]\\  & - \mathbb{P}(\sigma_{1}\neq (0.5,work)) \cdot \mathbb{E}_{y_{2:T}^{*},noreg,\sigma}[\sum_{t = 1}^{T}V(noreg^{^{\sigma}},\sigma,p_{t}^{\sigma},y_{t})] \\ & - \mathbb{P}(\sigma_{1} = (0.5,work)) \cdot \mathbb{E}_{y_{2:T}^{*},\mathcal{L},\sigma}[\sum_{t = 1}^{T}V(a_{t}^{\sigma},p_{t}^{\sigma},y_{t})] \\
& \geq \frac{3}{4} \mathbb{E}_{y_{2:T}^{*},a^{*},\sigma}[\sum_{t = 1}^{T}V(a_{t}^{\sigma},(0.5,work),y_{t})] - o(T) \\ & - \mathbb{P}(\sigma_{1}\neq (0.5,work)) \cdot \mathbb{E}_{y_{2:T}^{*},noreg,\sigma}[\sum_{t = 1}^{T}V(noreg^{^{\sigma}},\sigma,p_{t}^{\sigma},y_{t})] \\ & - \mathbb{P}(\sigma_{1} = (0.5,work)) \cdot \mathbb{E}_{y_{2:T}^{*},\mathcal{L},\sigma}[\sum_{t = 1}^{T}V(a_{t}^{\sigma},p_{t}^{\sigma},y_{t})] \tag{By Lemma~\ref{lem:balanced_close}}\\
& = \frac{3}{4}(\frac{T}{2} - \frac{1}{2}\cdot \frac{T}{2}) - \mathbb{P}(\sigma_{1}\neq (0.5,work)) \cdot \mathbb{E}_{y_{2:T}^{*},noreg,\sigma}[\sum_{t = 1}^{T}V(noreg^{^{\sigma}},\sigma,p_{t}^{\sigma},y_{t})] \\ & - \mathbb{P}(\sigma_{1} = (0.5,work)) \cdot \mathbb{E}_{y_{2:T}^{*},\mathcal{L},\sigma}[\sum_{t = 1}^{T}V(a_{t}^{\sigma},p_{t}^{\sigma},y_{t})] \tag{By the definition of $a^{*}$ over $y^{*}$} \\
& = \frac{3}{4}\cdot \frac{T}{4} - \mathbb{P}(\sigma_{1}\neq (0.5,work)) \cdot o(T) - \mathbb{P}(\sigma_{1} = (0.5,work)) \cdot \mathbb{E}_{y_{2:T}^{*},\mathcal{L},\sigma}[\sum_{t = 1}^{T}V(a_{t}^{\sigma},p_{t}^{\sigma},y_{t})] \tag{By Lemma~\ref{lem:PrincipalpayoffUB}} \\
& \geq \frac{3}{4} \cdot \frac{T}{4} - \mathbb{P}(\sigma_{1}\neq (0.5,work)) \cdot o(T) - \mathbb{P}(\sigma_{1} = (0.5,work)) \cdot (\frac{T}{4} + o(T)) \tag{By Lemma~\ref{lem:payoffUBextended}} \\
& \geq \frac{3}{4} \cdot \frac{T}{4} - \frac{1}{2} \cdot o(T) - \frac{1}{2} \cdot (\frac{T}{4} + o(T)) = \frac{3T}{16} - \frac{T}{8} - o(T) \\
& = \Omega(T) \\
\end{align*}

The expected total regret over this distribution of sequences against $\mathcal{L}$ is $\Omega(T)$. Therefore, there must be at least one sequence beginning with $M$ that has regret of $\Omega(T)$.

\item The mechanism provides the contract $0.6$ and the recommendation $work$ w.p. $\leq \frac{1}{2}$. Then, we will evaluate the expected regret of $\sigma$ over the distribution of nature states which begin with $y_{1} = H$ and then are distributed according to $y^{*}_{2:T}$. There is at least a $\frac{1}{2}$ probability that after the first round, the Agent immediately begins playing $noreg$. Alternately, if the Principal had played $(0.6,work)$ in the first round (and throughout the entire game), the Agent would have played $b^{*}$. We can compute the regret of the Principal to this alternate policy sequence, in expectation over $y^{*}_{2:T}$:

\begin{align*}
& \mathbb{E}_{y_{2:T}^{*},\mathcal{L},\sigma}[\sum_{t = 1}^{T}V(a_{t}^{\sigma},(0.6,work),y_{t})] - \mathbb{E}_{y_{2:T}^{*},\mathcal{L},\sigma}[\sum_{t = 1}^{T}V(a_{t}^{\sigma},p_{t}^{\sigma},y_{t})] \\
& = \mathbb{P}(balanced_{all})\mathbb{E}_{y_{2:T}^{*},\mathcal{L}}[\sum_{t = 1}^{T}V(\mathcal{L},(0.6,work),y_{t})|balanced_{all}]\\ & + \mathbb{P}(\neg balanced_{all})\mathbb{E}_{y_{2:T}^{*},\mathcal{L},\sigma}[\sum_{t = 1}^{T}V(a_{t}^{\sigma},(0.6,work),y_{t})|\neg balanced_{all}]  - \mathbb{E}_{y_{2:T}^{*},\mathcal{L},\sigma}[\sum_{t = 1}^{T}V(a_{t}^{\sigma},p_{t}^{\sigma},y_{t})] \\
& \geq \frac{3}{4}\mathbb{E}_{y_{2:T}^{*},\mathcal{L},\sigma}[\sum_{t = 1}^{T}V(a_{t}^{\sigma},(0.6,work),y_{t})|balanced_{all}] - \mathbb{E}_{y_{2:T}^{*},\mathcal{L},\sigma}[\sum_{t = 1}^{T}V(a_{t}^{\sigma},p_{t}^{\sigma},y_{t})] \tag{By Lemma~\ref{lem:iteratedlog}} \\
& \geq \frac{3}{4} \mathbb{E}_{y_{2:T}^{*},b^{*},\sigma}[\sum_{t = 1}^{T}V(a_{t}^{\sigma},(0.6,work),y_{t})|balanced_{all}]\\  & - \mathbb{P}(\sigma_{1}\neq (0.6,work)) \cdot \mathbb{E}_{y_{2:T}^{*},noreg,\sigma}[\sum_{t = 1}^{T}V(noreg^{^{\sigma}},\sigma,p_{t}^{\sigma},y_{t})] \\ & - \mathbb{P}(\sigma_{1} = (0.6,work)) \cdot \mathbb{E}_{y_{2:T}^{*},\mathcal{L},\sigma}[\sum_{t = 1}^{T}V(a_{t}^{\sigma},p_{t}^{\sigma},y_{t})] \\
& \geq \frac{3}{4} \mathbb{E}_{y_{2:T}^{*},b^{*},\sigma}[\sum_{t = 1}^{T}V(a_{t}^{\sigma},(0.6,work),y_{t})] - o(T) \\ & - \mathbb{P}(\sigma_{1}\neq (0.6,work)) \cdot \mathbb{E}_{y_{2:T}^{*},noreg,\sigma}[\sum_{t = 1}^{T}V(noreg^{^{\sigma}},\sigma,p_{t}^{\sigma},y_{t})] \\ & - \mathbb{P}(\sigma_{1} = (0.6,work)) \cdot \mathbb{E}_{y_{2:T}^{*},\mathcal{L},\sigma}[\sum_{t = 1}^{T}V(a_{t}^{\sigma},p_{t}^{\sigma},y_{t})] \tag{By Lemma~\ref{lem:balanced_close}}\\
& = \frac{3}{4}(\frac{T}{5}) - \mathbb{P}(\sigma_{1}\neq (0.6,work)) \cdot \mathbb{E}_{y_{2:T}^{*},noreg,\sigma}[\sum_{t = 1}^{T}V(noreg^{^{\sigma}},\sigma,p_{t}^{\sigma},y_{t})] \\ & - \mathbb{P}(\sigma_{1} = (0.6,work)) \cdot \mathbb{E}_{y_{2:T}^{*},\mathcal{L},\sigma}[\sum_{t = 1}^{T}V(a_{t}^{\sigma},p_{t}^{\sigma},y_{t})] \tag{By the definition of $b^{*}$ over $y^{*}$} \\
& = \frac{3}{4}\cdot \frac{T}{5} - \mathbb{P}(\sigma_{1}\neq (0.6,work)) \cdot o(T) - \mathbb{P}(\sigma_{1} = (0.6,work)) \cdot \mathbb{E}_{y_{2:T}^{*},\mathcal{L},\sigma}[\sum_{t = 1}^{T}V(a_{t}^{\sigma},p_{t}^{\sigma},y_{t})] \tag{By Lemma~\ref{lem:PrincipalpayoffUB}} \\
& \geq \frac{3T}{20} - \mathbb{P}(\sigma_{1}\neq (0.6,work)) \cdot o(T) - \mathbb{P}(\sigma_{1} = (0.6,work)) \cdot (\frac{T}{4} + o(T)) \tag{By Lemma~\ref{lem:payoffUBextended}} \\
& \geq \frac{3T}{20} - \frac{1}{2} \cdot o(T) - \frac{1}{2} \cdot (\frac{T}{4} + o(T)) = \frac{3T}{20} - \frac{T}{8} - o(T) \\
& = \Omega(T) \\
\end{align*}

For an equivalent argument to the first case, there must be at least one sequence beginning with $H$ that has regret of $\Omega(T)$

\end{itemize}
\end{proof}

Next, we show that these learning algorithm which can guarantee the Principal non-vanishing policy regret also satisfies our assumption that the Agent achieves no swap regret.

\begin{lemma}
$\mathcal{L}$ will have vanishing contextual swap regret  \label{lem:contextualnegative}
\end{lemma}

\begin{proof}

We will prove that $\mathcal{L}$ has vanishing contextual swap regret against any mechanism $\sigma$. Because this set of mechanisms includes all constant mechanisms, we now only need to prove this one stronger claim instead of two claims to satisfy assumption~\ref{asp:no-internal-reg}.
Let $t_{b}$ be the first round when the Agent defects to begin playing $noreg$. Then, the contextual swap regret of $\mathcal{L}$ against any sequence $y$ (not necessarily drawn from $y^{*}$) can be expressed as

\begin{align*}
&T \cdot \ir(y_{1:T},p_{1:T},r_{1:T}) = \E_{\cL,\sigma}\left[\max_{h:\cP\times \cA\times \cA\mapsto \cA}\sum_{t=1}^T (U(h(p_t^{\sigma},r_t^{\sigma},a_t^{\sigma}),p_t^{\sigma}, y_t) - U(a_t^{\sigma},p_t^{\sigma},y_t^{\sigma})) \right]\  \\
& = \E_{\cL,\sigma}\left[\max_{h:\cP\times \cA\times \cA\mapsto \cA}\sum_{t=1}^{t_{b}} (U(h(p_t^{\sigma},r_t^{\sigma},a_t^{\sigma}),p_t^{\sigma}, y_t) - U(a_t^{\sigma},p_t^{\sigma},y_t)) \right]\ + \\ & \E_{\cL,\sigma}\left[\max_{h:\cP\times \cA\times \cA\mapsto \cA}\sum_{t=t_{b}+1}^T (U(h(p_t^{\sigma},r_t^{\sigma},a_t^{\sigma}),p_t^{\sigma}, y_t) - U(a_t^{\sigma},p_t^{\sigma},y_t)) \right]\   \\
& = \E_{\cL,\sigma}\left[\max_{h:\cP\times \cA\times \cA\mapsto \cA}\sum_{t=1}^{t_{b}} (U(h(p_t^{\sigma},r_t^{\sigma},a_t^{\sigma}),p_t^{\sigma}, y_t) - U(a_t^{\sigma},p_t^{\sigma},y_t)) \right]\ + \\ & \E_{noreg,\sigma}\left[\max_{h:\cP\times \cA\times \cA\mapsto \cA}\sum_{t=t_{b}+1}^T (U(h(p_t^{\sigma},r_t^{\sigma},a_t^{\sigma}),p_t^{\sigma}, y_t) - U(a_t^{\sigma},p_t^{\sigma},y_t)) \right]\ \tag{By the fact that $\mathcal{L}$ begins playing $noreg$ at $t_{b}+1$}  \\
& \leq o(T) + \E_{noreg,\sigma}\left[\max_{h:\cP\times \cA\times \cA\mapsto \cA}\sum_{t=t_{b}+1}^T (U(h(p_t^{\sigma},r_t^{\sigma},a_t^{\sigma}),p_t^{\sigma}, y_t) - U(a_t^{\sigma},p_t^{\sigma},y_t)) \right]\ \tag{By Lemma~\ref{lem:boundreg}}  \\
& \leq o(T) \tag{By the fact that $noreg$ has bounded contextual swap regret} \\
\end{align*}

Thus, $\ir(y_{1:T},p_{1:T},r_{1:T}) \leq \frac{o(T)}{T} = o(1)$

\end{proof}

\begin{lemma}
As long as $noreg$ has vanishing negative regret, $\mathcal{L}$ will have vanishing negative regret. \label{lem:negative}
\end{lemma}
\begin{proof}

Let $t_{b}$ be the first round in which the Agent begins playing $noreg$.
We can split up the negative regret of the Agent as follows:

\begin{align*}
&T \cdot \nr(y_{1:T},p_{1:T}^\sigma,r^\sigma_{1:T}) =\EEs{\cL,\sigma}{\sum_{t=1}^T U(a_t^\sigma,p_t^\sigma,y_t)- \max_{h:\cP_0 \times \cA\mapsto \cA} (h(p_t^\sigma, r_t^\sigma),p_t^\sigma,y_t)} \\
& =\EEs{\mathcal{L},\sigma}\sum_{t=1}^{t_{b}} U(a_t^\sigma,p_t^\sigma,y_t)- \max_{h:\cP_0 \times \cA\mapsto \cA} {\sum_{t=1}^{t_{b}} U(h(p_t^\sigma, r_t^\sigma),p_t^\sigma,y_t)} \\
& + \EEs{noreg,\sigma}\sum_{t=t_{b}+1}^T U(a_t^\sigma,p_t^\sigma,y_t)- \max_{h:\cP_0 \times \cA\mapsto \cA} {\sum_{t=t_{b+1}}^{T} U(h(p_t^\sigma, r_t^\sigma),p_t^\sigma,y_t)} \\
& \leq \EEs{\mathcal{L},\sigma}\sum_{t=1}^{t_{b}} U(a_t^\sigma,p_t^\sigma,y_t)- \max_{h:\cP_0 \times \cA\mapsto \cA} {\sum_{t=1}^{t_{b}} U(h(p_t^\sigma, r_t^\sigma),p_t^\sigma,y_t)} + o(T) \tag{By the fact that $noreg$ has vanishing negative regret.} \\
& \leq o(T) \tag{
By Lemma~\ref{lem:boundreg}.}
\end{align*}

Thus, $\nr \leq \frac{o(T)}{T} = o(1)$.

\end{proof}

\label{app:imposs}
\begin{lemma}
$\mathbb{E}_{y^{*},\mathcal{L},\sigma}[\sum_{t = 1}^{T}V(a_{t}^{\sigma},(0.5,work),y_{t})] \leq \mathbb{E}_{y^{*},\mathcal{L},\sigma}[\sum_{t = 1}^{T}V(a_{t}^{\sigma},(0.5,work),y_{t})|balanced_{all}] + o(T)$
\label{lem:balanced_close}
\end{lemma}

\begin{proof}

In this proof we use the fact that the distributions $y^{*}|balanced_{all}$ and $y^{*}$ are very close to each other to show that the Principal's expected payoff must be similar under both.

\begin{align*}
&\mathbb{E}_{y^{*},\mathcal{L},\sigma}[\sum_{t = 1}^{T}V(a_{t}^{\sigma},(0.5,work),y_{t})] \\
& =\mathbb{P}(balanced_{all})\mathbb{E}_{y^{*},\mathcal{L},\sigma}[\sum_{t = 1}^{T}V(a_{t}^{\sigma},(0.5,work),y_{t})|balanced_{all}] \\ & + \mathbb{P}(\neg balanced_{all})\mathbb{E}_{y^{*},\mathcal{L},\sigma}[\sum_{t = 1}^{T}V(a_{t}^{\sigma},(0.5,work),y_{t})|\neg balanced_{all}] \\
& \leq \mathbb{P}(balanced_{all})\mathbb{E}_{y^{*},\mathcal{L},\sigma}[\sum_{t = 1}^{T}V(a_{t}^{\sigma},(0.5,work),y_{t})|balanced_{all}] \\ & + T^{-\frac{1}{10}}\mathbb{E}_{y^{*},\mathcal{L},\sigma}[\sum_{t = 1}^{T}V(a_{t}^{\sigma},(0.5,work),y_{t})|\neg balanced_{all}]  \tag{By Lemma~\ref{lem:iteratedlog}}\\ 
& \leq \mathbb{P}(balanced_{all})\mathbb{E}_{y^{*},\mathcal{L},\sigma}[\sum_{t = 1}^{T}V(a_{t}^{\sigma},(0.5,work),y_{t})|balanced_{all}] + T^{-\frac{1}{10}} \cdot T \\ 
& \leq \mathbb{E}_{y^{*},\mathcal{L},\sigma}[\sum_{t = 1}^{T}V(a_{t}^{\sigma},(0.5,work),y_{t})|balanced_{all}] + o(T) \\ 
\end{align*}
\end{proof}

\begin{lemma}
If $y_{2:T} \sim y^{*}_{2:T}$, with probability at least $1 - T^{\frac{1}{10}}$, $balanced_{all}=true$. Furthermore, $balanced_{all}$ implies that the difference between the number of $M$ and $H$ states is $o(T)$. \label{lem:iteratedlog}
\end{lemma}

\begin{proof}
Let us consider $y^{*}_{2:T}$ to be a sequence of independent, identically distributed random variables $S$, where the value is $1$ when the state is $M$ and $-1$ otherwise. Then they have mean $0$ and variance $1$. The absolute value of the difference between the number of $M$ states and the number of $H$ states is now exactly equal to $|S_{T}| = |\sum_{i=1}^{T}y_{i}|$.

This is now a Rademacher random walk. By an application of the nonasymptotic version of the Law of Iterated Logarithm in \cite{balsubramani2015sharp}, we have that with probability $\geq 1 - T^{-\frac{1}{10}}$, for all $t \leq T$  simultaneously,

\begin{align*}
& |S_{t}| \leq \sqrt{3t(2log(log(\frac{5}{2}t))+ log(2T^{\frac{1}{10}))}} \\
& \leq \sqrt{3T(2log(log(\frac{5}{2}T))+ log(2T^{\frac{1}{10}))}} = o(T) \\
\end{align*}

\end{proof}

\begin{lemma}
In expectation over $y^{*}$, the expected payoff of any mechanism $\sigma$ against $noreg$ is at most $o(T)$. \label{lem:PrincipalpayoffUB}
\end{lemma}

\begin{proof}
Let $s_{m,w}$ be the number of rounds in which the state is medium and the Agent works, and define $s_{h,w}$, $s_{m,s}$, and $s_{h,s}$ accordingly. Let us assume for contradiction that the Principal receives expected payoff of at least $c \cdot T$. Then,

\begin{align*}
&\mathbb{E}_{y_{*},noreg,\sigma}[\sum_{t = 1}^{T}V(a_{t}^{\sigma},p_{t}^{\sigma},y_{t})]\geq c\cdot T \\ & \Rightarrow \mathbb{E}_{y^{*},noreg,\sigma}[\sum_{t=1}^{T}((2 - 2p_{t}) \cdot \mathbbm{1}[m,w])] \geq c \cdot T  \\
& \Rightarrow \mathbb{E}_{y^{*},noreg,\sigma}[2s_{m,w}]  - c \cdot T \geq  \mathbb{E}_{y^{*},noreg,\sigma}[\sum_{t=1}^{T} 2p_{t} \cdot \mathbbm{1}[m,w])] \\
\end{align*}

However, by assumption, we also have that 

\begin{align*}
& \mathbb{E}_{y^{*},noreg,\sigma}[\sum_{t=1}^{T}U(a_{t}^{\sigma},p_{t}^{\sigma},y_{t})] \geq -o(T) \tag{By the fact that the Agent could play $shirk$ every round and get $0$} \\
& \Rightarrow \mathbb{E}_{y^{*},noreg,\sigma}[\sum_{t=1}^{T}(2p_{t} \cdot \mathbbm{1}[m,w]) - s_{m,w} - s_{h,w}] \geq -o(T)  \\
& \Rightarrow \mathbb{E}_{y^{*},noreg,\sigma}[2 \cdot s_{m,w} - c\cdot T - s_{m,w} - s_{h,w}] \geq -o(T)\tag{Using the Principal payoff expression} \\
& \Rightarrow \mathbb{E}_{y^{*},noreg,\sigma}[s_{m,w} - s_{h,w}] \geq c \cdot T  -o(T) > 0 \tag{For sufficiently large $T$} \\
\end{align*}

As $\sigma$ does not take the states of nature as input, we know that $y_{t} \sim y^{*}$ is independent of $(p_{t},r_{t})$. Furthermore, as $noreg$ does not take the states of nature as input, we know that $a_{t}$, conditioned on $(p_{t},r_{t})$, is independent of $y_{t} \sim y^{*}$. Putting these together, we get that $a_{t}$ is independent of $y_{t}$. Therefore,

\begin{align*}
& \mathbb{E}_{y^{*},noreg,\sigma}[s_{m,w} - s_{h,w}] \\
& = \sum_{t=1}^{T}\mathbb{P}_{y^{*},noreg,\sigma}(y_{t} = M, a_{t} = work) - \sum_{t=1}^{T}\mathbb{P}_{y^{*},noreg,\sigma}(y_{t} = H, a_{t} = work) \\
& = \sum_{t=1}^{T}\mathbb{P}_{y^{*}_{1:t-1},noreg,\sigma}(a_{t} = work) \cdot \mathbb{P}_{y^{*}_{t}}y_{t} = M) - \sum_{t=1}^{T}\mathbb{P}_{y^{*}_{1:t-1},noreg,\sigma}(a_{t} = work) \cdot \mathbb{P}_{y^{*}_{t}}y_{t} = H) \tag{By the independence of $a$ and $y$} \\
& = \frac{1}{2}\sum_{t=1}^{T}\mathbb{P}_{y^{*}_{1:t-1},noreg,\sigma}(a_{t} = work)  - \frac{1}{2}\sum_{t=1}^{T}\mathbb{P}_{y^{*}_{1:t-1},noreg,\sigma}(a_{t} = work)  = 0 \\
\end{align*}

This derives a contradiction, proving our claim.

\end{proof}

\begin{lemma}
If $y_{1} = M$ then no Principal mechanism can get expected payoff more than $\frac{T}{4} + o(T)$ payoff against $\mathcal{L}$, in expectation over $y^{*}_{2:T}$. If $y_{1} = H$ then no Principal mechanism can get expected payoff more than $\frac{T}{5} + o(T)$ payoff against $\mathcal{L}$, in expectation over $y^{*}_{2:T}$. 
\label{lem:payoffUBextended}
\end{lemma}

\begin{proof}
First, assume $y_{1} = M$. Furthermore, let $t'$ be the first round in which the Principal mechanism $\sigma$ does not play $(0.5,work)$. Then, the payoff of the Principal is

\begin{align*}
& \mathbb{E}_{y_{2:T}^{*},\cL,\sigma}[\sum_{t = 1}^{t'}V(a_{t}^{\sigma},(0.5,work),y_{t})] + \mathbb{E}_{y_{2:T}^{*},\cL,\sigma}[\sum_{t = t'}^{T}V(a_{t}^{\sigma},p_{t}^{\sigma},y_{t})] \\
& = \mathbb{E}_{y_{2:T}^{*},(a^{*})^{\sigma}}[\sum_{t = 1}^{t'}V(a_{t}^{\sigma},(0.5,work),y_{t})] + \mathbb{E}_{y_{2:T}^{*},noreg,\sigma}[\sum_{t = t'}^{T}V(noreg^{\sigma},\sigma,p_{t}^{\sigma},y_{t})] \\
& = \frac{t'}{2} - \frac{t'}{4} + \mathbb{E}_{y_{2:T}^{*},noreg,\sigma}[\sum_{t = t'}^{T}V(noreg^{\sigma},\sigma,p_{t}^{\sigma},y_{t})] \\
& = \frac{t'}{4} + o(T)  \tag{By Lemma~\ref{lem:PrincipalpayoffUB}}\\
& \leq \frac{T}{4} + o(T) \\
\end{align*}

The analysis is similar for $y_{1} = H$. Let $t'$ be the first round in which the Principal mechanism $\sigma$ does not play $(0.6,work)$. Then, the payoff of the Principal is

\begin{align*}
& \mathbb{E}_{y_{2:T}^{*},\cL,\sigma}[\sum_{t = 1}^{t'}V(a_{t}^{\sigma},(0.6,work),y_{t})] + \mathbb{E}_{y_{2:T}^{*},\cL,\sigma}[\sum_{t = t'}^{T}V(a_{t}^{\sigma},p_{t}^{\sigma},y_{t})] \\
& = \mathbb{E}_{y_{2:T}^{*},(b^{*})^{\sigma}}[\sum_{t = 1}^{t'}V(a_{t}^{\sigma},(0.6,work),y_{t})] + \mathbb{E}_{y_{2:T}^{*},noreg,\sigma}[\sum_{t = t'}^{T}V(noreg^{^{\sigma}},\sigma,p_{t}^{\sigma},y_{t})] \\
& = \frac{2t'}{5} - \frac{t'}{5} + \mathbb{E}_{y_{2:T}^{*},noreg,\sigma}[\sum_{t = t'}^{T}V(noreg^{^{\sigma}},\sigma,p_{t}^{\sigma},y_{t})] \\
& = \frac{t'}{5} + o(T)  \tag{By Lemma~\ref{lem:PrincipalpayoffUB}}\\
& \leq \frac{T}{5} + o(T) \\
\end{align*}

\end{proof}

\begin{lemma}
For any prefix of play of length $T' \leq T$, as long as $balanced_{all}=true$ and the Principal plays only $0.5,work$ for all $\sigma$,
$$\E_{a^{*},\sigma}[\ir(y_{1:T},p_{1:T},r_{1:T})] \leq o(T)$$
and 
$$\E_{a^{*},\sigma}[\nr(y_{1:T},p_{1:T}^\sigma,r^\sigma_{1:T})] \leq o(T)$$

Similarly, for any prefix of play of length $T' \leq T$, as long as $balanced_{all}=true$, the Principal plays only $0.6,work$, for all $\sigma$,
$$\E_{b^{*},\sigma}[\ir(y_{1:T},p_{1:T},r_{1:T})] \leq o(T)$$
and
$$\E_{b^{*},\sigma}[\nr(y_{1:T},p_{1:T}^\sigma,r^\sigma_{1:T})] \leq o(T)$$
\label{lem:boundreg}
\end{lemma}

\begin{proof}

For the first case of $(0.5,work)$ and $a^{*}$, the Agent is always mapping $M$ to $work$ and $H$ to $shirk$. As $work$ gets payoff $2p - 1 \geq 0$ under $m$ and $shirk$ gets $0$, while $work$ gets $-1$ under $H$ and $shirk$ gets $0$, this is the optimal mapping. Therefore the Contextual Swap Regret is $0$. Now we can upper bound the negative regret:

\begin{align*}
& \E_{a^{*},\sigma}\left[\max_{h:\cP\times \cA\mapsto \cA}\sum_{t=1}^{T'} U(h(p_t^{\sigma},r_t^{\sigma}),p_t^{\sigma}, y_t)) - U(a_{t}^{\sigma},p_t^{\sigma}, y_t)) \right] \\
& =\E_{a^{*},\sigma}\left[\max_{a \in \cA}\sum_{t=1}^{T'} U(h(0.5,work),0.5, y_t)) - U(a_{t}^{\sigma},0.5, y_t)) \right] \tag{By the fact that the Principal is making a fixed (policy, recommendation) pair across all $t\leq T'$} \\
&= \E_{a^{*},\sigma}\left[\max_{a \in \cA}\sum_{t=1}^{T'} U(h(0.5,work),0.5, y_t)) \right] \tag{By definition of $a^{*}$} \\
&= \max(\E_{a^{*},\sigma}[\sum_{t=1}^{T'} U(work,0.5, y_t)], \E_{a^{*},\sigma}[\sum_{t=1}^{T'} U(shirk,0.5, y_t)) \\
&= \max(\E_{a^{*},\sigma}[\frac{1}{2}m_{y,T'} - h_{y,T}], 0)  \\
&\leq \max(\E_{a^{*},\sigma}[\frac{1}{2}h_{y,T'} + o(T) - h_{y,T}], 0) \tag{By the fact that $balanced_{t}$ is true over the entire prefix.}  \\
& \leq o(T)
\end{align*}

For the second case of $(0.6, work)$ and $b^{*}$, let us use $m_{y,T'}$ to refer to the number of $m$ states in the sequence, and $h_{y,T'}$ to refer to the number of $h$ states:

\begin{align*}
& \E_{b^{*},\sigma}\left[\max_{h:\cP\times \cA\times \cA\mapsto \cA}\sum_{t=1}^{T'} U(h(p_t^{\sigma},r_t^{\sigma},a_t^{\sigma}),p_t^{\sigma}, y_t)) - U(a_{t}^{\sigma},p_t^{\sigma}, y_t)) \right] \\
& =\E_{b^{*},\sigma}\left[\max_{h: \cA\mapsto \cA}\sum_{t=1}^{T'} U(h(0.6,work,a_t^{\sigma}),0.6, y_t)) - U(a_{t}^{\sigma},0.6, y_t)) \right] \tag{By the fact that the Principal is making a fixed (policy, recommendation) pair across all $t\leq T'$} \\
&= \E_{b^{*},\sigma}\left[\max_{h: \cA\mapsto \cA}\sum_{t=1}^{T'} U(h(0.6,work,a_t^{\sigma}),0.6, y_t)) - \frac{1}{5}(m_{y,T'} - h_{y,T'}) \right] \tag{By definition of $b^{*}$} \\
&= \E_{b^{*},\sigma}[\max_{a \in \cA}\sum_{t=1}^{T'} U(a,0.6, y_t))\mathbbm{1}[a_{t}^{\sigma} = work]] \\& + \E_{b^{*},\sigma}[\max_{a \in \cA}\sum_{t=1}^{T'} U(a,0.6, y_t))\mathbbm{1}[a_{t}^{\sigma} = shirk]] - \E_{b^{*},\sigma}[\frac{1}{5}(m_{y,T'} - h_{y,T'}) ]  \\
&= \max(\E_{b^{*},\sigma}[\sum_{t=1}^{T'} U(work,0.6, H))\mathbbm{1}[a_{t}^{\sigma} = shirk]], 0) - \frac{1}{5}(m_{y,T'} - h_{y,T'})  \tag{By the fact that $b^{*}$ only shirks when $y = H$, and that shirking always guarantees payoff $0$.}  \\
&= \max(m_{y},T' - h_{y,T'}, 0) - \frac{1}{5}(m_{y,T'} - h_{y,T'})   \tag{By the distribution of the states conditioned on $b^{*}$ playing $work$} \\
& \leq o(T) \tag{By the fact that $balanced_{t}$ is true over the entire prefix.} 
\end{align*}

Thus, in the second case the Contextual Swap Regret is upper bounded. Finally, we need that the Negative Regret is upper bounded: 

\begin{align*}
& \E_{b^{*},\sigma}\left[\max_{h:\cP\times \cA\mapsto \cA}\sum_{t=1}^{T'} U(h(p_t^{\sigma},r_t^{\sigma}),p_t^{\sigma}, y_t)) - U(a_{t}^{\sigma},p_t^{\sigma}, y_t)) \right] \\
& =\E_{b^{*},\sigma}\left[\max_{a \in \cA}\sum_{t=1}^{T'} U(h(0.6,work),0.6, y_t)) - U(a_{t}^{\sigma},0.6, y_t)) \right] \tag{By the fact that the Principal is making a fixed (policy, recommendation) pair across all $t\leq T'$} \\
&= \E_{a^{*},\sigma}\left[\max_{a \in \cA}\sum_{t=1}^{T'} U(h(0.6,work),0.6, y_t)) - \frac{1}{5}(m_{y,T'} - h_{y,T'}) \right] \tag{By definition of $b^{*}$} \\
&= \max(\E_{b^{*},\sigma}[\sum_{t=1}^{T'} U(work,0.6, y_t)], \E_{a^{*},\sigma}[\sum_{t=1}^{T'} U(shirk,0.6, y_t)) - \E_{a^{*},\sigma}[(m_{y,T'} - h_{y,T'}) ]  \\
&= \max(\E_{b^{*},\sigma}[\frac{1}{2}m_{y,T'} - h_{y,T}], 0) - \E_{b^{*},\sigma}[(m_{y,T'} - h_{y,T'}) ]  \\
& \leq o(T) \tag{By the fact that $balanced_{t}$ is true over the entire prefix.}  \\
\end{align*}

\end{proof}
\chapter{Eliciting User Preferences for Personalized Multi-Objective Decision Making
through Comparative Feedback}
\label{app:momdp}
\section{Proof of Lemma~\ref{lmm:V1}}\label{app:V1}
\lmmVinit*
\begin{proof}
%
We first show that, according to our algorithm (lines~\ref{alg-line:initV1-1}-\ref{alg-line:initV1-5} of Algorithm~\ref{alg:policy-threshold-free}), the returned $\pi_1$ satisfies that
$$\inner{w^*}{V^{\pi_1}}\geq \max_{i\in [k]}\inner{w^*}{V^{\pi^{\be_i}}}-\epsilon\,.$$ 
This can be proved by induction over $k$. In the base case of $k=2$, it's easy to see that the returned $\pi_1$ satisfies the above inequality.
Suppose the above inequality holds for any $k\leq n-1$ and we prove that it will also hold for $k=n$.
After running the algorithm over $j=2,\ldots,k-1$ (line $2$), the returned policy $\pi_{e^*}$ satisfies that
$$\inner{w^*}{V^{\pi^{e^*}}}\geq \max_{i\in [k-1]}\inner{w^*}{V^{\pi^{\be_i}}}-\epsilon\,.$$
Then there are two cases, 
\begin{itemize}
    \item If $\inner{w^*}{V^{\pi^{e^*}}}< \inner{w^*}{V^{\pi^{\be_k}}}-\epsilon$, we will return $\pi_1 = \pi^{\be_k}$ and also, $\inner{w^*}{V^{\pi^{\be_k}}}\geq \inner{w^*}{V^{\pi^{\be_i}}}-\epsilon$ for all $i\in [k]$.
    \item If $\inner{w^*}{V^{\pi^{e^*}}}\geq \inner{w^*}{V^{\pi^{\be_k}}}-\epsilon$, then we will return $\pi^{e^*}$ and it satisfies $\inner{w^*}{V^{\pi^{e^*}}}\geq \max_{i\in [k]}\inner{w^*}{V^{\pi^{\be_i}}}-\epsilon$\,.
\end{itemize}


As $\pi^{\be_i}$ is the optimal personalized policy when the user's preference vector is $\be_i$, we have that $$v^* =\inner{w^*}{V^{\pi^*}} = \sum_{i=1}^k w^*_i \inner{V^{\pi^*}}{\be_i}\leq \sum_{i=1}^k w^*_i \inner{V^{\pi^{\be_i}}}{\be_i} \leq \inner{w^*}{\sum_{i=1}^k V^{\pi^{\be_i}}}\,,$$
where the last inequality holds because the entries of $V^{\pi^{\be_i}}$ and $w^*$ are non-negative.
Therefore, there exists $i\in [k]$ such that $\inner{w^*}{V^{\pi^{\be_i}}} \geq \frac{1}{k} \inner{w^*}{V^{\pi^*}} =\frac{1}{k}v^*$. 

Then we have 
$$\inner{w^*}{V^{\pi_1}}\geq \max_{i\in[k]}\inner{w^*}{V^{\pi^{\be_i}}} -\epsilon \geq \frac{1}{k} v^*-\epsilon \geq \frac{1}{2k} v^*\,,$$ when $\epsilon \leq \frac{v^*}{2k}$.
By rearranging terms, we have $\frac{v^*}{\inner{w^*}{V^{\pi_1}}} \leq 2k$.

By setting $\calpha = 2k$, we have $\abs{\hat \alpha_i - \alpha_i}\inner{w^*}{V^{\pi_1}} \leq \calpha\epsilon = 2k\epsilon$ and thus, $\abs{\hat \alpha_i - \alpha_i}\leq \frac{4k^2\epsilon}{v^*}$.
\end{proof}
\section{Pseudo Code of Computation of the Basis Ratios}\label{app:bin-search}
The pseudo code of searching $\hat\alpha_i$'s is described in Algorithm~\ref{alg:alpha}.
\begin{algorithm}[H]\caption{Computation of Basis Ratios}\label{alg:alpha}
    \begin{algorithmic}[1]
        \STATE \textbf{input:} $(V^{\pi_1},\ldots, V^{\pi_d})$ and $\calpha$
        \FOR{$i=1,\ldots, d-1$}
        \STATE let $l=0$, $h=2C_\alpha$ and $\hat \alpha_i = C_\alpha$
        \WHILE{True}
        \IF{$\hat \alpha_i>1$}
        \STATE compare $\pi_1$ and $\frac{1}{\hat \alpha_i}\pi_{i+1} + (1-\frac{1}{\hat \alpha_i})\pi_0$; 
        \textbf{if} $\pi_1 \succ \frac{1}{\hat \alpha_i}\pi_{i+1} + (1-\frac{1}{\hat \alpha_i})\pi_0$ \textbf{then} $h\leftarrow \hat \alpha_i$, $\hat \alpha_i\leftarrow \frac{l+h}{2}$; \textbf{if} $\pi_1 \prec \frac{1}{\hat \alpha_i}\pi_{i+1} + (1-\frac{1}{\hat \alpha_i})\pi_0$ \textbf{then} $l\leftarrow \hat \alpha_i$, $\hat \alpha_i\leftarrow \frac{l+h}{2}$  
        \ELSE
        \STATE compare $\pi_{i+1}$ and $\hat{\alpha}_i\pi_1 + (1-\hat{\alpha}_i)\pi_0$; \textbf{if} $\hat{\alpha}_i\pi_1 + (1-\hat{\alpha}_i)\pi_0\succ \pi_{i+1}$ \textbf{then} $h\leftarrow \hat \alpha_i$, $\hat \alpha_i\leftarrow \frac{l+h}{2}$; \textbf{if} $\hat{\alpha}_i\pi_1 + (1-\hat{\alpha}_i)\pi_0\prec \pi_{i+1}$ \textbf{then} $l\leftarrow \hat \alpha_i$, $\hat \alpha_i\leftarrow \frac{l+h}{2}$
        \ENDIF
        \IF{``indistinguishable'' is returned}
        \STATE break
        \ENDIF
        \ENDWHILE
        \ENDFOR
        \STATE \textbf{output: } $(\hat \alpha_1,\ldots,\hat \alpha_{d-1})$ 
    \end{algorithmic}
\end{algorithm}
\section{Proof of Theorem~\ref{thm:est-without-tau}}
\thmwithouttau*

\begin{proof}
Theorem~\ref{thm:est-without-tau} follows by setting $\ealpha = \frac{4k^2\epsilon}{v^*}$ and $C_\alpha = 2k$ as shown in Lemma \ref{lmm:V1} and combining the results of Lemma~\ref{lmm:gap-hatw-w} and \ref{lmm:est-without-tau} with the triangle inequality.

Specifically, for any policy $\pi$, we have
\begin{align}\label{eq:Lem1Helper}
\begin{split}
\;\abs{\inner{\hat w}{V^{\pi}} - \inner{w'}{V^{\pi}}} \leq & \abs{\inner{\hat w}{V^{\pi}} - \inner{\hat w^{(\delta)}}{V^{\pi}}}+\abs{\inner{\hat w^{(\delta)}}{V^{\pi}} - \inner{w'}{V^{\pi}}}  \\
\leq &\cO((\sqrt{k} +1)^{d-d_\delta} \calpha (\frac{\calpha\cv^4d_\delta^{\frac{3}{2}}\norm{w'}_2^2\ealpha }{\delta^2}+\sqrt{k} \delta \norm{w'}_2))\\
\leq & \cO((\sqrt{k} +1)^{d-d_\delta +3}\norm{w'}_2 (\frac{\cv^4 k^4 \norm{w'}_2\epsilon}{v^*\delta^2}+ \delta))\,.
\end{split}
\end{align}

Since $\norm{w'} = \frac{\norm{w^*}}{\inner{w^*}{V^{\pi_1}}}$ and $\inner{w^*}{V^{\pi_1}} \geq \frac{v^*}{2k}$ from 
Lemma~\ref{lmm:V1},
we derive 
\begin{align*}
    &v^*- \inner{w^*}{V^{\pi^{\hat w}}} = \inner{w^*}{V^{\pi_1}} \left(\inner{w'}{V^{\pi^*}} - \inner{w'}{V^{\pi^{\hat w}}}\right)
    \\
    \leq&  \inner{w^*}{V^{\pi_1}}\left(\inner{\hat w}{V^{\pi^*}} - \inner{\hat w}{V^{\pi^{\hat w}}} + \cO((\sqrt{k} +1)^{d-d_\delta +3}\norm{w'}_2 (\frac{\cv^4 k^4 \norm{w'}_2\epsilon}{v^*\delta^2}+ \delta)) \right)
    \\
    \leq&   \cO\left(\inner{w^*}{V^{\pi_1}}(\sqrt{k} +1)^{d-d_\delta +3}\norm{w'}_2 (\frac{\cv^4 k^4 \norm{w'}_2\epsilon}{v^*\delta^2}+ \delta)\right)\\
    =&   \cO\left((\sqrt{k} +1)^{d-d_\delta +3}\norm{w^*}_2 (\frac{\cv^4 k^5 \norm{w^*}_2\epsilon}{v^{*2}\delta^2}+ \delta)\right)\\
    =& \cO\left((\frac{\cv^2 \norm{w^*}_2^2}{v^*})^\frac{2}{3}(\sqrt{k} +1)^{d+ \frac{16}{3}}  \epsilon^\frac{1}{3}\right)\,.
\end{align*}
The first inequality follows from
\begin{align*}
    \inner{w'}{V^{\pi^*}} - \inner{w'}{V^{\pi^{\hat w}}}=& \inner{w'}{V^{\pi^*}}- \inner{w'}{V^{\pi^{\hat w}}}\\ &+\left(\inner{\hat w}{V^{\pi^*}}-\inner{\hat w}{V^{\pi^*}}\right)+\left(\inner{\hat w}{V^{\pi^{\hat w}}}-\inner{\hat w}{V^{\pi^{\hat w}}}\right), 
\end{align*}
and applying (\ref{eq:Lem1Helper}) twice- once for $\pi^*$ and once for $\pi^{\hat w}$. The  last inquality follows by setting $\delta = \left(\frac{\cv^4k^5 \norm{w^*}_2\epsilon}{v^{*2}}\right)^{\frac{1}{3}}$.
\end{proof}
\section{Proof of Lemma~\ref{lmm:gap-hatw-w}}
\lmmhatww*
To prove Lemma~\ref{lmm:gap-hatw-w}, we will first define a matrix, $ A^\textrm{(full)}$.

Given the output $(V^{\pi_1},\ldots, V^{\pi_{d}})$ of Algorithm~\ref{alg:policy-threshold-free}, we have $\rank(\Span(\{V^{\pi_1},\ldots, V^{\pi_{d_\delta}}\})) = d_\delta$.

Let $\basis_1,\ldots,\basis_{d-d_\delta}$ be a set of orthonormal vectors that are orthogonal to $\Span(V^{\pi_1},\ldots, V^{\pi_{d_\delta}})$ and together with $V^{\pi_1},\ldots, V^{\pi_{d_\delta}}$ form a basis for $\Span(\{V^\pi|\pi\in \Pi\})$.

We define $A^\textrm{(full)}\in \R^{d\times k}$ as the matrix of replacing the last $d-d_\delta$ rows of $A$ with $\basis_1,\ldots,\basis_{d-d_\delta}$, i.e.,
\begin{equation*}
    A^\textrm{(full)} = \begin{pmatrix}
    V^{\pi_1\top}\\
    (\alpha_1 V^{\pi_1} -V^{\pi_2})^\top\\
    \vdots\\
    (\alpha_{d_\delta-1} V^{\pi_1} -V^{\pi_{d_\delta}})^\top\\
    \basis_1^\top\\
    \vdots\\
    \basis_{d-d_\delta}^\top
    \end{pmatrix}\,.
\end{equation*}
\begin{observation}
We have that $\Span(A^\textrm{(full)}) = \Span(\{V^\pi|\pi\in \Pi\})$ and $\rank(A^\textrm{(full)}) =d$. 
\end{observation}

\begin{restatable}{lemma}{lmmwwprime}\label{lmm:gap-w-w'}
    For all $w \in \R^k$ satisfying $A^\textrm{(full)} w = \be_1$, we have $\abs{w\cdot V^\pi - w'\cdot V^\pi}\leq \sqrt{k} \delta \norm{w'}_2$ for all $\pi$.
\end{restatable}

We then show that there exists a $w \in \R^k$ satisfying $A^\textrm{(full)} w = \be_1$ such that $\abs{\hatwd\cdot V^\pi - w\cdot V^\pi}$ is small for all $\pi\in \Pi$.
\begin{restatable}{lemma}{lmmminnorm}\label{lmm:min_norm_hatw_delta}
    If $\abs{\hat \alpha_i-\alpha_i}\leq \ealpha$ and $\alpha_i\leq \calpha$ for all $i\in [d-1]$, 
    for every $\delta \geq 4 \calpha^{\frac{2}{3}}\cv d^{\frac{1}{3}}\ealpha^\frac{1}{3}$
    there exists a $w \in \R^k$ satisfying $A^\textrm{(full)} w = \be_1$ s.t. $\abs{\hatwd\cdot V^\pi - w\cdot V^\pi}\leq \cO( \frac{\calpha\cv^4d_\delta^{\frac{3}{2}}\norm{w'}_2^2\ealpha }{\delta^2})$ for all $\pi$.
\end{restatable}

We now derive Lemma~\ref{lmm:gap-hatw-w} using the above two lemmas.
\begin{proof}[Proof of Lemma~\ref{lmm:gap-hatw-w}]
Let $w$ be defined in Lemma~\ref{lmm:min_norm_hatw_delta}. Then for any policy $\pi$, we have
\begin{align*}
    &\abs{\hatwd  \cdot V^\pi - w'\cdot V^\pi} \leq \abs{\hatwd \cdot V^\pi - w\cdot V^\pi}+\abs{w\cdot V^\pi - w'\cdot V^\pi}
    \\
    \leq& \cO( \frac{\calpha\cv^4d_\delta^{\frac{3}{2}}\norm{w'}_2^2\ealpha }{\delta^2}+\sqrt{k} \delta \norm{w'}_2) \,,
\end{align*}
by applying Lemma~\ref{lmm:gap-w-w'} and \ref{lmm:min_norm_hatw_delta}.
\end{proof}

\section{Proofs of Lemma~\ref{lmm:gap-w-w'} and Lemma~\ref{lmm:min_norm_hatw_delta}}\label{app:tool-lmms}
\lmmwwprime*
\begin{proof}[Proof of Lemma~\ref{lmm:gap-w-w'}]
Since $\Span(A^\textrm{(full)}) = \Span(\{V^\pi|\pi\in \Pi\})$, for every policy $\pi$, the value vector can be represented as a linear combination of row vectors of $A^\textrm{(full)}$, i.e., there exists $a = (a_1,\dots,a_d)\in \R^d$ s.t. 
\begin{align}\label{eq:lem4Helper}
    V^\pi = \sum_{i=1}^d{a_i A^\textrm{(full)}_i} = A^{\textrm{(full)}\top} a\,.
\end{align}
Now, for any unit vector $\xi \in \Span(\basis_1,\ldots,\basis_{d-d_\delta})$, we have $\inner{V^\pi}{\xi} \leq \sqrt{k} \delta$.

The reason is that at each round ${d_\delta+1}$, we pick an orthonormal basis $\rho_1,\ldots,\rho_{k-{d_\delta}}$ of $\Span(V^{\pi_1},\ldots,V^{\pi_{d_\delta}})^\perp$  (line~\ref{alg-line:orthonormal-basis} in Algorithm~\ref{alg:policy-threshold-free}) and pick $u_{d_\delta+1}$ to be the one in which there exists a policy with the largest component as described in line~\ref{alg-line:largest-component}.
Hence, $\abs{\inner{\rho_j}{V^\pi}}\leq \delta$ for all $j\in [k-{d_\delta}]$.

It follows from  Cauchy-Schwarz inequality that  $\inner{\xi}{V^\pi}= \sum_{j=1}^{k-{d_\delta}} \inner{\xi}{\rho_j}\inner{\rho_j}{V^\pi} \leq \sqrt{k}\delta$.

Combining  with the observation that $b_1,\ldots,b_{d-d_\delta}$ are pairwise 
 orthogonal 
and that each of them is 
orthogonal to $\Span(V^{\pi_1},\ldots,V^{\pi_{d_\delta}})$ we have $$\sum_{i={d_\delta+1}}^d a_i^2 = \abs{\inner{V^\pi}{\sum_{i=d_\delta+1}^d a_{i} \basis_{i-d_\delta}}} \leq \sqrt{\sum_{i={d_\delta+1}}^d a_i^2} \sqrt{k}\delta\,,$$
which implies that
\begin{equation}
    \sqrt{\sum_{i={d_\delta+1}}^d a_i^2} \leq \sqrt{k}\delta\,.\label{eq:anorm}
\end{equation}

Since $w'$ satisfies $Aw' =\be_1$, we have $$A^\textrm{(full)} w' = (1,0,\ldots,0, \inner{\basis_1}{w'},\ldots,\inner{\basis_{d-d_\delta}}{w'})\,.$$
For any $w \in \R^k$ satisfying $A^\textrm{(full)} w = \be_1$, consider $\tilde w = w +\sum_{i= 1}^{d-d_\delta}\inner{\basis_{i}}{w'} \basis_{i}$. 
Then we have $A^\textrm{(full)}\tilde w = A^\textrm{(full)}w'$.

Thus, applying (\ref{eq:lem4Helper}) twice, we get
\begin{align*}
    \tilde w\cdot V^\pi =  \tilde w^\top A^{\textrm{(full)}\top}a = w'^\top A^{\textrm{(full)}\top}a = w'\cdot V^\pi\,.
\end{align*} 
Hence, 
\begin{align*}
    &\abs{w\cdot V^\pi- w'\cdot V^\pi} = \abs{w\cdot V^\pi- \tilde w\cdot V^\pi} \stackrel{(a)}=\abs{\sum_{i=1}^d a_i (w - \tilde w)\cdot A^\textrm{(full)}_i  }\\
    =& \abs{\sum_{i=d_\delta +1}^{d} a_i \inner{\basis_{i-d_\delta}}{w'}} 
    \stackrel{(b)}{\leq}  \sqrt{\sum_{i={d_\delta+1}}^d a_i^2} \norm{w'}_2 \stackrel{(c)}{\leq} \sqrt{k} \delta \norm{w'}_2\,,
\end{align*}
where Eq~(a) follows from (\ref{eq:lem4Helper}), inequality~(b) from Cauchy-Schwarz, and inequality~(c) from applying \eqref{eq:anorm}.

\end{proof}

\lmmminnorm*
Before proving Lemma~\ref{lmm:min_norm_hatw_delta}, we introduce some notations and a claim.
\begin{itemize}
    \item For any $x,y\in\R^k$, let $\theta(x,y)$ denotes the angle between $x$ and $y$.
    \item For any subspace $U\subset \R^k$, let $\theta(x,U) := \min_{y\in U}\theta(x,y)$.
    \item For any two subspaces $U,U'\subset \R^k$, we define $\theta(U,U')$ as $\theta(U,U') = \max_{x\in U} \min_{y\in U'}\theta(x,y)$.
    \item For any matrix $M$, let $M_i$ denote the $i$-th row vector of $M$, $M_{i:j}$ denote the submatrix of $M$ composed of rows $i,i+1,\ldots,j$, and $M_{i:}$ denote the submatrix composed of all rows $j\geq i$.
    \item Let $\Span(M)$ denote the span of the rows of $M$.
\end{itemize}

Recall that $\hat A\in \R^{d\times k}$ is defined as 
\begin{align*}
    \hat A = \begin{pmatrix}
    V^{\pi_1\top}\\
    (\hat \alpha_1 V^{\pi_1} -V^{\pi_2})^\top\\
    \vdots\\
    (\hat \alpha_{d-1} V^{\pi_1} -V^{\pi_d})^\top
    \end{pmatrix}\,,
\end{align*}
which is the approximation of matrix $A\in \R^{d\times k}$ defined by true values of $\alpha_i$, i.e.,
\begin{align*}
    A = \begin{pmatrix}
    V^{\pi_1\top}\\
    (\alpha_1 V^{\pi_1} -V^{\pi_2})^\top\\
    \vdots\\
    (\alpha_{d-1} V^{\pi_1} -V^{\pi_d})^\top
    \end{pmatrix}\,.
\end{align*}
We denote by $\hat A^{(\delta)} = \hat A_{1:d_\delta}, A^{(\delta)}=A_{1:d_\delta}\in \R^{d_\delta\times k}$ the sub-matrices comprised of the first $d_\delta$ rows of $\hat A$ and $A$ respectively.

\begin{restatable}{claim}{lmmspanangle}\label{clm:spanangle}
If $\abs{\hat \alpha_i-\alpha_i}\leq \ealpha$ and $\alpha_i\leq \calpha$ for all $i\in [d-1]$, 
    for every $\delta \geq 4 \calpha^{\frac{2}{3}}\cv d^{\frac{1}{3}}\ealpha^\frac{1}{3}$, we have
\begin{equation}
    \theta(\Span(A^{(\delta)}_{2:}), \Span(\hat A^{(\delta)}_{2:})) \leq \eta_{\ealpha,\delta}\,,\label{eq:theta-A-hatA}
\end{equation}
and
\begin{equation}
    \theta(\Span(\hat A^{(\delta)}_{2:}), \Span(A^{(\delta)}_{2:})) \leq 
\eta_{\ealpha,\delta}\,,\label{eq:theta-hatA-A}
\end{equation}
where $\eta_{\ealpha,\delta} = \frac{4\calpha\cv^2 d_\delta\ealpha }{\delta^2}$.
\end{restatable}
To prove the above claim, we use the following lemma by~\cite{Balcan2015}.

\begin{lemma}[Lemma 3 of~\cite{Balcan2015}]\label{lmm:Balcan}
Let $U_l = \Span(\xi_1,\ldots,\xi_l)$ and $\hat U_l = \Span(\hat \xi_1,\ldots,\hat \xi_l)$.
Let $\eacc, \gamma_\text{new}\geq 0$ and $\eacc\leq \gamma_\text{new}^2/(10l)$, and assume that $\theta(\hat \xi_i, \hat U_{i-1})\geq \gamma_\text{new}$ for $i=2,\ldots, l$, and that  $\theta(\xi_{i}, \hat \xi_{i})\leq \eacc$ for $i=1,\ldots, l$.

Then,
$$\theta(U_l, \hat U_l)\leq 2l\frac{\eacc}{\gamma_\text{new}}\,.$$
\end{lemma}
\begin{proof}[Proof of Claim~\ref{clm:spanangle}]
For all $2\leq i \leq d_\delta$, we have that
\begin{align*}
    \theta(\hat A^{(\delta)}_i, \Span(\hat A^{(\delta)}_{2:i-1})) \geq& \theta(\hat A^{(\delta)}_i, \Span(\hat A^{(\delta)}_{1:i-1})) \geq \sin( \theta(\hat A^{(\delta)}_i, \Span(\hat A^{(\delta)}_{1:i-1})))\\ \stackrel{(a)}{\geq}&\frac{\abs{\hat A^{(\delta)}_i \cdot u_i}}{\norm{\hat A^{(\delta)}_i}_2}\stackrel{(b)}{\geq}
    \frac{\delta}{\norm{\hat A^{(\delta)}_i}_2} = \frac{\delta}{\norm{\hat \alpha_{i-1} V^{\pi_1} -V^{\pi_i}}_2} \geq \frac{\delta}{(\calpha+1)\cv}
\,,
\end{align*}
where Ineq~(a) holds as $u_i$ is orthogonal to $\Span(\hat A^{(\delta)}_{1:i-1})$ according to line~\ref{alg-line:orthonormal-basis} of Algorithm~\ref{alg:policy-threshold-free} and Ineq~(b) holds due to $\abs{\hat A^{(\delta)}_i \cdot u_i} = \abs{V^{\pi_i}\cdot u_i}\geq \delta$. 
The last inequality holds due to $\norm{\hat \alpha_{i-1} V^{\pi_1} -V^{\pi_i}}_2 \leq \hat \alpha_{i-1} \norm{V^{\pi_1}}_2+\norm{V_i}_2 \leq (\calpha+1)\cv$.

Similarly, we also have
\[\theta(A^{(\delta)}_i, \Span(A^{(\delta)}_{2:i-1}))\geq \frac{\delta}{(\calpha+1)\cv}\,.\]

We continue by decomposing $V^{\pi_i}$ in the direction of $V^{\pi_1}$ and the direction perpendicular to $V^{\pi_1}$. 

For convince, we denote $v_i^\parallel: = V^{\pi_i}\cdot \frac{V^{\pi_1}}{\norm{V^{\pi_1}}_2}$, $V_i^\parallel :=v_i^\parallel \frac{V^{\pi_1}}{\norm{V^{\pi_1}}_2}$, $V_i^\perp :=V^{\pi_i} -V_i^\parallel$ and $v_i^\perp := \norm{V_i^\perp}_2$.

Then we have 
\[\theta( A^{(\delta)}_i,\hat A^{(\delta)}_i) = \theta(\alpha_{i-1} V^{\pi_1} -V^{\pi_i},\hat\alpha_{i-1} V^{\pi_1} -V^{\pi_i})= \theta(\alpha_{i-1} V^{\pi_1} -V_{i}^\parallel- V_i^\perp,\hat\alpha_{i-1} V^{\pi_1} -V_{i}^\parallel- V_i^\perp)\,.\]
If $(\hat \alpha_{i-1} V^{\pi_1} -V_{i}^\parallel)\cdot(\alpha_{i-1} V^{\pi_1} -V_{i}^\parallel) \geq 0$, i.e., $\hat \alpha_{i-1} V^{\pi_1} -V_{i}^\parallel $ and $\alpha_{i-1} V^{\pi_1} -V_{i}^\parallel$ are in the same direction, then
\begin{align}
    \theta(A^{(\delta)}_i,\hat  A^{(\delta)}_i) 
    &= \abs{\arctan \frac{\norm{\hat \alpha_{i-1} V^{\pi_1} -V_{i}^\parallel}_2}{v_i^\perp}-\arctan \frac{\norm{\alpha_{i-1} V^{\pi_1} -V_{i}^\parallel}_2}{v_i^\perp} }\nonumber\\
    & \leq \abs{\frac{\norm{\hat \alpha_{i-1} V^{\pi_1} -V_{i}^\parallel}_2}{v_i^\perp} - \frac{\norm{\alpha_{i-1} V^{\pi_1} -V_{i}^\parallel}_2}{v_i^\perp}}\label{eq:arctan}\\
    &=\frac{\abs{\hat \alpha_{i-1} - \alpha_{i-1}}\norm{V^{\pi_1}}_2}{v_i^\perp}\nonumber\\
    &\leq \frac{\ealpha \cv}{\delta}\,,\label{eq:applytau}
\end{align}
where Ineq~\eqref{eq:arctan} follows from the fact that the derivative of $\arctan$ is at most $1$, i.e., $\frac{\partial \arctan x}{\partial x} =\lim_{a\to x}\frac{\arctan a-\arctan x}{a-x}= \frac{1}{1+x^2}\leq 1$.

Inequality~\eqref{eq:applytau} holds since $v_i^\perp \geq \abs{\inner{V^{\pi_i}}{u_i}}\geq \delta$.

If $(\hat \alpha_{i-1} V^{\pi_1} -V_{i}^\parallel)\cdot(\alpha_{i-1} V^{\pi_1} -V_{i}^\parallel) < 0$, i.e., $\hat \alpha_{i-1} V^{\pi_1} -V_{i}^\parallel $ and $\alpha_{i-1} V^{\pi_1} -V_{i}^\parallel$ are in the opposite directions, then we have $\norm{\hat \alpha_{i-1} V^{\pi_1} -V_{i}^\parallel}_2+\norm{\alpha_{i-1} V^{\pi_1} -V_{i}^\parallel}_2 = \norm{(\hat \alpha_{i-1} - \alpha_{i-1}) V^{\pi_1}}_2\leq \ealpha \norm{V^{\pi_1}}_2$. 

Similarly, we have
\begin{align*}
    \theta(\hat A^{(\delta)}_i, A^{(\delta)}_i) 
    &= \abs{\arctan \frac{\norm{\hat \alpha_{i-1} V^{\pi_1} -V_{i}^\parallel}_2}{v_i^\perp}+\arctan \frac{\norm{\alpha_{i-1} V^{\pi_1} -V_{i}^\parallel}_2}{v_i^\perp} } \\
    & \leq \abs{\frac{\norm{\hat \alpha_{i-1} V^{\pi_1} -V_{i}^\parallel}_2}{v_i^\perp} + \frac{\norm{\alpha_{i-1} V^{\pi_1} -V_{i}^\parallel}_2}{v_i^\perp}}\\
    &\leq\frac{\ealpha \norm{V^{\pi_1}}_2}{\abs{v_i^\perp}}\nonumber\\
    &\leq \frac{\ealpha \cv}{\delta}\,.
\end{align*}
By applying Lemma~\ref{lmm:Balcan} with $\eacc =  \frac{\ealpha \cv}{\delta}$, $ \gamma_\text{new}=\frac{\delta}{(\calpha+1)\cv}$, $(\xi_i,\hat \xi_i) = (A_{i+1},\hat A_{i+1})$ (and $(\xi_i,\hat \xi_i) = (\hat A_{i+1}, A_{i+1})$), we have that when 
$\delta \geq 10^{\frac{1}{3}}(\calpha+1)^{\frac{2}{3}}\cv d_\delta^{\frac{1}{3}}\ealpha^\frac{1}{3}$,
\begin{equation*}
    \theta(\Span(A^{(\delta)}_{2:}), \Span(\hat A^{(\delta)}_{2:})) \leq \frac{2d_\delta(\calpha+1) \cv^2\ealpha }{\delta^2}=\eta_{\ealpha,\delta}\,,
\end{equation*}
and
\begin{equation*}
    \theta(\Span(\hat A^{(\delta)}_{2:}), \Span(A^{(\delta)}_{2:})) \leq 
\eta_{\ealpha,\delta}\,.
\end{equation*}
This completes the proof of Claim~\ref{clm:spanangle} since $C_\alpha\geq 1$.
\end{proof}

\begin{proof}[Proof of Lemma~\ref{lmm:min_norm_hatw_delta}]
Recall that $ \hatwd=\argmin_{\hat A^{(\delta)} x =\be_1}\norm{x}_2$ is the minimum norm solution to $\hat A^{(\delta)} x =\be_1$. 

Thus, $\inner{\hatwd}{\basis_i} = 0$ for all $i\in [d-d_\delta]$.

Let $\lambda_1,\ldots,\lambda_{d_\delta-1}$ be any orthonormal basis of $\Span(A^{(\delta)}_{2:})$.

We construct a vector $w$ satisfying $A^\textrm{(full)} w =\be_1$ by removing $\hatwd$'s component in $\Span(A^{(\delta)}_{2:})$ and rescaling.

Formally, 
\begin{equation}\label{eq:creatw}
    w := \frac{\hatwd - \sum_{i=1}^{{d_\delta}-1} \inner{\hatwd}{\lambda_i} \lambda_i}{1 -V^{\pi_1}\cdot( \sum_{i=1}^{{d_\delta}-1} \inner{\hatwd}{\lambda_i} \lambda_i)}\,.
\end{equation}

It is direct to verify that $A_1\cdot w = V^{\pi_1}\cdot w = 1$ and $A_i \cdot w = 0$ for $i=2,\ldots,d_\delta$. As a result, $A^{(\delta)}w =\be_1$.

Combining with the fact that $\hatwd$ has zero component in $\basis_i$ for all $i\in [d-d_\delta]$, we have $A^\textrm{(full)} w =\be_1$.

According to Claim~\ref{clm:spanangle}, we have $$\theta(\Span(A^{(\delta)}_{2:}), \Span(\hat A^{(\delta)}_{2:})) \leq  \eta_{\ealpha,\delta}.$$

Thus, there exist unit vectors $\tilde \lambda_1,\ldots,\tilde \lambda_{d_\delta-1}\in \Span(\hat A^{(\delta)}_{2:})$ such that $\theta(\lambda_i, \tilde \lambda_i)\leq \eta_{\ealpha,\delta}$.

Since $\hat A^{(\delta)} \hatwd = \be_1$, we have $\hatwd \cdot \tilde \lambda_i = 0$ for all 
$i=1,\ldots,d_\delta-1$, and therefore,
\begin{align*}
    \abs{\hatwd \cdot \lambda_i} =  \abs{\hatwd \cdot (\lambda_i-\tilde \lambda_i)} \leq \norm{\hatwd}_2 \eta_{\ealpha,\delta}\,.
\end{align*}
This implies that for any policy $\pi$,
\[\abs{V^\pi\cdot\sum_{i=1}^{d_\delta-1} (\hatwd\cdot \lambda_i) \lambda_i}\leq \norm{V^\pi}_2 \sqrt{d_\delta}\norm{\hatwd}_2\eta_{\ealpha,\delta} \leq \cv\sqrt{d_\delta}\norm{\hatwd}_2\eta_{\ealpha,\delta} \,.\]
Denote by $\gamma =V^{\pi_1}\cdot( \sum_{i=1}^{{d_\delta}-1} \inner{\hatwd}{\lambda_i} \lambda_i)$, which is no greater than $\cv\sqrt{d_\delta}\norm{\hatwd}_2\eta_{\ealpha,\delta} $.

We have that 
\begin{align}
    \abs{\hatwd \cdot V^\pi - w\cdot V^\pi}& \leq \abs{\hatwd \cdot V^\pi - \frac{1}{1-\gamma} \hatwd\cdot V^\pi} + \abs{\frac{1}{1-\gamma} \hatwd\cdot V^\pi - w\cdot V^\pi}\nonumber\\
    &\leq \frac{\gamma \norm{\hatwd}_2 \cv}{1-\gamma} +\frac{1}{1-\gamma}\abs{\sum_{i=1}^{d_\delta-1} (\hatwd\cdot \lambda_i) \lambda_i\cdot V^\pi}\nonumber\\
    &\leq \frac{\gamma \norm{\hatwd}_2 \cv}{1-\gamma} +\frac{\cv\sqrt{d_\delta}\norm{\hatwd}_2\eta_{\ealpha,\delta}}{1-\gamma}\nonumber\\
    &= 2(\cv \norm{\hatwd}_2 +1)\cv\sqrt{d_\delta}\norm{\hatwd}_2\eta_{\ealpha,\delta},\label{eq:hatwwintermediate}
\end{align}
where the last equality holds when $\cv\sqrt{d_\delta}\norm{\hatwd}_2\eta_{\ealpha,\delta}\leq \frac{1}{2}$.

Now we show that $\norm {\hatwd}_2 \leq C\norm{w'}_2$ for some constant $C$.

Since $\hatwd$ is the minimum norm solution to $\hat A^{(\delta)}x= \be_1$, we will construct another solution to $\hat A^{(\delta)}x= \be_1$, denoted by $\hatwo$ in a simillar manner to the construction in Eq~\eqref{eq:creatw}, and show that $\norm{\hatwo}_2\leq C\norm{w'}$.


Let $\xi_1,\ldots,\xi_{d_\delta-1}$ be any orthonormal basis of $\Span(\hat A^{(\delta)}_{2:})$.

We construct a $\hatwo$ s.t. $\hat A^{(\delta)}\hatwo = \be_1$ by removing the component of $w'$  in $\Span(\hat A^{(\delta)}_{2:})$ and rescaling.

Specifically, let 
\begin{equation}
    \hatwo = \frac{ w'- \sum_{i=1}^{{d_\delta}-1} \inner{w'}{\xi_i} \xi_i}{1-\inner{V^{\pi_1}}{( \sum_{i=1}^{{d_\delta}-1} \inner{w'}{\xi_i} \xi_i)}}\,.
\end{equation}
Since $A^{(\delta)}w' = \be_1$, 
it directly follows that $\inner{\hat A_1}{\hatwo} = \inner{V^{\pi_1}}{\hatwo} = 1$ and that $\inner{\hat A_i}{\hatwo} = 0$ for $i=2,\ldots,d_\delta$, i.e., $\hat A^{(\delta)}\hatwo =\be_1$.

Since Claim~\ref{clm:spanangle} implies that $\theta( \Span(\hat A^{(\delta)}_{2:}), \Span(A^{(\delta)}_{2:})) \leq  \eta_{\ealpha,\delta}$, there exist unit vectors $\tilde \xi_1,\ldots,\tilde \xi_{d_\delta-1}\in \Span(A^{(\delta)}_{2:})$ such that $\theta(\xi_i,\tilde \xi_i)\leq \eta_{\ealpha,\delta}$.

As $w'$ has zero component in $\Span(A^{(\delta)}_{2:})$, $w'$ should have a small component in $\Span(\hat A^{(\delta)}_{2:})$.

In particular,
\begin{align*}
     \abs{\inner{w'}{\xi_i}}= \abs{\inner{w'}{\xi_i-\tilde \xi_i}} \leq \norm{w'}_2 \eta_{\ealpha,\delta}\,,
\end{align*}
which implies that 
\begin{align*}
    \norm{\sum_{i=1}^{{d_\delta}-1} \inner{w'}{\xi_i} \xi_i}_2\leq \sqrt{d_\delta} \norm{w'}_2 \eta_{\ealpha,\delta}\,.
\end{align*}
Hence
\begin{equation*}
    \abs{\inner{V^{\pi_1}}{( \sum_{i=1}^{{d_\delta}-1} \inner{w'}{\xi_i} \xi_i)}} \leq \cv\sqrt{d_\delta} \norm{w'}_2 \eta_{\ealpha,\delta} \,.
\end{equation*}
As a result, $\norm{\hatwo}_2 \leq \frac{3}{2}\norm{w'}_2$ when $ \cv\sqrt{d_\delta} \norm{w'}_2 \eta_{\ealpha,\delta} \leq \frac{1}{3}$, which is true when $\ealpha$ is small enough.

According to Lemma~\ref{lmm:V1}, $\ealpha\leq \frac{4k^2 \epsilon}{v^*}$, $\ealpha\rightarrow 0$ as $\epsilon\rightarrow 0$.

Thus, we have $\norm{\hatwd}_2\leq \norm{\hatwo}_2\leq \frac{3}{2} \norm{w'}_2$ and $\cv\sqrt{d_\delta}\norm{\hatwd}_2\eta_{\ealpha,\delta}\leq \frac{1}{2}$.

Combined with Eq~\eqref{eq:hatwwintermediate}, we get
\begin{align*}
    \abs{\hatwd \cdot V^\pi - w\cdot V^\pi} = \cO\left((\cv \norm{w'}_2 +1)\cv\sqrt{d_\delta}\norm{w'}_2\eta_{\ealpha,\delta}\right)\,.
\end{align*}
Since $\cv \norm{w'}_2 \geq \abs{\inner{V^{\pi_1}}{w'}} = 1$, by taking $\eta_{\ealpha,\delta} = \frac{4\calpha\cv^2 d_\delta\ealpha }{\delta^2}$ into the above equation, we have
\begin{align*}
    \abs{\hatwd \cdot V^\pi - w\cdot V^\pi}  =\cO\left( \frac{\calpha\cv^4 d_\delta^{\frac{3}{2}}\norm{w'}_2^2\ealpha }{\delta^2}\right)\,,
\end{align*}
which completes the proof.
\end{proof}
\section{Proof of Lemma~\ref{lmm:est-without-tau}}\label{app:est-without-tau}
\lmmwithouttau*
\begin{proof}

Given the output $(V^{\pi_1},\ldots, V^{\pi_{d}})$ of Algorithm~\ref{alg:policy-threshold-free}, we have $\rank(\Span(\{V^{\pi_1},\ldots, V^{\pi_{d_\delta}}\})) = d_\delta$.

For $i=d_\delta+1,\ldots,d$, let $\psi_{i}$ be the normalized vector of
$V^{\pi_i}$'s projection into $\Span(V^{\pi_1},\ldots,V^{\pi_{i-1}})^\perp$ with $\norm{\psi_i}_2 = 1$. 

Then we have that $\Span(V^{\pi_1},\ldots,V^{\pi_{i-1}},\psi_i) = \Span(V^{\pi_1},\ldots,V^{\pi_{i}})$ and that $\{\psi_i|i=d_\delta+1,\ldots,d\}$ are orthonormal.

For every policy $\pi$, the value vector can be represented as a linear combination of $\hat A_1,\ldots,\hat A_{d_\delta}, \psi_{d_\delta+1},\ldots, \psi_d$, i.e., there exists a unique $a = (a_1,\dots,a_d)\in \R^d$ s.t.
$V^\pi = \sum_{i=1}^{d_\delta} a_i \hat A_i +\sum_{i=d_\delta+1}^{d} a_i \psi_i$.

Since $\psi_i$ is orthogonal to $\psi_j$ for all $j\neq i$ and $\psi_i$ is are orthogonal to $\Span(\hat A_1,\ldots,\hat A_{d_\delta})$, we have $a_i = \inner{V^\pi}{\psi_i}$ for $i\geq d_\delta+1$.

This implies that 
\begin{align*}
    \abs{\inner{\hat w}{V^{\pi}} - \inner{\hatwd}{V^{\pi}}} \leq & \underbrace{\abs{\sum_{i=1}^{d_\delta}a_i\left(\inner{\hat w}{\hat A_i} - \inner{w^{(\delta)}}{\hat A_i}\right)}}_{(a)} + \underbrace{\abs{\sum_{i=d_\delta+1}^{d} a_i \inner{\hat w}{\psi_i}}}_{(b)} \\
    &+ \underbrace{\abs{\sum_{i=d_\delta+1}^{d} a_i \inner{\hatwd}{ \psi_i}}}_{(c)}\,.
\end{align*}
Since $\hatAd \hat w = \hatAd \hatwd =\be_1$, we have term $(a) = 0$.

We move on to bound term (c).

Note that the vectors $\{\psi_i|i=d_\delta+1,\ldots,d\}$ are orthogonal to $\Span(V^{\pi_1},\ldots, V^{\pi_{d_\delta}})$ and that together with $V^{\pi_1},\ldots, V^{\pi_{d_\delta}}$ they form a basis for $\Span(\{V^\pi|\pi\in \Pi\})$.

Thus, we can let $\basis_i$ in the proof of Lemma~\ref{lmm:gap-hatw-w} be $ \psi_{i+d_\delta}$.

In addition, all the properties of $\{\basis_i|i\in [d-d_\delta]\}$ also apply to $\{\psi_i|i=d_\delta+1,\ldots,d\}$ as well.

Hence, similarly to Eq~\eqref{eq:anorm},
\begin{equation*}
    \sqrt{\sum_{i={d_\delta+1}}^d a_i^2} \leq \sqrt{k}\delta\,.
\end{equation*}

%
Consequentially, we can bound term $(c)$ is by
$$(c) \leq \sqrt{k}\delta \norm{\hatwd}_2 =\frac{3}{2} \sqrt{k}\delta\norm{w'}_2$$
since $\norm{\hatwd}_2\leq \frac{3}{2}\norm{w'}_2$ when $ \cv\sqrt{d_\delta} \norm{w'}_2 \eta_{\ealpha,\delta} \leq \frac{1}{3}$ as discussed in the proof of Lemma~\ref{lmm:min_norm_hatw_delta}.

Now all is left is to bound term (b).

We cannot bound term (b) in the same way as that of term (c) because $\norm{\hat w}_2$ is not guaranteed to be bounded by $\norm{w'}_2$.

For $i=d_\delta+1,\ldots, d$, we define 
\[\epsilon_i:= \abs{\inner{\psi_i}{\hat A_i}} \,.\]
For any $i,j=d_\delta+1,\ldots, d$, $\psi_i$ is perpendicular to $V^{\pi_1}$, thus
$\abs{\inner{\psi_i}{\hat A_j}} = \abs{\inner{\psi_i}{\hat \alpha_{j-1}V^{\pi_1} - V^{\pi_j}}}= \abs{\inner{\psi_i}{V^{\pi_j}}}$.
Especially, we have
\[
\epsilon_i = \abs{\inner{\psi_i}{\hat A_i}} = \abs{\inner{\psi_i}{V^{\pi_i}}}\,.
\]
Let $\hat A_i^\parallel := \hat A_i- \sum_{j=d_\delta+1}^{d}\inner{\hat  A_i}{\psi_j}\psi_j$ denote $\hat A_i$'s projection into $\Span(\hat A_1,\ldots,\hat A_{d_\delta})$.

Since $\hat A_i$ has zero component in direction $\psi_j$ for $j>i$, we have $\hat A_i^\parallel = \hat A_i- \sum_{j=d_\delta+1}^{i}\inner{\hat  A_i}{\psi_j}\psi_j$.

Then, we have
\begin{align*}
    0= \inner{\hat w}{\hat A_i} = \hat w\cdot \hat A_i^\parallel + \hat w\cdot \sum_{j=d_\delta+1}^i\inner{\hat  A_i}{\psi_j}\psi_j= \hat w\cdot \hat A_i^\parallel -  \sum_{j=d_\delta+1}^i\inner{V^{\pi_i}}{\psi_j}\inner{\hat w}{\psi_j}\,,
\end{align*}
where the first equation holds due to $\hat A \hat w = \be_1$.

By rearranging terms, we have
\begin{align}
    \inner{V^{\pi_i}}{\psi_i} \inner{ \hat w}{\psi_i} = \hat w\cdot \hat A_i^\parallel -  \sum_{j=d_\delta+1}^{i-1}\inner{V^{\pi_i}}{\psi_j}\inner{\hat w}{\psi_j}\,.\label{eq:wui}
\end{align}
Recall that at iteration $j$ of Algorithm~\ref{alg:policy-threshold-free}, in line~\ref{alg-line:orthonormal-basis} we pick an orthonormal basis $\rho_1,\ldots, \rho_{k+1-j}$ of $\Span(V^{\pi_1},\ldots,V^{\pi_{j-1}})^\perp$. Since $\psi_j$ is in $ \Span(V^{\pi_1},\ldots,V^{\pi_{j-1}})^\perp$ according to the definition of $\psi_j$,
$\abs{\inner{V^{\pi_i}}{\psi_j}}$ is no greater then the norm of $V^{\pi_i}$'s projection into $\Span(V^{\pi_1},\ldots,V^{\pi_{j-1}})^\perp$.

Therefore, we have
\begin{align}
    &\abs{\inner{V^{\pi_i}}{\psi_j}} \leq \sqrt{k} \max_{l\in [k+1-j]} \abs{\inner{V^{\pi_i}}{\rho_l}}\stackrel{(d)}{\leq} 
    \sqrt{k} \max_{l\in [k+1-j]} \max(\abs{\inner{V^{\pi^{\rho_l}}}{\rho_l}},\abs{\inner{V^{\pi^{-\rho_l}}}{-\rho_l}})\nonumber\\
    \stackrel{(e)}{=}&
    \sqrt{k} \abs{\inner{V^{\pi_j}}{u_j}}\stackrel{(f)}{\leq} \sqrt{k} \abs{\inner{V^{\pi_j}}{\psi_j}}=\sqrt{k}\epsilon_j\,,\label{eq:psi-u}
\end{align}
where inequality (d) holds because $\pi^{\rho_l}$ is the optimal personalized policy with respect to the preference vector $\rho_l$, and Equation (e) holds due to the definition of $u_j$ (line~\ref{alg-line:largest-component} of Algorithm~\ref{alg:policy-threshold-free}). 
Inequality (f) holds since $\inner{V^{\pi_j}}{\psi_j}$ is the norm of $V^{\pi_j}$'s projection in $\Span(V^{\pi_1},\ldots,V^{\pi_{j-1}})^\perp$ and $u_j$ belongs to $\Span(V^{\pi_1},\ldots,V^{\pi_{j-1}})^\perp$.

By taking absolute value on both sides of Eq~\eqref{eq:wui}, we have
\begin{align}
    \epsilon_i \abs{\inner{ \hat w}{\psi_i}}
    = \abs{\hat w\cdot \hat A_i^\parallel -  \sum_{j=d_\delta+1}^{i-1}\inner{V^{\pi_i}}{\psi_j}\inner{\hat w}{\psi_j}}
    \leq \abs{\hat w\cdot \hat A_i^\parallel} +  \sqrt{k}\sum_{j=d_\delta+1}^{i-1}\epsilon_j\abs{ \inner{\hat w}{\psi_j}}\,.\label{eq:inducteps}
\end{align}
We can now bound $\abs{\hat w\cdot \hat A_i^\parallel}$ as follows.
\begin{align}
    \abs{\hat w\cdot \hat A_i^\parallel} &= \abs{\hatwd\cdot \hat A_i^\parallel}\label{eq:hatwwdelta}\\
    &= \abs{\hatwd\cdot (\hat A_i- \sum_{j=d_\delta+1}^{d}\inner{\hat  A_i}{\psi_j}\psi_j)} = \abs{\hatwd\cdot \hat A_i}\label{eq:applyminnorm}\\
    &\leq \abs{\hatwd\cdot A_i} + \abs{\hatwd\cdot (\hat A_i-A_i)}\nonumber\\
    &\leq \abs{w'\cdot A_i} + \abs{(\hatwd-w')\cdot A_i} + \abs{\hatwd\cdot (\hat A_i-A_i)}\nonumber\\
    &\leq 0+ (\calpha+1)\sup_{\pi} \abs{(\hatwd-w')\cdot V^\pi} +  \cv \norm{\hatwd}_2 \ealpha\nonumber\\
    &\leq C'\calpha\epsilon^{(\delta)}\nonumber\,,
\end{align}
for some constant $C'>0$.

Eq~\eqref{eq:hatwwdelta} holds because $\hatAd \hat w = \hatAd \hatwd = \be_1$ and $\hat A_i^\parallel$ belongs to $\Span(\hatAd)$. 
Eq~\eqref{eq:applyminnorm} holds because $\hatwd$ is the minimum norm solution to $\hatAd x = \be_1$, which implies that $\hatwd \cdot \psi_i = 0$. 
The last inequality follows by applying Lemma~\ref{lmm:gap-hatw-w}.

We will bound $\epsilon_i \abs{\inner{ \hat w}{\psi_i}}$ by induction on $i=d_\delta+1,\ldots,d$.

In the base case of $i=d_\delta+1$, 
\begin{align*}
    \epsilon_{d_\delta+1}  \abs{\inner{ \hat w}{\psi_{d_\delta+1}}}\leq \abs{\hat w\cdot \hat A_{d_\delta+1}^\parallel} \leq  C'\calpha\epsilon^{(\delta)}\,.
\end{align*}
Then, by induction through Eq~\eqref{eq:inducteps}, we have for $i = d_\delta+2,\ldots,d$,
\[\epsilon_i \abs{\inner{ \hat w}{\psi_i}} \leq (\sqrt{k} +1)^{i-d_\delta-1} C'\calpha\epsilon^{(\delta)}\,.\]
Similar to the deviation of Eq~\eqref{eq:psi-u}, we pick an orthonormal basis $\rho_1,\ldots,\rho_{k+1-i}$ of $\Span(V^{\pi_1},\ldots,V^{\pi_{i-1}})^\perp$ at line~\ref{alg-line:orthonormal-basis} of Algorithm~\ref{alg:policy-threshold-free}, then we have that, for any policy $\pi$,
\begin{align*}
    \abs{\inner{V^\pi}{\psi_i}} \leq \sqrt{k} \max_{l\in [k+1-i]} \abs{\inner{V^{\pi}}{\rho_l}}\leq   \sqrt{k} \abs{\inner{V^{\pi_{i}}}{u_i}} \leq \sqrt{k}\abs{\inner{V^{\pi_{i}}}{\psi_i}} = \sqrt{k}\epsilon_i\,.
\end{align*}
Then we have that term (b) is bounded by 
\begin{align*}
    (b) =& \abs{\sum_{i=d_\delta+1}^{d} \inner{V^\pi}{\psi_i} \inner{\hat w}{\psi_i}}
    \leq \sum_{i=d_\delta+1}^{d} \abs{\inner{V^\pi}{\psi_i}}\cdot \abs{{\inner{\hat w}{\psi_i}}}\\
    \leq& \sqrt{k}\sum_{i=d_\delta+1}^{d} \epsilon_i \abs{{\inner{\hat w}{\psi_i}}}
    \leq (\sqrt{k} +1)^{d-d_\delta} C'\calpha\epsilon^{(\delta)}\,.
\end{align*}
Hence we have that for any policy $\pi$,
\begin{equation*}
    \abs{\inner{\hat w}{V^{\pi}} - \inner{\hatwd}{V^{\pi}}} \leq (\sqrt{k} +1)^{d-d_\delta} C'\calpha\epsilon^{(\delta)} +\frac{3}{2} \sqrt{k}\delta\norm{w'}_2\,.
\end{equation*}
\end{proof}
\section{Proof of Theorem~\ref{thm:est-with-threshold}}\label{app:alg_with_threshold}
\thmestwithtau*

\begin{proof}[Proof of Theorem~\ref{thm:est-with-threshold}]
As shown in Lemma~\ref{lmm:V1}, we set $C_\alpha = 2k$ and have
$\ealpha = \frac{4k^2\epsilon}{v^*}$.
We have $\norm{w'} = \frac{\norm{w^*}}{\inner{w^*}{V^{\pi_1}}}$ and showed that $\inner{w^*}{V^{\pi_1}} \geq \frac{v^*}{2k}$ in the proof of Lemma \ref{lmm:V1}.

By applying Lemma~\ref{lmm:gap-hatw-w} and setting $\delta = \left(\frac{\cv^4k^5 \norm{w^*}_2\epsilon}{v^{*2}}\right)^{\frac{1}{3}}$, 
we have

\begin{align*}
    &v^*- \inner{w^*}{V^{\pi^{\hatwd}}} = \inner{w^*}{V^{\pi_1}} \left(\inner{w'}{V^{\pi^*}} - \inner{w'}{V^{\pi^{\hatwd}}}\right)
    \\
    \leq&  \inner{w^*}{V^{\pi_1}}\left(\inner{\hatwd}{V^{\pi^*}} - \inner{\hatwd}{V^{\pi^{\hatwd}}} +\cO(\sqrt{k}\norm{w'}_2 (\frac{\cv^4 k^5 \norm{w^*}_2\epsilon}{v^{*2}})^\frac{1}{3})\right)\\
    =&\cO(\sqrt{k}\norm{w^*}_2 (\frac{\cv^4 k^5 \norm{w^*}_2\epsilon}{v^{*2}})^\frac{1}{3})\,.
\end{align*}
\end{proof}



\section{Dependency on $\epsilon$}\label{app:eps-dependence}
In this section, we would like to discuss a potential way of improving the dependency on $\epsilon$ in Theorems~\ref{thm:est-without-tau} and \ref{thm:est-with-threshold}.

Consider a toy example where the returned three basis policies are $\pi_1$ with $V^{\pi_1} =(1,0,0)$, $\pi_2$ with $V^{\pi_2} =(1,1,1)$ and $\pi_3$ with $V^{\pi_3} = (1, \eta,-\eta)$ for some $\eta>0$ and $w^* = (1,w_2,w_3)$ for some $w_2,w_3$.

The estimated ratio $\hat \alpha_1$ lies in $ [1+w_2+w_3-\epsilon,1+w_2+w_3+\epsilon]$, and $\hat \alpha_2$ lies in $ [1+\eta w_2-\eta w_3 - \epsilon,1+\eta w_2-\eta w_3 + \epsilon]$. 
Suppose that $\hat \alpha_1 = 1+w_2+w_3+\epsilon$ and $\hat \alpha_2 = 1+\eta w_2-\eta w_3 + \epsilon$.

By solving
\begin{equation*}
    \begin{pmatrix}
    1 & 0 & 0\\
    w_2+w_3+\epsilon & -1 & -1\\
    \eta w_2-\eta w_3 + \epsilon & -\eta & \eta
    \end{pmatrix}
    \hat w = \begin{pmatrix}
        1\\ 0 \\ 0
    \end{pmatrix}\,
\end{equation*}
we can derive $\hat w_2 = w_2 +\frac{\epsilon}{2}(1+\frac{1}{\eta})$ and $\hat w_3 = w_3 +\frac{\epsilon}{2}(1-\frac{1}{\eta})$.

The quantity measuring sub-optimality we care about is $\sup_\pi \abs{\inner{\hat w }{V^\pi} - \inner{w^*}{V^\pi}}$, which is upper bounded by $\cv \norm{\hat w - w^*}_2$. 
But the $\ell_2$ distance between $\hat w$ and $w^*$ depends on the condition number of $\hat A$, which is large when $\eta$ is small. 
To obtain a non-vacuous upper bound  in Section~\ref{sec:policy-level}, we introduce another estimate $\hat w^{(\delta)}$ based on the truncated version of $\hat A$ and then upper bound $\norm{\hat w^{(\delta)} - w^*}_2$ and $\sup_\pi \abs{\inner{\hat w }{V^\pi} - \inner{\hat w^{(\delta)}}{V^\pi}}$ separately.

However, it is unclear if $\sup_\pi \abs{\inner{\hat w }{V^\pi} - \inner{w^*}{V^\pi}}$ depends on the condition number of $\hat A$. 
Due to the construction of Algorithm~\ref{alg:policy-threshold-free}, we can obtain some extra information about the set of all policy values.

First, since we find $\pi_2$ before $\pi_3$, $\eta$ must be no greater than $1$. 
According to the algorithm, $V^{\pi_2}$ is the optimal policy when the preference vector is $u_2$ (see line~\ref{alg-line: returnedu} of Algorithm~\ref{alg:policy-threshold-free} for the definition of $u_2$) and $V^{\pi_3}$ is the optimal policy when the preference vector is $u_3 = (0,1,-1)$.
Note that the angle between $u_2$ and $V^{\pi_2}$ is no greater than 45 degrees according to the definition of $u_2$. 
Then the values of all policies can only lie in the small box $B = \{x\in \R^3| \abs{u_2^\top x}\leq \abs{\inner{u_2}{V^{\pi_2}}}, \abs{u_3^\top x}\leq \abs{\inner{u_3}{V^{\pi_3}}}\}$. 
It is direct to check that for any $x\in B$, $\abs{\inner{\hat w}{x} - \inner{w^*}{x}} < (1+\sqrt{2})\epsilon$.
This example illustrates that even when the condition number of $\hat A$ is large, $\sup_\pi \abs{\inner{\hat w }{V^\pi} - \inner{w^*}{V^\pi}}$ can be small.
It is unclear if this holds in general.
Applying this additional information to upper bound $\sup_\pi \abs{\inner{\hat w }{V^\pi} - \inner{w^*}{V^\pi}}$ directly instead of through bounding $\cv \norm{\hat w - w^*}_2$ is a possible way of improving the term $\epsilon^\frac{1}{3}$.

\section{Description of C4 and Proof of Lemma~\ref{thm:cara}}\label{app:Caratheodory}
\begin{algorithm}[H]\caption{$\text{C4}
$: Compress  Convex Combination using Carathéodory's theorem}\label{alg:Caratheodory}
    \begin{algorithmic}[1]
        \STATE \textbf{input} a set of $k$-dimensional vectors $M \subset \R^k$ and a distribution $p\in \spl^M$
        \WHILE{$\abs{M}>k+1$}
            \STATE arbitrarily pick $k+2$ vectors $\mu_1,\ldots,\mu_{k+2}$ from $M$\label{alg-line:anyvecs}
            \STATE solve for $x\in \R^{k+2}$ s.t. $\sum_{i=1}^{k+2} x_i (\mu_i\append 1) = \bZero$, where $\mu \append 1$ denote the vector of appending $1$ to $\mu$
            \STATE  $i_0 \leftarrow \argmax_{i\in [k+2]} \frac{\abs{x_i}}{p(\mu_i)}$
            \STATE \textbf{if} $x_{i_0}<0$\label{alg-line:flipsign} \textbf{then} $x\leftarrow -x$
            \STATE $\gamma \leftarrow \frac{p(\mu_{i_0})}{x_{i_0}}$ and $\forall i\in [k+2]$, $p(\mu_i)\leftarrow p(\mu_i) - \gamma x_i$ 
            \STATE remove $\mu_i$ with $p(\mu_i) =0$ from $M$
        \ENDWHILE
        \STATE \textbf{output} $M$ and $p$
    \end{algorithmic}
\end{algorithm}
\cara*
\begin{proof}
The proof is similar to the proof of Carathéodory's theorem.
Given the vectors $\mu_1,\ldots,\mu_{k+2}$ picked in line~\ref{alg-line:anyvecs} of Algorithm~\ref{alg:Caratheodory} and their probability masses $p(\mu_i)$, we solve $x\in \R^{k+2}$ s.t. $\sum_{i=1}^{k+2} x_i (\mu_i\append 1) = \bZero$ in the algorithm. 

Note that there exists a non-zero solution of $x$ because $\{\mu_i\append 1|i\in[k+2]\}$ are linearly dependent.
Besides, $x$ satisfies $\sum_{i=1}^{d+2} x_i = 0$.

Therefore, $$\sum_{i=1}^{d+2} (p(\mu_i) - \gamma x_i) = \sum_{i=1}^{d+2} p(\mu_i).$$

For all $i$, if $x_i<0$, $p(\mu_i) - \gamma x_i \geq 0$ as $\gamma>0$; if $x_i>0$, then $\frac{x_i}{p(\mu_i)}\leq \frac{x_{i_0}}{p(\mu_{i_0})}=\frac{1}{\gamma}$ and thus $p(\mu_i) - \gamma x_i\geq 0$.

Hence, after one iteration, the updated $p$ is still a probability over $M$ (i.e., $p(\mu)\geq 0$ for all $\mu\in M$ and $\sum_{\mu\in M}p(M)=1$).
Besides, $\sum_{i=1}^{d+2} (p(\mu_i) - \gamma x_i) \mu_i = \sum_{i=1}^{d+2} p(\mu_i)\mu_i - \gamma \sum_{i=1}^{d+2} x_i \mu_i =  \sum_{i=1}^{d+2} p(\mu_i)\mu_i$.

Therefore, after one iteration, the expected value $\EEs{\mu\sim p}{\mu}$ is unchanged.

When we finally output $(M',q)$, we have that $q$ is a distribution over $M$ and that $\EEs{\mu\sim q}{\mu} =\EEs{\mu\sim p}{\mu}$.

Due to line~\ref{alg-line:flipsign} of the algorithm, we know that $x_{i_0}>0$.
Hence $p(\mu_{i_0}) - \gamma x_{i_0} = p(\mu_{i_0}) - \frac{p(\mu_{i_0})}{x_{i_0}} x_{i_0} =0$. 

We remove at least one vector $\mu_{i_0}$ from $M$ and we will run for at most $\abs{M}$ iterations.

Finally, solving $x$ takes $\cO(k^3)$ time and thus, Algorithm~\ref{thm:cara} takes $\cO(\abs{M}k^3)$ time in total.
\end{proof}

\section{Flow Decomposition Based Approach}\label{app:approach-flow-based}
We first introduce an algorithm based on the idea of flow decomposition.

For that, we construct a layer graph $G=((\Lsup{0}\cup\ldots \cup\Lsup{H+1}), E)$ with $H+2$ pairwise disjoint layers $\Lsup{0},\ldots,\Lsup{H+1}$, where every layer $t\leq H$ contains a set of vertices labeled by the (possibly duplicated) states reachable at the corresponding time step $t$, i.e., $\{s\in \cS \vert \Pr(S_t =s \vert S_0 = s_0) >0\}$.

Let us denote by $\xsup{t}_s$ the vertex in $\Lsup{t}$ labeled by state $s$.

Layer $\Lsup{H+1} =\{\xsup{H+1}_*\}$ contains only an artificial vertex, $\xsup{H+1}_*$, labeled by an artificial state $*$.

For $t=0,\ldots,H-1$, for every $\xsup{t}_s\in \Lsup{t}$, $\xsup{t+1}_{s'}\in \Lsup{t+1}$, we connect $\xsup{t}_s$ and $\xsup{t+1}_{s'}$  by an edge labeled by $(s,s')$ if $\pssp>0$.
Every vertex $\xsup{H}_s$ in layer $H$ is connected to $\xsup{H+1}_*$ by one edge, which is labeled by $(s, *)$.

We denote by $\Esup{t}$ the edges between $\Lsup{t}$ and $\Lsup{t+1}$.
Note that every trajectory $\tau = (s_0,s_1,\ldots,s_H)$ corresponds to a single path $(\xsup{0}_{s_0},\xsup{1}_{s_1},\ldots, \xsup{H}_{s_H},\xsup{H+1}_*)$ of length $H+2$ from $\xsup{0}_{s_0}$ to $\xsup{H+1}_*$.

This is a one-to-one mapping and in the following, we use path and trajectory interchangeably. 

The policy corresponds to a $(\xsup{0}_{s_0},\xsup{H+1}_*)$-flow with flow value $1$ in the graph $G$. 
In particular, the flow is defined as follows.

When the layer $t$ is clear from the context, we actually refer to vertex $\xsup{t}_s$ by saying vertex $s$.

For $t=0,\ldots,H-1$, for any edge $(s,s') \in \Esup{t}$, let $f:E\rightarrow \mathbb R^+$ be defined as
\begin{equation}
    f(s,s') = \sum_{\tj: (s^\tau_t,s^\tau_{t+1}) = (s,s')} q^\pi(\tau)\,,\label{eq:fdef}
\end{equation}
where $q^\pi(\tau)$ is the probability of $\tau$ being sampled.

For any edge $(s, *)\in \Esup{H}$, let
$f(s, *) = \sum_{(s',s)\in \Esup{H-1}} f(s',s)$. 
It is direct to check that the function $f$ is a well-defined flow.
We can therefore compute $f$ by dynamic programming.

For all $(s_0,s)\in \Esup{0}$, we have $f(s_0,s) = P(s|s_0,\pi(s_0))$ and for $(s,s')\in \Esup{t}$,
\begin{equation}
    f(s,s')= \pssp \sum_{s'': (s'',s)\in \Esup{t-1}} f(s'',s)\,.\label{eq:flowinduction}
\end{equation}
Now we are ready to present our algorithm by decomposing $f$ in Algorithm~\ref{alg:flow-decomposition}.

Each iteration in Algorithm~\ref{alg:flow-decomposition} will zero out at least one edge and thus, the algorithm will stop within $\abs{E}$ rounds.
\begin{algorithm}[H]\caption{Flow decomposition based approach}\label{alg:flow-decomposition}
    \begin{algorithmic}[1]
        \STATE initialize $Q\leftarrow \emptyset$.
        \STATE calculate $f(e)$ for all edge $e\in E$ by dynamic programming according to Eq~\eqref{eq:flowinduction}
        \WHILE{$\exists e\in E$ s.t. $f(e)>0$}
            \STATE pick a path $\tau=(s_0,s_1,\ldots,s_H,*)\in \Lsup{0}\times\Lsup{1}\times\ldots\times\Lsup{H+1}$
            s.t. $f(s_i,s_{i+1})>0\; \forall i\geq 0$
            \label{alg-line:path}
            \STATE  $f_\tau \leftarrow \min_{e\text{ in }\tau} f(e)$
            \STATE $Q\leftarrow Q\cup\{(\tau,f_\tau)\}$, $f(e) \leftarrow f(e)-f_\tau$ for $e$ in $\tau$\label{alg-line:flowupdate}
        \ENDWHILE
        \STATE output $Q$
    \end{algorithmic}
\end{algorithm}

\begin{restatable}{theorem}{thmflow}\label{thm:flow-decomposition}
    Algorithm~\ref{alg:flow-decomposition} outputs $Q$ satisfying that $\sum_{(\tau,f_\tau) \in Q} f_\tau \Phi(\tau) = V^{\pi}$ in time $\cO( H^2\abs{\cS}^2)$.
\end{restatable}

%
The core idea of the proof is that for any edge $(s,s')\in \Esup{t}$, the flow on $(s,s')$ captures the probability of $S_t=s\wedge S_{t+1}=s'$ and thus, the value of the policy $V^\pi$ is linear in $\{f(e)|e\in E\}$.

The output $Q$ has at most $\abs{E}$ number of weighted paths (trajectories).
We can further compress the representation through C4, which takes $O(\abs{Q}k^3)$ time.
\begin{corollary}
Executing Algorithm~\ref{alg:flow-decomposition} with the output $Q$ first and then running C4 over $\{(\Phi(\tau),f_\tau)|(\tau,f_\tau)\in Q\}$ returns a $(k+1)$-sized weighted trajectory representation in time $\cO(H^2\abs{\cS}^2 +k^3H\abs{\cS}^2 )$.
\end{corollary}
We remark that the running time of this flow decomposition approach  underperforms that of the expanding and compressing approach (see Theorem~\ref{thm:traj-compression}) whenever $|\cS|H + |\cS|k^3 = \omega(k^4 + k|\cS|)$. 

\section{Proof of Theorem~\ref{thm:flow-decomposition}}\label{app:flow-decomposition}
\thmflow*
\begin{proof}
\textbf{Correctness: }
The function $f$ defined by Eq~\eqref{eq:fdef} is a well-defined flow 
since for all $t=1,\ldots, H$, for all $s\in \Lsup{t}$, we have that
\begin{align*}
    &\sum_{s'\in \Lsup{t+1}:(s,s')\in \Esup{t}} f(s,s') 
    =
    \sum_{s'\in \Lsup{t+1}:(s,s')\in \Esup{t}}\sum_{\tj: (s^\tau_t,s^\tau_{t+1}) = (s,s')} q^\pi(\tau)
    = \sum_{\tj: s^\tau_t = s}q^\pi(\tau)\\
    = &\sum_{s''\in \Lsup{t-1}:(s'',s)\in \Esup{t-1}} f(s'',s)\,.
\end{align*}
In the following, 
we first show that Algorithm~\ref{alg:flow-decomposition} will terminate with $f(e)=0$ for all $e\in E$.

First, after each iteration, $f$ is still a feasible $(\xsup{0},\xsup{H+1})$-flow feasible flow with the total flow out-of $\xsup{0}$ reduced by $f_\tau$.
Besides, for edge $e$ with $f(e)>0$ at the beginning, we have $f(e)\geq 0$ throughout the algorithm because we never reduce $f(e)$ by an amount greater than $f(e)$.

Then, since $f$ is a $(\xsup{0},\xsup{H+1})$-flow and $f(e)\geq 0$ for all $e\in E$, we can always find a path $\tau$ in line~\ref{alg-line:path} of Algorithm~\ref{alg:flow-decomposition}.

Otherwise, the set of vertices reachable from $\xsup{0}$ through edges with positive flow does not contain $\xsup{H+1}$ and the flow out of this set equals the total flow out-of $\xsup{0}$. But since other vertices are not reachable, there is no flow out of this set, which is a contradiction.

In line~\ref{alg-line:flowupdate}, there exists at least one edge $e$ such that $f(e)>0$ is reduced to $0$. 
Hence, the algorithm will run for at most $\abs{E}$ iterations and terminate with $f(e)=0$ for all $e\in E$. 

Thus we have that for any $(s,s')\in \Esup{t}$, $f(s,s')=\sum_{(\tau,f_\tau)\in Q: (s^\tau_t,s^\tau_{t+1}) = (s,s')}f_\tau$.

Then we have
\begin{align*}
    V^\pi &= \sum_{\tj}q^\pi(\tau) \Phi(\tau)
    = \sum_{\tj}q^\pi(\tau) \left( \sum_{t=0}^{H-1} R(s^\tau_t,\pi(s^\tau_t))\right) \\
    & =  \sum_{t=0}^{H-1} \sum_{\tj}q^\pi(\tau) R(s^\tau_t,\pi(s^\tau_t))\\
    & =  \sum_{t=0}^{H-1} \sum_{(s,s')\in \Esup{t}} R(s,\pi(s)) \sum_{\tj: (s^\tau_t,s^\tau_{t+1}) =(s,s')}q^\pi(\tau) \\
    & =  \sum_{t=0}^{H-1} \sum_{(s,s')\in \Esup{t}}R(s,\pi(s)) f(s,s')\\
    & =  \sum_{t=0}^{H-1} \sum_{(s,s')\in \Esup{t}}R(s,\pi(s)) \sum_{(\tau,f_\tau)\in Q: (s^\tau_t,s^\tau_{t+1}) = (s,s')}f_\tau \\
    & = \sum_{(\tau,f_\tau)\in Q}f_\tau ( \sum_{t=0}^{H-1} R(s^\tau_t,\pi(s^\tau_t)))\\
    & =\sum_{(\tau,f_\tau)\in Q}f_\tau \Phi(\tau)\,.
\end{align*}
\textbf{Computational complexity: }
Solving $f$ takes $\cO(\abs{E})$ time.
The algorithm will run for $\cO(\abs{E})$ iterations and each iteration takes $\cO(H)$ time.
Since $\abs{E} = \cO(\abs{\cS}^2 H)$, the total running time of Algorithm~\ref{alg:flow-decomposition} is $\cO(\abs{\cS}^2 H^2)$.
C4 will take $\cO(k^3\abs{E})$ time.
\end{proof}
\section{Proof of Theorem~\ref{thm:traj-compression}}\label{app:traj-compression}
\trajcompress*
\begin{proof}
\textbf{Correctness:} 
C4 guarantees that $\abs{\Qsup{H}}\leq k+1$. 

We will prove $\sum_{\tau \in \Qsup{H}} \betasup{H}_\tau \Phi(\tau)=V^\pi$ by induction on $t=1,\ldots,H$.

Recall that for any trajectory $\tau$ of length $h$, $J(\tau)=\Phi(\tau) + V(s^\tau_h, H-h)$ was defined as the expected return of trajectories (of length $H$) with the prefix being $\tau$.

In addition, recall that $J_{\Qsup{t}} = \{J(\tau\append s)|\tau\in \Qsup{t}, s\in \cS\}$ and $p_{\Qsup{t},\betasup{t}}$ was defined by letting $p_{\Qsup{t},\betasup{t}}(\tau\append s) = \betasup{t}(\tau) P(s \vert s^\tau_t, \pi(s^\tau_t))$.

For the base case, we have that at $t=1$
\begin{align*}
    V^\pi &=R(s_0,\pi(s_0)) + \sum_{s\in \cS}  P(s|s_0,\pi(s_0))V^\pi(s, H-1)
    = \sum_{s\in \cS} P(s|s_0,\pi(s_0)) J((s_0)\append s)\\
    &= \sum_{s\in \cS} p_{\Qsup{0},\betasup{0}}((s_0)\append s) J((s_0)\append s) =\sum_{\tau\in \Qsup{1}} \betasup{1}(\tau)J(\tau)\,.
\end{align*}
Suppose that $V^\pi = \sum_{\tau'\in \Qsup{t}} \betasup{t}(\tau')J(\tau')$ holds at time $t$,
then we prove the statement holds at time $t+1$.
\begin{align*}
    V^\pi&=\sum_{\tau'\in \Qsup{t}} \betasup{t}(\tau')J(\tau')=\sum_{\tau'\in \Qsup{t}} \betasup{t}(\tau')(\Phi(\tau') +  V^\pi(s^{\tau'}_t, H-t)) \\
    &=\sum_{\tau'\in \Qsup{t}} \betasup{t}(\tau')\left(\Phi(\tau') + \sum_{s\in \cS}P(s|s^{\tau'}_t,\pi(s^{\tau'}_t))\left( R(s^{\tau'}_t, \pi(s^{\tau'}_t)) + V^\pi(s, H-t)\right)\right)\\
    &= \sum_{\tau'\in \Qsup{t}}\sum_{s\in \cS} \betasup{t}(\tau')P(s|s^{\tau'}_t,\pi(s^{\tau'}_t))\left(\Phi(\tau') +  R(s^{\tau'}_t, \pi(s^{\tau'}_t)) + V^\pi(s, H-t)\right)\\
    &= \sum_{\tau'\in \Qsup{t}}\sum_{s\in \cS} \betasup{t}(\tau')P(s|s^{\tau'}_t,\pi(s^{\tau'}_t))\left(\Phi(\tau'\append s)  + V^\pi(s, H-t)\right)\\
    &= \sum_{\tau'\in \Qsup{t}}\sum_{s\in \cS} p_{\Qsup{t},\betasup{t}}(\tau'\append s)J(\tau'\append s)\\
    &=\sum_{\tau\in \Qsup{t+1}} \betasup{t+1}(\tau)J(\tau)\,.
\end{align*}
By induction, the statement holds at $t=H$ by induction, i.e., $V^\pi= \sum_{\tau \in \Qsup{H}} \betasup{H}_\tau J(\tau) = \sum_{\tau \in \Qsup{H}} \betasup{H}_\tau \Phi(\tau)$.

\noindent\textbf{Computational complexity:}  Solving $V^\pi(s,h)$ for all $s\in \cS, h\in [H]$ takes time $\cO(kH\abs{\cS}^2)$. In each round, we need to call C4 for $\leq (k+1)\abs{S}$ vectors, which takes $\cO(k^4\abs{\cS})$ time. Thus, we need $\cO(k^4H\abs{\cS} + kH\abs{\cS}^2)$ time in total.
\end{proof}
\section{Example of maximizing individual objective}\label{app:eg-linear-dependent}
\begin{observation}
Assume there exist $k>2$ policies that together assemble $k$ linear independent value vectors. Consider the $k$ different policies $\pi^*_1,\dots,\pi^*_k$ that each $\pi^*_i$ maximizes the objective $i\in[k]$. Then, their respective value vectors $V^*_1,\dots,V^*_k$ are not  necessarily linearly independent. Moreover, if $V^*_1,\dots,V^*_k$ are linearly depended it does not mean that $k$ linearly independent value vectors do not exists.
\end{observation}
\begin{proof}
For simplicity, we show an example with a horizon of $H=1$ but the results could be extended to any $H\geq 1$. We will show an example where there are $4$ different value vectors, where $3$ of them are obtained by the $k=3$ policies that maximize the $3$ objectives and have linear dependence. 

Consider an MDP with a single state (also known as Multi-arm Bandit)   with $4$ actions with deterministic   reward vectors (which are also the expected values of the $4$ possible policies in this case):
\[
r(1)=
\begin{pmatrix}
8\\4\\2
\end{pmatrix},
\quad
r(2)=
\begin{pmatrix}
1\\2\\3
\end{pmatrix},
\quad
r(3)=
\begin{pmatrix}
85/12\\25/6\\35/12
\end{pmatrix}\approx
\begin{pmatrix}
7.083\\4.167\\2.9167
\end{pmatrix},
\quad
r(4)=
\begin{pmatrix}
1\\3\\2
\end{pmatrix}.
\]
Denote $\pi^a$ as the fixed policy that always selects action $a$. 
Clearly, policy $\pi^1$ maximizes the first objective, policy  $\pi^2$ the third, and policy  $\pi^3$ the second ($\pi^4$ do not maximize any objective).
However,
\begin{itemize}
    \item $r(3)$ linearly depends on $r(1)$ and $r(2)$ as 
\[
\frac{5}{6}r(1)+\frac{5}{12}r(2)=r(3).
\]
\item In addition, $r(4)$ is linearly independent in $r(1),r(2)$: Assume not. Then, there exists $\beta_1,\beta_2\in \mathbb{R}$ s.t.:
\[
\beta_1\cdot r(1)+\beta_2\cdot r(2) =
\begin{pmatrix}
8\beta_1 +\beta_2 \\4\beta_1 +2\beta_2 \\2\beta_1 +3\beta_2 
\end{pmatrix}
=\begin{pmatrix}
1\\3\\2
\end{pmatrix}= r(4).
\]
Hence, the first equations imply $\beta_2=1-8\beta_1$, and $4\beta_1+2-16\beta_1=3$, hence $\beta_1=-\frac{1}{12}$ and $\beta_2=\frac{5}{3}$. Assigning in the third equation yields $-\frac{1}{6}+5=2$ which is a contradiction.
\end{itemize}
\end{proof}
\chapter{Online Learning with Primary and Secondary Loss}\label{app:primary}
\section{Proof of Theorem~\ref{thm:lb2}}\label{apd:lb2}
\restatelb*
\begin{proof}
We construct an adversary with oblivious primary losses and adaptive secondary losses to prove the theorem. The adversary is inspired by the proof of the lower bound by~\cite{altschuler2018online}. We divide $T$ into $T^{1-\alpha}$ epochs evenly and the primary losses do not change within each epoch. Let $\lceil t \rceil_e =\min_{m:mT^\alpha\geq t} mT^\alpha$ denote the last time step of the epoch containing time step $t$. For each expert $h\in\cH$, at the beginning of each epoch, we toss a fair coin and let $\loss{1}_{t,h}=0$ if it is head and $\loss{1}_{t,h}=1$ if it is tail. It is well-known that there exists a universal constant $a$ such that $\EE{\min_{h\in\cH} Z_h} = E/2- a\sqrt{E\log(K)}$ where $Z_h\sim \Bin(E,1/2)$. Then we have
\begin{align*}
    \EE{\min_{h\in\cH}\sum_{t=1}^T\loss{1}_{t,h}}\leq \frac{T}{2}-aT^{\frac{1+\alpha}{2}}\sqrt{\log(K)}\:.
\end{align*}

For algorithm $\cA$, let $\cA_t$ denote the selected expert at time $t$. Then we construct adaptive secondary losses as follows. First, for the first $T^\alpha$ rounds, $\loss{2}_{t,h} = c+\delta $ for all $h\in\cH$. For $t\geq T^\alpha +1$,
\begin{align*}
\loss{2}_{t,h}=\begin{cases}c & \text{if } h=\cA_{t-1}=\ldots= \cA_{t-T^\alpha}\\c+\delta & \text{otherwise}\end{cases}\:.
\end{align*}
This indicates that the algorithm can obtain $\loss{2}_{t,\cA_t}=c$ only by selecting the expert she has consecutively selected in the last $T^\alpha$ rounds and that each switching leads to $\loss{2}_{t,\cA_t}= c+\delta$. Let $S$ denote the total number of switchings and $\tau_1, \ldots,\tau_S$ denote the time steps $\cA$ switches. For notation simplicity, let $\tau_{S+1}=T+1$. If $\EE{\Loss{1}_{T,\cA}}\geq {T}/{2}-aT^{\frac{1+\alpha}{2}}\sqrt{{\log(K)}}/2$, then $\EE{\reg{1}}\geq aT^{\frac{1+\alpha}{2}}\sqrt{{\log(K)}}/2$; otherwise,
\begin{align*}
\frac{T}{2} - \frac{1}{2}\EE{\sum_{s=1}^{S} \min\left(\tau_{s+1}-\tau_s,\lceil \tau_s\rceil_e+1-\tau_s\right)}\labelrel\leq{myeq:a} \EE{\Loss{1}_{T,\cA}} < \frac{T}{2} - aT^{\frac{1+\alpha}{2}}\sqrt{{\log(K)}}/2\:,
\end{align*}
where Eq.~\eqref{myeq:a} holds due to that the $s$-th switching helps to decrease the expected primary loss by at most $\min\left(\tau_{s+1}-\tau_s,\lceil \tau_s\rceil_e+1-\tau_s\right)/2$. Since the $s$-th switching increases the secondary loss to $c+\delta$ for at least $\min(\tau_{s+1}-1-\tau_{s}, T^\alpha)$ rounds, then we have
\begin{align*}
\EE{\Loss{2}_{T,\cA}} \geq& cT+\delta \EE{\sum_{s=1}^{S} \min(\tau_{s+1}-\tau_{s}, T^\alpha)}\\
\geq &cT+\delta \EE{\sum_{s=1}^{S} \min\left(\tau_{s+1}-\tau_s,\lceil \tau_s\rceil_e+1-\tau_s\right)}\\
> &cT+\delta aT^{\frac{1+\alpha}{2}}\sqrt{{\log(K)}},
\end{align*}
which indicates that $\EE{\reg{2}_c}=\Omega(T^{\frac{1+\alpha}{2}})$. Therefore, $\EE{\max\left(\reg{1}, \reg{2}_c\right)}\geq \max\left(\EE{\reg{1}},\EE{\reg{2}_c}\right)=\Omega(T^{\frac{1+\alpha}{2}})$.
\end{proof}

\section{Proof of Theorem~\ref{thm:subopt}}\label{apd:subopt}
\restatesubopt*
\begin{proof}
We divide $T$ into $T^{1-\beta}$ intervals evenly with $\beta=\frac{1+\alpha}{2}$ and construct $T^{1-\beta}+1$ worlds with $2$ experts. For computation simplicity, we let $\delta =1/2$. The adversary selects a random world $W$ at the beginning. She selects world $0$ with probability ${1}/{2}$ and world $w$ with probability ${1}/{2T^{1-\beta}}$ for all $w\in[T^{1-\beta}]$. 

In world $0$,  we design the losses of experts as shown in Table~\ref{tab:alglb}. During the $w$-th interval with $w\in[T^{1-\beta}]$ being odd, we set $(\loss{1}_{t,h_1},\loss{2}_{t,h_1},\loss{1}_{t,h_2},\loss{2}_{t,h_2}) = (0,c+\delta T^{\alpha - \beta},1,c-\delta T^{\alpha - \beta})$ for the first $T^\beta/2$ rounds and $(\loss{1}_{t,h_1},\loss{2}_{t,h_1},\loss{1}_{t,h_2},\loss{1}_{t,h_2}) = (1,c,0,c)$ for the second $T^\beta/2$ rounds. For $w$ being even, we swap the losses of the two experts, i.e., $(\loss{1}_{t,h_1},\loss{2}_{t,h_2},\loss{1}_{t,h_2},\loss{2}_{t,h_2}) = (1,c-\delta T^{\alpha - \beta},0,c+\delta T^{\alpha - \beta})$ for the first $T^\beta/2$ rounds and $(\loss{2}_{t,h_1},\loss{2}_{t,h_1},\loss{1}_{t,h_2},\loss{2}_{t,h_2}) = (0,c,1,c)$ for the second $T^\beta/2$ rounds.

The intuition of constructing world $w\in[T^{1-\beta}]$ is described as below. In world $w$, the secondary loss is the same as that in world $0$. The primary losses of each expert $h\in\cH$ in the first $w-1$ intervals are an approximately random permutation of that in world $0$. Therefore, any algorithm will attain almost the same expected primary loss (around $(w-1)T^\beta/2$) in the first $w-1$ intervals of world $w$. The primary losses during the first $T^\beta/2$ rounds in the $w$-th interval are the same as those in world $0$. Therefore, the cumulative losses from the beginning to any time $t$ in the first half of the $w$-th interval are almost the same in world $0$ and world $w$, which makes the algorithm only dependent on the cumulative losses behave nearly the same during the first half of the $w$-th interval in two worlds. For $t=(w-1/2)T^\beta+1,\ldots, T$, we set $\loss{1}_{t,h}=1$ for all $h\in\cH$, which indicates that any algorithms are unable to improve their primary loss after $t=(w-1/2)T^\beta+1$. To prove the theorem, we show that if the algorithm selects expert $h$ with loss $(1,c-\delta T^{\alpha - \beta})$ during the first half of the $w$-th interval with large fraction, then $\reg{1}$ will be large in world $w$; otherwise, $\reg{2}_c$ will be large in world $0$. 

More specifically, for the first $w-1$ intervals in world $w$, we need to make the cumulative primary losses to be $(w-1)T^\beta/2$ with high probability. Let $t' =(w-1)T^\beta-2\sqrt{(w-1)T^\beta\log(T)}$. For $t=1,\ldots, t'$, $\loss{1}_{t,h}$ are i.i.d. samples from $\Ber({1}/{2})$ for all $h\in\cH$. We denote by $E_h^{(w)}$ the event of $\abs{\sum_{t=1}^{t'} (\loss{1}_{t,h} -1/2)}\leq \sqrt{(w-1)T^\beta\log(T)}$ and denote by $E$ the event of $\cap_{h\in\cH,w\in[T^{1-\beta}]} E_h^{(w)}$. If $E_{h_1}^{(w)}\cap E_{h_2}^{(w)}$ holds, we compensate the cumulative primary losses by assigning $\loss{1}_{t,h}=1$ for $(w-1)T^\beta/2 - \sum_{t=1}^{t'} \loss{1}_{t,h}$ rounds and $\loss{1}_{t,h}=0$ for the remaining rounds during $t=t'+1,\ldots, (w-1)T^\beta$ for all $h\in\cH$ such that the cumulative primary losses in the first $w-1$ intervals for both experts are $(w-1)T^\beta/2$ ; otherwise, we set $\loss{1}_{t,h} = 1$ for all $h\in\cH$ during $t=t'+1,\ldots, (w-1)T^\beta$. Hence, if $E_{h_1}^{(w)}\cap E_{h_2}^{(w)}$, the cumulative losses $\Loss{1}_{(w-1)T^\beta,h} = (w-1)T^\beta/2$ for all $h\in\cH$. To make it clearer, the values of the secondary losses in world $w$ for an even $w$ if $E_{h_1}^{(w)}\cap E_{h_2}^{(w)}$ holds are illustrated in Table~\ref{tab:alglb2}.
 
Let $q_w=2\sum_{t=(w-1)T^\beta+1}^{(w-1/2)T^\beta}\EE{\1(\loss{1}_{t,\cA_t}=0)}/T^\beta$ denote the expected fraction of selecting the expert with losses $(0,c+\delta T^{\alpha-\beta})$ in $w$-th interval in world $0$ as well as that in world $w$ when $E$ holds. We denote by $\reg{1,w}=\Loss{1,w}_{T,\cA}- \Loss{1,w}_{T,h_0}$ and $\reg{1,w}_{t'}=\Loss{1,w}_{t',\cA}- \Loss{1,w}_{t',h_0}$ with $h_0=\argmin_{h\in\cH} \Loss{1,w}_{T,h}$ being the best expert in hindsight the regret with respect to the primary loss for all times and the regret incurred during $t=1,\ldots,t'$ in world $w$. We denote by $\reg{2,w}_c$ the regret to $cT$ with respect to the secondary loss in world $w$. Then we have  
\begin{align*}
\EEc{\reg{2,W}_c}{W=0} = \sum_{w\in[T^{1-\beta}]} \frac{\delta(2q_w-1)T^\alpha}{2}\:,
\end{align*}
and for all $w\in[T^{1-\beta}]$,
\begin{align*}
\EEc{\reg{1,W}}{W=w,E} \geq& (1-q_w)\frac{T^\beta}{2} +\EEc{\reg{1,w}_{t'}}{W=w,E}-\left((w-1)T^\beta-t'\right) \\
\geq& (1-q_w)\frac{T^\beta}{2}-2\sqrt{(w-1)T^\beta\log(T)}\: .
\end{align*}
Due to Hoeffding's inequality and union bound, we have $\PP{\neg E_{h}^{(w)}}\leq \frac{2}{T^2}$ for all $h\in\cH$ and $w\in[T^{1-\beta}]$ and $\PP{\neg E}\leq \frac{4}{T^{1+\beta}}$. Let $Q=\frac{{\sum_{w=1}^{T^{1-\beta}} q_w}}{T^{1-\beta}}$ denote the average of $q_w$ over all $w\in[T^{1-\beta}]$. By taking expectation over the adversary, we have

\begin{align}
&\EE{\max\left(\reg{1}, \reg{2}_c\right)}\nonumber\\
\geq & \PP{E}\cdot\EEc{\max\left(\reg{1}, \reg{2}_c\right)}{E}\nonumber\\
\geq &\left(1-\frac{4}{T^{1+\beta}}\right)\left(\frac{1}{2T^{1-\beta}}\sum_{w=1}^{T^{1-\beta}} \EEc{\reg{1,W}}{W=w,E}+ \frac{1}{2}\EEc{\reg{2,W}_c}{W=0,E}\right)\nonumber\\
\geq& \frac{1}{2}\left(\frac{1}{2T^{1-\beta}}\left(\sum_{w} {(1-q_w)}\frac{T^\beta}{2} -2\sum_{w=1}^{T^{1-\beta}}\sqrt{(w-1)T^\beta\log(T)}\right)+ \frac{\delta}{4}{\sum_{w=1}^{T^{1-\beta}}(2q_w-1)}T^\alpha\right) \nonumber\\
\geq& \frac{1}{8}(1-Q)T^\beta  -\sqrt{T\log(T)} + \frac{\delta}{8}(2Q-1)T^{1-\beta+\alpha}\nonumber\\
\geq& \frac{1}{16}T^{\frac{1+\alpha}{2}}-\sqrt{T\log(T)},\label{eq:alglb}
\end{align}
where Eq.~\eqref{eq:alglb} holds by setting $\beta=\frac{1+\alpha}{2}$ and $\delta =1/2$.
\end{proof}

\begin{table}[H]
\caption{The losses in world $0$.}\label{tab:alglb}
\centering
\begin{tabular}{c|c|c|c|c|c|c|c}
\toprule
\multicolumn{2}{c|}{experts\textbackslash time} &$T^\beta/2$ &$T^\beta/2$ &$T^\beta/2$ &$T^\beta/2$ &$T^\beta/2$ & $\ldots$ \\
\cmidrule{1-8}
 \multirow{2}{*}{$h_1$} &$\loss{1}$& $0$ & $1$ & $1$ & $0$ & $0$ & $\ldots$\\
\cmidrule{2-8}
 & $\loss{2}$ & $c+\delta T^{\alpha-\beta}$ & $c$ &$c-\delta T^{\alpha-\beta}$ & $c$ & $c+\delta T^{\alpha-\beta}$& $\ldots$\\
\cmidrule{1-8}
 \multirow{2}{*}{$h_2$} &$\loss{1}$& $1$ & $0$ & $0$ & $1$ & $1$ & $\ldots$\\
\cmidrule{2-8}
& $\loss{2}$ & $c-\delta T^{\alpha-\beta}$ & $c$ &$c+\delta T^{\alpha-\beta}$ & $c$ & $c-\delta T^{\alpha-\beta}$& $\ldots$\\
\bottomrule
\end{tabular}
\end{table}

\begin{table}[H]
\caption{The primary losses in world $w$ (which is even) if $E_{h_1}^{(w)}\cap E_{h_2}^{(w)}$ holds.}\label{tab:alglb2}
\centering
\begin{tabular}{c|c|c|c|c|c|c}
\toprule
\multicolumn{2}{c|}{experts\textbackslash time} &$t'$ &$(w-1)T^\beta- t'$ &$T^\beta/2$ &$T^\beta/2$ &$T-T^\beta$ \\
\cmidrule{1-7}
 {$h_1$} &$\loss{1}$& i.i.d. from $\Ber({1}/{2})$ & compensate & $1$ & $1$ & $1$\\
\cmidrule{1-7}
{$h_2$} &$\loss{1}$& i.i.d. from $\Ber({1}/{2})$ & compensate & $0$ & $1$ & $1$\\
\bottomrule
\end{tabular}
\end{table}
\section{Proof of Theorem~\ref{thm:linK}}\label{apd:linK}
\restatelinK*
\begin{proof}
The idea is to construct an example in which the best expert with respect to the primary loss is deactivated sequentially while incurring an extra $\Theta(T^\alpha)$ secondary loss. In the example, we set $\cH=[K]$. Let $T_{k} = T^{\alpha+\frac{(k-1)(1-\alpha)}{K-1}}$ for $k\in[K]$ and $T_{0}=0$. For each expert $k\in\cH$, we set $(\loss{1}_{t,k},\loss{2}_{t,k}) = (1,c)$ for $t\leq T_{{k-1}}$ and $(\loss{1}_{t,k},\loss{2}_{t,k}) = (0,c+\frac{\delta T^\alpha}{T_{k}-T_{{k-1}}})$ for $t\geq T_{{k-1}}+1$. Then expert $k$ will be deactivate at time $t=T_{k}$. For any algorithm with $\sreg{1}_{k}=o(T_{k})$ for all $k \in \cH$, expert $k$ should be selected for $T_{k}-2T_{{k-1}}-o(T_{k})$ rounds during $t=T_{{k-1}}+1,\ldots,T_{k}$. Therefore, we have $\reg{2}_c \geq \sum_{k\in[K]}\frac{\delta T^\alpha}{T_{k}-T_{{k-1}}} (T_{k}-2T_{{k-1}}-o(T_{k}))= \Omega(KT^\alpha)$.
\end{proof}

\section{Proof of Theorem~\ref{thm:ub}}\label{apd:ub}
\restateub*

\begin{proof}
Let $\tLoss{1,h^*}_h = \sum_{m=1}^{T^{1-\alpha}}\ts(e_m)\tloss{1}_{e_m,h}$ and $\tLoss{1,h^*}_\cA = \sum_{m=1}^{T^{1-\alpha}}\ts(e_m)\tloss{1}_{e_m,\cA}$ denote the cumulative pseudo primary losses of expert $h$ and algorithm $\cA$ during the time when $h^*$ is active. First, since we update $w_{m+1,h}^{h^*} = w_{m,h}^{h^*}\eta^{\ts(e_m)(\tloss{1}_{e_m,h}-\eta \tloss{1}_{e_m,\cA})+1}\leq w_{m,h}^{h^*}$ with $\eta\in[1/\sqrt{2},1]$ and experts will not be reactivated between (not including) $t_n$ and $t_{n+1}$, the probability of following the first rule on Line~7 in Algorithm~\ref{alg:orc2}, which is $\frac{w_{m+1,h_{m}}}{w_{m,h_{m}}}$, is legal. Then we show that at each epoch $m$, the probability of getting $h_m = h$ is $\PP{h_m = h}=p_{m,h}$. The proof follows Lemma~1 by~\citep{geulen2010regret}. For an reactivating epoch $m\in\{(t_n-1)/T^\alpha+1\}_{n=0}^N$, $h_m$ is drawn from $p_m$ and thus, $\PP{h_m = h}=p_{m,h}$ holds. For other epochs $m\notin\{(t_n-1)/T^\alpha+1\}_{n=0}^N$, we prove it by induction. Assume that $\PP{h_{m-1} = h}=p_{m-1,h}$, then
\begin{align*}
\PP{h_m = h} &= \PP{h_{m-1} = h}\frac{w_{m,h}}{w_{m-1,h}} + p_{m,h} \sum_{h'\in \cH} \PP{h_{m-1} = h'}\left(1-\frac{w_{m,h'}}{w_{m-1,h'}}\right)\\
& = \frac{w_{m-1,h}}{W_{m-1}}\cdot\frac{w_{m,h}}{w_{m-1,h}} +\frac{w_{m,h}}{W_m}\left(1-  \sum_{h'\in \cH}\frac{w_{m-1,h'}}{W_{m-1}}\cdot \frac{w_{m,h'}}{w_{m-1,h'}}\right)\\
& =p_{m,h}\:.
\end{align*}
To prove the upper bound on sleeping regrets, we follow Claim~12 by \cite{blum2007external} to show that $\sum_{h,h^*} w_{m,h}^{h^*}\leq K \eta^{m-1}$ for all $m\in [T^{1-\alpha}]$.

First, we have
\begin{align}
W_m \tloss{1}_{e_m,\cA} = W_m \sum_{h\in\cH}p_{m,h} \tloss{1}_{e_m,h} = \sum_{h\in\cH}w_{m,h} \tloss{1}_{e_m,h}= \sum_{h\in\cH}\sum_{h^*\in\cH}\ts(e_m)w_{m,h}^{h^*} \tloss{1}_{e_m,h}\:. \label{eq:intm}
\end{align}
Then according to the definition of $w_{m,h}^{h^*}$, we have
\begin{align*}
&\sum_{h\in\cH,h^*\in\cH} w_{m+1,h}^{h^*}\\=&\sum_{h\in\cH,h^*\in\cH} w_{m,h}^{h^*}\eta^{\ts(e_m)(\tloss{1}_{e_m,h}-\eta \tloss{1}_{e_m,\cA})+1}\\
\leq& \eta\left(\sum_{h\in\cH,h^*\in\cH} w_{m,h}^{h^*} \left(1-(1-\eta)\ts(e_m)\tloss{1}_{e_m,h}\right)\left(1+(1-\eta)\ts(e_m)\tloss{1}_{e_m,\cA}\right)\right)\\
\leq& \eta\left(\sum_{h\in\cH,h^*\in\cH} w_{m,h}^{h^*} - (1-\eta)\left(\sum_{h\in\cH,h^*\in\cH} w_{m,h}^{h^*}\ts(e_m)\tloss{1}_{e_m,h} - W_m\tloss{1}_{e_m,\cA}\right)\right)\\\nonumber
=&\eta\sum_{h\in\cH,h^*\in\cH} w_{m,h}^{h^*}\:,\label{eq:sr}
\end{align*}
where the last inequality adopts Eq.~\eqref{eq:intm}. Combined with $w_{1,h}^{h^*} = \frac{1}{K}$ for all $h\in\cH,h^*\in\cH$, we have $\sum_{h,h^*} w_{m+1,h}^{h^*}\leq K \eta^{m}$. Since $w_{m+1,h}^{h^*} = w_{1,h}^{h^*}\eta^{\sum_{i=1}^{m} \ts(e_i)\tloss{1}_{e_i,h}-\eta\sum_{i=1}^{m} \ts(e_i)\tloss{1}_{e_i,\cA}+m} \leq K \eta^{m}$, we have
\begin{align*}
    \tLoss{1,h^*}_\cA -\tLoss{1,h^*}_h\leq \frac{(1-\eta)\tLoss{1,h^*}_h +\frac{2\log(K)}{\log(1/\eta)}}{\eta}\,.
\end{align*}
By setting $\eta = 1-\sqrt{2\log(K)/T^{1-\alpha}}$, we have $\sreg{1}(h^*) \leq 2\sqrt{\log(K)T^{1+\alpha}} +2T_{h^*}\sqrt{\log(K)T^{\alpha-1}}$.

To derive $\reg{2}_c$, we bound the number of switching times. We denote by $S_n$ the number of epochs in which some experts are deactivated during $(t_n-1)/T^\alpha+1 < m< (t_{n+1}-1)/T^\alpha+1$ and by $\tau_1,\ldots,\tau_{S_n}$ the deactivating epochs, i.e., $\Delta \cH_{\tau_i}\neq \emptyset$ for $i\in [S_n]$. We denote by $\alpha_m$ the probability of following the second rule at line~7 in Algorithm~\ref{alg:orc2}, which is getting $h_m$ from $p_m$. Then we have
\begin{align*}
\alpha_m = \sum_{h\in\cH} \PP{h_{m-1} = h}\left(1-\frac{w_{m,h}}{w_{m-1,h}}\right)=\sum_{h\in\cH} \frac{w_{m-1,h}}{W_{m-1}}\left(1-\frac{w_{m,h}}{w_{m-1,h}}\right) = \frac{W_{m-1}-W_{m}}{W_{m-1}}\:.
\end{align*}
Since $W_{\tau_{i+1}}/W_{\tau_i+1}\geq \eta^{2(\tau_{i+1}-\tau_i-1)}$, we have
\begin{align*}
    \sum_{m=\tau_i+1}^{\tau_{i+1}} \alpha_m &\leq 1-\sum_{m=\tau_i+2}^{\tau_{i+1}} \log(1-\alpha_m) = 1-\sum_{m=\tau_i+2}^{\tau_{i+1}} \log\left(\frac{W_m}{W_{m-1}}\right) = 1+\log\left(\frac{W_{\tau_{i+1}}}{W_{\tau_i+1}}\right)\\
    &\leq 1+2\sqrt{2}(\tau_{i+1}-\tau_i-1)(1-\eta) = 1+4(\tau_{i+1}-\tau_i-1)\sqrt{\log(K)/T^{1-\alpha}}\,.
\end{align*}
Therefore, during time $(t_n-1)/T^\alpha \leq m< (t_{n+1}-1)/T^\alpha$, the algorithm will switch at most $K+4(t_{n+1}-t_n)\sqrt{\log(K)/T^{1-\alpha}}+1$ times in expectation,
which results in $\reg{2}_c \leq 4\delta\sqrt{\log(K)T^{1+\alpha}}+\delta N(K+1)T^{\alpha} = O(\sqrt{\log(K)T^{1+\alpha}}+NKT^{\alpha})$
\end{proof}

\chapter{Robust Learning under Clean-Label Attack}\label{app:clean-label}
\section{Proof of results in Section~\ref{sec:examples}}\label{appx:examples}
\subsection{Hypothesis class with VC dimension $d=1$}\label{appx:vcd1}
For any $f\in \cH$, let $\max^{\leq^\cH_f}_{x\in S_0}x$ denote the maximal element w.r.t. the partial ordering $\leq^\cH_f$ in any non-empty ordered finite set $S_0$, i.e., $\forall x'\in S_0, x'\leq^\cH_f \left(\max^{\leq^\cH_f}_{x\in S_0}x\right)$. Then for any arbitratrily chosen but fixed $f\in \cH$, the algorithm 
(originally proposed by \citealp{ben20152}) is described as follows.
\begin{algorithm}[H]\caption{Robust algorithm for $\cH$ with VC dimension $d=1$}\label{alg:vc1}
  \begin{algorithmic}[1]
    \STATE \textbf{input}: data $S$
    \STATE If every $(x,y) \in S$ has $y=f(x)$, \textbf{return} $\hat{h}=f$
    \STATE Let $x_m = \max^{\leq^\cH_f}_{(x,y)\in S, y\neq f(x)} x$
    \STATE $\hat{h}(x)=1-f(x)$ for $x\leq^\cH_f x_m$ and $\hat{h}(x)=f(x)$ otherwise
    \STATE \textbf{return} $\hat{h}$
  \end{algorithmic}
\end{algorithm}
By Lemma~5 of~\cite{ben20152}, $\leq^\cH_f$ for $d=1$ is a tree ordering. Thus, all points labeled differently by $f$ and $h^*$ should lie on one path, i.e., for every $x,x'\in\cX$, if $h^*(x)\neq f(x)$ and $h^*(x')\neq f(x')$, then $x\po x'$ or $x' \po x$. Due to this structure property of hypothesis class with VC dimenison $1$, adding clean-label attacking points can only narrow down the error region of Algorithm~\ref{alg:vc1}.
\begin{thm}
  For any $\cH$ with VC dimension $d=1$, Algorithm~\ref{alg:vc1} can $(\epsilon,\delta)$-robustly learn $\cH$ using $m$ samples, where 
  \[
    m = \left\lceil \frac{2\ln(1/\delta)}{\epsilon} \right\rceil\,.
    \]
\end{thm}

\begin{proof}
  First, we prove that $X=\{x\in \cX|h^*(x)\neq f(x)\}$ is totally ordered by $\leq^\cH_f$. That is, for every $x,x'\in\cX$, if $h^*(x)\neq f(x)$ and $h^*(x')\neq f(x')$, then $x\po x'$ or $x' \po x$. If it is not true, then there exists $h_1,h_2\in\cH$ such that $h_1(x)\neq f(x), h_1(x')= f(x')$ and $h_2(x')\neq f(x'), h_2(x)= f(x)$. Then, $\{f,h^*,h_1,h_2\}$ shatters $\{x,x'\}$, which contradicts that $d=1$. Therefore the finite set $\{x|(x,y)\in S, h^*(x)\neq f(x)\}\subseteq \{x\in \cX|h^*(x)\neq f(x)\}$ is also an ordered set. If the set is not empty, $x_m$ in the algorithm is well-defined.
  
  Now note that either $\hat{h}=f$ 
  or else $x_m$ is defined and then every 
  $x$ with $\hat{h}(x) \neq f(x)$ has 
  $x \leq^\cH_f x_m$, which implies 
  $h^*(x) \neq f(x)$ as well (since $h^*(x_m) \neq f(x_m)$).
  In particular, if every $(x,y) \in S_\trn$ 
  has $y=f(x)$ then 
  the attackable region $\ATK(h^*, S_\trn,\A,\Adv) \subseteq X$.
  Otherwise let 
  $x_{S_\trn} = \max^{\leq^\cH_f}_{(x,y)\in S_\trn, y\neq f(x)} x$, 
  the maximal element in $\{x|(x,y)\in S_\trn, h^*(x)\neq f(x)\}$, 
  we would have that 
  $\ATK(h^*, S_\trn,\A,\Adv) \subseteq \{x|x_{S_\trn}\po x, h^*(x)\neq f(x)\}$.
  
  In particular, if 
  $\cP_\cD(X)\leq \epsilon$ 
  the above facts imply 
  $\atk_\cD(h^*, S_\trn,\A) \leq \epsilon$.
  Otherwise if $\cP_\cD(X) > \epsilon$, 
  then we let $X_\epsilon \subseteq X$ be any  minimal set such that $\cP_\cD(X_\epsilon)\geq\frac{\epsilon}{2}$ and for every $x'\in X\setminus X_\epsilon$ and every $x\in X_\epsilon$, $x' \po x$. If $\cP_\cD(X_\epsilon)\geq \epsilon$, there exists an element $x\in X_\epsilon$ with probability mass at least $\frac{\epsilon}{2}$. When $m \geq \frac{2\ln(1/\delta)}{\epsilon}$, with probability at least $1-\delta$, $x$ is in $S_\trn$ and therefore 
  $\ATK(h^*, S_\trn,\A,\Adv) \subseteq X_\epsilon \setminus \{x\}$, so that 
  $\atk_\cD(h^*, S_\trn,\A) \leq \frac{\epsilon}{2}$. Otherwise, if $\frac{\epsilon}{2} \leq \cP_\cD(X_\epsilon)< \epsilon$, then as long as $S_\trn$ contains at least one example from $X_\epsilon$, then $\ATK(h^*, S_\trn,\A,\Adv) \subseteq X_\epsilon$, so that $\atk_\cD(h^*, S_\trn,\A) \leq \cP_\cD(X_\epsilon)<\epsilon$. Since $S_\trn$ contains an example from $X_\epsilon$ with probability at least  $1 - (1-\frac{\epsilon}{2})^m$, when $m \geq \frac{2\ln(1/\delta)}{\epsilon}$ we have that with probability at least $1-\delta$, $\atk_\cD(h^*, S_\trn,\A) \leq \epsilon$.
\end{proof}


\vspace{-20pt}
\subsection{Union of intervals}\label{appx:unionintvls}
\begin{thm}
  The algorithm described in Section~\ref{sec:examples} can $(\epsilon,\delta)$-robustly learn union of intervals $\cH_k$ using $m$ samples, where
  \[
    m= O\!\left(\frac{1}{\epsilon}(k\log(1/\epsilon)+\log(1/\delta))\right)\,.
    \]
\end{thm}
\begin{proof}
  We denote the target function by $h^* = \ind{\cup_{i=1}^{k^*}(a_{2i-1},a_{2i})}$ with $0=a_0\leq a_1\leq \ldots\leq a_{2k^*+1}=1$ and $a_{2i-1}\neq a_{2i},\forall i\in [k^*]$ for some $0 \leq k^*\leq k$. In the following, we will construct two classifiers consistent with the training set and then prove that the attackable rate of our algorithm is upper bounded by the sum of the error rates of these two classifiers. For any $i\in[k^*]$, we define $c_i^+$ as the minimum consistent positive interval within $(a_{2i-1},a_{2i})$, i.e.,
  \begin{align*}
      c_i^+=
      \left\{
      \begin{array}{cl}
       \left[\min\limits_{x\in (a_{2i-1},a_{2i}):(x,y)\in S_\trn}x,\max\limits_{x\in (a_{2i-1},a_{2i}):(x,y)\in S_\trn}x\right] 
       & \quad\text{if }S_\trn\cap (a_{2i-1},a_{2i})\times \cY\neq \emptyset\,,\\
     \emptyset      & \quad\text{otherwise}\,.
      \end{array}
      \right.
  \end{align*}
  Similarly, for $i=0,\ldots,k^*$, we define $c_i^-$ as the minimum consistent negative interval within $[a_{2i},a_{2i+1}]$. Since $x=0$ and $x=1$ are labeled as $0$ by every hypothesis in $\cH_k$, we would like $c_0^-$ to include $x=0$ and $c_{k^*}-$ to include $x=1$. Then we denote by $\bar {S_\trn} = S_\trn\cup \{(0,0),(1,0)\}$ and then define $c_i^-$ as
  \begin{align*}
      c_i^-=
      \left\{
      \begin{array}{cl}
       \left[\min\limits_{x\in [a_{2i},a_{2i+1}]:(x,y)\in \bar {S_\trn}}x,\max\limits_{x\in [a_{2i},a_{2i+1}]:(x,y)\in \bar {S_\trn}}x\right] 
       & \quad\text{if }(\bar {S_\trn})\cap [a_{2i},a_{2i+1}]\times \cY\neq \emptyset\,,\\
     \emptyset      & \quad\text{otherwise}\,.
      \end{array}
      \right.
  \end{align*}  
  Let us define two classifiers: $h_c^+ = \ind{\cup_{i=1}^{k^*}c_i^+}$ and $h_c^- = 1- \ind{\cup_{i=0}^{k^*}c_i^-}$. Then we extend $\cH_k$ to $\bar{\cH}_k = \cup_{k'\leq k}\{\ind{\cup_{i=1}^{k'} [a_i,b_i]}|0\leq a_i< b_i\leq 1,\forall i\in [k']\}\cup \cH_k$ by including union of closed intervals. Since both $h_c^+,h_c^-\in \bar{\cH}_k$ are consistent with $S_\trn$ and the VC dimension of $\bar{\cH}_k$ is $2k$, by classic uniform convergence results \citep{vapnik:74,blumer1989learnability}, for any data distribution $\cD$, with probability at least $1-\delta$ over $S_\trn\sim \cD^m$, $\err(h_c^+)\leq \frac{\epsilon}{2}$ and $\err(h_c^-)\leq \frac{\epsilon}{2}$ where $m=O(\frac{1}{\epsilon}(2k\log(1/\epsilon)+\log(1/\delta)))$.


  It is easy to see that the algorithm (even under attack) will always predicts $1$ over $c_i^+$ as the attacker cannot add negative instances into $c_i^+$. Then for any attacker $\Adv$ and any $i\in[k^*]$, for any $x\in \ATK(h^*, S_\trn,\A,\Adv)\cap (a_{2i-1},a_{2i})$, we will have $x\notin c_i^+$, which is classified $0$ by $h_c^+$. Therefore, $\ATK(h^*, S_\trn,\A,\Adv)\cap \{x|h^*(x)=1\}\subseteq \{x|h_c^+(x)=0,h^*(x)=1\}$. We can prove a similar result for $\ATK(h^*, S_\trn,\A,\Adv)\cap \{x|h^*(x)=0\}$. If the algorithm (under attack) predicts $1$ on any $x\in[a_{2i},a_{2i+1}]$ for $i=1,\ldots,k^*-1$, then $c_i^-=\emptyset$ and $h_c^-(x)=1\neq h^*(x)$. Note that the algorithm always correctly labels points in $[0,a_1]$ and $[a_{2k^*},1]$ as there are no positively-labeled points in these two intervals. Therefore, $\ATK(h^*, S_\trn,\A,\Adv)\cap \{x|h^*(x)=0\}\subseteq \{x|h_c^-(x)=1,h^*(x)=0\}$. Then for any point in the attackable region, it is either in the error region of $h_c^+$ or in the error region of $h_c^-$. That is, $\ATK(h^*, S_\trn,\A,\Adv)\subseteq \{x|h_c^+(x)\neq h^*(x)\}\cup \{x|h_c^-(x)\neq h^*(x)\}$. Hence, the attackable rate $\atk(h^*, S_\trn,\A)\leq \err(h_c^+)+\err(h_c^-)\leq \epsilon$.
\end{proof}


\vspace{-20pt}
\subsection{Hypothesis class with infinite hollow star number}\label{appx:hollow}
\begin{thm}\label{thm:hollow}
  For any hypothesis class $\cH$ with hollow star number $k_o\geq 3$, for any consistent proper learner $\A$, there exists a target function $h^*$ and a data distribution $\cD$ such that if $m\leq \floor{(k_o-1)/2}$, then $\EEs{S_\trn\sim \cD^m}{\atk(h^*,S_\trn,\A)}\geq 1/4$ and $\PPs{S_\trn\sim \cD^m}{\atk(h^*,S_\trn,\A)>1/8}\geq 1/7$.
\end{thm}
\begin{proof}
  For $\cH$ with hollow star number $k_o$, there is an unrealizable set $S=\{(x_1,y_1),\ldots, (x_{k_o},y_{k_o})\}$, such that for each $i\in \{1,\ldots,k_o\}$, there exists $h_i\in \cH$ with $\{j\in[k_o]|h_i(x_j)\neq y_j\} = \{i\}$. The target function $h^*=h_{i^*}$. For notation simplicity, we let $i$ represent point $x_i$ and then $h(i)=h(x_i)$ for any classifier $h$ and any $i\in [k_o]$. The marginal data distribution $\cD_\cX$ is a uniform distribution over $[k_o]\setminus\{i^*\}$. We define an attacker which injects all points except $i,i^*$ to attack point $i$ if it is unseen. Specifically, let $S_{i^*,i}=\{(j,h_{i^*}(j))|j\in[k_o]\setminus\{i^*,i\}\}$ denotes the set of all points excluding $i,i^*$ and we define the attacker as
  \begin{align*}
    \Adv(h_{i^*},S_\trn,i)=\begin{cases}
      S_{i^*,i}& \text{if } (i,y_i)\notin S_\trn\,,\\
      \emptyset & \text{else}\,. 
    \end{cases}
  \end{align*}
  Given the input data set $S_\trn\cup \Adv(h_{i^*},S_\trn,i)$, if $(i,y_i)$ is not sampled in the training set, then any algorithm cannot tell whether the true target function is $h_i$ or it is $h_{i^*}$. Since $\{(1,y_1),\ldots,(k_o,y_{k_o})\}$ is unrealizable, any proper classifier consistent with $S_{i,i^*}$ cannot predict $i$ as $h_{i^*}(i)=y_i$ and $i^*$ as $h_i(i^*) =y_{i^*}$ at the same time. Let $S_\trn$ be $m\leq\floor{k_o/2}$ i.i.d. samples from $\cD$ and then we have
  \begin{align*}
    & \sup_{i^*\in [k_o]}\EEs{S_\trn\sim \cD^m}{\atk(h^*,S_\trn,\A)}\\
    \geq &\EEs{i^*\sim \Unif([k_o]), S_\trn\sim \cD^m}{\atk(h^*,S_\trn,\A)}\\
    \geq & \EEs{i^*\sim \Unif([k_o]), S_\trn\sim \cD^m,(i,y_i)\sim \cD,\A}{\ind{\A(S_\trn\cup S_{i^*,i}, i)\neq h_{i^*}(i)\cap i\notin S_{\trn,\cX}}}\\
    \geq &\EEs{i^*, i\sim \Unif([k_o]\setminus\{i^*\} ) }{\EEs{S_\trn\sim\cD^m,\A}{\ind{\A(S_\trn\cup S_{i^*,i}, i)\neq h_{i^*}(i)|i\notin S_{\trn,\cX}}}\cdot \PP{i\notin S_{\trn,\cX}}  }\\
    \geq &\frac{1}{(k_o-1)k_o}\sum_{i^*=1}^{k_o}\sum_{i\neq i^*} \EEs{S_\trn\sim\cD^m,\A}{\ind{\A(S_\trn\cup S_{i^*,i}, i)\neq h_{i^*}(i)|i\notin S_{\trn,\cX}}}\cdot\frac{1}{2}\\
    = &\frac{1}{2(k_o-1)k_o}\sum_{i^*=1}^{k_o}\sum_{i\neq i^*} \EEs{S_\trn\sim\Unif^m(S_{i^*,i}),\A}{\ind{\A(S_\trn\cup S_{i^*,i}, i)\neq h_{i^*}(i)}}\\
    = &\frac{\sum_{i^*<i} \EEs{S_\trn\sim\Unif^m(S_{i^*,i}),\A}{\ind{\A(S_\trn\cup S_{i^*,i}, i)\neq h_{i^*}(i)}+\ind{\A(S_\trn\cup S_{i^*,i}, i^*)\neq h_{i}(i^*)}}}{2(k_o-1)k_o}\\
    \geq &\frac{(k_o-1)k_o}{4(k_o-1)k_o}=\frac{1}{4}\,. 
  \end{align*}
  
  For the second part, by Markov's inequality, we have
  \begin{align*}
      &\PPs{S_\trn\sim \cD^m}{\atk(h^*,S_\trn,\A)>1/8}=
      1-\PPs{S_\trn\sim \cD^m}{\atk(h^*,S_\trn,\A)\leq 1/8}\\
      \geq & 1- \frac{1-\EE{\atk(h^*,S_\trn,\A)}}{7/8}=\frac{1}{7}\,,
  \end{align*}
  which completes the proof.
\end{proof}

\begin{thm}
If $k_o = \infty$, then 
for any consistent \emph{proper} learning algorithm $\A$, for every $m \in \nats$, $\exists h^* \in \cH$ and distribution 
$\cD$ on $D_{h^*}$ such that $\E_{S_\trn \sim \cD^m}[ \atk_{\cD}(h^*,S_\trn,\A) ] \geq 1/4$.
\end{thm}
\begin{proof}
  For any hypothesis class with $k_o=\infty$, there exists a sequence of hollow star set $\{S_i\}_{i=1}^\infty$ with increasing size $\{k_i\}_{i=1}^\infty$ with $k_1\geq 3$. Then following the proof of Theorem~\ref{thm:hollow}, for any $m\leq \floor{(k_i-1)/2}$ for some $i$, there exists a target function and a data distribution such that the expected attackable rate is at least $1/4$ with sample size $m$. Since $k_i\rightarrow \infty$ as $i\rightarrow \infty$, this theorem is proved.
\end{proof}

\vspace{-20pt}
\subsection{Proof of Theorem~\ref{thm:opt-pac}}\label{appx:opt-pac}

Here we present the proof of Theorem~\ref{thm:opt-pac} 
establishing that any deterministic 
robust learner necessarily obtains a 
sample complexity with $O(1/\epsilon)$ 
dependence on $\epsilon$.

\begin{proof}[of Theorem~\ref{thm:opt-pac}]
Without loss of generality, we suppose 
$R(m) \leq 1$ and $R(m)$ is nonincreasing, 
since we can always replace it with 
$\sup_{m' \geq m} \min\{ R(m'), 1 \}$, 
which is monotone and inherits the other assumed 
properties of $R$.
For convenience, 
let us also extend the function $R(m)$ to non-integer values of $m$ by defining $R(\alpha) = R(\lfloor \alpha \rfloor)$, 
and defining $R(0) = 1$.
Also define $\Log(x) = \lceil \log_{2}(x) \rceil$ for any $x \geq 1$.

Fix any $h^* \in \cH$.
Since $\A$ is deterministic, note that for any finite multiset 
$S \subseteq D_{h^*}$ there is a set $\ATK_{S} \subseteq \cX$ corresponding to the points that would be attackable for $\A$ if $S_\trn = S$.
Moreover, we may note that the set $\ATK_{S}$ is \emph{non-increasing} in $S$ 
(subject to $S \subseteq D_{h^*}$), 
since adding any $(x,y) \in D_{h^*}$ to $S$ is equivalent to constraining the adversary to include these points in its attack set. 

Now we argue that $R\!\left(\frac{m}{\Log(1/\delta)}\right)$ 
is a $1-\delta$ confidence bound on 
$\atk_{\cD}(h^*,S_\trn,\A)$.
For any distribution $\cD$ on $D_{h^*}$, and any $\delta \in (0,1)$, 
if $m < \Log(1/\delta)$ then we trivially have $\atk_{\cD}(h^*,S_\trn,\A) \leq R\!\left(\frac{m}{\Log(1/\delta)} \right)$.
Otherwise, if $m \geq \Log(1/\delta)$, then letting $S_\trn \sim \cD^m$, 
letting $S_1$ be the first $\left\lfloor \frac{m}{\Log(1/\delta)} \right\rfloor$ elements of $S_\trn$, 
$S_2$ the next $\left\lfloor \frac{m}{\Log(1/\delta)} \right\rfloor$ elements of $S_\trn$, and so on up to $S_{\Log(1/\delta)}$, 
each $i \leq \Log(1/\delta)$ has, independently, probability at least $\frac{1}{2}$ of 
$\atk_{\cD}(h^*,S_i,\A) \leq R\!\left(\frac{m}{\Log(1/\delta)}\right)$.
In particular, this implies that, with probability at least $1 - (1/2)^{\Log(1/\delta)} \geq 1-\delta$, 
at least one $i \leq \Log(1/\delta)$ will satisfy this inequality.
Moreover, by the monotonicity property of $\ATK$, we know that 
$\ATK_{S_{\trn}} \subseteq \bigcap_{i \leq \Log(1/\delta)} \ATK_{S_{i}}$.
Thus, with probability at least $1-\delta$, 
\begin{align*} 
\atk_{\cD}(h^*,S_\trn,\A) 
& = \P_{(x,y) \sim \cD}( x \in \ATK_{S_{\trn}} ) 
\leq \min_{i \leq \Log(1/\delta)} \P_{(x,y) \sim \cD}( x \in \ATK_{S_i} ) 
\\ & = \min_{i \leq \Log(1/\delta)} \atk_{\cD}(h^*,S_i,\A) 
\leq R\!\left(\frac{m}{\Log(1/\delta)}\right).
\end{align*}

The remainder of the proof follows a familiar ``conditioning'' argument from the literature on log factors in the sample complexity of PAC learning 
\citep*[e.g.,][]{hanneke:thesis,hanneke2016refined}.
Fix any distribution $\cD$ over $D_{h^*}$.
We proceed by induction on $m$, establishing for each $m$ 
that $\forall \delta \in (0,1)$, for $S_{\trn} \sim \cD^m$, with probability at least $1-\delta$, 
$\atk_{\cD}(h^*,S_{\trn},\A) \leq \frac{c}{m}\Log\!\left(\frac{1}{\delta}\right)$, where $c$ is a finite $R$-dependent constant.
Note that this suffices to establish the theorem by taking $c_{R} = c$ 
(assuming base $2$ in the $\log$). 
The claim is trivially satisfied for $m < 3\Log\!\left(\frac{1}{\delta}\right)$, as the claimed bound is vacuous (taking any $c \geq 3$).
Now as an inductive hypothesis suppose $m \geq 3\Log\!\left(\frac{1}{\delta}\right)$ is such that, 
for every $m' < m$, for $S_\trn \sim \cD^{m'}$, 
for any $\delta \in (0,1)$,  
with probability at least $1-\delta$, 
$\atk_{\cD}(h^*,S_\trn,\A) \leq \frac{c}{m'}\Log\!\left(\frac{1}{\delta}\right)$.

Fix any $\delta \in (0,1)$ and let $S_\trn \sim \cD^m$.
Note that $\atk_{\cD}(h^*,S_\trn,\A) = \P_{(x,y) \sim \cD}( x \in \ATK_{S_{\trn}} )$.
Let $S_{\lfloor m/2 \rfloor}$ be the first $\lfloor m/2 \rfloor$ of the data points in $S_\trn$, 
and let $T = (S \setminus S_{\lfloor m/2 \rfloor}) \cap (\ATK_{S_{\lfloor m/2 \rfloor}} \times \cY)$: 
that is, $T$ are the samples in the last $\lceil m/2 \rceil$ points in $S_\trn$ 
that are in the attackable region when $\A$ has training set $S_{\lfloor m/2 \rfloor}$. 

Since, conditioned on $S_{\lfloor m/2 \rfloor}$ and $|T|$, the examples in $T$ 
are conditionally i.i.d.\ 
with each sample having distribution $\cD( \cdot | \ATK_{S_{\lfloor m/2 \rfloor}} \times \cY )$ on $D_{h^*}$, 
the property of $R(\cdot)$ established above 
implies that with conditional (given $S_{\lfloor m/2 \rfloor}$ and $|T|$) 
probability at least $1-\frac{\delta}{3}$, we have 
$\P_{(x,y) \sim \cD}( x \in \ATK_{T} | x \in \ATK_{S_{\lfloor m/2 \rfloor}} ) \leq R\!\left(\frac{|T|}{\Log(3/\delta)}\right)$. 
By the law of total probability, this inequality holds with (unconditional) probability at least $1-\frac{\delta}{3}$.

Furthermore, a Chernoff bound (applied under the conditional distribution given $S_{\lfloor m/2 \rfloor}$) 
and the law of total probability imply that, with probability at least $1-\frac{\delta}{3}$, 
if $\P_{(x,y) \sim \cD}( x \in \ATK_{S_{\lfloor m/2 \rfloor}} ) \geq \frac{16}{m} \ln\frac{3}{\delta}$, 
then $|T| \geq \P_{(x,y) \sim \cD}( x \in \ATK_{S_{\lfloor m/2 \rfloor}} ) \frac{m}{4}$.
Combining these two events with monotonicity of $R$, 
by the union bound, with probability at least $1-\frac{2}{3}\delta$, 
either 
$\P_{(x,y) \sim \cD}( x \in \ATK_{S_{\lfloor m/2 \rfloor}} )  < \frac{16}{m} \ln\frac{3}{\delta}$
or 
$\P_{(x,y) \sim \cD}( x \in \ATK_{T} | x \in \ATK_{S_{\lfloor m/2 \rfloor}} ) 
\leq R\!\left( \P_{(x,y) \sim \cD}( x \in \ATK_{S_{\lfloor m/2 \rfloor}} ) \frac{m}{4 \Log(3/\delta)} \right)$. 

Next, by monotonicity of $\ATK_S$, we have 
$\ATK_{S_{\trn}} \subseteq \ATK_{S_{\lfloor m/2 \rfloor}} \cap \ATK_{T}$.
Therefore, 
$\atk_{\cD}(h^*,S_\trn,\A)$ $\leq \P_{(x,y) \sim \cD}( x \in \ATK_{S_{\lfloor m/2 \rfloor}} ) \P_{(x,y) \sim \cD}( x \in \ATK_{T} | x \in \ATK_{S_{\lfloor m/2 \rfloor}} )$.
Thus, on the above event of probability at least $1-\frac{2}{3}\delta$,
either 
$\atk_{\cD}(h^*,S_\trn,\A)  < \frac{16}{m} \ln\frac{3}{\delta}$
or 
\begin{align*} 
\atk_{\cD}(h^*,S_\trn,\A) 
& \leq \P_{(x,y) \sim \cD}( x \in \ATK_{S_{\lfloor m/2 \rfloor}} ) R\!\left( \P_{(x,y) \sim \cD}( x \in \ATK_{S_{\lfloor m/2 \rfloor}} ) \frac{m}{4\Log(3/\delta)} \right) 
\\ & = \atk_{\cD}(h^*,S_{\lfloor m/2 \rfloor},\A) R\!\left( \atk_{\cD}(h^*,S_{\lfloor m/2 \rfloor},\A) \frac{m}{4\Log(3/\delta)} \right).
\end{align*}

By the inductive hypothesis, with probability at least $1-\frac{\delta}{3}$, we have that 
$\atk_{\cD}(h^*,S_{\lfloor m/2 \rfloor},\A) \leq \frac{c}{\lfloor m/2 \rfloor}\Log\!\left(\frac{3}{\delta}\right) \leq \frac{3 c}{m} \Log\!\left(\frac{3}{\delta}\right)$.
For any $\alpha \geq 1$, define $R'(\alpha) = \frac{1}{\alpha} \sup_{1 \leq \alpha' \leq \alpha} \alpha' R(\alpha')$,  
and note that $R(\alpha) \leq R'(\alpha)$ for all $\alpha \geq 1$, 
and $\alpha R'(\alpha)$ is nondecreasing in $\alpha \geq 1$.
Therefore, on the above event, 
\begin{align*}
& \atk_{\cD}(h^*,S_{\lfloor m/2 \rfloor},\A) R\!\left( \atk_{\cD}(h^*,S_{\lfloor m/2 \rfloor},\A) \frac{m}{4 \Log(3/\delta)} \right)
\\ & \leq \frac{3 c}{m} \Log\!\left(\frac{3}{\delta}\right) R'\!\left( \frac{3 c}{4} \right)
\leq \frac{9 c}{m} \Log\!\left(\frac{1}{\delta}\right) R'\!\left( \frac{3 c}{4} \right).
\end{align*}
Now note that $\lim_{\alpha \to \infty} R'(\alpha) = 0$. 
To see this, for the sake of contradiction, suppose $\exists \epsilon > 0$ and 
a strictly increasing sequence $\alpha_{t} \geq 1$ with $\alpha_{t} \to \infty$ such that $R'(\alpha_{t}) \geq \epsilon$, 
and let $\alpha'_{t}$ be any sequence with $1 \leq \alpha'_{t} \leq \alpha_{t}$ 
and $\frac{1}{\alpha_{t}} \alpha'_{t} R(\alpha'_{t}) \geq R'(\alpha_{t})/2 \geq \epsilon/2$.
If there exists an infinite subsequence $t_{i}$ with $\alpha'_{t_{i}}$ bounded above by some finite $\bar{\alpha}$, 
then $\lim_{i \to \infty} \frac{1}{\alpha_{t_{i}}} \alpha'_{t_{i}} R\!\left(\alpha'_{t_{i}}\right) \leq \lim_{i \to \infty} \frac{\bar{\alpha}}{\alpha_{t_{i}}}=0$: 
a contradiction.  Otherwise, we have $\alpha'_{t} \to \infty$, so that 
$\lim_{t \to \infty} \frac{1}{\alpha_{t}} \alpha'_{t} R(\alpha'_{t}) \leq \lim_{t \to \infty} R(\alpha'_{t}) = 0$: 
again, a contradiction.
Thus, since we have just established that $\lim_{\alpha \to \infty} R'(\alpha) = 0$, 
there exists a sufficiently large choice of $c$ 
for which $R'\!\left( \frac{3 c}{4} \right) \leq \frac{1}{9}$, 
so that  $\frac{9 c}{m} \Log\!\left(\frac{1}{\delta}\right) R'\!\left( \frac{3 c}{4} \right) \leq \frac{c}{m} \Log\!\left(\frac{1}{\delta}\right)$.

Altogether, by the union bound, we have 
established that 
with probability at least $1-\delta$, 
either \linebreak
$\atk_{\cD}(h^*,S_\trn,\A) < \frac{16}{m} \ln\frac{3}{\delta}$
or 
$\atk_{\cD}(h^*,S_\trn,\A) \leq \frac{c}{m} \Log\!\left(\frac{1}{\delta}\right)$.
Taking $c$ sufficiently large so that 
$c \geq 16 \ln(3e)$, 
both cases imply that
$\atk_{\cD}(h^*,S_\trn,\A) \leq \frac{c}{m} \Log\!\left(\frac{1}{\delta}\right)$.
The theorem now follows by the principle of induction.
\end{proof}

\vspace{-20pt}
\section{Proof of Theorem~\ref{thm:lblinear}}\label{appx:lb_linear}
In this section, we first formally prove the statement in the case of $n=3$, for which we already provided a proof sketch in Section~\ref{sec:linear}. In the proof sketch, we relax the definition of linear hypothesis class by allowing the decision boundary to be either positive or negative. Here, we adopt the convention that the boundary is only allowed to be positive. Then we prove the statement in the case of $n=2$, which requires a more delicate construction. Before proving Theorem~\ref{thm:lblinear}, we first introduce a lemma.

\begin{lm}\label{lmm:exp2hp}
For any hypothesis class $\cH$, any algorithm $\A$ and any $m>0$, if there exists a  universal constant $c>0$,
a distribution $\mu$ over $\cH$,
and a set of distributions $\cD(h)$ over $D_{h}$ for every $h\in \supp(\mu)$,
such that $\EEs{h\sim \mu, S_\trn \sim \cD(h)^m}{\atk_{\cD(h)}(h, S_\trn,\A)}\geq 2c$, then $\cH$ is not $(\epsilon,\delta)$-robust learnable.
\end{lm}
\begin{proof}
First, take $\eps = c$, we have by the definition of $\sup$, there exists an $h^*\in \cH$ such that,
\begin{align*}
    &\EEs{S_\trn \sim \cD(h^*)^m}{\atk_{\cD(h^*)}(h^*, S_\trn,\A)} \\
    \geq& \sup_{h\in \cH} \EEs{S_\trn \sim \cD(h)^m}{\atk_{\cD(h)}(h, S_\trn,\A)} - \eps \\
    \geq \textbf{}& \EEs{h\sim \mu, S_\trn \sim \cD(h)^m}{\atk_{\cD(h)}(h, S_\trn,\A)} - \eps\\
    \geq& c\,.
\end{align*}

Then by Markov's inequality,
\begin{align*}
    &\PP{\atk_{\cD(h^*)}(h^*, S_\trn,\A)>c/2}=1-\PP{\atk_{\cD(h^*)}(h^*, S_\trn,\A)\leq c/2}\\
    \geq& 1-\frac{1-\EE{\atk_{\cD(h^*)}(h^*, S_\trn,\A)}}{1-c/2}\geq \frac{c}{2-c}\,.
\end{align*}
Hence, $\cH$ is not $(\epsilon,\delta)$-robust learnable.
\end{proof}

\begin{proof}[of Theorem~\ref{thm:lblinear} in $n=3$] We divide the proof into three parts: a) the construction of the target function and the data distribution, b) the construction of the attacker and c) the analysis of the attackable rate.

\paragraph{The target function and the data distribution. }
We denote by $\sph = \sph^3(\bZero,1)$ the sphere of the $3$-dimensional unit ball centered at the origin. For some small $0<\eta<1/6$, let $\cH_\eta = \{h(x) = \ind{\inner{w}{x}-\frac{1}{2}\geq 0}|\norm{w}=1\}\cup \{h(x) = \ind{\inner{w}{x}-\frac{1-\eta}{2}\leq 0}|\norm{w}=1\}$ denote a set of linear classifiers with boundary $1/2$ or $\frac{1-\eta}{2}$ away from the origin. Let $K_w= \{x|\inner{w}{x}-\frac{1}{2}=0\}$ denote the hyperplane of the boundary of $h=\ind{\inner{w}{x}-\frac{1}{2}\geq 0}$ and $C_w = K_w\cap \sph$ denote the intersection of $K_w$ and $\sph$, which is a circle with radius $\sqrt{3}/2$ centered at $w/2$. 


We consider the target function $h^*$ selected uniformly at random from $\cH_\eta$, which is equivalent to: randomly picking $w\sim \Unif(\sph)$ and randomly picking $j\sim \Ber(1/2)$; if $j=1$, letting $h^* = h_{w,j}^* = \ind{\inner{w}{x}-\frac{1}{2}\geq 0}$; otherwise letting $h^* = h^*_{w,j} = \ind{\inner{w}{x}-\frac{1-\eta}{2}\leq 0}$. If the target function $h^*=h^*_{w,j}$, the data distribution $\cD = \cD_{w,j}$ is the uniform distribution over $C_w\times \{j\}$. Note that all instances on the circle $C_w$ are labeled as $j$ by $h^*_{w,j}$. We will show that the expected attackable rate $\EEs{h^*\sim \Unif(\cH_\eta), S_\trn\sim \cD^m}{\atk_\cD(h^*, S_\trn, \A)}\geq 1/2$. Combining with Lemma~\ref{lmm:exp2hp}, we prove Theorem~\ref{thm:lblinear} in $n=3$.

\paragraph{The attacker. }
Then we define the attacker $\Adv$ in the following way. We first define a map $m_{x_0}: \sph\mapsto \sph$ for some $x_0\in \sph$ such that $m_{x_0}(x) = {2\inner{x_0}{x}}x_0-x$. Here, $m_{x_0}(x)$ is the reflection of $x$ through the line passing the origin and $x_0$. Note that $m_{x_0}(m_{x_0}(x))=x$. This symmetric property will help to confuse algorithms such that no algorithm can distinguish the training data and the poisoning data. For $S_\trn\sim \cD_{w,j}^m$, we define $m_{x_0}(S_\trn) = \{ (m_{x_0}(x),1-y)|(x,y)\in S_\trn \}$, and let
  \begin{align*}
    \Adv(h^*_{w,j},S_\trn,x_0)=\begin{cases}
      m_{x_0}(S_\trn)& \text{if }S_{\trn,\cX}\cap \cB(x_0,\sqrt{3\eta/2})=\emptyset\,,\\
      \emptyset & \text{else.}
    \end{cases}
  \end{align*}
  Now we show that $\Adv$ is a clean-label attacker. In the second case of $\Adv(h^*_{w,j},S_\trn,x_0)=\emptyset$, it is clean-labeled trivially. In the first case of $S_{\trn,\cX}\cap \cB(x_0,\sqrt{3\eta/2})=\emptyset$, we discuss two cases:
  \begin{itemize}
    \item The target function $h^*=h_{w,j}^* = \ind{\inner{w}{x}-\frac{1}{2}\geq 0}$ has its decision boundary $\frac{1}{2}$ away from the origin, i.e., $j=1$. Then every training instance is labeled by $1$ and for any training instance $x$, $\inner{w}{m_{x_0}(x)}-\frac{1}{2} = \inner{x_0}{x}-1<0$. Hence $\Adv(h^*_{w,j},S_\trn,x_0)$ is clean-labeled.
    
    \item The target function $h^*=h_{w,j}^* = \ind{\inner{w}{x}-\frac{1-\eta}{2}\leq 0}$ has its decision boundary $\frac{1-\eta}{2}$ away from the origin, i.e., $j=0$. For each training instance $x$, since $x\notin \cB(x_0,\sqrt{3\eta/2})$, we have $\norm{x-x_0}_2^2\geq \frac{3\eta}{2}$ and thus, $\inner{x}{x_0}\leq 1-\frac{3\eta}{4}$. Then $\inner{w}{m_{x_0}(x)}-\frac{1-\eta}{2} = \inner{x_0}{x}-(1-\frac{\eta}{2})\leq 1-\frac{3\eta}{4}-(1-\frac{\eta}{2})<0$. Hence $\Adv(h^*_{w,j},S_\trn,x_0)$ is clean-labeled.
  \end{itemize}
  
  \paragraph{Analysis. }
  Let $\cE_1(h^*_{w,j},S_\trn,x_0)$ denote the event of $\{\A(S_\trn\cup \Adv(h^*_{w,j},S_\trn,x_0), x_0)\neq h^*_{w,j}(x_0)\}$ 
  and $\cE_2(S_\trn,x_0)$ denote the event of $S_{\trn,\cX}\cap \cB(x_0,\sqrt{3\eta/2})=\emptyset$. 
  It is not hard to check that $\cE_2(S_\trn,x_0) =\cE_2(m_{x_0}(S_\trn),x_0)$ due to the symmetrical property of the reflection. 
  Besides, conditional on $\cE_2(S_\trn,x_0)$, 
  the poisoned data set $S_\trn\cup \Adv(h^*_{w,j},S_\trn,x_0) = m_{x_0}(S_\trn)\cup \Adv(h^*_{m_{x_0}(w),1-j},m_{x_0}(S_\trn),x_0)$ 
  and thus, any algorithm $\A$ will behave the same (under attack) at test instance $x_0$ given training set $S_\trn$ or $m_{x_0}(S_\trn)$. 
  Since $h^*_{w,j}(x_0)\neq h^*_{m_{x_0}(w),1-j}(x_0)$, 
  we know that $\ind{\cE_1(h^*_{w,j},S_\trn,x_0)} = \ind{\neg\cE_1(h^*_{m_{x_0}(w),1-j},m_{x_0}(S_\trn),x_0)}$ conditional on $\cE_2(S_\trn,x_0)$. 
  Let $f_{w,j}(x)$ denote the probability density function of the marginal distribution of $\cD_{w,j}$ 
  (i.e., the uniform distribution over $C_w$) 
  and then we have $f_{w,j}(x) = f_{m_{x_0}(w),1-j}(m_{x_0}(x))$. 
  For any fixed $x_0$, the distributions of $w$ and $m_{x_0}(w)$ and the distributions of $j$ and $1-j$ are the same respectively. 
  The training set $S_\trn$ are samples drawn from $\cD_{w,j}$, and hence we can view $m_{x_0}(S_\trn)$ as samples drawn from $\cD_{m_{x_0}(w),1-j}$. Then for any algorithm $\A$, we have
  \begingroup
  \allowdisplaybreaks
  \begin{align}
    &\EEs{h^*_{w,j}\sim \Unif(\cH_\eta),S_\trn\sim \cD_{w,j}^m}{\atk_\cD(h^*_{w,j}, S_\trn, \A)}\nonumber\\
    \geq &\EEs{w\sim \Unif(\sph), j\sim\Ber(1/2),S_\trn\sim \cD_{w,j}^m,(x,y)\sim \cD_{w,j},\cA}{\ind{\cE_1(h^*_{w,j},S_\trn,x)\cap \cE_2(S_\trn,x)}}\nonumber\\
    =&\int_{x\in \sph}\EEs{w\sim \Unif(\sph), j\sim\Ber(1/2),S_\trn\sim \cD_{w,j}^m,\cA}{f_{w,j}(x)\ind{\cE_1(h^*_{w,j},S_\trn,x)\cap  \cE_2(S_\trn,x)}}dx\label{eq:toadd}\\
    =&\int\limits_{x\in \sph}\underset{{w, j,S_\trn\sim \cD_{w,j}^m,\cA}}{\E}{\left[f_{m_{x}(w),1-j}(x)\ind{\neg\cE_1(h^*_{m_x(w),1-j},m_x(S_\trn),x)\cap \cE_2(m_x(S_\trn),x)}\right]}dx\label{eq:sameevent}\\ 
    =&\int_{x\in \sph}\EEs{w, j,S_\trn\sim \cD_{w,j}^m,\cA}{f_{w,j}(x)\ind{\neg\cE_1(h^*_{w,j},S_\trn,x)\cap  \cE_2(S_\trn,x)}}dx\label{eq:samedist}\\
    =&\frac{1}{2}\int_{x\in \sph}\EEs{w\sim \Unif(\sph), j\sim \Ber(1/2),S_\trn\sim \cD^m,\cA}{f_{w,j}(x)\ind{\cE_2(S_\trn,x)}}dx
    \xrightarrow{\eta\rightarrow 0^+} \frac{1}{2}\,,\label{eq:addition}
  \end{align}
  \endgroup
  where Eq.~\eqref{eq:sameevent} uses the fact $\cE_2(S_\trn,x_0) =\cE_2(m_{x_0}(S_\trn),x_0)$ and that conditional on $\cE_2(S_\trn,x_0)$, $\ind{\cE_1(h^*_{w,j},S_\trn,x_0)} = \ind{\neg\cE_1(h^*_{m_{x_0}(w),1-j},m_{x_0}(S_\trn),x_0)}$; Eq.~\eqref{eq:samedist} uses the fact that for any fixed $x_0$, the distributions of $w$ and $m_{x_0}(w)$ and the distributions of $j$ and $1-j$ are the same respectively; and Eq.~\eqref{eq:addition} is the average of Eq.~\eqref{eq:toadd} and Eq.~\eqref{eq:samedist}.
\end{proof}

\begin{proof}[of Theorem~\ref{thm:lblinear} in $n=2$]
Again, we divide the proof into three parts.

\paragraph{The target function and the data distribution.}
In $2$-dimensional space, we denote by $w = (\cos\theta,\sin\theta)$ and represent the target function $h^* = \ind{\inner{(\cos\theta^*,\sin\theta^*)}{x}+b^*\geq 0}$ by $(\theta^*,b^*)$. Then the target function is selected in the following way: uniformly at random selecting a point $o$ from a $2$-dimensional ball centered at $\bZero$ with some large enough radius $r\!\geq\! 4$, i.e., $o\sim\! \Unif(\cB^2(\bZero,\!r))$, then randomly selecting a direction $\theta^* \sim \Unif([0,2\pi))$, and letting the target function be $h^* = \ind{\inner{(\cos \theta^*,\sin \theta^*)}{x-o}\geq 0}$. Then for any $m\in \NN$, we construct the data distribution over $2m$ discrete points, where all points are labeled the same and the distance between every two instances is independent of $h^*$. Specifically, the data distribution $\cD$ is described as follows. 
  \begin{itemize}
    \item We randomly draw $s \sim \Ber(1/2)$. We define two unit vectors $v_1 = (\sin \theta^*, -\cos \theta^*)$ and $v_2 =(2s-1)\cdot (\cos \theta^*,\sin \theta^*)$. Here $v_1$ is perpendicular to $w^*$ and $v_2$ is in the same direction as $w^*$ if $s=1$ and in the opposite direction of $w^*$ if $s=0$.
    \item Let $\cX_m = \{x_1,\ldots,x_{2m}\}$ be a set of $2m$ points. For notation simplicity, we also define $x_0$ and $x_{2m+1}$. Let $x_0 = o$ and for all $i\in [2m+1]$, let $x_{i} = x_{i-1}+l\cos(\beta_{i-1})v_1+ l\sin(\beta_{i-1})v_2$, where $\beta_{i}=7\beta_{i-1}$, $\beta_0 = 7^{-2m}\cdot \frac{\pi}{6}$ and $l=\frac{1}{2m}$.
    \item Let the marginal data distribution be a uniform distribution over $\cX_m$. Note that if $s=1$, all training points lie on the positive side of the decision boundary and are labeled by $1$; if $s=0$, all training points lie on the negative side and are labeled by $0$.
  \end{itemize}
  Here $(o,v_1,v_2)$ constructs a new coordinate system. For any $x\in \R^2$, we use $\tilde{x}= ((x-o)^\top v_1, (x-o)^\top v_2)$ to represent $x$ in this new coordinate system. Then the decision boundary of the target function is represented as $\inner{\tilde{x}}{\tilde{v_2}}=0$ and for any $x\in \cX_m$ we have $\inner{\tilde{x}}{\tilde{v_2}}>0$. It is worth noting that for any $i\in[2m]$, if the positions of three points $x_{i-1},x_i,x_{i+1}$ are fixed, then $o,s,v_1,v_2$ are all fixed.
  
  \paragraph{The attacker.}
  For any ${x_i}\in \cX_m$, we let $b_i=l\sum_{j=0}^i \cos \beta_j -\frac{l(\cos \beta_{i-1} +\cos \beta_i)\sum_{j=0}^i \sin \beta_j}{\sin \beta_{i-1}+\sin \beta_i}$ such that $z_i = {b_iv_1+o}$, ${x_{i-1}}$ and ${x_{i+1}}$ are collinear. We denote by $L_{x_i}$ the line passing $z_i$ and $x_{i}$. Then for any $i\in [2m]$, let
  \[
    \tilde{\Rfl_{L_{x_i}}(x)}=\frac{2\inner{\tilde{x}-\tilde{z_i}}{\tilde{x_i}-\tilde{z_i}}}{\norm{\tilde{x_i}-\tilde{z_i}}_2^2}(\tilde{x_i}-\tilde{z_i}) -\tilde{x}+2\tilde{z_i}\,,\]
  be the reflection of $x$ across $L_{x_i}$ as illustrated in Fig.~\ref{fig:lblin2d}. 
  
  \begin{figure}[H]
      \centering
      \includegraphics[width= 0.5\textwidth]{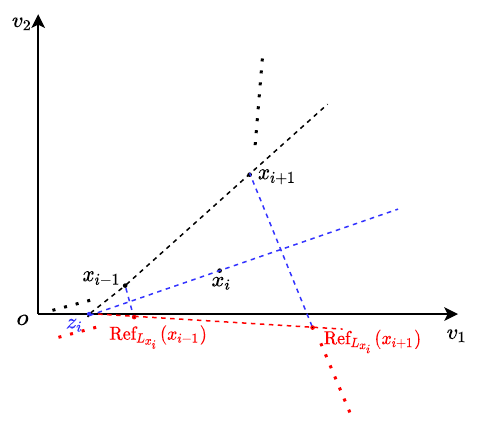}
      \caption{Illustration of $\Rfl_{L_{x_i}}(\cdot)$.}
      \label{fig:lblin2d}
\end{figure}

For $S_\trn\sim \cD^m$, we let $U=\cX_m\setminus S_{\trn,\cX}$ denote the set of points not sampled in $\cX_m$, where $\abs{U}\geq m$. Then we define $\Adv$ as
  \begin{align*}
      \Adv(h^*,S_\trn, x_i)=\begin{cases}
      \{(\Rfl_{L_{x_i}}(x),1-y)|(x,y)\in S_\trn\}&\text{if } x_i\in U\,,\\
      \emptyset & \text{else.}
      \end{cases}
  \end{align*}
  Now we need to show that $\Adv$ is a clean-label attacker. We will show that $\inner{\tilde{\Rfl_{L_{x_i}}(x_j)}}{\tilde{v_2}}<0$ for all $j\neq i\in [2m]$, which implies that the poison data is correctly labeled when $x_i\in U$. First, we claim that for any $j\neq i$, $x_j$ lie in the polytope above the line passing $x_{i-1},x_{i+1}$, the line passing $x_{i-1},x_i$ and the line passing $x_{i},x_{i+1}$. Formally speaking, for any $j\neq i$, $x_j$ satisfies
  \begin{align*}
      \begin{array}{l}
      \inner{(-\sin \beta_{i-1}-\sin \beta_i,\cos \beta_{i-1}+\cos \beta_i)}{\tilde{x_j}-\tilde{x_{i+1}}}\geq 0\,,\\
      \inner{(-\sin \beta_{i-1},\cos \beta_{i-1})}{\tilde{x_j}-\tilde{x_{i}}}\geq 0\,,\\
      \inner{(-\sin \beta_i,\cos \beta_i)}{\tilde{x_j}-\tilde{x_{i}}}\geq 0\,.
      \end{array}
  \end{align*}
  This claim is not hard to prove. At a high level, we prove this claim by that $\{\beta_i\}_{i=0}^{2m}$ is monotonically increasing and that the polygon defined by connecting every pair of neighboring points in $\cX_m\cup \{x_0,x_{2m+1}\}$ is convex. For the first constraint, it is satisfied trivially when $j=i-1,i+1$. If $j\geq i+2$, by direct calculation, we have 
  \begingroup
  \allowdisplaybreaks
  \begin{align*}
      &\inner{(-\sin \beta_{i-1}-\sin \beta_i,\cos \beta_{i-1}+\cos \beta_i)}{\tilde{x_j}-\tilde{x_{i+1}}}\\
      =&\inner{(-\sin \beta_{i-1}-\sin \beta_i,\cos \beta_{i-1}+\cos \beta_i)}{\left(\sum_{k=i+1}^{j-1}\cos \beta_k, \sum_{k=i+1}^{j-1}\sin \beta_k\right)}\\
      = &\sum_{k=i+1}^{j-1}\cos \beta_k\inner{(-\sin \beta_{i-1}-\sin \beta_i,\cos \beta_{i-1}+\cos \beta_i)}{\left(1, \frac{\sum_{k=i+1}^{j-1}\sin \beta_k}{\sum_{k=i+1}^{j-1}\cos \beta_k}\right)}\\
      \geq &\sum_{k=i+1}^{j-1}\cos \beta_k\inner{(-\sin \beta_{i-1}-\sin \beta_i,\cos \beta_{i-1}+\cos \beta_i)}{\left(1, \tan \beta_{i+1} \right)}\\
      \geq &\frac{\sum_{k=i+1}^{j-1}\cos \beta_k}{\cos \beta_{i+1}}(\sin(\beta_{i+1}-\beta_{i-1}) +\sin(\beta_{i+1}-\beta_{i}))\\
      \geq & 0\,.
  \end{align*}
  \endgroup
  Similarly, if $j\leq i-2$, then
  \begin{align*}
      &\inner{(-\sin \beta_{i-1}-\sin \beta_i,\cos \beta_{i-1}+\cos \beta_i)}{\tilde{x_j}-\tilde{x_{i+1}}}\\
      =&\inner{(-\sin \beta_{i-1}-\sin \beta_i,\cos \beta_{i-1}+\cos \beta_i)}{\tilde{x_j}-\tilde{x_{i-1}}}\\
      =&-\inner{(-\sin \beta_{i-1}-\sin \beta_i,\cos \beta_{i-1}+\cos \beta_i)}{\left(\sum_{k=j}^{i-2}\cos \beta_k, \sum_{k=j}^{i-2}\sin \beta_k\right)}\\
      = &\sum_{k=j}^{i-2}\cos \beta_k\inner{(\sin \beta_{i-1}+\sin \beta_i,-\cos \beta_{i-1}-\cos \beta_i)}{\left(1, \frac{\sum_{k=j}^{i-2}\sin \beta_k}{\sum_{k=j}^{i-2}\cos \beta_k}\right)}\\
      \geq &\sum_{k=j}^{i-2}\cos \beta_k\inner{(\sin \beta_{i-1}+\sin \beta_i,-\cos \beta_{i-1}-\cos \beta_i)}{\left(1, \tan \beta_{i-2} \right)}\\
      \geq &\frac{\sum_{k=j}^{i-2}\cos \beta_k}{\cos \beta_{i-2}}(\sin(\beta_{i-1}-\beta_{i-2}) +\sin(\beta_{i}-\beta_{i-2}))\\
      \geq & 0\,.
  \end{align*}
  It is easy to check that $x_j$ satisfies the second and the third constraints using the same way of computation, which is omitted here. Based on that $x_j$ lie in the polytope for all $j\neq i$, then we only need to prove that 
  $\inner{\tilde{\Rfl_{L_{x_i}}(x)}}{\tilde{v_2}}<0$ for the points lying on the faces of the polytope, which are $\tilde{x}\in \{\tilde{x_{i+1}} +\eta (\cos \beta_{i},\sin \beta_{i})|\eta\geq 0\}$, $\tilde{x}\in \{\tilde{x_{i-1}} -\eta (\cos \beta_{i-1},\sin \beta_{i-1})|\eta\geq 0\}$ and $\tilde{x}\in \{\eta \tilde{x_{i-1}} +(1-\eta)\tilde{x_{i+1}}|\eta\in [0,1]\}$. 
  Since $\tilde{\Rfl_{L_{x_i}}(\cdot)}$ is a linear transform, if we can show $\inner{\tilde{\Rfl_{L_{x_i}}(x_{i-1})}}{\tilde{v_2}}<0$ and 
  $\inner{\tilde{\Rfl_{L_{x_i}}(x_{i+1})}}{\tilde{v_2}}<0$, 
  then we have $\inner{\tilde{\Rfl_{L_{x_i}}(x)}}{\tilde{v_2}}<0$ for all points on the third face 
  $\{\eta \tilde{x_{i-1}} +(1-\eta)\tilde{x_{i+1}}|\eta\in [0,1]\}$. 
  Hence, we only need to prove the statement for points lying on the first two faces.
  
  For any two vectors $u,v$, we denote by $\theta(u,v)$ the angle between $u$ and $v$. Then let us denote by $\theta_1 = \theta(\tilde{x_{i+1}}-\tilde{z_i},\tilde{x_{i}}-\tilde{z_i})$ the angle between $\tilde{x_{i+1}}-\tilde{z_i}$ and  and $\tilde{x_{i}}-\tilde{z_i}$ and $\theta_2 = \theta(\tilde{x_{i}}-\tilde{z_i},\tilde{v_1})$ the angle between $\tilde{x_{i}}-\tilde{z_i}$ and $\tilde{v_1}$. Then we have both $\theta_1\leq \beta_{i}\leq \frac{\pi}{6}$ and $\theta_2\leq \beta_{i}\leq \frac{\pi}{6}$. Then since $\norm{\tilde{x_i}-\tilde{x_{i-1}}} = \norm{\tilde{x_{i+1}}-\tilde{x_{i}}} = l$ and $\theta(\tilde{x_{i+1}}-\tilde{x_{i-1}},\tilde{x_i}-\tilde{x_{i-1}}) = (\beta_i-\beta_{i-1})/2$ due to the construction, we have
  \begin{align*}
    &\sin \theta_1 = \frac{\norm{\tilde{x_{i}}-(\tilde{x_{i+1}}+\tilde{x_{i-1}})/2}}{\norm{\tilde{x_{i}}-\tilde{z_i}}}=\frac{l\sin((\beta_i-\beta_{i-1})/2)}{\norm{\tilde{x_{i}}-\tilde{z_i}}}
    =\frac{l\sin(3\beta_{i-1})}{\norm{\tilde{x_{i}}-\tilde{z_i}}}\geq \frac{3\beta_{i-1}l}{2\norm{\tilde{x_{i}}-\tilde{z_i}}}\,,
  \end{align*}
  where the last inequality is by $3\beta_{i-1}\leq \frac{\pi}{6}$. On the other hand, we have
  \begin{align*}
    \sin \theta_2 = \frac{l\sum_{k=0}^{i-1}\sin \beta_k}{\norm{\tilde{x_{i}}-\tilde{z_i}}}\leq \frac{l\sum_{k=0}^{i-1}\beta_k}{\norm{\tilde{x_{i}}-\tilde{z_i}}}\leq \frac{7\beta_{i-1}l}{6\norm{\tilde{x_{i}}-\tilde{z_i}}}\,.
  \end{align*} 
  Combining these two equations, we have $\sin\theta_1>\sin \theta_2$, which indicates that $\theta_1>\theta_2$. Then since $\beta_i-\theta_2 = \theta(\tilde{x_{i+1}}-\tilde{x_i},\tilde{x_{i}}-\tilde{z_i}) \geq \theta_1$, we have $\beta_i> 2\theta_2$. Let $\tilde{w_i} = \frac{\tilde{x_{i}}-\tilde{z_i}}{\norm{\tilde{x_{i}}-\tilde{z_i}}}$ denote the unit vector in the direction of $\tilde{x_{i}}-\tilde{z_i}$. For $\tilde{x}=\tilde{x_{i+1}} +\eta (\cos \beta_{i},\sin \beta_{i})$ with $\eta\geq 0$,
  \begin{align*}
      &\inner{\tilde{\Rfl_{L_{x_i}}(x)}}{\tilde{v_2}}\\
      =&{2\inner{\tilde{x}-\tilde{z_i}}{\tilde{w_i}}}\inner{\tilde{w_i}}{\tilde{v_2}} -\inner{\tilde{x}-\tilde{z_i}}{\tilde{v_2}}\\
      =&{2\inner{\tilde{x_{i+1}}-\tilde{z_i}+\eta (\cos \beta_{i},\sin \beta_{i})}{\tilde{w_i}}}\inner{\tilde{w_i}}{\tilde{v_2}} -\inner{\tilde{x_{i+1}}-\tilde{z_i}+\eta (\cos \beta_{i},\sin \beta_{i})}{\tilde{v_2}}\\
      =&2\norm{\tilde{x_{i+1}}-\tilde{z_i}}\cos \theta_1\sin \theta_2-\norm{\tilde{x_{i+1}}-\tilde{z_i}}\sin(\theta_1+\theta_2) +2\eta\cos(\beta_i-\theta_2)\sin \theta_2-\eta \sin\beta_i\\
      = &\norm{\tilde{x_{i+1}}-\tilde{z_i}}\sin(\theta_2-\theta_1) +\eta \sin (2\theta_2-\beta_i)\\
      < & 0\,.
  \end{align*}
  It is easy to check that $\beta_{i-1}\leq \theta_2$ (let $\tilde{p}$ denote the intersection of the line passing $\tilde{x_{i-1}}$ and $\tilde{x_i}$ and the line $\inner{\tilde{x}}{\tilde{v_2}}=0$, $\beta_{i-1} = \theta(\tilde{x_i}-\tilde{p}, \tilde{v_1})$ and $\theta_2$ is the external angle of triangle with vertices $\tilde{p}, \tilde{z_i}$ and $\tilde{x_i}$). Then for $\tilde{x}=\tilde{x_{i-1}} -\eta (\cos \beta_{i-1},\sin \beta_{i-1})$ with $\eta\geq 0$,
  \begin{align*}
      &\inner{\tilde{\Rfl_{L_{x_i}}(x)}}{\tilde{v_2}}\\
      =&{2\inner{\tilde{x}-\tilde{z_i}}{\tilde{w_i}}}\inner{\tilde{w_i}}{\tilde{v_2}} -\inner{\tilde{x}-\tilde{z_i}}{\tilde{v_2}}\\
      =&{2\inner{\tilde{x_{i-1}}-\tilde{z_i}-\eta (\cos \beta_{i-1},\sin \beta_{i-1})}{w_i}}\inner{\tilde{w_i}}{\tilde{v_2}} -\inner{\tilde{x_{i-1}}-\tilde{z_i}-\eta (\cos \beta_{i-1},\sin \beta_{i-1})}{\tilde{v_2}}\\
      =&2\norm{\tilde{x_{i-1}}-\tilde{z_i}}\cos \theta_1\sin \theta_2-\norm{\tilde{x_{i-1}}-\tilde{z_i}}\sin(\theta_1+\theta_2) -2\eta\cos(\beta_{i-1}-\theta_2)\sin \theta_2+\eta \sin\beta_{i-1}\\
      = &\norm{\tilde{x_{i-1}}-\tilde{z_i}}\sin(\theta_2-\theta_1) +\eta \sin (\beta_{i-1}-2\theta_2)\\
      < & 0\,.
  \end{align*}
  Now we complete the proof of $\inner{\tilde{\Rfl_{L_{x_i}}(x_j)}}{\tilde{v_2}}<0$ for all $j\neq i$ and that $\Adv$ is a clean-label attacker. It is worth noting that $L_{x_i'}=L_{x_i}$, where $L_{x_i'}$ is defined over $\{x_j'|j\in [2m]\}$ in the same way as $L_{x_i}$ defined over $\{x_j|j\in [2m]\}$. This is because reflections of $z_i$ and $x_i$ over $L_{x_i}$ are themselves. This symmetric property plays an important role in the analysis.

\paragraph{Analysis.}
Our probabilistic construction of the target function $h^*$ and the data distribution $\cD$ and the random sampling process of drawing $m$ i.i.d. samples from $\cD$ is equivalent to: sampling a multiset of indexes $I_\trn\sim \Unif([2m])$ first; then selecting the target function and the data distribution to determine the positions of the $m$ training points; mapping $I_\trn$ to $S_\trn$ by adding instance-label pair $(x_i,h^*(x_i))$ to $S_\trn$ for each $i$ in $I_\trn$. We let $I_u = [2m]\setminus I_\trn$ denote the indexes not sampled. As we know from the construction, for any $i\in[2m]$, once $s$ and the positions of $o$ and $x_i$ is determined, the positions of other points in $\cX_m$ and $h^*$ are determined. Then we consider an equivalent way of determining the target function and the data distribution. That is, randomly selecting the position of $x_i$ (dependent on the randomness of $o,\theta^*,s$) and then considering the following two different processes of selecting $s$ and $o$.
  \begin{itemize}
      \item Given a fixed $x_i$, randomly select $s\sim \cD(s|x_i)$ and select $o\sim \cD(o|s,x_i)$, where $\cD(s|x_i)$ and $\cD(o|s,x_i)$ denote the conditional distributions of $s$ and $o$ respectively. Note that when $x_i$ satisfies $\norm{x_i}\leq r-2$, $\cD(s|x_i) = \Ber(1/2)$ and $\cD(o|s,x_i)=\Unif(\sph^2(x_i,r_i))$ is a uniform distribution over the circle with radius $r_i$ centered at $x_i$, where $r_i$ is the distance between $o$ and $x_i$ and is a constant according to the definition.
      
      \item Given $(x_i,s,o)$ selected in the above process, if $\norm{x_i}\leq r-2$, we let $s'=1-s$ and $o'=\Rfl_{L_{x_i}}(o)$ (where $x_{i+1}$ and $x_{i-1}$ is determined by $(x_i,s,o)$); otherwise we let $s'=s$ and $o'=o$. It is easy to check that the distribution of $s'$ conditional on $x_i$ is $\Ber(1/2)$ and the distribution of $o'$ conditional on $x_i$ is $\Unif(\sph^2(x_i,r_i))$ if $\norm{x_i}\leq r-2$.
  \end{itemize}
  Therefore, the distributions of $s$ and $s'$ and the distributions of $o$ and $o'$ are the same given $x_i$ respectively. 
  Our following analysis depends on the event of $\norm{x_i}\leq r-2$, 
  the probability of which is $\PP{\norm{x_i}\leq r-2}\geq \PP{\norm{o}\leq r-3} = \frac{(r-3)^2}{r^2}$. 
  We let $S_\trn(x_i,s,o)$ denote the training set by mapping $I_\trn$ to the positions determined by $(x_i,s,o)$ 
  and let $h^*(x_i,s,o)$ denote the target function determined by $(x_i,s,o)$. 
  Note that when $\norm{x_i}\leq r-2$ and $x_i$ is not in the training set, the poisoned data sets with the training sets generated in the above two different processes are the same, i.e., 
  $S_\trn(x_i,s,o)\cup \Adv(h^*(x_i,s,o),S_\trn(x_i,s,o), x_i) = S_\trn(x_i,s',o')\cup \Adv(h^*(x_i,s',o'),S_\trn(x_i,s',o'), x_i)$.
  This is due to the symmetric property of the attacker. 
  Hence, any algorithm will behave the same at point $x_i$ no matter whether the training set is $S_\trn(x_i,s,o)$ or $S_\trn(x_i,s',o')$. 
  In addition, the target functions produced in the two different processes classify $x_i$ differently when $\norm{x_i}\leq r-2$. Let $\cE_2(x_i)$ denote the event of $\{\norm{x_i}\leq r-2\}$ and $\cE(h^*,\A,S_\trn,i)$ denote the event of $\A(S_\trn\cup \Adv(h^*,S_\trn,x_i), x_i)\neq h^*(x_i)$. Then for any $i\in I_u$, conditional on $\cE_2(x_i)$, 
  for any algorithm $\A$, we have 
  \begingroup
  \allowdisplaybreaks
  \begin{align}
      \ind{\cE(h^*(x_i,s,o),\A,S_\trn(x_i,s,o),i)} = \ind{\neg \cE(h^*(x_i,s',o'),\A,S_\trn(x_i,s',o'),i)}\,.\label{eq:s0}
  \end{align}
  Similar to the proof in the case of $n=3$, we have the expected attackable rate
    \begin{align}
      &\EEs{h^*,s,S_\trn\sim \cD^m}{\atk(h^*, S_\trn,\A)}\nonumber\\
      \geq &\frac{1}{2m}\EEs{o\sim \Unif(\{x:\norm{x}\leq r\}),\theta^* \sim \Unif(2\pi),s\sim \Ber(1/2),S_\trn\sim \cD^m,\A}{\sum_{x_i\in U}\ind{\cE(h^*,\A,S_\trn,i)}}\nonumber\\
      = &\frac{1}{2m}\EEs{I_\trn\sim \Unif([2m])}{\sum_{i\in I_u}\EEs{x_i,s,o,\A}{\ind{\cE(h^*(x_i,s,o),\A,S_\trn(x_i,s,o),i)}}}\nonumber\\
      \geq &\frac{1}{2m}\EEs{I_\trn}{\sum_{i\in I_u}\EEs{x_i}{\EEs{s,o,\A}{\ind{\cE(h^*(x_i,s,o),\A,S_\trn(x_i,s,o),i)}|x_i}\ind{\cE_2(x_i)}}}\nonumber\\
      = &\frac{1}{4m}\left(\EEs{I_\trn}{\sum_{i\in I_u}\EEs{x_i}{\EEs{s,o,\A}{\ind{\cE(h^*(x_i,s,o),\A,S_\trn(x_i,s,o),i)}|x_i}\ind{\cE_2(x_i)}}}\right.\nonumber\\
      +& \left.\EEs{I_\trn}{\sum_{i\in I_u}\EEs{x_i}{\EEs{s',o',\A}{\ind{\neg\cE(h^*(x_i,s',o'),\A,S_\trn(x_i,s',o'),i)}|x_i}\ind{\cE_2(x_i)}}}\right)\label{eq:s1}\\
      = &\frac{1}{4m}\EEs{I_\trn}{\sum_{i\in I_u}\EEs{x_i}{\ind{\norm{x_i}\leq r-2}}}\label{eq:s2}\\
      \geq&\frac{m(r-3)^2}{4mr^2}\nonumber\\
      \geq& \frac{1}{64},\nonumber
    \end{align}
    \endgroup
    when $r\geq 4$. Here Eq.~\eqref{eq:s1} holds due to Eq.~\eqref{eq:s0} and Eq.~\eqref{eq:s2} holds since the distributions of $s$ and $s'$ and the distributions of $o$ and $o'$ are the same given $x_i$ respectively. Combining with Lemma~\ref{lmm:exp2hp}, we complete the proof.
\end{proof}
\vspace{-20pt}
\section{Proof of Theorem~\ref{thm:lin2d}}\label{appx:lin2d}
\begin{proof} The proof contains three steps. For notation simplicity, we sometimes use $(\beta,b)$ to represent the linear classifier $h_{\beta,b}$. We say an angle $\beta$ is consistent with a data set $S$ if there exists $b\in [-2,2]$ such that $(\beta,b)$ is consistent with $S$. Also, for an fixed $\beta$, we say an offset $b$ is consistent with $S$ if $(\beta,b)$ is consistent with $S$.
\paragraph{Step 1: For any fixed $\beta$, the probability mass of union of error region is bounded.} For any $\beta$, if there exists any $b\in[-2,2]$ such that $(\beta,b)$ is consistent with $S_\trn$, from this set of consistent $b$'s, 
we denote by $b_{\sup}(\beta)$ the superior value of this set and $b_{\inf}(\beta)$ the inferior value of this set. 
By uniform convergence bound in PAC learning~\citep{blumer1989learnability}, when $m\geq \frac{4}{\epsilon'} \log \frac{2}{\delta} +  \frac{24}{\epsilon'} \log \frac{13}{\epsilon'}$,
we have that with probability at least $1-\delta$, every linear classifier $(\beta,b)$ consistent with $S_\trn$ has $\err(h_{\beta,b})\leq \epsilon'$.
Then the probability mass of the union of error region of all $(\beta,b)$ consistent with $S_\trn$ for a fixed $\beta$ is
\begin{align*}
    &\cP(\{x| \exists b\in [-2,2], h_{\beta,b}(x)\neq h^*(x), \forall (x',y')\in S_\trn, y'=h_{\beta,b}(x') \})\\
    = &\cP( \bigcup\limits_{b_{\inf}(\beta) \leq b\leq b_{\sup}(\beta) } \{x| b \text{~is consistent}~\&~ h_{\beta,b}(x)\neq h^*(x) \}  )\\
    =& \lim\limits_{\delta \to 0^+}  \cP( \{x| h_{\beta,b_{\inf}(\beta)+ \delta}(x)\neq h^*(x) \}\cup\{x|b_{\inf}(\beta)\text{~is consistent}~\&~ h_{\beta,b_{\inf}(\beta)}(x)\neq h^*(x) \} \\
    &\quad\cup  \{x|  h_{\beta,b_{\sup}(\beta)- \delta}(x)\neq h^*(x) \} \cup  \{x|b_{\sup}(\beta)\text{~is consistent} ~\&~ h_{\beta,b_{\sup}(\beta)}(x)\neq h^*(x) \} ) \\
    \leq &2\epsilon'\,.
\end{align*}
If there does not exist any consistent $b\in[-2,2]$ for $\beta$, then $\cP(\{x| \exists b\in [-2,2], h_{\beta,b}(x)\neq h^*(x), \forall (x',y')\in S_\trn, y'=h_{\beta,b}(x') \})=0$.

\paragraph{Step 2: The binary-search path of $\beta$ is unique and adding clean-label points can only change the depth of the search.} That is, for any fixed target function $(\beta^*,b^*)$, for $h-l=2\pi,\pi,\frac{\pi}{2},\ldots$, 
if $\frac{l+h}{2}$ is not consistent with the input (poisoned or not) data set $S$,
then there cannot exist $\beta$ consistent with the input data set $S$ in both two intervals $(l,\frac{l+h}{2})$ and $(\frac{l+h}{2},h)$. 
Since $\beta^*$ is always consistent with $S$, only the interval containing $\beta^*$ will contain $\beta$ consistent with $S$. 
To prove this statement, assume that any $\beta \in\{ l,h, \frac{l+h}{2}\}$ is not consistent with $S$ and there exists $(\beta_1,b_1)$ with $\beta_1\in (l,\frac{l+h}{2})$ and $(\beta_2,b_2)$ with $\beta_2\in (\frac{l+h}{2},h)$ consistent with $S$. If $\beta_2-\beta_1\leq \pi$, let $\beta_3 =\frac{l+h}{2}$; otherwise, let $\beta_3=l$. 
Since $(\beta_1,b_1)$ and $(\beta_2,b_2)$ are consistent classifiers,
for any $\alpha_1,\alpha_2\geq 0$, 
$\ind{(\alpha_1(\cos\beta_1,\sin \beta_1)+\alpha_2(\cos\beta_2,\sin \beta_2))\cdot x+ \alpha_1 b_1+\alpha_2 b_2\geq 0}$ is also a consistent classifier.
By setting $\alpha_1 = \frac{\sin (\beta_2-\beta_3)}{\sin (\beta_2-\beta_1)}$ 
and $\alpha_2 = \frac{\sin (\beta_3-\beta_1)}{\sin (\beta_2-\beta_1)}$, 
we have $(\beta_3, \frac{\sin (\beta_2-\beta_3)b_1+\sin (\beta_3-\beta_1)b_2}{\sin(\beta_2-\beta_1)})$ is consistent. If $\frac{\sin (\beta_2-\beta_3)b_1+\sin (\beta_3-\beta_1)b_2}{\sin(\beta_2-\beta_1)}\in [-2,2]$, this contradicts that any $\beta_3$ is not consistent with $S$; else, since $\cX\subseteq\cB^n(\bZero,1)$, there must exist $b\in[-2,2]$ such that $(\beta_3, b)$ is consistent, which is a contradiction.
  
\paragraph{Step 3: When $h-l< \arctan(f(\epsilon'')/2)$, the attackable rate caused by deeper search is at most $2\epsilon''$.} We consider two cases: $\abs{b^*}> 1$ and $\abs{b^*}\leq 1$. In the case of $\abs{b^*}> 1$, the target function classifies $\cX$ all positive or all negative and thus, there always exists a consistent $b$ for $\beta=0$. The binary-search for $\beta$ will not search in depth and output $(0,b)$ for some consistent $b$. Therefore, $\atk(h^*, S_\trn, \cA)\leq \cP(\{x| \exists b\in [-2,2], h_{0,b}(x)\neq h^*(x), \forall (x',y')\in S_\trn, y'=h_{0,b}(x') \})\leq 2\epsilon'$. 

In the case of $\abs{b^*}\leq 1$, let $A_{z}\eqdef \{x|(\cos \beta^*,\sin \beta^*)\cdot x +b^*\in [-z,z]\}$ for $z\geq 0$. We will show that when $h-l\leq {\arctan(z/2)}$ for $z\leq 2$, the classifier boundary $K_{\beta,b}$ must lie in $A_z$, i.e., $K_{\beta,b}\eqdef \{x|(\cos \beta,\sin \beta)\cdot x+b=0, \norm{x}\leq 1\}\subseteq A_z$. This indicates that when $h-l\leq {\arctan(z/2)}$, the algorithm can only make mistakes at points inside $A_z$. 
If it is false, let $\{q_1,q_2\}=\{x|(\cos \beta^*,\sin \beta^*)\cdot x+b^*=0\}\cap \sph(\bZero,1)$ with 
$\beta^*-\theta((q_2-q_1), e_1) = \frac{\pi}{2}\pmod{2\pi}$ 
if the intersection of the target function boundary and the unit circle has two different points; or let $q_1=q_2=q$ where $\{q\}=\{x|(\cos \beta^*,\sin \beta^*)\cdot x+b^*=0\}\cap \sph(\bZero,1)$ if intersection of the target function boundary and the unit circle has only one point. 
Similarly, we denote by $\{p_1,p_2\}$ the intersection of of $K_{\beta,b}$ and the unit circle. Since $h-l\leq\arctan(1)=\frac{\pi}{4}$, the input data set must contain both positive points and negative points and hence, $K_{\beta,b}\cap \sph(\bZero,1)$ is not empty. We let $\{p_1,p_2\}=\{x|(\cos \beta,\sin \beta)\cdot x+b=0\}\cap \sph(\bZero,1)$ with $\beta-\theta((p_2-p_1), e_1)  = \frac{\pi}{2} \pmod{2\pi}$
if the intersection of the boundary of classifier $(\beta,b)$ and the unit circle has two different points; or let $p_1=p_2=p$ where $\{p\}=\{x|(\cos \beta,\sin \beta)\cdot x+b=0\}\cap \sph(\bZero,1)$ if intersection of the boundary of classifier $(\beta,b)$ and the unit circle has only one point. The definitions of $q_1,q_2,p_1,p_2$ are illustrated in Fig.~\ref{fig:lin2d}. Then at least one of $\{p_1,p_2\}$ is not in $A_z$ and w.l.o.g., assume that $p_2\notin A_z$. 
It is easy to check that $p_1,p_2$ must lie on the same side of the boundary of the target function, otherwise $\abs{\beta-\beta^*}=\theta((p_2-p_1),(q_2-q_1))>\arctan(z/2)$, which contradicts $h-l\leq {\arctan(z/2)}$. Then there exists a consistent classifier $h'=\ind{\inner{w}{x-p_2}\geq 0}$ with $w = (q_{12}-p_{22}, p_{21}-q_{11})$ (whose boundary is the line passing $q_1$ and $p_2$). Then let $\beta'$ denote the direction of $h'$, i.e., $\cos\beta' =\frac{q_{12}-p_{22}}{\norm{w}}$ and 
$\sin\beta' = \frac{p_{21}-q_{11}}{\norm{w}}$, 
and we have $\abs{\beta'-\beta^*}= \theta(p_2-q_1,q_2-q_1)> \arctan(z/2)$, which contradicts that $h-l\leq {\arctan(z/2)}$.
\begin{figure}[H]
    \centering
    \includegraphics[width = 0.5\textwidth]{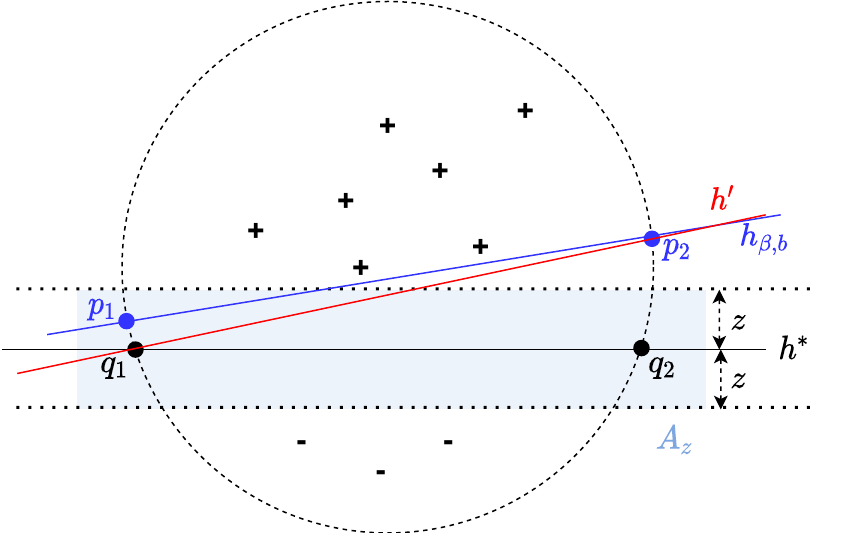}
    \caption{Illustration of $p_1,p_2,q_1,q_2,h'$.}
    \label{fig:lin2d}
\end{figure}
We let $z = f(\epsilon'')$. When $h-l\leq\arctan(f(\epsilon'')/2)$, the classifier can only make mistakes inside $A_{f(\epsilon'')}$ and thus the attackable rate is upper bounded by $2\epsilon''$. 
In addition, when $h-l>\arctan(f(\epsilon'')/2)$, the binary-search has searched to at most the $\floor{\log_2(\frac{2\pi}{\arctan (f(\epsilon'')/2)})}+1$-th depth, which leads to at most $2\epsilon'(\floor{\log_2(\frac{2\pi}{\arctan (f(\epsilon'')/2)})}+1)$ attackable rate. Combining these results together, when $m\geq \frac{4}{\epsilon'} \log \frac{2}{\delta} +  \frac{24}{\epsilon'} \log \frac{13}{\epsilon'}$, with probability at least $1-\delta$ over $S_\trn\sim \cD^m$, we have
  \begin{align}
    \atk(h^*, S_\trn, \cA) \leq &2\epsilon'\left(\floor{\log_2\left(\frac{2\pi}{\arctan (f(\epsilon'')/2)}\right)}+1\right) + 2\epsilon''\nonumber\\
    \leq& 2\epsilon'\log_2\left(\frac{4\pi}{\arctan (f(\epsilon'')/2)}\right) + 2\epsilon''\nonumber\\
    \leq &\log_2(\frac{4\pi}{\pi/8 (f(\epsilon'')\wedge 2)}) 2\epsilon' + 2\epsilon'' \label{eq:ss0}\\
    = &\log_2(\frac{32}{f(\epsilon'')\wedge 2})2\epsilon' + 2\epsilon'' \nonumber\,,
  \end{align}
  where Eq.~\eqref{eq:ss0} holds because $\arctan (x/2)\geq {\pi}/{4}$ when $x\geq 2$ and $\arctan (x/2)\geq {\pi x}/{8}$ when $x\in [0,2]$.
\end{proof}
\vspace{-20pt}
\section{Proof of Theorem~\ref{thm:linSVM}}\label{appx:linSVM}

To prove Theorem~\ref{thm:linSVM}, we first introduce a lemma on the behavior of uniform distribution on a unit sphere. 
\begin{lm}[Lemma~2.2 by \cite{ball1997elementary}]\label{lmm:ballconc}
    For any $a\in [0,1)$, for $x\sim \Unif(\sph^n(\bZero,1))$, with probability at least $1-e^{-na^2/2}$, we have $\inner{x}{e_1}\leq a$.
\end{lm}

\begin{proof}[of Theorem~\ref{thm:linSVM}] One essential hardness for robust learning of linear classifiers in high dimension is that for a fixed test instance $x_0$ on the sphere of a unit ball, with high probability over the selection of a set of training points from the uniform distribution over the sphere, every training instance has small component in the direction of $x_0$ as shown by Lemma~\ref{lmm:ballconc}. Taking advantage of this, the attacker can add a point (labeled differently from $x_0$) which is considerably closer to $x_0$ than all of the training instances, thus altering the behavior of SVM at $x_0$ as he wishes. In the following, we prove the theorem based on this idea. We divide the proof into three parts: a) the construction of the target function and the data distribution, b) the construction of the attacker and c) the analysis of the attackable rate. 
\paragraph{The target function and the data distribution.} 
The target function is $h^* = \ind{\inner{w^*}{x}\geq -\gamma/2}$ with $w^* = e_1$ and margin $\gamma= 1/8$. We define the marginal data distribution $\cD_{\cX}$ by putting probability mass $1-8\epsilon$ on $-e_1$ and putting probability mass $8\epsilon$ uniformly on the half sphere of a unit ball $\{x|\norm{x}=1,\inner{x}{e_1}\geq 0\}$. 
We let $\cD^+$ denote the uniform distribution over this positive half sphere. We draw $m$ i.i.d. training samples $S_\trn$ from $\cD$ and then let $S_\trn^+$ denote the positive training samples. Let $m^+ = \abs{S_\trn^+}$ denote the number of positive training samples.

\paragraph{The attacker.} For a given test instance $x_0\in \{x|\norm{x}=1,\inner{x}{e_1}\geq 0\}$, we define two base vectors $v_1 = e_1$ and $v_2 = \frac{x_0-\inner{x_0}{e_1}e_1}{\norm{x_0-\inner{x_0}{e_1}e_1}_2}$. Note that $v_2$ is well-defined almost surely. Then we define an attacker $\Adv$ as
\begin{align*}
    \Adv(h^*,S_\trn,x_0)=\begin{cases}
      \{(-v_2,1)\}&\text{if } m^+=0,\\
      \{(-\gamma v_1+\sqrt{1-\gamma^2}v_2,0)\}& \text{else}\,.
    \end{cases}
\end{align*}
Since $\inner{w^*}{v_2} = 0>-\frac{\gamma}{2}$ and $ \inner{w^*}{-\gamma v_1 + \sqrt{1-\gamma^2}v_2}=-\gamma <-\frac{\gamma}{2}$, $\Adv$ is a clean-label attacker.

\paragraph{Analysis.}
For $n\leq 128$, if $m<\frac{1}{8\epsilon}\vee \frac{e^{n/128}}{768\epsilon} = \frac{1}{8\epsilon} $ and $\eps < 1/16$, then $\PP{m^+=0}=(1-8\epsilon)^m\geq \frac{1}{4}$. Furthermore, if $m^+=0$, SVM can only observe instance-label pairs of $(-v_1,0)$ and $(-v_2,1)$ and then output $\hat{h}(x) =\ind{\inner{v_1-v_2}{x}\geq 0}$. Therefore, if $\inner{e_1}{x_0}<\frac{1}{\sqrt{2}}$, then $x_0$ is attackable. Therefore, for $m<\frac{1}{8\epsilon}$, we have 
  \begin{align*}
    \EEs{S_\trn\sim \cD^m}{\atk(h^*,S_\trn,\SVM)}\geq 8\epsilon\PPs{x\sim \cD^+}{\inner{x}{e_1}<\frac{1}{\sqrt{2}}} \PP{m^+=0} \geq \epsilon\,.
  \end{align*}
  For $n> 128$, if $m < \frac{1}{8\epsilon}$, the analysis above works as well. Else suppose $\frac{1}{8\epsilon} \leq m \leq \frac{e^{n/128}}{768\epsilon} $, we know that $\EE{m^+} = 8m\epsilon$, thus by Chernoff bounds, we have $\PP{m^+>32m\epsilon} \leq e^{-24m\epsilon}\leq e^{-3}$. Furthermore, by Lemma~\ref{lmm:ballconc} and the union bound, drawing $m_0$ i.i.d. samples $S_0\sim (\cD^{+})^{m_0}$, with probability at least $1-3m_0e^{-n/128}$, every instance $x\in S_0$ satisfies $\inner{x}{v_1}\leq \frac{1}{8}$ and $\inner{x}{v_2}\leq \frac{1}{8}$. Let $\cE$ denote the event of $\{\forall (x,y)\in S_\trn^+,\inner{x}{v_1}\leq \frac{1}{8},\inner{x}{v_2}\leq \frac{1}{8},1\leq m^+ \leq 32m\epsilon\}$. If $\cE$ holds, then there is a linear separator
  \[\left(\sqrt{\frac{1+\gamma}{2}}v_1 -\sqrt{\frac{1-\gamma}{2}}v_2\right)^\top x+\frac{1}{2}\sqrt{\frac{1+\gamma}{2}}+\frac{1}{16}\sqrt{\frac{1-\gamma}{2}}\geq 0\,,\]
  such that the distance between any point in $S_\trn \cup \Adv(h^*,S_\trn,x_0)$ and the linear separator is no smaller than $\frac{3}{8}-\frac{\sqrt{7}}{64}\geq \frac{1}{4}$. Hence the distance between the points and the seperator output by SVM is also no smaller than $\frac{1}{4}$. When the test instance $x_0$ satisfies $\inner{x_0}{v_1}\leq \frac{1}{8}$, we have $\norm{x_0-(-\gamma v_1+\sqrt{1-\gamma^2}v_2)}\leq \frac{1}{4}$ and then $x_0$ is misclassified as negative by SVM. Hence, we have
  \begin{align*}
    &\EEs{S_\trn\sim \cD^m}{\atk(h^*,S_\trn,\SVM)}\\
    \geq & 8\epsilon \EEs{S_\trn\sim \cD^m}{\PPs{x\sim \cD^+}{\SVM(S_\trn \cup \Adv(h^*,S_\trn,x),x)\neq h^*(x)}}\\
    \geq & 8\epsilon \EEs{x\sim \cD^+, S_\trn\sim \cD^m}{\ind{\forall (x',y')\in S_\trn^+, \inner{x'}{v_1}\leq \frac{1}{8},\inner{x'}{v_2}\leq \frac{1}{8},\inner{x}{v_1}\leq \frac{1}{8}}}\\
    \geq & 8\epsilon \EEs{x\sim \cD^+}{\EEs{S_\trn}{\ind{\cE}|x}\ind{\inner{x}{v_1}\leq \frac{1}{8}}}\\
    \geq & 8\epsilon (1-2e^{-n/128})(1-e^{-3}-e^{-8m\epsilon})(1-96m\epsilon e^{-n/128})\\
    \geq &\epsilon\,,
  \end{align*} 
  when $\frac{1}{8\epsilon} \leq m\leq \frac{e^{n/128}}{768\epsilon}$ and $n> 128$.  Thus in all we have shown that if $m< \frac{1}{8\epsilon} \vee \frac{e^{n/128}}{768\epsilon} $ then $\EEs{S_\trn\sim \cD^m}{\atk(h^*,S_\trn,\SVM)} > \epsilon$.
\end{proof}
\section{Proof of Theorem~\ref{thm:lblinmrg}}\label{appx:lb_linmrg}
\begin{proof}[of Theorem~\ref{thm:lblinmrg}]
The proof combines the idea of constructing a set of symmetrical poisoning instances in the proof of Theorem~\ref{thm:lblinear} and the idea that the training instances are far away from the test point in the proof of Theorem~\ref{thm:linSVM}. Again, we divide the proof into three parts as we did in the previous proofs.
\paragraph{The target function and the data distribution.}

We denote every point in $\R^n$ by $(x,z)$ for $x\in \R^{n-1}$ and $z\in \R$. The target function $h^*$ is selected uniformly at random from $\cH^*$, where $\cH^* = \{\ind{\inner{(jw^*,1)}{(x,z)}\geq j\gamma/2}|j\in\{\pm 1\}, w^*\in \sph^{n-1}(\bZero,1)\}$. Let $\gamma=\frac{1}{8}$. For target function $h^* = h_{w^*,j} = \ind{\inner{(jw^*,1)}{(x,z)}\geq j\gamma/2}$, the marginal data distribution $\cD_{w^*,j,\cX}$ puts probability mass $1-8\epsilon$ on the point $e_{n}$, then put the remaining $8\epsilon$ probability uniformly over the half sphere of a $(n-1)$-dimensional unit ball $\sph_{w^*,\gamma}\times\{0\}$, where $\sph_{w^*,\gamma}=\sph^{n-1}(\gamma w^*,1)\cap \{x|\inner{w^*}{x}\geq \gamma\}$. Then since every hypothesis in $\cH^*$ predicts $e_n$ positively, we only need to focus on the half sphere in the lower dimension. Note that the label of every point on the half sphere is determined by $j$. We sample $S_{\trn}\sim \cD^m_{w^*,j}$ and $S_{\trn,w^*,j}$ denote the samples on $\sph_{w^*,\gamma}\times\{0\}\times \cY$. Let $m_{w^*,j} = \abs{S_{\trn,w^*,j}}$.

\paragraph{The attacker.}

For any $u_1,u_2\in \R^{n-1}\setminus \{\bZero\}$, let $K_{u_1}(u_2) = \{x| \norm{u_1}^2 \inner{u_2}{x} - \inner{u_1}{u_2}\inner{u_1}{x} = 0 \}$ denote the homogeneous (passing through the origin) hyperplane perpendicular to the vector $u_2 - \inner{u_1}{u_2} \frac{u_1}{\norm{u_1}^2}$. For any given test instance $(x_0,0)$ with $x_0\in \sph_{w^*,\gamma}$, we define two base vectors $v_1=w^*$ and $v_2 = \frac{x_0-\inner{x_0}{w^*}w^*}{\norm{x_0-\inner{x_0}{w^*}w^*}}$. Note that $v_2$ is well-defined almost surely. Denote $K_{x_0} = K_{x_0}(v_1)$.
Let $x_\parallel= \inner{x}{v_1}v_1 + \inner{x}{v_2}v_2$ denote $x$'s component on the hyperplane defined by $v_1,v_2$ and $x_\perp = x-x_\parallel$ denote the component perpendicular to $v_1,v_2$ and then we define $\Rfl_{K_{x_0}}(x)\eqdef x_\perp + \frac{2\inner{x_\parallel}{x_0}}{\norm{x_0}^2}x_0  - x_\parallel$ as the reflection of $x$ through $K_{x_0}$.
Then we define an attacker $\Adv$ as
\begin{align*}
  &\Adv(h_{w^*,j},S_\trn,(x_0,0)) \\
  =&\begin{cases}
    \{((\Rfl_{K_{x_0}}(x),0),1-y)|((x,0),y)\in S_\trn\} & \text{if } \cE_1(w^*,S_\trn,x_0, m_{w^*,j})\,,\\
    \emptyset & \text{else}\,,
  \end{cases}
\end{align*}
where
\begin{align*}
    \cE_1(w^*,S_\trn,x_0, m_{w^*,j} ) = &\left\{\forall (x,0)\in S_{\trn,\cX}, \inner{x}{w^*}\leq \frac{1}{8} + \gamma, \inner{x}{\frac{x_0-\inner{x_0}{w^*}w^*}{\norm{x_0-\inner{x_0}{w^*}w^*}}}\leq \frac{1}{8}\right\} \\
    &\quad \cap \{\inner{x_0}{w^*}\leq \frac{1}{8}+\gamma\}\cap \{m_{w^*,j} \leq 32m\epsilon\}\,.
\end{align*}
Here $\cE_1(w^*,S_\trn,x_0, m_{w^*,j} )$ is thought as a condition to attack $x_0$. Then we show that $\Adv$ is a clean-label attacker.
If $\cE_1(w^*,S_\trn,x_0, m_{w^*,j})$ holds, we have
\begin{align*}
  &\inner{\Rfl_{K_{x_0}}(x)}{w^*}\\
  =& \inner{x_\perp + \frac{2\inner{x_\parallel}{x_0}}{\norm{x_0}^2}x_0  - x_\parallel}{v_1} \\
  =& \inner{x  - 2 \inner{x}{v_1}v_1 - 2\inner{x}{v_2}v_2 + 2 (\inner{x}{v_1}\inner{x_0}{v_1} + \inner{x}{v_2}\inner{x_0}{v_2}) \frac{x_0}{\norm{x_0}^2}}{v_1}\\
  =& -\inner{x}{v_1} + 2\inner{x}{v_1}\inner{x_0}{v_1}^2 \frac{1}{\norm{x_0}^2} + 2 \inner{x}{v_2}\inner{x_0}{v_2}\inner{x_0}{v_1} \frac{1}{\norm{x_0}^2}\\
  \leq& -\inner{x}{v_1} + 2(\frac{1}{8}+\gamma)^2 \inner{x}{v_1} + 2 \cdot \frac{1}{8} \inner{x_0}{v_1}\\
  \leq & \left(2(\frac{1}{8}+\gamma)^2-1\right)\gamma + \frac{1}{4}(\frac{1}{8}+\gamma)\\
  <& 0\,,
  \end{align*}
where the last inequality holds since $\gamma =\frac{1}{8}$. Therefore, $(\Rfl_{K_{x_0}}(x),0)$ is labeled different from $(x,0)$ and $\Adv$ is a clean-label attacker.

\paragraph{Analysis.}
Observe that the probabilistic construction of the target function and the data distribution along with the random sampling of a test instance and the training set can be viewed in an equivalent way: first drawing the number of training samples on the half sphere $m'$ from a binomial distribution $\Bin(m,8\epsilon)$; then on a fixed known half sphere, drawing a test instance and the training set with $m'$ samples on the half sphere and $m-m'$ samples on $e_n$; and finally randomly selecting a coordinate system to decide the position of the true sphere and selecting a $j$ to decide the labels of the training samples. Formally, let us fix a half sphere $\Gamma^{n-1}_+ = \{ x\in \Gamma^{n-1}(\bZero,1)|\inner{x}{e_1} \geq 0\}$ and then sample $m'\sim \Bin(m,8\epsilon)$, 
$t_0\sim \Unif(\Gamma^{n-1}_+)$ and 
$Q_\trn \sim \Unif(\Gamma^{n-1}_+)^{m'}$. 
We denote by $\cE_3(Q_\trn,t_0)$ the event of $\left\{\forall q\in Q_\trn, \inner{q}{e_1}\leq \frac{1}{8} , \inner{q}{\frac{t_0 - \inner{t_0}{e_1}e_1}{\norm{t_0 - \inner{t_0}{e_1}e_1}} }\leq \frac{1}{8}\right\}\cap \{\inner{t_0}{e_1}\leq \frac{1}{8}\} \cap\{m' = |Q_\trn| \leq 32m\epsilon\} $. Then we sample $T_{n-1}\sim \Unif(O(n-1))$, where $O(n-1)$ is the orthogonal group. Finally we sample $j\sim \Unif(\{\pm 1\})$. We denote by $R_{t_0}$ the linear isometry that reflects across the hyperplane $K_{\gamma e_1 + t_0}(e_1)$ in $\R^{n-1}$, i.e., $R_{t_0} u = \Rfl_{K_{\gamma e_1+t_0}(e_1)}(u)$, for any $u\in \R^n$. 

Conditional on $m', t_0$ and $Q_\trn$ sampled in the above process, we consider two different coordinate systems and $j$'s, which lead to two groups of random variables $(j, T_{n-1}, w^*, x_0, S_{\trn,\cX}, m_{w^*,j})$ and $(\wt{j}, \wt{T}_{n-1},\wt{w}^*,\wt{x}_0,\wt{S}_{\trn,\cX}, \wt{m}_{\wt{w}^*,\wt{j}})$. Here for any random variable in the first group, we add a tilde to represent the corresponding random variable in the second group.
\begin{itemize}
    \item In the first group, we have $j,T_{n-1},w^* = T_{n-1}e_1,  x_0 = T_{n-1}(\gamma e_1 + t_0),S_{\trn,\cX} = T_{n-1}(\gamma e_1 +  Q_\trn) \times \{0\} \cup \{e_n\}^{m-m'}$ and $m_{w^*,j} = m' $.
    \item In the second group, we let $\wt{j} = -j,\wt{T}_{n-1} = T_{n-1}R_{t_0}, \wt{w}^* = \wt{T}_{n-1}e_1, \wt{x}_0 =\wt{T}_{n-1}(\gamma e_1 +t_0) , \wt{S}_{\trn,\cX} = \wt{T}_{n-1}(\gamma e_1 + Q_\trn)\times \{0\}\cup \{e_n\}^{m-m'}$ and $\wt{m}_{\wt{w}^*,\wt{j}} = m'$.
\end{itemize}
The above two groups provide two ways of realizing the random process of selecting $h^*$, $(x_0,0)$ and $S_{\trn,\cX}$: 
namely, $h^* = h_{w^*,j} = \ind{\inner{(jw^*,1)}{(x,z)}\geq j\gamma/2}, (x_0,0), S_{\trn,\cX}$ and $h^* = h_{\wt{w}^*,\wt{j}} =\ind{\inner{(\wt{j}\wt{w}^*,1)}{(x,z)}\geq \wt{j}\gamma/2}, (\wt{x}_0,0),\wt{S}_{\trn,\cX}$. 
Let $\wt{S}_{\trn} = \{ (x, h_{\wt{w}^*,\wt{j}}(x))| x\!\in\!  \wt{S}_{\trn,\cX}  \}$ denote the data set of instances in $\wt{S}_{\trn,\cX}$ labeled by $h_{\wt{w}^*,\wt{j}}$. 
Note that $(w^*,j,S_\trn,x_0)$ and $(\wt{w}^*,\wt{j}, \wt{S}_\trn,\wt{x}_0)$ are identical in distribution and that $x_0= T_{n-1}(\gamma e_1 + t_0) = T_{n-1} R_{t_0}(\gamma e_1 + t_0)  = \wt{x}_0$. 
We now argue that $S_\trn\cup \Adv( h_{w^*,j},S_\trn, (x_0,0))$ and $\wt{S}_\trn\cup \Adv(h_{\wt{w}^*,\wt{j}},\wt{S}_\trn, (\wt{x}_0,0) )$ are identical conditional on $\cE_3( Q_\trn,\gamma e_1 + t_0)$. To prove this, we propose and prove the following three claims.

\paragraph{Claim I} $\cE_3(Q_\trn, t_0) \Leftrightarrow \cE_1(w^*, S_\trn,x_0, m_{w^*,j}) \Leftrightarrow \cE_1(\wt{w}^*,\wt{S}_{\trn,\cX}, \wt{x}_0, \wt{m}_{\wt{w}^*.\wt{j}})$.
\begin{proof}[of Claim I]
This is true since $T_{n-1}, R_{t_0}\in O(n-1)$, thus they keep all the inner product properties. In particular, for any $(x,0)\in S_\trn$, $x = T_{n-1} (\gamma e_1 + q)$ for some $q\in Q_\trn$ by definition of $S_\trn$. Furthermore,
\begin{gather*}
    \inner{x}{w^*} = \inner{T_{n-1}(\gamma e_1 + q)}{T_{n-1} e_1} = \gamma + \inner{q}{e_1}\,.
\end{gather*}
Thus $\inner{x}{w^*} \leq \frac{1}{8} + \gamma \Leftrightarrow \inner{q}{e_1}\leq \frac{1}{8} $. All the other equivalences can be derived similarly, thus omitted here.
\end{proof}

\paragraph{Claim II} For any homogeneous hyperplane $L_{u_1}\in \R^{n-1}$ with normal vector $u_1$, for any $u_2 \in \R^{n-1}$ , we have $\Rfl_{T_{n-1} L_{u_1}} (T_{n-1}u_2) = T_{n-1} \Rfl_{L_{u_1}}(u_2)$.
\begin{proof}[of Claim II]
We consider two cases. If $u_2 \in L_{u_1}$, then we have,
\begin{align*}
    \Rfl_{T_{n-1} L_{u_1}} (T_{n-1}u_2) = T_{n-1}u_2 = T_{n-1} \Rfl_{L_{u_1}}(u_2).
\end{align*}
Else if $u_2\notin L_{u_1}$, we 
denote by $u_3 = \Rfl_{T_{n-1} L_{u_1}} (T_{n-1}u_2)$. Thus $u_3$ is the only point such that $u_3\neq  T_{n-1}u_2$, $\inner{u_3 - T_{n-1}u_2}{T_{n-1}u_1} = 0 $ and 
$ \norm{u_3} = \norm{T_{n-1} u_2}$. 
These immediately give us $T_{n-1}^\top u_3 \neq u_2$, $\inner{T_{n-1}^\top u_3 - u_2}{u_1} = 0 $ and 
$\norm{T_{n-1}^\top u_3} = \norm{u_2}$, which means $T_{n-1}^\top u_3 = \Rfl_{L_{u_1}}(u_2)$. Thus
\begin{align*}
     \Rfl_{T_{n-1} L_{u_1}} (T_{n-1}u_2) = T_{n-1} T_{n-1}^\top u_3 = T_{n-1} \Rfl_{L_{u_1}}(u_2)\,,
\end{align*}
which completes the proof.
\end{proof}

\paragraph{Claim III} Conditional on $\cE_3(Q_\trn, t_0)$, the poisoned datasets $S_\trn\cup \Adv( h_{w^*,j},S_\trn, (x_0,0))$ and $\wt{S}_\trn\cup \Adv(h_{\wt{w}^*,\wt{j}},\wt{S}_\trn, (\wt{x}_0,0) )$ are identical.
\begin{proof}[of Claim III]
Denote by $K = K_{\gamma e_1 +t_0}(e_1)$ the homogeneous hyperplane perpendicular to $e_1 - \inner{\gamma e_1 + t_0}{e_1} \frac{\gamma e_1 + t_0}{\norm{\gamma e_1 + t_0}^2}$. 
Thus we have,
\begin{align*}
    T_{n-1} K &= \left\{ T_{n-1}x \bigg| \inner{x}{e_1 - \inner{\gamma e_1 + t_0}{e_1} \frac{\gamma e_1 + t_0}{\norm{\gamma e_1 + t_0}^2}} = 0 \right\}  \\
        &= \left\{ x \bigg| \inner{x}{T_{n-1} \left( e_1 - \inner{\gamma e_1 + t_0}{e_1} \frac{\gamma e_1 + t_0}{\norm{\gamma e_1 + t_0}^2}\right)} = 0 \right\}\\
        &= \left\{ x \bigg| \inner{x}{w^* - \inner{x_0}{w^*} \frac{x_0}{\norm{x_0}^2}} = 0 \right\}\\
        &= K_{x_0}(w^*)\,.
\end{align*}
By Claim I, we know that $\cE_3(Q_\trn, t_0)$ and $\cE_1(w^*, S_\trn,x_0, m_{w^*,j})$ are equivalent. Thus, conditional on $\cE_3(Q_\trn,t_0)$, 
    \begin{align}
        \Adv( h_{w^*,j},S_\trn, (x_0,0)) &= \{((\Rfl_{K_{x_0}(w^*)}(x),0),1-j)|((x,0),j)\in S_\trn\}\nonumber\\
        &=\{((\Rfl_{T_{n-1} K}(T_{n-1}(\gamma e_1+ q)),0),1-j)|q \in Q_\trn\}\nonumber \\
        &= \{(( T_{n-1}\Rfl_{K}(\gamma e_1+ q),0),1-j)|q \in Q_\trn\} \label{eq:applyclm2}\\
        &= \{(( T_{n-1}R_{t_0}(\gamma e_1+ q),0),1-j)|q \in Q_\trn\}\nonumber\\
        &= \wt{S}_\trn \setminus  \{(e_n,1)\}^{m-m'}\nonumber\,,
    \end{align}
    where Eq.~\eqref{eq:applyclm2} holds by applying Claim II. 
Similarly, for $\Adv(h_{\wt{w}^*,\wt{j}},\wt{S}_\trn, (x_0,0))$, the plane of reflection is $K_{x_0}(\wt{w}^*) =  K_{\wt{T}_{n-1}(\gamma e_1 + t_0)}(\wt{T}_{n-1}e_1) = \wt{T}_{n-1}K_{\gamma e_1 +t_0}(e_1) = \wt{T}_{n-1} K$ and
\begin{align}
    \Adv(h_{\wt{w}^*,\wt{j}},\wt{S}_\trn, (x_0,0)) &= \{((\Rfl_{K_{x_0}(\wt{w}^*)}(x),0),1-j)|((x,0),\wt{j})\in \wt{S}_\trn\} \nonumber\\
    &=\{((\Rfl_{\wt{T}_{n-1} K}(\wt{T}_{n-1}(\gamma e_1+ q)),0),j)|q \in Q_\trn\} \nonumber\\
    &= \{(( T_{n-1}(\gamma e_1+ q),0),j)|q \in Q_\trn\}\label{eq:applyclm22}\\
    &= S_\trn \setminus  \{(e_n,1)\}^{m-m'}\nonumber\,,
\end{align}
where Eq.~\eqref{eq:applyclm22} holds by applying Claim II and $\wt{T}_{n-1} = {T}_{n-1}R_{t_0}$.
Thus,
\begin{align*}
    S_\trn\cup \Adv( h_{w^*,j},S_\trn, (x_0,0)) &= S_\trn\cup \wt{S}_\trn \setminus  \{(e_n,1)\}^{m-m'} \\  &=\wt{S}_\trn\cup \Adv(h_{\wt{w}^*,\wt{j}},\wt{S}_\trn, (\wt{x}_0,0) )\,.
\end{align*}
\end{proof}
Now we have proved that $S_\trn\cup \Adv( h_{w^*,j},S_\trn, (x_0,0))$ and $\wt{S}_\trn\cup \Adv(h_{\wt{w}^*,\wt{j}},\wt{S}_\trn, (\wt{x}_0,0) )$ are identical conditional on $\cE_3( Q_\trn,\gamma e_1 + t_0)$. Hence in this case any algorithm will behave the same given the input data being either $S_\trn$ or $\wt{S}_\trn$. 
Let $\cE_2(\A,S_\trn,\Adv,h^*,x_0)$ denote the event $\A(S_\trn\cup \Adv(h^*,S_\trn,$ $(x_0,0)),(x_0,0))\neq h^*((x_0,0))$. Since $h_{w^*,j}((x_0,0)) \neq  h_{\wt{w}^*,\wt{j}}((x_0,0))$, 
then conditional on $\cE_3(Q_\trn,t_0)$, for any algorithm $\A$, 
we have 
$$\ind{\cE_2(\A,S_\trn,\Adv,h_{w^*,j},x_0)} = \ind{\neg\cE_2(\A,\wt{S}_\trn ,\Adv,h_{\wt{w}^*,\wt{j}},\wt{x}_0)}\,.$$ 

If $m <  \frac{1}{8\epsilon}$, then we have $ \mathbb{E}_{t_0,m', Q_\trn}[ \ind{\cE_3(Q_\trn,t_0)\cup\{m'=0\}}] \geq \PP{m'=0} = (1-8\epsilon)^m > \frac{1}{4}$ when $\epsilon\leq 1/16$. Else since $\EE{m'} = 8m\epsilon$, by Chernoff bounds, we have $\PP{m'>32m\epsilon} \leq e^{-24m\epsilon}\leq e^{-3}$. Furthermore, by Lemma~\ref{lmm:ballconc} and the union bound, drawing $m'$ i.i.d. samples $S_0\sim \Unif(\Gamma^{n-1}_+\times\{0\})^{m'}$, with probability at least $1-3m'e^{-\frac{n-1}{128}}$, every $(x,0)\in S_0$ satisfy $\inner{x}{e_1}\leq \frac{1}{8}$ and $\inner{x}{\frac{t_0 - \inner{t_0}{e_1}e_1}{\norm{t_0 - \inner{t_0}{e_1}e_1}} }\leq \frac{1}{8}$. Thus in all, we have, for any algorithm $\A$,
\begin{align}
&\mathbb{E}_{w^*,j,S_{\trn}\sim \cD_{w^*,j}^m, (x,y)\sim \cD_{w^*,j},\cA}[\atk(h^*,S_\trn, \cA)] \nonumber \\
=&\mathbb{E}_{w^*,j,S_{\trn}\sim \cD_{w^*,j}^m, (x,y)\sim \cD_{w^*,j},\cA}[\ind{\A(S_\trn\cup \Adv(h^*,S_\trn,x),x)\neq h^*(x)}] \nonumber \\
\geq& 8 \epsilon\mathbb{E}_{w^*,j,S_{\trn}, x_0\sim \Unif(\Gamma_{w^*,\gamma}) ,\A}[\ind{\cE_2(\A,S_\trn,\Adv,h_{w^*,j},x_0)}\nonumber\\
&\cdot \ind{\cE_1(w^*,S_\trn,x_0,m_{w^*,j})\cup \{m_{w^*,j}=0\}}]   \nonumber \\
=&8\epsilon \mathbb{E}_{t_0, m',Q_\trn}[ \ind{\cE_3(Q_\trn,t_0)\cup\{m'=0\}} \mathbb{E}_{T_{n-1},j,\cA} [ \ind{\cE_2(\A,S_\trn,\Adv,h_{w^*,j},x_0)}  ]]\nonumber\\
=&4\epsilon \mathbb{E}_{t_0,m', Q_\trn}[ \ind{\cE_3(Q_\trn,t_0)\cup\{m'=0\}} \mathbb{E}_{T_{n-1},j,\cA} [ \ind{\cE_2(\A,S_\trn,\Adv,h_{w^*,j},x_0)}  ]]\nonumber\\
&+4\epsilon \mathbb{E}_{t_0,m', Q_\trn}[ \ind{\cE_3(Q_\trn,t_0)\cup\{m'=0\}} \mathbb{E}_{T_{n-1},j,\cA} [ \ind{\neg\cE_2(\A,\wt{S}_\trn ,\Adv,h_{\wt{w}^*,\wt{j}},\wt{x}_0)}  ]] \label{myeq1}\\
=&4\epsilon \mathbb{E}_{t_0,m', Q_\trn}[ \ind{\cE_3(Q_\trn,t_0)\cup\{m'=0\}} \mathbb{E}_{T_{n-1},j,\cA} [ \ind{\cE_2(\A,S_\trn,\Adv,h_{w^*,j},x_0)}  ]]\nonumber\\
&+4\epsilon \mathbb{E}_{t_0,m', Q_\trn}[ \ind{\cE_3(Q_\trn,t_0)\cup\{m'=0\}} \mathbb{E}_{T_{n-1},j,\cA} [ \ind{\neg\cE_2(\A, S_\trn ,\Adv,h_{w^*,j},x_0)}  ]]\label{myeq2}\\
=& 4\epsilon \mathbb{E}_{t_0,m', Q_\trn}[ \ind{\cE_3(Q_\trn,t_0)\cup\{m'=0\}}] \nonumber\\
\geq&\begin{cases}
4 \epsilon (1 -2e^{-\frac{n-1}{128}})(1-e^{-3})(1-96m\epsilon e^{-\frac{n-1}{128}})& \text{when }m \geq \frac{1}{8\epsilon}\\
4 \epsilon(1-8\epsilon)^m &\text{when }m < \frac{1}{8\epsilon}
\end{cases}
\nonumber \\
>& \epsilon,\nonumber 
\end{align}
when $ m \leq \frac{e^\frac{n-1}{128}}{192\epsilon}$ and $n\geq 257$. Here Eq.~\eqref{myeq1} holds due to the fact that $\Adv$ will make $S_\trn\cup \Adv( h_{w^*,j},S_\trn, (x_0,0))$ and $\wt{S}_\trn\cup S_\Adv(h_{\wt{w}^*,\wt{j}},\wt{S}_\trn, (\wt{x}_0,0) )$ identical conditional on $\cE_3(Q_\trn,t_0)$ and Eq.~\eqref{myeq2} holds because $(w,j,S_\trn,x_0)$ is identical to $(\wt{w}^*,\wt{j}, \wt{S}_\trn,\wt{x}_0)$ in distribution. Thus in all, we have shown that for $n\geq 256$, if $ m \leq \frac{e^\frac{n-1}{128}}{192\epsilon}$ then for all algorithm $\cA$,   the expected attackable rate is $ \EEs{w,j,S_\trn\sim \cD^m}{\atk(h^*,S_\trn,\cA)} > \epsilon$. Thus there exists a target function $h^*\in\cH$ and a distribution $\cD$ over $D_{h^*}$ with margin $\gamma = 1/8$ such that $\EEs{S_\trn\sim \cD^m}{\atk_\cD(h^*, S_\trn,\A)}$ $> \epsilon$.
\end{proof}
\section{Proof of Theorem~\ref{thm:timproper}}\label{appx:improper_finite}
We first introduce two lemmas for the proof of the theorem.
\begin{lm}
\label{lmm:bennett}
For any hypothesis class $\cH$ with finite VC dimensional $d$, any distribution $\cD$, with probability at least $1-\delta$ over $S_\trn\sim \cD^m$, for all $h\in \cH$,
\begin{align*}
\err(h) - \err_{S_\trn}(h) \leq \sqrt{\frac{18d(1-\err_{S_\trn}(h))\err_{S_\trn}(h)\ln (em/\delta)}{m-1}} + \frac{15d\ln(em/\delta)}{m-1}\,.
\end{align*}  
\end{lm}

\begin{proof}
This lemma is a direct result of empirical Bennett's inequality (Theorem~6 by~\cite{maurer2009empirical}) and Sauer's lemma. Let $\Lambda(\cdot)$ denote the growth function of $\cH$. Then by empirical Bennett's inequality (Theorem~6 by~\cite{maurer2009empirical}), we have with probability at least $1-\delta$ over $S_\trn\sim \cD^m$,
\begin{align*}
      \err(h)-\err_{S_\trn}(h)\leq \sqrt{\frac{18(1-\err_{S_\trn}(h))\err_{S_\trn}(h)\ln(\Lambda(m)/\delta)}{m-1}} +\frac{15\ln (\Lambda(m)/\delta)}{m-1},\forall h\in\cH\,.
\end{align*}
By Sauer's lemma, $\Lambda(m)\leq (\frac{em}{d})^d$, which completes the proof.
\end{proof}

\begin{lm}
\label{lmm:subsample}
For any hypothesis class $\cH$ with finite VC dimensional $d$, a fixed data set $S$ with $m$ elements, realizable by some $h^*\in \cH$. Let $S_0$ be a set with size $m_0<m$ drawn from $S$ uniformly at random without replacement. Then with probability at least $1-\delta$, for all $h\in \cH$ with $\err_{S_0}(h) = 0$, we have
\begin{align*}
\err_{S}(h)\leq \frac{d\ln(em/d)+\ln(1/\delta)}{m_0}\,.
\end{align*} 
\end{lm}

\begin{proof}
For any $h\in \cH$, we have
  \begin{align*}
      &\PP{ \err_{S}(h)> \epsilon, \err_{S_{0}}(h)=0}
      \leq \frac{{m-k\choose m_0}}{{m\choose m_0}}
      \leq (1-k/m)^{m_0}\,,
  \end{align*}
  where $k = \ceil{\epsilon\cdot m}$.
By Sauer's lemma, $\Lambda(m)\leq (\frac{em}{d})^d$. Taking the union bound completes the proof.
\end{proof}

\begin{proof}[of Theorem~\ref{thm:timproper}]
Let $m=|S_\trn|$ be the number of training samples. Let $h_{i}=\cL(S^{(i)})$ denote the output hypothesis of block $i$. Let $N_{c} =\{i_1,\ldots,i_{n_c}\}\subseteq [10t+1]$ denote the set of index of non-contaminated blocks without poisoning points with $n_c = \abs{N_{c}}$. Each block has $m_0=\floor{\frac{m}{10t+1}}$ or $m_0 = \ceil{\frac{m}{10t+1}}$ data points (dependent on the actual number of poison points injected by the attacker) and at least $9t+1$ blocks do not contain any poison points, i.e., $n_c\geq 9t+1$. If a point $x$ is predicted incorrectly, then it is predicted incorrectly by more than $4t+1$ non-contaminated classifiers. Given training data $S_\trn$, for any $x\in\cX$, any $m_0$ and any $t$-point attacker $\Adv$ to make each block has $m_0$ points, we have
\begin{align*}
      &\PPs{\A}{\A(S_\trn\cup \Adv(S_\trn,h^*,x),x)\neq h^*(x)} \\
      =& \PPs{\A}{\sum_{i=1}^{10t+1}\ind{h_{i}(x)\neq h^*(x)}\geq 5t+1}\\
      \leq& \PPs{\A}{\sum_{i\in N_c}\ind{h_{i}(x)\neq h^*(x)}\geq 4t+1}\\
      \leq& \frac{1}{4t+1}\EEs{N_c}{\EEs{\A}{\sum_{i\in N_c}\ind{h_{i}(x)\neq h^*(x)}\big|N_c}}\\
      \leq& 2.5\EEs{N_c}{\EEs{\A}{\ind{h_{i_1}(x)\neq h^*(x)}|N_c}}\,.
\end{align*}
Notice here, if $m_0$ is fixed, 
the randomness of $\A$ can be regarded as selecting $N_c$ first and drawing $n_cm_0$ samples uniformly at random from $S_\trn$ without replacement to construct $S^{(i_1)}, S^{(i_2)}\,\ldots,S^{(i_{n_c})}$.
More specifically, conditioned on $N_c$, the randomness of $\cA$ on $h_{i_1}$ is only through drawing $S^{(i_1)}$, i.e., drawing $m_0$ samples without replacement from the clean training examples $S_\trn$. The important thing is that if $m_0$ is fixed, this distribution does not depend on the attacker. Hence,
\begin{align}\label{eq:atkvserr}
    &\atk(t,h^*,S_\trn,\A)\nonumber\\
    \leq& \EEs{(x,y)\sim \cD}{\sup_{\Adv} \PPs{\cA}{\A(S_\trn\cup \Adv(S_\trn,h^*,x),x)\neq h^*(x)}}\nonumber\\
    \leq & \EEs{(x,y)\sim \cD}{\sup_{\Adv: m_0=\floor{\frac{m}{10t+1}}} \PPs{\cA}{\A(S_\trn\cup \Adv(S_\trn,h^*,x),x)\neq h^*(x)}}\nonumber\\
    &+ \EEs{(x,y)\sim \cD}{\sup_{\Adv: m_0=\ceil{\frac{m}{10t+1}}} \PPs{\cA}{\A(S_\trn\cup \Adv(S_\trn,h^*,x),x)\neq h^*(x)}}\nonumber\\
    \leq& 2.5  \EEs{(x,y)\sim \cD}{  \sup_{\Adv: m_0=\floor{\frac{m}{10t+1}}} \EEs{N_c}{\EEs{\A}{\ind{h_{i_1}(x)\neq h^*(x)}|N_c}} }\\ 
    &+ 2.5  \EEs{(x,y)\sim \cD}{  \sup_{\Adv: m_0=\ceil{\frac{m}{10t+1}}} \EEs{N_c}{\EEs{\cA}{\ind{h_{i_1}(x)\neq h^*(x)}|N_c}} }\\
    \leq& 2.5\EEs{N_c}{\EEs{S^{(i_1)}}{\err\left(h_{i_1}\right)\Big|N_c}; m_0 =  \floor{\frac{m}{10t+1}}}\nonumber\\
    &+ 2.5\EEs{N_c}{\EEs{S^{(i_1)}}{\err\left(h_{i_1}\right)\Big|N_c}; m_0 =  \ceil{\frac{m}{10t+1}}}\,.
\end{align}
In the following, we will bound the error of $h_{i_1}$ for each value of $m_0$. Let $\cE$ denote the event of $\err_{S_\trn}(h_{i_1})\leq \frac{(d+1)\ln(em/d)}{m_0}$ and by Lemma~\ref{lmm:subsample} we have $\PPs{\A}{\neg \cE|N_c}\leq \frac{d}{em}$. Then with probability at least $1-\delta$ over the choice of $S_\trn$, for each fixed value of $m_0$, we have
\begin{align}
    &\EEs{\cA}{\err(h_{i_1})|N_c}\nonumber\\
    = & \EEs{\cA}{\err(h_{i_1})\ind{\cE}|N_c} + \EEs{\cA}{\err(h_{i_1})\ind{\neg \cE}|N_c}\nonumber\\
    = & \EEs{\cA}{(\err(h_{i_1})-\err_{S_\trn}(h_{i_1})+\err_{S_\trn}(h_{i_1}))\ind{\cE}|N_c} +  \PPs{\cA}{\neg \cE|N_c}\nonumber\\
    \leq &\sqrt{\frac{18d\ln (em/\delta)(d+1)\ln(em/d)}{(m-1)m_0}} + \frac{15d\ln(em/\delta)}{m-1} +\frac{(d+1)\ln(em/d)}{m_0}+\frac{d}{em}\label{eq:applybennett} \\
    \leq &\frac{1}{2} \left(  \frac{6d\ln(em/\delta)}{m-1} + \frac{6d\ln(em/d)}{m_0} \right) + \frac{15d\ln(em/\delta)}{m-1} +\frac{(d+1)\ln(em/d)}{m_0}+\frac{d}{em}\nonumber \\
    \leq & \frac{24d\ln(em)}{m_0} + \frac{19d\ln(1/\delta)}{10tm_0}\,, \label{eq:errimproper}
\end{align}
where Eq.~\eqref{eq:applybennett} applies Lemma~\ref{lmm:bennett}. Then when $m_0\geq  \frac{960 d}{\epsilon} \ln \frac{2640 e td}{\epsilon}  + \frac{19d\ln(1/\delta)}{\epsilon t}  $, we have that $\EEs{\cA}{\err(h_{i_1})|N_c}\leq \frac{24d\ln(em)}{m_0} + \frac{19d\ln(1/\delta)}{10tm_0}\leq 0.2\epsilon$. Combined with Eq.~\eqref{eq:atkvserr}, we have that when $m\geq (10t+1)(\frac{960 d}{\epsilon} \ln \frac{2640 e td}{\epsilon} + \frac{19d\ln(1/\delta)}{\epsilon t}+1)  $, the $t$-point attackable rate is $\atk(t,h^*,S_\trn,\A)\leq \epsilon$.
\end{proof}
\section{Proofs and discussions for $(t,\epsilon,\delta)$-robust proper learners}\label{appx:proper_finite}
\subsection{Proof of Theorem~\ref{thm:tpropproj}}
\begin{proof}
Similar to the proof of Theorem~\ref{thm:timproper}, we let $m=|S_\trn|$ be the number of training samples, and let $N_{c} =\{i_1,\ldots,i_{n_c}\}\subseteq [10tk_p+1]$ denote the set of index of blocks without poisoning points. Each block has $m_0=\floor{\frac{m}{10tk_p+1}}$ or $m_0=\ceil{\frac{m}{10tk_p+1}}$ data points and at least $t(10k_p-1)+1$ blocks do not contain any attacking points, i.e., $n_c\geq t(10k_p-1)+1$. 
Given any fixed $x\in\cX$, if $\sum_{i=1}^{10tk_p+1}\ind{h_{i}(x)\neq h^*(x)}\leq 10t < \frac{10tk_p+1}{k_p}$, then $x\in \cX_{\cH',k_p}$, thus $\hat{h}(x) = \Major(\cH',x) = h^*(x)$. Thus we have, given training data $S_\trn$, for any $x\in\cX$, 
any $m_0$ and any $t$-point attacker $\Adv$ to make each block has $m_0$ points, we have
\begin{align*}
      &\PPs{\A}{\A(S_\trn\cup \Adv(S_\trn,h^*,x),x)\neq h^*(x)} \\
      \leq& \PPs{\A}{\sum_{i=1}^{10tk_p+1}\ind{h_{i}(x)\neq h^*(x)}\geq 10t+1}\\
      \leq& \PPs{\A}{\sum_{i\in N_c}\ind{h_{i}(x)\neq h^*(x)}\geq 9t+1}\\
      \leq& \frac{1}{9t+1}\EEs{N_c}{\EEs{\A}{\sum_{i\in N_c}\ind{h_{i}(x)\neq h^*(x)}\Big|N_c}}\\
      \leq& \frac{10}{9}k_p\EEs{N_c}{\EEs{\A}{\ind{h_{i_1}(x)\neq h^*(x)}\Big|N_c}}\,,
\end{align*}
which indicates 
\begin{align*}
    \atk(t,h^*,S_\trn,\A)&\leq \frac{10}{9}k_p \EEs{N_c}{\EEs{\cA}{\err\!\left(h_{i_1}\right)|N_c}; m_0\!=\!\floor{\frac{m}{10tk_p+1}}} \\
    &\quad + \frac{10}{9}k_p\EEs{N_c}{ \EEs{\cA}{\err\!\left(h_{i_1} \right)|N_c};m_0\!=\!\ceil{\frac{m}{10tk_p+1}}}. 
\end{align*}
Then we bound $\EEs{\cA}{\err(h_{i_1})|N_c}$ in the same way as the proof of Theorem~\ref{thm:timproper}. Following the same calculation process of Eq.~\eqref{eq:errimproper}, we have with probability at least $1-\delta$, $\EEs{\cA}{\err(h_{i_1})|N_c}\leq \frac{24d\ln(em)}{m_0} + \frac{19d\ln(1/\delta)}{10k_ptm_0}$ by using Lemma~\ref{lmm:subsample}
and Lemma~\ref{lmm:bennett}. Then when $m_0\geq \frac{960 dk_p}{\epsilon} \ln \frac{2640 e tdk_p^2}{\epsilon} + \frac{19d\ln(1/\delta)}{\epsilon t} $, we have $\EEs{\cA}{\err(h_{i_1})|N_c}\leq \frac{24d\ln(em)}{m_0} + \frac{19d\ln(1/\delta)}{10k_ptm_0}\leq  \frac{0.2}{k_p}\epsilon $ . Therefore, we have that when $m\geq (10k_pt+1)(\frac{960 dk_p}{\epsilon} \ln \frac{2640 e tdk_p^2}{\epsilon} + \frac{19d\ln(1/\delta)}{\epsilon t}+1)$, the $t$-point attackable rate is $\atk(t,h^*,S_\trn,\A)\leq \epsilon$.
\end{proof}

\subsection{A proper learner for hypothesis class with no limitation over $k_p$}
\begin{algorithm}[H] \caption{A robust proper learner for $t$-point attacker}\label{alg:propert}
  {\begin{algorithmic}[1]
      \STATE \textbf{input}: a proper ERM learner $\cL$, data $S$
      \STATE uniformly at random pick $\floor{\frac{\abs{S}}{3t/\epsilon}}$ points $S_0$ from $S$ \emph{with replacement}
      \STATE \textbf{return} $\cL(S_0)$
  \end{algorithmic}}
\end{algorithm}

\begin{thm}
  For any hypothesis class with VC dimension $d$, with any proper ERM learner $\cL$, Algorithm~\ref{alg:propert} can $(t,\epsilon,\delta)$-robustly learn $\cH$ using $m$ samples where
  \[
      m=O\left(\frac{dt}{\epsilon^2} \log \frac{d}{\epsilon}+\frac{d}{\epsilon}\log \frac{1}{\delta}\right) \,.
  \]
\end{thm}
\begin{proof}
  Let $\cE$ denote the event that every point in $S_0$ is selected from the training data $S_\trn$. Let $m=\abs{S_\trn}$ and $h_0=\cL(S_0)$. Let us denote the size of $S_0$ by $m_0 = \floor{\frac{\abs{S}}{3t/\epsilon}}$, which can be $\floor{\frac{m}{3t/\epsilon}}$ or $\ceil{\frac{m}{3t/\epsilon}}$. Since $\PPs{\A}{\cE} \geq (1-\frac{t}{3tm_0/\epsilon})^{m_0}\geq 1-\frac{\ln 4}{3}\epsilon$, we have $\PPs{\A}{\neg \cE}\leq \frac{\ln 4}{3}\epsilon  $. Then for any $t$-point attacker $\Adv$, we have
  \begin{align*}
    &\PPs{\A}{\A(S_\trn\cup \Adv(h^*,S_\trn,x),x)\neq h^*(x)}\\
    = &\PPs{\A}{h_0(x)\neq h^*(x)\cap \cE}+\PPs{\A}{h_0(x)\neq h^*(x)\cap \neg \cE}\\
    \leq &\PPs{\A}{h_0(x)\neq h^*(x)| \cE}+\PPs{\A}{\neg \cE}\\
    \leq &\PPs{\A}{h_0(x)\neq h^*(x)| \cE}+ \frac{\ln 4}{3}\epsilon\,,
  \end{align*}
  which indicates $\atk(t,h^*,S_\trn,\A)\leq \EEs{\cA}{\err(h_{0})\Big|\cE; m_0 \!=\! \floor{\frac{m}{3t/\epsilon}}}+\EEs{\cA}{\err(h_{0})\Big|\cE; m_0 \!=\! \ceil{\frac{m}{3t/\epsilon}}}$ $+ \frac{\ln 4}{3}\epsilon$. Conditioned on $\cE$, $S_0$ is a set of i.i.d. samples uniformly drawn from $S_\trn$. By classic uniform convergence bound, $\err_{S_\trn}(h_0)\leq \frac{2}{m_0}(d\log(2em_0/d)+\log(2/\delta_0))$ with probability at least $1-\delta_0$ over the choice of $S_0$ (for a fixed $S_\trn$). Let $\cE_1$ denote the event of $\err_{S_\trn}(h_0)\leq \frac{2}{m_0}(d+1)\log(2em_0/d)$ and it is easy to check that $\PPs{\A}{\neg \cE_1}\leq \frac{d}{em_0}$. Similar to the proof of Theorem~\ref{thm:timproper}, with probability at least $1-\delta$, we have 
  \begin{align}
      &\EEs{\cA}{\err(h_{0})|\cE}\nonumber \\
      = & \EEs{\cA}{\err(h_{0})\ind{\cE_1}|\cE} + \EEs{\cA}{\err(h_{0})\ind{\neg \cE_1}|\cE}\nonumber\\
      \leq &\sqrt{\frac{36d\ln (em/\delta)(d+1)\log(2em_0/d)}{(m-1)m_0}} + \frac{15d\ln(em/\delta)}{m-1} +\frac{2}{m_0}(d+1)\log\frac{2em_0}{d}+\frac{d}{em_0}\label{eq:applybennett2}\\
      \leq & \frac{1}{2}  \left(    \frac{6d\ln(em/\delta)}{m-1} + \frac{12d\ln(2em_0/d)}{m_0}    \right) +  \frac{15d\ln(em/\delta)}{m-1} +   \frac{6d \ln (2em_0/d)}{m_0}   +\frac{d}{em_0} \nonumber\\
    \leq & \frac{13d\ln(2em_0/d)}{m_0} + \frac{18d\ln(em/\delta)}{m-1}\nonumber\\
    \leq & \frac{31d\ln(2em_0)}{m_0} + \frac{18d\ln(1/\delta)}{(3t/\epsilon-1)m_0}\,,\nonumber
  \end{align}
  where Eq.~\eqref{eq:applybennett2} adopts Lemma~\ref{lmm:bennett}. When $m_0\geq  \frac{1120d}{\epsilon} \ln  \frac{560e d}{\epsilon} + \frac{72 d\ln (1/\delta)}{t} $,  $\EEs{\cA}{\err(h_{0})|\cE} \leq 0.25\epsilon$. Hence, with probability at least $1-\delta$, the $t$-point attackable rate is $\atk(t,h^*,S_\trn,\A)\leq \epsilon$ by using $m$ training samples where
  \[m= \frac{3t}{\epsilon}\left(\frac{1120d}{\epsilon} \ln  \frac{560e d}{\epsilon} + \frac{72 d\ln (1/\delta)}{t}+1\right)\,.\]
\end{proof}

\section{Proof of Theorem~\ref{thm:lbfnt}}\label{appx:lb_finite}
\begin{proof}[of Theorem~\ref{thm:lbfnt}]
Now we show that for any sample size $m>0$, there exists a hypothesis class $\cH$ with VC dimension $5d$, a target function $h^*\in \cH$ and a data distribution $\cD$ on $D_{h^*}$ such that $\EEs{S_\trn\sim \cD^m}{\atk(t,h^*,S_\trn, \A)}\geq \min(\frac{3td}{64m},\frac{3}{8})$. We start with proving this statement in the base case of $d=1$ and then extend it to $d\geq 2$. We divide the proof into four parts: a) construction of the hypothesis class, the target function and the data distribution in $d=1$, b) computation of the VC dimension of the hypothesis class, c) construction of the attacker, and d) generalization to $d\geq 1$.
  
\paragraph{The hypothesis class, the target function and the data distribution.}
  
We denote by $\sph = \sph^3(\bZero,1)$ the sphere of the $3$-d unit ball centered at the origin. First, consider a base case where the domain $\cX = \sph\cup \bZero$, which is the union of the sphere of a unit ball centered at the origin and the origin. 
For any point $q\in \sph$, We let $C_q=\sph^3(q,1)\cap \sph$ denote the circle of intersection of the sphere of two unit balls. Then we define $h_{q,1}=\ind{C_q}$, which only classifies the circle $C_q$ positive and $h_{q,0}=\ind{\sph\setminus C_q}$ only classifies the circle and the origin negative. Our hypothesis class is $\cH = \{h_{q,j}|q\in \sph, j\in\{0,1\}\}$. We draw our target $h^*$ uniformly at random from $\cH$, which is equivalent to drawing $q\sim \Unif(\sph)$ and $j\sim \Ber(1/2)$. 
  The marginal data distribution $\cD_{q,j,\cX}$ puts probability mass $\zeta\in (0,\frac{t}{8m}]$ uniformly on the circle $C_q$ and puts the remaining probability mass on $\bZero$, where the value of $\zeta$ is determined later. We draw $S_\trn\sim \cD_{q,j}^m$. 
  

\paragraph{The VC dimension of the hypothesis class.} Then we show that the VC dimension of $\cH$ is $5$. Since all classifiers in $\cH$ will classify $\bZero$ as negative, $\bZero$ cannot be shattered and thus, we only need to find shattered points on the sphere. Then we show that $\cH$ can shatter $5$ points. It is not hard to check that the following set of $5$ points can be shattered: $\left\{(\frac{1}{2}, \frac{\sqrt{3}}{2} \cos(\frac{2k\pi}{5}), \frac{\sqrt{3}}{2} \sin(\frac{2k\pi}{5})) \right\}_{k=1}^5$.

Then we show that $\cH$ cannot shatter $6$ points. For any $6$ points $P=\{p_1,\ldots,p_6\}$, if the $6$ points can be shattered, then for any subset $P_1\subseteq P$ with size $3$, there exists a hypothesis classifying $P_1$ as $0$s and $P\setminus P_1$ as $1$s. That is, there exists a circle of radius $\frac{\sqrt{3}}{2}$ such that either only $P_1$ is on the circle or only $P\setminus P_1$ is on the circle. Then we claim that no $4$ points can be on a circle of radius $\frac{\sqrt{3}}{2}$. If there are $4$ points, w.l.o.g., $\{p_1,p_2,p_3,p_4\}$ on a circle of radius $\frac{\sqrt{3}}{2}$, then $\{p_i, p_5, p_6\}$ has to be on a circle $C_i$ of radius $\frac{\sqrt{3}}{2}$, where $1\leq j\neq i\leq 4$, $p_j\notin C_i$. But since the radius is fixed, there are only two different circles passing through $\{p_5,p_6\}$. Hence, there exists $1\leq i\neq j \leq 4$ such that $C_i = C_j$, which contradicts that $p_j\notin C_i$. 
  
Then w.l.o.g., if $\{p_1,p_2,p_3\}$ is on a circle of radius $\frac{\sqrt{3}}{2}$. Consider $\{p_1,p_2,p_4\}$ and $\{p_3,p_5,p_6\}$, if $\{p_1,p_2,p_4\}$ is on a circle of radius $\frac{\sqrt{3}}{2}$, then $\{p_3,p_4,p_5,p_6\}$ is on a circle of radius $\frac{\sqrt{3}}{2}$ (to label $\{p_1,p_2\}$ different from $\{p_3,p_4,p_5,p_6\}$); if $\{p_3,p_5,p_6\}$ is on a circle of radius $\frac{\sqrt{3}}{2}$, then there are three sub-cases: $\{p_1,p_3,p_5\}$ is on a circle of radius $\frac{\sqrt{3}}{2}$, $\{p_2,p_3,p_5\}$ is on a circle of radius $\frac{\sqrt{3}}{2}$ and both $\{p_2,p_4,p_6\},\{p_1,p_4,p_6\}$ are on two circles of radius $\frac{\sqrt{3}}{2}$. For the first case, $\{p_1,p_2,p_4,p_6\}$ is on a circle of radius $\frac{\sqrt{3}}{2}$ (to label $\{p_3,p_5\}$ different from $\{p_1,p_2,p_4,p_6\}$). For the second case, similarly $\{p_1,p_2,p_4,p_6\}$ is on a circle of radius $\frac{\sqrt{3}}{2}$. For the third case, $\{p_1,p_2,p_3,p_5\}$ is on a circle of radius $\frac{\sqrt{3}}{2}$. Therefore, any $6$ points cannot be shattered.

\paragraph{The attacker.}
We adopt the reflection function $m_{x_0}(\cdot)$ defined in the proof of Theorem~\ref{thm:lblinear} where $m_{x_0}(x) = {2\inner{x_0}{x}}x_0-x$ for $x\in \sph$. For $S_\trn\sim \cD_{q,j}^m$, we let $S_q=C_q\cap S_{\trn,\cX}$ denote the training instances in $C_q$ (with replicants) and we further define $m_{x_0}(S_\trn) = \{ (m_{x_0}(x),1-y)|(x,y)\in S_q\times \cY \}$, and let
\begin{align*}
    \Adv(h^*,S_\trn,x_0)=\begin{cases}
      m_{x_0}(S_\trn)& \text{if } x_0\notin S_{\trn,\cX}, \abs{S_q}\leq t\,,\\
      \emptyset & \text{else}\,.
    \end{cases}
\end{align*}
If $x_0\notin S_{\trn,\cX}$, then $h_{q,j}$ is consistent with $S_\trn\cup \Adv(h_{q,j},S_\trn,x_0)$. That is, $\Adv(h_{q,j},S_\trn,x_0)$ is clean-labeled.

\paragraph{Analysis.}
Due to the construction, we have
\begin{align*}
      \EEs{S_\trn\sim \cD_{q,j}^m}{\abs{S_q}} = m\zeta\,.
\end{align*}
Then by Markov's inequality, we have
\begin{align*}
      \PPs{{S_\trn\sim \cD_{q,j}^m}}{\abs{S_q}\geq t}\leq \frac{m\zeta}{t}< \frac{1}{4}\,.
\end{align*}
Let $\cE_1(\A,\Adv,h_{q,j},S_\trn,x_0)$ denote the event of $\{\A(S_\trn\cup \Adv(h_{q,j},S_\trn,x_0), x_0)\neq h_{q,j}(x_0)\}$ and let $\cE_2(S_\trn,x_0,q)$ denote the event of $\{\abs{S_q}\leq t \cap x_0\notin S_{\trn,\cX}\}$. It is easy to check that $\cE_2(S_\trn,x_0,q) =\cE_2(m_{x_0}(S_\trn),x_0,m_{x_0}(q))$. Besides, conditional on $\cE_2(S_\trn,x_0,q)$, we have the poisoned data set $S_\trn\cup \Adv(h_{q,j},S_\trn,x_0) = m_{x_0}(S_\trn)\cup \Adv(h_{m_{x_0}(q),1-j},m_{x_0}(S_\trn),x_0)$ and thus, any algorithm $\A$ will behave the same at the test instance $x_0$ no matter whether the training set is $S_\trn$ or $m_{x_0}(S_\trn)$. Since $h_{q,j}(x_0)\neq h_{m_{x_0}(q),1-j}(x_0)$, we have $\ind{\cE_1(\A,\Adv,h_{q,j},S_\trn,x_0)} = \ind{\neg\cE_1(\A,\Adv,h_{m_{x_0}(q),1-j},m_{x_0}(S_\trn),x_0)}$ conditional on $\cE_2(S_\trn,x_0,q)$. 
Let $f_{q}(x)$ denote the probability density function of $\Unif(C_q)$ and then we have $f_{q}(x) = f_{m_{x_0}(q)}(m_{x_0}(x))$. 
For any fixed $x_0$, the distributions of $q$ and $m_{x_0}(q)$ and the distributions of $j$ and $1-j$ are the same respectively. 
Since $S_\trn$ are samples drawn from $\cD_{q,j}^m$, $m_{x_0}(S_\trn)$ are actually samples drawn from $\cD_{m_{x_0}(q),1-j}^m$. Then we have
\begin{align*}
    &\EEs{h^*\sim \Unif(\cH),S_\trn\sim \cD^m}{\atk_\cD(t,h^*,S_\trn, \A)}\\
    =&\zeta\EEs{q\sim \Unif(\sph), j\sim\Ber(\frac{1}{2}),S_\trn\sim \cD_{q,j}^m,x\sim \Unif(C_q),\A}{\ind{\cE_1(\A,\Adv,h_{q,j},S_\trn,x)}}\\
    \geq &\zeta\EEs{q\sim \Unif(\sph), j\sim\Ber(\frac{1}{2}),S_\trn\sim \cD_{q,j}^m,x\sim \Unif(C_q),\A}{\ind{\cE_1(\A,\Adv,h_{q,j},S_\trn,x)\cap \cE_2(S_\trn,x,q)}}\\
    =&\zeta\int_{x\in \sph}\EEs{q\sim \Unif(\sph), j\sim\Ber(\frac{1}{2}),S_\trn\sim \cD_{q,j}^m,\A}{f_{q}(x)\ind{\cE_1(\A,\Adv,h_{q,j},S_\trn,x)\cap \cE_2(S_\trn,x,q)}}dx\\
    =&\zeta\int_{x\in \sph}{\E}_{q\sim \Unif(\sph), j\sim\Ber(\frac{1}{2}),S_\trn\sim \cD_{q,j}^m,\A} [f_{m_x(q)}(x)\ind{\neg \cE_1(\A,\Adv,h_{m_x(q),1-j},m_x(S_\trn),x)}\\
    &\cdot \ind{\cE_2(m_x(S_\trn),x,m_{x}(q))}]dx\\
    =&\zeta\int_{x}\EEs{q\sim \Unif(\sph), j\sim\Ber(\frac{1}{2}),S_\trn\sim \cD_{q,j}^m,\A}{f_{q}(x)\ind{\neg \cE_1(\A,\Adv,h_{q,j},S_\trn,x)} \cdot \ind{\cE_2(S_\trn,x,q)}}dx\\
    =&\frac{\zeta}{2}\int_{x}\EEs{q\sim \Unif(\sph), j\sim \Ber(\frac{1}{2}),S_\trn\sim \cD_{q,j}^m}{f_{q}(x)\ind{\cE_2(S_\trn,x,q)}}dx\\
    >&\frac{3\zeta}{8}\,,
\end{align*}
which completes the proof for $d=1$ by setting $\zeta = \min(\frac{t}{8m},1)$. 
  
\paragraph{Extension to general $d\geq 1$.}
To extend the base case to $d>1$, we construct $d$ separate balls and repeat the above construction on each ball individually. For $i\in[d]$, let $\sph_i = \sph^3(3ie_1,1)$ denote the sphere of a ball with radius $1$ centered at $3ie_1$. 
Consider the domain $\cX = \cup_{i\in [d]}\sph_i\cup \{\bZero\}$ as the union of $d$ non-overlapping unit balls and the origin. 
For $q_i\in \sph_i$, let $h^1_{q_i}=\ind{\sph^3(q_i,1)\cap \sph_i}$ denote the hypothesis classifying only points on the circle of $\sph^3(q_i,1)\cap \sph_i$ positive and $h^0_{q_i}=\ind{\sph_i\setminus \sph^3(q_i,1)}$ denote the hypothesis classifying only points on $\sph_i$ positive except the circle $\sph^3(q_i,1)\cap \sph_i$. Let $h^s_{q_1,\ldots,q_{d}}=\sum_{i\in[d]}h^{s_i}_{q_i}$, where $s\in\{0,1\}^{d}$ denote the hypothesis combining all $d$ balls and $\cH = \{h^s_{q_1,\ldots,q_{d}}|q_i\in \sph_i,\forall i\in [d],s\in\{0,1\}^{d}\}$. We have the VC dimension of $\cH$ is $5d$. Our target function is selected uniformly at random from $\cH$ and similar to the case of $d=1$, we assign probability $\zeta = \min(\frac{1}{d},\frac{t}{8m})$ to each circle on the balls and the remaining probability mass on the origin. Since every ball is independent with other balls and thus, we have $\EEs{h^*\sim \Unif(\cH),S_\trn\sim \cD^m}{\atk_\cD(t, h^*, S_\trn, \A)}> \frac{3d\zeta}{8}=  \min(\frac{3td}{64m},\frac{3}{8})$. 
  
 In all, there exists a target function $h^*\in \cH$ and a data distribution $\cD$ over $D_{h^*}$ such that $\EEs{S_\trn}{\atk_\cD(t, h^*, S_\trn, \A)}> \epsilon$ when $m<\frac{3td}{64\epsilon}$ for $\epsilon\leq \frac{3}{8}$.
\end{proof}
\chapter{Learning under Transformation Invariances}\label{app:transformation}
\section{Proof of Theorem~\ref{thm:inv-da-ub}}\label{app:inv-da-ub}
\begin{proof}
    Let $d = \vcao(\cH,\cG)$. 
    According to the definition of DA and the invaraintly realizable setting, given the input $S\sim \cD^m$, the output of DA $\hat h\in \cH$ satisfies $\err_{\cG S}(\hat h) = 0$, i.e., $\hat h(x) = y$ for all $(x,y)\in \cG S$.
    Consider two sets $S$ and $S'$ of $m$ i.i.d. samples drawn from the data distribution $\cD$ each.
    We denote $A_{S}$ the event of $\{\exists h\in \cH, \err_\cD(h)\geq \epsilon, \err_{\cG S}(h)=0\}$ and $B_{S,S'}$ the event of $\{\exists h\in \cH, \err_{S'}(h)\geq \frac{\epsilon}{2},  \err_{\cG S}(h)=0\}$.
    By Chernoff bound, we have $\Pr(B_{S,S'})\geq \Pr(A_S)\cdot \Pr(B_{S,S'}|A_S)\geq \frac{1}{2}\Pr(A_S)$ when $m\geq \frac{8}{\epsilon}$.
    The sampling process of $S$ and $S'$ is equivalent to drawing $2m$ i.i.d. samples and then randomly partitioning into $S$ and $S'$ of $m$ each.
    For any fixed $S''$, for any $h$ with $\err_{S'}(h)\geq \frac{\epsilon}{2}$ and $\err_{\cG S}(h)=0$, if $h$ misclassifies some $(x,y)\in S''$, then all examples in the orbit of $x$, i.e., $\cG \{(x,y)\} \cap S''$, must go to $S'$. 
    Now to prove the theorem, we divide $S''$ into two categories in terms of the number of examples in each orbit. 
    Let $R_1 = \{x|\abs{\cG x \cap S''_{\cX}}\geq \log^2 m, x\in S''_\cX\}$ and $R_2 = S''_{\cX}\setminus R_1$. 
    For any $h$ making at least $\frac{\epsilon m}{2}$ mistakes in $S''_\cX$, $h$ either makes at least $\frac{\epsilon m}{4}$ mistakes in $R_1$ or makes at least $\frac{\epsilon m}{4}$ mistakes in $R_2$.
    Then let $\cH_0\subset \cH$ denote the set of hypotheses making at least $\frac{\epsilon m}{2}$ mistakes in $S''_\cX$ and divide $\cH_0$ into two sub-classes as follows.
    \begin{itemize}
        \item Let $\cH_1=\{h\in \cH_0| h \text{ makes at least } \frac{\epsilon m}{4} \text{ mistakes in } R_1\}$.
        For any $h\in \cH_1$, we let $X(h)\subset R_1$ denote a minimal set of examples in $R_1$ (breaking ties arbitrarily but in a fixed way) such that $h$ misclassify $X(h)$ and
        $\abs{(\cG X(h))\cap S''_\cX}\geq \frac{\epsilon m}{4}$ where $\cG X(h) = \{g x |x\in X(h), g\in \cG\}$ is the set of all examples lying in the orbits generated from $X(h)$. 
        Let $K(h) = \cG X(h)$ and $\cK = \{K(h)|h\in \cH_1\}$ the collection of all such sets.
        Notice that each example in $X(h)$ must belong to different orbits, otherwise it is not minimal. 
        Besides, each orbit in $K(h)$ contains at least $\log^2 m$ examples from $S''_\cX$ according to the definition of $R_1$.
        Hence, we have $\abs{X(h)}\leq \frac{\epsilon m}{4 \log^2 m}$.
        Since there are at most $\frac{2m}{\log^2m}$ orbits generated from $R_1$, we have $\abs{\cK} \leq \sum_{i=1}^{\frac{\epsilon m}{4 \log^2 m}} {\frac{2m}{\log^2 m}\choose i} \leq \left(\frac{8e}{\epsilon}\right)^{\frac{\epsilon m}{4 \log^2 m}}$. 
        Recall that $\err_{\cG S}(h)=0$ iff. $h(x)=y$ for all $(x,y) \in \cG S$.
        Since $h$ misclassify $X(h)$, all examples in their orbits must go to $S'$ to guarantee $\err_{\cG S}(h)=0$.
        Thus, we have
        \begin{align*}
            &\Pr(\exists h\in \cH_1, \err_{S'}(h)\geq \frac{\epsilon}{2},  \err_{\cG S}(h)=0)\\
            \leq &\Pr(\exists h\in \cH_1, K(h)\cap S_\cX=\emptyset)
            = \Pr(\exists K\in \cK, K\cap S_\cX=\emptyset)\\
            \leq &\sum_{K\in \cK} 2^{-\frac{\epsilon m}{4}} \leq \left(\frac{8e}{\epsilon}\right)^{\frac{\epsilon m}{4 \log^2 m}}\cdot 2^{-\frac{\epsilon m}{4}} = 2^{-\frac{\epsilon m}{4}(1-\frac{\log(8 e/\epsilon)}{\log^2 m})}\leq 2^{-\frac{\epsilon m}{8}}\,,
        \end{align*}
        when $m\geq \frac{8e}{\epsilon}+ 4$.

        \item Let $\cH_2 = \cH_0\setminus \cH_1$. 
        That is to say, for all $h\in \cH_2$, $h$ will make at least $\frac{\epsilon m}{4}$ mistakes in $R_2$. 
        Since $\vcao(\cH,\cG) =d$ and every orbit generated from $R_2$ contains fewer than $\log^2 m$ examples in $S''_\cX$, 
        the number of examples in $R_2$ that can be shattered by $\cH$ is no greater than $d\log^2 m$. 
        Thus the number of ways labeling examples in $R_2$ is upper bounded by $(\frac{2em}{d})^{d\log^2 m}$ by Sauer's lemma. 
        Hence, we have
        \begin{align*}
            \Pr(\exists h\in \cH_2, \err_{S'}(h)\geq \frac{\epsilon}{2},  \err_{\cG S}(h)=0)\leq (\frac{2em}{d})^{d\log^2 m}\cdot 2^{-\frac{\epsilon m}{4}} = 2^{-\frac{\epsilon m}{4}+ d\log^2 m \log(2em/d)}\,.
        \end{align*}
    \end{itemize}
    Combining the results for $\cH_1$ and $\cH_2$, we have
    \begin{align*}
        \Pr(B_{S,S'}) \leq &\Pr(\exists h\in \cH_1, \err_{S'}(h)\geq \frac{\epsilon}{2},  \err_{\cG S}(h)=0) + \Pr(\exists h\in \cH_2, \err_{S'}(h)\geq \frac{\epsilon}{2},  \err_{\cG S}(h)=0)\\
        \leq & 2^{-\frac{\epsilon m}{8}} + 2^{-\frac{\epsilon m}{4}+ d\log^2 m \log(2em/d)}\\
        \leq & \frac{\delta}{2},
    \end{align*}
    when $m\geq \frac{8}{\epsilon}(d\log^2 m\log \frac{2em}{d}+\log \frac 4 \delta + e) + 4$.
\end{proof}
\section{Proof of Theorem~\ref{thm:inv-da-lb}}\label{app:inv-da-lb}
\begin{proof}
    For any $d>0$, for any $\cX,\cH,\cG$ satisfying that there exists a subset $X=\{x_0,x_1,\ldots,x_{2d}\}\subset \cX$ such that
    \begin{itemize}
        \item their orbits are pairwise disjoint;
        \item for all $u\subset \{1,\ldots,2d\}$ with $\abs{u} =d$, there exists an $h_u\in \cH$ such that $h_u(g x_i)=1$ for all $g\in \cG, i\in u\cup\{0\}$ and $h_u(x_i) =0$ for $i\in \{0,\ldots,2d\}\setminus u$,
    \end{itemize}
    we will prove the theorem for $\cH' = \{h_u|u\subset \{1,\ldots,2d\}, \abs{u}=d\}$.
    For $\cX,\cH,\cG$ satisfying the above conditions, we have $\vcao(\cH',\cG)\geq d$ and $\vco(\cH',\cG)$ between $0$ and $\vcao(\cH',\cG)$.
    Then consider that the target function $h^*=h_{u^*}$ is chosen uniformly at random from $\cH'$. 
    The marginal data distribution $\cD_\cX$ puts probability mass $1-16\epsilon$ on $x_0$ and $\frac{16\epsilon}{d}$ on each point in $X_{u^*}:= \{x_i|i\in u^*\}$. 
    Then the target function $h^*$ is $(\cG, \cD_\cX)$-invariant. 
    
    Let the sample size $m = \frac{d}{64\epsilon}$.
    Given the training set $S_\trn \sim \cD^m$, the expected number of sampled examples in $X_{u^*}$ is $\frac{d}{4}$.
    By Markov's inequality, with probability greater than $1/2$, we observed fewer than $\frac{d}{2}$ points of $X_{u^*}$ in $S_\trn$ (denoted as event $B$). 
    Let $\cA$ be any proper learner, which means $\cA$ must output a hypothesis in $\cH'$.
    For any $h\in \cH'$ consistent with $\cG S_\trn$, $h$ must predict $d$ unobserved points in $\{x_1,\ldots,x_{2d}\}$ as $0$.
    Since for each unobserved point in $\{x_1,\ldots,x_{2d}\}$ labeled as $0$ by $h$, conditioned on event $B$, this point has probability greater than $\frac{1}{3}$ to be in $X_{u^*}$, which implies it is misclassified by $h$.
    By following the stardard technique, let $\err'(h) = \Pr_{(x,y)\sim \cD}(h(x)\neq y \wedge x\in X_{u^*})$, which is no greater than $\err(h)$ for any predictor $h$.
    Hence,
    \begin{align*}
        \EEs{h^*,S_\trn}{\err'(\cA(S_\trn))|B}
        =& 
        \EEs{S_\trn}{\EEs{h^*}{\err'(\cA(S_\trn))|S_\trn,B}|B}\\
        = &\EEs{S_\trn}{\EEs{h^*}{\frac{16\epsilon}{d}\sum_{i\in [2d]:\cA(S_\trn,x_i)=0}\ind{i\in u^*}|S_\trn,B}|B}\\
        = &\frac{16\epsilon}{d}\EEs{S_\trn}{\sum_{i\in [2d]:\cA(S_\trn,i)=0}\EEs{h^*}{\ind{i\in u^*}|S_\trn,B}|B}\\
        > &\frac{16\epsilon}{d}\EEs{S_\trn}{\sum_{i\in [2d]:\cA(S_\trn,i)=0}\frac{1}{3}|B}\\
        =&\frac{16\epsilon}{3}\,.
    \end{align*}
    Then we have
    \[\EEs{h^*,S_\trn}{\err'(\cA(S_\trn))}\geq \EEs{h^*,S_\trn}{\err'(\cA(S_\trn))|B} \cdot \Pr(B) > \frac{8\epsilon}{3}\,.\]
    Thus, for any proper learner $\cA$, there exists a target hypothesis $h^*\in \cH'$ and a data distribution s.t. $\EEs{S_\trn}{\err'(\cA(S_\trn)))}> 8\epsilon/3$. 
    Since $\err'(h)\leq 16\epsilon$ for any predictor $h$, with probability greater than $\frac{1}{9}$, 
    $\err(\cA(S_\trn))\geq \err'(\cA(S_\trn)) >\epsilon$.

    Here is an example of $\cX,\cH',\cG$ satisfying the above conditions.
    Let $\cX = \{0,\pm 1, \pm 2, \ldots, \pm 2d\}$. 
    The group $\cG$ is defined as $\cG_d=\{e, -e\}$ where $e$ is the identity element.
    Thus $\cX$ can be divided into $2d+1$ pairwise disjoint orbits, $\{\{0\}, \{\pm 1\},\ldots,\{\pm 2d\}\}$.
    For any $u\subset [2d]$ with $\abs{u}=d$, define $h_u := 1-\ind{[2d]\setminus u}$, which labels $[2d]\setminus u$ by $0$ and the other points by $1$. 
    Then we define the hypothesis class $\cH' = \{h_u|\abs{u}=d\}$.
    Since $\{1,\ldots,d\}$ can be shattered and no $d+1$ points can be shattered by $\cH'$,
    we have $\vcao(\cH',\cG)=d$.
    And since $\{0,-1,\ldots,-2d\}$ can only be labeled as $1$ by $\cH'$, we have $\vco(\cH',\cG)=0$. 
\end{proof}
\section{Proof of Theorem~\ref{thm:inv-da-help}}\label{app:inv-da-help}
\begin{proof}
    Let $d = \vcd(\cH)$ and $X=\{x_1,x_2,\ldots,x_d\}$ be a set of examples shattered by $\cH$.
    Let $\cS_{d-1}$ denote the permutation group acting on $d-1$ objects.
    Then for any partition $(A,[d-1]\setminus A)$ of $[d-1]$, we let $\cG(A) := \{\sigma\in \cS_{d-1}| 
    \forall i\in A, \sigma(i)\in A\}$.
    By acting $\cG(A)$ on $X$, then $X$ are partitioned into three orbits: $\{x_i|i\in A\}, \{x_i|i\notin A\}$ and $\{x_d\}$.
    
    For convenience, we first consider the case of the instance space being $X$.
    Since there are only three orbits and every labeling of $X$ is realized by $\cH$, $\vcao(\cH,\cG(A))= \vco(\cH,\cG(A)) =3$ for all $A\subset [d-1]$.
    Consider that we pick a set $A^*$ uniformly at random from $2^{[d-1]}$.
    Let $h^* = \ind{\{x_i\}_{i\in A^*}}$ and $\cG= \cG(A^*)$.
    The data distribution put probability mass $1-16\epsilon$ on $x_d$ and the remaining $16\epsilon$ uniformly over $\{x_i|i\in [d-1]\}$.
    Then for the training set size $m=\frac{d-1}{64\epsilon}$, with probability at least $1/2$, at most half of $\{x_1,x_2,\ldots,x_{d-1}\}$ is sampled in the training set $S_\trn$.
    For any algorithm $\cA$ not knowing $\cG$, $\cA$ will output a hypothesis $\hat h = \cA(S_\trn)$, which does not depend on $\cG$.
    For each unobserved point $x$ in $\{x_1,x_2,\ldots,x_{d-1}\}$, $\cA$ has probability $1/2$ to misclassify $x$.
    Following the standard technique, let $\err'(h) := \Pr_{(x,y)\sim \cD}(h(x)\neq y \wedge x\in \{x_1,x_2,\ldots,x_{d-1}\})$ and then we have
    \begin{align*}
        \EEs{A^*,S_\trn}{\err'(\cA(S_\trn))} \geq 2\epsilon\,, 
    \end{align*} 
    which implies $\cM_\inv(\epsilon,\delta;\cH,\cG,\cA)=\Omega(\frac{d}{\epsilon} +\frac{1}{\epsilon}\log \frac{1}{\delta})$ by applying the standard technique in proving a lower bound of sample complexity in standard PAC learning.

    In the case where the instance space not being $X$, we modify $\cG(A^*)$ a little by arranging all points in $\{x|h^*(x)=1\}\setminus X$ in one orbit and all points in $\{x|h^*(x)=0\}\setminus X$ in another orbit. 
    Then there are at most $5$ orbits, and $\vcao(\cH,\cG)= \vco(\cH,\cG) \leq 5$.
\end{proof}
\section{Proof of Theorem~\ref{thm:inv-opt}}\label{app:inv-opt}

\begin{proof}
For any $k\leq \vco(\cH,\cG)$, let $X_k = \{x_1,\ldots,x_k\}$ be a set shattered in the way defined in Definition~\ref{def:vco}.
Then $X_k$ can be shattered by $\cH(X_k)=\{h_{|X_k}|h\in \cH \text{ is } (\cG,X_k) \text{-invariant}\}$ and $\vcd(\cH(X_k)) = k$.
Since any data distribution $\cD$ with $\cD_\cX(X_k)=1$ is $\cG$-invariant,  
any lower bound on the sample complexity of PAC learning of $\cH(X_k)$ also lower bounds the sample complexity of invariantly realizable PAC learning of $\cH$.
Then the lower bound follows by standard arguments from~\cite{vapnik:74,blumer1989learnability,ehrenfeucht1989general}.

For the upper bound, recall that the algorithm $\cA$ is defined by letting $\cA(S)(x) = Q_{\cH(X_S\cup \{x\}), X_{S}\cup \{x\}}(S, x)$ if $\cH(X_S\cup \{x\})\neq \emptyset$ and predicting arbitrarily if $\cH(X_S\cup \{x\})= \emptyset$ in Section~\ref{subsec:inv-opt}.
Due to the invariantly-realizable setting, if $S$ and the test point $(x,y)$ are i.i.d. from the data distribution, $h^*_{|X_S\cup \{x\}}$ is in $\cH(X_S\cup \{x\})$ a.s. and then, $\cH(X_S\cup \{x\})$ is nonempty.
Following the analogous proof by~\cite{haussler1994predicting} for standard PAC learning, we have
\begin{align}
    \EEs{S\sim \cD^t}{\err(\cA(S))} =& \EEs{(x_i,y_i)_{i\in[t+1]}\sim \cD^{t+1}}{\ind{\A(\{x_{i},y_{i}\}_{i\in [t]},x_{t+1})\neq y_{t+1}}}\nonumber \\
    =& \frac{1}{(t+1)!} \sum_{\sigma\in \text{Sym}(t+1) } \EE{\ind{\A(\{x_{\sigma(i)},y_{\sigma(i)}\}_{i\in [t]},x_{\sigma(t+1)})\neq y_{\sigma(t+1)}}}\nonumber\\
    =& \EE{\frac{1}{(t+1)!} \sum_{\sigma\in \text{Sym}(t+1) } \ind{\A(\{x_{\sigma(i)},y_{\sigma(i)}\}_{i\in [t]},x_{\sigma(t+1)})\neq y_{\sigma(t+1)}}}\nonumber\\
    \leq &  \frac{\EE{\vcd(\cH(\{x_i|i\in [t+1]\}))}}{t+1}\label{eq:apply-lmm1}\\
    \leq & \frac{\vco(\cH,\cG)}{t+1}\label{eq:apply-defvco}\,,
\end{align}
where Eq~\eqref{eq:apply-lmm1} adopts Lemma~\ref{lmm:1-inclusion} and Eq~\eqref{eq:apply-defvco} holds due to the definition of $\vco(\cH,\cG)$.
To convert this algorithm, guaranteeing the expected error upper bounded by $\vco(\cH,\cG)$, into an algorithm with high probability $1-\delta$, we again follow an argument of~\cite{haussler1994predicting}.
Specifically, the algorithm runs $\cA$ for $\ceil{\log(2/\delta)}$ times, each time using a new sample of size $\ceil{4\vco(\cH,\cG)/\epsilon}$.
Then the algorithm selects the hypothesis from the outputs with the minimal error on a new sample of size $\ceil{32/\epsilon(\ln(2/\delta)+\ln(\ceil{\log(2/\delta)}+1))}$.
\end{proof}
\section{Proof of Theorem~\ref{thm:re-da}}\label{app:re-da}
We first introduce a useful lemma about a well-known Boosting algorithm, known as $\alpha$-Boost.
Given access to a weak learning algorithm, it can output a hypothesis with strong learning guarantee.
See~\cite{schapire2012boosting} for a proof.
\begin{lemma}[Boosting]\label{lmm:boosting}
    For any $k, n\in \NN$ and multiset $(x_1,y_1),\ldots,(x_n,y_n)\in \cX\times\cY$, suppose $\cA_0$ is an algorithm that, for any distribution $\cP$ on $\cX\times\cY$ with $\cP(\{(x_1,y_1),\ldots,(x_n,y_n)\})=1$, there exists $S_\cP\in \{(x_1,y_1),\ldots,(x_n,y_n)\}^k$ with $\err_\cP(\cA_0(S_\cP))\leq 1/3$. Then there is a numerical constant $c\geq 1$ such that, for $T=\ceil{c\log(n)}$, there exists multisets $S_1,\ldots,S_T\in \{(x_1,y_1),\ldots,(x_n,y_n)\}^k$ such that, for $\hat h(\cdot) = \Majority(\cA_0(S_1)(\cdot),\ldots,\cA_0(S_T)(\cdot))$, it holds that $\hat h(x_i)=y_i$ for all $i\in [n]$.
\end{lemma}
Part of the proof relies on a well-known generalization bound for compression schemes.
The following is the classic result due to \cite{littlestone1986relating}.

\begin{lemma}[Consistent compression generalization bound]\label{lmm:re-compression}
    There exists a finite numerical constant $c>0$ such that, for any compression scheme $(\kappa,\rho)$, for any $n\in \NN$ and $\delta\in (0,1)$, for any distribution $\cD$ on $\cX\times \cY$, for $S\sim \cD^n$,
    with probability at least $1-\delta$, if $\err_S(\rho(\kappa(S)))=0$, then
    \begin{equation*}
        \err_\cD(\rho(\kappa(S)))\leq \frac{c}{n-\abs{\kappa(S)}}(\abs{\kappa(S)} \log(n)+\log(1/\delta))\,.
    \end{equation*}
\end{lemma}

\begin{proof}[Proof of the first part of Theorem~\ref{thm:re-da}]
    The proof is inspired by the idea that representing algorithms by an orientation in a 1-inclusion graph in the transductive setting by~\cite{daniely2014optimal}.
    We will first prove a lower bound in the transductive setting and then extend the result to the inductive setting.
    For any $t\geq 2$, denote $\mu = \mu(\cH,\cG,t)$ and let $\bphi$ with $\abs{\bphi}=t$ and $P\in \Delta(B(\cH,\cG,\bphi))$ be given such that $\mu(\cH,\cG,\bphi,P)\geq \frac{\mu}{2}$.
    A augmented dataset $\cG S = \{(gx,y)|(x,y)\in S, g\in \cG\}$ is the same as a multiset of labeled orbits $\{\cG x\times \{y\}|(x,y)\in S\}$ up to different data formats. 
    For convenience, we overload the notation a little by also allowing $\cA$ being a mapping from a multiset of labeled orbits to a hypothesis.
    Then we construct a 1-inclusion graph $G_{\cH,\cG}(\bphi) = \{V,E\}$ as introduced in Section~\ref{sec:re} and define a mapping $w_\cA\in W$ as follows.
    For any edge $e = \{\bff,\bg,x_i\}\in E$, let
    \[w_\cA(\{\bff,\bg,x_i\},\bff) = \Pr(\cA(\bphi_{-i},\bff_{-i},x_i) = g_i)\,,\]
    and 
    \[w_\A(\{\bff,\bg,x_i\},\bg) = \Pr(\cA(\bphi_{-i},\bff_{-i},x_i) = f_i)\,,\]
    which is well-defined as $\Pr(\cA(\bphi_{-i},\bff_{-i},x_i) = g_i)+\Pr(\cA(\bphi_{-i},\bff_{-i},x_i) = f_i) =1$.
    Suppose our target function and the instance sequence $(\bff, \bx)$ is drawn from the distribution $P$.
    Then the expected number of mistakes in the transductive learning setting is
    \begin{align}
        &\EEs{(\bff,\bx)\sim P,\cA}{\sum_{i=1}^t \ind{\cA(\bphi_{-i},\bff_{-i},x_i)\neq f_i}}\nonumber\\
        \geq & \EEs{(\bff,\bx)\sim P}{\sum_{i\in [t]: \exists e\in E, \{\bff,x_i\}\subset e} w_\A(e(\bff,x_i),\bff)}\nonumber\\
        \geq &\min_{w}\EEs{(\bff,\bx)\sim P}{\sum_{i\in [t]: \exists e\in E, \{\bff,x_i\}\subset e} w(e(\bff,x_i),\bff)}\nonumber\\
        = & \mu(\cH,\cG,\bphi,P)\,,\label{eq:lb-mu-trans}
    \end{align}
    where Eq~\eqref{eq:lb-mu-trans} holds due to the definition of $\mu(\cH,\cG,\bphi,P)$.
    
    Now we prove the lower bound in the inductive setting based on the similar idea. 
    Let $\bphi$ and $P$ be the same as those in the transductive setting.
    For any $\epsilon <\frac{\mu}{16(t-1)}$, we draw $(\bff,\bx)\sim P$ and then let $\bff$ be our target function and let the marginal data distribution be like,
    putting probability mass $1-\frac{16(t-1)\epsilon}{\mu}$ on $x_t$ and the remaining probability mass uniformly over  $\{x_1,\ldots,x_{t-1}\}$.
    Denote this data distribution by $\cD_{\bff,\bx}$.
    For any fixed $i\in [t-1]$, $\Pr(x_i \notin S_{\trn,\cX}) = (1-\frac{16\epsilon}{\mu})^m\geq (\frac{1}{4})^{\frac{16m\epsilon}{\mu}}\geq \frac{1}{2}$ when $m\leq \frac{\mu}{32\epsilon}$.
    For any hypothesis $h$, let $\err'_\cD(h) = \Pr_{(x,y)\sim \cD}(h(x)\neq y \text{ and } x\in \{x_i\}_{i\in [t-1]})$ and we always have $\err(h)\geq \err'(h)$. Then we have
    \begin{align}
        &\EEs{(\bff,\bx)\sim P, S_\trn\sim \cD_{\bff,\bx}^m, \cA}{\err'(\cA(\cG S_\trn))}\nonumber\\
        = & \frac{16\epsilon}{\mu}\sum_{i=1}^{t-1}\EEs{(\bff,\bx)\sim P, S_\trn\sim \cD_{\bff,\bx}^m, \cA}{\ind{\cA(\cG S_\trn,x_i)\neq f_i}}\nonumber\\
        \geq& \frac{16\epsilon}{\mu}\sum_{i=1}^{t-1}\EEs{(\bff,\bx)\sim P}{\Pr_{S_\trn\sim \cD_{\bff,\bx}^m, \cA}(\cA(\cG S_\trn,x_i)\neq f_i|x_i\notin S_{\trn,\cX})\Pr(x_i\notin S_{\trn,\cX})}\nonumber\\
        \geq & \frac{8\epsilon}{\mu}\EEs{(\bff,\bx)\sim P}{\sum_{i=1}^{t-1}\Pr_{S_\trn\sim \cD_{\bff,\bx}^m, \cA}(\cA(\cG S_\trn,x_i)\neq f_i|x_i\notin S_{\trn,\cX})}\label{eq:eq:lb-mu-indc}\,.
    \end{align}
    For all $i\in[t-1]$, for all $z\in \phi_i$, if there is an edge $e=\{\bff,\bg, z\}\in E$, we let
    \begin{equation*}
        \tilde w(e,\bff) = \Pr_{S_\trn\sim \cD_{\bff,\bx}^m, \cA}(\cA(\cG S_\trn,z)\neq f_i|z\notin S_{\trn,\cX})\,,
    \end{equation*}
    where $\bx$ is an arbitrary sequence in $\cU_{\bff}(\bphi)$ satisfying that $x_i = z$.
    Then $\tilde w(e,\bff)$ is well-defined since the distribution of $\cG S_\trn$ conditioned on $x_i\notin S_{\trn,\cX}$ is the same for all $\bx\in \cU_{\bff}(\bphi)$ with $x_i = z$.
    Actually, conditioned on $x_i\notin S_{\trn,\cX}$, the distribution of $\cG S_\trn$ is also the same when $S_\trn$ is sampled from $\cD_{\bg,\bx'}$ where $\bx'$ is an arbitrary sequence in  $\cU_{\bg}(\bphi)$ satisfying that $x_i' = z$.
    Hence, $\tilde w(e,\bff) + \tilde w(e,\bg)=1$.
    By letting $\tilde w(e,\bh)=0$ for all $\bh\notin e$, $\tilde w$ is in $W$.
    Then we have 
    \begin{align*}
        \text{Eq~\eqref{eq:eq:lb-mu-indc}} &\geq \frac{8\epsilon}{\mu}\EEs{(\bff,\bx)\sim P}{\sum_{i\in [t-1]: \exists e\in E, \{\bff,x_i\}\subset e}\tilde w(e,\bff)}\\
        &\geq \frac{8\epsilon}{\mu}\min_{w\in W}\EEs{(\bff,\bx)\sim P}{\sum_{i\in [t]: \exists e\in E, \{\bff,x_i\}\subset e} w(e,\bff)-1}\\
        &\geq 4\epsilon(1-2/\mu)\geq 2\epsilon,
    \end{align*}
    when $\mu\geq 4$. Hence, there exists a labeling function $\bff$ (i.e., there exists a target function $h^*$) and a data distribution $\cD_{\bff,\bx}$ such that $\EEs{S_\trn,\cA}{\err'(\cA(\cG S_\trn))}\geq 2\epsilon$. Since $\err'(h)\leq \frac{16(t-1)\epsilon}{\mu}$ for all hypothesis $h$, $\Pr(\err(\cA(\cG S_\trn)) >\epsilon)\geq \Pr(\err'(\cA(\cG S_\trn)) >\epsilon)>\frac{\mu}{16(t-1)}$.
\end{proof}

\begin{proof}[Proof of the second and third parts of Theorem~\ref{thm:re-da}]
    For any $n\in \NN$, for any given sample $\{(x_1,y_1),\ldots,(x_{n+1},y_{n+1})\}$, let $\bphi =\{\phi_1,\ldots,\phi_{n+1}\} =\{\cG x_1,\ldots, \cG x_{n+1}\}$ denote the multi-set of $t+1$ orbits and construct the one-inclusion graph $G_{\cH,\cG}(\bphi)$.
    As mentioned in Definition~\ref{def:mu}, every $w\in W$ defines a randomized orientation of each edge in graph $G_{\cH,\cG}(\bphi)$.
    That is, for any fixed $w\in W$, for every edge $e=\{\bff,\bg,x\}$, $w$ defines a probability over $\{\bff, \bg\}$.
    Then we can construct an algorithm $\A_w$ for each $w\in W$.
    
    Given the input $(\bphi_{-i},\vec y_{-i}, x_i)$, $\A_w$ finds the subset of vertices $\{(\bff,x_i)|\bff_{-i} = \vec y_{-i}\}$ whose labelings are consistent with $(\bphi_{-i},\vec y_{-i})$. 
    If there exist two such vertices, $(\bff,x_i)$ and $(\bg,x_i)$, they must be connected by $e=(\bff,\bg,x_i)$ due to the definition of $G_{\cH,\cG}(\bphi)$.
    Then $\A_w$ will predict $x_i$ as $f_i$ with probability $w(e,\bg)$ and as $g_i$ with probability $w(e,\bff)$.
    If only one such vertex $(\bff,x_i)$ exists, $\A_w$  predicts the label of $x_i$ by $f_i$. 
    Due to the realizable setting, there must exist at least one such vertex and the algorithm's prediction must be correct when only one vertex exists.
    
    To complete the algorithm, the remaining part is how to choose a good $w\in W$. 
    For any true labeling $\bff^*$ and any sequence of natural data $\bx^*\in \cU_{\bff^*}(\bphi)$, for each $i\in [n+1]$, if there is an edge $e\supset \{\bff^*, x_i^*\}$, it means the algorithm possibly misclassify $x_i^*$ (with the probability dependent on $w$); if there is no such an edge, it means the algorithm will not misclassify $x_i^*$ no matter what $w$ is. 
    For any labeling $\bff\in \Pi_\cH(\bphi)$ and a sequence of natural data $\bx\in \cU_{f}(\bphi)$, 
    we can represent the subset of the points in $\bx$ that the algorithm is uncertain about by a mapping $a_{\bff,\bx}: {E\times \Pi_\cH(\bphi)}\mapsto \{0,1\}$ where $a_{\bff,\bx}(e,\bg) = 1$ iff. $\bg=\bff$ and there exists $i\in [n+1]$ s.t. $\{\bff,x_i\}\in e$.
    Due to the definition, $a_{\bff,\bx}$ has at most $n+1$ non-zero entries.
    Let $A = \{a_{\bff,\bx}|\bff\in \Pi_\cH(\bphi), \bx\in \cU_{f}(\bphi)\}$ denote the set of all such mappings. 
    We now first consider the case where $\abs{E}<\infty$.
    Then for a training set of size $n$, we can rewrite the expected error as
    \begin{align}
        &\EEs{S\sim \cD^n}{\err(\A_w(\cG S))}\nonumber\\
        =&\EEs{S\sim \cD^n, (x,y)\sim \cD,\A_w}{\ind{\A_w(\cG S, x)\neq y}}\nonumber\\
        =&\EEs{(\bx,\vec y)\sim \cD^{n+1},\A_w}{\frac{1}{n+1}\sum_{i=1}^{n+1}\ind{\A_w(\bphi_{-i}, \vec y_{-i}, x_i)\neq y_i}}\nonumber\\
        =&\EEs{(\bx,\vec y)\sim \cD^{n+1},\A_w}{\frac{1}{n+1}\sum_{i=1}^{n+1}\ind{\A_w(\bphi_{-i},\vec y_{-i}, x_i)\neq y_i}\ind{\exists e\supset \{x_i,\bff^*\}}}\nonumber\\
        =&\EEs{(\bx,\vec y)\sim \cD^{n+1}}{\frac{1}{n+1}\sum_{e\in E}w(e,\bff^*)a_{\bff^*,\bx}(e,\bff^*)}\nonumber\\
        =&\EEs{(\bx,\vec y)\sim \cD^{n+1}}{\frac{1}{n+1}\sum_{e\in E, \bff\in \Pi_\cH(\bphi)}w(e,\bff)a_{\bff^*,\bx}(e,\bff)}\,.\label{eq:err-ub-harm}
    \end{align}
    where the last equality holds due to $a_{\bff^*,\bx}(e,\bff)=0$ for all $\bff\neq \bff^*$.
    Since $\bff^*$ and $\bx$ is unknown, our goal is to find a $w$ with $\sum_{e\in E, \bff\in \Pi_\cH(\bphi)}w(e,\bff)a_{\bff^*,\bx}(e,\bff)$ upper bounded for all $\bff^*$ and $\bx$. 
    The algorithm picks $w^* = \argmin_{w\in W}\max_{a\in A} \sum_{e\in E, \bff\in \Pi_{\cH}(\bphi)}w(e,\bff)a(e,\bff)$. 
    Then we have
    \begin{align}
        &\max_{a\in A}\sum_{e\in E, \bff\in \Pi_\cH(\bphi)}w^*(e,\bff)a(e,\bff)= \min_{w\in W} \max_{a\in A} \sum_{e\in E, \bff\in \Pi_\cH(\bphi)}w(e,\bff)a(e,\bff)\nonumber\\
        = & \min_{w\in W} \max_{a\in \conv(A)} \sum_{e\in E, \bff\in \Pi_\cH(\bphi)}w(e,\bff)a(e,\bff)= \max_{a\in \conv(A)}\min_{w\in W} \sum_{e\in E, \bff\in \Pi_\cH(\bphi)}w(e,\bff)a(e,\bff)\label{eq:finite-minimax}\,,
    \end{align}
    where the last equality is due to Minimax theorem.
    Since the optimal solution $a^*$ to Eq.~\eqref{eq:finite-minimax} is in the convex hull of $A$, 
    there is a distribution $P^*\in \Delta(B(\cH,\cG,\bphi))$ such that $a^* = \EEs{(\bff^*,\bx)\sim P^*}{a_{\bff^*,\bx}}$.
    Then we have
    \begin{align*}
        \text{Eq~\eqref{eq:finite-minimax}} = &\min_{w\in W}\sum_{e\in E, \bff\in \Pi_\cH(\bphi)}w(e,\bff) \EEs{(\bff^*,\bx)\sim P^*}{a_{\bff^*,\bx}(e,\bff)}\\
        = &\min_{w\in W}\EEs{(\bff^*,\bx)\sim P^*}{\sum_{e\in E, \bff\in \Pi_\cH(\bphi)}w(e,\bff) a_{\bff^*,\bx}(e,\bff)}\\
        = &\min_{w\in W}\EEs{(\bff^*,\bx)\sim P^*}{\sum_{i\in [n+1]: \exists e\in E, \{\bff^*,x_i\}\subset e} w(e(\bff^*,x_i),\bff^*)}\\
        =& \mu(\cH,\cG,\bphi, P^*)\,.
    \end{align*}
    Combined with Eq~\eqref{eq:err-ub-harm}, we have $\EEs{S \sim \cD^n}{\err(\A_{w^*}(\cG S))}\leq \frac{\mu(\cH,\cG,\bphi,n+1)}{n+1}$.
    
    For $\abs{E}=\infty$, $E\times \Pi_\cH(\bphi)$ could be infinite dimensional and thus, we need to use Sion's minimax theorem.
    The details of how to apply Sion's minimax theorem are described as follows.
    For all $w\in W$, for any edge $e = \{\bff,\bg, x\}$, we have $w(e,\bff) + w(e,\bg) =1$ and thus,
    there exists a one-to-one mapping $\beta: W\mapsto [0,1]^E$ where $\beta(w)(e) = w(e,\bff_{e, 0})$ where $\bff_{e, 0}$ is the labeling in $e$ predicting $x$ as zero.
    In the following, we will overload the notation by using $w$ to represent $\beta(w)$ when it is clear from the context that it is in the space $[0,1]^E$.
    Then we define a mapping $\BL(\cdot,\cdot): [0,1]^E\times A$ by 
    \begin{equation*}
        \BL(w,a):=\sum_{e:a(e,\bff_{e,0})=1}w(e) +\sum_{e:a(e,\bff_{e,1})=1}(1-w(e))\,,
    \end{equation*}
    where $\bff_{e, 0}, \bff_{e, 1}$ are labelings in $e = \{\bff_{e, 0}, \bff_{e, 1},x\}$ and they label $x$ as $0$ and $1$ respectively.
    Then similar to Eq~\eqref{eq:err-ub-harm}, we can represent the expected error as
    \begin{align*}
        \EEs{S \sim \cD^n}{\err(\A_w (\cG S))} = \EEs{(\bx,\vec y)\sim \cD^{n+1}}{\frac{1}{n+1}\BL(w,a_{\bff^*,\bx})}\,.
    \end{align*}
    Now we want to upper bound $\min_{w\in [0,1]^{E}} \sup_{a\in A}\BL(w,a)$.
    $\BL(\cdot,\cdot)$ can be extended to $\R^E\times A$ by letting $\BL(w,a)=\sum_{e:a(e,\bff_{e,0})=1}w(e) +\sum_{e:a(e,\bff_{e,1})=1}(1-w(e))$ for $w\in \R^E$.
    Since $a$ has at most $n+1$ non-zeros entries, $\abs{\BL(w,a)}\leq (n+1)\max(\norm{w}_\infty, \norm{\bOne - w}_\infty)$.
    Let $\tilde A=\{\sum_{i=1}^N c_i a_i|\forall i, a_i\in A, c_i>0,\sum_{i=1}^N c_i =1, N\in \NN\}$ be the set of all finite convex combination of elements in $A$.
    We extend $\BL$ from $R^{E} \times A$ to $R^{E}\times \tilde A$ by defining $\BL(w,a) = \sum_{i=1}^N c_i \BL(w,a_i)$ for $a=\sum_{i=1}^N c_i a_i\in \tilde A$.

    We define a metric $d$ in $\tilde A$: for $a=\sum_{i=1}^N c(a_i) a_i$ and $a'  =\sum_{i=1}^{N'} c'(a_i') a_i'$, the distance between $a$ and $a'$ is $d(a,a') := \sum_{\alpha \in A:c(\alpha)\neq 0 \text{ or } c'(\alpha)\neq 0}\abs{c(\alpha) - c'(\alpha)}$.
    For any fixed $w\in \R^E$, for any $a\in \tilde A$,
    for every open ball $\cB_r(\BL(w,a))$ centered at $\BL(w,a)$ with radius $r>0$, 
    there is an open ball $\cB_{r'}(a)$ in the metric space $(\tilde A,d)$ with $r' = \frac{r}{(n+1)(\max(\norm{w}_{\infty}, \norm{\bOne - w}_{\infty}))}$ such that for all $a'\in \cB_{r'}(a)$,
    $\abs{\BL(w,a')-\BL(w,a)} \leq \sum_{\alpha \in \{a_i\}_{i\in [N]}\cup \{a_i'\}_{i\in [N']}}\abs{c(\alpha)-c'(\alpha)}\abs{\BL(w,\alpha)}< r'\cdot \abs{\BL(w,\alpha)}\leq r$. 
    Hence, $\BL(w, \cdot)$ is continuous for all $w\in \R^E$.

    Consider the the standard topology in $\R$ and then $[0,1]$ is compact.
    Then let $\cT$ be the product topology of $\R^E$.
    Then by Tychonoff theorem, $[0,1]^E$ is compact in $\R^E$.
    For any fixed $a = \sum_{i=1}^N c_i a_i \in \tilde A$, there are at most $N(n+1)$ non-zero entries. 
    Then for any $w\in \R^E$,
    for every open ball $\cB_r(\BL(w,a))$ centered at $\BL(w,a)$ with radius $r>0$, 
    then there is a neighborhood $U= \prod_{e\in E} S_e$, where $S_e = (w_e-\frac{r}{N(n+1)},w_e+\frac{r}{N(n+1)})$ if at least one of $a(e,\bff_{e,0}), a(e,\bff_{e,1})$ is non-zero and $S_e = \R$ for other $e\in E$, such that $\abs{\BL(w',a)- \BL(w,a)}<r$ for all $w'\in U$. 
    That is, $\BL(U)\subset \cB_r(\BL(w,a))$.
    Hence, $\BL(\cdot, a)$ is continuous for all $a\in \tilde A$.

    It is easy to check that $\BL(\cdot, a)$ and $\BL(w, \cdot)$ are linear for all $a\in \tilde A$, $w\in W$.
    Then by Sion's minimax theorem, we have 
    \begin{align*}
        \min_{w\in [0,1]^E} \sup_{a\in A}\BL(w,a)
        \leq  \min_{w\in [0,1]^E} \sup_{a\in \tilde A}\BL(w,a)
        = \sup_{a\in \tilde A}\min_{w\in [0,1]^E}\BL(w,a)=:v^*\,.
    \end{align*}
    Let $w^*=\argmin_{w\in [0,1]^E} \sup_{a\in A}\BL(w,a)$.
    There exists a sequence $a_1,a_2,\ldots$ in $\tilde A$ such that $\lim_{k\rightarrow\infty} \min_{w\in [0,1]^E}\BL(w,a_k) = v^*$.
    For each $a_k$, we let $P_k\in \Delta(B(\cH,\cG,\bphi))$ be the distribution over $B(\cH,\cG,\bphi)$ such that $a_k = \EEs{(\bff^*,\bx)\sim P_k}{a_{\bff^*,\bx}}$.
    Due to the definition of $\tilde A$, we know that $P_k$ is a discrete distribution with finite support.
    Then we have
    \begin{align*}
        \sup_{a\in A}\BL(w^*,a)\leq &\min_{w\in [0,1]^E} \sup_{a\in \tilde A}\BL(w,a)\\
        = &\sup_{a\in \tilde A}\min_{w\in [0,1]^E}\BL(w,a)\\
        = &\lim_{k\rightarrow\infty} \min_{w\in [0,1]^E}\BL(w,\EEs{(\bff^*,\bx)\sim P_k}{a_{\bff^*,\bx}})\\
        = &\lim_{k\rightarrow\infty} \min_{w\in [0,1]^E}\EEs{(\bff^*,\bx)\sim P_k}{\BL(w,a_{\bff^*,\bx})}\\
        \leq &\sup_{P\in \Delta(B(\cH,\cG,\bphi))}\min_{w\in [0,1]^E}\EEs{(\bff^*,\bx)\sim P}{\BL(w,a_{\bff^*,\bx})}\\
        = & \mu(\cH,\cG,\bphi,n+1)\,.
    \end{align*}
    Hence, $\EEs{S \sim \cD^n}{\err(\A_{w^*}(\cG S))}\leq \frac{\mu(\cH,\cG,\bphi,n+1)}{n+1}$.
    \paragraph{The first upper bound}  We can convert the above bound into a high probability bound by $\alpha$-Boost.
    Let $\cA_0 = \cA_{w^*}$ as defined above.
    Let $t$ be any positive integer such that $\frac{1}{6}\leq \frac{\mu(\cH,\cG,t)}{t}\leq \frac{1}{3}$.
    As established above, for $S_\trn = \{(x_1,y_1),\ldots,(x_m,y_m)\}\sim \cD^m$ and any distribution $\cP$ supported on $\{(x_1,y_1),\ldots,(x_m,y_m)\}$,
    for $k=t-1$ and $S\sim \cP^k$, $\EE{\err_{\cP}(\cA_0(S))}\leq 1/3$.
    Thus given $S_\trn$ and $\cP$, there exists a deterministic choice of $S_\cP\in \{(x_1,y_1),\ldots,(x_m,y_m)\}^k$ with $\err_{\cP}(\cA_0(S_\cP))\leq 1/3$.
    Then Lemma~\ref{lmm:boosting} implies that for a value $T=\ceil{c_1 \log m}$ (for numerical constant $c_1\geq 1$), 
    there exists $S_1,\ldots,S_T\in \{(x_1,y_1),\ldots,(x_m,y_m)\}^k$ such that, for $\hat h(\cdot) = \Majority(\cA_0(S_1)(\cdot),\ldots,\cA_0(S_T)(\cdot))$, 
    it holds that $\hat h(x_i)=y_i$ for all $i\in [m]$.
    Note that $\hat h$ can be expressed as a compression scheme.
    By Lemma~\ref{lmm:re-compression},
    with probability at least $1-\delta$,
    \[\err(\hat h)\leq \frac{c_2}{m-kT}\left(kT\log m + \log \frac{1}{\delta}\right)\,,\]
    for a numerical constant $c_2\geq 1$. 
    Thus, for any given $\epsilon\in (0,1)$, the right hand side can be made less than $\epsilon$ for an appropriate choice of
    \[m = O( \frac{1}{\epsilon} (k\log^2\frac{k}{\epsilon} + \log \frac{1}{\delta}))\,,\]
    where $k = t-1\leq 6\mu(\cH,\cG,t) -1$.
    \paragraph{The second upper bound} Again, using the same standard technique as we used in Theorem~\ref{thm:inv-opt} to convert an algorithm with expected error upper bound to an algorithm with high probability guarantee. 
    The algorithm runs $\cA_{w^*}$ for $\ceil{\log(2/\delta)}$ times, each time using a new sample of size $\ceil{4\mu(\cH,\cG)/\epsilon}$.
    Then the algorithm selects the hypothesis from the outputs with the minimal error on a new sample of size $\ceil{32/\epsilon(\ln(2/\delta)+\ln(\ceil{\log(2/\delta)}+1))}$.
\end{proof}
\section{Proof of Theorem~\ref{thm:re-mu-dim}}\label{app:re-mu-dim}
\begin{proof}
    Let $d = \dim(\cH,\cG)$.
    Let $\bphi$ and the corresponding $B = \{(\bff,\bx_\bff)\}_{\bff\in \cY^d}$ be given.
    Let $P$ be the uniform distribution over $B$.
    Then
    \begin{align}
        &\mu(\cH,\cG,\bphi,P)=\min_{w\in W} \EEs{(\bff,\bx_\bff)\sim P}{\sum_{i\in [d]:\exists e\in E, \{\bff,x_{\bff, i}\}\subset e} w(e,\bff)}\nonumber \\
        =& \min_w  \frac{1}{2^d}\sum_{\bff\in \cY^d}\sum_{i\in [d]:\exists e\in E, \{\bff,x_{\bff, i}\}\subset e} w(e,\bff)\nonumber\\
        =& \frac{1}{2^d} \min_w \sum_{\bff\in \cY^d}\sum_{i=1}^d \frac{1}{2}\left(w(e(\bff, x_{\bff, i}),\bff) + w(e(\bff, x_{\bff, i}),(1-f_i,\bff_{-i}))\right)\label{eq:apply-def-dim}\\
        =&\frac{1}{2^d}\cdot \frac{2^d\cdot d}{2}= \frac{d}{2}\nonumber\,,
    \end{align}
    where Eq~\eqref{eq:apply-def-dim} holds since $x_{\bff,i} = x_{\bg,i}$ if $\bff \oplus \bg = \be_i$ due to the definition of $\dim(\cH,\cG)$.
\end{proof}
\section{Proof of Theorem~\ref{thm:re-opt}}\label{app:re-opt}
\begin{proof}[Proof of the lower bound] 
    Let $d = \vcao(\cH,\cG)$, let $X_d = \{x_1,\ldots,x_d\}$ be a set shattered in the way defined in Definition~\ref{def:vcao} and let $\cH':= \{h_{|X_d}|h\in \cH\}$.
    Then $\vcd(\cH') = d$ and any lower bound on the sample complexity of PAC learning of $\cH'$ is also a lower bound on the sample complexity of relaxed realizable PAC learning of $\cH$.
    The lower bound follow by standard arguments from~\cite{vapnik:74,blumer1989learnability,ehrenfeucht1989general}.
\end{proof}

\begin{proof}[Proof of the upper bound of the algorithm ERM-INV]
    The algorithm ERM-INV works as follows.
    ERM-INV first applies ERM over the original data set.
    Then for every test instance $x$, 
    if $x$ lies in the orbits generated by the original data, 
    the algorithm predicts $x$ by the label of the training instance in the same orbit; 
    otherwise, the algorithm predicts according to the ERM output.
    Specifically, given the training set $S_\trn=\{(x_1,y_1),\ldots,(x_m,y_m)\}\sim \cD^m$, the algorithm finds a hypothesis $h\in \cH$ consistent with $S_\trn$ and then outputs $f_{h,S_\trn}$ defined by
    \begin{align}
        f_{h,S_\trn}(x) = 
        \begin{cases}
        y_i& \text{if there exists }i\in[m] \st x\in \cG x_i\,,\\
        h(x)& \text{o.w.}
        \end{cases}\label{eq:erm-inv}
    \end{align}
    The function $f_{h,S_\trn}$ is well-defined a.s. when the data distribution $\cD$ is $\cG$-invariant since if there exists $i\neq j$ such that $x_i \in \cG x_j$, then $y_i = y_j = h^*(x_j)$.

    The proof idea is similar to that of Theorem~\ref{thm:inv-da-ub}.
    Let $d = \vcao(\cH,\cG)$.
    Consider two sets $S$ and $S'$ of $m$ i.i.d. samples drawn from the data distribution $\cD$ each.
    We denote $A_{S}$ the event of $\{\exists h\in \cH, \err_{S}(h)=0, \err_\cD(f_{h,S})\geq \epsilon\}$ and $B_{S,S'}=\{\exists h\in \cH, \err_{S }(h)=0, \err_{S'}(f_{h,S})\geq \frac{\epsilon}{2}\}$.
    By Chernoff bound, we have $\Pr(B_{S,S'})\geq \Pr(A_S)\cdot \Pr(B_{S,S'}|A_S)\geq \frac{1}{2}\Pr(A_S)$ when $m\geq \frac{8}{\epsilon}$.
    The sampling process of $S$ and $S'$ is equivalent to drawing $2m$ i.i.d. samples and then randomly partitioning into $S$ and $S'$ of $m$ each.
    For any fixed $S''$, let us divide $S''$ into two categories in terms of the number of examples in each orbit.
    Let $R_1 = \{x|\abs{\cG x \cap S''_{\cX}}\geq \log^2 m\}$ and $R_2 = S''_{\cX}\setminus R_1$.
    Let $\cH_0\subset \cH$ denote the set hypotheses making at least $\frac{\epsilon m}{2}$ mistakes in $S''$.
    Now we divide $\cH_0$ into two sub-classes as follows.
    \begin{itemize}
        \item  Let $\cH_1=\{h\in \cH_0| h \text{ makes fewer than }\frac{\epsilon m}{4} \text{ mistakes in } R_2\}$.
        Let $\cF_{S''}(\cH_1) := \{f_{h,T}| h\in \cH_1, T\subset S'', \abs{T}=m\}$. 
        For any $h$, if $S$ is correctly labeled, then $\err_{\cG S}(f_{h,S})=0$.
        Thus we have
        \begin{align*}
            &\Pr(\exists h\in \cH_1, \err_{S}(h)=0, \err_{S'}(f_{h,S})\geq \frac{\epsilon}{2})\\
            \leq &\Pr(\exists h\in \cH_1, \err_{\cG S}(f_{h,S})=0, \err_{S'}(f_{h,S})\geq \frac{\epsilon}{2})\\
            \leq &\Pr(\exists f\in \cF_{S''}(\cH_1), \err_{\cG S}(f)=0, \err_{S'}(f)\geq \frac{\epsilon}{2})\,.
        \end{align*}
        Since every $h\in \cH_1$ makes fewer than $\frac{\epsilon m}{4}$ mistakes in $R_2$, for all $f\in \cF_{S''}(\cH_1)$, 
        if $f$ makes at least $\frac{\epsilon m}{2}$ mistakes in $S''$, 
        it must makes at least $\frac{\epsilon m}{4}$ mistakes in $R_1$. 
        Similar to the case~1 in the proof of Theorem~\ref{thm:inv-da-ub}, for any $f\in \cF_{S''}(\cH_1)$, 
        we let $X(f)\subset R_1$ denote a minimal set of examples in $R_1$ (breaking ties arbitrarily but in a fixed way) such that $f$ misclassify $X(f)$ and $\abs{(\cG X(f))\cap S''_\cX}\geq \frac{\epsilon m}{4}$ where $\cG X(f) = \{g x |x\in X(f), g\in \cG\}$ is the set of all examples lying in the orbits generated from $X(f)$. 
        Let $K(f) = \cG X(f)$ and $\cK = \{K(f)|f\in \cF_{S''}(\cH_1)\}$.
        Notice that each example in $X(f)$ must belong to different orbits, otherwise it is not minimal. 
        Besides, each orbit contains at least $\log^2 m$ examples from $S''_\cX$.
        Hence, $\abs{X(f)}\leq \frac{\epsilon m}{4 \log^2 m}$. 
        Since there are at most $\frac{2m}{\log^2 m}$ orbits generated from $R_1$, we have $\abs{\cK} \leq \sum_{i=1}^{\frac{\epsilon m}{4 \log^2 m}} {\frac{2m}{\log^2 m}\choose i} \leq \left(\frac{8e}{\epsilon}\right)^{\frac{\epsilon m}{4 \log^2 m}}$.
        Since $f$ misclassify $X(f)$, all examples in $K(f)$ must go to $S'$ to guarantee $\err_{\cG S}(f)=0$ and thus, 
        we have
        \begin{align*}
            &\Pr(\exists f\in \cF_{S''}(\cH_1), \err_{\cG S}(f)=0, \err_{S'}(f)\geq \frac{\epsilon}{2})\\
            \leq & \Pr(\exists f\in \cF_{S''}(\cH_1), K(f)\cap S_\cX=\emptyset)=\Pr(\exists K\in \cK, K\cap S_\cX=\emptyset)\\
            \leq &\sum_{K\in \cK} 2^{-\frac{\epsilon m}{4}} \leq \left(\frac{8e}{\epsilon}\right)^{\frac{\epsilon m}{4 \log^2 m}}\cdot 2^{-\frac{\epsilon m}{4}} = 2^{-\frac{\epsilon m}{4}(1-\frac{\log(8 e/\epsilon)}{\log^2 m})}\leq 2^{-\frac{\epsilon m}{8}}\,,
        \end{align*}
        when $m\geq \frac{8e}{\epsilon}+4$.
        
        \item Let $\cH_2 = \cH_0\setminus \cH_1$.
        Now we will bound $\Pr(\exists h\in \cH_2, \err_{S}(h)=0, \err_{S'}(f_{h,S})\geq \frac{\epsilon}{2})$.
        Similar to the case~2 in Theorem~\ref{thm:inv-da-ub}, every $h\in \cH_2$ will make at least $\frac{\epsilon m}{4}$ mistakes in $R_2$. 
        Since $\vcao(\cH,\cG) = d$ and every orbit generated from $R_2$ contains fewer than $\log^2 m$ examples, the number of examples in $R_2$ that can be shattered by $\cH$ is no greater than $d\log^2 m$. 
        Thus, the number of ways labeling examples in $R_2$ is upper bounded by $(\frac{2em}{d})^{d\log^2 m}$ by Sauer's lemma. 
        For any multi-subset $X\subset S''_\cX$ and hypothesis $h$, we denote by $\hat M_{X}(h)$ the number of instances in $X$ misclassified by $h$.
        Hence, we have
    \begin{align*}
        &\Pr(\exists h\in \cH_2, \err_{S}(h)=0, \err_{S'}(f_{h,S})\geq \frac{\epsilon}{2})\\
        \leq & \Pr(\exists h\in \cH_2, \hat M_{S_\cX\cap R_2}(h)=0, \hat M_{S'_\cX\cap R_2}(h)\geq \frac{\epsilon m}{4})\\
        \leq &(\frac{2em}{d})^{d\log^2 m}\cdot 2^{-\frac{\epsilon m}{4}} = 2^{-\frac{\epsilon m}{4}+ d\log^2 m \log(2em/d)}
    \end{align*}
    \end{itemize}
    Combining the results for $\cH_1$ and $\cH_2$, we have
    \begin{align*}
        \Pr(B_{S,S'}) \leq  2^{-\frac{\epsilon m}{8}} + 2^{-\frac{\epsilon m}{4}+ d\log^2 m \log(2em/d)}\leq  \frac{\delta}{2},
    \end{align*}
    when $m\geq \frac{8}{\epsilon}(d\log^2 m\log \frac{2em}{d} + \log \frac 4 \delta + e) + 4$.
\end{proof}
\begin{proof}[Proof of the upper bound of the 1-inclusion-graph predictor]
    The algorithm is similar to the 1-inclusion-graph predictor in Theorem~\ref{thm:inv-opt}.
    For any $t\in \NN$ and $S=\{(x_1,y_1),\ldots, (x_t,y_t)\}$, let $X_S$ be the set of different elements in $S_{\cX}$ and $\cH'(X_S):= \{h_{|X_S}|\forall x',x\in X_S, x'\in \cG x \text{ implies } h(x')=h(x)\}$.
    Here $\cH'(X_S)$ is different from $\cH(X_S)$ defined in Theorem~\ref{thm:inv-opt} in the sense that every hypothesis in $\cH'(X_S)$ is not $(\cG,X_S)$-invariant but only predict the observed examples in the same orbit in the same way.
    Note that $h^*_{|X_S}$ is in $\cH'(X_S)$ if $S$ is realized by $h^*$.
    Let $Q_{\cH'(X_S),X_S}$ be the function guaranteed by Eq~\eqref{eq:1-inclusion} for the instance space $X_S$ and hypothesis class $\cH'(X_S)$.
    Given a set $S$ of $t$ i.i.d. samples, we let $\cA(S)$ be defined as $\cA(S)$ be defined as $\cA(S, x) = Q_{\cH'(X_{S}\cup \{x\}), X_{S}\cup \{x\}}(S, x)$ if $\cH'(X_{S}\cup \{x\})$ is nonempty and predicting arbitrarily if it is empty.
    Following the analogous proof of Theorem~\ref{thm:inv-opt}, we have
    \begin{align*}
        \EEs{S\sim \cD^t}{\err(\cA(S))} =& \EEs{(x_i,y_i)_{i\in[t+1]}\sim \cD^{t+1}}{\ind{\A(\{x_{i},y_{i}\}_{i\in [t]},x_{t+1})\neq y_{t+1}}} \\
        =& \frac{1}{(t+1)!} \sum_{\sigma\in \text{Sym}(t+1) } \EE{\ind{\A(\{x_{\sigma(i)},y_{\sigma(i)}\}_{i\in [t]},x_{\sigma(t+1)})\neq y_{\sigma(t+1)}}}\\
        =& \EE{\frac{1}{(t+1)!} \sum_{\sigma\in \text{Sym}(t+1) } \ind{\A(\{x_{\sigma(i)},y_{\sigma(i)}\}_{i\in [t]},x_{\sigma(t+1)})\neq y_{\sigma(t+1)}}}\\
        \leq &  \frac{\EE{\vcd(\cH'(\{x_i|i\in [t+1]\}))}}{t+1}\leq \frac{\vcao(\cH,\cG)}{t+1}\,.
    \end{align*}
Again, we use the same method as that in Theorem~\ref{thm:inv-opt} to convert this algorithm, guaranteeing the expected error upper bounded by $\vcao(\cH,\cG)$, into an algorithm with high probability $1-\delta$.
Specifically, the algorithm runs $\cA$ for $\ceil{\log(2/\delta)}$ times, each time using a new sample of size $\ceil{4\vcao(\cH,\cG)/\epsilon}$.
Then the algorithm selects the hypothesis from the outputs with the minimal error on a new sample of size $\ceil{32/\epsilon(\ln(2/\delta)+\ln(\ceil{\log(2/\delta)}+1))}$.
\end{proof}
\section{Proof of Theorem~\ref{thm:ag-opt}}\label{app:ag-opt}
To prove the theorem, we will use a generalization bound for agnostic compression scheme by~\cite*{graepel2005pac}.
\begin{lemma}[Agnostic compression generalization bound]\label{lmm:ag-compression}
    There exists a finite numerical constant $c>0$ such that, for any compression scheme $(\kappa,\rho)$, for any $n\in \NN$ and $\delta\in (0,1)$, for any distribution $\cD$ on $\cX\times \cY$, for $S\sim \cD^n$,
    letting $B(S,\delta):=\frac{1}{n}(\abs{\kappa(S)}\log(n)+\log(1/\delta))$,
    with probability at least $1-\delta$, then
    \begin{equation*}
        \abs{\err_\cD(\rho(\kappa(S)))-\err_S(\rho(\kappa(S)))}\leq c\sqrt{\err_S(\rho(\kappa(S)))B(S,\delta)} + c B(S,\delta)\,.
    \end{equation*}
\end{lemma}

\begin{proof}[Proof of the lower bound]
    For the lower bound, our construction follows \cite{ben2014sample}.
    For any $\cX,\cH$, let $A_1,\ldots,A_d$ be subsets of $\cX$ such that the orbits of every two different elements $x,x'\in \cup_{i\in[d]} A_i$ are disjoint, i.e., $\forall x,x'\in \cup_{i\in[d]} A_i, \cG x\cap \cG x' =\emptyset$.
    We say $\cH$ set-shatters $A_1,\ldots,A_d$ if for every binary vector $\by\in \cY^d$, there exists some $h_\by\in \cH$ such that for all $i\in [d]$ and $x\in \cX$, if $x\in A_i$ then $h_\by(x) = y_i$.
    Then by following Theorem~7 of \cite{ben2014sample}, if $\cH$ set-shatters $A_1,\ldots,A_d$ for some infinite subsets $A_1,\ldots,A_d$ of $\cX$, the standard agnostic PAC sample complexity of learning $\cH$ under deterministic labels for instance space being $\cup_{i\in [d]} A_i$ is lower bounded by $\frac{d}{\epsilon^2}$ for all $\delta < 1/32$.
    Since any data distribution $\cD$ with $\cD_\cX(\cup_{i\in [d]} A_i)=1$ is $\cG$-invariant, the above lower bound also lower bounds $\cM_\ag(\epsilon,\delta;\cH,\cG)$ for all $\delta < 1/32$.
    Let $h(x) = 0$ for all $x\notin \cup_{i\in[k]} A_i$ and for all $h\in \cH$, $\vcao(\cH,\cG) = d$.
    Thus, $\cM_\ag(\epsilon,1/64;\cH,\cG) = \Omega(\frac{\vcao(\cH,\cG)}{\epsilon^2})$.
    The construction above works for any $d>0$.
\end{proof}
\begin{proof}[Proof of the upper bound of ERM-INV]
    Let $d = \vcao(\cH,\cG)$.
    Consider two sets $S$ and $S'$ of $m$ i.i.d. samples drawn from the data distribution $\cD$ each.
    We denote $A_{S}$ the event of $\{\exists h\in \cH, \err_\cD(f_{h,S})\geq \err_{S}(h)+\epsilon\}$ and $B_{S,S'}=\{\exists h\in \cH, \err_{S'}(f_{h,S})\geq \err_{S}(h)+\frac{\epsilon}{2}\}$.
    By Hoeffding bound, we have $\Pr(B_{S,S'})\geq \Pr(A_S)\cdot \Pr(B_{S,S'}|A_S)\geq \frac{1}{2}\Pr(A_S)$ when $m\geq \frac{2}{\epsilon^2}$.
    The sampling process of $S$ and $S'$ is equivalent to drawing $2m$ i.i.d. samples and then randomly partitioning into $S$ and $S'$ of $m$ each.
    For any fixed $S''$, let us divide $S''$ into two categories in terms of the number of examples in each orbit.
    Let $R_1 = \{x|\abs{\cG x \cap S''_{\cX}}\geq \log^2 m\}$ and $R_2 = S''_\cX\setminus R_1$.
    For any multi-subset $X\subset S''_\cX$ and hypothesis $h$, we denote by $\hat M_{X}(h)$ the number of instances in $X$ misclassified by $h$.
    Let $\cF_{S''}(\cH) := \{f_{h,T}| h\in \cH, T\subset S'', \abs{T}=m\}$. 
    Then we have 
    \begin{align*}
        &\Pr(B_{S,S'}) \\
        = &\Pr(\exists h\in \cH, \err_{S'}(f_{h,S})\geq \err_{S}(h)+\frac{\epsilon}{2})\\
        \leq & \Pr(\exists h\in \cH, (\hat M_{S'_\cX \cap R_1}(f_{h,S})\geq \hat M_{S_\cX \cap R_1}(h)+\frac{\epsilon m}{4}) \vee (\hat M_{S'_\cX \cap R_2}(f_{h,S})\geq \hat M_{S_\cX \cap R_2}(h)+\frac{\epsilon m}{4}))\\
        \leq & \underbrace{\Pr(\exists h\in \cH, \hat M_{S'_\cX \cap R_1}(f_{h,S})\geq \hat M_{S_\cX \cap R_1}(h)+\frac{\epsilon m}{4})}_{(a)}\\
        &+ \underbrace{\Pr(\exists h\in \cH,\hat M_{S'_\cX \cap R_2}(f_{h,S})\geq \hat M_{S_\cX \cap R_2}(h)+\frac{\epsilon m}{4})}_{(b)}\,.
    \end{align*}
    For the first term, since $\err_{\cG S}(f_{h,S})=0$ according to the definition of $f_{h,S}$,
    \begin{align*}
        (a) \leq &\Pr(\exists f\in \cF_{S''}(\cH), \hat M_{S'_\cX \cap R_1}(f)\geq \frac{\epsilon m}{4}, \err_{\cG S}(f)=0)\,.
    \end{align*}
    Again, similar to case~1 in Theorem~\ref{thm:inv-da-ub} (also the first upper bound in Theorem~\ref{thm:re-opt}), 
    for any $f\in \cF_{S''}(\cH)$, 
    we let $X(f)\subset R_1$ denote a minimal set of examples in $R_1$ (breaking ties arbitrarily but in a fixed way) such that $f$ misclassify $X(f)$ and $\abs{(\cG X(f))\cap S''_\cX}\geq \frac{\epsilon m}{4}$ where $\cG X(f) = \{g x |x\in X(f), g\in \cG\}$ is the set of all examples lying in the orbits generated from $X(f)$. 
    Let $K(f) = \cG X(f)$ and $\cK = \{K(f)|f\in \cF_{S''}(\cH)\}$.
    Notice that each example in $X(f)$ must belong to different orbits, otherwise it is not minimal. 
    Besides, each orbit contains at least $\log^2 m$ examples.
    Hence, $\abs{X(f)}\leq \frac{\epsilon m}{4 \log^2 m}$. 
    Since there are at most $\frac{2m}{\log^2 m}$ orbits generated from $R_1$, we have $\abs{\cK} \leq \sum_{i=1}^{\frac{\epsilon m}{4 \log^2 m}} {\frac{2m}{\log^2 m}\choose i} \leq \left(\frac{8e}{\epsilon}\right)^{\frac{\epsilon m}{4 \log^2 m}}$.
    Since $f$ misclassify $X(f)$, all examples in $K(f)$ must go to $S'$ to guarantee $\err_{\cG S}(f)=0$ and thus, 
    we have
    \begin{align*}
        &\Pr(\exists f\in \cF_{S''}(\cH),\hat M_{S'_\cX \cap R_1}(f)\geq \frac{\epsilon m}{4}, \err_{\cG S}(f)=0)\\
        \leq & \Pr(\exists f\in \cF_{S''}(\cH), K(f)\cap S_\cX=\emptyset)=\Pr(\exists K\in \cK, K\cap S_\cX=\emptyset)\\
        \leq &\sum_{K\in \cK} 2^{-\frac{\epsilon m}{4}} \leq \left(\frac{8e}{\epsilon}\right)^{\frac{\epsilon m}{4 \log^2 m}}\cdot 2^{-\frac{\epsilon m}{4}} = 2^{-\frac{\epsilon m}{4}(1-\frac{\log(8 e/\epsilon)}{\log^2 m})}\leq 2^{-\frac{\epsilon m}{8}}\,,
    \end{align*}
    when $m\geq \frac{8e}{\epsilon}$.
    For the second term, since $\hat M_{S'_\cX \cap R_2}(h)\geq \hat M_{S'_\cX \cap R_2}(f_{h,S})$, then we have
    \begin{align}
        (b) \leq & \Pr(\exists h\in \cH, \hat M_{S'_\cX \cap R_2}(h)\geq \hat M_{S_\cX \cap R_2}(h)+\frac{\epsilon m}{4})\nonumber\\
        \leq & (\frac{2em}{d})^{d\log^2 m}\cdot e^{-2m'(\frac{\epsilon m}{8m'})^2} \label{eq:apply-hoeffding}\\
        \leq & e^{-\frac{\epsilon^2m}{32}+ d\log^2 m \ln(2em/d)}\nonumber\,,
    \end{align}
    where Eq~\eqref{eq:apply-hoeffding} adopts Hoeffding bound and Sauer's lemma (the number points in $R_2$ that can be shattered by $\cH$ is at most $d\log^2 m$; otherwise $\vcao(\cH,\cG)> d$) and $m' = \max_{h\in \cH} M_{ R_2}(h)$.
    By setting $m\geq \frac{32}{\epsilon^2}(d\log^2 m \ln(2em/d) + \ln(8/\delta)+1) + \frac{8}{\epsilon}\log(8/\delta)$, we have $\Pr(A_S)\leq \delta/2$.
    By Hoeffding bound, with probability at least $1-\delta/2$, $\err_S(h^*)\leq \err_\cD(h^*)+\frac{\epsilon}{2}$.
    By a union bound, we have that with probability at least $1-\delta$, $\err_\cD(f_{ 
    h,S})\leq \err_S(h) + \epsilon \leq \err_S(h^*) + \epsilon \leq \err_\cD(h^*)+ 3\epsilon/2$.
\end{proof}

\begin{proof}[Proof of the upper bound of the 1-inclusion-graph predictor]
    Here we provide another algorithm based on the technique of reduction-to-realizable of \cite{david2016supervised} and the 1-inclusion-graph predictor in the relaxed realizable setting.
    Following the argument in Theorem~\ref{thm:re-opt}, given a sample $S=\{(x_1,y_1),\ldots, (x_n,y_n)\}$, let $X_S$ be the set of different elements in $S_{\cX}$ and define $\cH'(X_S):= \{h_{|X_S}|\forall x',x\in X_S, x'\in \cG x \text{ implies } h(x')=h(x)\}$.
    Then we define an algorithm $\cA_0$ as 
    \begin{align}
        \cA_0(S,x) = 
        \begin{cases}
            Q_{\cH'(X_{S}\cup \{x\}), X_{S}\cup \{x\}}(S, x)
            &\text{if } \cH'(X_{S}\cup \{x\})\neq \emptyset\,,\\
            0 &\text{ o.w.}
        \end{cases}\label{eq:def-of-A}
    \end{align}
    where $Q_{\cH'(X_{S}\cup \{x\}), X_{S}\cup \{x\}}$ is the function guaranteed by Eq~\eqref{eq:1-inclusion} for hypothesis class $\cH'(X_{S}\cup \{x\})$ and instance space $ X_{S}\cup \{x\}$.
    In the relaxed realizable setting, if $S\sim \cD^n$ and $x\sim \cD_\cX$, $\cH'(X_{S}\cup \{x\})\neq \emptyset$ a.s. as it contains $h^*_{|X_{S}\cup \{x\}}$. 
    While in the agnostic setting, it is not the case.
    Even if $S\sim \cD^n$ and $x\sim \cD_\cX$, $\cH'(X_{S}\cup \{x\})$ could be empty as there might exist an instance $x'\in X_S \cap \cG x$ such that no hypothesis in $\cH$ labeling them in the same way.
    
    Let $d = \vcao(\cH,\cG)$.
    If a sample $\{(x_1,y_1),\ldots, (x_{n+1},y_{n+1})\}$ is realizable by $\cH$, then by Lemma~\ref{lmm:1-inclusion}, we have
    \begin{align}
        \frac{1}{(n+1)!}\sum_{\sigma\in \Sym(n+1)}\ind{\cA_0(\{x_{\sigma(i)},y_{\sigma(i)}\}_{i\in [n]},x_{\sigma(n+1)})\neq y_{\sigma(n+1)}} \leq \frac{d}{n+1}\,.\label{eq:apply-ag-1-inclusion}
    \end{align}
    
    Now we use $\cA_0$ as a weak learner to construct a compression scheme by following the construction \cite{david2016supervised}.
    Given a training set $S_\trn=\{(x_1,y_1),\ldots,(x_m,y_m)\}$, let $R$ denote the largest submulitset of $S_\trn$ that is realizable w.r.t. $\cH$.
    If $\abs{R} = 0$, then define $\hat h$ as the all-0 function $\hat h(x)=0$.
    Otherwise, if $\abs{R}>0$, 
    for any distribution $\cP$ on $R$,
    by Eq~\eqref{eq:apply-ag-1-inclusion}, we have that
    \begin{align*}
        &\EEs{S\sim \cP^{3d}}{\err_\cP(\cA_0(S))}\\
        =&\EEs{(x_i,y_i)_{i\in[3d+1]}\sim  \cP^{3d}}{\frac{1}{(3d+1)!}\sum_{\sigma\in \Sym(3d+1)}\ind{\cA_0(\{x_{\sigma(i)},y_{\sigma(i)}\}_{i\in [3d]},x_{\sigma(3d+1)})\neq y_{\sigma(3d+1)}} }\\
        \leq& \frac{1}{3}\,.
    \end{align*}
    Hence, there exists $S_\cP\in R^{3d}$ with $\err_\cP(\cA_0(S_\cP))\leq 1/3$.
    Thus, the algorithm $\cA_0$ can serve as a weak learner.
    By Lemma~\ref{lmm:boosting}, for $T=\ceil{c\log \abs{R}}$ (for a numerical constant $c>0$), there exist $S_1,\ldots, S_T\in R^{3d}$ such that, letting $\hat h (\cdot)= \Majority(\cA_0(S_1)(\cdot),\ldots,\cA_0(S_T)(\cdot))$, we have $\err_{R}(\hat h)=0$.
    Thus, $\err_{S_\trn}(\hat h)\leq \inf_{h\in \cH} \err_{S_\trn}(h)$.
    Here $\hat h$ is the output of the compression scheme that selects $\kappa(S_\trn) = (S_1,\ldots, S_T)$ and $\rho(\kappa(S_\trn))=\hat h$.
    By Lemma~\ref{lmm:ag-compression}, with probability at least $1-\delta/2$,
    \begin{align*}
        \err_\cD(\hat h) \leq \err_{S_\trn}(\hat h) + c'\sqrt{\err_{S_\trn}(\hat h)B(S_\trn,\delta/2)} + c' B(S_\trn,\delta/2)\,,
    \end{align*}
    for a numerical constant $c'>0$.
    Following the same argument of \cite{david2016supervised} (Lemma~3.2), we have that 
    \begin{equation*}
        \Pr_{S_\trn\sim \cD^m}(\err_S(\hat h)\geq \inf_{h\in \cH} \err_{\cD}(h) + \sqrt{\log(2/\delta)/m})\leq \delta/2\,.
    \end{equation*}
    By taking a union bound, then we have that with probability at least $1-\delta$,
    \begin{align*}
        \err_\cD(\hat h) \leq &\inf_{h\in \cH} \err_{\cD}(h) + c'\sqrt{(\inf_{h\in \cH} \err_{\cD}(h) + \sqrt{\log(2/\delta)/m})\frac{1}{m}(3d\ceil{c\log\abs{R}}\log m + \log1/\delta)} \\
        &+\frac{c'}{m}(3d\ceil{c\log\abs{R}}\log m + \log(1/\delta))\,.
    \end{align*}
    Hence,
    \begin{equation*}
        \cM_\ag(\epsilon,\delta;\cH,\cG) = O\left(\frac{d}{\epsilon^2}\log^2\left(\frac{d}{\epsilon}\right)+\frac{1}{\epsilon^2}\log(1/\delta)\right)\,.
    \end{equation*}
\end{proof}
\section{Adaptive algorithms}
For any hypothesis $h\in \cH$, we say $h$ is $(1-\eta)$-invariant over the distribution $\cD_{\cX}$ for some $\eta\in [0,1]$ if $\PPs{x\sim \cD_{\cX}}{\exists x'\in \cG x, h(x')\neq h(x)}= \eta$.
We call $\eta(h)=\eta$ the invariance parameter of $h$ with respect to $\cD_{\cX}$.
In the relaxed realizable setting, the problem degenerates into the invariantly realizable setting when $\eta(h^*)=0$.
This implies that we can benefit from transformation invariances more when $\eta(h^*)$ is smaller.
The case is similar in the agnostic setting.
In this section, we discuss adaptive learning algorithms for different levels of invariance of the target function.

\subsection{An adaptive algorithm in the relaxed realizable setting}\label{app:unified-re}
There might exist more than one hypotheses in $\cH$ with zero error and we let the target function $h^*$ be the one with the smallest invariance parameter (breaking ties arbitrarily). 
For any multiset $X$ in $\cX$, for any hypothesis $h$, denote $h_{|X}$ the restriction of $h$ on the set of different elements in $X$.
Then we introduce a distribution-dependent dimension as follows.

\begin{definition}[approximate $(1-\eta)$-invariant VC dimension]\label{def:re-appx-dim}
    For any $\eta\in [0,1]$ and finite multi-subset $X\subset \cX$,
    let $\cH^{\eta}(X) := \{h_{|X}|\frac{1}{\abs{X }}\sum_{x\in X}{\ind{\exists x'\in \cG x, h(x')\neq h(x)}}\leq \eta \wedge \forall x,x'\in \cG x \cap X, h(x')=h(x)\}$.
    For any $m\in \NN$, marginal distribution $\cD_\cX$ and target function $h^*$, the approximate $(1-\eta)$-invariant VC dimension is defined as
    \[\vco^{\eta}(m,h^*,\cH,\cG,\cD_\cX):= \EEs{X\sim \cD_{\cX}^{m+1}}{\vcd(\cH^{\eta}(X))|h^*_{|X}\in \cH^{\eta}(X)}\,,\]
    when $\Pr(h^*_{|X}\in \cH^{\eta}(X))>0$ and $\vco^{\eta}(m,h^*,\cH,\cG,\cD_\cX) =0$ when $\Pr(h^*_{|X}\in \cH^{\eta}(X))=0$.
    By taking supremum over $m$, 
    \[\vco^{\eta}(h^*,\cH,\cG,\cD_\cX) = \sup_{m\in \NN}\vco^{\eta}(m,h^*,\cH,\cG,\cD_\cX)\,.\]
\end{definition}
In this definition, all hypotheses in $\cH^{\eta}(X)$ need to satisfy two constraints: a) the empirical invariance parameter is less than or equal to $\eta$, and b) the prediction over instances in $X$ is invariant over orbits.
Here the second constraint arises due to the fact that $h^*$ satisfies this constraint.
Note that $\vco^{\eta}(h^*,\cH,\cG,\cD_\cX)$ is monotonic increasing in $\eta$.
For all $\cD_\cX, h^*$, we have $\vco^{0}(h^*,\cH,\cG,\cD_\cX)\leq \vco(\cH,\cG)$ and $\vco^{\eta}(h^*,\cH,\cG,\cD_\cX)\leq \vcao(\cH,\cG)$ for all $\eta\in [0,1]$.
We will use the approximate $(1-\eta)$-invariant VC dimension to characterize the sample complexity dependent on $\eta$.
Ideally, it is heuristic to adopt a notion like $\vcao(\{h\in \cH|\eta(h) = \eta(h^*)\})$.
But this is impossible to achieve as we cannot obtain an accurate estimate of $\eta(h^*)$ via finite data points.

\begin{proposition}\label{prop:re-eta-known-ub}
    If it is known that $\eta(h^*)\leq \eta$ for some $\eta\in [0,1]$, for any training sample size $m\in \NN$, for any $\Delta \geq \sqrt{\frac{\ln (n+1)}{2(n+1)}}$ with $n=\Theta(\frac{m}{\log(1/\delta)})$, there is an algorithm achieving error $O(\frac{\vco^{\eta+\Delta}(h^*,\cH,\cG,\cD_\cX)\log(1/\delta)}{m})$ with probability at least $1-\delta$.
\end{proposition}
Given a sample $S=\{(x_1,y_1),\ldots,(x_n,y_n)\}$ and a test instance $x$, 
let $X_S$ denote the set of different elements in $S_\cX$ and $X=\{x_1,\ldots, x_{n},x\}$ the multiset of all unlabeled instances from both the training set and the test instance.
The algorithm $\cA$ is defined by 
\begin{equation*}
    \cA(S, x) = Q_{\cH^{\eta+\Delta}(X),X_S\cup\{x\}}(S,x)\,,
\end{equation*}
where $Q_{\cH^{\eta+\Delta}(X),X_S\cup\{x\}}$ is the function guaranteeed by Eq~\eqref{eq:1-inclusion} for the instance space $X_S\cup\{x\}$ and the concept class $\cH^{\eta+\Delta}(X)$.
Similar to Theorem~\ref{thm:inv-opt}, algorithm $\cA$ can achieve expected error $\frac{\vco^{\eta+\Delta}(h^*,\cH,\cG,\cD_\cX)+1}{n+1}$.
Then by following the same confidence boosting argument, we run $\cA$ for $\ceil{\log (2/\delta)}$ times on independent new samples and select the output hypothesis with minimum error a new sample.
The details of the algorithm and the proof of Proposition~\ref{prop:re-eta-known-ub} are deferred to Appendix~\ref{app:re-eta-known-ub}. 

In the more general case where $\eta$ is unknown, we build an algorithm based on the algorithm for known $\eta$ above.
Denote $\cA_{\eta,\Delta}$ the algorithm satisfying the guarantee in Proposition~\ref{prop:re-eta-known-ub} with probability $1-\delta/2$ for hyperparameters $\eta$ and $\Delta$.
Then we divide $[0,1]$ into uniform intervals and then search for the interval where $\eta(h^*)$ lies in.
The detailed algorithm is provided in Algorithm~\ref{alg:unif-re}.

\begin{algorithm}[t]\caption{An adaptive algorithm in the relaxed realizable setting}\label{alg:unif-re}
    \begin{algorithmic}[1]
    \STATE Input: a labeled sample $S_\trn$ of size $m$, $m_1\in \NN$, $\Delta\in [0,1]$
    \STATE Randomly partition $S$ into $S_1$ with $\abs{S_1}=m_1$ and $S_2$ with $\abs{S_2}=m-m_1$
    \FOR{$i=0,1,\ldots,\ceil{1/(2\Delta)}$}
    \STATE Let $h_i = \cA_{(2i-1)\Delta,\Delta}(S_1)$
    \STATE Return $\hat h = \argmin_{h\in \{h_i|i\in \{0,\ldots,\ceil{1/(2\Delta)}\}\}}\err_{S_2}(h)$
    \ENDFOR
    \end{algorithmic}
  \end{algorithm}

\begin{theorem}\label{thm:re-eta-unknown-ub}
    Set $\Delta = \sqrt{\frac{\ln (n+1)}{2(n+1)}}$ for $n=\Theta(\frac{m}{\log (m) \log(1/\delta))})$ and $\abs{S_1} = \Theta(\frac{m}{\log m})$.
    Let $i^*\geq 0$ be the smallest integer $i$ such that $\max((2i-1)\Delta,0)\geq \eta(h^*)$. 
    Then Algorithm~\ref{alg:unif-re} achieves error $O(\frac{\vco^{2i^* \Delta}(h^*,\cH,\cG,\cD_\cX)\log(1/\delta)\log(m)}{m})$ with probability at least $1-\delta$.
\end{theorem}
The proof is deferred to Appendix~\ref{app:re-eta-unknown-ub}.
The algorithm above perform close to optimally in both the invariant realizable setting and the relaxed realizable setting.
Intuitively, the algorithm above outperforms PAC-optimal algorithms in Theorem~\ref{thm:re-opt} when $\eta(h^*)$ is small.
Below is an example showing the advantage of this adaptive algorithm in the extreme case of $\eta(h^*) = 0$.
\begin{example}
Consider the construction of $\cH_d,\cG_d, h^*, \cD$ in the proof of Theorem~\ref{thm:inv-da-lb}. 
In this example, we have $i^*=0$ and $\cH^0(X)=\{\bOne\}$ for any $X$.
Thus, $\vco^0(h^*,\cH,\cG,\cD_\cX) = 0$.
Hence, the algorithm above only requires sample complexity $O(1)$ to output a zero-error predictor.
However, algorithms only caring the worst case upper bound may require much more samples.
For example, ERM-INV predicts exactly the same as standard ERM in this example and thus, it requires $\Omega(\frac{\vcao(\cH,\cG)}{\epsilon})$ samples to achieve $\epsilon$ error.
\end{example}

\paragraph{Open question:} It is unclear whether this adaptive algorithm is optimal and whether the approximate $(1-\eta)$-invariant VC dimension is the best $\eta$-dependent measure to characterize the sample complexity.

\subsection{An adaptive algorithm in the agnostic setting}\label{app:unified-ag}
In the agnostic setting, we assume that there exists an optimal hypothesis $h^*\in \cH$ such that $\err(h^*) = \inf_{h\in \cH}\err(h)$.
Then similar to the realizable setting, we can design algorithms that adapt to $\eta(h^*)$.
However, it is more challenging to design an adaptive algorithm in the agnostic setting than in the relaxed realizable setting.
One of the most direct ideas is to combine agnostic compression scheme with the adaptive algorithm in the realizable setting.
One possible way of combination is finding the largest realizable subset of the data and applying the adaptive algorithm in the relaxed realizable setting.
However, this does not work since the realizable subset is not i.i.d. and the empirical invariance parameter calculated based on this subset is biased.
Another possible way of combination is calculating the empirical invariance parameter over the whole data set, reducing the hypothesis class based on this empirical value and then run the compression scheme in Theorem~\ref{thm:ag-opt} based on this reduced hypothesis class.
This does not work either because the predictor depends on the whole data set now and the compression size is too large.
Hence, there is a significant barrier that has arisen as a result of estimating the invariance parameter while obtaining low error at the same time.

To get around this obstacle, we provide an approach of using two independent data sets.
Specifically, we partition the hypothesis class into subclasses with different empirical invariance parameters based on a data set first.
Notice that in this step, we only need an unlabeled data set.
Then we run the compression scheme in Theorem~\ref{thm:ag-opt} for each subclass and return the one with the small validation error.
The detailed algorithm is presented in Algorithm~\ref{alg:unif-ag}.

\begin{algorithm}[t]\caption{An adaptive algorithm in the agnostic case}\label{alg:unif-ag}
  \begin{algorithmic}[1]
  \STATE Input: a unlabeled data set $U$ of size $u$ and a labeled data set $S$ of size $m$,$m_1\in \NN$, $\Delta>0$
  \STATE Randomly divide $S$ into $S_1$ of size $m_1$ and $S_2$ of size $m-m_1$
  \STATE Use $U$ to partition $\cH$ into $\hat \cH_1,\ldots,\hat \cH_K$ with $\hat \cH_i = \{h\in \cH| \frac{1}{u}\sum_{x\in U} \ind{\exists x'\in \cG x, h(x')\neq h(x)} \in (2(i-1)\Delta, 2i\Delta]\}$ and $K = \ceil{\frac{1}{2\Delta}}$
  \FOR{$i=0,1,2,\ldots,K$}
  \STATE Run the algorithm $\cA$ in Theorem~\ref{thm:ag-opt} over $S_1$ for hypothesis class $\hat \cH_i$ and output $h_i$
  \STATE Return $\hat h = \argmin_{h\in \{h_i|i\in \{0,\ldots,K\}\}} \err_{S_2}(h)$
  \ENDFOR
  \end{algorithmic}
\end{algorithm}

\begin{theorem}\label{thm:ag-eta}
    For each $h\in \cH$, the invariance indicator function of $h$ is a mapping $\iota_h: \cX\mapsto \{0,1\}$ such that $\iota_h(x) = \ind{\exists x'\in \cG x, h(x')\neq h(x)}$.
    Denote $\cI = \{\iota_h|h\in \cH\}$ the set of invariance indicator functions for all $h\in \cH$.
    Then for any $\Delta \in (0,1]$, Algorithm~\ref{alg:unif-ag} can achieve $\err(\hat h) \leq \err(h^*) + O\left(\sqrt{\frac{\vcao(\cH^*)\log^2 m + \log(1/\delta) + \log(1/\Delta)}{m}}\right)$, where $\cH^* = \{h| \eta(h)\in (\eta(h^*) -2\Delta -2 \Delta',\eta(h^*) +2\Delta + 2\Delta')\}$ with $\Delta' = \Theta(\sqrt{\frac{\vcd(\cI)\log(u)+\log(1/\delta)}{u}})$.
\end{theorem}
The detailed proof of Theorem~\ref{thm:ag-eta} is deferred to Appendix~\ref{app:ag-eta}.
The upper bound in Theorem~\ref{thm:ag-eta} depends on $\vcd(\cI)$, which can be arbitrarily larger than $\vcd(\cH)$.
For example, for any $d>0$, let $\cX = \{0,1\}^d\times [d]$.
For each $\bb\in \{0,1\}^d$, define a hypothesis $h_\bb$ by letting $h_\bb((\bb,i)) = b_i$ and $h_\bb(x) = 0$ for all other $x\in \cX$.
Let the hypothesis class $\cH = \{h_\bb|\bb\in \{0,1\}^d\}$.
Then it is direct to check that $\vcd(\cH) = 1$ but $\vcd(\cI) = d$.
It is unclear how to design an adaptive algorithm with theoretical guarantee independent of $\vcd(\cI)$.
The $\eta$-dependent dimension we adopt in the agnostic setting is different from that in the relaxed realizable setting.
It is also unclear what is the best way to characterize the dependence on $\eta$ in the agnostic setting.

\section{Proof of Proposition~\ref{prop:re-eta-known-ub}}\label{app:re-eta-known-ub}
\begin{proof}
    The proof follows that of Theorem~\ref{thm:inv-opt}.
    Given a training sample of size $m$, let $n$ be the largest integer such that $m \geq n\ceil{\log(2/\delta)} + \ceil{8(n+1)(\ln(2/\delta)+\ln(\ceil{\log(2/\delta)}+1))}$.
    For any sample $S=\{(x_1,y_1),\ldots,(x_n,y_n)\}$ and a test instance $x_{n+1}=x$,
    if $h^*$ is in $\cH^{\eta+\Delta}(X)$, by Lemma~\ref{lmm:1-inclusion}, we have that
    \begin{align*}
        \frac{1}{(n+1)!}\sum_{\sigma\in \text{Sym}(n+1) } \ind{\cA(\{x_{\sigma(i)},y_{\sigma(i)}\}_{i\in [n]},x_{\sigma(n+1)})\neq y_{\sigma(n+1)}}\leq \frac{\vcd(\cH^{\eta+\Delta}(X))}{n+1}\,.
    \end{align*}
    Thus,
    \begin{align}
        &\EEs{S\sim \cD^n}{\err(\cA(S))} \nonumber\\
        =& \EEs{(x_i,y_i)_{i\in[n+1]}\sim \cD^{n+1}}{\ind{\A(\{x_{i},y_{i}\}_{i\in [n]},x_{n+1})\neq y_{n+1}}} \nonumber\\
        =& \frac{1}{(n+1)!} \sum_{\sigma\in \text{Sym}(n+1) } \EE{\ind{\A(\{x_{\sigma(i)},y_{\sigma(i)}\}_{i\in [n]},x_{\sigma(n+1)})\neq y_{\sigma(n+1)}}}\nonumber\\
        =& \EE{\frac{1}{(n+1)!} \sum_{\sigma\in \text{Sym}(n+1) } \ind{\A(\{x_{\sigma(i)},y_{\sigma(i)}\}_{i\in [n]},x_{\sigma(n+1)})\neq y_{\sigma(n+1)}}}\nonumber\\
        \leq & \EEc{\frac{1}{(n+1)!} \sum_{\sigma\in \text{Sym}(n+1) } \ind{\A(\{x_{\sigma(i)},y_{\sigma(i)}\}_{i\in [n]},x_{\sigma(n+1)})\neq y_{\sigma(n+1)}}}{h^*_{|X}\in \cH^{\eta+\Delta}(X)} \nonumber\\
        & \cdot\Pr(h^*_{|X}\in \cH^{\eta+\Delta}(X))+ \Pr(h^*_{|X}\notin \cH^{\eta+\Delta}(X))\nonumber\\
        \leq & \frac{\EEs{X\sim \cD_{\cX}^{n+1}}{\vcd(\cH^{\eta+\Delta}(X))|h^*_{|X}\in \cH^{\eta+\Delta}(X)}}{n+1}\Pr(h^*_{|X}\in \cH^{\eta+\Delta}(X))  \nonumber\\
        &+ \Pr(h^*_{|X}\notin \cH^{\eta+\Delta}(X))\nonumber\\
        \leq & \frac{\vco^{\eta+\Delta}(h^*,\cH,\cG,\cD_\cX)}{n+1}\Pr(h^*_{|X}\in \cH^{\eta+\Delta}(X)) + \Pr(h^*_{|X}\notin \cH^{\eta+\Delta}(X))\nonumber\\
        \leq & \frac{\vco^{\eta+\Delta}(h^*,\cH,\cG,\cD_\cX)+1}{n+1}\,,\label{eq:unif-exp-er}
    \end{align}
    where Eq~\eqref{eq:unif-exp-er} holds due to $\Pr(h^*_{|X}\notin \cH^{\eta+\Delta}(X))\leq \Pr(\frac{1}{\abs{X }}\sum_{x\in X}{\ind{\exists x'\in \cG x, h^*(x')\neq h^*(x)}}-\eta(h^*)\geq \Delta)\leq \exp(-2(n+1)\Delta^2)$ by Hoeffding bound.

    Then we again follow the classic technique to boost the confidence.
    The algorithm runs $\cA$ for $\ceil{\log(2/\delta)}$ times, each time using a new sample $S_i$ of size $n$ for $i=1,\ldots,\ceil{\log(2/\delta)}$.
    Let $h_i = \cA(S_i)$ and then selects the hypothesis $\hat h$ from $\{h_i|i\in [\ceil{\log(2/\delta)}]\}$ with the minimal error on a new sample $S_0$ of size $t=\ceil{8(n+1)(\ln(2/\delta)+\ln(\ceil{\log(2/\delta)}+1))}$.

    Denote $\epsilon = \frac{\vco^{\eta+\Delta}(h^*,\cH,\cG,\cD_\cX)+1}{n+1}$.
    For each $i$, by Eq~\eqref{eq:unif-exp-er}, we have $\EE{\err(h_i)}\leq \epsilon$. By Markov's inequality, with probability at least $\frac{1}{2}$, $\err(h_i)\leq 2\epsilon$.
    Since $h_i$ are independent, we have that with probability at least $1-\frac{\delta}{2}$, at least one of $\{h_i|i\in [\ceil{\log(2/\delta)}]\}$ has error smaller than $2\epsilon$.
    Then by Chernoff bound, for each $i$, on the event $\err(h_i)\leq 2\epsilon$,
    \begin{equation*}
        \Pr(\err_{S_0}(h_i)> 3\epsilon|h_i)< e^{-\frac{t\epsilon}{6}}\,.
    \end{equation*}
    Also, on the event $\err(h_i)> 4\epsilon$,
    \begin{equation*}
        \Pr(\err_{S_0}(h_i)\leq 3\epsilon|h_i)\leq e^{-\frac{t\epsilon}{8}}\,.
    \end{equation*}
    Thus, by the law of total probability and a union bound, with probability at least $1-\frac{\delta}{2}$, if any $i$ has $\err(h_i)\leq 2\epsilon$, then the returned the hypothesis $\hat h$ has $\err(\hat h)\leq 4\epsilon$.
    By a union bound, the proof is completed.
\end{proof}
\section{Proof of Theorem~\ref{thm:re-eta-unknown-ub}}\label{app:re-eta-unknown-ub}
\begin{proof}
    Given a training sample of size $m$, let $\abs{S_1} = m_1$ and $\abs{S_2}=m_2=m-m_1$.
    The values of $m_1,m_2$ are determined later.
    Let $n$ be the largest integer such that $m_1 \geq n\ceil{\log(2/\delta)} + \ceil{8(n+1)(\ln(2/\delta)+\ln(\ceil{\log(2/\delta)}+1))}$.
    Let $\Delta = \sqrt{\frac{\ln (n+1)}{2(n+1)}}$ and then the number of rounds we run $\cA_{\eta, \Delta}$ as a subroutine is upper bounded by $\ceil{1/(2\Delta)}+1=\ceil{\sqrt{(n+1)/(2\ln(n+1))}}+1\leq m$.
    According to Proposition~\ref{prop:re-eta-known-ub}, there is a numerical constant $c>0$ such that with probability $1-\delta/2$, $\err(h_{i^*})\leq \frac{c \vco^{2i^*\Delta}(h^*,\cH,\cG,\cD_\cX)\ln(2/\delta)}{m_1}$.
    Denote $\epsilon = \frac{c \vco^{2i^*\Delta}(h^*,\cH,\cG,\cD_\cX)\ln(2/\delta)}{m_1}$.
    By Chernoff bound, for each $i \neq i^*$, on the event $\err(h_i)>4\epsilon$,
    \begin{equation*}
        \Pr(\err_{S_2}(h_i)\leq 2\epsilon|h_i)\leq e^{-\frac{m_2\epsilon}{2}}\,.
    \end{equation*}
    Also, on the event $\err(h_{i^*})\leq \epsilon$,
    \begin{equation*}
        \Pr(\err_{S_2}(h_{i^*})> 2\epsilon|h_{i^*})< e^{-\frac{m_2\epsilon}{3}}\,.
    \end{equation*}
    Let $m_1 = \frac{cm}{3\ln m +c}$ and $m_2 = \frac{3m\ln m}{3\ln m +c}$. 
    Then with probability at least $1-m e^{-m_2\epsilon/3}\geq 1-\delta/2$, if $\err(h_{i^*})\leq \epsilon$, then the returned classifier has error smaller than $4\epsilon$.
    By taking a union bound, the proof is completed.
\end{proof}
\section{Proof of Theorem~\ref{thm:ag-eta}}\label{app:ag-eta}
\begin{proof}
    Given a labeled data set $S$ of size $m$, let $\abs{S_1} = m_1$ and $\abs{S_2}=m_2=m-m_1$.
    The values of $m_1, m_2$ are determined later.
    Let $\hat i$ be the $i\in [K]$ such that $h^*\in \hat \cH_i$.
    By Theorem~\ref{thm:ag-opt}, we have that there exists a numerical constant $c>0$ such that with probability at least $1-\delta/3$,
    \[\err(h_{\hat i})\leq \err(h^*) + c\sqrt{\frac{\vcao(\hat \cH_{\hat i})\log^2m_1 + \log(1/\delta)}{m_1}}\,.\]
    Then by Hoeffding bound, for any $\epsilon>0$, for each $i\in K$, with probability at least $1-\delta/(3K)$,
    \[\abs{\err_{S_2}(h_i)-\err(h_i)} \leq \epsilon\,,\]
    when $m_2 \geq \frac{\ln (6K/\delta)}{2{\epsilon}^2}$.
    Then by taking a union bound, with probability at least $1-\delta/3$, for all $i\in K$ we have
    \[\abs{\err_{S_2}(h_i)-\err(h_i)} \leq \epsilon\,.\]
    Hence, with probability at least $1-2\delta/3$,
    \[\err(\hat h) \leq \err(h_{\hat i}) + 2 \epsilon \leq \err(h^*) + c\sqrt{\frac{\vcao(\hat \cH_{\hat i})\log^2m_1  + \log(1/\delta)}{m_1}} + 2\epsilon\,.\]
    By uniform convergence bound, with probability at least $1-\delta/3$,
    \[\abs{\frac{1}{u}\sum_{x\in U} \ind{\exists x'\in \cG x, h(x')\neq h(x)} - \eta(h)} \leq \Delta', \forall h\in \cH\,.\]
    It follows that $\hat \cH_{\hat i} \subset \cH^*$ and thus, $\vcao(\hat \cH_{\hat i}) \leq \vcao(\cH^*)$.
    The proof is completed by letting $\epsilon = \sqrt{\frac{\vcao(\cH^*) + \log(1/\delta)}{m_1}}$ and $m_1 = \frac{2(\vcao(\cH^*) + \log(1/\delta))m}{2(\vcao(\cH^*) + \log(1/\delta)) + \ln(6K/\delta)}$.
\end{proof}

\printbibliography

@article{gofer16, author = {Gofer, Eyal and Mansour, Yishay}, title = {Lower Bounds on Individual Sequence Regret}, year = {2016}, issue_date = {April 2016}, publisher = {Kluwer Academic Publishers}, address = {USA}, volume = {103}, number = {1}, issn = {0885-6125}, journal = {Mach. Learn.}, month = {apr}, pages = {1–26}, numpages = {26} }

@misc{balsubramani2015sharp,
      title={Sharp Finite-Time Iterated-Logarithm Martingale Concentration}, 
      author={Akshay Balsubramani},
      year={2015},
      eprint={1405.2639},
      archivePrefix={arXiv},
      primaryClass={math.PR}
}

@inproceedings{camara2020mechanisms,
  title={Mechanisms for a no-regret agent: Beyond the common prior},
  author={Camara, Modibo K and Hartline, Jason D and Johnsen, Aleck},
  booktitle={2020 ieee 61st annual symposium on foundations of computer science (focs)},
  pages={259--270},
  year={2020},
  organization={IEEE}
}

@article{carroll2021contract,
  title={Contract Theory},
  author={Carroll, Gabriel},
  year={2021}
}

@article{carroll2015robustness,
  title={Robustness and linear contracts},
  author={Carroll, Gabriel},
  journal={American Economic Review},
  volume={105},
  number={2},
  pages={536--563},
  year={2015},
  publisher={American Economic Association 2014 Broadway, Suite 305, Nashville, TN 37203}
}

@article{blum2014learning,
  title={Learning optimal commitment to overcome insecurity},
  author={Blum, Avrim and Haghtalab, Nika and Procaccia, Ariel D},
  journal={Advances in Neural Information Processing Systems},
  volume={27},
  year={2014}
}

@inproceedings{blum2008regret,
  title={Regret minimization and the price of total anarchy},
  author={Blum, Avrim and Hajiaghayi, MohammadTaghi and Ligett, Katrina and Roth, Aaron},
  booktitle={Proceedings of the fortieth annual ACM symposium on Theory of computing},
  pages={373--382},
  year={2008}
}

@article{cai2023selling,
  title={Selling to Multiple No-Regret Buyers},
  author={Cai, Linda and Weinberg, S Matthew and Wildenhain, Evan and Zhang, Shirley},
  journal={arXiv preprint arXiv:2307.04175},
  year={2023}
}

@article{brown2023learning,
  title={Is Learning in Games Good for the Learners?},
  author={Brown, William and Schneider, Jon and Vodrahalli, Kiran},
  journal={arXiv preprint arXiv:2305.19496},
  year={2023}
}

@article{kolumbus2022and,
  title={How and why to manipulate your own agent: On the incentives of users of learning agents},
  author={Kolumbus, Yoav and Nisan, Noam},
  journal={Advances in Neural Information Processing Systems},
  volume={35},
  pages={28080--28094},
  year={2022}
}

@inproceedings{mansour2022strategizing,
  title={Strategizing against learners in bayesian games},
  author={Mansour, Yishay and Mohri, Mehryar and Schneider, Jon and Sivan, Balasubramanian},
  booktitle={Conference on Learning Theory},
  pages={5221--5252},
  year={2022},
  organization={PMLR}
}

@inproceedings{lykouris2016learning,
  title={Learning and efficiency in games with dynamic population},
  author={Lykouris, Thodoris and Syrgkanis, Vasilis and Tardos, {\'E}va},
  booktitle={Proceedings of the twenty-seventh annual ACM-SIAM symposium on Discrete algorithms},
  pages={120--129},
  year={2016},
  organization={SIAM}
}

@inproceedings{braverman2018selling,
  title={Selling to a no-regret buyer},
  author={Braverman, Mark and Mao, Jieming and Schneider, Jon and Weinberg, Matt},
  booktitle={Proceedings of the 2018 ACM Conference on Economics and Computation},
  pages={523--538},
  year={2018}
}

@inproceedings{nekipelov2015econometrics,
  title={Econometrics for learning agents},
  author={Nekipelov, Denis and Syrgkanis, Vasilis and Tardos, Eva},
  booktitle={Proceedings of the sixteenth acm conference on economics and computation},
  pages={1--18},
  year={2015}
}

@article{roughgarden2015intrinsic,
  title={Intrinsic robustness of the price of anarchy},
  author={Roughgarden, Tim},
  journal={Journal of the ACM (JACM)},
  volume={62},
  number={5},
  pages={1--42},
  year={2015},
  publisher={ACM New York, NY, USA}
}

@article{roth2020multidimensional,
  title={Multidimensional dynamic pricing for welfare maximization},
  author={Roth, Aaron and Slivkins, Aleksandrs and Ullman, Jonathan and Wu, Zhiwei Steven},
  journal={ACM Transactions on Economics and Computation (TEAC)},
  volume={8},
  number={1},
  pages={1--35},
  year={2020},
  publisher={ACM New York, NY, USA}
}

@inproceedings{roth2016watch,
  title={Watch and learn: Optimizing from revealed preferences feedback},
  author={Roth, Aaron and Ullman, Jonathan and Wu, Zhiwei Steven},
  booktitle={Proceedings of the forty-eighth annual ACM symposium on Theory of Computing},
  pages={949--962},
  year={2016}
}

@article{chen2020learning,
  title={Learning strategy-aware linear classifiers},
  author={Chen, Yiling and Liu, Yang and Podimata, Chara},
  journal={Advances in Neural Information Processing Systems},
  volume={33},
  pages={15265--15276},
  year={2020}
}

@inproceedings{dong2018strategic,
  title={Strategic classification from revealed preferences},
  author={Dong, Jinshuo and Roth, Aaron and Schutzman, Zachary and Waggoner, Bo and Wu, Zhiwei Steven},
  booktitle={Proceedings of the 2018 ACM Conference on Economics and Computation},
  pages={55--70},
  year={2018}
}

@book{bolton2004contract,
  title={Contract theory},
  author={Bolton, Patrick and Dewatripont, Mathias},
  year={2004},
  publisher={MIT press}
}

@inproceedings{dughmi2016algorithmic,
  title={Algorithmic bayesian persuasion},
  author={Dughmi, Shaddin and Xu, Haifeng},
  booktitle={Proceedings of the forty-eighth annual ACM symposium on Theory of Computing},
  pages={412--425},
  year={2016}
}

@inproceedings{cohen2019optimal,
  title={Optimal algorithm for bayesian incentive-compatible exploration},
  author={Cohen, Lee and Mansour, Yishay},
  booktitle={Proceedings of the 2019 ACM Conference on Economics and Computation},
  pages={135--151},
  year={2019}
}

@inproceedings{sellke2021price,
  title={The price of incentivizing exploration: A characterization via thompson sampling and sample complexity},
  author={Sellke, Mark and Slivkins, Aleksandrs},
  booktitle={Proceedings of the 22nd ACM Conference on Economics and Computation},
  pages={795--796},
  year={2021}
}

@inproceedings{zu2021learning,
  title={Learning to persuade on the fly: Robustness against ignorance},
  author={Zu, You and Iyer, Krishnamurthy and Xu, Haifeng},
  booktitle={Proceedings of the 22nd ACM Conference on Economics and Computation},
  pages={927--928},
  year={2021}
}

@inproceedings{haghtalab2022learning,
  title={Learning in stackelberg games with non-myopic agents},
  author={Haghtalab, Nika and Lykouris, Thodoris and Nietert, Sloan and Wei, Alexander},
  booktitle={Proceedings of the 23rd ACM Conference on Economics and Computation},
  pages={917--918},
  year={2022}
}

@inproceedings{bernasconi2023optimal,
  title={Optimal rates and efficient algorithms for online Bayesian persuasion},
  author={Bernasconi, Martino and Castiglioni, Matteo and Celli, Andrea and Marchesi, Alberto and Trov{\`o}, Francesco and Gatti, Nicola},
  booktitle={International Conference on Machine Learning},
  pages={2164--2183},
  year={2023},
  organization={PMLR}
}

@article{bernasconi2022sequential,
  title={Sequential information design: Learning to persuade in the dark},
  author={Bernasconi, Martino and Castiglioni, Matteo and Marchesi, Alberto and Gatti, Nicola and Trov{\`o}, Francesco},
  journal={Advances in Neural Information Processing Systems},
  volume={35},
  pages={15917--15928},
  year={2022}
}

@article{wu2022sequential,
  title={Sequential information design: Markov persuasion process and its efficient reinforcement learning},
  author={Wu, Jibang and Zhang, Zixuan and Feng, Zhe and Wang, Zhaoran and Yang, Zhuoran and Jordan, Michael I and Xu, Haifeng},
  journal={arXiv preprint arXiv:2202.10678},
  year={2022}
}

@inproceedings{gan2022bayesian,
  title={Bayesian persuasion in sequential decision-making},
  author={Gan, Jiarui and Majumdar, Rupak and Radanovic, Goran and Singla, Adish},
  booktitle={Proceedings of the AAAI Conference on Artificial Intelligence},
  volume={36},
  number={5},
  pages={5025--5033},
  year={2022}
}

@article{mansour2022bayesian,
  title={Bayesian exploration: Incentivizing exploration in Bayesian games},
  author={Mansour, Yishay and Slivkins, Aleksandrs and Syrgkanis, Vasilis and Wu, Zhiwei Steven},
  journal={Operations Research},
  volume={70},
  number={2},
  pages={1105--1127},
  year={2022},
  publisher={INFORMS}
}

@article{zhu2022sample,
  title={The sample complexity of online contract design},
  author={Zhu, Banghua and Bates, Stephen and Yang, Zhuoran and Wang, Yixin and Jiao, Jiantao and Jordan, Michael I},
  journal={arXiv preprint arXiv:2211.05732},
  year={2022}
}

@inproceedings{collina2023efficient,
  title={Efficient Stackelberg Strategies for Finitely Repeated Games},
  author={Collina, Natalie and Arunachaleswaran, Eshwar Ram and Kearns, Michael},
  booktitle={Proceedings of the 2023 International Conference on Autonomous Agents and Multiagent Systems},
  pages={643--651},
  year={2023}
}

@misc{haghtalab2023calibrated,
      title={Calibrated Stackelberg Games: Learning Optimal Commitments Against Calibrated Agents}, 
      author={Nika Haghtalab and Chara Podimata and Kunhe Yang},
      year={2023},
      eprint={2306.02704},
      archivePrefix={arXiv},
      primaryClass={cs.GT}
}

@inproceedings{ho2014adaptive,
  title={Adaptive contract design for crowdsourcing markets: Bandit algorithms for repeated principal-agent problems},
  author={Ho, Chien-Ju and Slivkins, Aleksandrs and Vaughan, Jennifer Wortman},
  booktitle={Proceedings of the fifteenth ACM conference on Economics and computation},
  pages={359--376},
  year={2014}
}

@article{holmstrom1979moral,
  title={Moral hazard and observability},
  author={Holmstr{\"o}m, Bengt},
  journal={The Bell journal of economics},
  pages={74--91},
  year={1979},
  publisher={JSTOR}
}

@incollection{grossman1992analysis,
  title={An analysis of the principal-agent problem},
  author={Grossman, Sanford J and Hart, Oliver D},
  booktitle={Foundations of Insurance Economics: Readings in Economics and Finance},
  pages={302--340},
  year={1992},
  publisher={Springer}
}

@article{holmstrom1987aggregation,
  title={Aggregation and linearity in the provision of intertemporal incentives},
  author={Holmstrom, Bengt and Milgrom, Paul},
  journal={Econometrica: Journal of the Econometric Society},
  pages={303--328},
  year={1987},
  publisher={JSTOR}
}

@inproceedings{dutting2022combinatorial,
  title={Combinatorial contracts},
  author={D{\"u}tting, Paul and Ezra, Tomer and Feldman, Michal and Kesselheim, Thomas},
  booktitle={2021 IEEE 62nd Annual Symposium on Foundations of Computer Science (FOCS)},
  pages={815--826},
  year={2022},
  organization={IEEE}
}

@article{chassang2013calibrated,
  title={Calibrated incentive contracts},
  author={Chassang, Sylvain},
  journal={Econometrica},
  volume={81},
  number={5},
  pages={1935--1971},
  year={2013},
  publisher={Wiley Online Library}
}

@misc{RothNotes,
  author        = {Aaron Roth},
  title         = {Uncertain: Modern Topics in Uncertainty Estimation},
  month         = {September},
  year          = {2023},
  publisher={University of Pennsylvania}
}

@article{foster1998asymptotic,
  title={Asymptotic calibration},
  author={Foster, Dean P and Vohra, Rakesh V},
  journal={Biometrika},
  volume={85},
  number={2},
  pages={379--390},
  year={1998},
  publisher={Oxford University Press}
}

@article{foster2018smooth,
  title={Smooth calibration, leaky forecasts, finite recall, and nash dynamics},
  author={Foster, Dean P and Hart, Sergiu},
  journal={Games and Economic Behavior},
  volume={109},
  pages={271--293},
  year={2018},
  publisher={Elsevier}
}

@inproceedings{GopalanKRSW22,
  author    = {Parikshit Gopalan and
               Adam Tauman Kalai and
               Omer Reingold and
               Vatsal Sharan and
               Udi Wieder},
  editor    = {Mark Braverman},
  title     = {Omnipredictors},
  booktitle = {13th Innovations in Theoretical Computer Science Conference, {ITCS}
               2022, January 31 - February 3, 2022, Berkeley, CA, {USA}},
  series    = {LIPIcs},
  volume    = {215},
  pages     = {79:1--79:21},
  publisher = {Schloss Dagstuhl - Leibniz-Zentrum f{\"{u}}r Informatik},
  year      = {2022},
  doi       = {10.4230/LIPIcs.ITCS.2022.79},
  bibsource = {dblp computer science bibliography, https://dblp.org}
}

@inproceedings{GHK23,
  author       = {Ira Globus{-}Harris and
                  Declan Harrison and
                  Michael Kearns and
                  Aaron Roth and
                  Jessica Sorrell},
  editor       = {Andreas Krause and
                  Emma Brunskill and
                  Kyunghyun Cho and
                  Barbara Engelhardt and
                  Sivan Sabato and
                  Jonathan Scarlett},
  title        = {Multicalibration as Boosting for Regression},
  booktitle    = {International Conference on Machine Learning, {ICML} 2023, 23-29 July
                  2023, Honolulu, Hawaii, {USA}},
  series       = {Proceedings of Machine Learning Research},
  volume       = {202},
  pages        = {11459--11492},
  publisher    = {{PMLR}},
  year         = {2023},
  bibsource    = {dblp computer science bibliography, https://dblp.org}
}

@article{zhao2021calibrating,
  title={Calibrating predictions to decisions: A novel approach to multi-class calibration},
  author={Zhao, Shengjia and Kim, Michael and Sahoo, Roshni and Ma, Tengyu and Ermon, Stefano},
  journal={Advances in Neural Information Processing Systems},
  volume={34},
  pages={22313--22324},
  year={2021}
}

@inproceedings{gopalan2023loss,
  title={Loss Minimization Through the Lens Of Outcome Indistinguishability},
  author={Gopalan, Parikshit and Hu, Lunjia and Kim, Michael P and Reingold, Omer and Wieder, Udi},
  booktitle={14th Innovations in Theoretical Computer Science Conference (ITCS 2023)},
  year={2023},
  organization={Schloss Dagstuhl-Leibniz-Zentrum f{\"u}r Informatik}
}

@inproceedings{GJRR23,
  author       = {Sumegha Garg and
                  Christopher Jung and
                  Omer Reingold and
                  Aaron Roth},
  title        = {Oracle Efficient Online Multicalibration and Omniprediction},
  booktitle      = {ACM-SIAM Symposium on Discrete Algorithms},
  year         = {2024}
}

@article{gopalan2023characterizing,
  title={Characterizing notions of omniprediction via multicalibration},
  author={Gopalan, Parikshit and Kim, Michael P and Reingold, Omer},
  journal={arXiv preprint arXiv:2302.06726},
  year={2023}
}

@inproceedings{hebert2018multicalibration,
  title={Multicalibration: Calibration for the (computationally-identifiable) masses},
  author={H{\'e}bert-Johnson, Ursula and Kim, Michael and Reingold, Omer and Rothblum, Guy},
  booktitle={International Conference on Machine Learning},
  pages={1939--1948},
  year={2018},
  organization={PMLR}
}

@article{kakade2008deterministic,
  title={Deterministic calibration and Nash equilibrium},
  author={Kakade, Sham M and Foster, Dean P},
  journal={Journal of Computer and System Sciences},
  volume={74},
  number={1},
  pages={115--130},
  year={2008},
  publisher={Elsevier}
}

@article{foster1999regret,
  title={Regret in the on-line decision problem},
  author={Foster, Dean P and Vohra, Rakesh},
  journal={Games and Economic Behavior},
  volume={29},
  number={1-2},
  pages={7--35},
  year={1999},
  publisher={Elsevier}
}

@article{NRRX23,
  title={High-Dimensional Prediction for Sequential Decision Making},
  author={Noarov, Georgy and Ramalingam, Ramya and Roth, Aaron and Xie, Stephan},
  journal={arXiv preprint arXiv:2310.17651},
  year={2023}
}

@article{blum2007external,
  title={From external to internal regret.},
  author={Blum, Avrim and Mansour, Yishay},
  journal={Journal of Machine Learning Research},
  volume={8},
  number={6},
  year={2007}
}

@article{castiglioni2021bayesian,
  title={Bayesian Agency: Linear versus Tractable Contracts},
  author={Castiglioni, Matteo and Marchesi, Alberto and Gatti, Nicola},
  journal={arXiv e-prints},
  pages={arXiv--2106},
  year={2021}
}

@inproceedings{cohen2022learning,
  title={Learning approximately optimal contracts},
  author={Cohen, Alon and Deligkas, Argyrios and Koren, Moran},
  booktitle={International Symposium on Algorithmic Game Theory},
  pages={331--346},
  year={2022},
  organization={Springer}
}

@misc{deng2019strategizing,
      title={Strategizing against No-regret Learners}, 
      author={Yuan Deng and Jon Schneider and Balusubramanian Sivan},
      year={2019},
      eprint={1909.13861},
      archivePrefix={arXiv},
      primaryClass={cs.GT}
}

@article{Balcan2015CommitmentWR,
  title={Commitment Without Regrets: Online Learning in Stackelberg Security Games},
  author={Maria-Florina Balcan and Avrim Blum and Nika Haghtalab and Ariel D. Procaccia},
  journal={Proceedings of the Sixteenth ACM Conference on Economics and Computation},
  year={2015}
}

@inproceedings{dutting19, author = {D\"{u}tting, Paul and Roughgarden, Tim and Talgam-Cohen, Inbal}, title = {Simple versus Optimal Contracts}, year = {2019}, isbn = {9781450367929}, publisher = {Association for Computing Machinery}, address = {New York, NY, USA}, booktitle = {Proceedings of the 2019 ACM Conference on Economics and Computation}, pages = {369–387}, numpages = {19}, location = {Phoenix, AZ, USA}, series = {EC '19} }

@article{kamenica2011bayesian,
  title={Bayesian persuasion},
  author={Kamenica, Emir and Gentzkow, Matthew},
  journal={American Economic Review},
  volume={101},
  number={6},
  pages={2590--2615},
  year={2011},
  publisher={American Economic Association}
}

@inproceedings{bousquet2020proper,
  title={Proper Learning, {H}elly Number, and an Optimal {SVM} Bound},
  author={Bousquet, Olivier and Hanneke, Steve and Moran, Shay and Zhivotovskiy, Nikita},
  booktitle={Proceedings of the 33rd Annual Conference on Learning Theory},
  year={2020}
}

@article{littlestone1986relating,
  title={Relating data compression and learnability},
  author={Littlestone, Nick and Warmuth, Manfred},
  year={1986},
  publisher={Technical report, University of California Santa Cruz}
}

@article{hanneke:16a,
author = {Hanneke, Steve},
title = {The Optimal Sample Complexity of {PAC} Learning},
journal = {Journal of Machine Learning Research},
volume = 17,
number = 38, 
pages = "1--15",
year = 2016
}

@article{hanneke2016refined,
  title={Refined error bounds for several learning algorithms},
  author={Hanneke, Steve},
  journal={The Journal of Machine Learning Research},
  volume={17},
  number={1},
  pages={4667--4721},
  year={2016},
  publisher={JMLR. org}
}

@phdthesis{hanneke:thesis,
title = {Theoretical Foundations of Active Learning},
author = {Hanneke, Steve},
school = {Machine Learning Department, School of Computer Science, Carnegie Mellon University},
year = 2009
}

@article{maurer2009empirical,
  title={Empirical Bernstein bounds and sample variance penalization},
  author={Maurer, Andreas and Pontil, Massimiliano},
  journal={arXiv preprint arXiv:0907.3740},
  year={2009}
}

@article{ball1997elementary,
  title={An elementary introduction to modern convex geometry},
  author={Ball, Keith and others},
  journal={Flavors of geometry},
  volume={31},
  pages={1--58},
  year={1997}
}

@article{ben20152,
  title={2 Notes on Classes with {V}apnik-{C}hervonenkis Dimension 1},
  author={Ben-David, Shai},
  journal={arXiv preprint arXiv:1507.05307},
  year={2015}
}

@article{hanneke2015minimax,
  title={Minimax analysis of active learning},
  author={Hanneke, Steve and Yang, Liu},
  journal={The Journal of Machine Learning Research},
  volume={16},
  number={1},
  pages={3487--3602},
  year={2015},
  publisher={JMLR. org}
}

@article{auer:07,
author = {Auer, Peter and Ortner, Ronald},
title = {A New {PAC} bound for Intersection-Closed Concept Classes},
journal = {Machine Learning}, 
volume = 66,
number = {2-3},
pages = "151--163",
year = 2007
}

@article{helmbold:90,
author = {D. Helmbold and R. Sloan and M. Warmuth},
title = {Learning Nested Differences of Intersection-Closed Concept Classes},
journal = {Machine Learning},
volume = 5,
number = 2,
year = 1990,
pages = "165--196"
}

@inproceedings{steinhardt2017certified,
  title={Certified defenses for data poisoning attacks},
  author={Steinhardt, Jacob and Koh, Pang Wei W and Liang, Percy S},
  booktitle={Advances in neural information processing systems},
  pages={3517--3529},
  year={2017}
}

@inproceedings{koh2017understanding,
  title={Understanding Black-box Predictions via Influence Functions},
  author={Koh, Pang Wei and Liang, Percy},
  booktitle={International Conference on Machine Learning},
  pages={1885--1894},
  year={2017}
}

@inproceedings{shafahi2018poison,
  title={Poison frogs! targeted clean-label poisoning attacks on neural networks},
  author={Shafahi, Ali and Huang, W Ronny and Najibi, Mahyar and Suciu, Octavian and Studer, Christoph and Dumitras, Tudor and Goldstein, Tom},
  booktitle={Advances in Neural Information Processing Systems},
  pages={6103--6113},
  year={2018}
}

@inproceedings{suciu2018does,
  title={When does machine learning {FAIL}? generalized transferability for evasion and poisoning attacks},
  author={Suciu, Octavian and Marginean, Radu and Kaya, Yigitcan and Daume III, Hal and Dumitras, Tudor},
  booktitle={27th {USENIX} Security Symposium ({USENIX} Security 18)},
  pages={1299--1316},
  year={2018}
}

@article{levine2020deep,
  title={Deep partition aggregation: Provable defense against general poisoning attacks},
  author={Levine, Alexander and Feizi, Soheil},
  journal={arXiv preprint arXiv:2006.14768},
  year={2020}
}

@article{xu2012robustness,
  title={Robustness and generalization},
  author={Xu, Huan and Mannor, Shie},
  journal={Machine learning},
  volume={86},
  number={3},
  pages={391--423},
  year={2012},
  publisher={Springer}
}

@inproceedings{bubeck2019adversarial,
  title={Adversarial examples from computational constraints},
  author={Bubeck, S{\'e}bastien and Lee, Yin Tat and Price, Eric and Razenshteyn, Ilya},
  booktitle={International Conference on Machine Learning},
  pages={831--840},
  year={2019},
  organization={PMLR}
}

@inproceedings{cullina2018pac,
  title={{PAC}-learning in the presence of adversaries},
  author={Cullina, Daniel and Bhagoji, Arjun Nitin and Mittal, Prateek},
  booktitle={Advances in Neural Information Processing Systems},
  pages={230--241},
  year={2018}
}

@inproceedings{montasser2019vc,
  title={{VC} Classes are Adversarially Robustly Learnable, but Only Improperly},
  author={Montasser, Omar and Hanneke, Steve and Srebro, Nathan},
  booktitle={Conference on Learning Theory},
  pages={2512--2530},
  year={2019}
}

@article{goodfellow2014explaining,
  title={Explaining and harnessing adversarial examples},
  author={Goodfellow, Ian J and Shlens, Jonathon and Szegedy, Christian},
  journal={arXiv preprint arXiv:1412.6572},
  year={2014}
}

@article{szegedy2013intriguing,
  title={Intriguing properties of neural networks},
  author={Szegedy, Christian and Zaremba, Wojciech and Sutskever, Ilya and Bruna, Joan and Erhan, Dumitru and Goodfellow, Ian and Fergus, Rob},
  journal={arXiv preprint arXiv:1312.6199},
  year={2013}
}

@book{quionero2009dataset,
  title={Dataset shift in machine learning},
  author={Quionero-Candela, Joaquin and Sugiyama, Masashi and Schwaighofer, Anton and Lawrence, Neil D},
  year={2009},
  publisher={The MIT Press}
}

@article{montasser2020efficiently,
  title={Efficiently Learning Adversarially Robust Halfspaces with Noise},
  author={Montasser, Omar and Goel, Surbhi and Diakonikolas, Ilias and Srebro, Nathan},
  journal={arXiv preprint arXiv:2005.07652},
  year={2020}
}

@Article{haussler:94,
author = {Haussler, David and Littlestone, Nick and Warmuth, Manfred},
title = {Predicting $\{0,1\}$-Functions on Randomly Drawn Points},
journal = {Information and Computation},
year = {1994},
volume = {115},
number = 2,
pages = "248--292"
}

@article{darnstadt:15,
author = {Darnst\"{a}dt, Malte},
title = {The Optimal {PAC} Bound for Intersection-Closed Concept Classes},
journal = {Information Processing Letters},
volume = 115,
number = 4,
year = 2015,
pages = "458--461"
}

@book{vapnik:74,
author = {V. Vapnik and A. Chervonenkis},
title = {Theory of Pattern Recognition},
publisher = {Nauka, Moscow},
year = 1974
}

@article{blumer1989learnability,
  title={Learnability and the {Vapnik-Chervonenkis} dimension},
  author={Blumer, Anselm and Ehrenfeucht, Andrzej and Haussler, David and Warmuth, Manfred K},
  journal={Journal of the ACM (JACM)},
  volume={36},
  number={4},
  pages={929--965},
  year={1989},
  publisher={ACM New York, NY, USA}
}

@inproceedings{ma2019data,
  title={Data Poisoning against Differentially-Private Learners: Attacks and Defenses},
  author={Ma, Yuzhe and Zhu, Xiaojin Zhu and Hsu, Justin},
  booktitle={International Joint Conference on Artificial Intelligence},
  year={2019}
}

@inproceedings{gao2021learning,
  title={Learning and Certification under Instance-targeted Poisoning},
  author={Gao, Ji and Karbasi, Amin and Mahmoody, Mohammad},
  booktitle = {Conference on Uncertainty in Artificial Intelligence},
  year={2021}
}

@inproceedings{mahloujifar2017blockwise,
  title={Blockwise p-tampering attacks on cryptographic primitives, extractors, and learners},
  author={Mahloujifar, Saeed and Mahmoody, Mohammad},
  booktitle={Theory of Cryptography Conference},
  pages={245--279},
  year={2017},
  organization={Springer}
}

@inproceedings{mahloujifar2018learning,
  title={Learning under $ p $-Tampering Attacks},
  author={Mahloujifar, Saeed and Diochnos, Dimitrios I and Mahmoody, Mohammad},
  booktitle={Algorithmic Learning Theory},
  pages={572--596},
  year={2018},
  organization={PMLR}
}

@inproceedings{mahloujifar2019can,
  title={Can Adversarially Robust Learning Leverage Computational Hardness?},
  author={Mahloujifar, Saeed and Mahmoody, Mohammad},
  booktitle={Algorithmic Learning Theory},
  pages={581--609},
  year={2019},
  organization={PMLR}
}

@inproceedings{mahloujifar2019curse,
  title={The curse of concentration in robust learning: Evasion and poisoning attacks from concentration of measure},
  author={Mahloujifar, Saeed and Diochnos, Dimitrios I and Mahmoody, Mohammad},
  booktitle={Proceedings of the AAAI Conference on Artificial Intelligence},
  volume={33},
  number={01},
  pages={4536--4543},
  year={2019}
}

@inproceedings{mahloujifar2019universal,
  title={Universal multi-party poisoning attacks},
  author={Mahloujifar, Saeed and Mahmoody, Mohammad and Mohammed, Ameer},
  booktitle={International Conference on Machine Learing (ICML)},
  year={2019}
}

@inproceedings{etesami2020computational,
  title={Computational concentration of measure: Optimal bounds, reductions, and more},
  author={Etesami, Omid and Mahloujifar, Saeed and Mahmoody, Mohammad},
  booktitle={Proceedings of the Fourteenth Annual ACM-SIAM Symposium on Discrete Algorithms},
  pages={345--363},
  year={2020},
  organization={SIAM}
}

@inproceedings{biggio2012poisoning,
  title={Poisoning attacks against support vector machines},
  author={Biggio, Battista and Nelson, Blaine and Laskov, Pavel},
  booktitle={Proceedings of the 29th International Coference on International Conference on Machine Learning},
  pages={1467--1474},
  year={2012}
}

@inproceedings{barreno2006can,
  title={Can machine learning be secure?},
  author={Barreno, Marco and Nelson, Blaine and Sears, Russell and Joseph, Anthony D and Tygar, J Doug},
  booktitle={Proceedings of the 2006 ACM Symposium on Information, computer and communications security},
  pages={16--25},
  year={2006}
}

@article{papernot2016towards,
  title={Towards the science of security and privacy in machine learning},
  author={Papernot, Nicolas and McDaniel, Patrick and Sinha, Arunesh and Wellman, Michael},
  journal={arXiv preprint arXiv:1611.03814},
  year={2016}
}

@article{ehrenfeucht1989general,
  title={A general lower bound on the number of examples needed for learning},
  author={Ehrenfeucht, Andrzej and Haussler, David and Kearns, Michael and Valiant, Leslie},
  journal={Information and Computation},
  volume={82},
  number={3},
  pages={247--261},
  year={1989},
  publisher={Elsevier}
}

@inproceedings{daniely2014optimal,
  title={Optimal learners for multiclass problems},
  author={Daniely, Amit and Shalev-Shwartz, Shai},
  booktitle={Conference on Learning Theory},
  pages={287--316},
  year={2014},
  organization={PMLR}
}

@article{david2016supervised,
  title={Supervised learning through the lens of compression},
  author={David, Ofir and Moran, Shay and Yehudayoff, Amir},
  journal={Advances in Neural Information Processing Systems},
  volume={29},
  year={2016}
}

@article{graepel2005pac,
  title={PAC-Bayesian compression bounds on the prediction error of learning algorithms for classification},
  author={Graepel, Thore and Herbrich, Ralf and Shawe-Taylor, John},
  journal={Machine Learning},
  volume={59},
  number={1-2},
  pages={55--76},
  year={2005},
  publisher={Springer}
}

@book{schapire2012boosting,
  title={Boosting: Foundations and Algorithms},
  author={Schapire, Robert E and Freund, Yoav},
  year={2012},
  publisher={MIT Press}
}

@inproceedings{ben2014sample,
  title={The sample complexity of agnostic learning under deterministic labels},
  author={Ben-David, Shai and Urner, Ruth},
  booktitle={Conference on Learning Theory},
  pages={527--542},
  year={2014},
  organization={PMLR}
}

@article{haussler1994predicting,
  title={Predicting $\{$0, 1$\}$-functions on randomly drawn points},
  author={Haussler, David and Littlestone, Nick and Warmuth, Manfred K},
  journal={Information and Computation},
  volume={115},
  number={2},
  pages={248--292},
  year={1994},
  publisher={Elsevier}
}

@article{hopkins2021realizable,
  title={Realizable Learning is All You Need},
  author={Hopkins, Max and Kane, Daniel and Lovett, Shachar and Mahajan, Gaurav},
  journal={arXiv preprint arXiv:2111.04746},
  year={2021}
}

@article{Lyle2020,
  title={On the benefits of invariance in neural networks},
  author={Lyle, Clare and van der Wilk, Mark and Kwiatkowska, Marta and Gal, Yarin and Bloem-Reddy, Benjamin},
  journal={arXiv preprint arXiv:2005.00178},
  year={2020}
}

@inproceedings{Dao2019,
  title={A kernel theory of modern data augmentation},
  author={Dao, Tri and Gu, Albert and Ratner, Alexander and Smith, Virginia and De Sa, Chris and R{\'e}, Christopher},
  booktitle={International Conference on Machine Learning},
  pages={1528--1537},
  year={2019},
  organization={PMLR}
}

@article{Chen2020,
  title={A Group-Theoretic Framework for Data Augmentation},
  author={Chen, Shuxiao and Dobriban, Edgar and Lee, Jane H},
  journal={Journal of Machine Learning Research},
  volume={21},
  pages={1--71},
  year={2020}
}

@article{wood1996representation,
  title={Representation theory and invariant neural networks},
  author={Wood, Jeffrey and Shawe-Taylor, John},
  journal={Discrete applied mathematics},
  volume={69},
  number={1-2},
  pages={33--60},
  year={1996},
  publisher={Elsevier}
}

@article{bloem2020probabilistic,
  title={Probabilistic Symmetries and Invariant Neural Networks.},
  author={Bloem-Reddy, Benjamin and Teh, Yee Whye},
  journal={J. Mach. Learn. Res.},
  volume={21},
  pages={90--1},
  year={2020}
}

@inproceedings{ravanbakhsh2017equivariance,
  title={Equivariance through parameter-sharing},
  author={Ravanbakhsh, Siamak and Schneider, Jeff and Poczos, Barnabas},
  booktitle={International Conference on Machine Learning},
  pages={2892--2901},
  year={2017},
  organization={PMLR}
}

@inproceedings{kondor2018generalization,
  title={On the generalization of equivariance and convolution in neural networks to the action of compact groups},
  author={Kondor, Risi and Trivedi, Shubhendu},
  booktitle={International Conference on Machine Learning},
  pages={2747--2755},
  year={2018},
  organization={PMLR}
}

@inproceedings{fawzi2016adaptive,
  title={Adaptive data augmentation for image classification},
  author={Fawzi, Alhussein and Samulowitz, Horst and Turaga, Deepak and Frossard, Pascal},
  booktitle={2016 IEEE international conference on image processing (ICIP)},
  pages={3688--3692},
  year={2016},
  organization={Ieee}
}

@article{cubuk2018autoaugment,
  title={Autoaugment: Learning augmentation policies from data},
  author={Cubuk, Ekin D and Zoph, Barret and Mane, Dandelion and Vasudevan, Vijay and Le, Quoc V},
  journal={arXiv preprint arXiv:1805.09501},
  year={2018}
}

@article{chatzipantazis2021learning,
  title={Learning Augmentation Distributions using Transformed Risk Minimization},
  author={Chatzipantazis, Evangelos and Pertigkiozoglou, Stefanos and Dobriban, Edgar and Daniilidis, Kostas},
  journal={arXiv preprint arXiv:2111.08190},
  year={2021}
}

@incollection{fukushima1982neocognitron,
  title={Neocognitron: A self-organizing neural network model for a mechanism of visual pattern recognition},
  author={Fukushima, Kunihiko and Miyake, Sei},
  booktitle={Competition and cooperation in neural nets},
  pages={267--285},
  year={1982},
  publisher={Springer}
}

@article{lecun1989backpropagation,
  title={Backpropagation applied to handwritten zip code recognition},
  author={LeCun, Yann and Boser, Bernhard and Denker, John S and Henderson, Donnie and Howard, Richard E and Hubbard, Wayne and Jackel, Lawrence D},
  journal={Neural computation},
  volume={1},
  number={4},
  pages={541--551},
  year={1989},
  publisher={MIT Press}
}

@inproceedings{cohen2016group,
  title={Group equivariant convolutional networks},
  author={Cohen, Taco and Welling, Max},
  booktitle={International conference on machine learning},
  pages={2990--2999},
  year={2016},
  organization={PMLR}
}

@inproceedings{dieleman2016exploiting,
  title={Exploiting cyclic symmetry in convolutional neural networks},
  author={Dieleman, Sander and De Fauw, Jeffrey and Kavukcuoglu, Koray},
  booktitle={International conference on machine learning},
  pages={1889--1898},
  year={2016},
  organization={PMLR}
}

@inproceedings{worrall2017harmonic,
  title={Harmonic networks: Deep translation and rotation equivariance},
  author={Worrall, Daniel E and Garbin, Stephan J and Turmukhambetov, Daniyar and Brostow, Gabriel J},
  booktitle={Proceedings of the IEEE Conference on Computer Vision and Pattern Recognition},
  pages={5028--5037},
  year={2017}
}

@article{gontijo2020affinity,
  title={Affinity and diversity: Quantifying mechanisms of data augmentation},
  author={Gontijo-Lopes, Raphael and Smullin, Sylvia J and Cubuk, Ekin D and Dyer, Ethan},
  journal={arXiv preprint arXiv:2002.08973},
  year={2020}
}

@article{krizhevsky2012imagenet,
  title={Imagenet classification with deep convolutional neural networks},
  author={Krizhevsky, Alex and Sutskever, Ilya and Hinton, Geoffrey E},
  journal={Advances in neural information processing systems},
  volume={25},
  year={2012}
}

@inproceedings{elesedy2021linear,
  title={Provably strict generalisation benefit for equivariant models},
  author={Elesedy, Bryn and Zaidi, Sheheryar},
  booktitle={International Conference on Machine Learning},
  pages={2959--2969},
  year={2021},
  organization={PMLR}
}

@article{elesedy2021kernel,
  title={Provably Strict Generalisation Benefit for Invariance in Kernel Methods},
  author={Elesedy, Bryn},
  journal={Advances in Neural Information Processing Systems},
  volume={34},
  year={2021}
}

@inproceedings{elesedy2022group,
  title={Group Symmetry in PAC Learning},
  author={Elesedy, Bryn},
  booktitle={ICLR 2022 Workshop on Geometrical and Topological Representation Learning},
  year={2022}
}

@inproceedings{mei2021learning,
  title={Learning with invariances in random features and kernel models},
  author={Mei, Song and Misiakiewicz, Theodor and Montanari, Andrea},
  booktitle={Conference on Learning Theory},
  pages={3351--3418},
  year={2021},
  organization={PMLR}
}

@article{bietti2021sample,
  title={On the Sample Complexity of Learning under Geometric Stability},
  author={Bietti, Alberto and Venturi, Luca and Bruna, Joan},
  journal={Advances in Neural Information Processing Systems},
  volume={34},
  year={2021}
}

@article{shen2022data,
  title={Data Augmentation as Feature Manipulation: a story of desert cows and grass cows},
  author={Shen, Ruoqi and Bubeck, S{\'e}bastien and Gunasekar, Suriya},
  journal={arXiv preprint arXiv:2203.01572},
  year={2022}
}

@inproceedings{wang2020tent,
  author    = {Dequan Wang and
               Evan Shelhamer and
               Shaoteng Liu and
               Bruno A. Olshausen and
               Trevor Darrell},
  title     = {Tent: Fully Test-Time Adaptation by Entropy Minimization},
  booktitle = {International Conference on Learning Representations},
  year      = {2021}
}

@article{kang2019incentive,
  title={Incentive mechanism for reliable federated learning: A joint optimization approach to combining reputation and contract theory},
  author={Kang, Jiawen and Xiong, Zehui and Niyato, Dusit and Xie, Shengli and Zhang, Junshan},
  journal={IEEE Internet of Things Journal},
  volume={6},
  number={6},
  pages={10700--10714},
  year={2019},
  publisher={IEEE}
}

@inproceedings{zeng2020fmore,
  title={Fmore: An incentive scheme of multi-dimensional auction for federated learning in mec},
  author={Zeng, Rongfei and Zhang, Shixun and Wang, Jiaqi and Chu, Xiaowen},
  booktitle={2020 IEEE 40th International Conference on Distributed Computing Systems (ICDCS)},
  pages={278--288},
  year={2020},
  organization={IEEE}
}

@article{tian2021contract,
  title={A Contract Theory based Incentive Mechanism for Federated Learning},
  author={Tian, Mengmeng and Chen, Yuxin and Liu, Yuan and Xiong, Zehui and Leung, Cyril and Miao, Chunyan},
  journal={arXiv e-prints},
  pages={arXiv--2108},
  year={2021}
}

@article{zhang2022enabling,
  title={Enabling long-term cooperation in cross-silo federated learning: A repeated game perspective},
  author={Zhang, Ning and Ma, Qian and Chen, Xu},
  journal={IEEE Transactions on Mobile Computing},
  year={2022},
  publisher={IEEE}
}

@article{liu2020incentives,
  title={Incentives for federated learning: a hypothesis elicitation approach},
  author={Liu, Yang and Wei, Jiaheng},
  journal={arXiv preprint arXiv:2007.10596},
  year={2020}
}

@inproceedings{zhang2021incentive,
  title={Incentive mechanism for horizontal federated learning based on reputation and reverse auction},
  author={Zhang, Jingwen and Wu, Yuezhou and Pan, Rong},
  booktitle={Proceedings of the Web Conference 2021},
  pages={947--956},
  year={2021}
}

@article{cong2020vcg,
  title={A VCG-based fair incentive mechanism for federated learning},
  author={Cong, Mingshu and Yu, Han and Weng, Xi and Qu, Jiabao and Liu, Yang and Yiu, Siu Ming},
  journal={arXiv preprint arXiv:2008.06680},
  year={2020}
}

@article{karimireddy2022mechanisms,
  title={Mechanisms that Incentivize Data Sharing in Federated Learning},
  author={Karimireddy, Sai Praneeth and Guo, Wenshuo and Jordan, Michael I},
  journal={arXiv preprint arXiv:2207.04557},
  year={2022}
}

@article{zhan2020learning,
  title={A learning-based incentive mechanism for federated learning},
  author={Zhan, Yufeng and Li, Peng and Qu, Zhihao and Zeng, Deze and Guo, Song},
  journal={IEEE Internet of Things Journal},
  volume={7},
  number={7},
  pages={6360--6368},
  year={2020},
  publisher={IEEE}
}

@inproceedings{blum2021one,
  title={One for one, or all for all: Equilibria and optimality of collaboration in federated learning},
  author={Blum, Avrim and Haghtalab, Nika and Phillips, Richard Lanas and Shao, Han},
  booktitle={International Conference on Machine Learning},
  pages={1005--1014},
  year={2021},
  organization={PMLR}
}

@article{woodworth2020minibatch,
  title={Minibatch vs local sgd for heterogeneous distributed learning},
  author={Woodworth, Blake E and Patel, Kumar Kshitij and Srebro, Nati},
  journal={Advances in Neural Information Processing Systems},
  volume={33},
  pages={6281--6292},
  year={2020}
}

@article{dekel2012optimal,
  title={Optimal Distributed Online Prediction Using Mini-Batches.},
  author={Dekel, Ofer and Gilad-Bachrach, Ran and Shamir, Ohad and Xiao, Lin},
  journal={Journal of Machine Learning Research},
  volume={13},
  number={1},
  year={2012}
}

@article{mcmahan2016communication,
  title={Communication-efficient learning of deep networks from decentralized data (2016)},
  author={McMahan, H Brendan and Moore, Eider and Ramage, Daniel and Hampson, S and y Arcas, B Ag{\"u}era},
  journal={arXiv preprint arXiv:1602.05629},
  year={2016}
}

@article{kairouz2019advances,
  title={Advances and open problems in federated learning. CoRR},
  author={Kairouz, Peter and McMahan, H Brendan and Avent, Brendan and Bellet, Aur{\'e}lien and Bennis, Mehdi and Bhagoji, Arjun Nitin and Bonawitz, Keith and Charles, Zachary and Cormode, Graham and Cummings, Rachel and others},
  journal={arXiv preprint arXiv:1912.04977},
  year={2019}
}

@article{myerson1981optimal,
  title={Optimal auction design},
  author={Myerson, Roger B},
  journal={Mathematics of operations research},
  volume={6},
  number={1},
  pages={58--73},
  year={1981},
  publisher={INFORMS}
}

@article{huber2001gaining,
  title={Gaining competitive advantage through customer value oriented management},
  author={Huber, Frank and Herrmann, Andreas and Morgan, Robert E},
  journal={Journal of consumer marketing},
  volume={18},
  number={1},
  pages={41--53},
  year={2001},
  publisher={MCB UP Ltd}
}

@book{borgers2015introduction,
  title={An introduction to the theory of mechanism design},
  author={B{\"o}rgers, Tilman},
  year={2015},
  publisher={Oxford University Press, USA}
}

@article{mcmahan2016federated,
  title={Federated learning of deep networks using model averaging},
  author={McMahan, H Brendan and Moore, Eider and Ramage, Daniel and y Arcas, Blaise Ag{\"u}era},
  journal={arXiv preprint arXiv:1602.05629},
  year={2016},
  publisher={Technical report}
}

@article{karimireddy2020mime,
  title={Mime: Mimicking centralized stochastic algorithms in federated learning},
  author={Karimireddy, Sai Praneeth and Jaggi, Martin and Kale, Satyen and Mohri, Mehryar and Reddi, Sashank J and Stich, Sebastian U and Suresh, Ananda Theertha},
  journal={arXiv preprint arXiv:2008.03606},
  year={2020}
}

@article{shi2022fedfaim,
  title={FedFAIM: A Model Performance-based Fair Incentive Mechanism for Federated Learning},
  author={Shi, Zhuan and Zhang, Lan and Yao, Zhenyu and Lyu, Lingjuan and Chen, Cen and Wang, Li and Wang, Junhao and Li, Xiang-Yang},
  journal={IEEE Transactions on Big Data},
  year={2022},
  publisher={IEEE}
}

@article{xu2021gradient,
  title={Gradient driven rewards to guarantee fairness in collaborative machine learning},
  author={Xu, Xinyi and Lyu, Lingjuan and Ma, Xingjun and Miao, Chenglin and Foo, Chuan Sheng and Low, Bryan Kian Hsiang},
  journal={Advances in Neural Information Processing Systems},
  volume={34},
  pages={16104--16117},
  year={2021}
}

@article{xu2020reputation,
  title={A reputation mechanism is all you need: Collaborative fairness and adversarial robustness in federated learning},
  author={Xu, Xinyi and Lyu, Lingjuan},
  journal={arXiv preprint arXiv:2011.10464},
  year={2020}
}

@inproceedings{ghorbani2019data,
  title={Data shapley: Equitable valuation of data for machine learning},
  author={Ghorbani, Amirata and Zou, James},
  booktitle={International Conference on Machine Learning},
  pages={2242--2251},
  year={2019},
  organization={PMLR}
}

@article{krizhevsky2009learning,
  title={Learning multiple layers of features from tiny images},
  author={Krizhevsky, Alex and Hinton, Geoffrey and others},
  year={2009},
  journal={Citeseer}
}

@article{bergen2012genome,
  title={Genome-wide association studies of schizophrenia: does bigger lead to better results?},
  author={Bergen, Sarah E and Petryshen, Tracey L},
  journal={Current opinion in psychiatry},
  volume={25},
  number={2},
  pages={76--82},
  year={2012},
  publisher={LWW}
}

@inproceedings{hu2020trading,
  title={Trading data for learning: Incentive mechanism for on-device federated learning},
  author={Hu, Rui and Gong, Yanmin},
  booktitle={GLOBECOM 2020-2020 IEEE Global Communications Conference},
  pages={1--6},
  year={2020},
  organization={IEEE}
}

@inproceedings{sim2020collaborative,
  title={Collaborative machine learning with incentive-aware model rewards},
  author={Sim, Rachael Hwee Ling and Zhang, Yehong and Chan, Mun Choon and Low, Bryan Kian Hsiang},
  booktitle={International Conference on Machine Learning},
  pages={8927--8936},
  year={2020},
  organization={PMLR}
}

@inproceedings{tang2021incentive,
  title={An incentive mechanism for cross-silo federated learning: A public goods perspective},
  author={Tang, Ming and Wong, Vincent WS},
  booktitle={IEEE INFOCOM 2021-IEEE Conference on Computer Communications},
  pages={1--10},
  year={2021},
  organization={IEEE}
}

@article{zheng2021fl,
  title={FL-Market: Trading Private Models in Federated Learning},
  author={Zheng, Shuyuan and Cao, Yang and Yoshikawa, Masatoshi and Li, Huizhong and Yan, Qiang},
  journal={arXiv e-prints},
  pages={arXiv--2106},
  year={2021}
}

@article{cho2022federate,
  title={To Federate or Not To Federate: Incentivizing Client Participation in Federated Learning},
  author={Cho, Yae Jee and Jhunjhunwala, Divyansh and Li, Tian and Smith, Virginia and Joshi, Gauri},
  journal={arXiv preprint arXiv:2205.14840},
  year={2022}
}

@incollection{richardson2020budget,
  title={Budget-bounded incentives for federated learning},
  author={Richardson, Adam and Filos-Ratsikas, Aris and Faltings, Boi},
  booktitle={Federated Learning},
  pages={176--188},
  year={2020},
  publisher={Springer}
}

@article{blum2017collaborative,
  title={Collaborative PAC learning},
  author={Blum, Avrim and Haghtalab, Nika and Procaccia, Ariel D and Qiao, Mingda},
  journal={Advances in Neural Information Processing Systems},
  volume={30},
  year={2017}
}

@inproceedings{li2019privacy,
	title={Privacy-preserving federated brain tumour segmentation},
	author={Li, Wenqi and Milletari, Fausto and Xu, Daguang and Rieke, Nicola and Hancox, Jonny and Zhu, Wentao and Baust, Maximilian and Cheng, Yan and Ourselin, Sebastien and Cardoso, M Jorge and others},
	booktitle={International workshop on machine learning in medical imaging},
	pages={133--141},
	year={2019},
	organization={Springer}
}

@article{ezzeldin2021fairfed,
  title={Fairfed: Enabling group fairness in federated learning},
  author={Ezzeldin, Yahya H and Yan, Shen and He, Chaoyang and Ferrara, Emilio and Avestimehr, Salman},
  journal={arXiv preprint arXiv:2110.00857},
  year={2021}
}

@inproceedings{roth2020federated,
  title={Federated learning for breast density classification: A real-world implementation},
  author={Roth, Holger R and Chang, Ken and Singh, Praveer and Neumark, Nir and Li, Wenqi and Gupta, Vikash and Gupta, Sharut and Qu, Liangqiong and Ihsani, Alvin and Bizzo, Bernardo C and others},
  booktitle={Domain Adaptation and Representation Transfer, and Distributed and Collaborative Learning: Second MICCAI Workshop, DART 2020, and First MICCAI Workshop, DCL 2020, Held in Conjunction with MICCAI 2020, Lima, Peru, October 4--8, 2020, Proceedings 2},
  pages={181--191},
  year={2020},
  organization={Springer}
}

@article{powell2019nvidia,
	title={Nvidia clara federated learning to deliver ai to hospitals while protecting patient data},
	author={Powell, Kimberly},
	journal={Nvidia Blog},
	year={2019}
}

@misc{mcmahan_ramage_2017, 
	title={Federated Learning: Collaborative Machine Learning without Centralized Training Data}, 
	url={https://ai.googleblog.com/2017/04/federated-learning-collaborative.html}, 
	journal={Google AI Blog}, 
	author={McMahan, Brendan and Ramage, Daniel}, 
	year={2017}, 
	month={Apr}
}

@misc{apple, 
	title={Designing for Privacy - WWDC19 - Videos}, 
	url={https://developer.apple.com/videos/play/wwdc2019/708}, 
	journal={Apple Developer}, 
	author={Apple}
}

@article{paulik2021federated,
  title={Federated evaluation and tuning for on-device personalization: System design \& applications},
  author={Paulik, Matthias and Seigel, Matt and Mason, Henry and Telaar, Dominic and Kluivers, Joris and van Dalen, Rogier and Lau, Chi Wai and Carlson, Luke and Granqvist, Filip and Vandevelde, Chris and others},
  journal={arXiv preprint arXiv:2102.08503},
  year={2021}
}

@inproceedings{patel2022distributed,
  title={Federated Online and Bandit Convex Optimization},
  author={Patel, Kumar Kshitij and Wang, Lingxiao and Saha, Aadirupa and Srebro, Nathan},
  booktitle={International Conference on Machine Learning},
  year={2023}
}

@article{huang2021federated,
  title={Federated linear contextual bandits},
  author={Huang, Ruiquan and Wu, Weiqiang and Yang, Jing and Shen, Cong},
  journal={Advances in neural information processing systems},
  volume={34},
  pages={27057--27068},
  year={2021}
}

@article{patel2022towards,
  title={Towards Optimal Communication Complexity in Distributed Non-Convex Optimization},
  author={Patel, Kumar Kshitij and Wang, Lingxiao and Woodworth, Blake E and Bullins, Brian and Srebro, Nati},
  journal={Advances in Neural Information Processing Systems},
  volume={35},
  pages={13316--13328},
  year={2022}
}

@article{wei2023incentivized,
  title={Incentivized communication for federated bandits},
  author={Wei, Zhepei and Li, Chuanhao and Xu, Haifeng and Wang, Hongning},
  journal={arXiv preprint arXiv:2309.11702},
  year={2023}
}

@article{wang2022friends,
  title={Friends to help: Saving federated learning from client dropout},
  author={Wang, Heqiang and Xu, Jie},
  journal={arXiv preprint arXiv:2205.13222},
  year={2022}
}

@article{xu2023stabilizing,
  title={Stabilizing and Improving Federated Learning with Non-IID Data and Client Dropout in IoT Systems},
  author={Xu, Jian and Yang, Meiling and Ding, Wenbo and Huang, Shao-Lun},
  journal={arXiv preprint arXiv:2303.06314},
  year={2023}
}

@article{huang2023evaluating,
  title={Evaluating and Incentivizing Diverse Data Contributions in Collaborative Learning},
  author={Huang, Baihe and Karimireddy, Sai Praneeth and Jordan, Michael I},
  journal={arXiv preprint arXiv:2306.05592},
  year={2023}
}

@article{haghtalab2022demand,
  title={On-demand sampling: Learning optimally from multiple distributions},
  author={Haghtalab, Nika and Jordan, Michael and Zhao, Eric},
  journal={Advances in Neural Information Processing Systems},
  volume={35},
  pages={406--419},
  year={2022}
}

@article{nguyen2018improved,
  title={Improved algorithms for collaborative PAC learning},
  author={Nguyen, Huy and Zakynthinou, Lydia},
  journal={Advances in Neural Information Processing Systems},
  volume={31},
  year={2018}
}

@inproceedings{donahue2021model,
  title={Model-sharing games: Analyzing federated learning under voluntary participation},
  author={Donahue, Kate and Kleinberg, Jon},
  booktitle={Proceedings of the AAAI Conference on Artificial Intelligence},
  volume={35},
  number={6},
  pages={5303--5311},
  year={2021}
}

@article{littlestone1988learning,
  title={Learning quickly when irrelevant attributes abound: A new linear-threshold algorithm},
  author={Littlestone, Nick},
  journal={Machine learning},
  volume={2},
  number={4},
  pages={285--318},
  year={1988},
  publisher={Springer}
}

@inproceedings{ben2009agnostic,
  title={Agnostic Online Learning.},
  author={Ben-David, Shai and P{\'a}l, D{\'a}vid and Shalev-Shwartz, Shai},
  booktitle={COLT},
  volume={3},
  pages={1},
  year={2009}
}

@article{dasgupta2004analysis,
  title={Analysis of a greedy active learning strategy},
  author={Dasgupta, Sanjoy},
  journal={Advances in neural information processing systems},
  volume={17},
  year={2004}
}

@inproceedings{kosaraju2002optimal,
  title={On an optimal split tree problem},
  author={Kosaraju, S Rao and Przytycka, Teresa M and Borgstrom, Ryan},
  booktitle={Algorithms and Data Structures: 6th International Workshop, WADS’99 Vancouver, Canada, August 11--14, 1999 Proceedings},
  pages={157--168},
  year={2002},
  organization={Springer}
}

@article{golovin2010near,
  title={Near-optimal bayesian active learning with noisy observations},
  author={Golovin, Daniel and Krause, Andreas and Ray, Debajyoti},
  journal={Advances in Neural Information Processing Systems},
  volume={23},
  year={2010}
}

@inproceedings{chakaravarthy2007decision,
  title={Decision trees for entity identification: Approximation algorithms and hardness results},
  author={Chakaravarthy, Venkatesan T and Pandit, Vinayaka and Roy, Sambuddha and Awasthi, Pranjal and Mohania, Mukesh},
  booktitle={Proceedings of the twenty-sixth ACM SIGMOD-SIGACT-SIGART symposium on Principles of database systems},
  pages={53--62},
  year={2007}
}

@inproceedings{blum2020advancing,
  author =	{Avrim Blum and Thodoris Lykouris},
  title =	{{Advancing Subgroup Fairness via Sleeping Experts}},
  booktitle =	{11th Innovations in Theoretical Computer Science Conference (ITCS 2020)},
  pages =	{55:1--55:24},
  series =	{Leibniz International Proceedings in Informatics (LIPIcs)},
  ISBN =	{978-3-95977-134-4},
  ISSN =	{1868-8969},
  year =	{2020},
  volume =	{151},
  editor =	{Thomas Vidick},
  publisher =	{Schloss Dagstuhl--Leibniz-Zentrum fuer Informatik},
  address =	{Dagstuhl, Germany}
}

@inproceedings{ben-porat23,
author = {Ben-Porat, Omer and Torkan, Rotem},
title = {Learning with Exposure Constraints in Recommendation Systems},
year = {2023},
booktitle = {Proceedings of the ACM Web Conference 2023},
series = {WWW '23}
}

@inproceedings{Zhang22,
author = {Zhang, Hanrui and Cheng, Yu and Conitzer, Vincent},
title = {Efficient Algorithms for Planning with Participation Constraints},
year = {2022},
series = {EC '22}
}

@inproceedings{hardt2016strategic,
  title={Strategic classification},
  author={Hardt, Moritz and Megiddo, Nimrod and Papadimitriou, Christos and Wootters, Mary},
  booktitle={Proceedings of the 2016 ACM conference on innovations in theoretical computer science},
  pages={111--122},
  year={2016}
}

@InProceedings{MansourSW18,
  author =	{Yishay Mansour and Aleksandrs Slivkins and Zhiwei Steven Wu},
  title =	{{Competing Bandits: Learning Under Competition}},
  booktitle =	{Innovations in Theoretical Computer Science Conference (ITCS 2018)},
  year={2018},
   ISBN =	{978-3-95977-060-6},
  ISSN =	{1868-8969},
  URN =		{urn:nbn:de:0030-drops-83341}
}

@inproceedings{MansourSS15,
 author = {Mansour, Yishay and Slivkins, Aleksandrs and Syrgkanis, Vasilis},
 title = {Bayesian Incentive-Compatible Bandit Exploration},
 booktitle = {Proceedings of the Sixteenth ACM Conference on Economics and Computation},
 series = {EC '15},
 year = {2015},
 isbn = {978-1-4503-3410-5},
 location = {Portland, Oregon, USA},
 pages = {565--582},
}

@inproceedings{MansourSSW16,
 author = {Mansour, Yishay and Slivkins, Aleksandrs and Syrgkanis, Vasilis and Wu, Zhiwei Steven},
 title = {Bayesian Exploration: Incentivizing Exploration in Bayesian Games},
 booktitle = {Proceedings of the 2016 ACM Conference on Economics and Computation},
 series = {EC '16},
 year = {2016},
 isbn = {978-1-4503-3936-0},
 location = {Maastricht, The Netherlands},
 pages = {661--661},
 numpages = {1},
 acmid = {2940755},
 publisher = {ACM},
 address = {New York, NY, USA},
}

@article{Che-13,
  title={Optimal Design for Social Learning},
  author={Che, Yeon-Koo and H{\"o}rner, Johannes},
  year={2013}
}

@inproceedings{Sellke21,
author = {Sellke, Mark and Slivkins, Aleksandrs},
title = {The Price of Incentivizing Exploration: A Characterization via Thompson Sampling and Sample Complexity},
year = {2021},
location = {Budapest, Hungary},
series = {EC '21}
}

@inproceedings{Bahar2019FiduciaryB,
  title={Fiduciary Bandits},
  author={Gal Bahar and Omer Ben-Porat and Kevin Leyton-Brown and Moshe Tennenholtz},
  booktitle={International Conference on Machine Learning},
  year={2019}
}

@inproceedings{Bahar19,
author = {Bahar, Gal and Smorodinsky, Rann and Tennenholtz, Moshe},
title = {Social Learning and the Innkeeper's Challenge},
year = {2019},
series = {EC '19}
}

@inproceedings{Immorlica19,
author = {Immorlica, Nicole and Mao, Jieming and Slivkins, Aleksandrs and Wu, Zhiwei Steven},
title = {Bayesian Exploration with Heterogeneous Agents},
year = {2019},
series = {WWW '19}
}

@article{Kremer-JPE14,
    author = {Ilan Kremer and Yishay Mansour and Motty Perry},
    title = {Implementing the "Wisdom of the Crowd"},
    journal = {J. of Political Economy},
    volume = {122},
    issue  = {5},
    pages = {988--1012},
    month = Oct,
    year = {2014},
    note = {Preliminary version appeared in \emph{ACM Conf. on Economics and Computation}, 2014.}
}

@inproceedings{BaharST16,
  author    = {Gal Bahar and
               Rann Smorodinsky and
               Moshe Tennenholtz},
  title     = {Economic Recommendation Systems: One Page Abstract},
  booktitle = {Proceedings of the 2016 {ACM} Conference on Economics and Computation,
               {EC}},
  pages     = {757},
  year      = {2016},
  comment	={(See https://arxiv.org/abs/1507.07191)}
}

@article{Slivkins17,
  author    = {Aleksandrs Slivkins},
  title     = {Incentivizing exploration via information asymmetry},
  journal   = {{ACM} Crossroads},
  volume    = {24},
  number    = {1},
  pages     = {38--41},
  year      = {2017}
}

@article{echenique2019incentive,
  title={Incentive Compatible Active Learning},
  author={Echenique, Federico and Prasad, Siddharth},
  journal={arXiv preprint arXiv:1911.05171},
  year={2019}
}

@book{slivkins2019introduction,
  title={Introduction to Multi-Armed Bandits},
  author={Slivkins, A.},
  isbn={9781680836219},
  series={Foundations and Trends in Machine Learning Series},
  url={https://books.google.com/books?id=6ViCzQEACAAJ},
  year={2019},
  publisher={Now Publishers}
}

@misc{simchowitz2023exploration,
      title={Exploration and Incentives in Reinforcement Learning}, 
      author={Max Simchowitz and Aleksandrs Slivkins},
      year={2023},
      eprint={2103.00360},
      archivePrefix={arXiv},
      primaryClass={cs.LG}
}

@inproceedings{Immorlica20,
author = {Immorlica, Nicole and Mao, Jieming and Slivkins, Aleksandrs and Wu, Zhiwei Steven},
title = {Incentivizing Exploration with Selective Data Disclosure},
year = {2020},
series = {EC '20}
}

@misc{banihashem2023bandit,
      title={Bandit Social Learning: Exploration under Myopic Behavior}, 
      author={Kiarash Banihashem and MohammadTaghi Hajiaghayi and Suho Shin and Aleksandrs Slivkins},
      year={2023},
      eprint={2302.07425},
      archivePrefix={arXiv},
      primaryClass={cs.GT}
}

@Inbook{Lyu2020,
author="Lyu, Lingjuan
and Xu, Xinyi
and Wang, Qian
and Yu, Han",
editor="Yang, Qiang
and Fan, Lixin
and Yu, Han",
title="Collaborative Fairness in Federated Learning",
bookTitle="Federated Learning: Privacy and Incentive",
year="2020",
}

@inproceedings{BlumIHPPV17kidney,
author = {Blum, Avrim and Caragiannis, Ioannis and Haghtalab, Nika and Procaccia, Ariel D. and Procaccia, Eviatar B. and Vaish, Rohit},
title = {Opting into Optimal Matchings},
year = {2017},
series = {SODA '17}
}

@inproceedings{CamilleriWMJJ22,
  author       = {Romain Camilleri and
                  Andrew Wagenmaker and
                  Jamie H. Morgenstern and
                  Lalit Jain and
                  Kevin G. Jamieson},
  title        = {Active Learning with Safety Constraints},
  booktitle    = {NeurIPS},
  year         = {2022},
}

@inproceedings{Donahue22Fair,
author = {Donahue, Kate and Chouldechova, Alexandra and Kenthapadi, Krishnaram},
title = {Human-Algorithm Collaboration: Achieving Complementarity and Avoiding Unfairness},
year = {2022},
series = {FAccT '22}
}

@article{Paxlovid22,
    author = {Najjar-Debbiny, Ronza and Gronich, Naomi and Weber, Gabriel and Khoury, Johad and Amar, Maisam and Stein, Nili and Goldstein, Lee Hilary and Saliba, Walid},
    title = "{Effectiveness of Paxlovid in Reducing Severe Coronavirus Disease 2019 and Mortality in High-Risk Patients}",
    journal = {Clinical Infectious Diseases},
    year = {2022},
}

@inproceedings{Donahue23,
author = {Donahue, Kate and Kleinberg, Jon},
title = {Fairness in Model-Sharing Games},
year = {2023},
series = {WWW '23}
}

@inproceedings{DonahueK21,
  author       = {Kate Donahue and
                  Jon M. Kleinberg},
  title        = {Optimality and Stability in Federated Learning: {A} Game-theoretic
                  Approach},
  booktitle    = {Advances in Neural Information Processing Systems 34: Annual Conference
                  on Neural Information Processing Systems 2021, NeurIPS 2021, December
                  6-14, 2021, virtual},
  year         = {2021},
}

@inproceedings{BlumG21,
  author       = {Avrim Blum and
                  Paul G{\"{o}}lz},
    title        = {Incentive-Compatible Kidney Exchange in a Slightly Semi-Random Model},
  booktitle    = {{EC} '21: The 22nd {ACM} Conference on Economics and Computation,
                  Budapest, Hungary, July 18-23, 2021},
   publisher    = {{ACM}},
  year         = {2021},
}

@article{BlumMansour20,
  author       = {Avrim Blum and
                  Yishay Mansour},
  title        = {Kidney exchange and endless paths: On the optimal use of an altruistic
                  donor},
  journal      = {CoRR},
  volume       = {abs/2010.01645},
  year         = {2020},
  url          = {https://arxiv.org/abs/2010.01645},
  eprinttype    = {arXiv},
  eprint       = {2010.01645},
  bibsource    = {dblp computer science bibliography, https://dblp.org}
}

@article{DickersonPS19,
  author       = {John P. Dickerson and
                  Ariel D. Procaccia and
                  Tuomas Sandholm},
  title        = {Failure-Aware Kidney Exchange},
  journal      = {Manag. Sci.},
  year         = {2019},
}

@inproceedings{AshlagiR11,
  author       = {Itai Ashlagi and
                  Alvin E. Roth},
  title        = {Individual rationality and participation in large scale, multi-hospital
                  kidney exchange},
  booktitle    = {Proceedings 12th {ACM} Conference on Electronic Commerce (EC-2011)},
  year         = {2011},
}

@article{Roth04,
    author = {Roth, Alvin E. and Sönmez, Tayfun and Ünver, M. Utku},
    title = "{Kidney Exchange*}",
    journal = {The Quarterly Journal of Economics},
    year = {2004},
}

@inproceedings{ShenCW22,
  author       = {Jie Shen and
                  Nan Cui and
                  Jing Wang},
  title        = {Metric-Fair Active Learning},
  booktitle    = {International Conference on Machine Learning, {ICML} 2022, 17-23 July
                  2022, Baltimore, Maryland, {USA}},
  series       = {Proceedings of Machine Learning Research},
  year         = {2022},
}

@article{FreundSST97,
  author       = {Yoav Freund and
                  H. Sebastian Seung and
                  Eli Shamir and
                  Naftali Tishby},
  title        = {Selective Sampling Using the Query by Committee Algorithm},
  journal      = {Mach. Learn.},
  year         = {1997},
}

@article{Hanneke14survey,
year = {2014},
volume = {7},
journal = {Foundations and Trends® in Machine Learning},
title = {Theory of Disagreement-Based Active Learning},
author = {Steve Hanneke}
}

@inproceedings{Buchfuhrer10,
  author       = {David Buchfuhrer and
                  Shaddin Dughmi and
                  Hu Fu and
                  Robert Kleinberg and
                  Elchanan Mossel and
                  Christos H. Papadimitriou and
                  Michael Schapira and
                  Yaron Singer and
                  Christopher Umans},
  title        = {Inapproximability for VCG-Based Combinatorial Auctions},
  booktitle    = {Proceedings of the Twenty-First Annual {ACM-SIAM} Symposium on Discrete
                  Algorithms, {SODA} 2010, Austin, Texas, USA, January 17-19, 2010},
  publisher    = {{SIAM}},
  year         = {2010},
}

@inproceedings{Baek21,
  author       = {Jackie Baek and
                  Vivek F. Farias},
  title        = {Fair Exploration via Axiomatic Bargaining},
  booktitle    = {Advances in Neural Information Processing Systems 34: Annual Conference
                  on Neural Information Processing Systems 2021, NeurIPS 2021, December
                  6-14, 2021, virtual},
  pages        = {22034--22045},
  year         = {2021},
}

@article{aussel2020combining,
  title={Combining federated and active learning for communication-efficient distributed failure prediction in aeronautics},
  author={Aussel, Nicolas and Chabridon, Sophie and Petetin, Yohan},
  journal={arXiv preprint arXiv:2001.07504},
  year={2020}
}

@article{shen2016distributed,
  title={Distributed active learning},
  author={Shen, Pengcheng and Li, Chunguang and Zhang, Zhaoyang},
  journal={IEEE Access},
  volume={4},
  pages={2572--2579},
  year={2016},
  publisher={IEEE}
}

@inproceedings{xu2023fair,
  author       = {Xinyi Xu and
                  Zhaoxuan Wu and
                  Arun Verma and
                  Chuan Sheng Foo and
                  Bryan Kian Hsiang Low},
  title        = {{FAIR:} Fair Collaborative Active Learning with Individual Rationality
                  for Scientific Discovery},
  booktitle    = {International Conference on Artificial Intelligence and Statistics,
                  25-27 April 2023, Palau de Congressos, Valencia, Spain},
  year         = {2023},
}

@article{Camilleri23,
  author       = {Romain Camilleri and
                  Andrew Wagenmaker and
                  Jamie Morgenstern and
                  Lalit Jain and
                  Kevin Jamieson},
  title        = {Fair Active Learning in Low-Data Regimes},
  journal      = {CoRR},
  volume       = {abs/2312.08559},
  year         = {2023},
  eprinttype    = {arXiv},
  eprint       = {2312.08559}
}

@article{wang2023framework,
  title={A Framework for Incentivized Collaborative Learning},
  author={Wang, Xinran and Le, Qi and Khan, Ahmad Faraz and Ding, Jie and Anwar, Ali},
  journal={arXiv preprint arXiv:2305.17052},
  year={2023}
}

@book{Shalev-Shwartz_Ben-David_2014, place={Cambridge}, title={Understanding Machine Learning: From Theory to Algorithms}, publisher={Cambridge University Press}, author={Shalev-Shwartz, Shai and Ben-David, Shai}, year={2014}}

@inproceedings{Blum17,
 author = {Blum, Avrim and Haghtalab, Nika and Procaccia, Ariel D and Qiao, Mingda},
 booktitle = {Advances in Neural Information Processing Systems},
 editor = {I. Guyon and U. Von Luxburg and S. Bengio and H. Wallach and R. Fergus and S. Vishwanathan and R. Garnett},
 pages = {},
 publisher = {Curran Associates, Inc.},
 title = {Collaborative PAC Learning},
 volume = {30},
 year = {2017}
}

@string{iclr ={Proceedings of the International Conference on Learning Representations (ICLR)}}

@string{iros = {Proceedings of the IEEE/RSJ International Conference on Intelligent Robots and Systems (IROS)}}

@string{ijrr="International Journal of Robotics Research"}

@string{rss="Proceedings of Robotics: Science and Systems (RSS)"}

@string{aaai="Proceedings of the National Conference on Artificial Intelligence (AAAI)"}

@string{uai="Proceedings of the Conference on Uncertainty in Artificial Intelligence (UAI)"}

@string{icml="Proceedings of the International Conference on Machine Learning (ICML)"}

@string{jmlr="Journal of Machine Learning Research"}

@string{ijcai="Proceedings of the International Joint Conference on Artificial Intelligence (IJCAI)"}

@string{neurips="Advances in Neural Information Processing Systems (NeurIPS)"}

@string{corl="Proceedings of the Conference on Robot Learning (CoRL)"}

@string{colt="Proceedings of the Annual Conference on Learning Theory (COLT)"}

@string{ecml-pkdd="Proceedings of the Joint European Conference on Machine Learning and Knowledge Discovery in Databases (ECML PKDD)"}

@string{jair="Journal of Artificial Intelligence Research"}

@inproceedings{saha2021dueling,
  title={Dueling convex optimization},
  author={Saha, Aadirupa and Koren, Tomer and Mansour, Yishay},
  booktitle={International Conference on Machine Learning},
  pages={9245--9254},
  year={2021},
  organization={PMLR}
}

@inproceedings{kane2017active,
  title={Active classification with comparison queries},
  author={Kane, Daniel M and Lovett, Shachar and Moran, Shay and Zhang, Jiapeng},
  booktitle={2017 IEEE 58th Annual Symposium on Foundations of Computer Science (FOCS)},
  pages={355--366},
  year={2017},
  organization={IEEE}
}

@inproceedings{Balcan2015,
  author    = {Maria{-}Florina Balcan and
               Avrim Blum and
               Santosh S. Vempala},
  title     = {Efficient Representations for Lifelong Learning and Autoencoding},
  booktitle = colt,
  OPTseries    = {{JMLR} Workshop and Conference Proceedings},
  pages     = {191--210},
  year      = {2015},
}

@inproceedings{sadigh17,
	OPTaddress = {Cambridge, MA},
	author = {Sadigh, Dorsa and Dragan, Anca D and Sastry, Shankar and Seshia, Sanjit A},
	booktitle = rss,
	month = jul,
	title = {Active preference-based learning of reward functions},
	year = {2017}}

@inproceedings{rothkopf2011preference,
  title={Preference elicitation and inverse reinforcement learning},
  author={Rothkopf, Constantin A and Dimitrakakis, Christos},
  booktitle={Proceedings of the Joint European Conference on Machine Learning and Knowledge Discovery in Databases},
  pages={34--48},
  year={2011},
}

@inproceedings{wirth2016model,
  title={Model-free preference-based reinforcement learning},
  author={Wirth, Christian and F{\"u}rnkranz, Johannes and Neumann, Gerhard},
  booktitle=aaai,
  year={2016}
}

@inproceedings{akrour2012april,
  title={{APRIL}: {A}ctive preference learning-based reinforcement learning},
  author={Akrour, Riad and Schoenauer, Marc and Sebag, Mich{\`e}le},
  booktitle={Proceedings of the Joint European Conference on Machine Learning and Knowledge Discovery in Databases},
  pages={116--131},
  year={2012},
}

@article{jain15,
	author = {Jain, Ashesh and Sharma, Shikhar and Joachims, Thorsten and Saxena, Ashutosh},
	journal = ijrr,
	number = {10},
	pages = {1296--1313},
	title = {Learning preferences for manipulation tasks from online coactive feedback},
	volume = {34},
	year = {2015}}

@inproceedings{wilson2012bayesian,
  title={A {B}ayesian approach for policy learning from trajectory preference queries},
  author={Wilson, Aaron and Fern, Alan and Tadepalli, Prasad},
  booktitle=neurips,
  OPTvolume={25},
  year={2012}
}

@article{furnkranz12,
	author = {F{\"u}rnkranz, Johannes and H{\"u}llermeier, Eyke and Cheng, Weiwei and Park, Sang-Hyeun},
	journal = {Machine Learning},
	number = {1},
	pages = {123--156},
	title = {Preference-based reinforcement learning: {A} formal framework and a policy iteration algorithm},
	volume = {89},
	year = {2012}}

@article{ailon12,
	author = {Ailon, Nir},
	journal = jmlr,
	month = jan,
	number = {1},
	pages = {137--164},
	title = {An Active Learning Algorithm for Ranking from Pairwise Preferences with an Almost Optimal Query Complexity},
	volume = {13},
	year = {2012}}

@inproceedings{biyik18,
	OPTaddress = {Zurich, Switzerland},
	author = {Biyik, Erdem and Sadigh, Dorsa},
	booktitle = corl,
	month = oct,
	pages = {519--528},
	title = {Batch active preference-based learning of reward functions},
	year = {2018}}

@inproceedings{christiano2017deep,
	OPTaddress = {Long Beach, CA},
	author = {Christiano, Paul F and Leike, Jan and Brown, Tom and Martic, Miljan and Legg, Shane and Amodei, Dario},
	booktitle = neurips,
	month = dec,
	title = {Deep reinforcement learning from human preferences},
	year = {2017}}

@inproceedings{ibarz18,
	OPTaddress = {Montreal, Canada},
	author = {Ibarz, Borja and Leike, Jan and Pohlen, Tobias and Irving, Geoffrey and Legg, Shane and Amodei, Dario},
	booktitle = neurips,
	month = dec,
	title = {Reward learning from human preferences and demonstrations in {Atari}},
	year = {2018}}

@article{lee21,
	author = {Lee, Kimin and Smith, Laura and Abbeel, Pieter},
	journal = {arXiv preprint arXiv:2106.05091},
	title = {{PEBBLE}: {F}eedback-efficient interactive reinforcement learning via relabeling experience and unsupervised pre-training},
	year = {2021}}

@inproceedings{cheng11,
	OPTaddress = {Athens, Greece},
	author = {Cheng, Weiwei and F{\"u}rnkranz, Johannes and H{\"u}llermeier, Eyke and Park, Sang-Hyeun},
	booktitle = ecml-pkdd,
	month = sep,
	pages = {312--327},
	title = {Preference-based policy iteration: {L}everaging preference learning for reinforcement learning},
	year = {2011}}

@article{knox22,
	author = {Knox, W Bradley and Hatgis-Kessell, Stephane and Booth, Serena and Niekum, Scott and Stone, Peter and Allievi, Alessandro},
	journal = {arXiv preprint arXiv:2206.02231},
	title = {Models of human preference for learning reward functions},
	year = {2022}}

@inproceedings{mannor01,
	OPTaddress = {Vancouver, B.C., Canada},
	author = {Mannor, Shie and Shimkin, Nahum},
	booktitle = neurips,
	month = dec,
	title = {The steering approach for multi-criteria reinforcement learning},
	year = {2001}}

@article{mannor04,
	author = {Mannor, Shie and Shimkin, Nahum},
	journal = jmlr,
	month = apr,
	pages = {325--360},
	title = {A geometric approach to multi-criterion reinforcement learning},
	volume = {5},
	year = {2004}}

@inproceedings{cheung19,
	OPTaddress = {Vancouver, B.C., Canada},
	author = {Cheung, Wang Chi},
	booktitle = neurips,
	title = {Regret minimization for reinforcement learning with vectorial feedback and complex objectives},
	year = {2019}}

@inproceedings{barrett08,
	OPTaddress = {Helsinki, Finland},
	author = {Barrett, Leon and Narayanan, Srini},
	booktitle = icml,
	month = jul,
	pages = {41--47},
	title = {Learning all optimal policies with multiple criteria},
	year = {2008}}

@inproceedings{chatterjee06,
	OPTaddress = {Marseille, France},
	author = {Chatterjee, Krishnendu and Majumdar, Rupak and Henzinger, Thomas A},
	booktitle = {Proceedings of the Annual Symposium on Theoretical Aspects of Computer Science (STACS)},
	month = feb,
	pages = {325--336},
	title = {Markov decision processes with multiple objectives},
	year = {2006}}

@inproceedings{chatterjee07,
	OPTaddress = {New Delhi, India},
	author = {Chatterjee, Krishnendu},
	booktitle = {Proceedings of the International Conference on Foundations of Software Technology and Theoretical Computer Science (FSTTCS)},
	month = dec,
	pages = {473--484},
	title = {Markov decision processes with multiple long-run average objectives},
	year = {2007}}

@article{roijers13,
	author = {Roijers, Diederik M and Vamplew, Peter and Whiteson, Shimon and Dazeley, Richard},
	journal = jair,
	month = oct,
	pages = {67--113},
	title = {A survey of multi-objective sequential decision-making},
	volume = {48},
	year = {2013}}

@book{mcauley2022,
      title     = "Personalized Machine Learning",
      author    = "McAuley, Julian",
      year      = 2022,
      note      = "in press",
      publisher = "Cambridge University Press"
    }

@article{HayesRBKMRVZDHH22MORLsurvey,
  author    = {Conor F. Hayes and
               Roxana Radulescu and
               Eugenio Bargiacchi and
               Johan K{\"{a}}llstr{\"{o}}m and
               Matthew Macfarlane and
               Mathieu Reymond and
               Timothy Verstraeten and
               Luisa M. Zintgraf and
               Richard Dazeley and
               Fredrik Heintz and
               Enda Howley and
               Athirai A. Irissappane and
               Patrick Mannion and
               Ann Now{\'{e}} and
               Gabriel de Oliveira Ramos and
               Marcello Restelli and
               Peter Vamplew and
               Diederik M. Roijers},
  title     = {A practical guide to multi-objective reinforcement learning and planning},
  journal   = {Autonomous Agents and Multi-Agent Systems},
  year      = {2022},
}

@inproceedings{chen19a,
	OPTaddress = {Macau, China},
	author = {Chen, Xi and Ghadirzadeh, Ali and Bj{\"o}rkman, M{\aa}rten and Jensfelt, Patric},
	booktitle = iros,
	month = nov,
	pages = {977--983},
	title = {Meta-learning for multi-objective reinforcement learning},
	year = {2019}}

@inproceedings{bhatia20,
	author = {Bhatia, Kush and Pananjady, Ashwin and Bartlett, Peter and Dragan, Anca and Wainwright, Martin J},
	booktitle = neurips,
	month = dec,
	pages = {7413--7424},
	title = {Preference learning along multiple criteria: {A} game-theoretic perspective},
	year = {2020}}

@article{doumpos07,
	author = {Doumpos, Michael and Zopounidis, Constantin},
	journal = {Computational Optimization and Applications},
	number = {1},
	pages = {61--80},
	title = {Regularized estimation for preference disaggregation in multiple criteria decision making},
	volume = {38},
	year = {2007}}

@ARTICLE{Alharin20SurveyInt,
  author={Alharin, Alnour and Doan, Thanh-Nam and Sartipi, Mina},
  journal={IEEE Access}, 
  title={Reinforcement Learning Interpretation Methods: A Survey}, 
  year={2020},
  volume={8},
  number={},
  pages={171058-171077},
  doi={10.1109/ACCESS.2020.3023394}}

@article{jamieson2011active,
  title={Active ranking using pairwise comparisons},
  author={Jamieson, Kevin G and Nowak, Robert},
  journal={Advances in neural information processing systems},
  volume={24},
  year={2011}
}

@inproceedings{SG18,
  title={Battle of Bandits},
  author={Saha, Aadirupa and Gopalan, Aditya},
  booktitle={Uncertainty in Artificial Intelligence},
  year={2018}
}

@inproceedings{SGwin18, 
  title={{PAC Battling Bandits in the Plackett-Luce Model}}, 
  author={Saha, Aadirupa and Gopalan, Aditya}, 
  booktitle={Algorithmic Learning Theory}, 
  pages={700--737}, 
  year={2019} 
}

@inproceedings{Ailon+14,
  title={Reducing Dueling Bandits to Cardinal Bandits.},
  author={Ailon, Nir and Karnin, Zohar Shay and Joachims, Thorsten},
  booktitle={ICML},
  volume={32},
  pages={856--864},
  year={2014}
}

@inproceedings{Zoghi+14RUCB,
  title={Relative upper confidence bound for the $k$-armed dueling bandit problem},
  author={Zoghi, Masrour and Whiteson, Shimon and Munos, Remi and Rijke, Maarten de and others},
  booktitle={JMLR Workshop and Conference Proceedings},
  number={32},
  pages={10--18},
  year={2014},
  organization={JMLR}
}

@inproceedings{Sui+17,
  title={Multi-dueling bandits with dependent arms},
  author={Yanan Sui and Vincent Zhuang and Joel Burdick and Yisong Yue},
  booktitle={Conference on Uncertainty in Artificial Intelligence},
  series={UAI'17},
  year={2017},
}

@inproceedings{sui18survey, 
  title={Advancements in Dueling Bandits.},
  author={Sui, Yanan and Zoghi, Masrour and Hofmann, Katja and Yue, Yisong},
  booktitle={IJCAI},
  pages={5502--5510},
  year={2018}
}

@article{Ren+18,
  title={P{AC} Ranking from Pairwise and Listwise Queries: Lower Bounds and Upper Bounds},
  author={Ren, Wenbo and Liu, Jia and Shroff, Ness B},
  journal={arXiv preprint arXiv:1806.02970},
  year={2018}
}

@article{bengs2021preference,
  title={Preference-based online learning with dueling bandits: A survey.},
  author={Bengs, Viktor and Busa-Fekete, R{\'o}bert and El Mesaoudi-Paul, Adil and H{\"u}llermeier, Eyke},
  journal={J. Mach. Learn. Res.},
  year={2021}
}

@article{WirthRlSurvey2017,
author = {Wirth, Christian and Akrour, Riad and Neumann, Gerhard and F\"{u}rnkranz, Johannes},
title = {A Survey of Preference-Based Reinforcement Learning Methods},
year = {2017},
journal = {J. Mach. Learn. Res.},
}

@inproceedings{BalcanVW16,
	author = {Maria-Florina Balcan and
               Ellen Vitercik and
               Colin White},
	booktitle = COLT,
	title = {Learning Combinatorial Functions from Pairwise Comparisons},
	year = {2016}}

@inproceedings{YonaMEG22,
	author = {Gal Yona and
               Shay Moran and
               Gal Elidan and
               Amir Globerson},
	booktitle = UAI,
	title = {Active learning with label comparisons},
	year = {2022}}

@inproceedings{PacchianoDuelingRL,
  author    = {Aldo Pacchiano and
               Aadirupa Saha and
               Jonathan Lee},
               booktitle = AISTATS,
  title     = {Dueling {RL:} Reinforcement Learning with Trajectory Preferences},
  year      = {2022},
}

@inproceedings{mannor2014approachability,
  title={Approachability in unknown games: Online learning meets multi-objective optimization},
  author={Mannor, Shie and Perchet, Vianney and Stoltz, Gilles},
  booktitle={Conference on Learning Theory},
  pages={339--355},
  year={2014},
  organization={PMLR}
}

@inproceedings{wang2022imo,
  title={IMO\^{} 3: Interactive Multi-Objective Off-Policy Optimization},
  author={Wang, Nan and Wang, Hongning and Karimzadehgan, Maryam and Kveton, Branislav and Boutilier, Craig},
  booktitle={Proceedings of the Thirty-First International Joint Conference on Artificial Intelligence, IJCAI-22},
  year={2022}
}

@article{Marinescu_Razak_Wilson_2017, 
title={Multi-Objective Influence Diagrams with Possibly Optimal Policies},
journal={Proceedings of the AAAI Conference on Artificial Intelligence}, 
author={Marinescu, Radu and Razak, Abdul and Wilson, Nic}, year={2017} }

@article{VARIAN197463,
title = {Equity, envy, and efficiency},
journal = {Journal of Economic Theory},
volume = {9},
number = {1},
pages = {63-91},
year = {1974},
issn = {0022-0531},
author = {Hal R Varian}
}

@article{foley1967resource,
  title={Resource allocation and the public sector.},
  author={Foley, Duncan Karl},
  journal = {Yale economic essays},
  volume = {7},
  number = {1},
  pages = {45-98},
  year={1967}
}

@inproceedings{jia2019towards,
  title={Towards efficient data valuation based on the shapley value},
  author={Jia, Ruoxi and Dao, David and Wang, Boxin and Hubis, Frances Ann and Hynes, Nick and G{\"u}rel, Nezihe Merve and Li, Bo and Zhang, Ce and Song, Dawn and Spanos, Costas J},
  booktitle={The 22nd International Conference on Artificial Intelligence and Statistics},
  pages={1167--1176},
  year={2019},
  organization={PMLR}
}

@inproceedings{balkanski2017statistical,
  title={Statistical cost sharing},
  author={Balkanski, Eric and Syed, Umar and Vassilvitskii, Sergei},
  booktitle={Proceedings of the 31st International Conference on Neural Information Processing Systems},
  pages={6222--6231},
  year={2017}
}

@inproceedings{agarwal2019marketplace,
  title={A marketplace for data: An algorithmic solution},
  author={Agarwal, Anish and Dahleh, Munther and Sarkar, Tuhin},
  booktitle={Proceedings of the 2019 ACM Conference on Economics and Computation},
  pages={701--726},
  year={2019}
}

@misc{NVIDIA2,
	Author = {Kimberly Powell},
	Howpublished = {\url{https://blogs.nvidia.com/blog/2019/12/01/clara-federated-learning/}},
	Title = {{NVIDIA} Clara Federated Learning to Deliver AI to Hospitals While Protecting Patient Data},
	Year = 2019
}

@article{10.2307/3690406,
 ISSN = {0364765X, 15265471},
 URL = {http://www.jstor.org/stable/3690406},
 author = {Serge A. Plotkin and David B. Shmoys and Éva Tardos},
 journal = {Mathematics of Operations Research},
 number = {2},
 pages = {257--301},
 publisher = {INFORMS},
 title = {Fast Approximation Algorithms for Fractional Packing and Covering Problems},
 volume = {20},
 year = {1995}
}

@techreport{wen2019federated,
  title={Federated Learning powered by {NVIDIA} {Clara}},
  author={Wen, Yuhong and Li, Wenqi and Roth, Holger and Dogra, Prerna},
  journal={{NVIDIA} Developer Blog},
  year = {2019},
  month = {December},
  url = {https://developer.nvidia.com/blog/federated-learning-clara/} 
}

@article{arora_multiplicative_nodate,
	title = {The {Multiplicative} {Weights} {Update} {Method}: a {Meta} {Algorithm} and {Applications}},
	language = {en},
	author = {Arora, Sanjeev and Hazan, Elad and Kale, Satyen},
	  journal={Theory of Computing},
  year={2012},
	pages = {31}
}

@inproceedings{luo2015achieving,
  title={Achieving all with no parameters: Adanormalhedge},
  author={Luo, Haipeng and Schapire, Robert E},
  booktitle={Conference on Learning Theory},
  pages={1286--1304},
  year={2015},
  organization={PMLR}
}

@inproceedings{mcmahan2017communication,
  title={Communication-efficient learning of deep networks from decentralized data},
  author={McMahan, Brendan and Moore, Eider and Ramage, Daniel and Hampson, Seth and y Arcas, Blaise Aguera},
  booktitle={Artificial Intelligence and Statistics},
  pages={1273--1282},
  year={2017},
  organization={PMLR}
}

@incollection{chen:tight2018,
	title = {Tight {Bounds} for {Collaborative} {PAC} {Learning} via {Multiplicative} {Weights}},
	booktitle = {Advances in {Neural} {Information} {Processing} {Systems} 31},
	publisher = {Curran Associates, Inc.},
	author = {Chen, Jiecao and Zhang, Qin and Zhou, Yuan},
	editor = {Bengio, S. and Wallach, H. and Larochelle, H. and Grauman, K. and Cesa-Bianchi, N. and Garnett, R.},
	year = {2018},
	pages = {3598--3607}
	}

@article{donahue2020model,
  title={Model-sharing Games: Analyzing Federated Learning Under Voluntary Participation},
  author={Donahue, Kate and Kleinberg, Jon},
  journal={Proceedings of the AAAI Conference on Artificial Intelligence},
  year={2021},
  volume={33},
}

@inproceedings{li2019fair,
  title={Fair Resource Allocation in Federated Learning},
  author={Li, Tian and Sanjabi, Maziar and Beirami, Ahmad and Smith, Virginia},
  booktitle={International Conference on Learning Representations},
  year={2019}
}

@incollection{lyu2020collaborative,
  title={Collaborative fairness in federated learning},
  author={Lyu, Lingjuan and Xu, Xinyi and Wang, Qian and Yu, Han},
  booktitle={Federated Learning},
  pages={189--204},
  year={2020},
  publisher={Springer}
}

@inproceedings{yu2020fairness,
  title={A fairness-aware incentive scheme for federated learning},
  author={Yu, Han and Liu, Zelei and Liu, Yang and Chen, Tianjian and Cong, Mingshu and Weng, Xi and Niyato, Dusit and Yang, Qiang},
  booktitle={Proceedings of the AAAI/ACM Conference on AI, Ethics, and Society},
  pages={393--399},
  year={2020}
}

@article{lin2019free,
  title={Free-riders in federated learning: Attacks and defenses},
  author={Lin, Jierui and Du, Min and Liu, Jian},
  journal={arXiv preprint arXiv:1911.12560},
  year={2019}
}

@article{zhang2020hierarchically,
  title={Hierarchically fair federated learning},
  author={Zhang, Jingfeng and Li, Cheng and Robles-Kelly, Antonio and Kankanhalli, Mohan},
  journal={arXiv preprint arXiv:2004.10386},
  year={2020}
}

@article{nash1951non,
  title={Non-cooperative games},
  author={Nash, John},
  journal={Annals of mathematics},
  pages={286--295},
  year={1951},
  publisher={JSTOR}
}

@article{caragiannis2012efficiency,
  title={The efficiency of fair division},
  author={Caragiannis, Ioannis and Kaklamanis, Christos and Kanellopoulos, Panagiotis and Kyropoulou, Maria},
  journal={Theory of Computing Systems},
  volume={50},
  number={4},
  pages={589--610},
  year={2012},
  publisher={Springer}
}

@article{anshelevich2008price,
  title={The price of stability for network design with fair cost allocation},
  author={Anshelevich, Elliot and Dasgupta, Anirban and Kleinberg, Jon and Tardos, {\'E}va and Wexler, Tom and Roughgarden, Tim},
  journal={SIAM Journal on Computing},
  volume={38},
  number={4},
  pages={1602--1623},
  year={2008},
  publisher={SIAM}
}

@article{DBLP:journals/corr/CohenATS17,
  author    = {Gregory Cohen and
               Saeed Afshar and
               Jonathan Tapson and
               Andr{\'{e}} van Schaik},
  title     = {{EMNIST:} an extension of {MNIST} to handwritten letters},
  journal   = {CoRR},
  volume    = {abs/1702.05373},
  year      = {2017},
  url       = {http://arxiv.org/abs/1702.05373},
  archivePrefix = {arXiv},
  eprint    = {1702.05373},
  bibsource = {dblp computer science bibliography, https://dblp.org}
}

@inproceedings{kingma15,
  author    = {Diederik P. Kingma and
               Jimmy Ba},
  title     = {Adam: {A} Method for Stochastic Optimization},
  booktitle={International Conference on Learning Representations},
  year      = {2015}
}

@article{srivastava2014dropout,
  title={Dropout: a simple way to prevent neural networks from overfitting},
  author={Srivastava, Nitish and Hinton, Geoffrey and Krizhevsky, Alex and Sutskever, Ilya and Salakhutdinov, Ruslan},
  journal={The journal of machine learning research},
  volume={15},
  number={1},
  pages={1929--1958},
  year={2014},
  publisher={JMLR. org}
}

@inproceedings{altschuler2018online,
  title={Online learning over a finite action set with limited switching},
  author={Altschuler, Jason and Talwar, Kunal},
  booktitle={Conference On Learning Theory},
  pages={1569--1573},
  year={2018}
}

\end{document}